\documentclass[final,hidelinks,onefignum,onetabnum]{siamart220329}

\usepackage{lipsum}
\usepackage{amsfonts}
\usepackage{graphicx}
\usepackage{epstopdf}
\usepackage{algorithmic}
\ifpdf
  \DeclareGraphicsExtensions{.eps,.pdf,.png,.jpg}
\else
  \DeclareGraphicsExtensions{.eps}
\fi

\newsiamremark{remark}{Remark}
\newsiamremark{hypothesis}{Hypothesis}
\crefname{hypothesis}{Hypothesis}{Hypotheses}
\newsiamthm{claim}{Claim}

\headers{SELTO: Sample-Efficient Learned Topology Optimization}{Sören Dittmer, David Erzmann, Henrik Harms, Peter Maass}

\title{SELTO: Sample-Efficient Learned\linebreak Topology Optimization\thanks{The authors would like to thank the Federal Ministry for Economic Affairs
and Climate Action of Germany (BMWK) and the German Aerospace Center (DLR) Space Agency for supporting this work (grant no. 50 RL 2060)}}

\author{Sören Dittmer\footnotemark[2]~\footnotemark[3]~\footnotemark[4]
\and David Erzmann\footnotemark[2]~\footnotemark[4]
\and Henrik Harms\footnotemark[2]
\and Peter Maass\footnotemark[2]}

\usepackage{amsopn}

\ifpdf
\hypersetup{
  pdftitle={SELTO: Sample-Efficient Learned Topology Optimization},
  pdfauthor={Sören Dittmer, David Erzmann, Henrik Harms, Peter Maass}
}
\fi

\usepackage{caption}
\usepackage{subcaption}
\usepackage{amsmath, amssymb, amsfonts}
\usepackage{multirow}
\usepackage{makecell}
\usepackage{cellspace} 
\usepackage[section]{placeins}
\usepackage{tabularx,booktabs}
\newcolumntype{Y}{>{\centering\arraybackslash}X}
\usepackage{array}
\usepackage{stackengine}
\usepackage{enumitem}
\usepackage{siunitx}

\newcommand\xrowht[2][0]{\addstackgap[.5\dimexpr#2\relax]{\vphantom{#1}}}

\begin{document}

\maketitle

\begin{abstract}
Recent developments in Deep Learning (DL) suggest a vast potential for Topology Optimization (TO). However, while there are some promising attempts, the subfield still lacks a firm footing regarding basic methods and datasets. We aim to address both points. First, we explore physics-based preprocessing and equivariant networks to create sample-efficient components for TO DL pipelines. We evaluate them in a large-scale ablation study using end-to-end supervised training. The results demonstrate a drastic improvement in sample efficiency and the predictions' physical correctness. Second, to improve comparability and future progress, we publish the two first TO datasets containing problems and corresponding ground truth solutions.
\end{abstract}



\def\thefootnote{$\dagger$}\footnotetext{Center for Industrial Mathematics, University of Bremen, Germany (\email{sdittmer@math.uni-bremen.de}, \email{erzmann@uni-bremen.de}, \email{hharms@uni-bremen.de}, \email{pmaass@uni-bremen.de})}
\def\thefootnote{$\ddagger$}\footnotetext{Cambridge Image Analysis, Centre for Mathematical Sciences, University of Cambridge, UK}
\def\thefootnote{$\S$}\footnotetext{Equal contribution}
\renewcommand{\thefootnote}{\arabic{footnote}}

\section{Introduction}
\label{sec:introduction}
The computational discipline of Topology Optimization (TO) generates mechanical structures. Increased computational power made TO an integral tool for engineers in fields ranging from heat transfer~\cite{dede2009multiphysics} and acoustics~\cite{duhring2008acoustic, yoon2007topology} to fluid~\cite{borrvall2003topology} and solid mechanics~\cite{eschenauer2001topology}. Still, TO remains computationally costly and often time-consuming~\cite{aage2015topology, jensen2021topology}. Recently, Deep Learning (DL) approaches tried to address this. While the approaches are promising, the subfield lacks foundations. This paper aims to establish these foundations, focusing on linear elasticity problems.

The lack of foundations shows most in two places. First, we see a need for more involvement of physics priors in the DL pipeline and the evaluation of their efficacy. Second, to the authors' knowledge, no publicly available TO datasets exist. The release of public datasets often proved to be the deciding spark for the flowering of DL subfields. Here we aim to address both points.

We will now give a high-level overview of the classical SIMP~\cite{bendsoe2003topology} method to frame the problem DL methods try to solve. \textit{Solid Isotropic Material with Penalization} (SIMP)~\cite{bendsoe2003topology}, arguably the most important classical TO method, uses a density-based setup. Here one discretizes a given design domain into voxels, each having a density value between $0$ and $1$. These values represent the amount of material in the voxel, thereby defining a mechanical structure. One employs the partial differential equation (PDE) governing the physical phenomenon of interest to determine the performance based on specified constraints and cost functions. One then adjusts the voxel densities via iterative optimization methods to improve performance. From a mathematical perspective, this constitutes a PDE parameter identification problem.

While much progress has been made, classical methods' iterative nature and the in resolution superlinear computational PDE-cost makes classical methods highly computationally demanding -- often to the point of practical impossibility~\cite{aage2015topology}. Recent research tries to overcome these challenges using deep learning (DL), i.e., neural networks~\cite{bengio2017deep}, to speed up and improve the optimization process.

While these DL methods can often solve TO problems in less than a second, they still lack a solid foundation regarding architecture, data preprocessing, evaluation criteria, benchmarks, and datasets. Overall, the subdiscipline of applying DL to TO is in its infancy, with most papers focusing on two-dimensional settings~\cite{lee2020cnn, nie2021topologygan, oh2019deep, qian2021accelerating, sosnovik2019neural, wang2021deep, zhang2019deep, zhang2021tonr}. In particular, the lack of datasets drastically limits the evaluation and comparability of new approaches.

This paper establishes and compares a set of tools for DL-based TO, including the choice of network architecture, data preprocessing techniques, and the incorporation of physical priors. The paper also publishes two three-dimensional datasets containing almost \num{10000} TO problems and corresponding solutions.

As in DL, not only the amount but also the similarity of the training data to one's problem plays a crucial role; we also study the so-called \textit{generalization}~\cite{kawaguchi2017generalization} of our tools. Generalization is critical for TO, as large-scale training data generation costs can be prohibitive.

Our main contributions are:
\begin{itemize}
    \item We develop and evaluate physical priors in the DL pipeline, e.g., PDE-based preprocessing and equivariant models. 
    \item We provide a large-scale ablation study to assess the sample-efficiency of different architectures and preprocessings, i.e., we study the efficacy in the small data setting.
    \item We provide the first publicly available TO datasets containing problems and corresponding ground truth solutions.
\end{itemize}
We believe the publication of datasets will improve the field's comparability and incorporating physical laws into the data pipeline marks a significant step toward real-world applicability.

\section{Preliminaries and related work}
\label{sec:Preliminaries}
We now briefly introduce classical TO, followed by a review of the current literature on DL for TO.

\subsection{Density-based topology optimization}
We start by discussing density-based TO, arguably the most common classical TO approach -- also used to generate our datasets (see \Cref{3_sec:datasets}).

Density-based TO~\cite{bendsoe2003topology} aims to minimize a cost or objective function by adjusting the material's density distribution ${\rho:\Omega\to\{\rho_{\min},1\}}$ over a fixed domain $\Omega\subset\mathbb{R}^d$, typically $d\in\{2, 3\}$. Here, $0<\rho_\text{min}\ll1$ defines a minimal density value. While the final objective is to obtain binary densities $\rho(\cdot)\in \{0,1\}$, setting $\rho_\text{min} > 0$ is a numerical necessity for solving the governing PDE. Additionally, this optimization is subject to physical constraints. The specified objective function and constraints may vary depending on the user's needs.

This paper focuses on compliance minimization, the most common setting for mechanical problems. The corresponding optimization problem reads as follows:
\begin{subequations}
\begin{alignat}{2}
&\!\min_{\rho} &\qquad& F^Tu(\rho) \label{eq:compliance}\\
& \text{subject to} & & K(\rho)u=F, \label{eq:pde}\\
& & & \|\rho\|_1 \leq V_{\max}, \label{eq:volume}\\
& & & \sigma_\text{vM}(u) \leq \sigma_\text{ys}. \label{eq:yield}
\end{alignat}
\end{subequations}
Here \eqref{eq:compliance} is the compliance objective function, $F$ represents the global load distribution, $u$ are the displacement and $K$ is the symmetric positive operator of linear elasticity. $K$ includes the characteristic properties of the used material described by Young's modulus and Poisson's ratio, which we denote by $E\in \mathbb{R}$ and $\nu\in [0,0.5]$, respectively. One constrains the amount of allowed material by $V_{\max}$~\eqref{eq:volume}, and one may include a stress constraint~\eqref{eq:yield}. This is done to ensure that the maximal von Mises stress $\sigma_\text{vM}$ is below the yield stress $\sigma_\text{ys}$ of the material. The von Mises stresses are used to predict mechanical yielding and are derived non-linearly from the displacements $u$.

In practice, the SIMP~\cite{bendsoe2003topology} method relaxes the problem. Instead of strictly binary density values, one allows $\rho:\Omega\to[\rho_{\min},1]$ and encourages near-binary densities by extending Young's modulus over $\rho$'s interval via $E(\rho) = E_0\rho^p$, where $E_0$ is the original (isotropic) material's modulus. The exponent $p$ controls the penalization of non-binary densities, usually $p=3$. For implementation purposes, one discretizes the design space $\Omega$ into a regular voxel grid and iteratively updates $\rho$ until a user-specified convergence criterion is met.

As discussed in the introduction, despite several advancements in structural TO, one of the main challenges is the high computational cost. The main bottleneck in classical iterative approaches is that each iteration uses the displacements and stresses for the current density and therefore has to solve the PDE for linear elasticity in \eqref{eq:pde}. Due to this computational challenge, TO at high resolutions, i.e., over many voxels, can take hours, even days~\cite{aage2015topology}. This inspired researchers to develop DL-based TO methods to reduce or eliminate the need to solve PDEs.

\subsection{Neural networks for topology optimization}
One can broadly classify advances in TO using DL into three categories~\cite{zhang2021tonr}.

\textbf{Reduce or eliminate SIMP iterations:} The first serious attempts of TO via DL aimed to reduce the number of SIMP iterations~\cite{banga20183d, sosnovik2019neural, xue2021efficient}. In 2017, Sosnovik et al.~\cite{sosnovik2019neural} interpreted two-dimensional TO problems as image-to-image regression problems and were the first to apply \textit{convolutional neural networks} (CNNs) to TO. Following well-known image processing approaches, they trained a UNet model~\cite{ronneberger2015u} to map from intermediate SIMP iterations to the final structure. In 2018, Banga et al.~\cite{banga20183d} transferred these ideas to the three-dimensional case. Xue et al.~\cite{xue2021efficient} made each SIMP iteration cheaper by running it on a coarse resolution. They then applied a DL-based super-resolution method to increase the structure's final granularity. In 2020, Abueidda et al.~\cite{abueidda2020topology} were the first to use \textit{residual neural networks} (ResNets)~\cite{zhang2018road} and to consider two-dimensional nonlinear elasticity.

In 2019, Yu et al.~\cite{yu2019deep} developed the first end-to-end learning routine that directly predicts the final density without performing any SIMP iterations. They also created a generative framework to increase the resolution of their predicted designs. This formed the basis for a series of publications \cite{nie2021topologygan,oh2019deep,rawat2019novel,shen2019new} on \textit{generative adversarial network} (GAN)-based TO algorithms~\cite{goodfellow2014generative}.

Compared to previous research, Nie et al.~\cite{nie2021topologygan} and Zhang et al.~\cite{zhang2019deep} achieved a better generalization by not directly giving the network the boundary conditions as input. Instead, they passed displacements and von Mises stresses into the network. They argue that neural networks have difficulties extending to previously unseen boundary conditions if the input data is very sparse since the high sparsity of the input matrices leads to high variance of the mapping function.

\textbf{Substitute SIMP's PDE solver:} These methods aim to remove classical PDE solvers from the SIMP algorithm, removing its primary bottleneck. Qian et al.~\cite{qian2021accelerating} proposed a dual-model neural network using a forward model to compute the compliance of the structure and an adjoint model to determine the derivatives with respect to the density of each voxel. Similarly, Chi et al.~\cite{chi2021universal} and Lee et al.~\cite{lee2020cnn} used neural networks to replace the gradient and objective function computation.

\textbf{Neural reparameterization:} Using implicit neural representation for complex signals is an ongoing research topic in computer vision~\cite{chen2021nerv} and engineering~\cite{mrowca2018flexible}. Several TO publications~\cite{deng2020topology,hoyer2019neural,zehnder2021ntopo,zhang2021tonr} feature neural networks to reparameterize the TO's density field. While Hoyer et al.~\cite{hoyer2019neural} mapped latent vectors to discrete grid densities, Chandrasekhar \& Suresh~\cite{chandrasekhar2021tounn} used multilayer perceptrons to learn a continuous mapping from spatial locations to density values. Since these models are mesh-independent, they can represent the density function at arbitrary resolutions. However, one usually still requires PDE evaluations for training, which is computationally demanding.

While most DL approaches to TO are entirely oblivious to the underlying physics, there is a small number of approaches that have started to include physics-inspired properties into the training process: Banga et al.~\cite{banga20183d} and Rade et al.~\cite{rade2020physics} augmented the dataset by including rotations and mirrors of given loads and boundary conditions to encourage equivariance. Nie et al.~\cite{nie2021topologygan} and Zhang et al.~\cite{zhang2018road} involved physics by feeding the network strain and stress information as inputs. Cang et al.~\cite{cang2019one} and Zhang et al.~\cite{zhang2021tonr} introduced physics via the loss function design, comparable to \textit{physics-informed neural networks} (PINNs)~\cite{raissi2019physics}.

To our knowledge, no literature incorporated the problem's underlying physics by modifying the architecture of the DL model itself. We demonstrate that we can dramatically improve \textit{sample efficiency}, i.e., the model's performance when trained on a few training samples, and facilitate geometric reasoning by restricting the hypothesis space to group equivariant models. Cohen \& Welling~\cite{cohen2016group} introduced the first group equivariant CNNs in 2016. Nowadays, their application ranges from chemistry~\cite{weiler20183d} and physics~\cite{bogatskiy2020lorentz} to a wide range of tasks in geometric DL~\cite{thomas2018tensor,gerken2021geometric}. For image processing tasks, Dumont et al.~\cite{dumont2018robustness} showed that enforcing relevant equivariances can improve generalization performance and 
\section{Datasets}
\label{3_sec:datasets}
We find a substantial lack in the availability of public three-dimensional TO datasets, i.e., to the best of our knowledge there is none. This is problematic for several reasons. First, it is hard to reproduce other published results. Second, it requires each researcher to generate their own datasets, which is time-consuming and computationally demanding. Lastly, the lack of established TO datasets impedes the comparability of results throughout the community. To alleviate this issue, we publish two three-dimensional TO datasets containing samples of mounting brackets, which we call \textit{disc dataset} and \textit{sphere dataset}, refering to the shape of their respective design spaces. Both datasets are publicly available at \url{https://doi.org/10.5281/zenodo.7034898}~\cite{selto_dataset}. 

Each dataset consists of TO problems and associated ground truth density distributions. We generated the samples in cooperation with the  \href{https://www.ariane.group/en/}{ArianeGroup} and \href{https://www.synera.io/}{Synera} using the  \href{https://www.altair.com/optistruct/}{Altair OptiStruct} implementation of SIMP within the Synera software. The ArianeGroup designed the mounting brackets in the datasets to be of practical use, though real-world aerospace applications would require more complex load cases. The samples are discretized on a $n_x\times n_y\times n_z$ voxel grid, where the choice of $n_x,n_y,n_z\in \mathbb{N}$ varies depending on the dataset. Both datasets have fixed Dirichlet boundary conditions but variable force positions and magnitudes. 
\newpage
One can uniquely characterize each TO problem via the following properties:
\vspace{2mm}
\begin{enumerate}
    \item The number of voxels $(n_x, n_y, n_z)$ and the voxel size in millimeters in each direction.
    \item Material properties, given by Young's modulus $E$, Poisson's ratio $\nu$ and a yield stress criterion $\sigma_{\text{ys}}$. We choose $E=70$ GPa, $\nu=0.3$ and $\sigma_{\text{ys}}=450$ MPa for both datasets.
    \item A binary ($3\times n_x\times n_y \times n_z$)-tensor $\omega_{\text{Dirichlet}}$ to encode the presence of directional homogeneous Dirichlet boundary conditions for every voxel. $1$s indicate the presence, and $0$s the absence of homogeneous Dirichlet boundary conditions.
    \item A real-valued ($3\times n_x\times n_y\times n_z$)-tensor $F$ to encode external forces, given in $\text{N}/\text{m}^3$. The three channels correspond to the force magnitudes in each spacial dimension.
    \item A ($1\times n_x\times n_y\times n_z$)-tensor $\omega_\text{design}$ containing values $\in\lbrace -1, 0,1\rbrace$ to encode design space information. We use $0$s and $1$s to constrain voxel densities to be $0$ or $1$, respectively. Entries of $-1$s indicate a lack of density constraints, which signifies that the density in that voxel can be freely optimized. This naturally defines the voxel sets $\Omega_{-1}, \Omega_0,$ and $\Omega_1$. For voxels that have Dirichlet boundary conditions or loads assigned to them we enforce the density value to be $1$ by setting $\omega_\text{design} = 1$.
\end{enumerate}
\vspace{2mm}
All tensors are defined voxel-wise, including $\omega_\text{Dirichlet}$ and $F$. This makes our datasets easy to use in DL applications as it allows for a shape-consistent tensor representation.

The SIMP method does not always provide a physically plausible solution for a TO problem, i.e., some solutions break under their load cases. Therefore, we clean both datasets after the dataset generation process by rejecting failing samples. This leaves a total count of almost \num{10000} problem-ground truth pairs. Both datasets can be split into subsets with load cases of one or two points of attack. We call these subsets \textit{simple} and \textit{complex}, respectively. We refer to the combination of the simple and complex dataset as the \textit{disc combined} and \textit{sphere combined} dataset.

\begin{figure}
    \centering
    \begin{subfigure}{0.22\textwidth}
    \centering
        \includegraphics[width=0.9\textwidth]{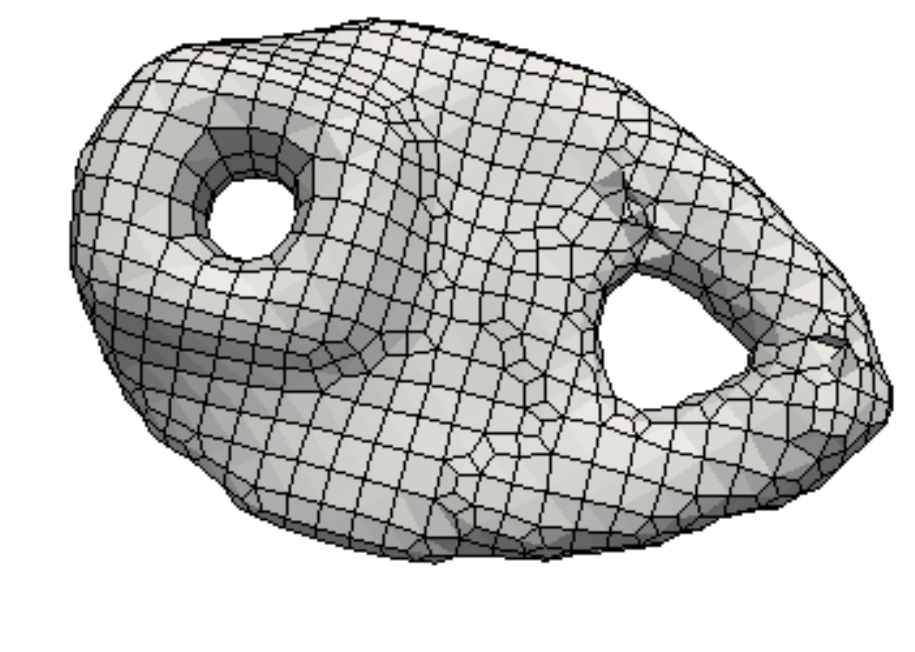}
        \subcaption{disc simple}
        \label{3_fig:samples_disc_simple}
    \end{subfigure}
    \begin{subfigure}{0.22\textwidth}
    \centering
        \includegraphics[width=0.7\textwidth]{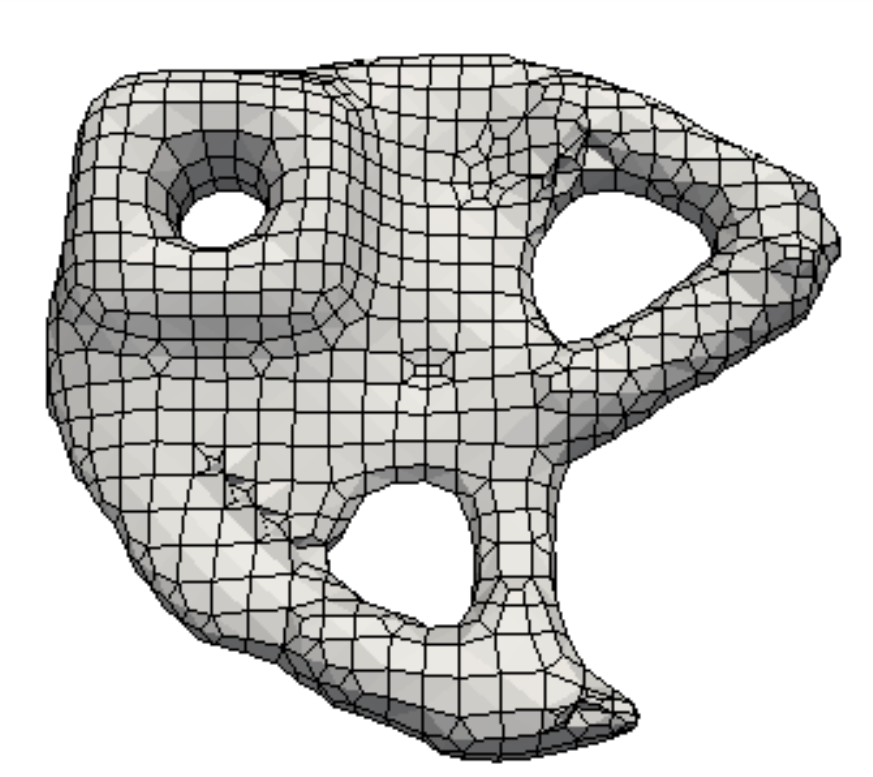}
        \subcaption{disc complex}
        \label{3_fig:samples_disc_complex}
    \end{subfigure}
    \begin{subfigure}{0.22\textwidth}
    \centering
        \includegraphics[width=0.7\textwidth]{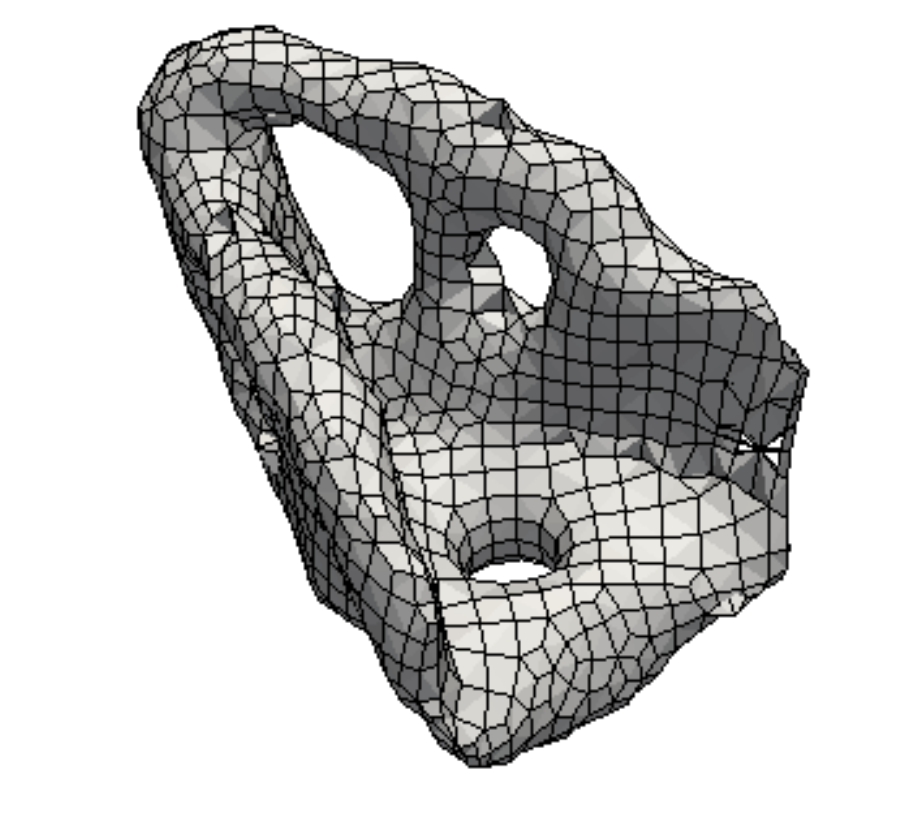}
        \subcaption{sphere simple}
        \label{3_fig:samples_sphere_simple}
    \end{subfigure}
    \begin{subfigure}{0.24\textwidth}
    \centering
        \includegraphics[width=0.9\textwidth]{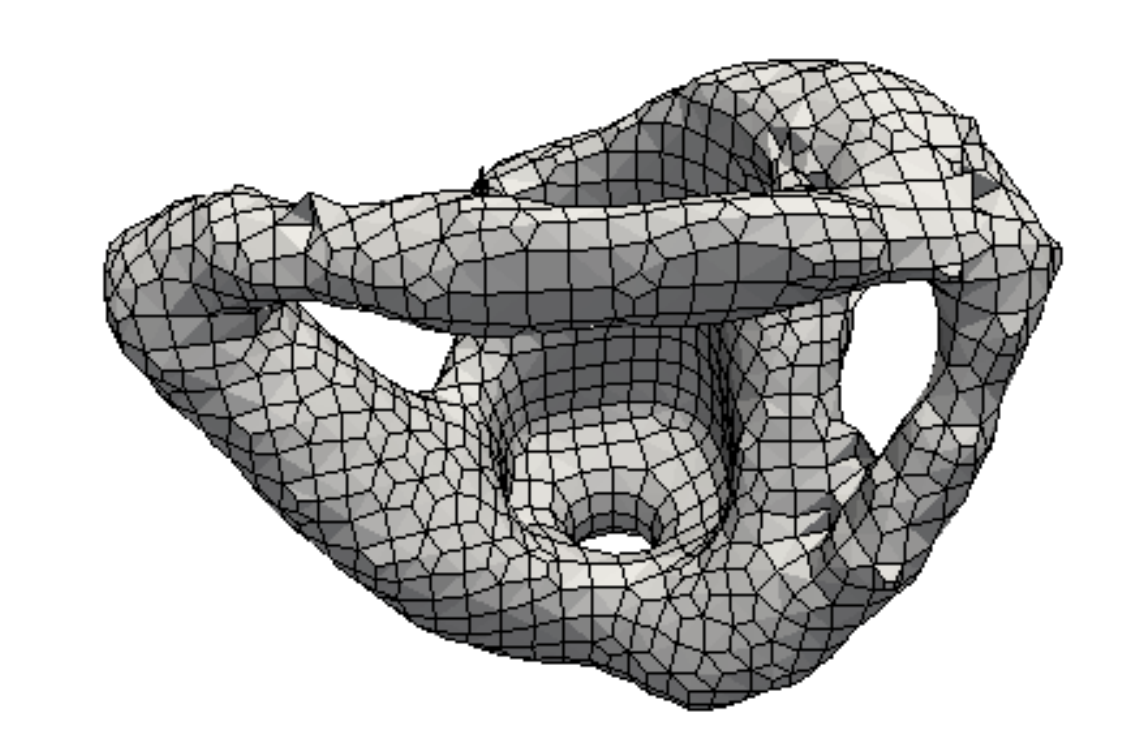}
        \subcaption{sphere complex}
        \label{3_fig:samples_sphere_complex}
    \end{subfigure}
    \caption{Ground truth examples from the disc and sphere dataset~\cite{selto_dataset}. The densities are defined on a voxel grid and are smoothed for visualization purposes using Taubin smoothing~\cite{taubin1995curve}.}
    \label{3_fig:samples}
\end{figure}

See~\cref{tab_selto_datasets} for an overview of both datasets and~\cref{3_fig:samples} for ground truth examples. More samples can be found in the \textit{ground truth}-columns of~\cref{appendix_b_samples}, where we present a total of $40$ randomly chosen samples.

\begin{table}[]
\centering
\footnotesize
\begin{tabular}{|c|c|c|c|c|c|c|}
\hline
dataset                                                                     & \# samples    & shape                                                                      & subsets & \# loads & \begin{tabular}[c]{@{}c@{}}\# training\\ samples\end{tabular} & \begin{tabular}[c]{@{}c@{}}\# validation\\ samples\end{tabular} \\ \hline \hline
\multirow{2}{*}{\begin{tabular}[c]{@{}c@{}}disc \\ (combined)\end{tabular}}  & \multirow{2}{*}{9246} & \multirow{2}{*}{\begin{tabular}[c]{@{}c@{}}$39\times 39\times 21$\\ voxels\end{tabular}}  & simple  & 1             & 1509                & 200                   \\ \cline{4-7} 
                                                                             &                       &                                                                            & complex & 2             & 7337                & 200                   \\ \hline\hline
\multirow{2}{*}{\begin{tabular}[c]{@{}c@{}}sphere\\ (combined)\end{tabular}} & \multirow{2}{*}{602}  & \multirow{2}{*}{\begin{tabular}[c]{@{}c@{}}$39\times 39\times 21$\\ voxels\end{tabular}} & simple  & 1             & 150                 & 36                    \\ \cline{4-7} 
                                                                             &                       &                                                                            & complex & 2             & 380                 & 36                    \\ \hline
\end{tabular}
\caption{Overview of our datasets, called \textit{disc} and \textit{sphere}, with the names referring to the shape of their design spaces. Both datasets can be split into subsets with load cases of one or two points of attack. We call these subsets \textit{simple} and \textit{complex}, respectively. When we refer to the mixed datasets, i.e., the combination of simple and complex, we sometimes call these \textit{disc combined} and \textit{sphere combined}.}
\label{tab_selto_datasets}
\end{table}

\section{Methods}
\label{section_methods}
This section introduces the DL pipelines we evaluate and analyze in \Cref{sec_numerical_experiments}. We examine different input preprocessing strategies and the effects of physically motivated group averaging~\cite{puny2021frame}. We use an end-to-end learning approach, i.e., train a neural network $f$ to map preprocessed TO problems to a optimized density distributions provided by the datasets.

\subsection{Preprocessings}
\label{section_preprocessings}
\begin{figure}
    \centering
    \includegraphics[width=0.6\textwidth]{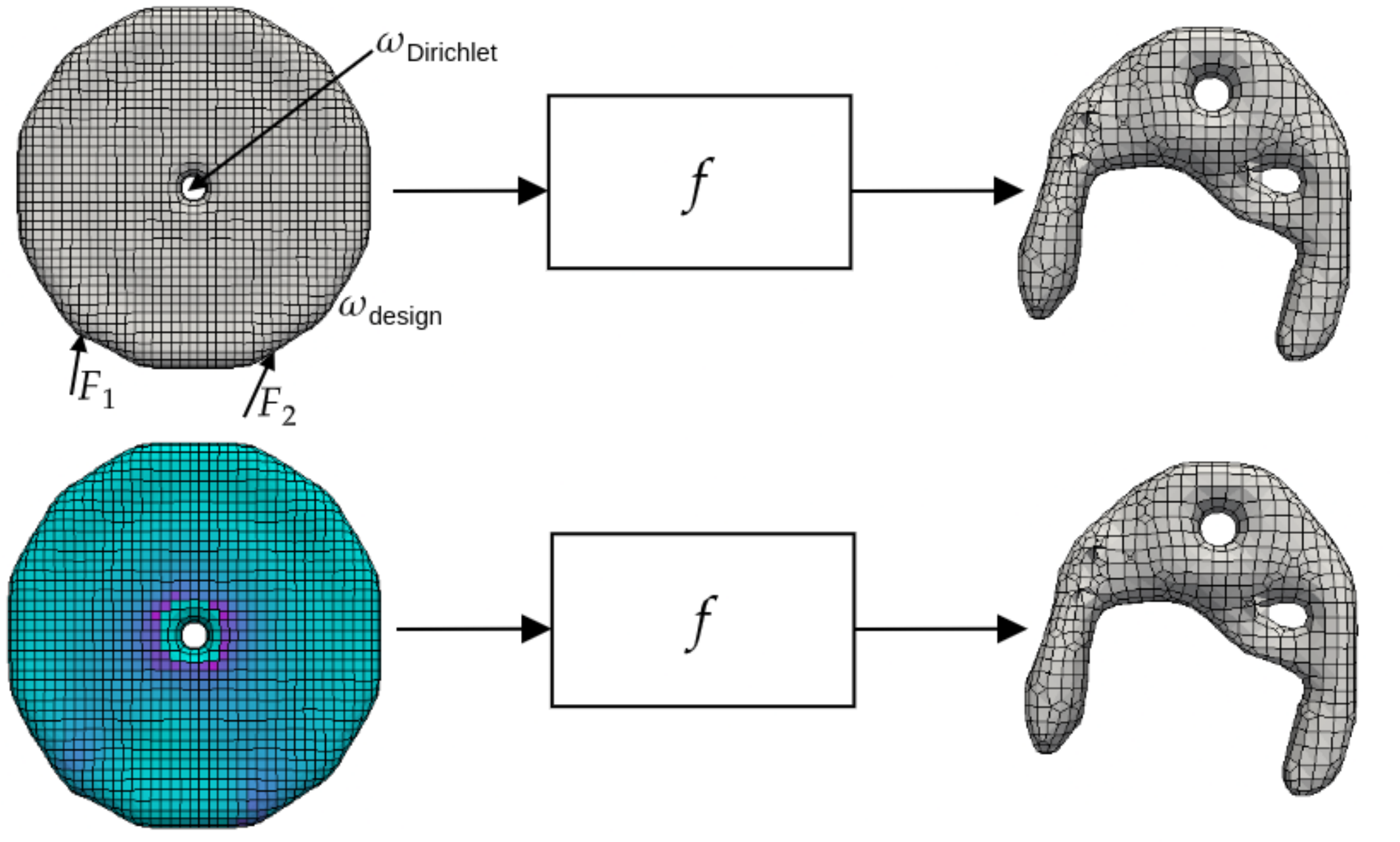}
    \caption{Illustration of our model pipeline, using an end-to-end learning approach, where the function $f$ represents the neural network. Top: Trivial preprocessing. We concatenate $\omega_\text{Dirichlet}$, $\omega_\text{design}$, and forces $F=F_1\cup F_2$ into a $7$-channel tensor, which after normalization, we use as input for $f$. Bottom: PDE preprocessing. We compute an initial density distribution $\rho_\text{init}$ and corresponding initial displacements and stresses. The image displays the initial von Mises stress distribution, which is a $1$-channel tensor.}
    \label{4_fig:pipeline}
\end{figure}

Choosing a suitable input preprocessing strategy is crucial for DL. We now present the two main preprocessings we use in this paper. It is possible to combine these via simple channel-wise concatenation of their outputs.
\vspace{2mm}
\begin{enumerate}
    \item \textbf{Trivial preprocessing.} The input of the neural network is a $7$-channel tensor which results from the channel-wise concatenation of Dirichlet boundary conditions $\omega_\text{Dirichlet}$, loads $F$, and design space information $\omega_\text{design}$. Additionally, we normalize each sample's $F$ via the mean $\| F \|_\infty$ over all training samples. This is arguably the most straightforward and intuitive type of preprocessing for learned end-to-end TO.
    \item \textbf{PDE preprocessing.} As proposed by Zhang et al. (2019)~\cite{zhang2019deep}, we first define an \textit{initial density distribution} $\rho_\text{init}$ that is $1$ on $\Omega_1$ and $\Omega_{-1}$, and $0$ on $\Omega_0$. For $\rho_\text{init}$, we then compute the \textit{initial von Mises stresses}, which we obtain by solving the PDE for linear elasticity. We normalize the resulting tensor analogously to the normalization of $F$ above. These initial von Mises stresses are then used as a $1$-channel input to the neural network. Analogously, it would also be possible to use the full initial stress tensor or the initial displacements as network input.
    \end{enumerate}
   \vspace{2mm} 
We illustrate both preprocessing strategies and our model pipeline in \cref{4_fig:pipeline}.

\subsection{Architecture}
We choose a UNet~\cite{ronneberger2015u} as our neural network architecture, which is a convolutional encoder-decoder network. The encoder consists of repeated application of convolutions, each followed by a rectified linear unit (ReLU) activation function and a max pooling operation. During the encoding, the encoder of the UNet reduces the spatial information while it increases the feature information. The decoder then extends the feature and spatial information through convolution and upsampling steps and concatenations with high-resolution features from the encoder. See \cref{appendix_a_network} for more details on the UNet architecture.
\newpage
\subsection{Equivariance}
\begin{figure}
    \centering
    \includegraphics[width=0.55\textwidth]{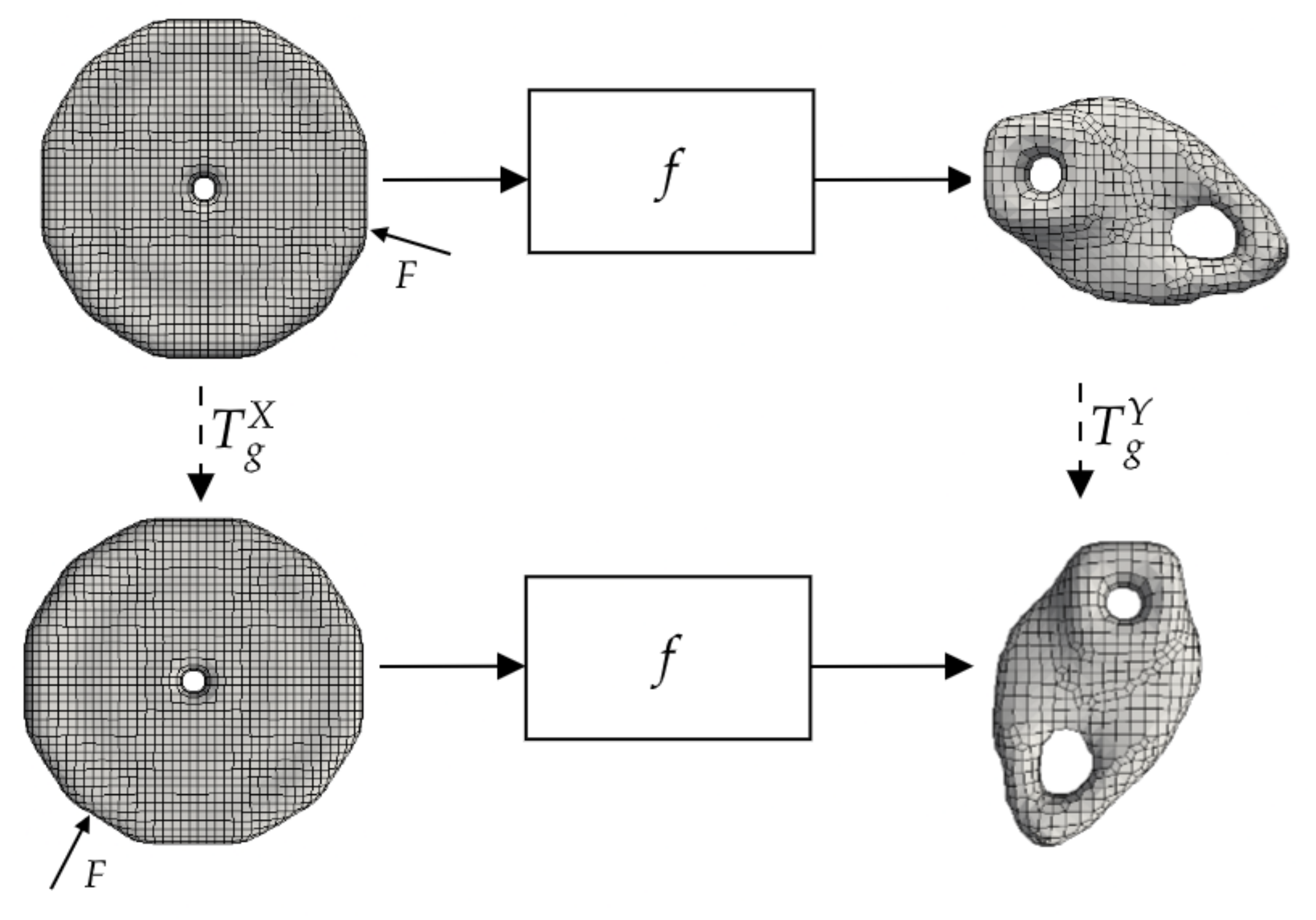}
    \includegraphics[width=0.43\textwidth]{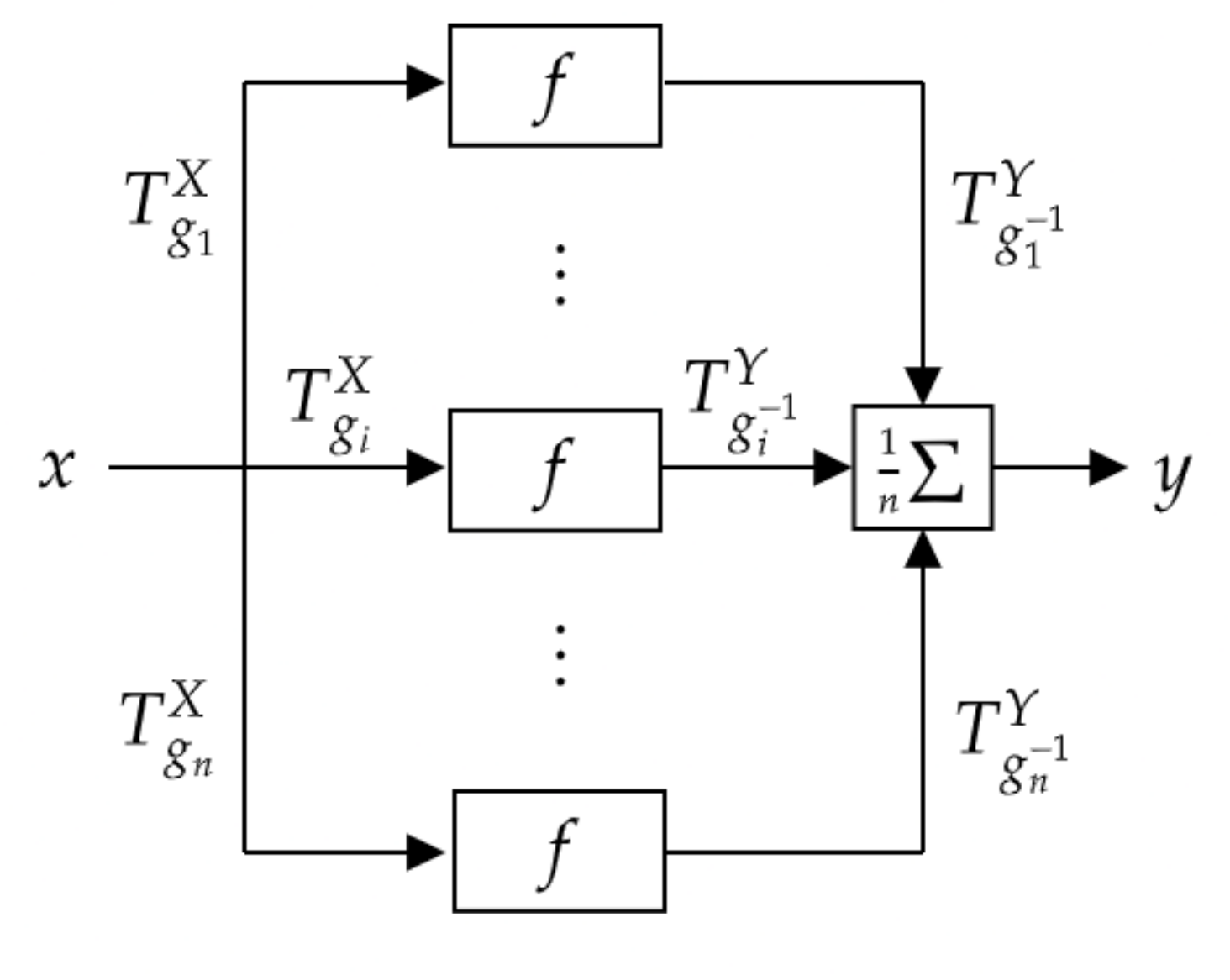}
    \caption{Left: If $f:X\rightarrow Y$ is $G$-equivariant, then for each $g\in G$ applying $f \circ T^X_g$ is equivalent to $T_g^Y \circ f$. Here, $g$ is represented by a 90° clockwise rotation. Right: Illustration of the equivariance wrapper $F_G^f$. We transform the input $x$ via group actions $T_{g_i}^X$ and feed the transformed data into the model $f$. We then reverse the initial transformation on the outputs via $T^Y_{g_i^{-1}}$ and aggregate the results. This process is called \textit{group averaging}. The choice of the (finite) transformation group $G$ is entirely free.}
    \label{4_fig:equivariance}
\end{figure}

Equivariance is the property of a function to commute with the actions of a symmetry group acting on its domain and range. For a given transformation group $G$, we say that a function $f:X\rightarrow Y$ is ($G$-)equivariant if
\begin{align}
    &f(T^X_g(x)) = T^Y_g(f(x)) &\forall g\in G, \, x\in X,
    \label{4_eq:equivariance}
\end{align}
where $T^{X}_g$ and $T^{Y}_g$ denote linear \textit{group actions} in the corresponding spaces $X$ and $Y$ \cite{cohen2016group}. That is, transforming an input $x$ by a transformation $g$ and then passing it through $f$ should give the same result as first mapping $x$ through $f$ and then transforming the output $f(x)$ (see left image of \cref{4_fig:equivariance}). In many machine learning tasks, we possess prior knowledge about equivariances our predictor should have. Including such knowledge directly into the model can significantly facilitate learning by freeing up model capacity for other factors of variation \cite{weiler2018learning}. Since $G$ is a group, it also contains the identity transformation and a unique inverse transformation $g^{-1}$ for each $g\in G$. Therefore, we can reformulate \eqref{4_eq:equivariance} as
\begin{align*}
    f(x) = T^Y_{g^{-1}} \left[ f\left(T_g^X(x)\right)\right],
\end{align*}
allowing the implementation of equivariance via \textit{group averaging}~\cite{murphy2018janossy, puny2021frame} by defining an equivariance wrapper $F_G^f$ as 
\begin{align*}
    F_G^f(x) := \frac{1}{| G|} \sum_{g\in G} T_{g^{-1}}^Y \left[ f(T_g^X(x))\right].
\end{align*}
$F_G^f$ is equivariant with respect to $G$ since for each $h\in G$ it holds that
\begin{align*}
    T_{h^{-1}}^Y \left[F_G^f\left(T_h^X(x)\right)\right] &= \frac{1}{| G|} \sum_{g\in G} T_{(gh)^{-1}}^Y \left[f\left(T_{gh}^X(x)\right)\right] \\
    &= \frac{1}{| G|} \sum_{gh\in G} T_{g^{-1}}^Y \left[f\left(T_{g}^X(x)\right)\right] \\
    &= \frac{1}{| G|} \sum_{g\in G} T_{g^{-1}}^Y \left[f\left(T_{g}^X(x)\right)\right] \\
    &= F_G^f(x).
\end{align*}
For an illustration of $F_G^f$ see the right image of \cref{4_fig:equivariance}.

The most popular approach to enforcing group equivariance in DL is enforcing invariance of the network's convolutional filters and activation functions~\cite{cohen2016group}. However, performing group averaging has several advantages. First, it is straightforward to implement. Second, its plug-and-play nature allows the trivial application to any (finite) transformation group $G$ and any model $f$. In particular, the applicability to any $f$ allows for direct comparisons between any non-equivariant model and its equivariant counterpart.

Natural choices of transformation groups for our applications are the dihedral symmetry group $D_4$ and the octahedral symmetry group $O_h$, which describe all combinations of 90° rotations and reflections in $\mathbb{R}^2$ and $\mathbb{R}^3$, respectively. Thus the equivariance wrapper applies 8 transformations in the $D_4$ and 48 in the $O_h$ case. Note that group actions on vector fields affect spacial locations and vector directions, unlike on scalar fields. The necessity to account for vector directions makes equivariance wrappers especially suited and flexible.

We want to clarify that the equivariance wrapper fundamentally differs from \textit{data augmentation}, i.e., augmenting the dataset with rotations and reflections. Augmentation averages over the objective, whereas the wrapper averages over the model. This difference endows the wrapper with several significant advantages, most notably: The model does not need to learn about equivariance as it is hard-coded. Consequently, the wrapper is not only approximately equivariant but exactly -- importantly, this holds for any input, even for inputs drastically different from the training data.
\section{Numerical experiments}
\label{sec_numerical_experiments}

In this section, we conduct several numerical experiments to illustrate the effectiveness of adding physics-based information via PDE preprocessing and equivariance to our DL models.
\subsection{Training}
We train and compare different combinations of preprocessings described in \Cref{section_methods}, each with and without equivariance. Due to the reduced number of voxels in the $z$-direction and for simplicity, we chose the dihedral symmetry group, $D_4$, as the transformation group for all our experiments. All models are implemented in PyTorch~\cite{paszke2017automatic}. We determine the batch sizes individually, depending on the memory capacity and comparability (see \cref{appendix_a_network}). We choose the weighted \textit{binary cross-entropy} (BCE) as our loss function, and calculate the weighting factor based on the training dataset. We use the Adam optimizer~\cite{kingma2014adam} with a learning rate of $10^{-3}$ in all our experiments.

We train all models until their improvement on the validation set stalled for $100$~epochs and then pick the model corresponding to the best validation epoch; this stopping criterion results in models trained for $100$ to $1000$ epochs. 

We also experimented with different network architectures, transformation groups and preprocessing strategies. However, they led to worse performances than the ones presented here. Regardless, we think that these experiments can still be of interest for the research community (see~\cref{appendix_ablations}).

\subsection{Evaluation}
We now discuss our evaluation methodology for the analysis of our DL models. In particular, we want to compare the impact of different preprocessing strategies and equivariances on our models. We begin by defining our evaluation criteria.

\subsubsection{Evaluation criteria}
We now introduce the criteria we use to evaluate our models over the validation datasets. Before we apply the criteria we first binarize the densities produced by the DL models, such that they only contain $0$s and $1$s. We make use of the following two criteria:
\vspace{2mm}
\begin{itemize}
\item \textbf{IoU:} In contrast to our loss, the BCE, which is distribution-based, the \textit{Intersection over Union} (IoU) is region-based. It is defined as
\begin{align*}
    \text{IoU} = \frac{\text{TP}}{\text{TP} + \text{FN} + \text{FP}},
\end{align*}
where TP, FN and FP denote the number of true positive, false negative and false positive voxel predictions. We limit the evaluation of the IoU to the editable design space $\Omega_{-1}$. Following Goodhart's law~\cite{goodhart1975problems} and established practices in the segmentation community, we use the IoU as our primary evaluation metric~\cite{ma2020segmentation, ronneberger2015u}.
\item \textbf{Fail percentage:} We consider a prediction failed if the von Mises stress in any voxel exceeds the yield stress by more than 10\% or if any voxel with a load case does not connect to one containing Dirichlet conditions. The fail percentage is the fraction of failed parts. This criterion makes us the first in the DL for TO community who verify their predictions' mechanical integrity.
\end{itemize}
\vspace{2mm}
A typical evaluation criterion in the DL for TO literature is the (balanced) accuracy metric. However, we found that in most cases, IoU yields comparable values but, in general, reflects the quality of its input more appropriately when used across different datasets. Given that we can interpret the binary densities of our structures as segmentation masks, IoU's applicability is not surprising as it is the most common metric for semantic segmentation. Therefore we will use IoU as our main evaluation criterion.

\subsubsection{Sample efficiency}
One of the main impediments to learned supervised TO is the high cost associated with generating new datasets. The necessity to generate thousands of problems with corresponding ground truth solutions often hinders practical applicability in real-world situations since the generation of large datasets can take days, even weeks. Consequently, reducing the number of required training samples is highly beneficial, e.g., by modifying the DL model design. Therefore, we put a particular emphasis on the visualization and measurement of our models' \textit{sample efficiency}, i.e., the model's performance when trained on few training samples. 

We visualize the sample efficiency of a model with what we call \textit{sample efficiency curves} (SE curves). For each SE curve, we train separate instances of a given model setup on subsets of the original training dataset of varying sizes. We then evaluate and compare the performance of these models on a fixed validation dataset, using IoU and fail percentage as our evaluation criteria. For the disc dataset, we choose training subsets of sizes $2$, $4$, $10$, $50$, $100$, $150$, $250$, $500$, $1000$, and $1500$. For the sphere dataset we train on $2$, $4$, $10$, $50$, $100$, and $150$ samples. This way, we obtain an individual SE~curve for each evaluation criterion. In order to compare different SE curves of the same criterion, we report two metrics: 
\vspace{2mm}
\begin{enumerate}
    \item The normalized \textit{area under the curve} (AUC) of that criterion up to a training sample size of $150$, denoted by $\text{AUC}_{150}$. We use this metric to quantify sample efficiency.
    \item The \textit{final score}, which is the value of that criterion achieved by the model trained on the largest training subset.
\end{enumerate}

\subsection{Results}
This section gives an overview of our numerical results and compares the performance of different models. We begin with our main results, followed by an analysis of our models' generalization capabilities. Finally, we discuss model alterations that did not yield improvements but might still be helpful for further understanding and research.

\subsubsection{Main results}
\label{sec_main_results}
\begin{figure}[t]
    \centering
    \includegraphics[trim={0cm 19.3cm 0cm 0cm},clip, width=0.7\textwidth]{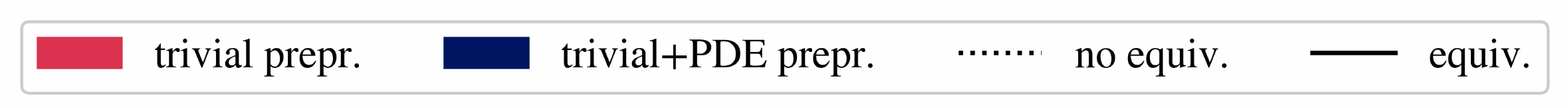}
     \begin{subfigure}[b]{0.49\textwidth}
         \centering
         \includegraphics[width=\textwidth]{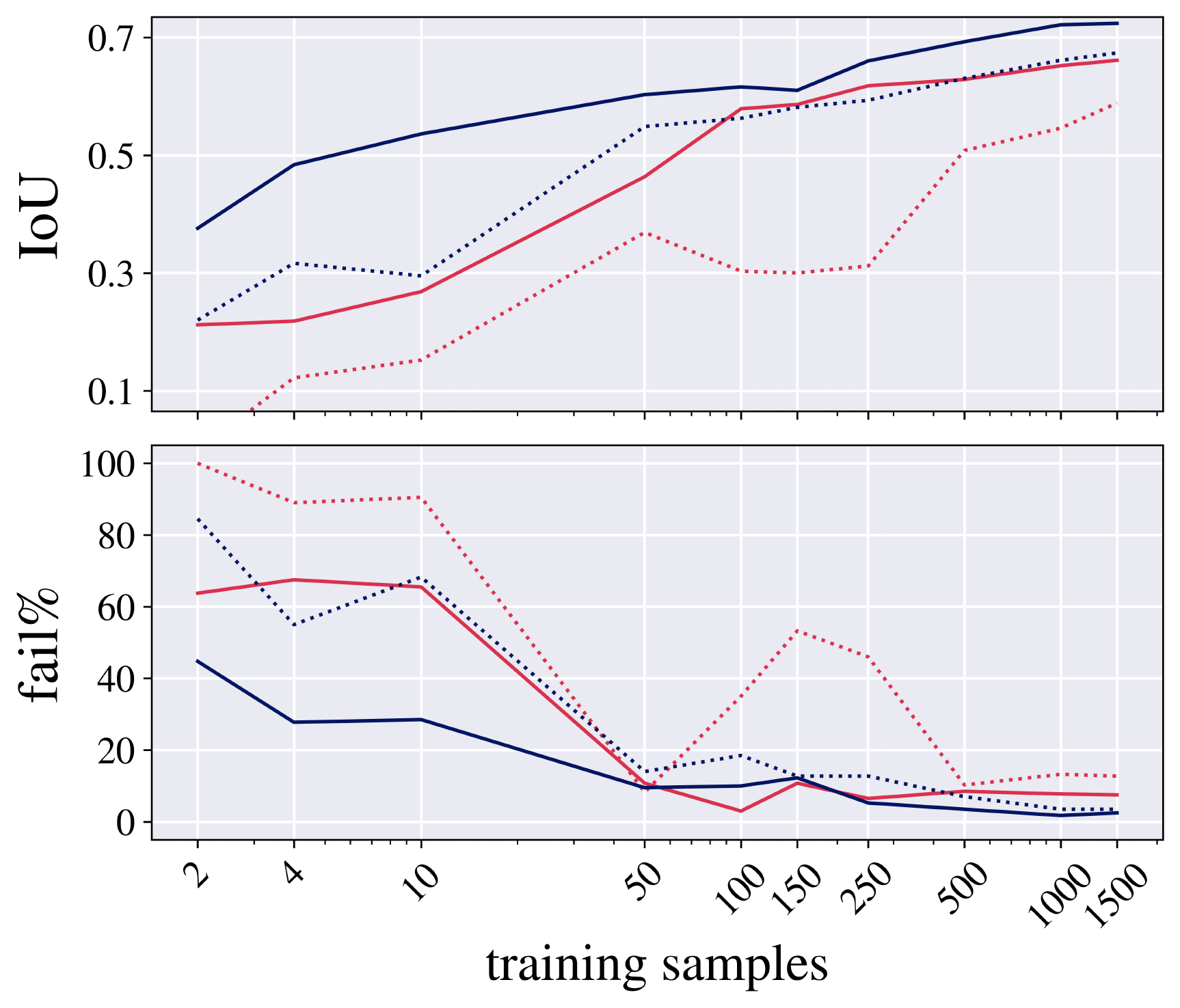}
         \caption{disc combined}
     \label{5_fig:se_curve_disc_combined}
     \end{subfigure}
     \hfill
     \begin{subfigure}[b]{0.49\textwidth}
         \centering
         \includegraphics[width=\textwidth]{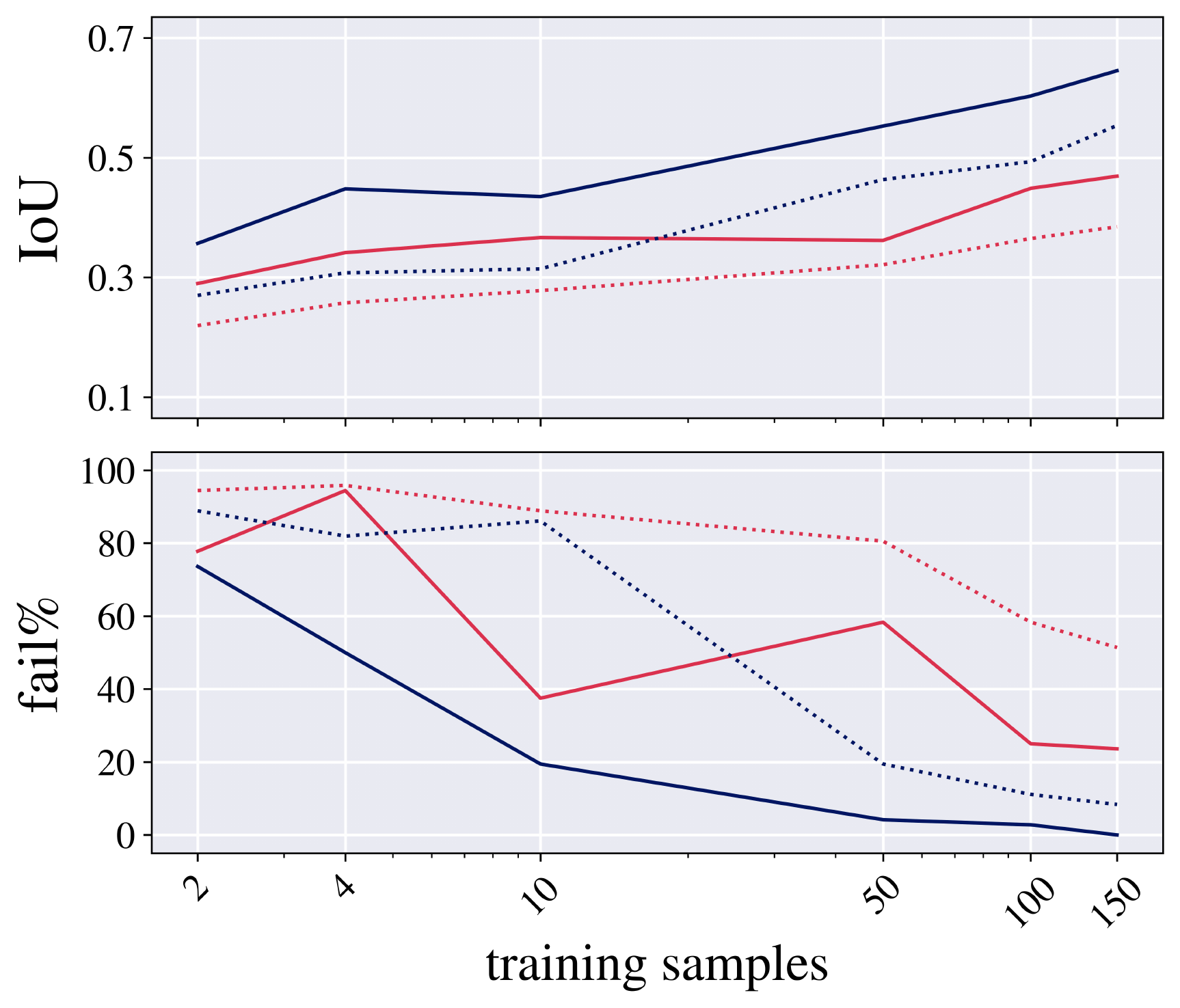}
         \caption{sphere combined}
    \label{5_fig:se_curve_sphere_combined}
    \end{subfigure}
    \caption{Sample efficiency curves, trained and evaluated on disc combined (left) and sphere combined (right). The x-axis shows the size of the training dataset for the different models on a logarithmic scale. On the $y$-axis we see performance of the criteria IoU and fail percentage. For each criterion we evaluate models using trivial preprocessing (red), trivial+PDE preprocessing (blue), equivariance (solid line) and no equivariance (dashed line).}
    \label{5_fig:se_curves}
\end{figure}

As expected, model performance tends to improve with increased training data. We also observe dramatic boosts in the UNet's performance when incorporating physics via equivariance and trivial+PDE preprocessing. These improvements are especially visible for low numbers of training samples; see the SE curves in \cref{5_fig:se_curves} and $\text{AUC}_{150}$ scores in \cref{5_tab:auc_scores_x_eval}. From \cref{5_fig:se_curves}, we observe that using trivial+PDE preprocessing and equivariance in combination leads to a reduction in the required training samples by \textit{two orders of magnitude} while maintaining equivalent IoU scores. Additionally, we notice a reduction in the fail percentage to almost $0\%$.

The improvements are also evident when visually comparing ground truths with predictions; see \cref{5_fig:disc_simple_predictions,5_fig:disc_complex_predictions,5_fig:sphere_simple_predictions,5_fig:sphere_complex_predictions}. Further, we observe that adding PDE preprocessing and equivariance leads to predictions closer to the ground truth densities, which is especially noticeable on small training sets. Moreover, applying our equivariance wrapper reduces the necessary number of epochs but increases overall training duration, see \cref{7_tab:nr_parameters} in \cref{appendix_a_network}.

For an unbiased impression of our model predictions and to avoid cherry-picking, we display 20 random samples and predictions from both datasets in \cref{7_tab:random_samples_disc,7_tab:random_samples_sphere} in \cref{appendix_b_samples}. We show two predictions for each sample, one using trivial preprocessing only and one using trival+PDE preprocessing and equivariance. We trained the UNets on 1500 and 150 samples for disc and sphere, respectively.

\subsubsection{Generalization}
As in all machine learning, generalization is also a problem in its application to TO. The problem led to some publications addressing the issue~\cite{nie2021topologygan, zhang2019deep}. We quantify our models' generalizability by cross-evaluating them on different data subsets not used during training, i.e., we evaluate all models trained on either disc simple, disc complex, or disc combined on each of the others. We proceed analogously for the sphere models and datasets.

We compare the $\text{AUC}_{150}$ and final scores for each model based on the corresponding SE curve. For both IoU and fail percentage, we present the results for the $\text{AUC}_{150}$ score in~\cref{5_tab:auc_scores_x_eval} and the final score in~\cref{5_tab:final_scores_x_eval}. We observe that the addition of PDE preprocessing and equivariance improves the generalization capabilities of our models considerably for both IoU and fail percentage, especially on small training sets.

\section{Conclusion}
We aimed to provide a strong foundation for future research in DL for TO. On the one hand, we proposed and analyzed basic components and principles for designing DL pipelines for TO. On the other hand, we provide two TO datasets enabling the training and comparability of DL methods.

More specifically, method-wise, we focused on physical correctness and sample-efficiency. In practice, the existence of appropriate large-scale training data is costly or simply not realistic; hence it is our conviction that sample-efficiency is a crucial component of any future DL approach to TO.

To achieve this, we developed a physics-inspired approach. We conducted a large-scale ablation study to prove the effectiveness of the two critical components of our approach -- on the one hand, PDE-based preprocessing and, on the other, mirror and rotation equivariant architectures. These two key elements drastically improve the overall predictions' physical correctness and the sample efficiency, i.e., the model performance when trained on only a few samples.

On the data side, we publish two three-dimensional TO datasets with a total count of almost \num{10000} problem-ground truth pairs. To our knowledge, these are the first publicly available datasets for TO.
\begin{table}[h]
    \caption{Tables containing $\text{AUC}_{150}$ scores for different combinations of preprocessings and equivariances. We (cross-)evaluate performance of models trained on the disc and sphere datasets and highlight the row-wise best scores in bold.}
    \label{5_tab:auc_scores_x_eval}
    \centering
    \begin{subtable}{\textwidth}
    \centering
    \resizebox{!}{5cm}{\begin{tabularx}{\textwidth}{c *{7}{Y}}
\toprule
& &
 \multicolumn{2}{c}{trivial prepr.}  
 & \multicolumn{2}{c}{trivial+PDE prepr.}\\
\cmidrule(lr){3-4} \cmidrule(l){5-6}
eval on & train on & no equiv. & equiv. & no equiv. & equiv.\\
\midrule \midrule
& simple & 0.76 & 0.86 & 0.84 & \textbf{0.87}\\
disc simple & complex & 0.35 & 0.59 & 0.57 & \textbf{0.63}\\
& combined & 0.63 & 0.80 & 0.78 & \textbf{0.81}\\\midrule
& simple & 0.35  & 0.40 & 0.43 & \textbf{0.44}\\
disc complex & complex & 0.28 & 0.44 & 0.48 & \textbf{0.58}\\
& combined & 0.33 & 0.45 & 0.48 & \textbf{0.56}\\\midrule
& simple & 0.55 & 0.63 & 0.64 & \textbf{0.65}\\
disc combined & complex & 0.31 & 0.52 & 0.52 & \textbf{0.61}\\
& combined & 0.48 & 0.63 & 0.63 & \textbf{0.69}\\
\midrule \midrule
& simple & 0.28  & 0.41 &  0.50 &  \textbf{0.61}\\
sphere simple & complex & 0.28 & 0.32 & 0.39 & \textbf{0.44}\\
& combined & 0.30 & 0.37 & 0.43 & \textbf{0.56}\\\midrule
& simple & 0.20 & 0.27 & 0.29 & \textbf{0.38}\\
sphere complex & complex & 0.45 & 0.47 & 0.56 & \textbf{0.59}\\
& combined & 0.38 & 0.45 & 0.49 & \textbf{0.57}\\\midrule
& simple & 0.24  & 0.34 &  0.40 &  \textbf{0.49}\\
sphere combined & complex & 0.36 & 0.39 & 0.48 & \textbf{0.52}\\
& combined & 0.34 & 0.41 & 0.46 & \textbf{0.56}\\
\bottomrule
\end{tabularx}}
    \subcaption{$\text{AUC}_{150}$ scores for the IoU criterion. Higher values indicate better results.}
    \end{subtable}
    \vfill
    \begin{subtable}{\textwidth}
    \centering
    \resizebox{!}{5cm}{\begin{tabularx}{\textwidth}{c *{7}{Y}}
\toprule
& &
 \multicolumn{2}{c}{trivial prepr.}  
 & \multicolumn{2}{c}{trivial+PDE prepr.}\\
\cmidrule(lr){3-4} \cmidrule(l){5-6}
eval on & train on & no equiv. & equiv. & no equiv. & equiv.\\
\midrule \midrule
& simple & 1.71  & 0.10 &  0.33 &  \textbf{0.01}\\
disc simple & complex & 27.21 & 1.35 & 1.11 & \textbf{1.10}\\
& combined & 2.95 & 0.76 & 0.89 & \textbf{0.05}\\\midrule
& simple & 39.11 & \textbf{10.21} & 22.30 & 14.50\\
disc complex & complex & 49.39 & 16.13 & 9.37 & \textbf{5.07}\\
& combined & 37.58 & 17.07 & 14.95 & \textbf{8.47}\\\midrule
& simple & 20.41 & \textbf{5.16} & 16.31 & 12.26\\
disc combined & complex & 38.30 & 8.74 & 5.24 & \textbf{2.69}\\
& combined & 20.27 & 8.92 & 7.92 & \textbf{4.26}\\
\midrule \midrule
& simple & 48.03 & 12.16 & 9.55 & \textbf{4.11}\\
sphere simple & complex & 59.52 & 36.84 & 8.41 & \textbf{6.06}\\
& combined & 56.51 & 24.02 & 27.93 & \textbf{7.73}\\\midrule
& simple & 94.37 & 68.17 & 72.60 & \textbf{34.95}\\
sphere complex & complex & 63.53 & 37.44 & 14.62 & \textbf{3.55}\\
& combined & 83.33 & 54.13 & 26.61 & \textbf{6.42}\\\midrule
& simple & 71.20 & 40.17 & 41.08 & \textbf{19.53}\\
sphere combined & complex & 61.52 & 37.14 & 11.51 & \textbf{4.80}\\
& combined & 69.92 & 39.08 & 27.27 & \textbf{7.08}\\
\bottomrule
\end{tabularx}}
    \subcaption{$\text{AUC}_{150}$ scores for the fail\% criterion.}
    \end{subtable}   
\end{table}

\begin{table}[h]
    \caption{Tables containing the final scores for different combinations of preprocessings and equivariances. We (cross-)evaluate performance of models trained on the disc and sphere datasets and highlight the row-wise best scores in bold.}
    \label{5_tab:final_scores_x_eval}
    \centering
    \begin{subtable}{\textwidth}
    \centering
    \resizebox{!}{5cm}{\begin{tabularx}{\textwidth}{c *{7}{Y}}
\toprule
& &
 \multicolumn{2}{c}{trivial prepr.}  
 & \multicolumn{2}{c}{trivial+PDE prepr.}\\
\cmidrule(lr){3-4} \cmidrule(l){5-6}
eval on & train on & no equiv. & equiv. & no equiv. & equiv.\\
\midrule \midrule
& simple & 0.87 & 0.89 & 0.90 & \textbf{0.91}\\
disc simple & complex & 0.52 & 0.61 & 0.61 & \textbf{0.62}\\
& combined & 0.77 & 0.84 & 0.83 & \textbf{0.85}\\\midrule
& simple & 0.37 & 0.40 & 0.44 & \textbf{0.45}\\
disc complex & complex & 0.40 & 0.48 & 0.52 & \textbf{0.62}\\
& combined & 0.40 & 0.48 & 0.52 & \textbf{0.59}\\\midrule
& simple & 0.62 & 0.65 & 0.67 & \textbf{0.68}\\
disc combined & complex & 0.46 & 0.55 & 0.57 & \textbf{0.62}\\
& combined & 0.59 & 0.66 & 0.67 & \textbf{0.72}\\
\midrule \midrule
& simple & 0.26 & 0.46 & 0.58 & \textbf{0.67}\\
sphere simple & complex & 0.29 & 0.35 & 0.39 & \textbf{0.48}\\
& combined & 0.34 & 0.44 & 0.52 & \textbf{0.66}\\\midrule
& simple & 0.18 & 0.23 & 0.35 & \textbf{0.40}\\
sphere complex & complex & 0.46 & 0.55 & 0.64 & \textbf{0.67}\\
& combined & 0.43 & 0.50 & 0.58 & \textbf{0.63}\\\midrule
& simple & 0.22 & 0.34 & 0.46 & \textbf{0.53}\\
sphere combined & complex & 0.37 & 0.45 & 0.51 & \textbf{0.57}\\
& combined & 0.38 & 0.47 & 0.55 & \textbf{0.65}\\
\bottomrule
\end{tabularx}}
    \subcaption{Final scores of the IoU criterion. Higher values indicate better results.}
    \end{subtable}
    \vfill
    \begin{subtable}{\textwidth}
    \centering
    \resizebox{!}{5cm}{\begin{tabularx}{\textwidth}{c *{7}{Y}}
\toprule
& &
 \multicolumn{2}{c}{trivial prepr.}  
 & \multicolumn{2}{c}{trivial+PDE prepr.}\\
\cmidrule(lr){3-4} \cmidrule(l){5-6}
eval on & train on & no equiv. & equiv. & no equiv. & equiv.\\
\midrule \midrule
& simple & \textbf{0.00} & \textbf{0.00} & \textbf{0.00} & \textbf{0.00}\\
disc simple & complex & \textbf{0.00} & \textbf{0.00} & \textbf{0.00} & \textbf{0.00}\\
& combined & \textbf{0.00} & \textbf{0.00} & \textbf{0.00} & \textbf{0.00}\\\midrule
& simple & 32.50 & 15.50 & 24.50 & \textbf{13.50}\\
disc complex & complex & 12.50 & 11.00 & 5.00 & \textbf{3.50}\\
& combined & 25.50 & 15.00 & 7.00 & \textbf{5.00}\\\midrule
& simple & 16.25 & 7.75 & 12.25 & \textbf{6.75}\\
disc combined & complex & 6.25 & 5.50 & 2.50 & \textbf{1.75}\\
& combined & 12.75 & 7.50 & 3.50 & \textbf{2.50}\\
\midrule \midrule
& simple & 52.78 & 8.33 & \textbf{0.00} & \textbf{0.00}\\
sphere simple & complex & 69.44 & 13.89 & \textbf{0.00} & \textbf{0.00}\\
& combined & 33.33 & 8.33 & 8.33 & \textbf{0.00}\\\midrule
& simple & 99.97 & 61.11 & 38.89 & \textbf{13.89}\\
sphere complex & complex & 66.67 & 13.89 & \textbf{0.00} & \textbf{0.00}\\
& combined & 69.44 & 38.89 & 8.33 & \textbf{0.00}\\\midrule
& simple & 76.39 & 34.72 & 19.44 & \textbf{6.94}\\
sphere combined & complex & 68.06 & 13.89 & \textbf{0.00} & \textbf{0.00}\\
& combined & 51.39 & 23.61 & 8.33 & \textbf{0.00}\\
\bottomrule
\end{tabularx}}
    \subcaption{Final scores of the fail\% criterion. Lower values indicate better results.}
    \end{subtable}   
\end{table}


\begin{table}[]
\caption{Model predictions of two different problems from the \textbf{disc simple} validation dataset, using the UNet with different preprocessings and equivariances. We train the models on subsets of the dataset and vary the training size along the columns of the table. At the boxes below the tables we show the corresponding ground truth density for each problem.}
\label{5_fig:disc_simple_predictions}
\begin{subtable}[h]{0.99\textwidth}
    \centering\setcellgapes{3pt}\makegapedcells
    \setlength\tabcolsep{3.5pt}
    \begin{tabular}{c|c||ScScScScSc}
    \multicolumn{2}{c||}{} & \multicolumn{5}{c}{training samples} \\\hline
     prepr. & equiv. & 10 & 50 & 100 & 500 & 1500 \\\hline
     \multirow{2}{*}{\rotatebox{90}{trivial}} & & \parbox[m]{6em}{\includegraphics[trim={0cm 0cm 0cm 0cm},clip, width=0.12\textwidth]{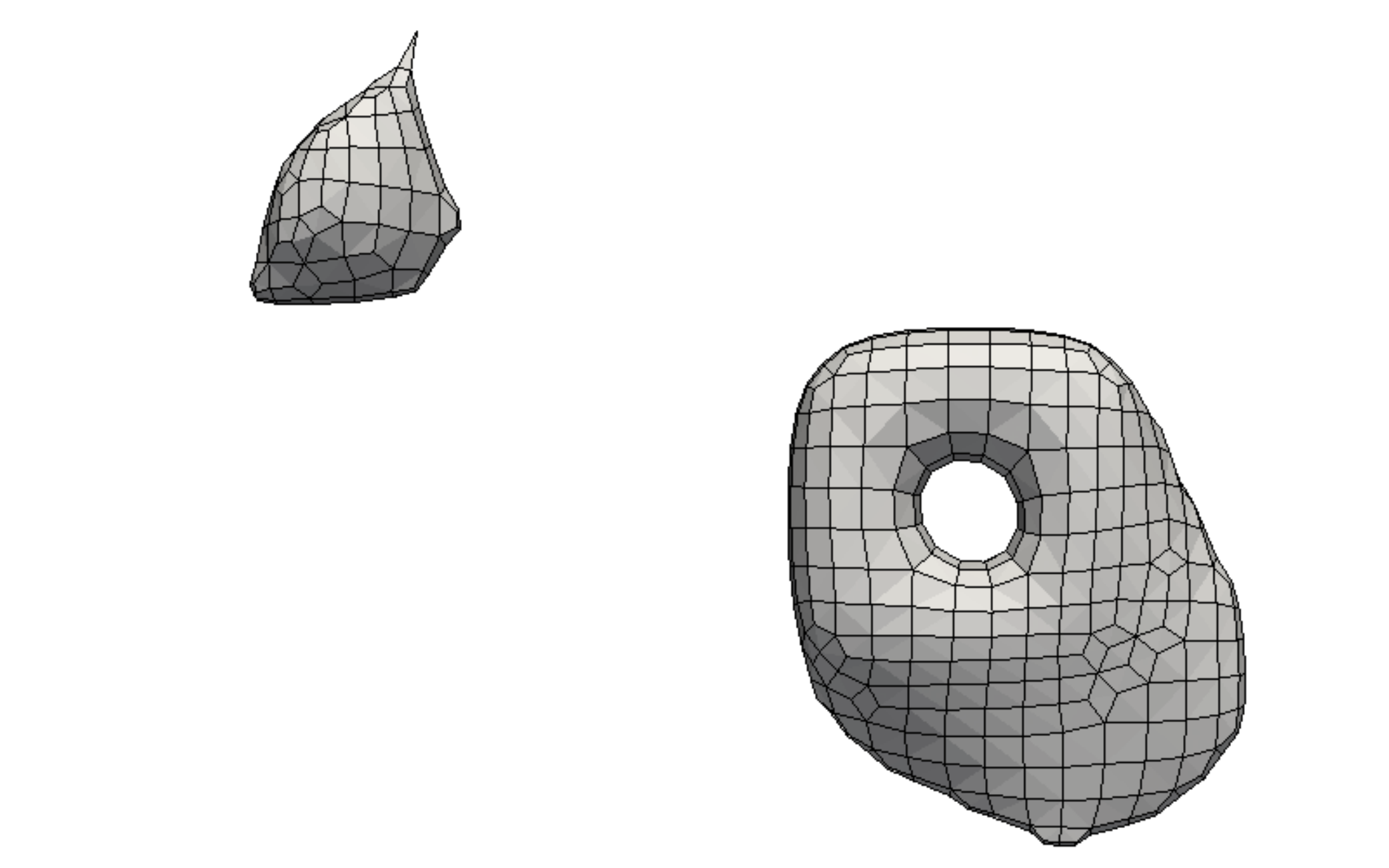}} & \parbox[m]{6em}{\includegraphics[trim={0cm 0cm 0cm 0cm},clip, width=0.12\textwidth]{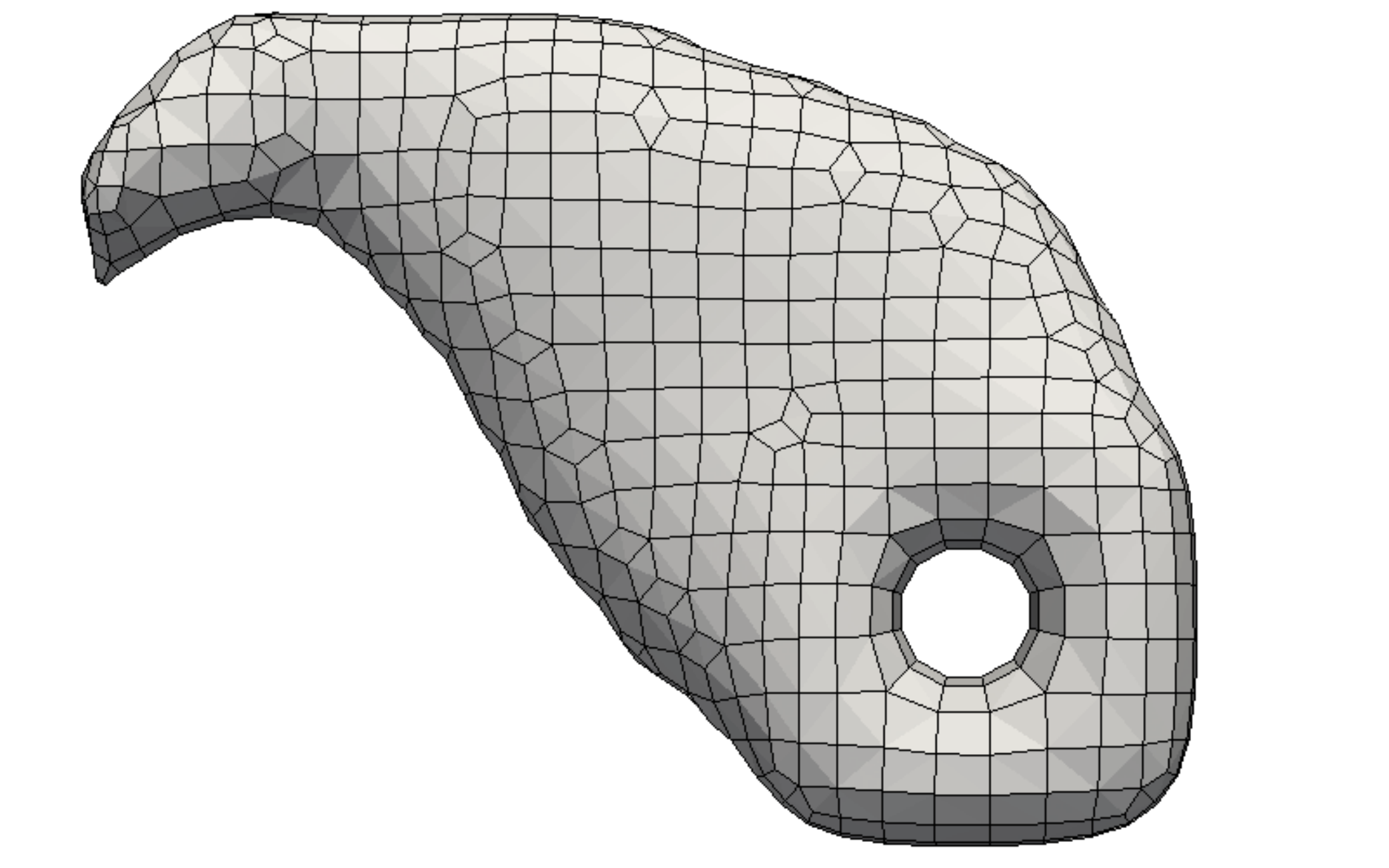}} & \parbox[m]{6em}{\includegraphics[trim={0cm 0cm 0cm 0cm},clip, width=0.12\textwidth]{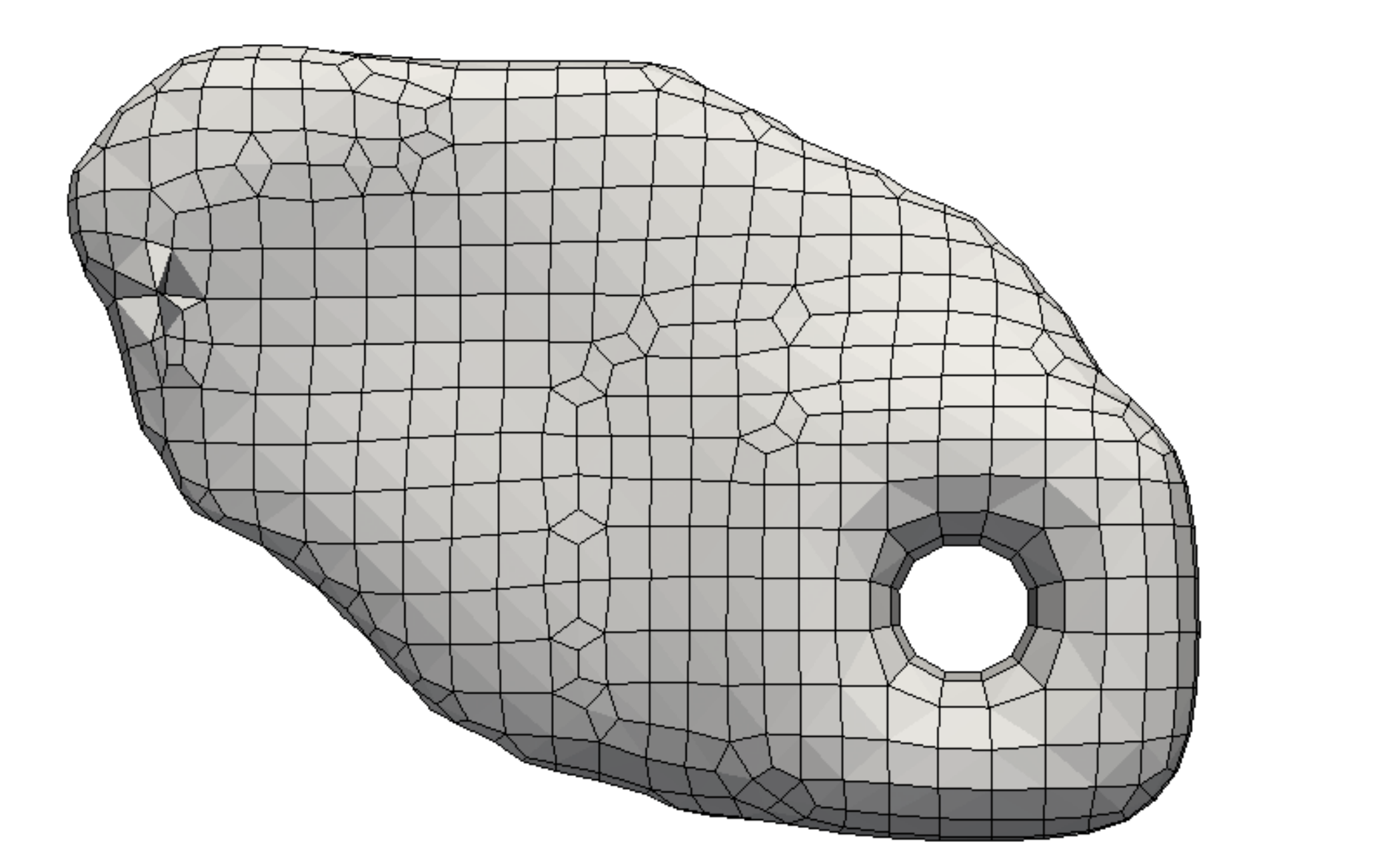}} & \parbox[m]{6em}{\includegraphics[trim={0cm 0cm 0cm 0cm},clip, width=0.12\textwidth]{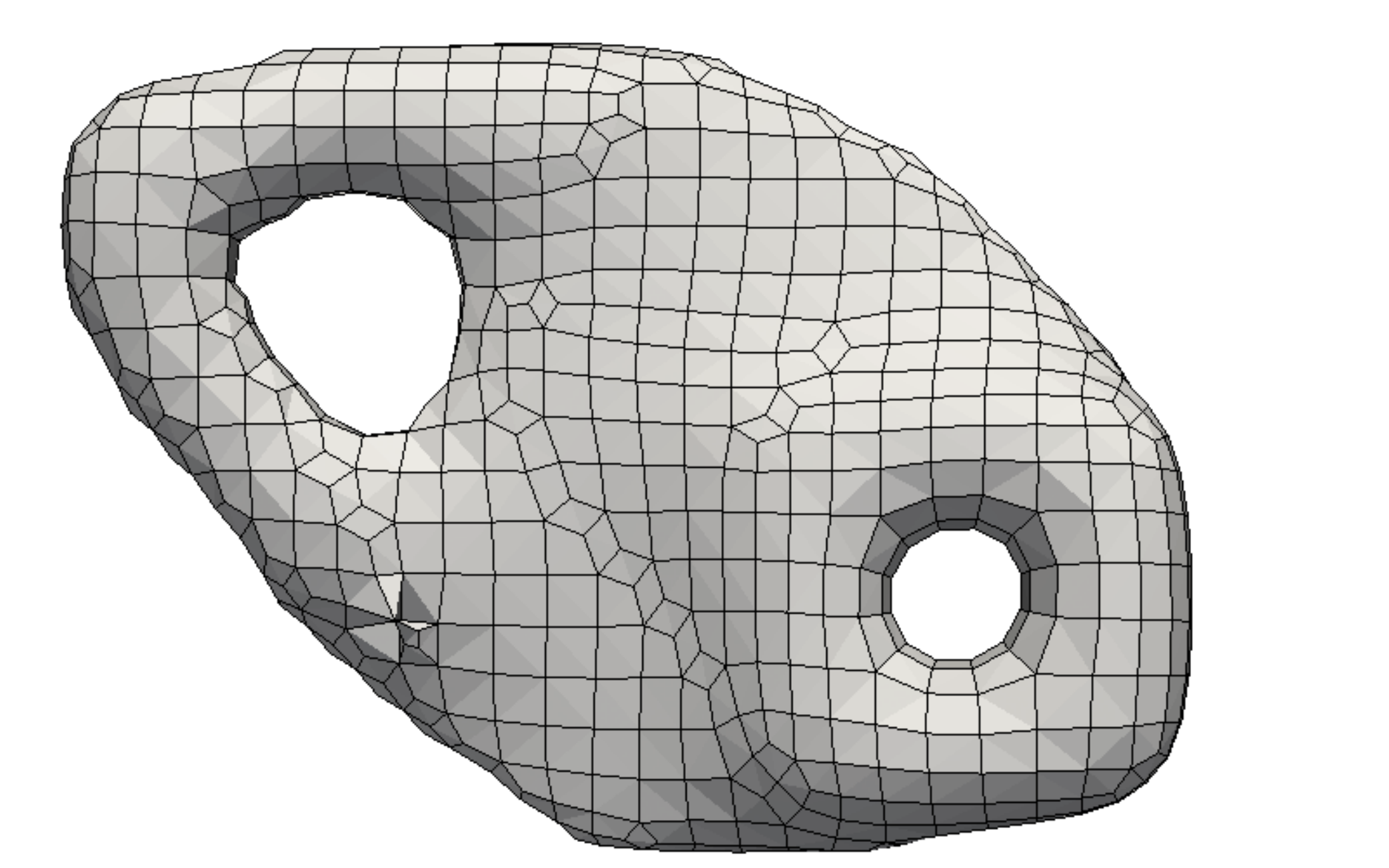}} & \parbox[m]{6em}{\includegraphics[trim={0cm 0cm 0cm 0cm},clip, width=0.12\textwidth]{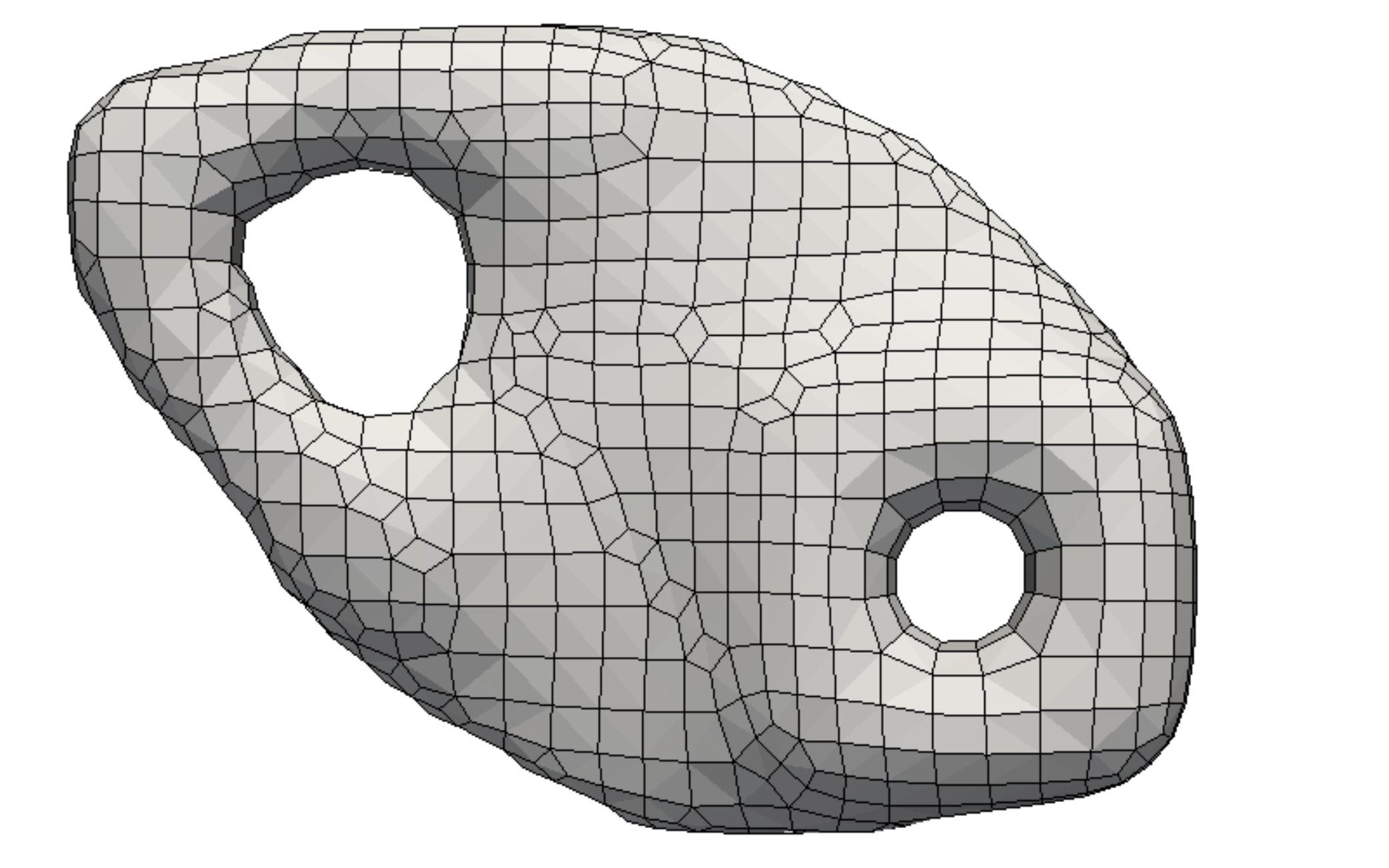}} \\\cline{2-7}
     & \checkmark
     & \parbox[m]{6em}{\includegraphics[trim={0cm 0cm 0cm 0cm},clip, width=0.12\textwidth]{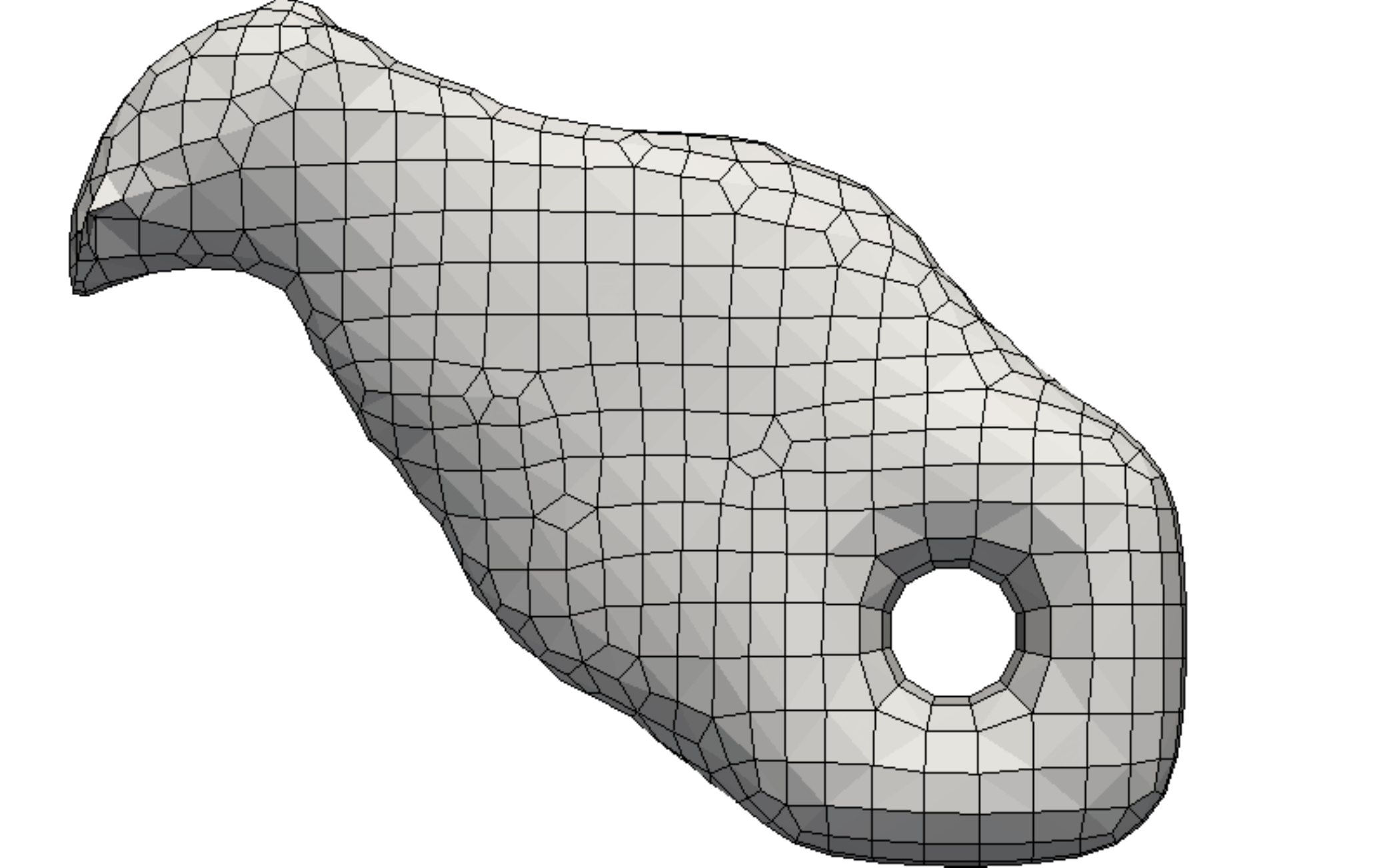}} & \parbox[m]{6em}{\includegraphics[trim={0cm 0cm 0cm 0cm},clip, width=0.12\textwidth]{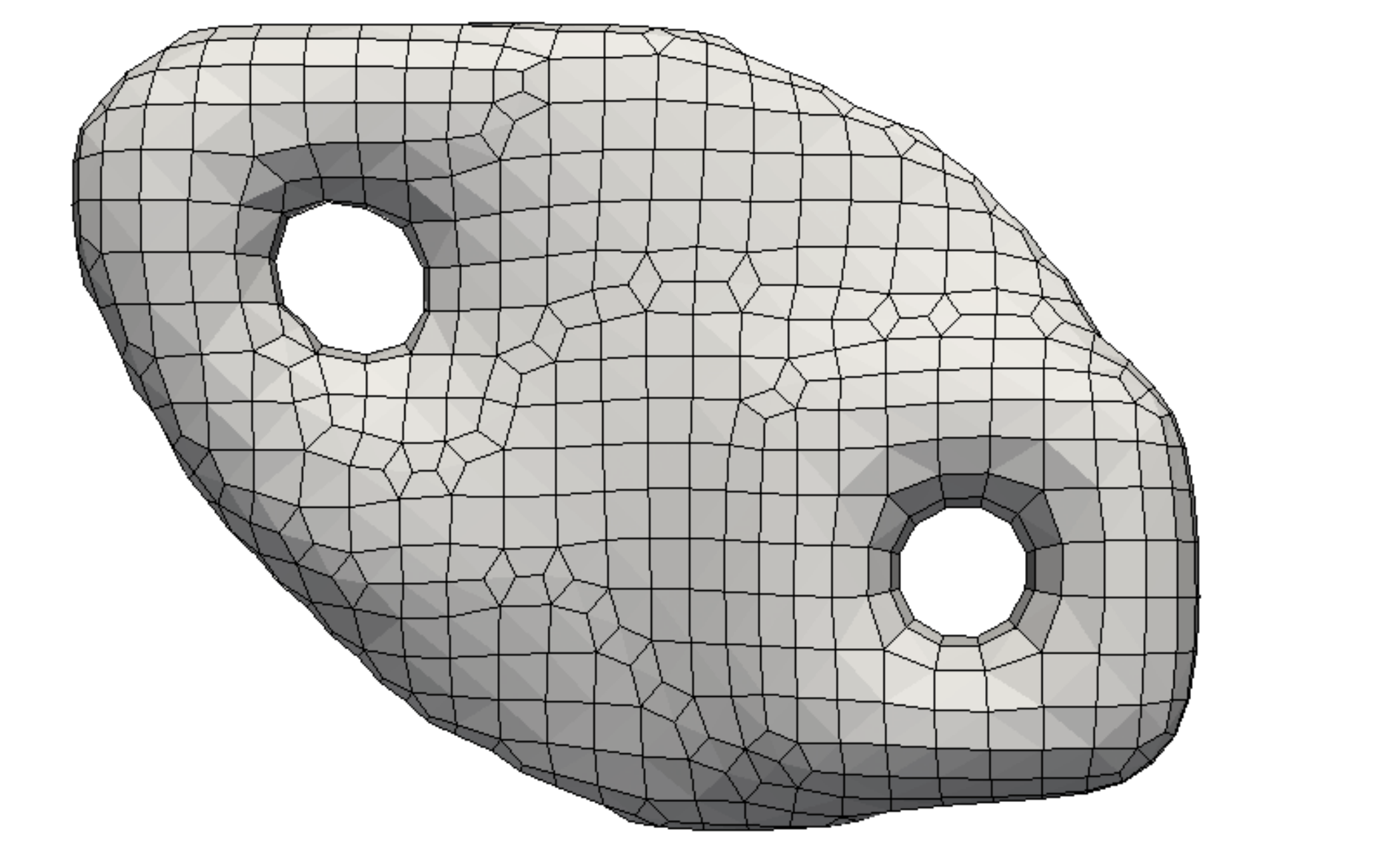}} & \parbox[m]{6em}{\includegraphics[trim={0cm 0cm 0cm 0cm},clip, width=0.12\textwidth]{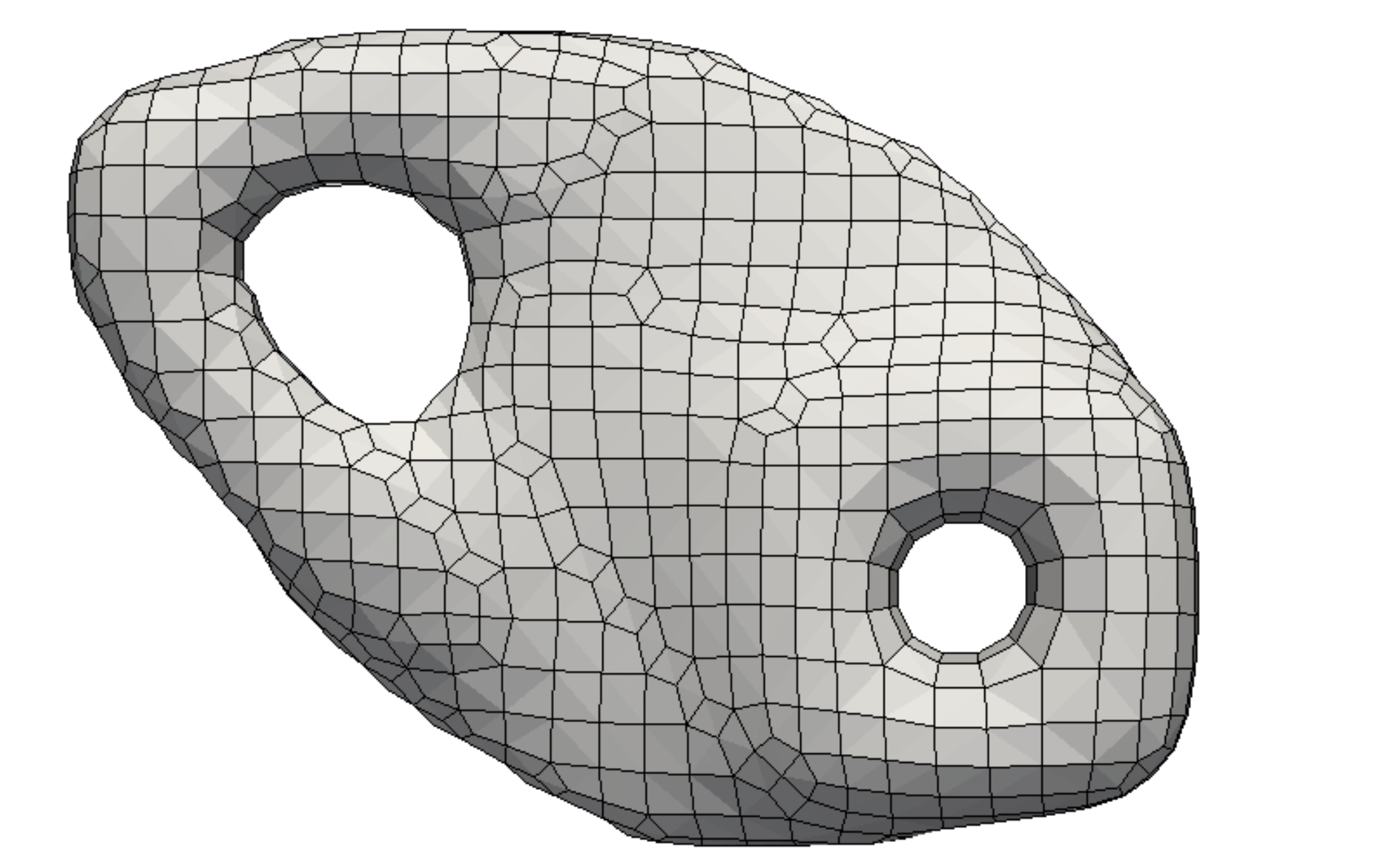}} & \parbox[m]{6em}{\includegraphics[trim={0cm 0cm 0cm 0cm},clip, width=0.12\textwidth]{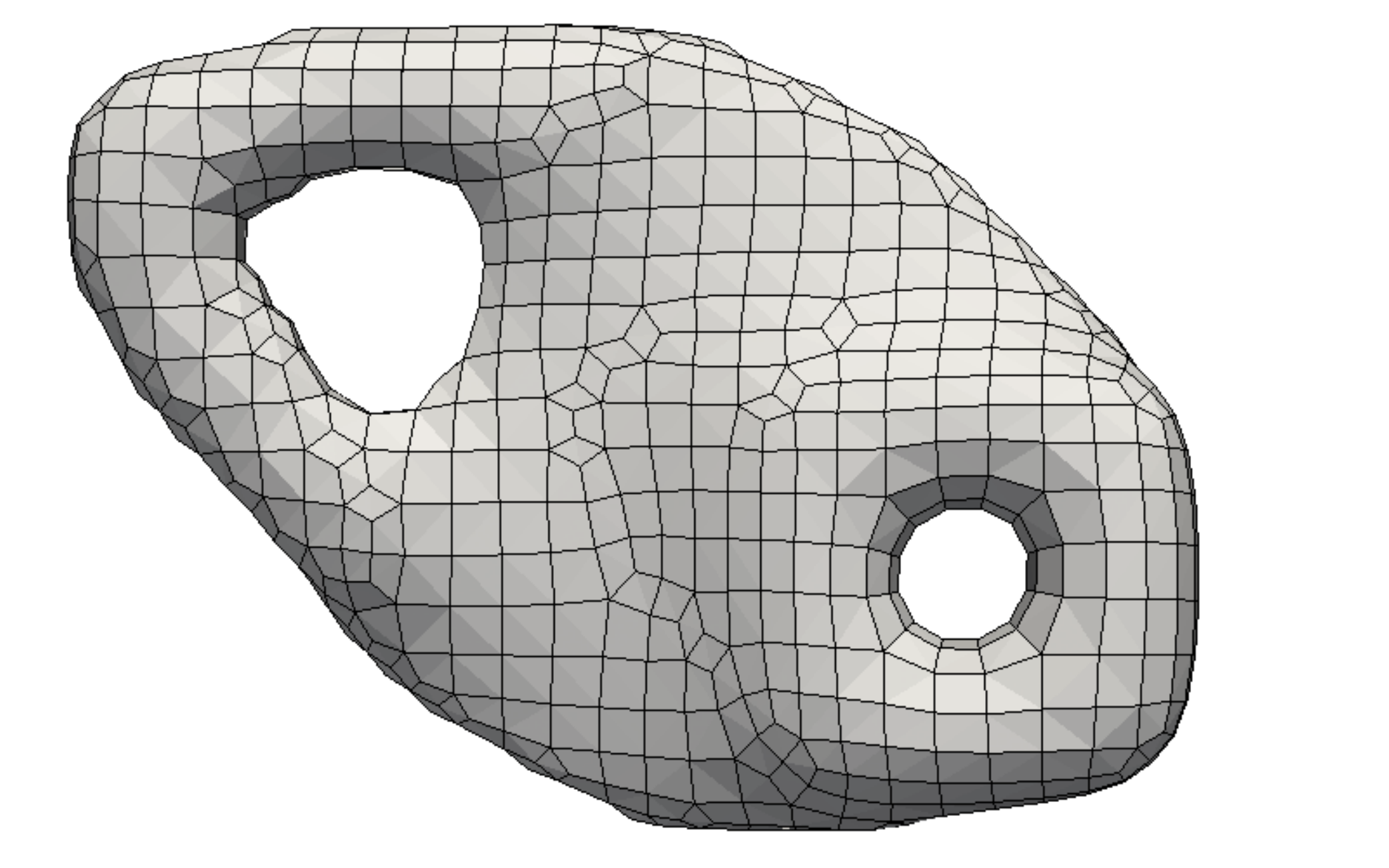}} & \parbox[m]{6em}{\includegraphics[trim={0cm 0cm 0cm 0cm},clip, width=0.12\textwidth]{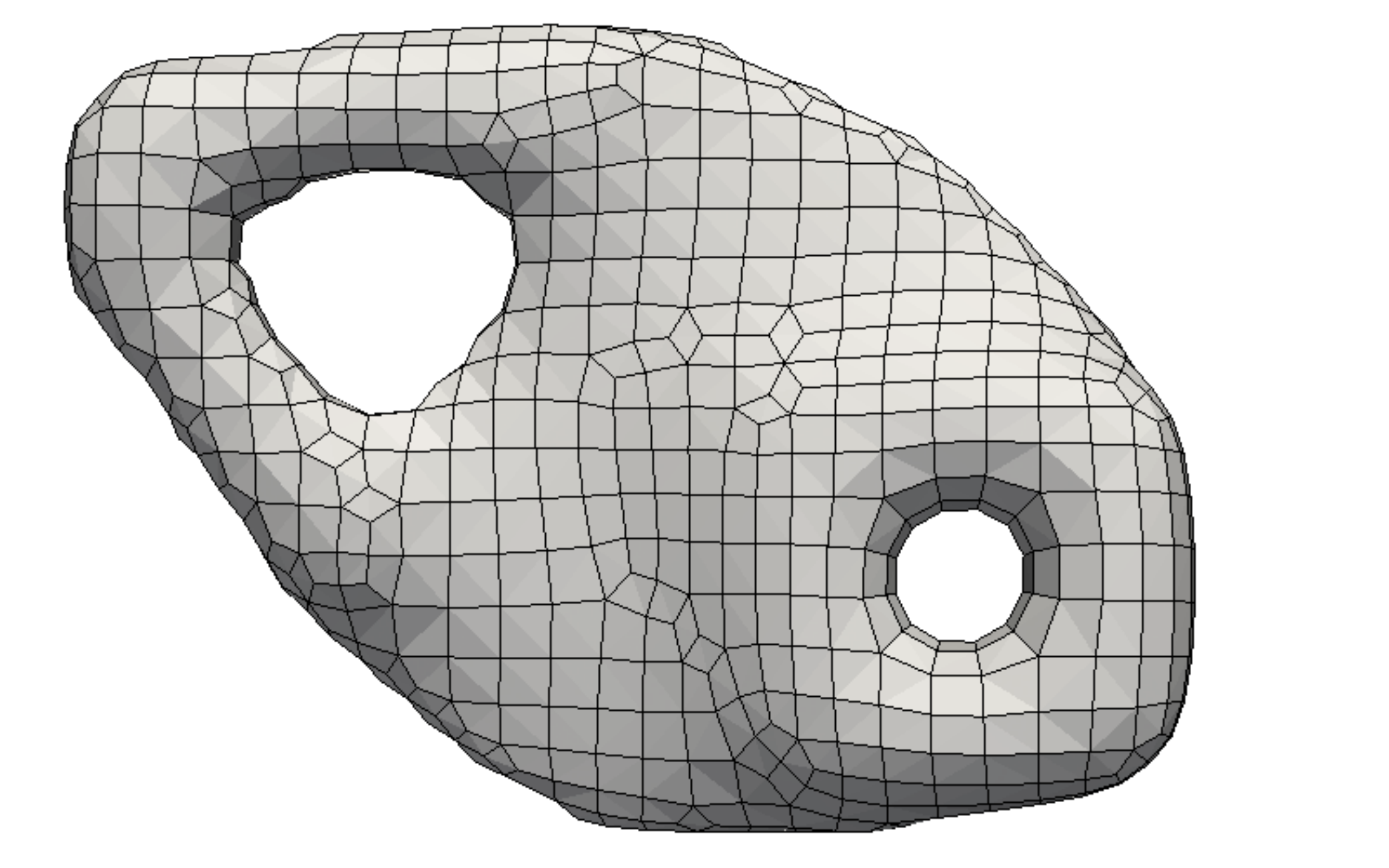}} \\\hline
     \multirow{2}{*}{\rotatebox{90}{trivial+PDE}} & & \parbox[m]{6em}{\includegraphics[trim={0cm 0cm 0cm 0cm},clip, width=0.12\textwidth]{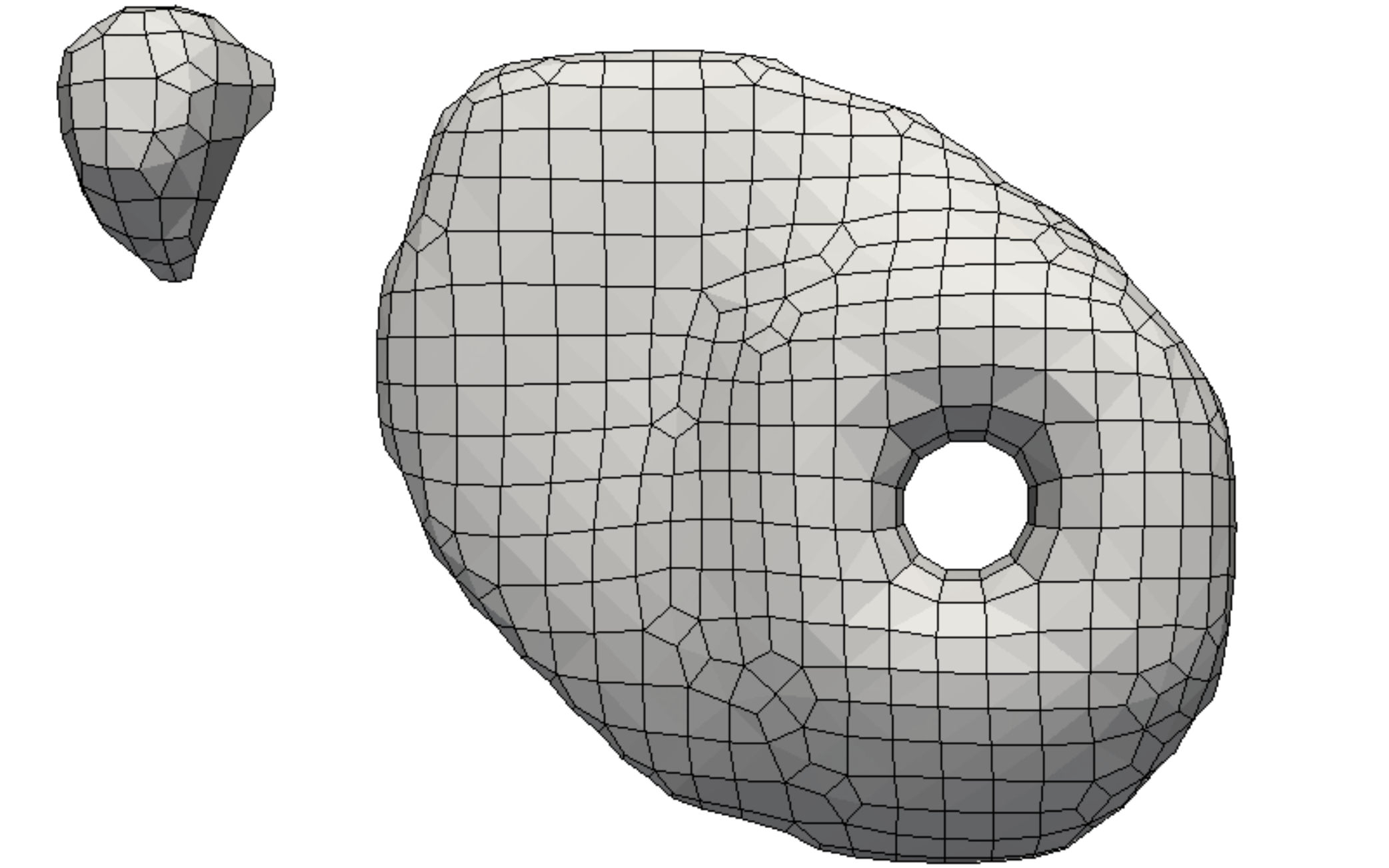}} & \parbox[m]{6em}{\includegraphics[trim={0cm 0cm 0cm 0cm},clip, width=0.12\textwidth]{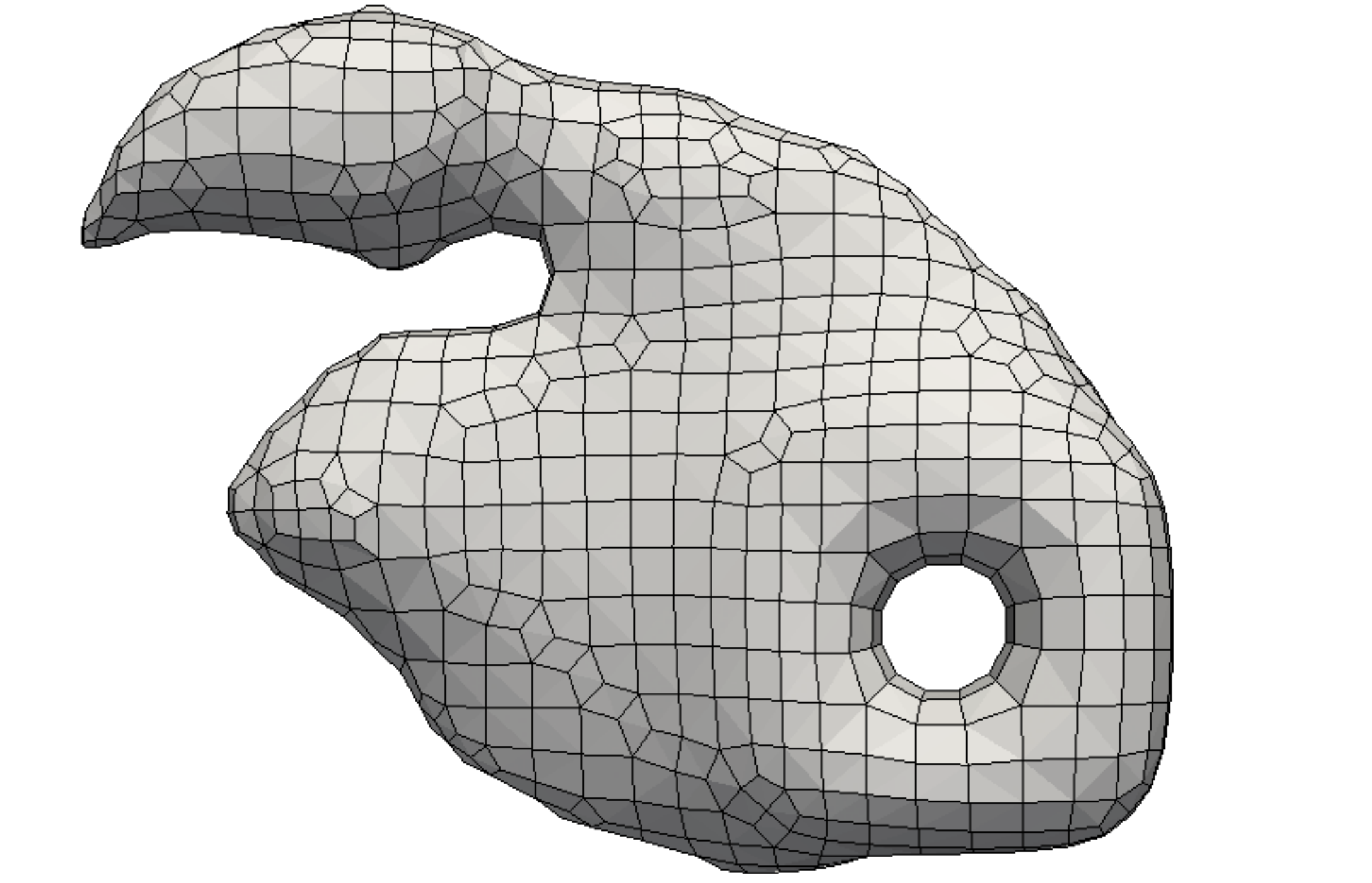}} & \parbox[m]{6em}{\includegraphics[trim={0cm 0cm 0cm 0cm},clip, width=0.12\textwidth]{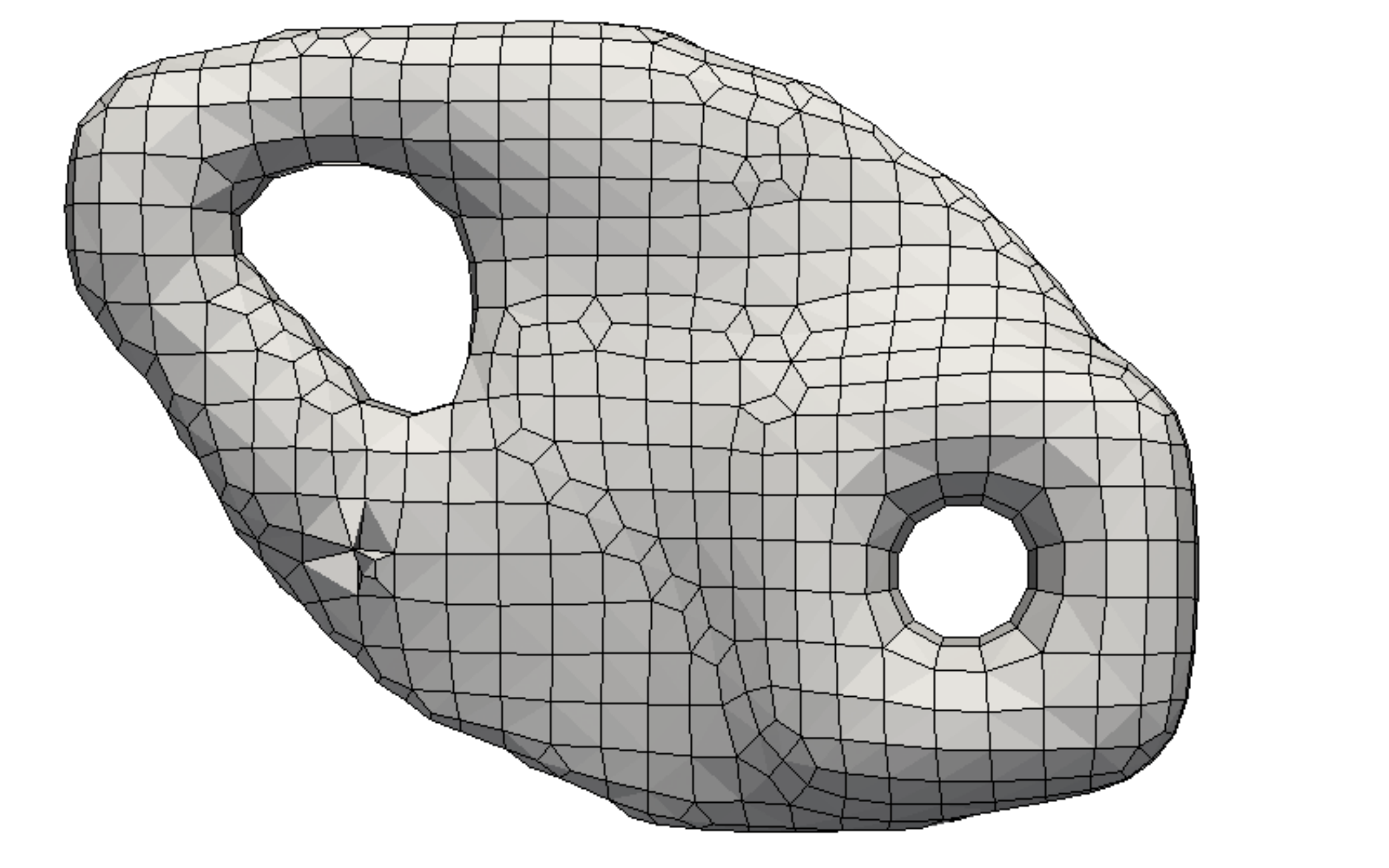}} & \parbox[m]{6em}{\includegraphics[trim={0cm 0cm 0cm 0cm},clip, width=0.12\textwidth]{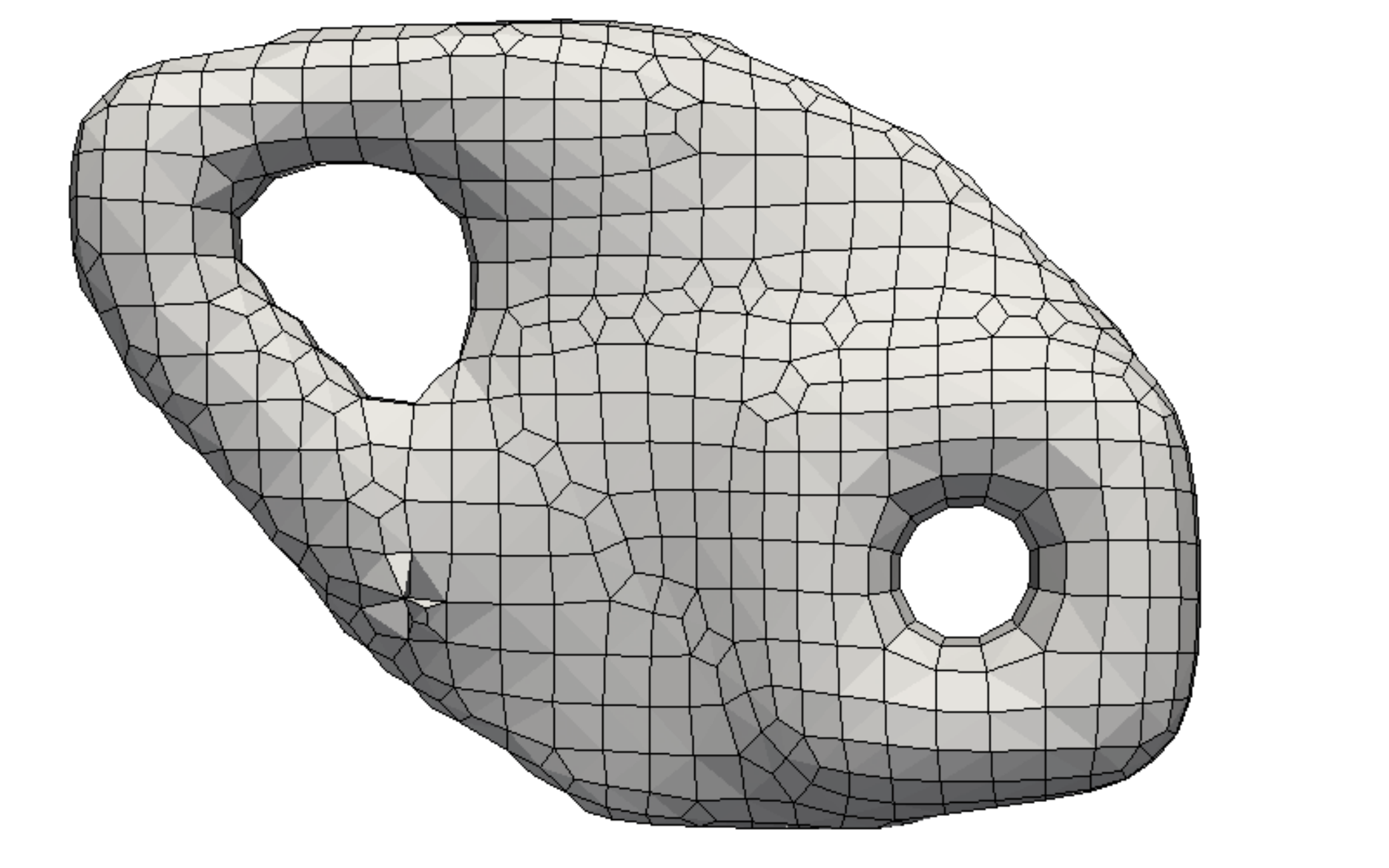}} & \parbox[m]{6em}{\includegraphics[trim={0cm 0cm 0cm 0cm},clip, width=0.12\textwidth]{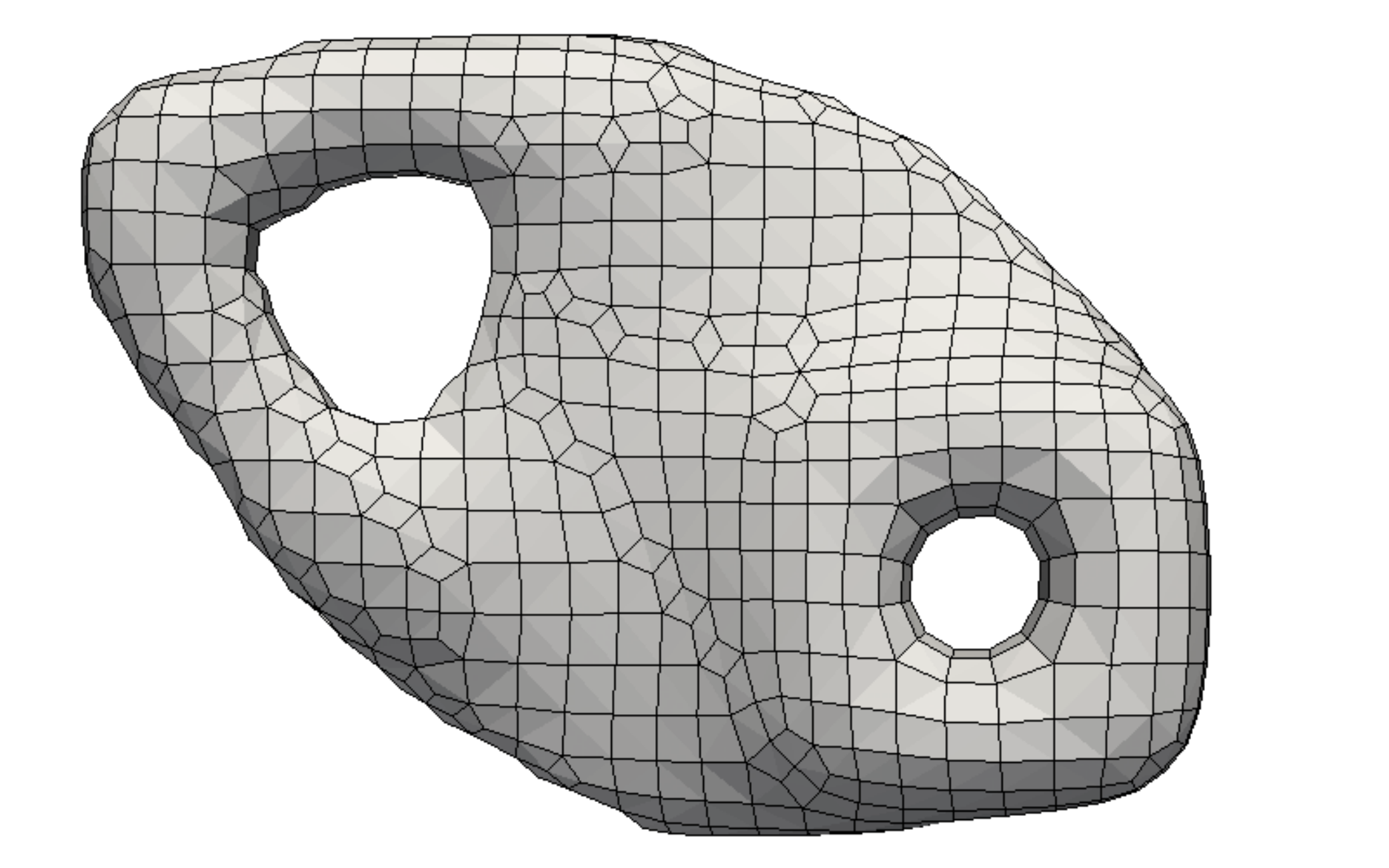}} \\\cline{2-7}
     & \checkmark & \parbox[m]{6em}{\includegraphics[trim={0cm 0cm 0cm 0cm},clip, width=0.12\textwidth]{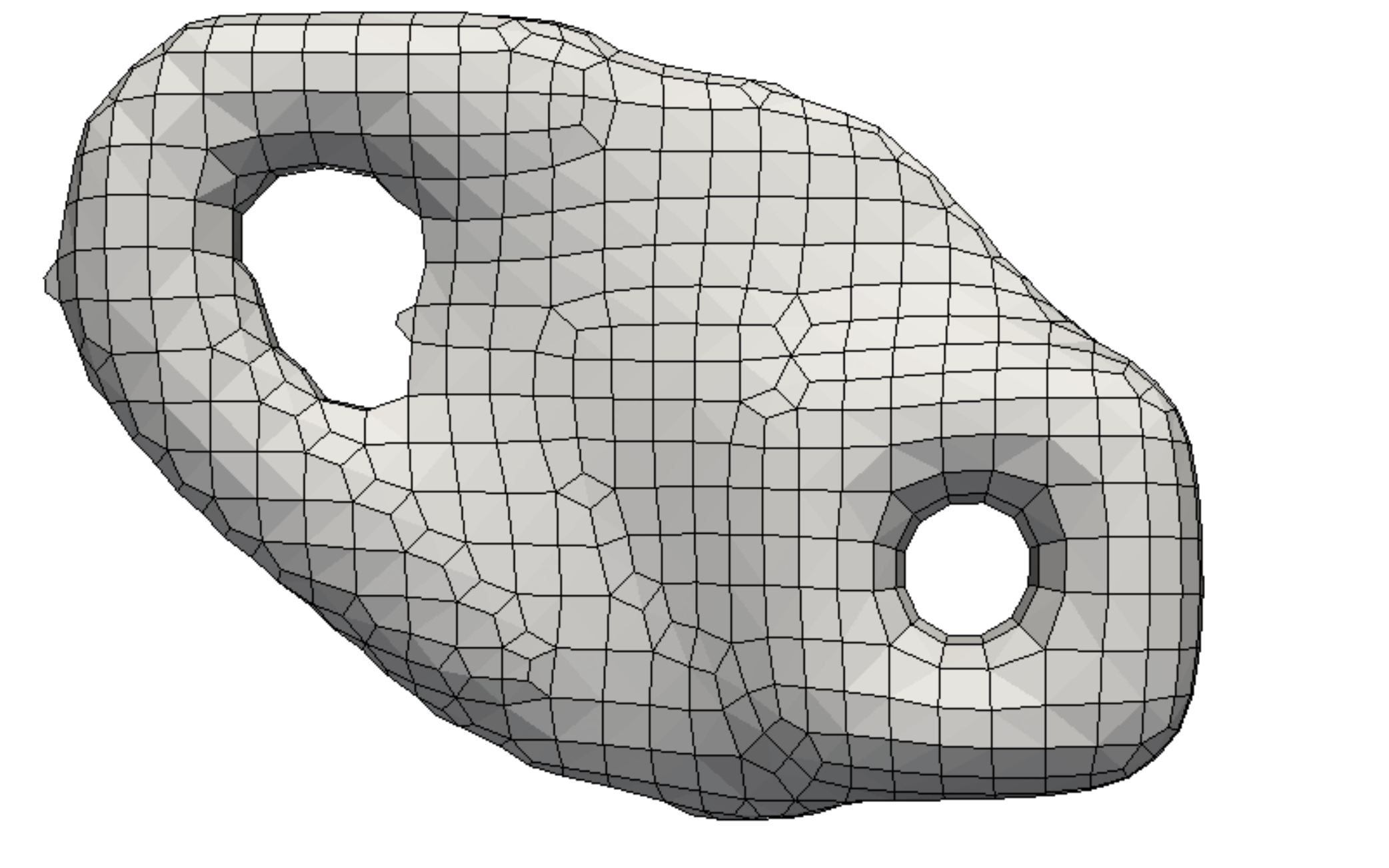}} & \parbox[m]{6em}{\includegraphics[trim={0cm 0cm 0cm 0cm},clip, width=0.12\textwidth]{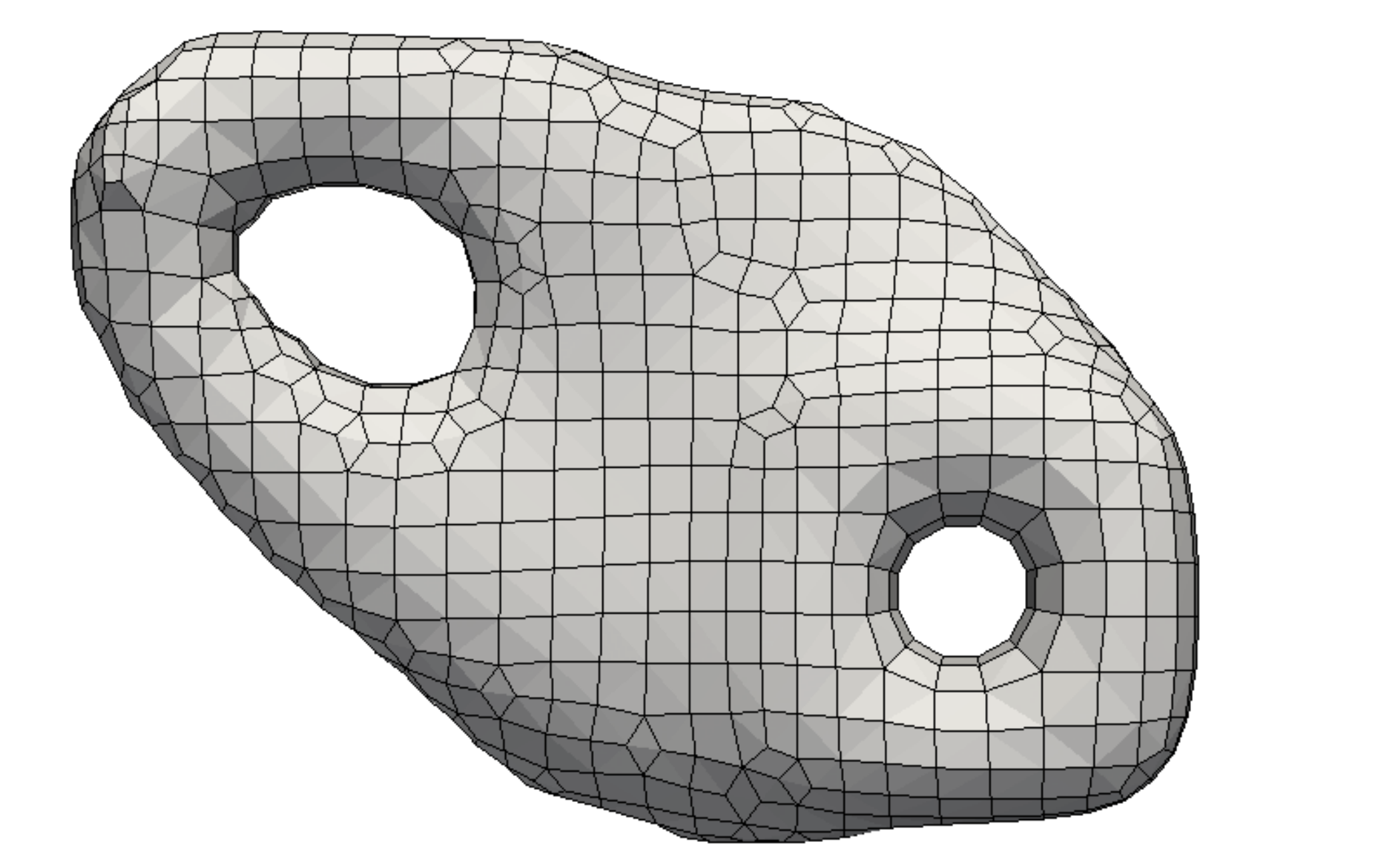}} & \parbox[m]{6em}{\includegraphics[trim={0cm 0cm 0cm 0cm},clip, width=0.12\textwidth]{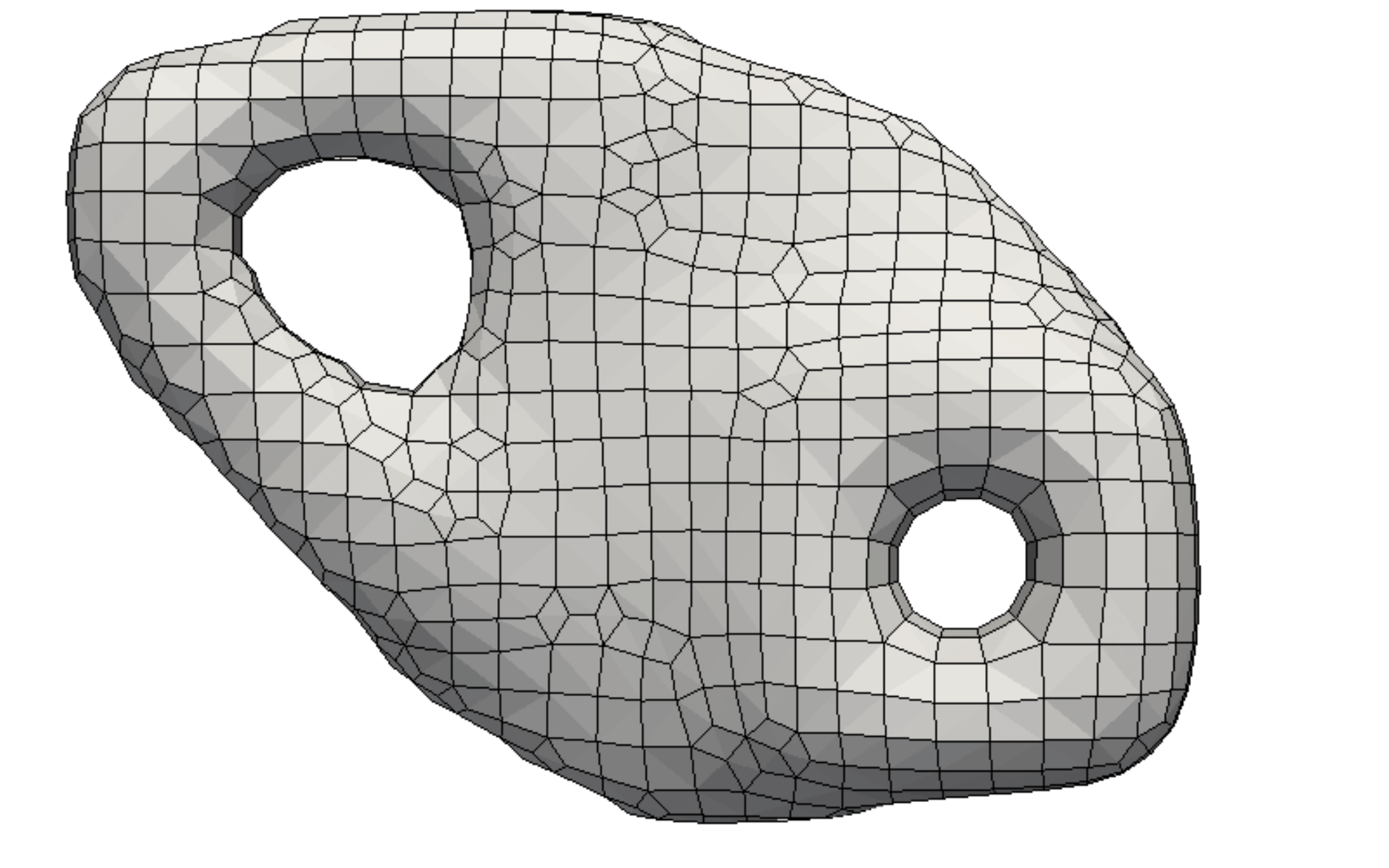}} & \parbox[m]{6em}{\includegraphics[trim={0cm 0cm 0cm 0cm},clip, width=0.12\textwidth]{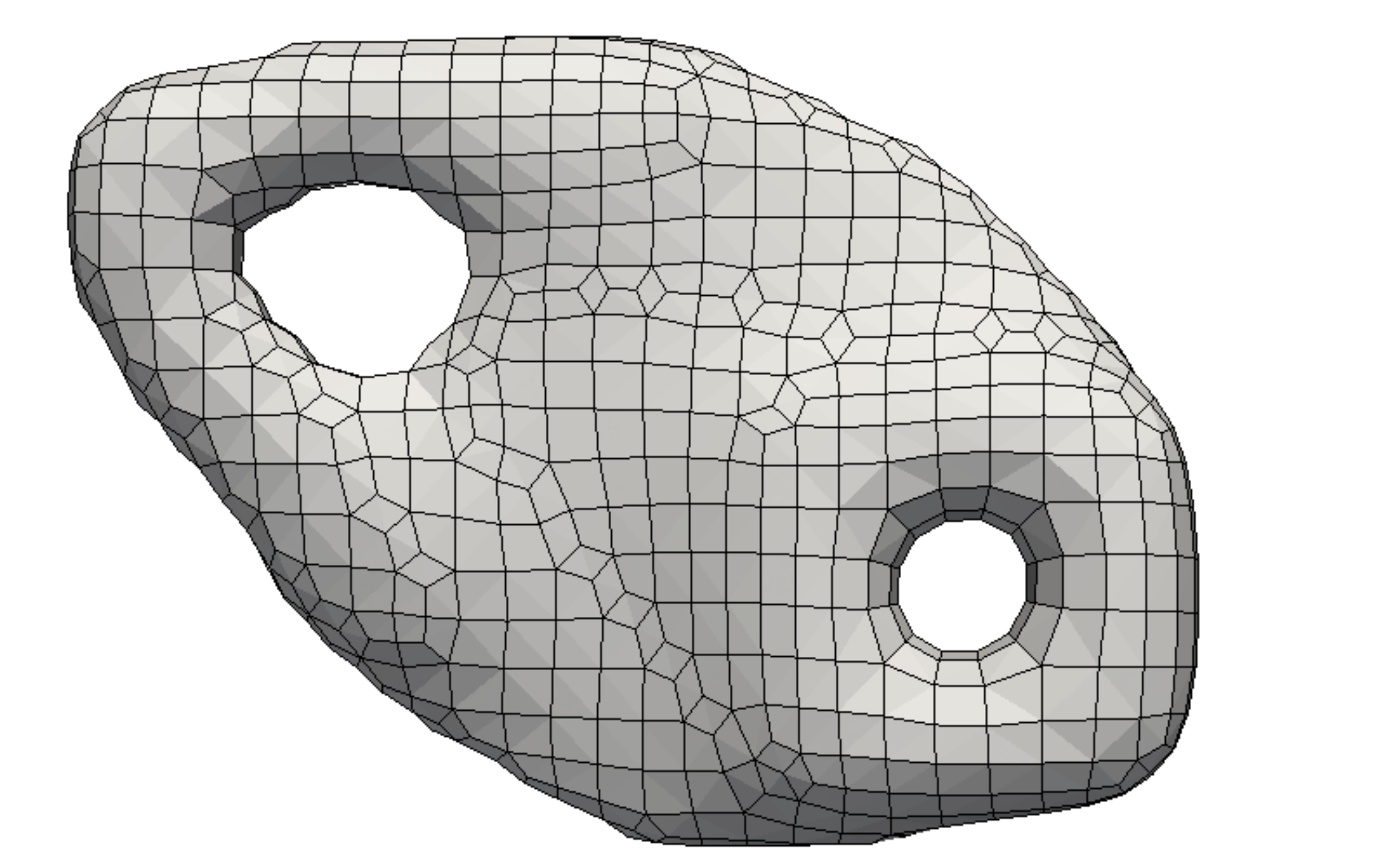}} & \parbox[m]{6em}{\includegraphics[trim={0cm 0cm 0cm 0cm},clip, width=0.12\textwidth]{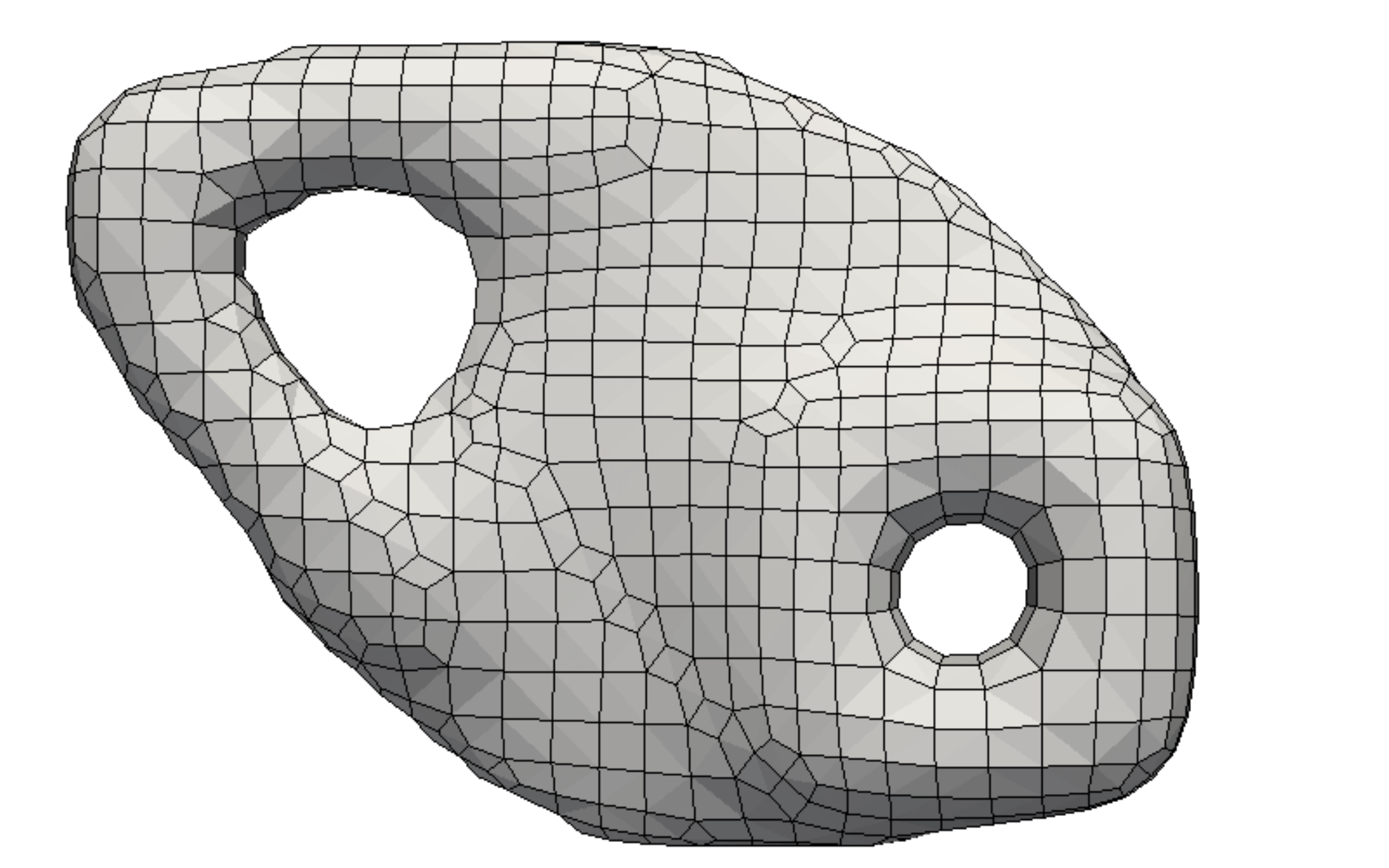}}\\
     \multicolumn{7}{c}{\fbox{\hspace{0.3cm}\parbox[m]{6em}{\includegraphics[trim={0cm 0cm 0cm 0cm},clip, width=0.12\textwidth]{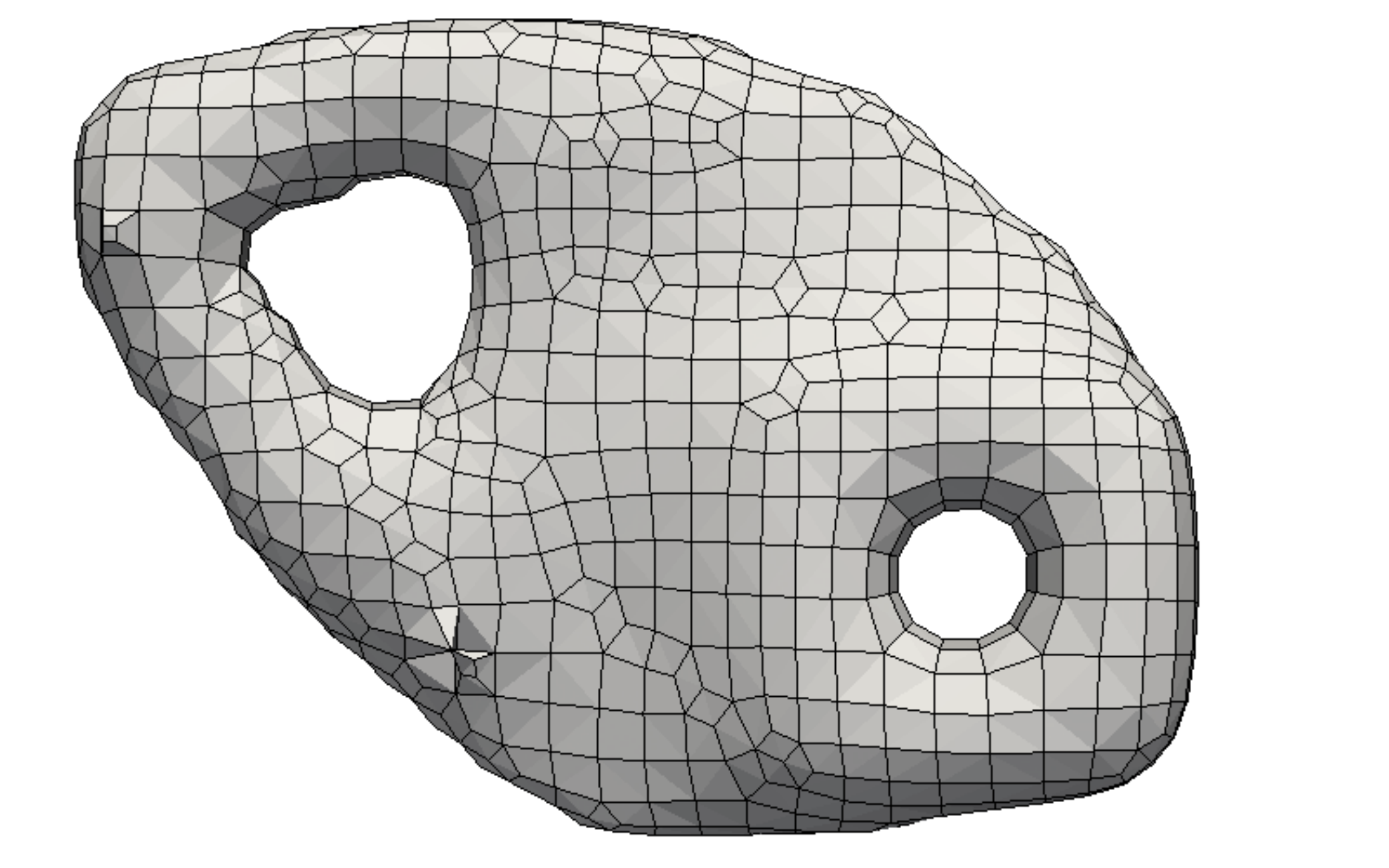}}}}
\end{tabular}
\end{subtable}
\begin{subtable}[h]{0.99\textwidth}
    \vspace{5mm}%
    \centering\setcellgapes{3pt}\makegapedcells
    \setlength\tabcolsep{3.5pt}
    \begin{tabular}{c|c||ScScScScSc}
    \multicolumn{2}{c||}{} & \multicolumn{5}{c}{training samples} \\\hline
     prepr. & equiv. & 10 & 50 & 100 & 500 & 1500 \\\hline
     \multirow{2}{*}{\rotatebox{90}{trivial}} & & \parbox[m]{6em}{\includegraphics[trim={0cm 0cm 0cm 0cm},clip, width=0.12\textwidth]{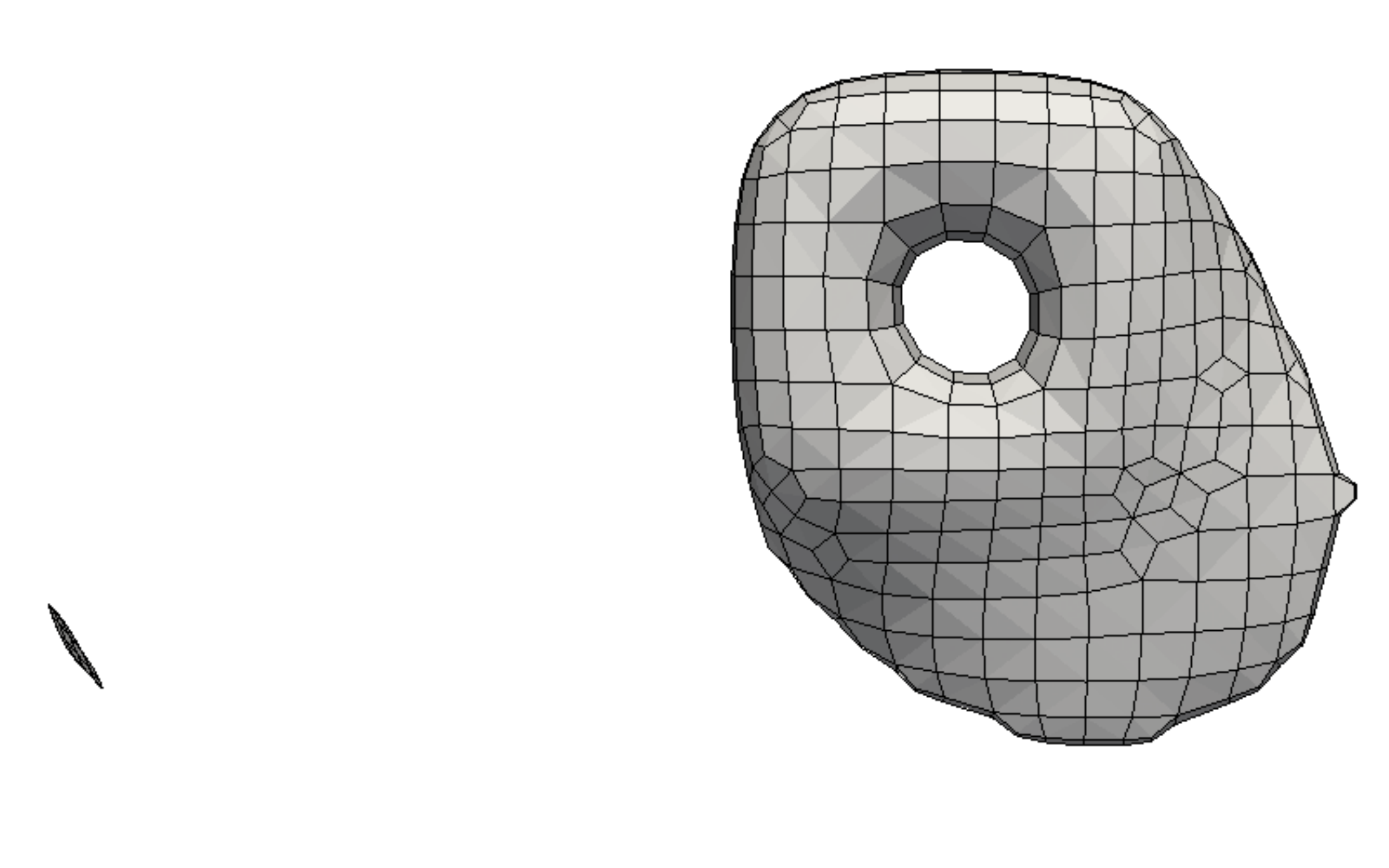}} & \parbox[m]{6em}{\includegraphics[trim={0cm 0cm 0cm 0cm},clip, width=0.12\textwidth]{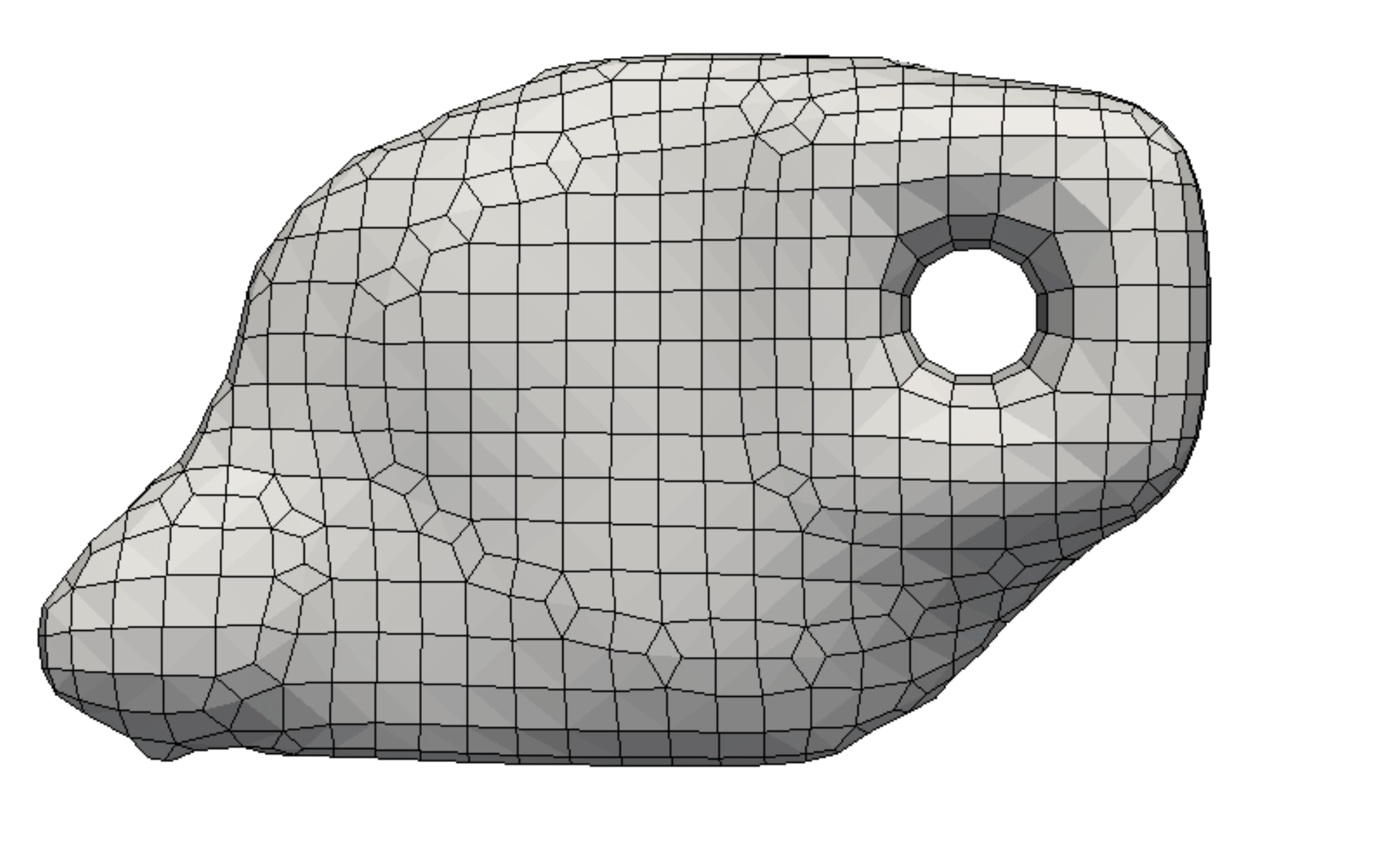}} & \parbox[m]{6em}{\includegraphics[trim={0cm 0cm 0cm 0cm},clip, width=0.12\textwidth]{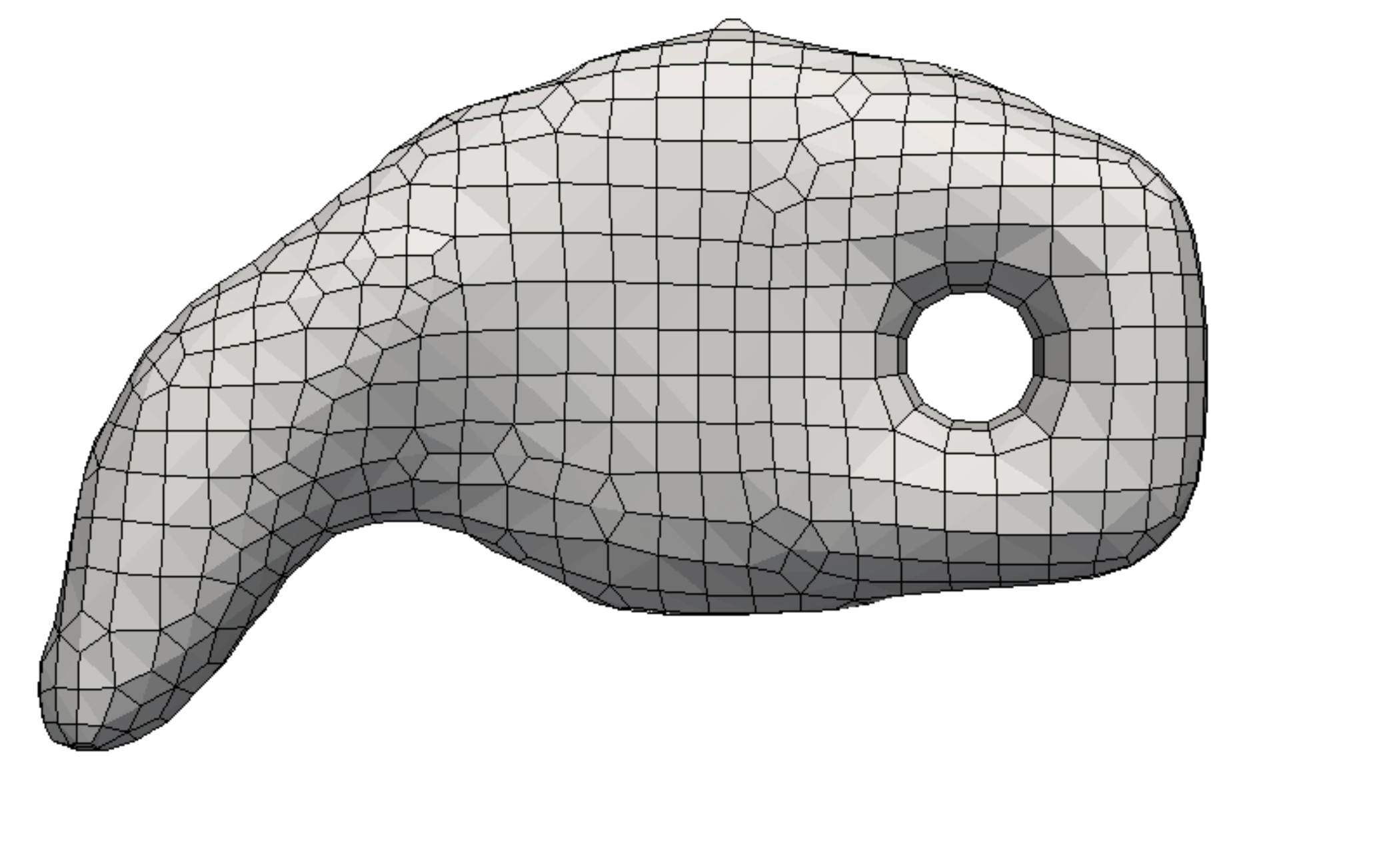}} & \parbox[m]{6em}{\includegraphics[trim={0cm 0cm 0cm 0cm},clip, width=0.12\textwidth]{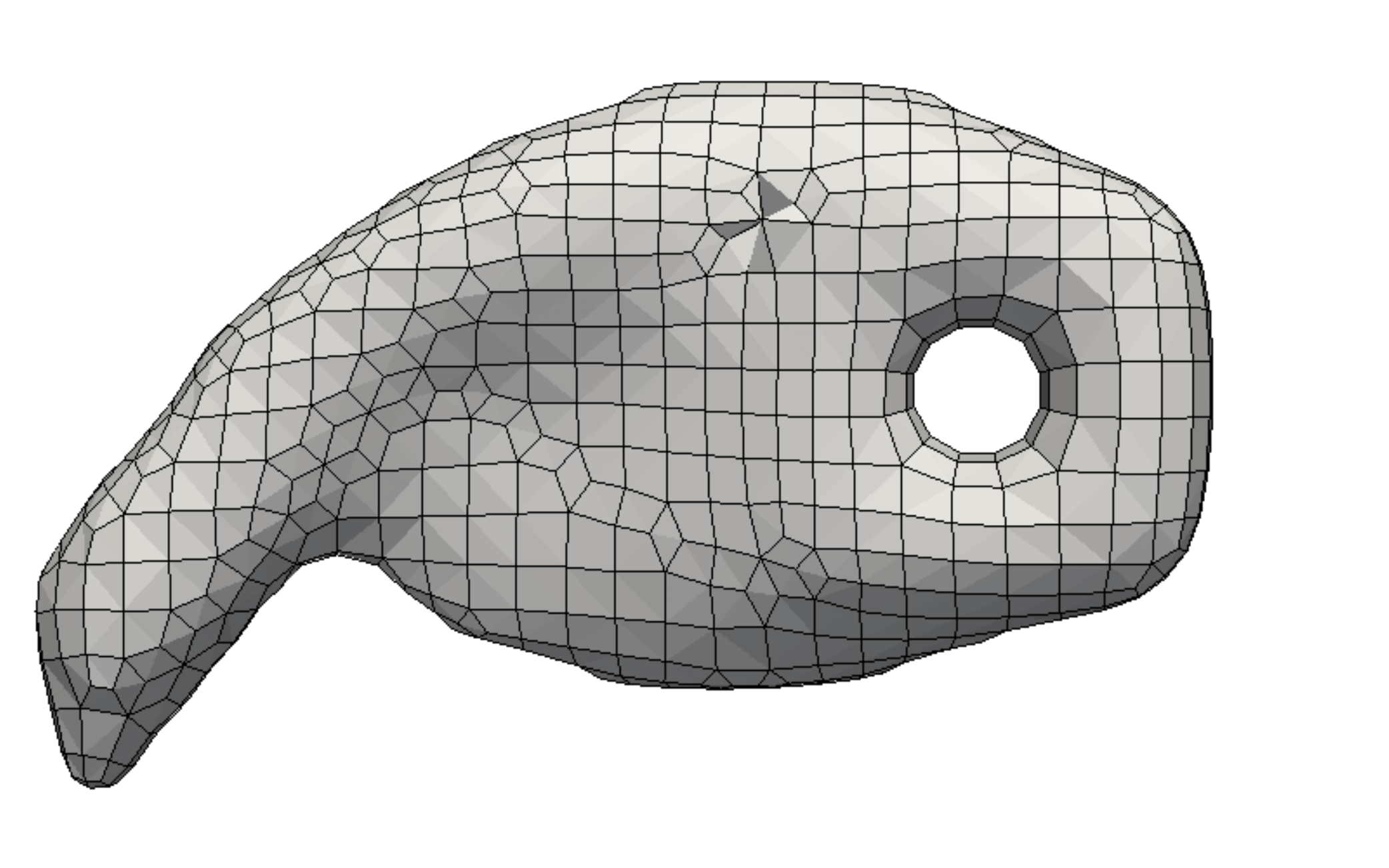}} & \parbox[m]{6em}{\includegraphics[trim={0cm 0cm 0cm 0cm},clip, width=0.12\textwidth]{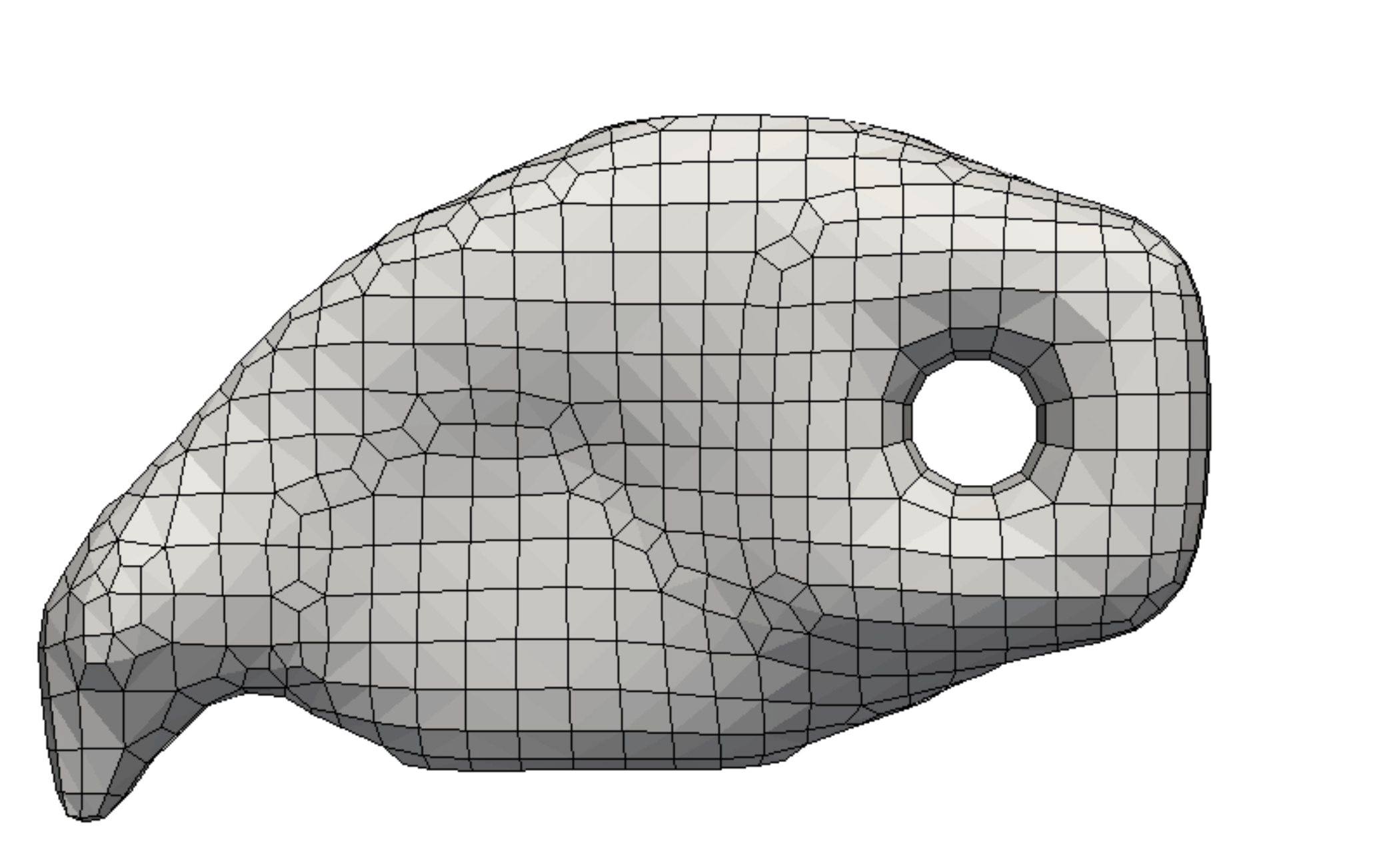}} \\\cline{2-7}
     & \checkmark
     & \parbox[m]{6em}{\includegraphics[trim={0cm 0cm 0cm 0cm},clip, width=0.12\textwidth]{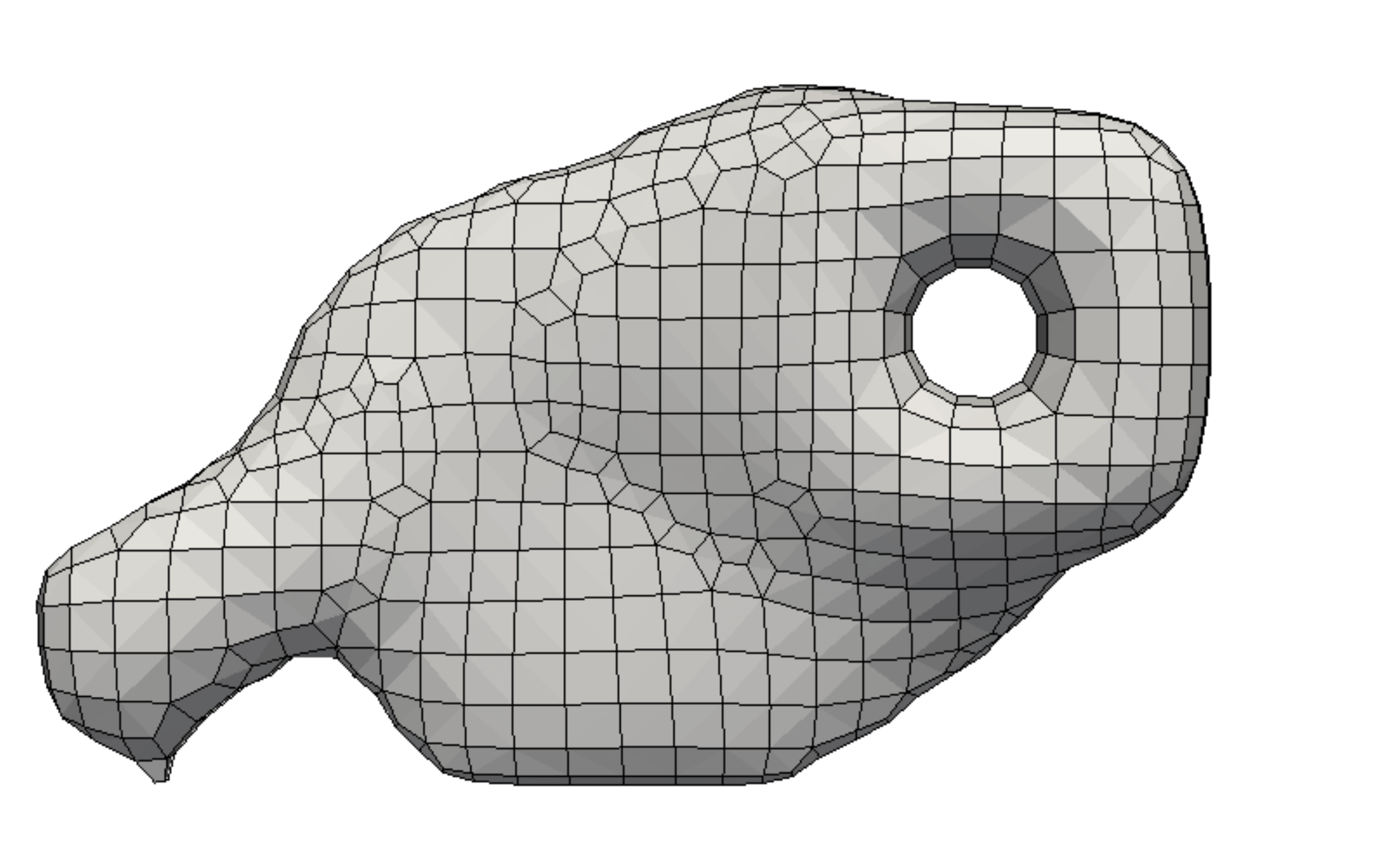}} & \parbox[m]{6em}{\includegraphics[trim={0cm 0cm 0cm 0cm},clip, width=0.12\textwidth]{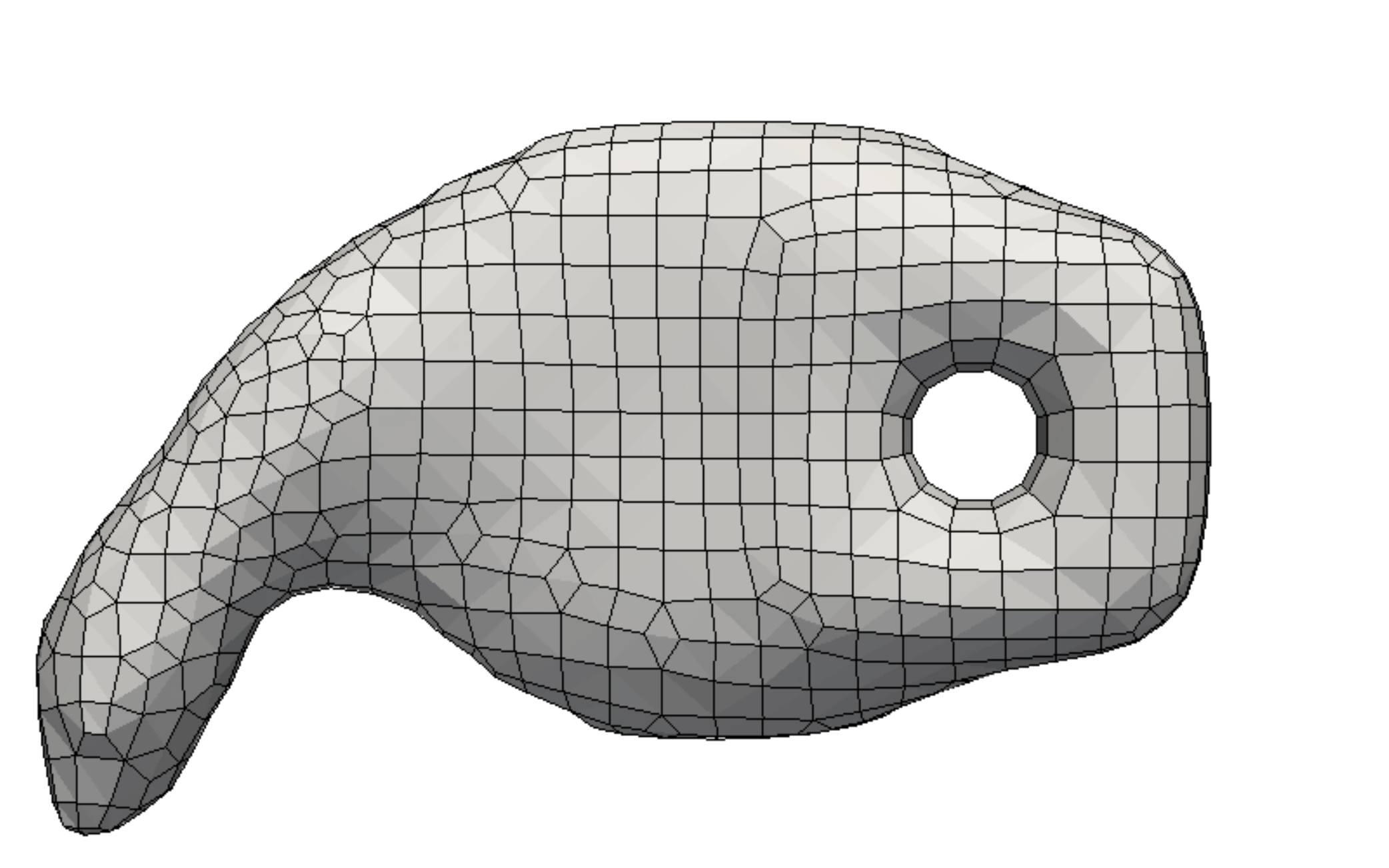}} & \parbox[m]{6em}{\includegraphics[trim={0cm 0cm 0cm 0cm},clip, width=0.12\textwidth]{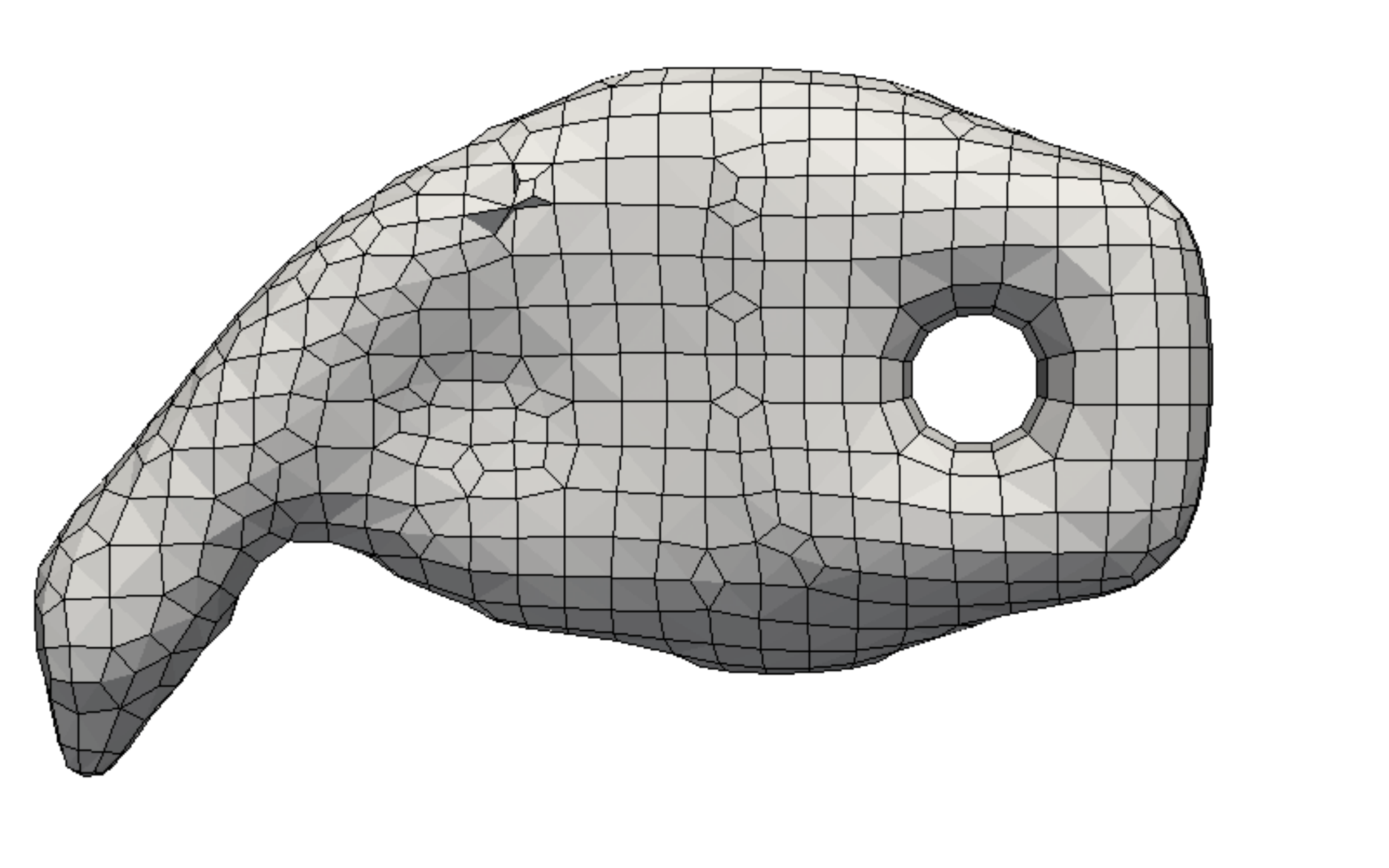}} & \parbox[m]{6em}{\includegraphics[trim={0cm 0cm 0cm 0cm},clip, width=0.12\textwidth]{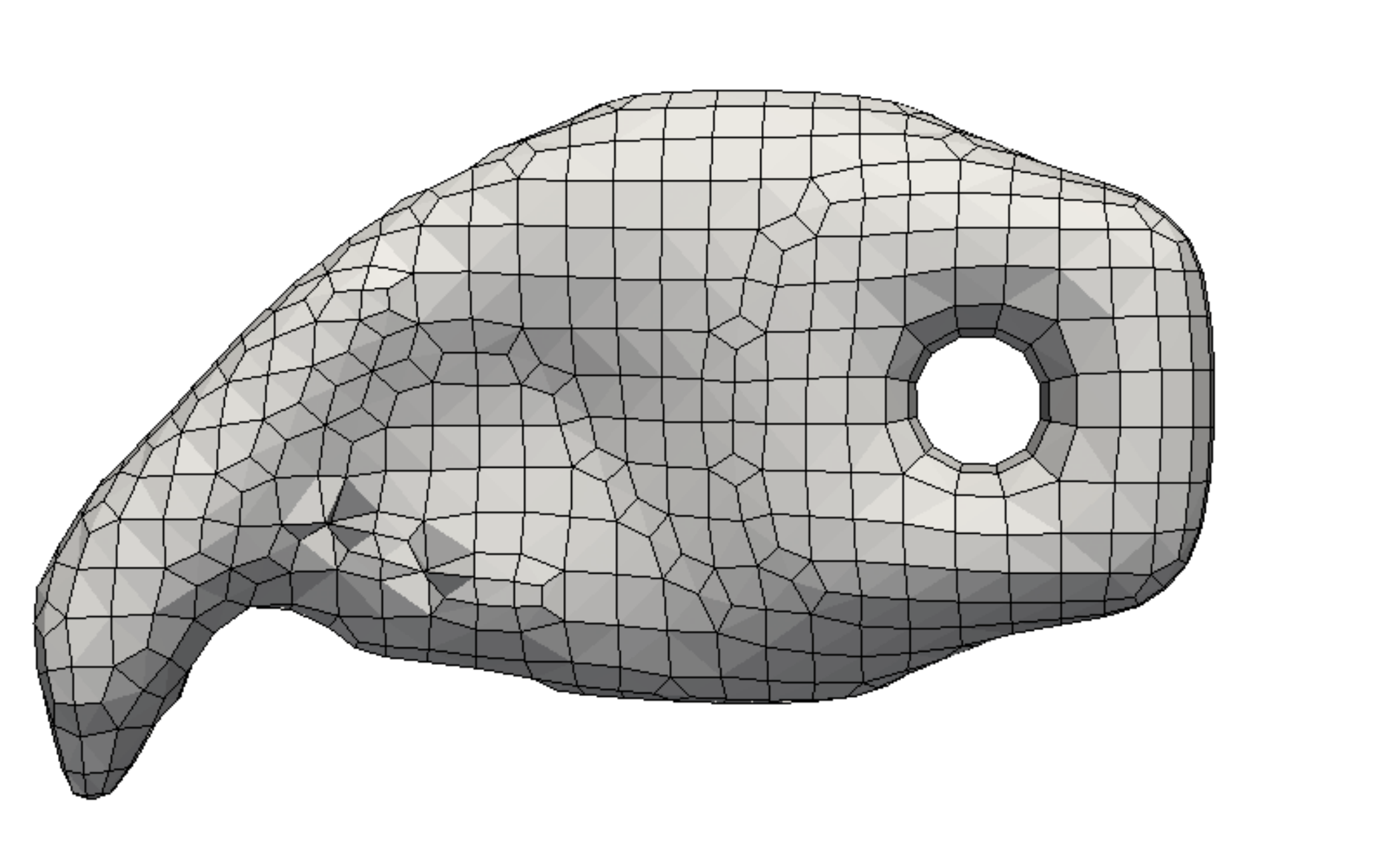}} & \parbox[m]{6em}{\includegraphics[trim={0cm 0cm 0cm 0cm},clip, width=0.12\textwidth]{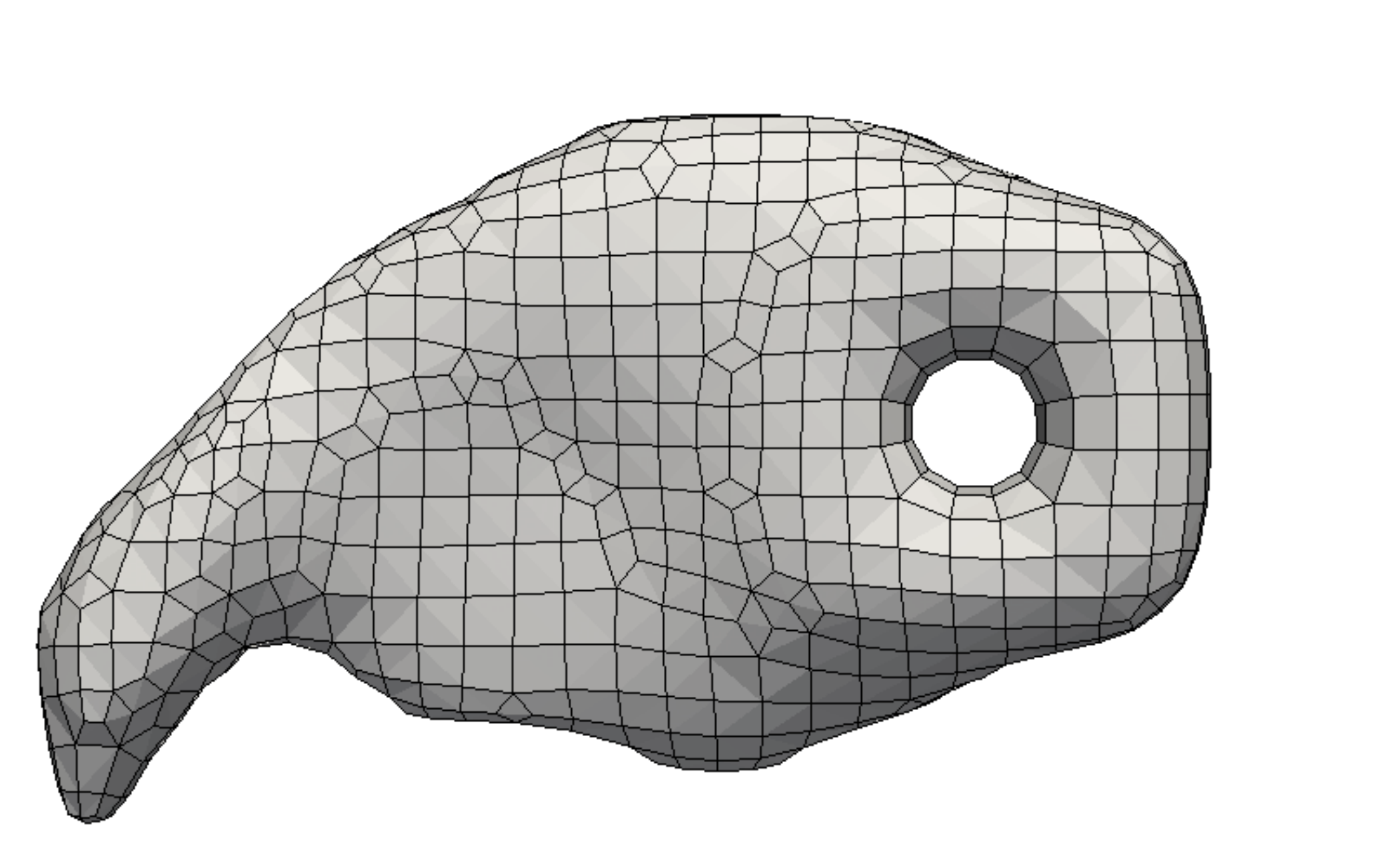}} \\\hline
     \multirow{2}{*}{\rotatebox{90}{trivial+PDE}} & & \parbox[m]{6em}{\includegraphics[trim={0cm 0cm 0cm 0cm},clip, width=0.12\textwidth]{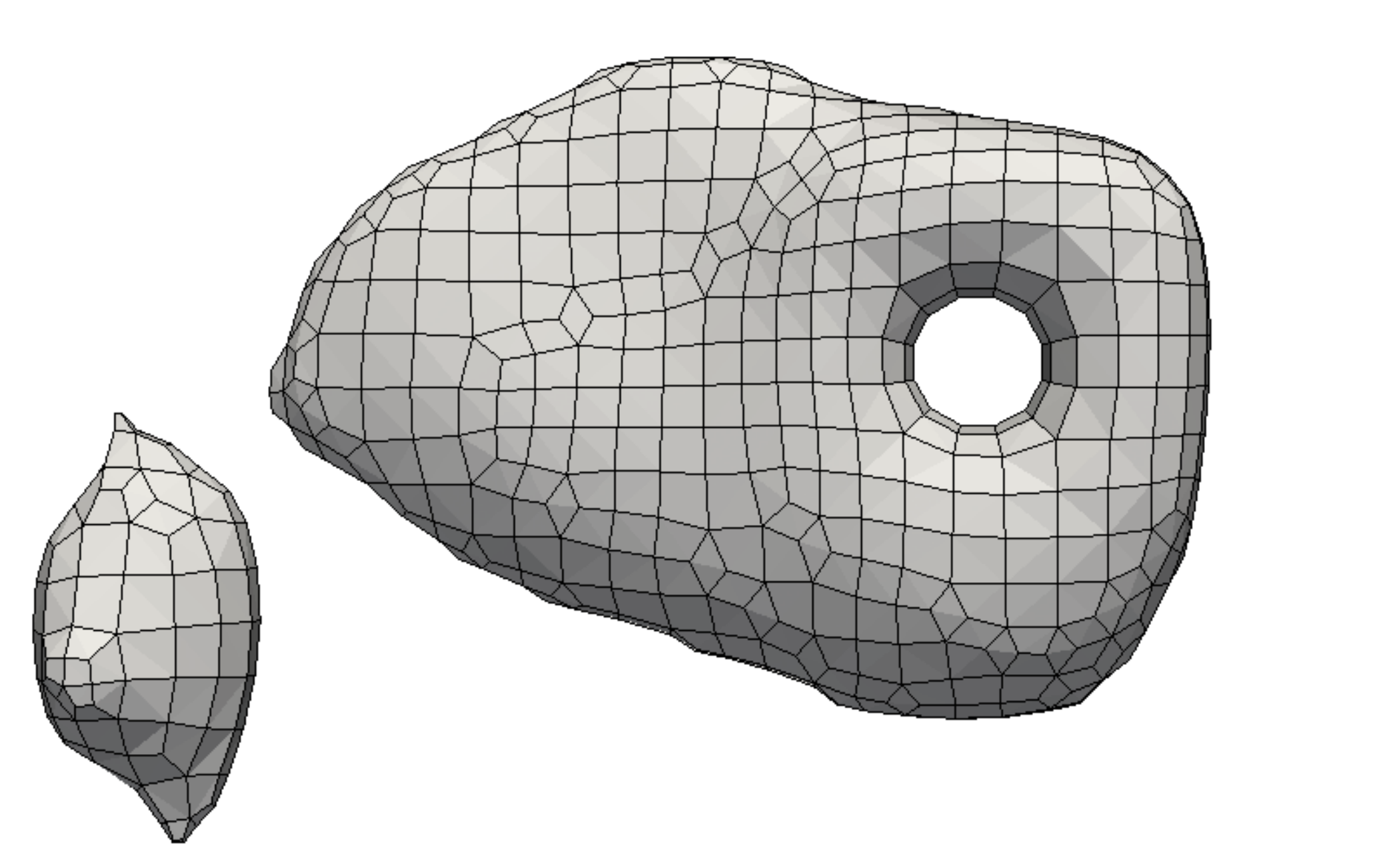}} & \parbox[m]{6em}{\includegraphics[trim={0cm 0cm 0cm 0cm},clip, width=0.12\textwidth]{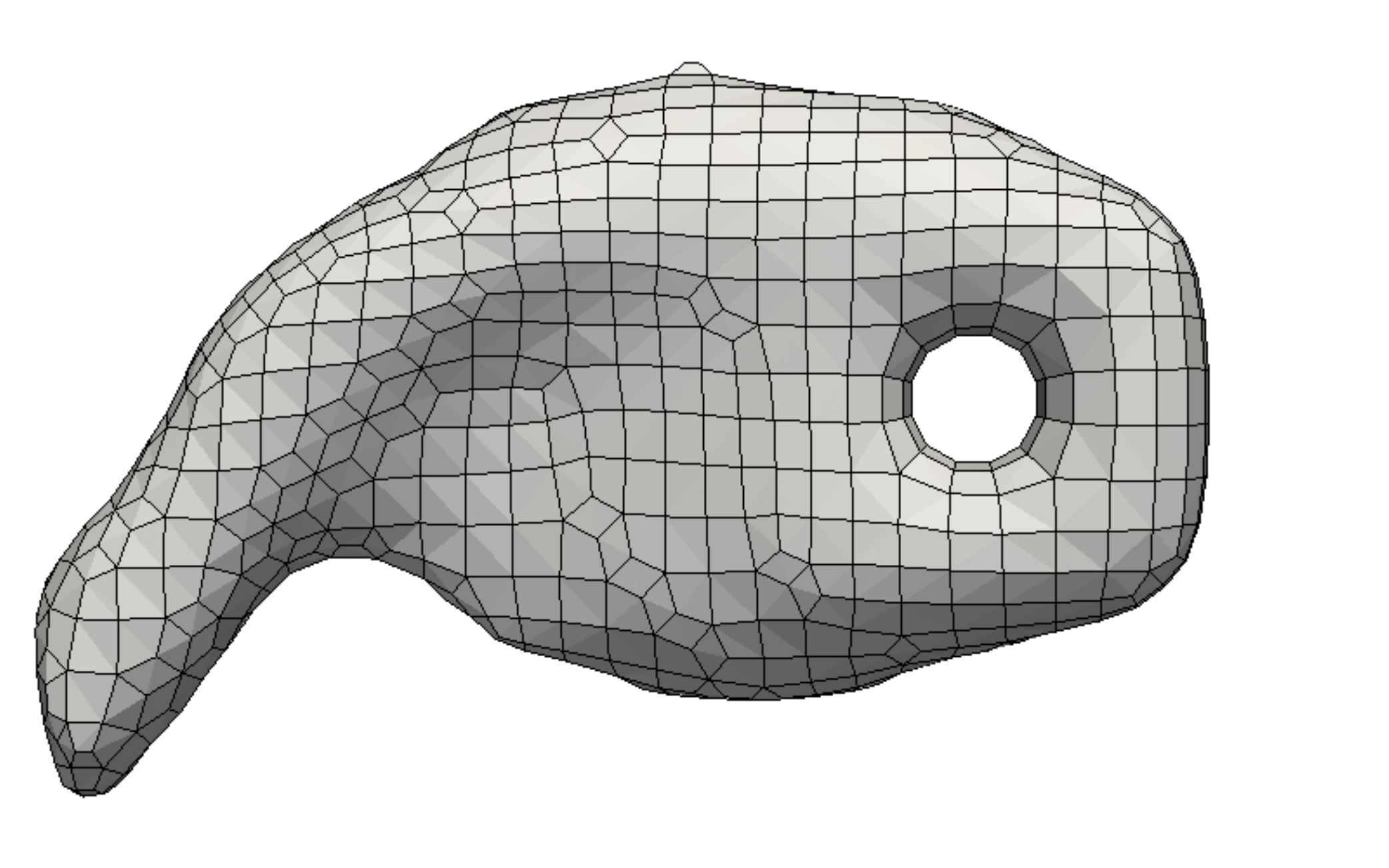}} & \parbox[m]{6em}{\includegraphics[trim={0cm 0cm 0cm 0cm},clip, width=0.12\textwidth]{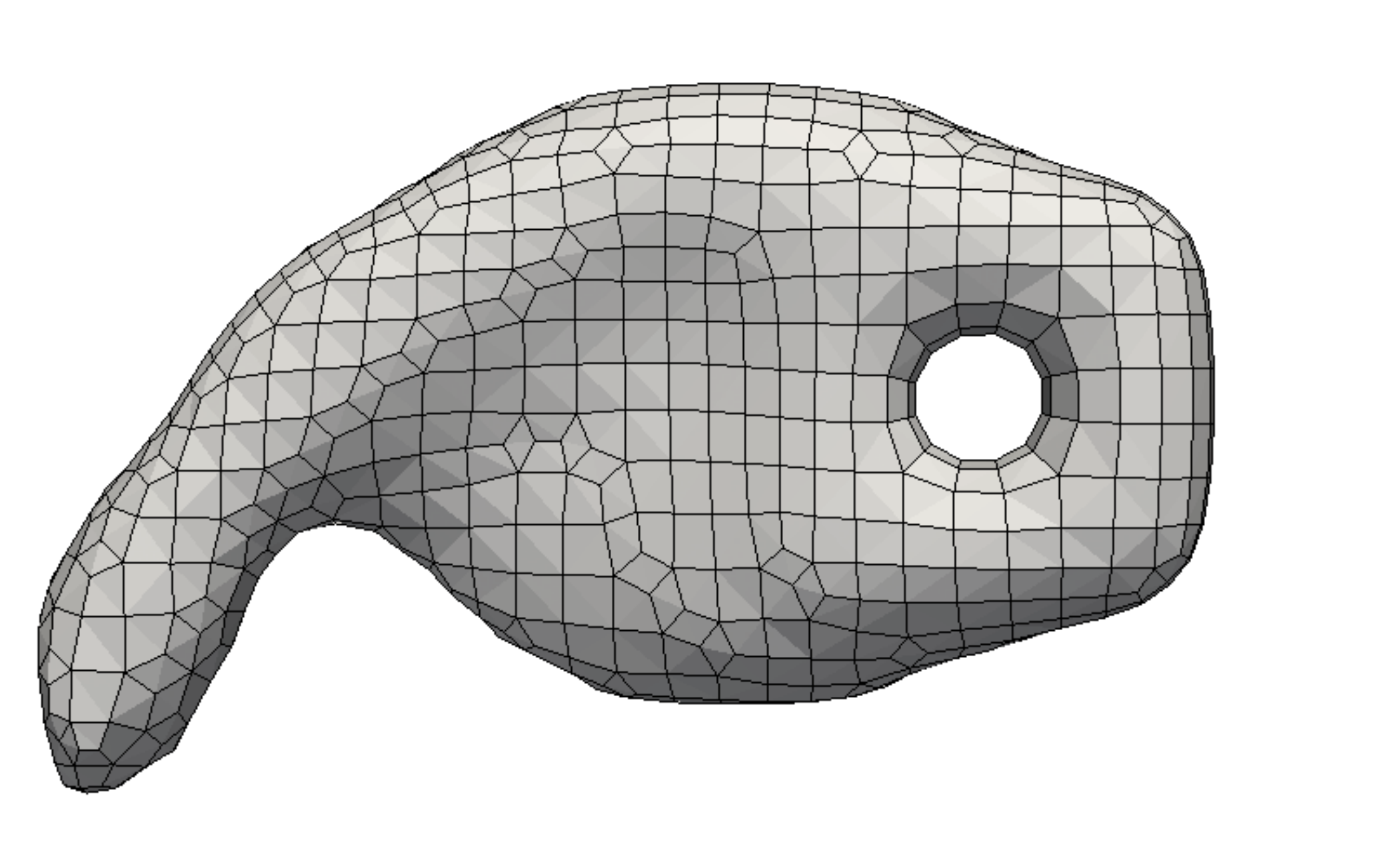}} & \parbox[m]{6em}{\includegraphics[trim={0cm 0cm 0cm 0cm},clip, width=0.12\textwidth]{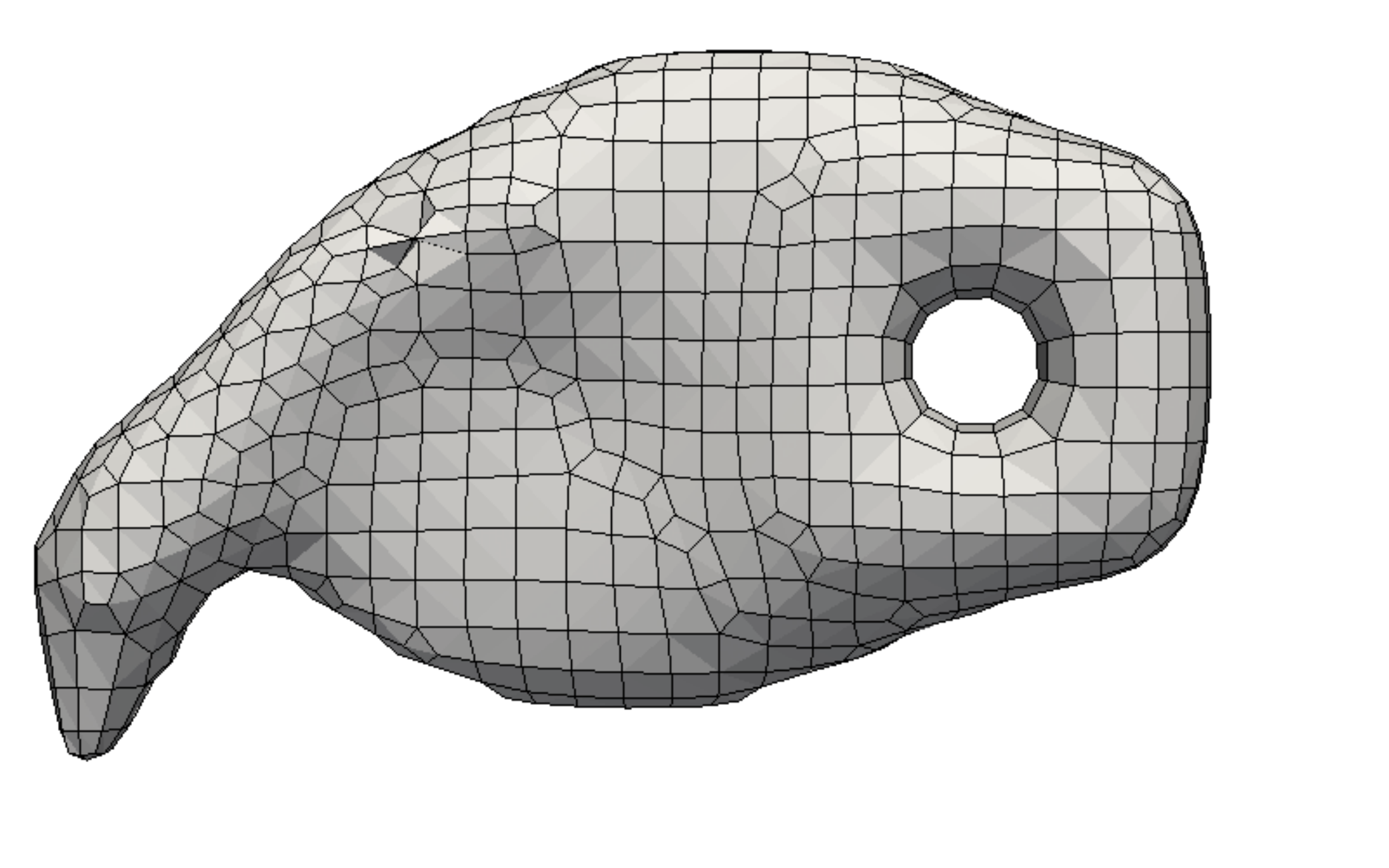}} & \parbox[m]{6em}{\includegraphics[trim={0cm 0cm 0cm 0cm},clip, width=0.12\textwidth]{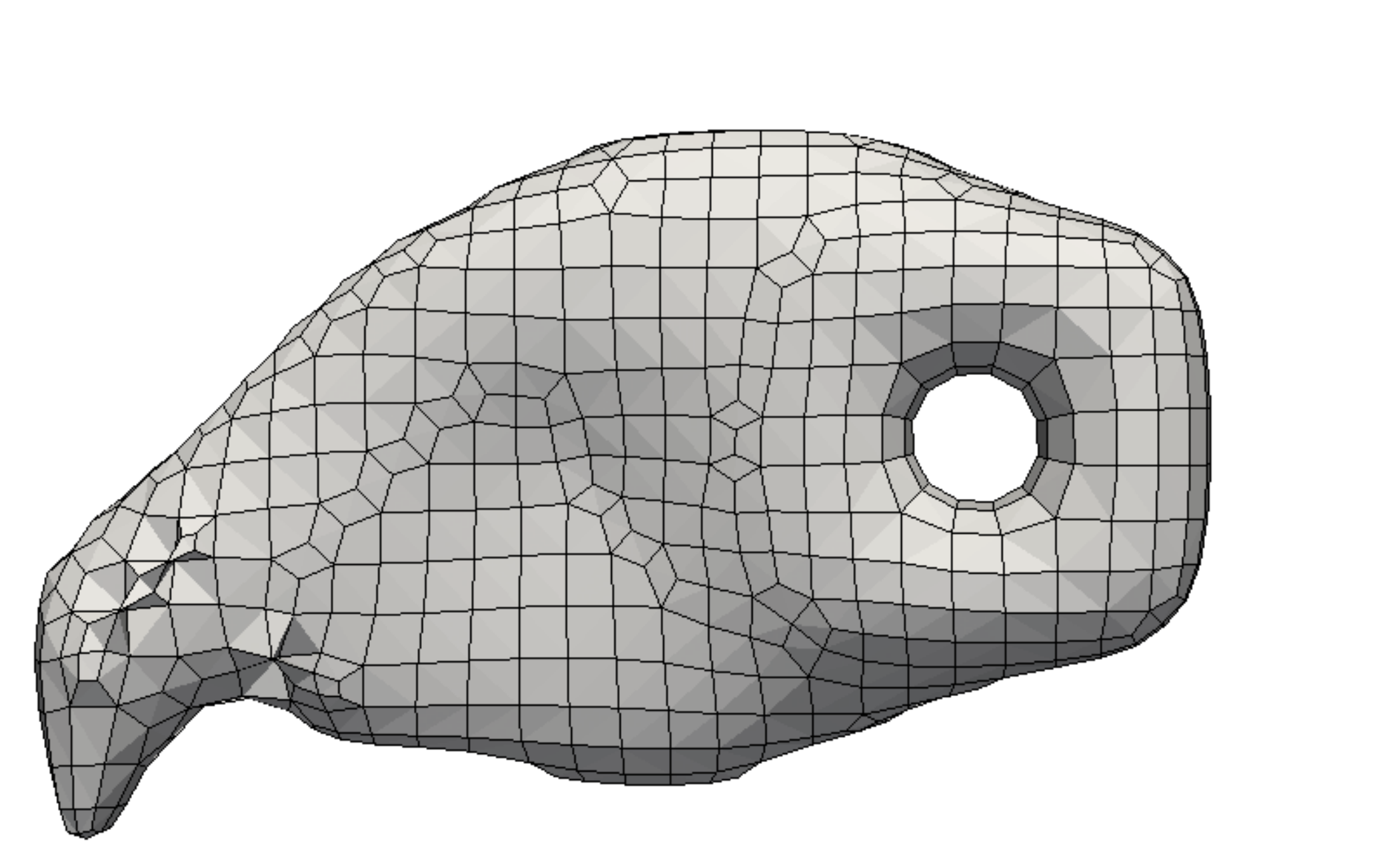}} \\\cline{2-7}
     & \checkmark & \parbox[m]{6em}{\includegraphics[trim={0cm 0cm 0cm 0cm},clip, width=0.12\textwidth]{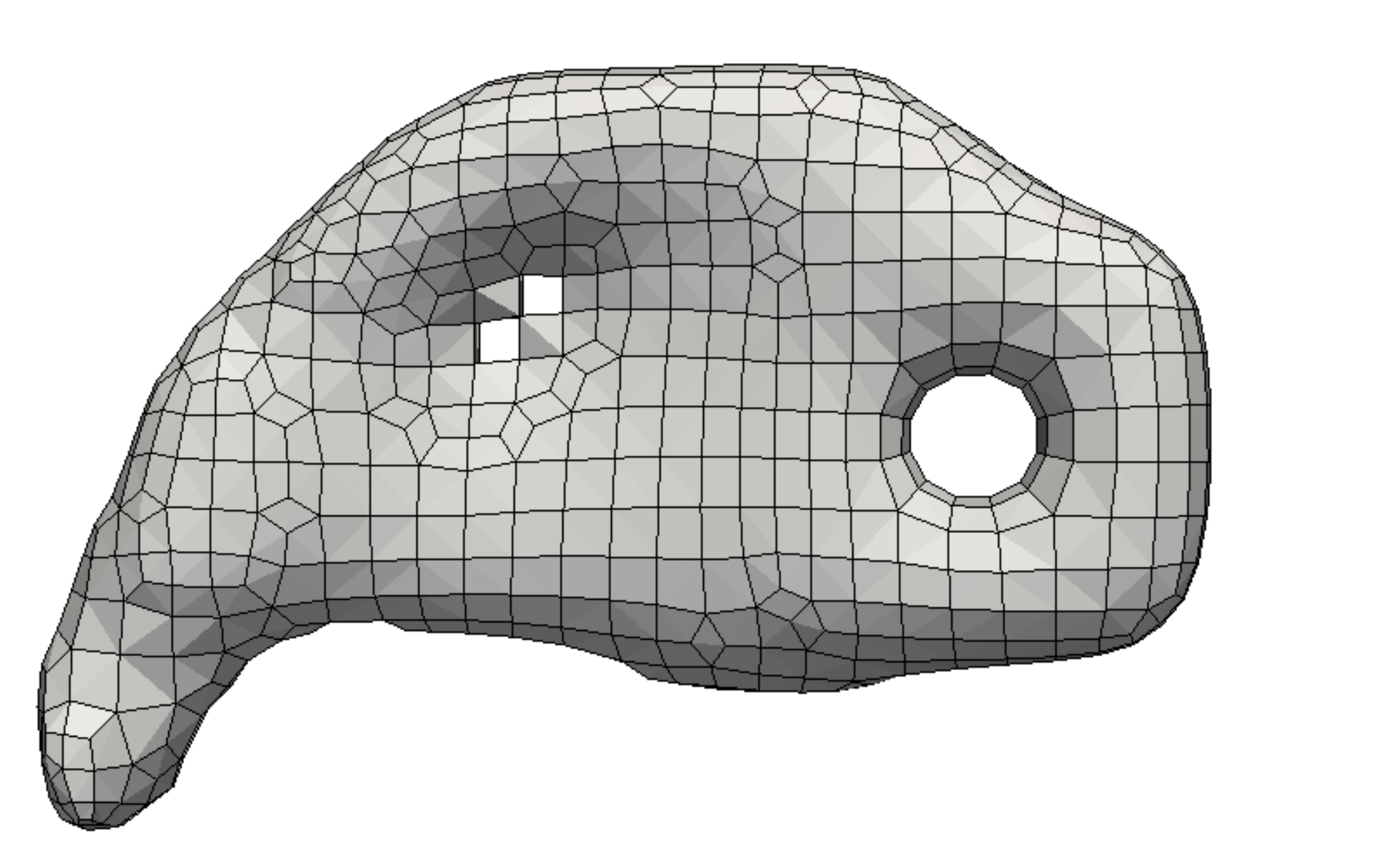}} & \parbox[m]{6em}{\includegraphics[trim={0cm 0cm 0cm 0cm},clip, width=0.12\textwidth]{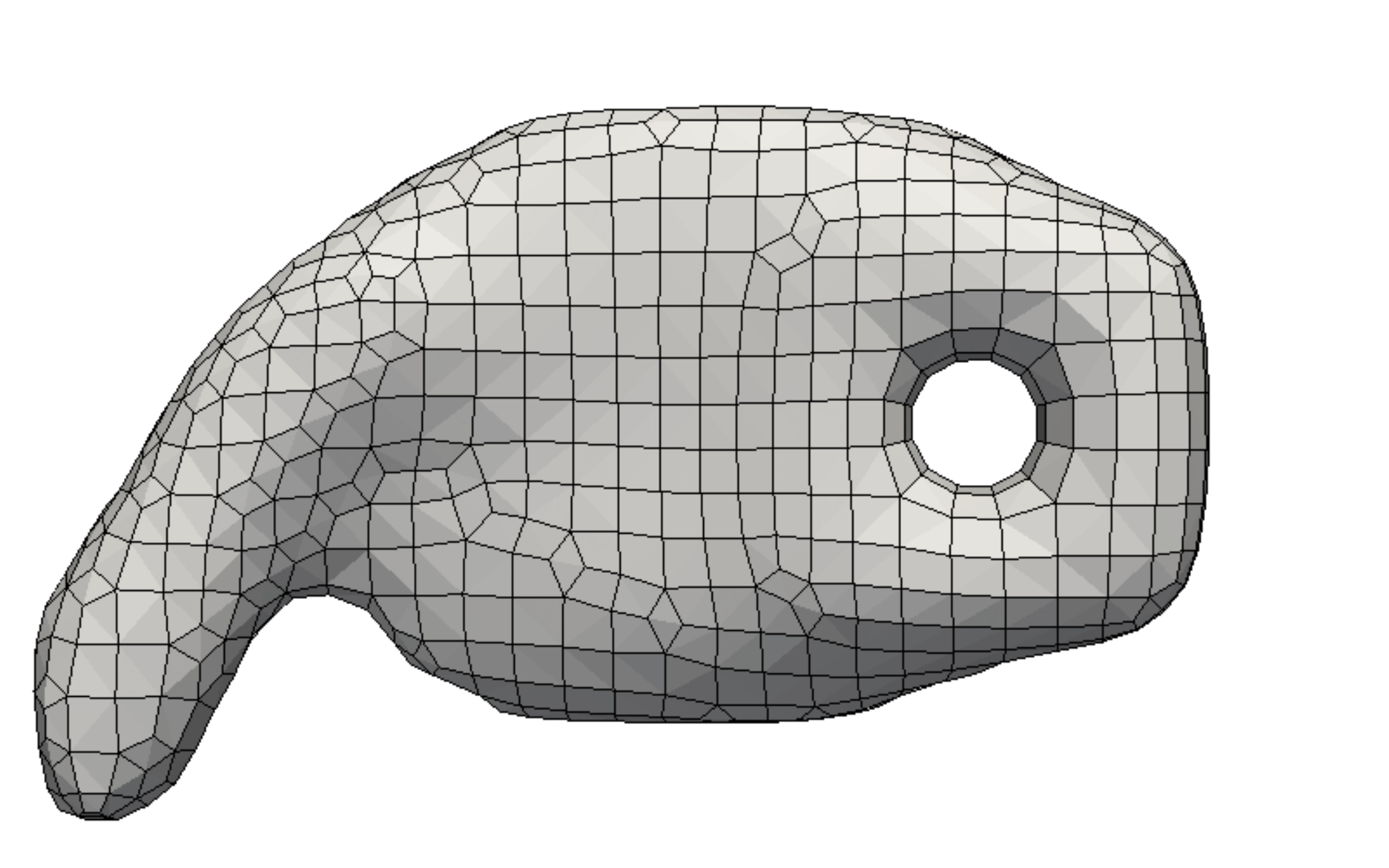}} & \parbox[m]{6em}{\includegraphics[trim={0cm 0cm 0cm 0cm},clip, width=0.12\textwidth]{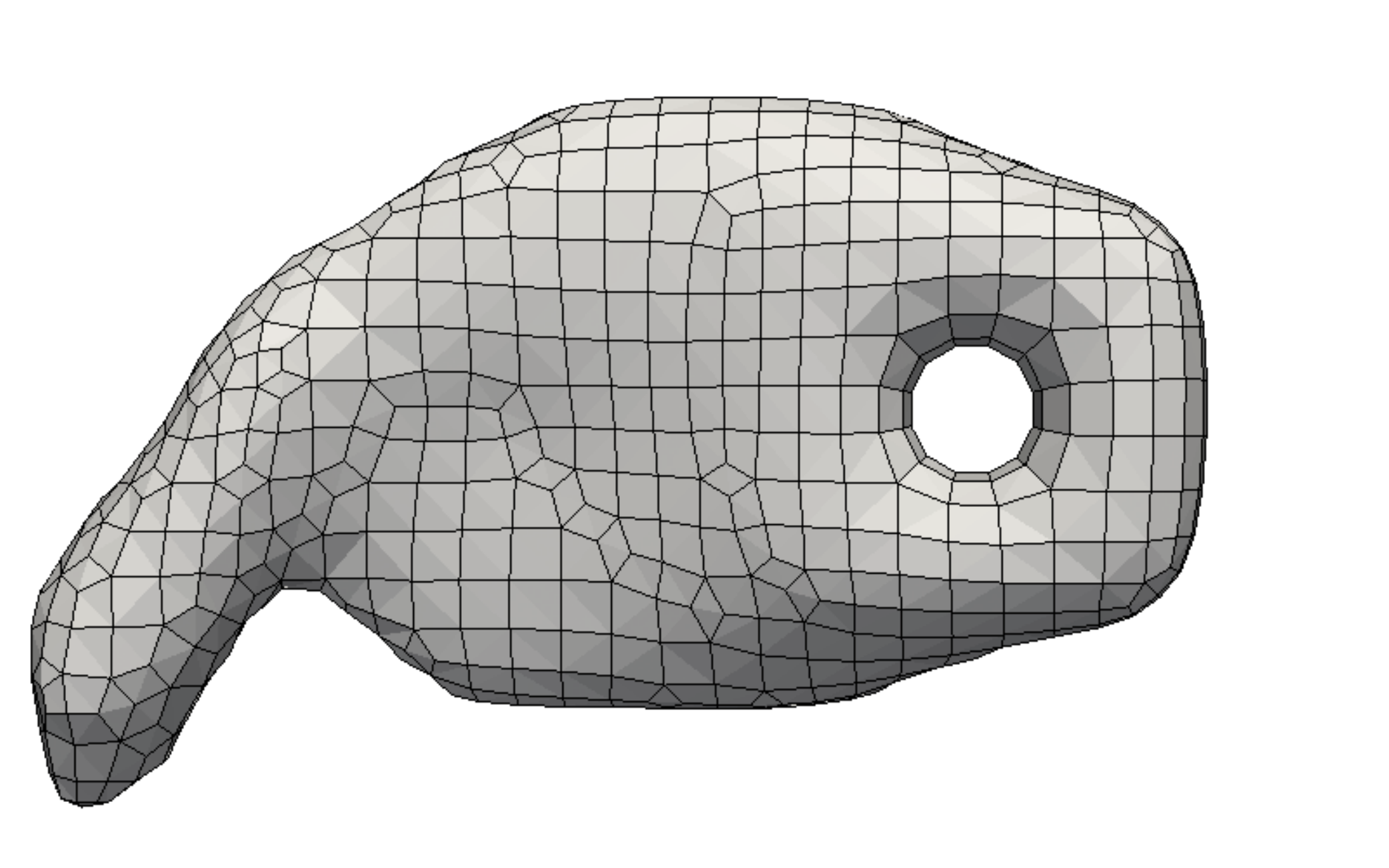}} & \parbox[m]{6em}{\includegraphics[trim={0cm 0cm 0cm 0cm},clip, width=0.12\textwidth]{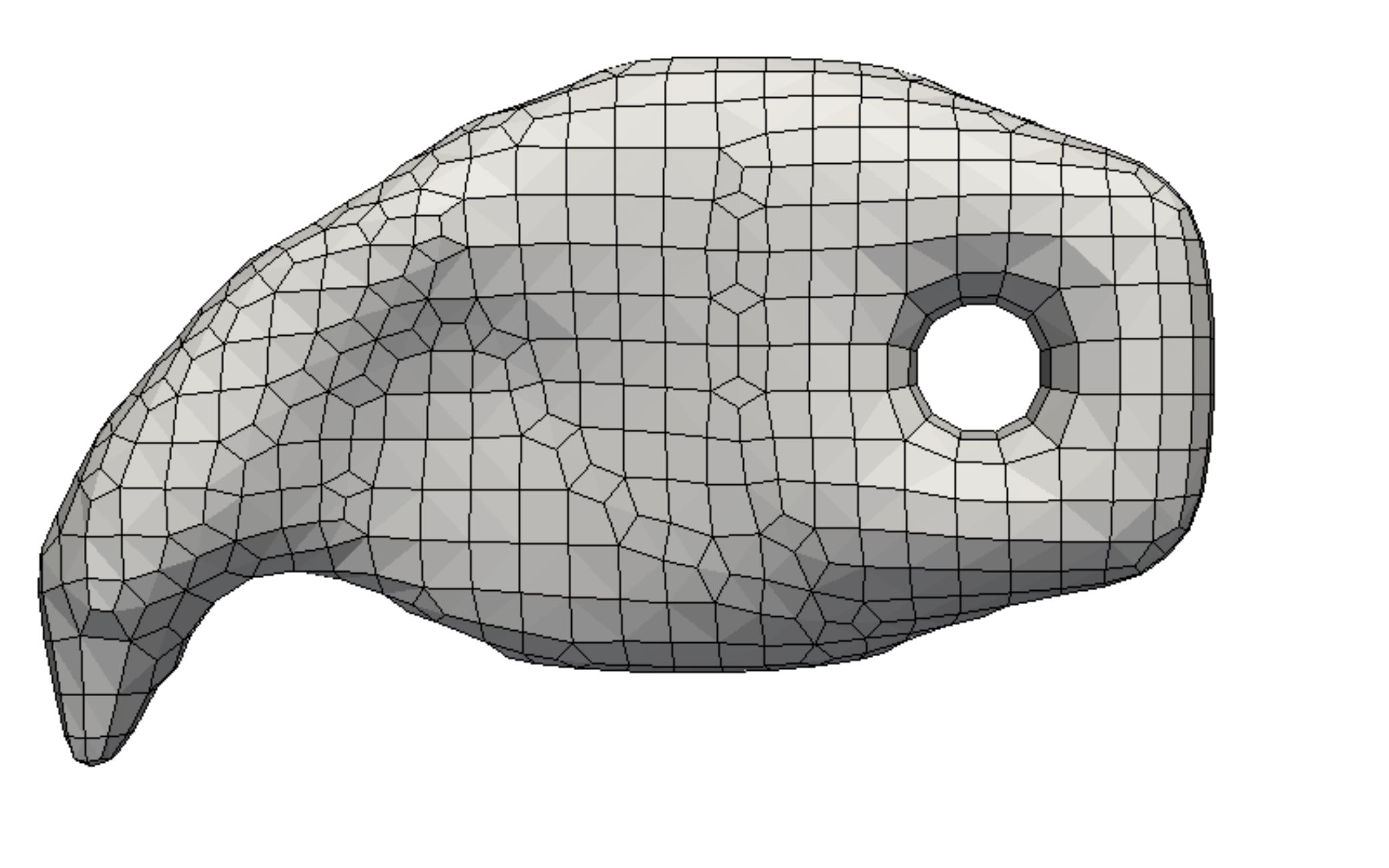}} & \parbox[m]{6em}{\includegraphics[trim={0cm 0cm 0cm 0cm},clip, width=0.12\textwidth]{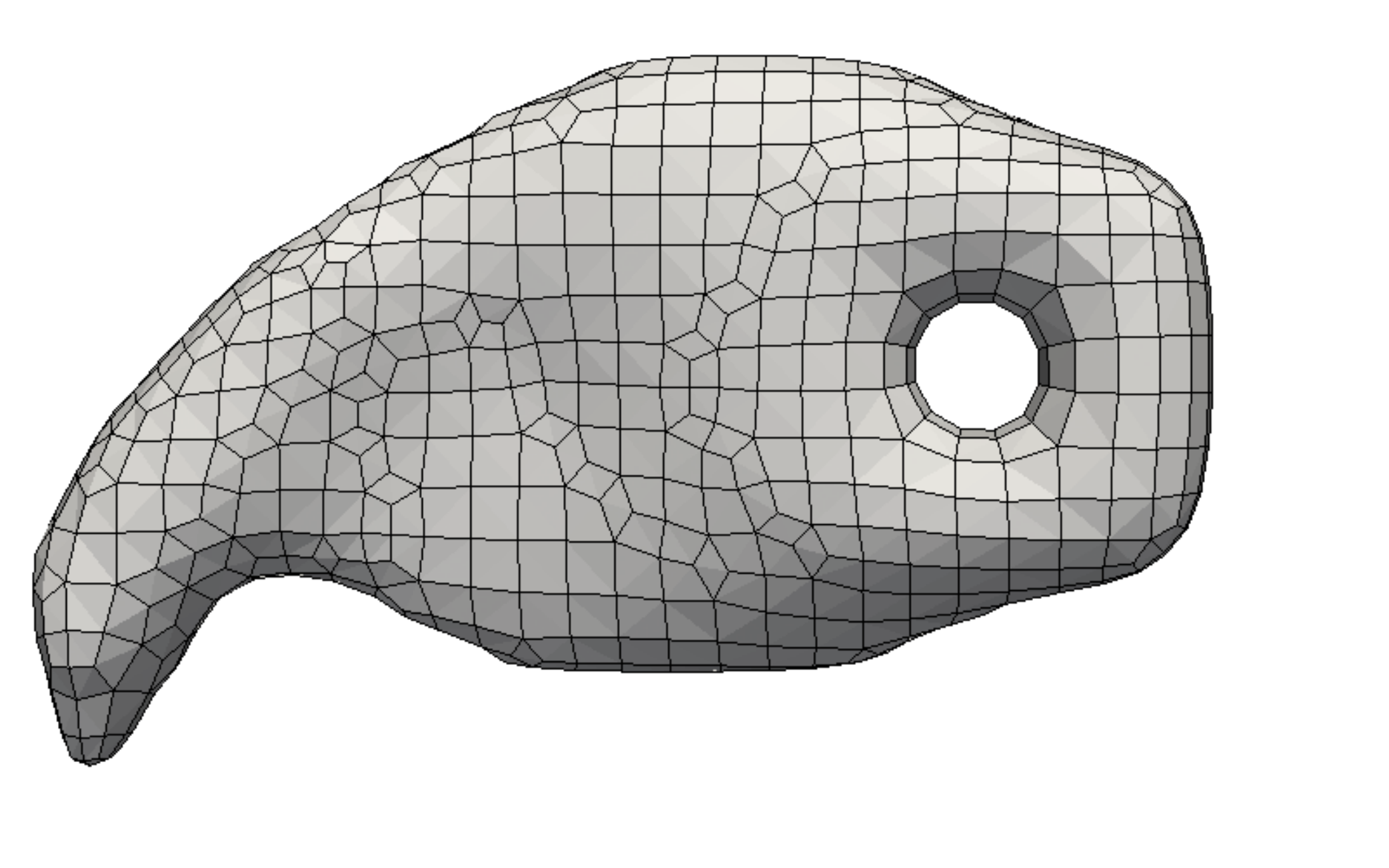}}\\
     \multicolumn{7}{c}{\fbox{\hspace{0.3cm}\parbox[m]{6em}{\includegraphics[trim={0cm 0cm 0cm 0cm},clip, width=0.12\textwidth]{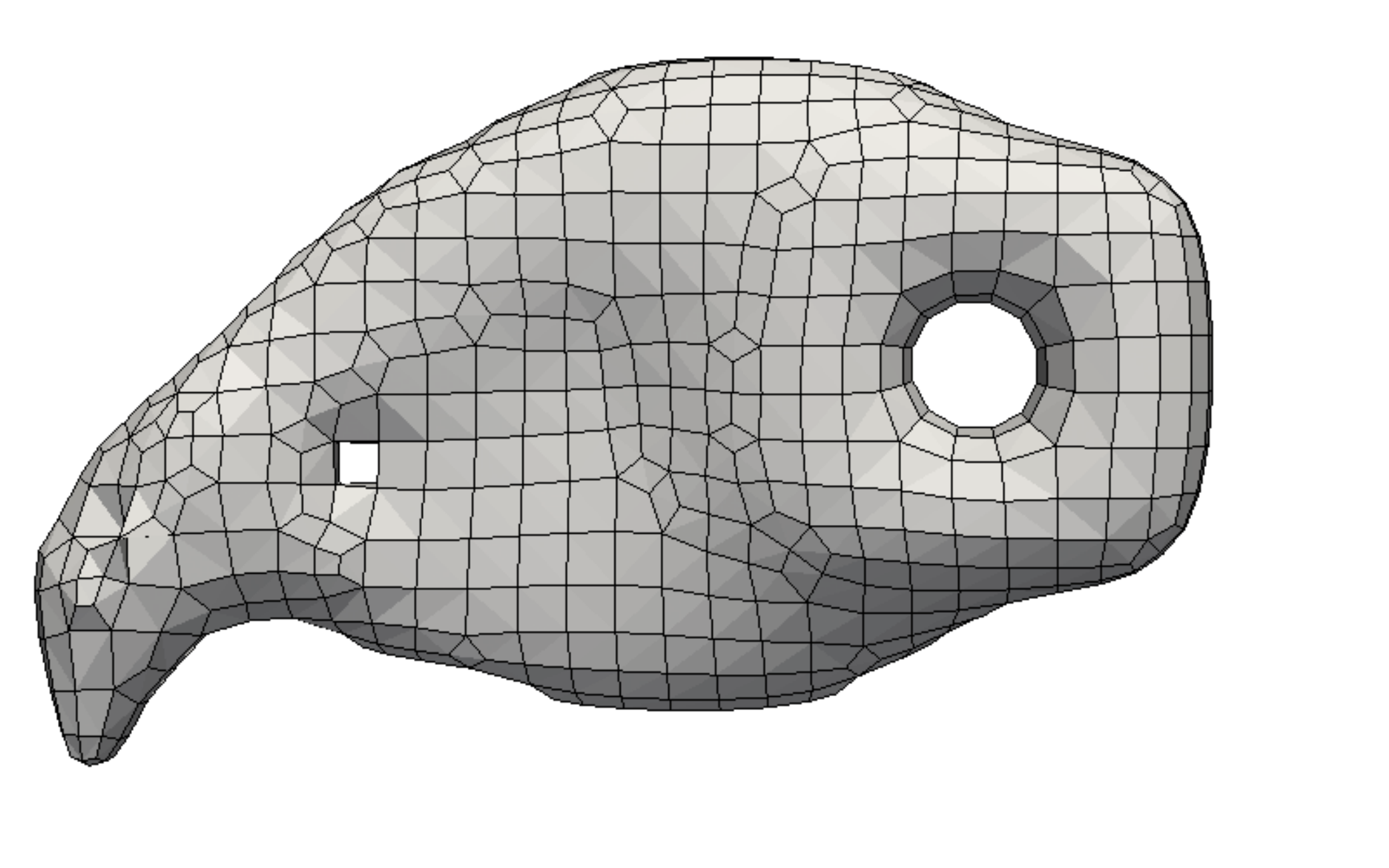}}}}
\end{tabular}
\end{subtable}
\end{table}

\begin{table}[]
\caption{Model predictions of two different problems from the \textbf{disc complex} validation dataset, using the UNet with different preprocessings and equivariances. We train the models on subsets of the dataset and vary the training size along the columns of the table. At the boxes below the tables we show the corresponding ground truth density for each problem.}
\label{5_fig:disc_complex_predictions}
\begin{subtable}[h]{0.99\textwidth}
    \centering\setcellgapes{3pt}\makegapedcells
    \setlength\tabcolsep{3.5pt}
    \begin{tabular}{c|c||ScScScScSc}
    \multicolumn{2}{c||}{} & \multicolumn{5}{c}{training samples} \\\hline
     prepr. & equiv. & 10 & 50 & 100 & 500 & 1500 \\\hline
     \multirow{2}{*}{\rotatebox{90}{trivial}} & & \parbox[m]{6em}{\includegraphics[trim={0cm 0cm 0cm 0cm},clip, width=0.12\textwidth]{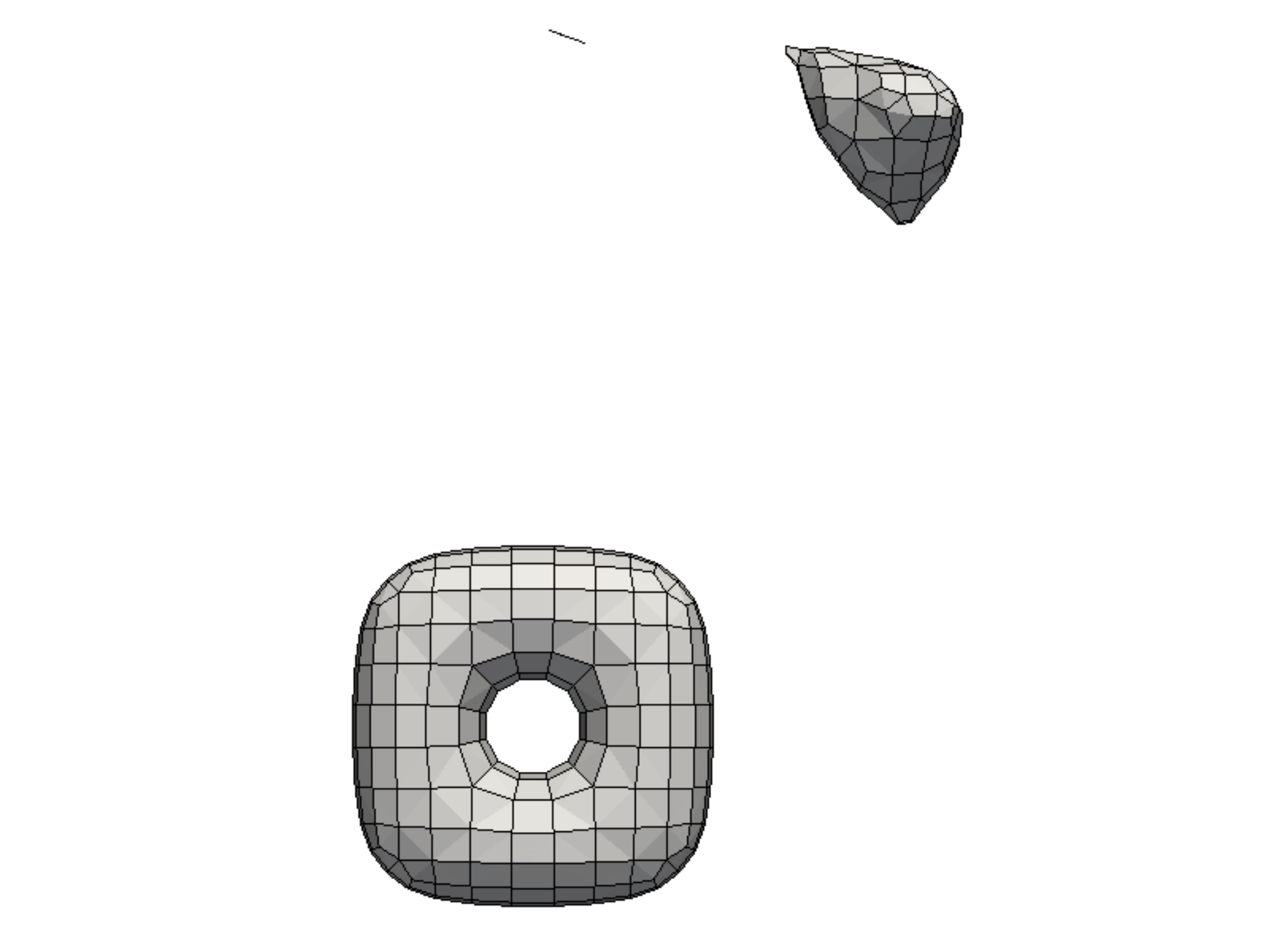}} & \parbox[m]{6em}{\includegraphics[trim={0cm 0cm 0cm 0cm},clip, width=0.12\textwidth]{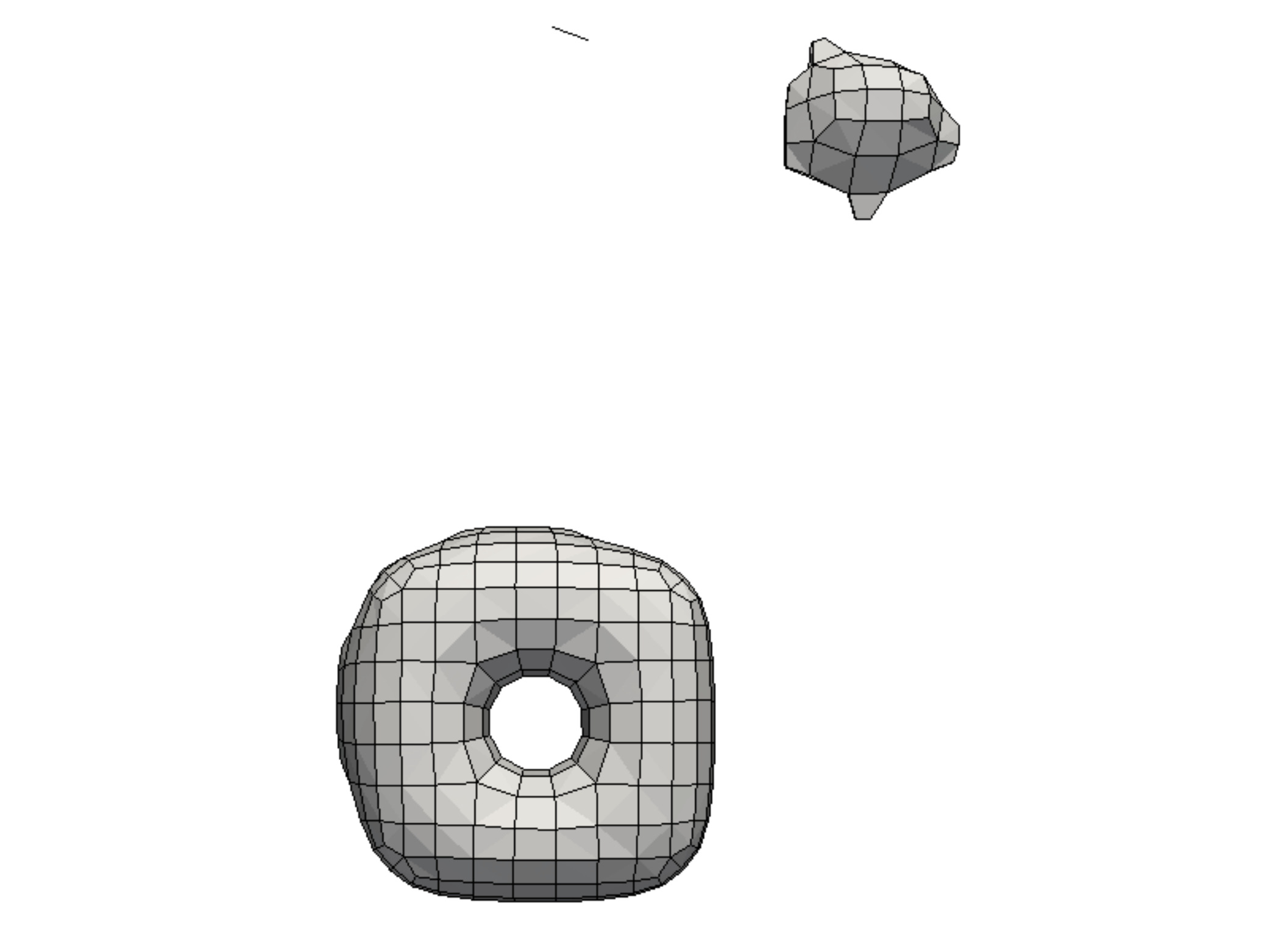}} & \parbox[m]{6em}{\includegraphics[trim={0cm 0cm 0cm 0cm},clip, width=0.12\textwidth]{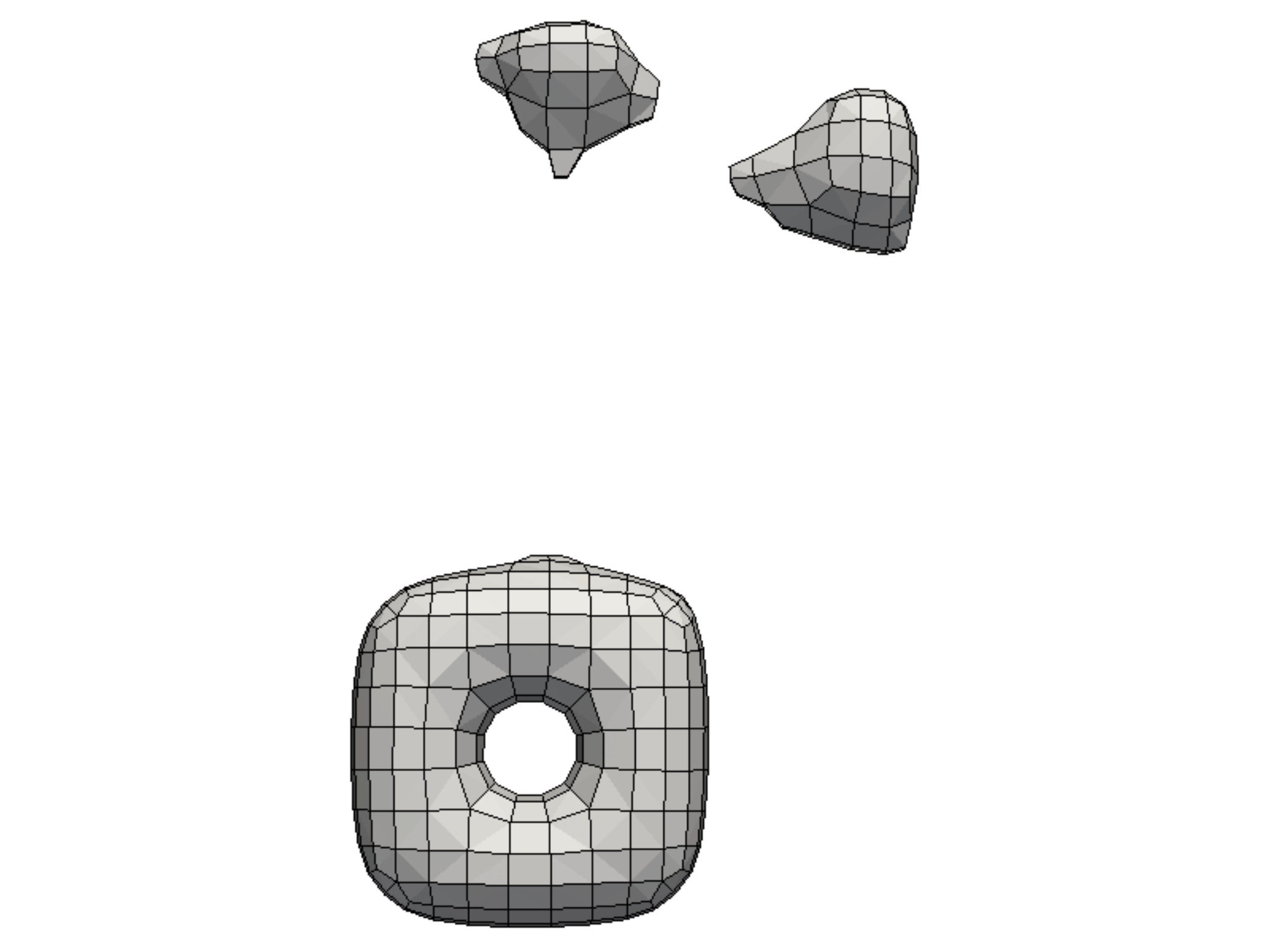}} & \parbox[m]{6em}{\includegraphics[trim={0cm 0cm 0cm 0cm},clip, width=0.12\textwidth]{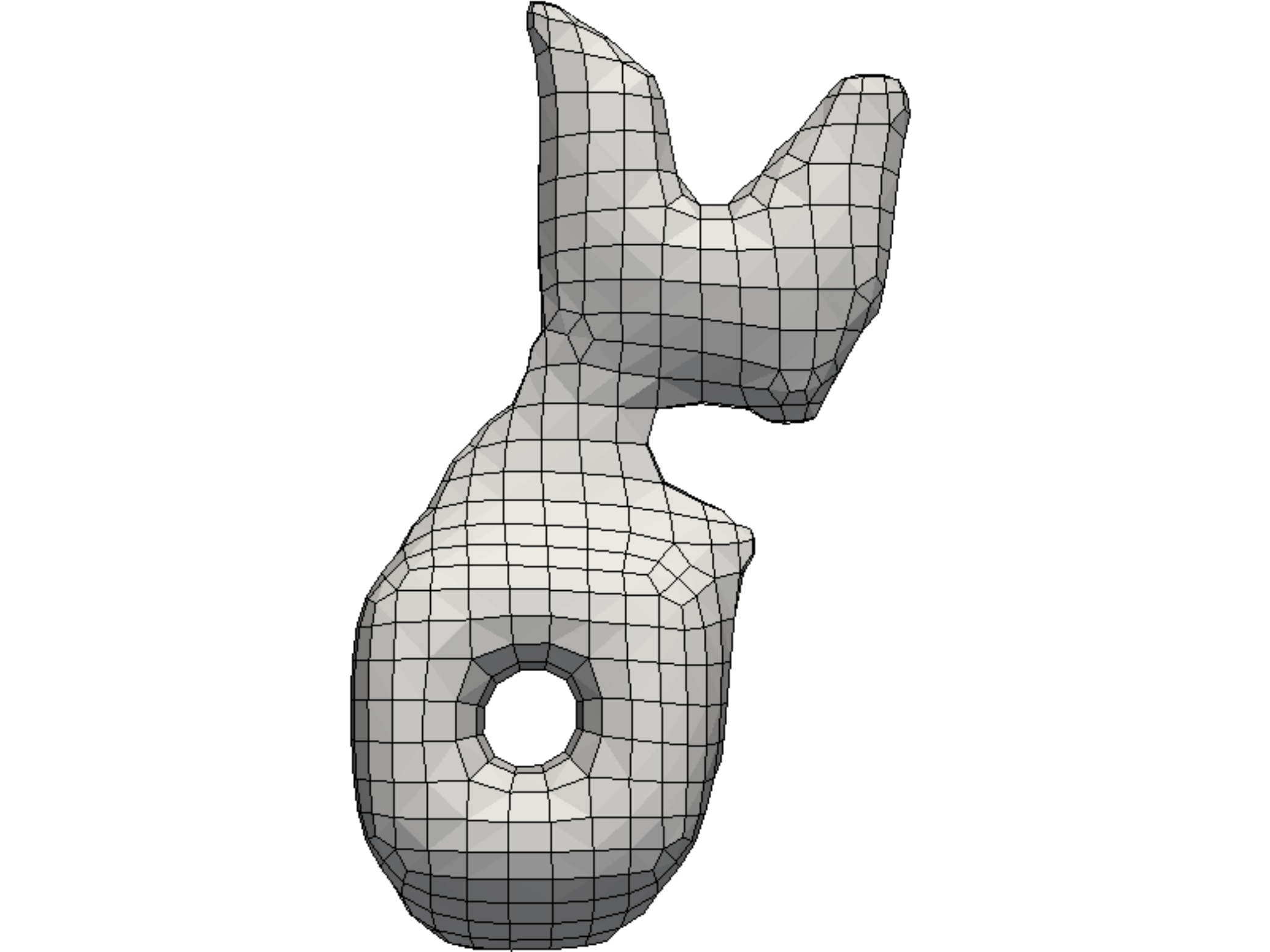}} & \parbox[m]{6em}{\includegraphics[trim={0cm 0cm 0cm 0cm},clip, width=0.12\textwidth]{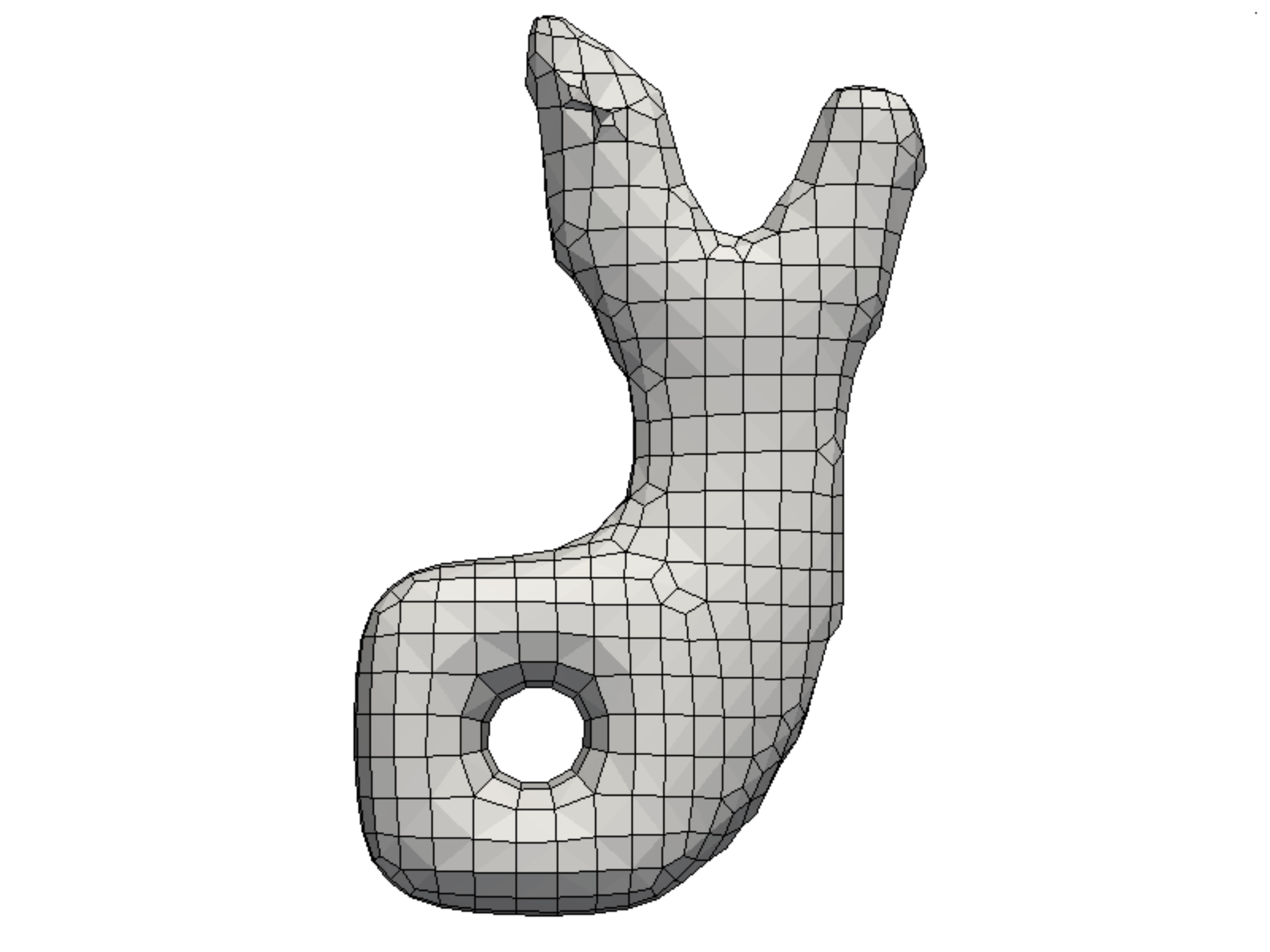}} \\\cline{2-7}
     & \checkmark
     & \parbox[m]{6em}{\includegraphics[trim={0cm 0cm 0cm 0cm},clip, width=0.12\textwidth]{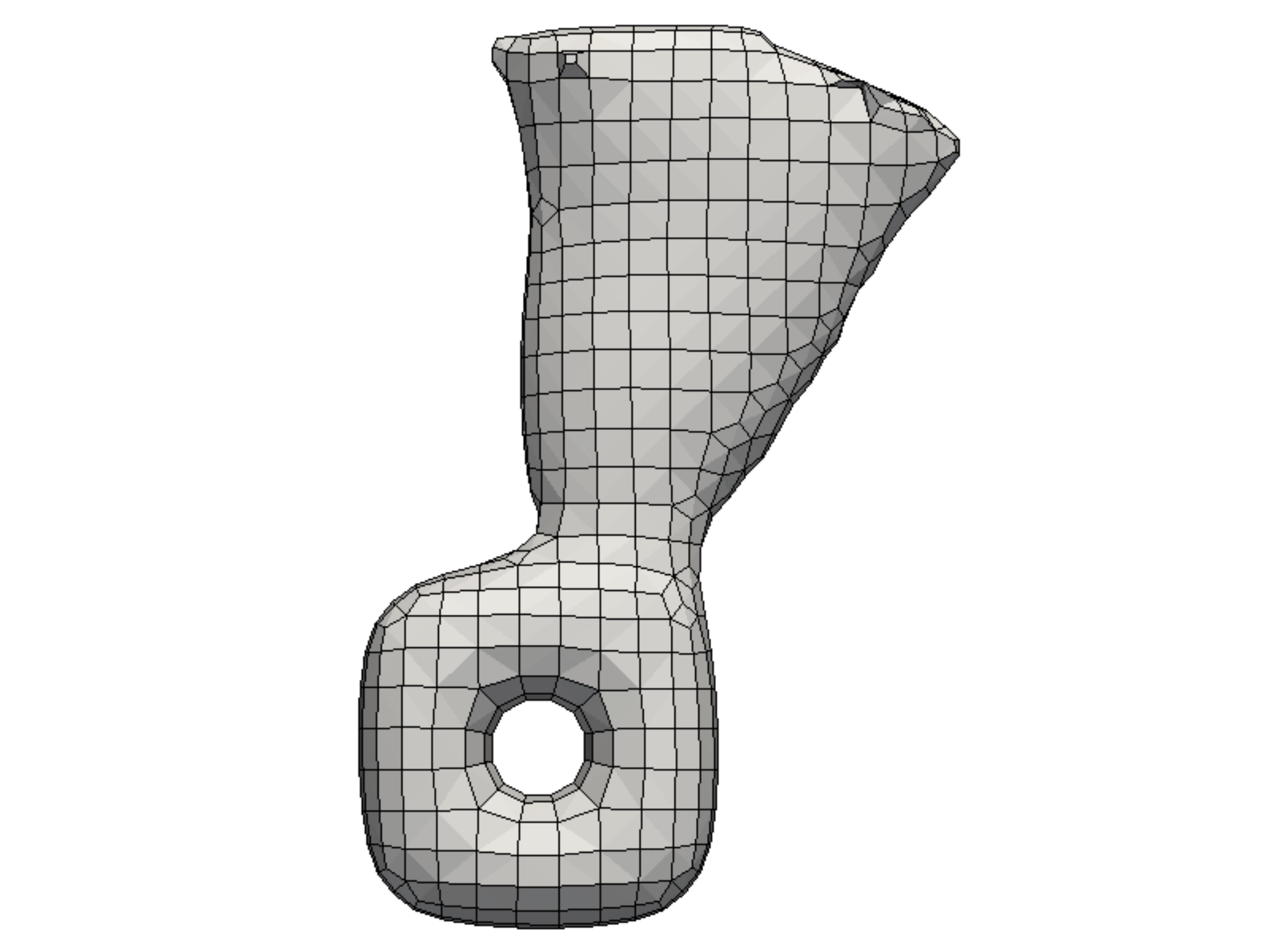}} & \parbox[m]{6em}{\includegraphics[trim={0cm 0cm 0cm 0cm},clip, width=0.12\textwidth]{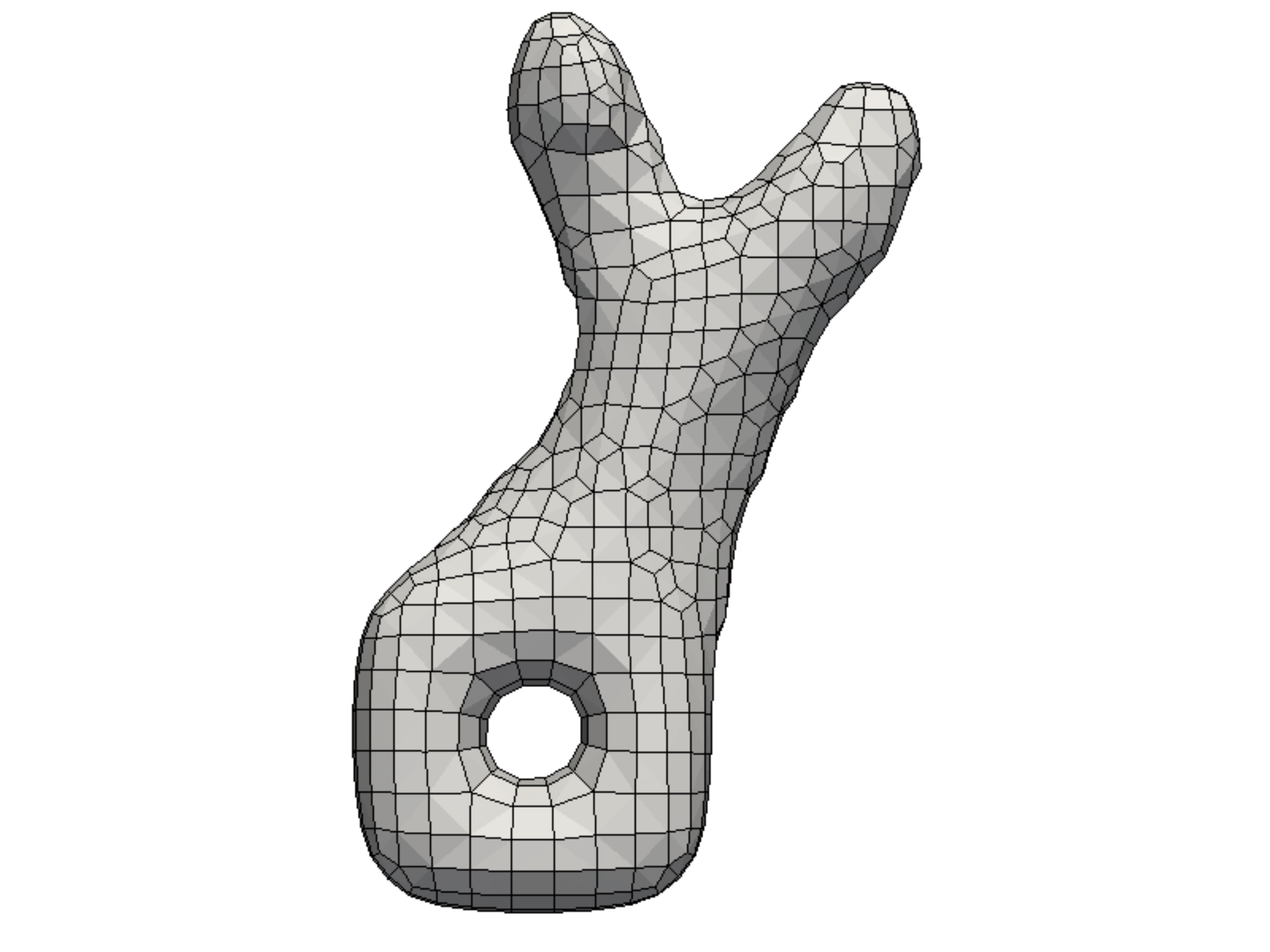}} & \parbox[m]{6em}{\includegraphics[trim={0cm 0cm 0cm 0cm},clip, width=0.12\textwidth]{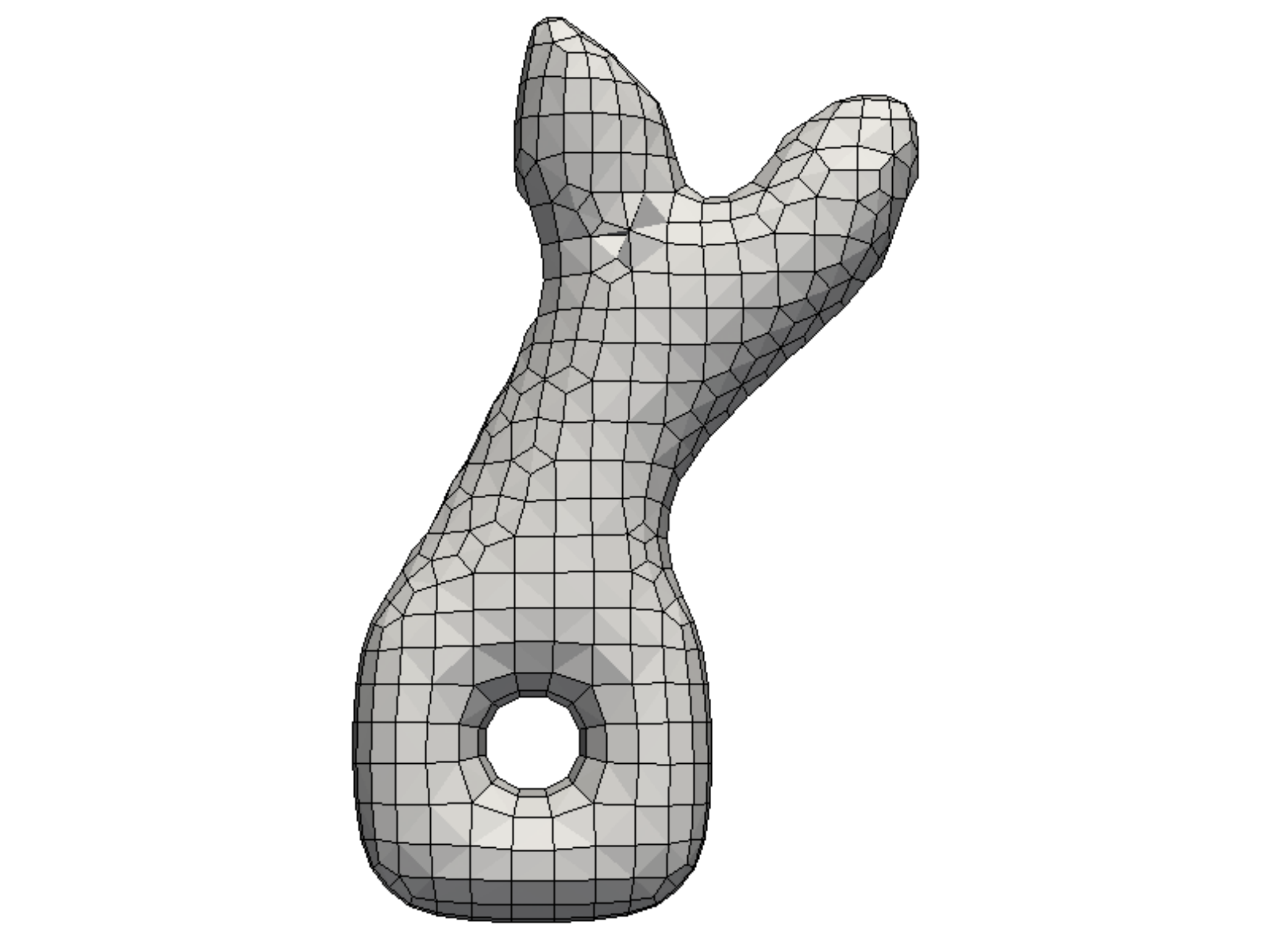}} & \parbox[m]{6em}{\includegraphics[trim={0cm 0cm 0cm 0cm},clip, width=0.12\textwidth]{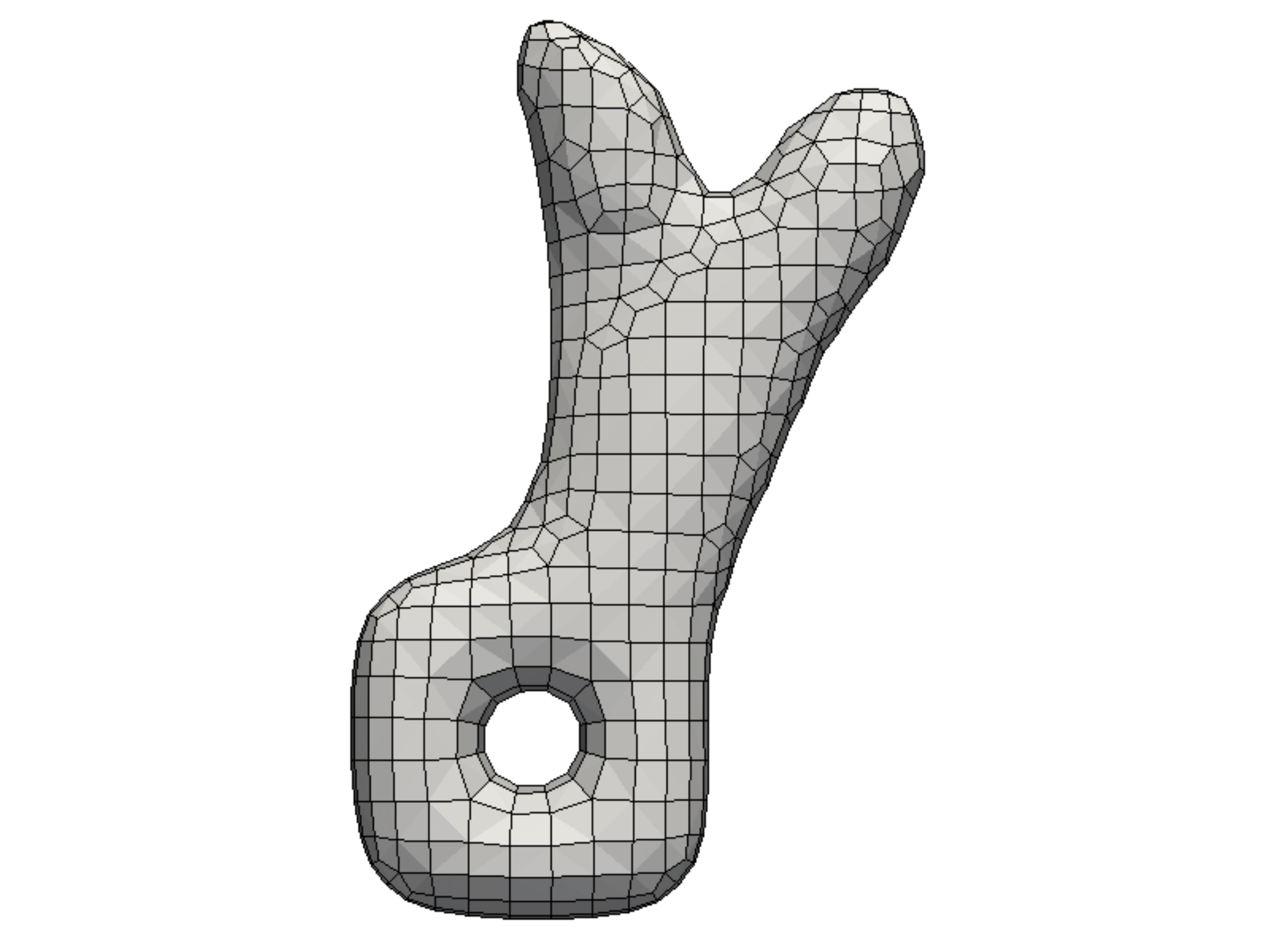}} & \parbox[m]{6em}{\includegraphics[trim={0cm 0cm 0cm 0cm},clip, width=0.12\textwidth]{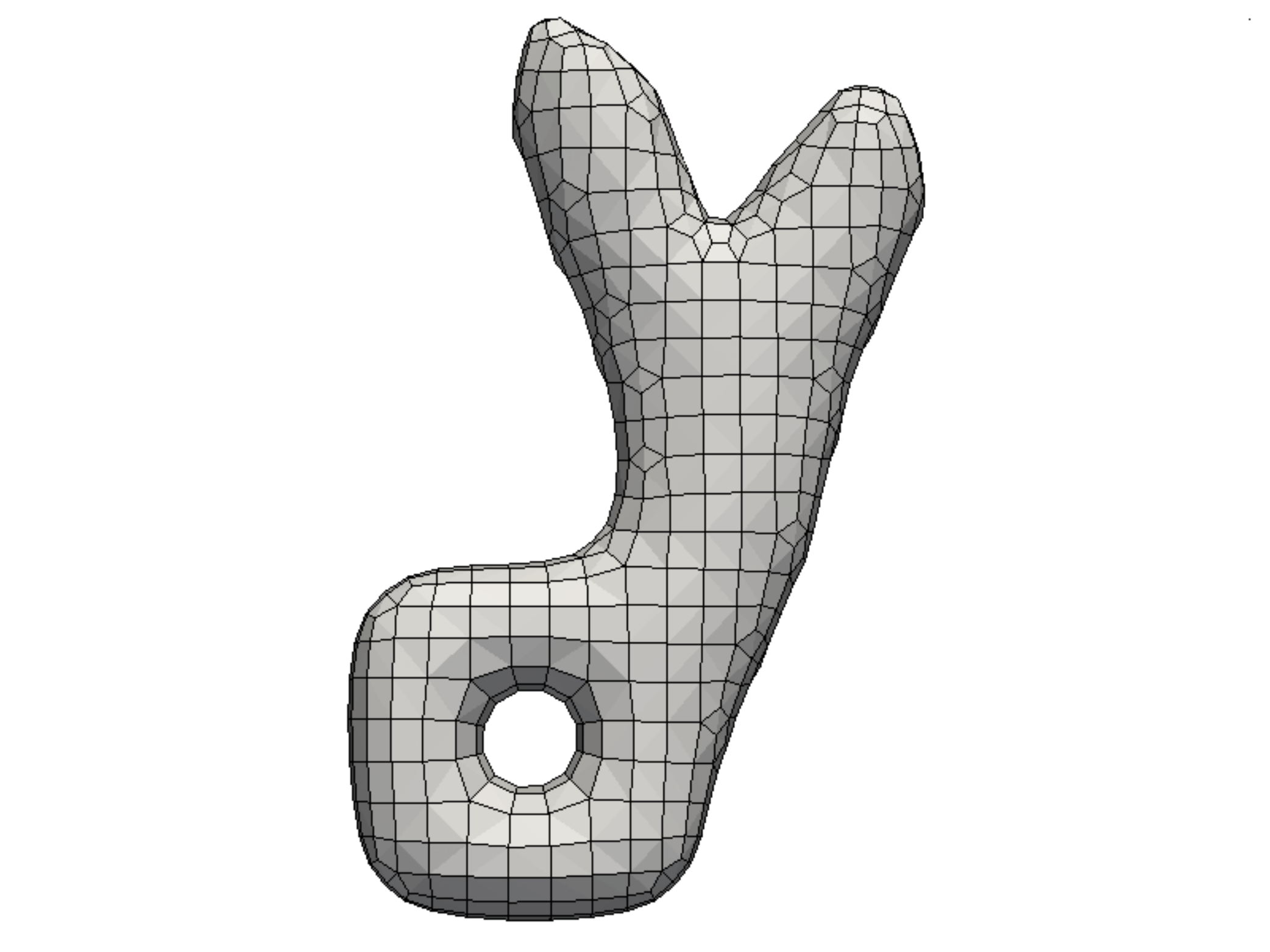}} \\\hline
     \multirow{2}{*}{\rotatebox{90}{trivial+PDE}} & & \parbox[m]{6em}{\includegraphics[trim={0cm 0cm 0cm 0cm},clip, width=0.12\textwidth]{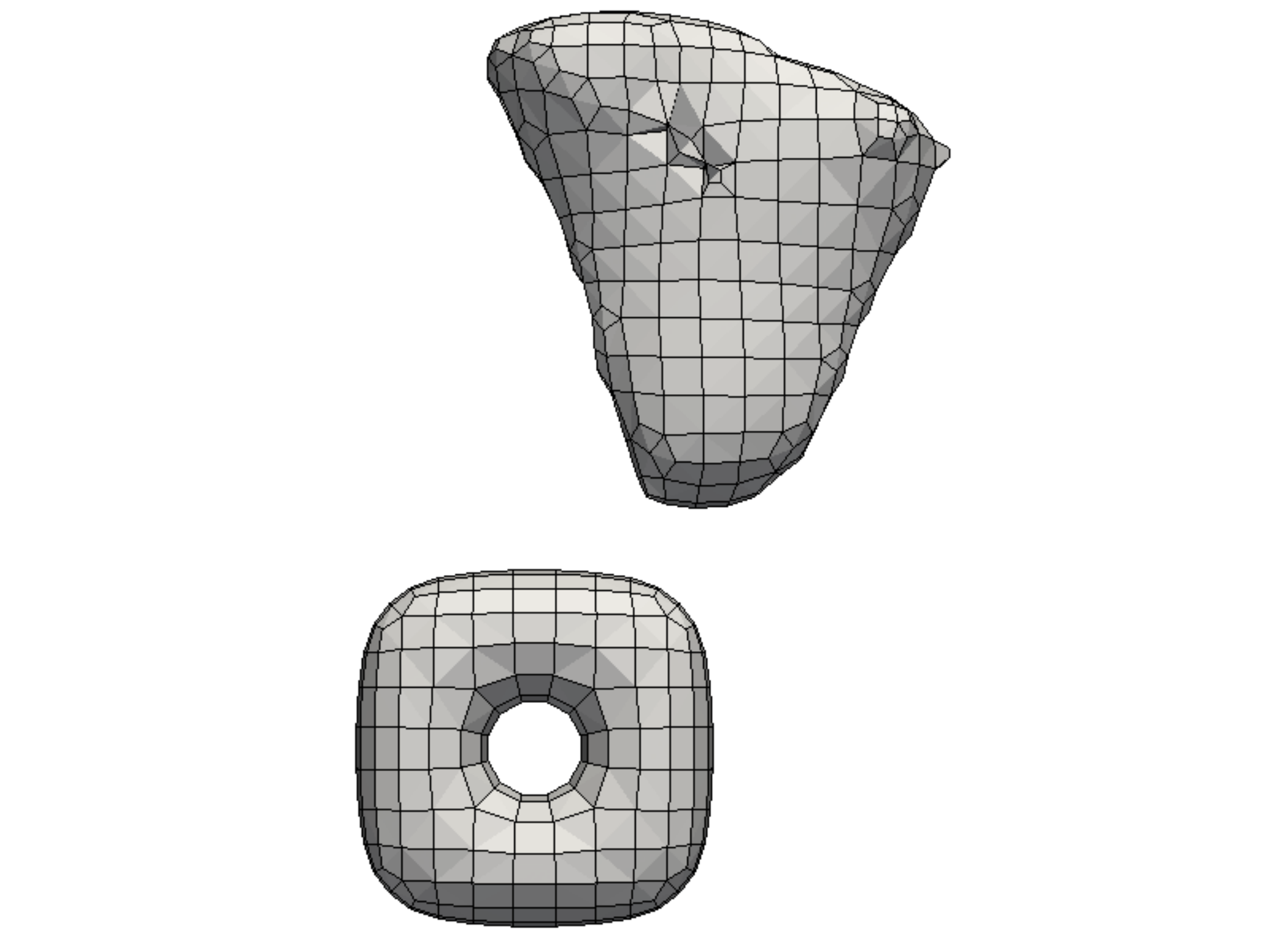}} & \parbox[m]{6em}{\includegraphics[trim={0cm 0cm 0cm 0cm},clip, width=0.12\textwidth]{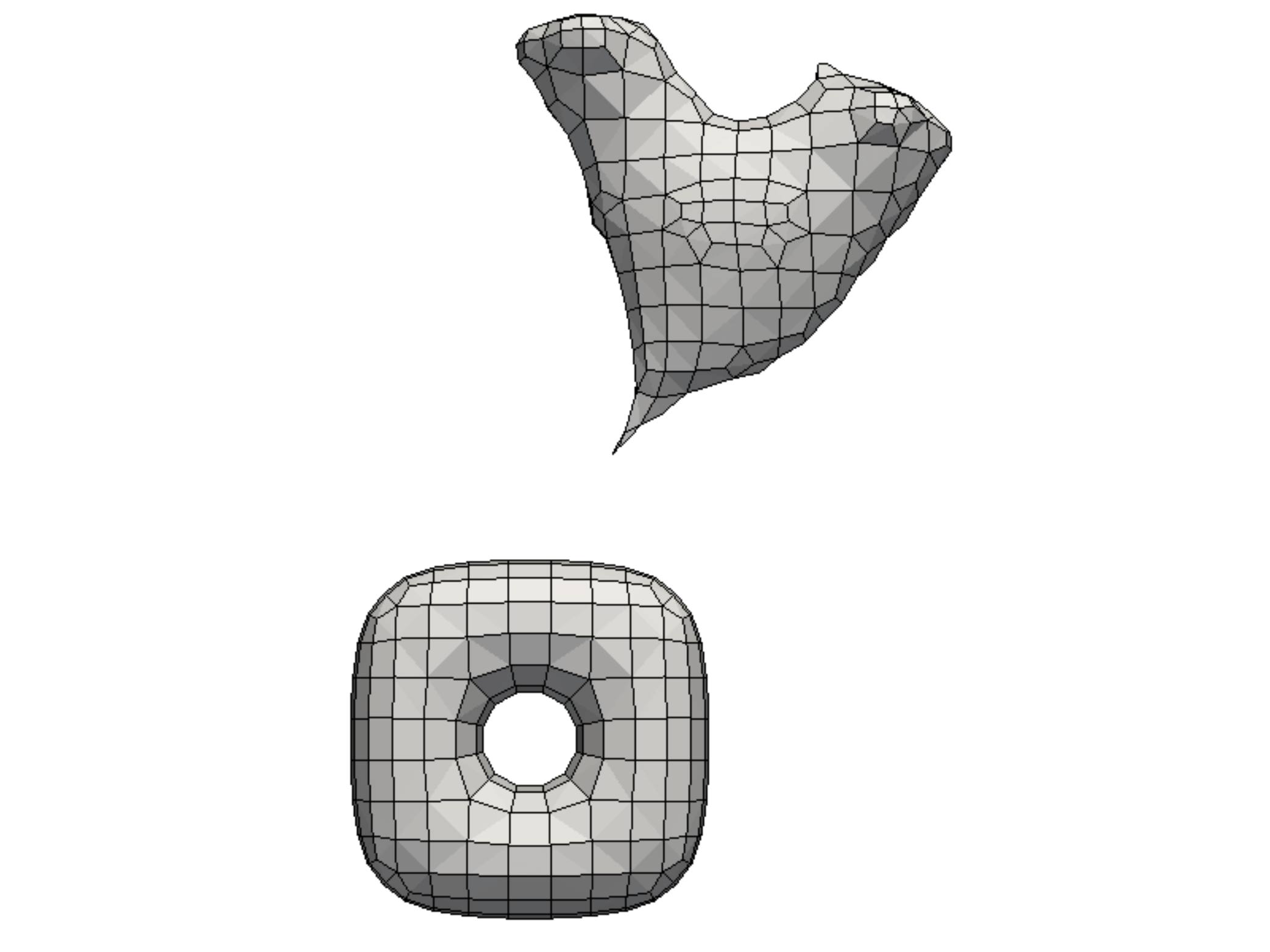}} & \parbox[m]{6em}{\includegraphics[trim={0cm 0cm 0cm 0cm},clip, width=0.12\textwidth]{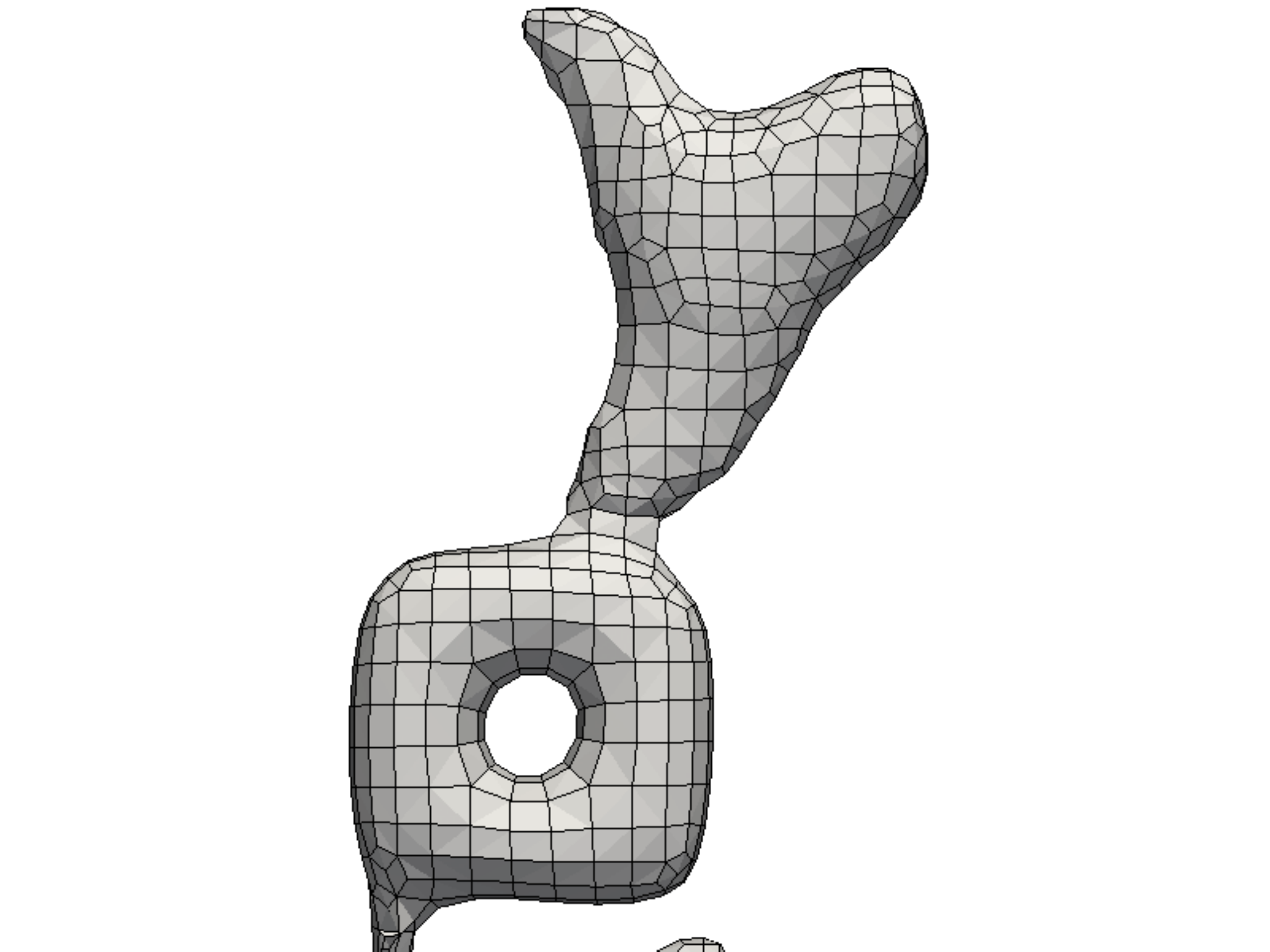}} & \parbox[m]{6em}{\includegraphics[trim={0cm 0cm 0cm 0cm},clip, width=0.12\textwidth]{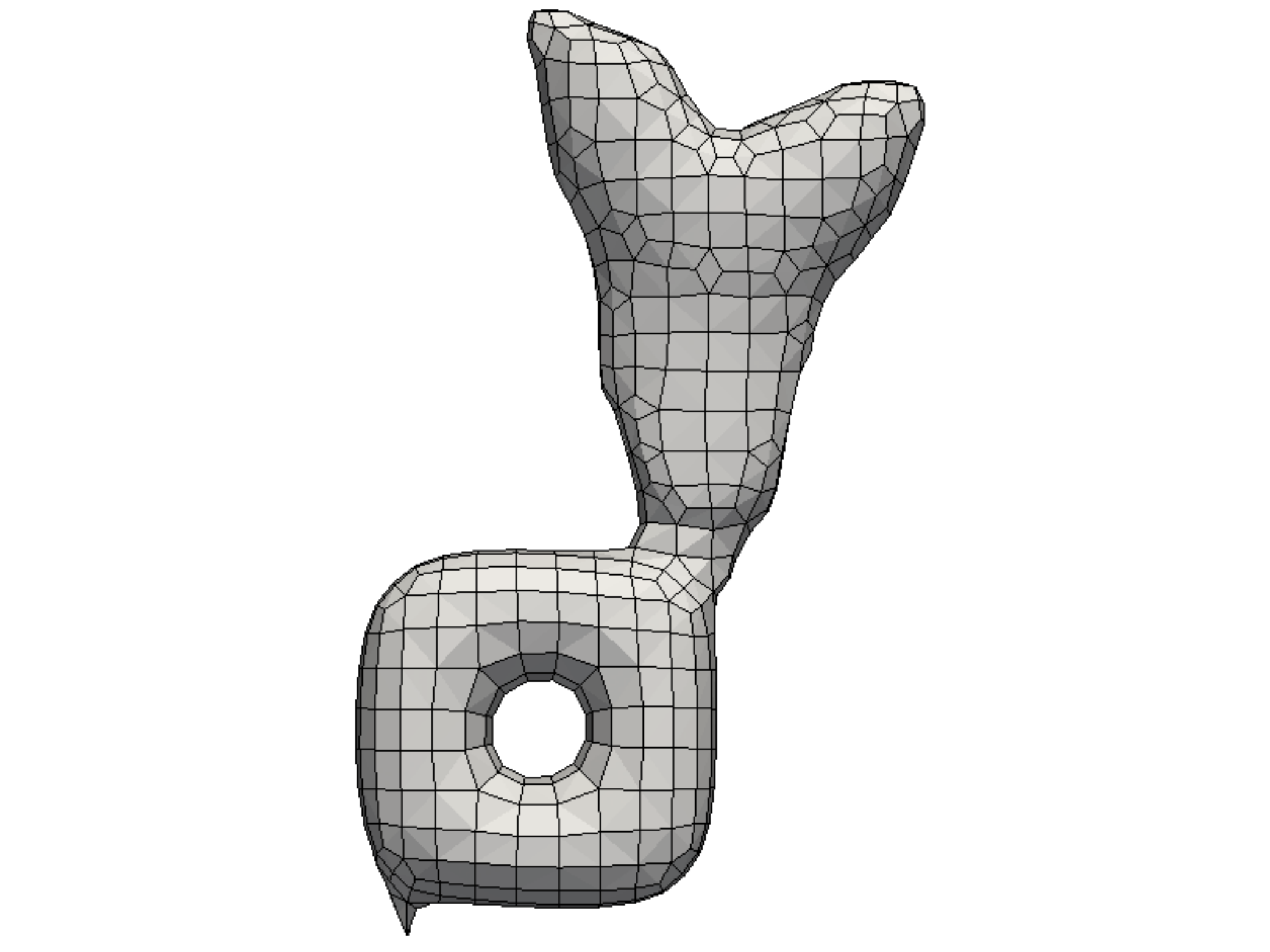}} & \parbox[m]{6em}{\includegraphics[trim={0cm 0cm 0cm 0cm},clip, width=0.12\textwidth]{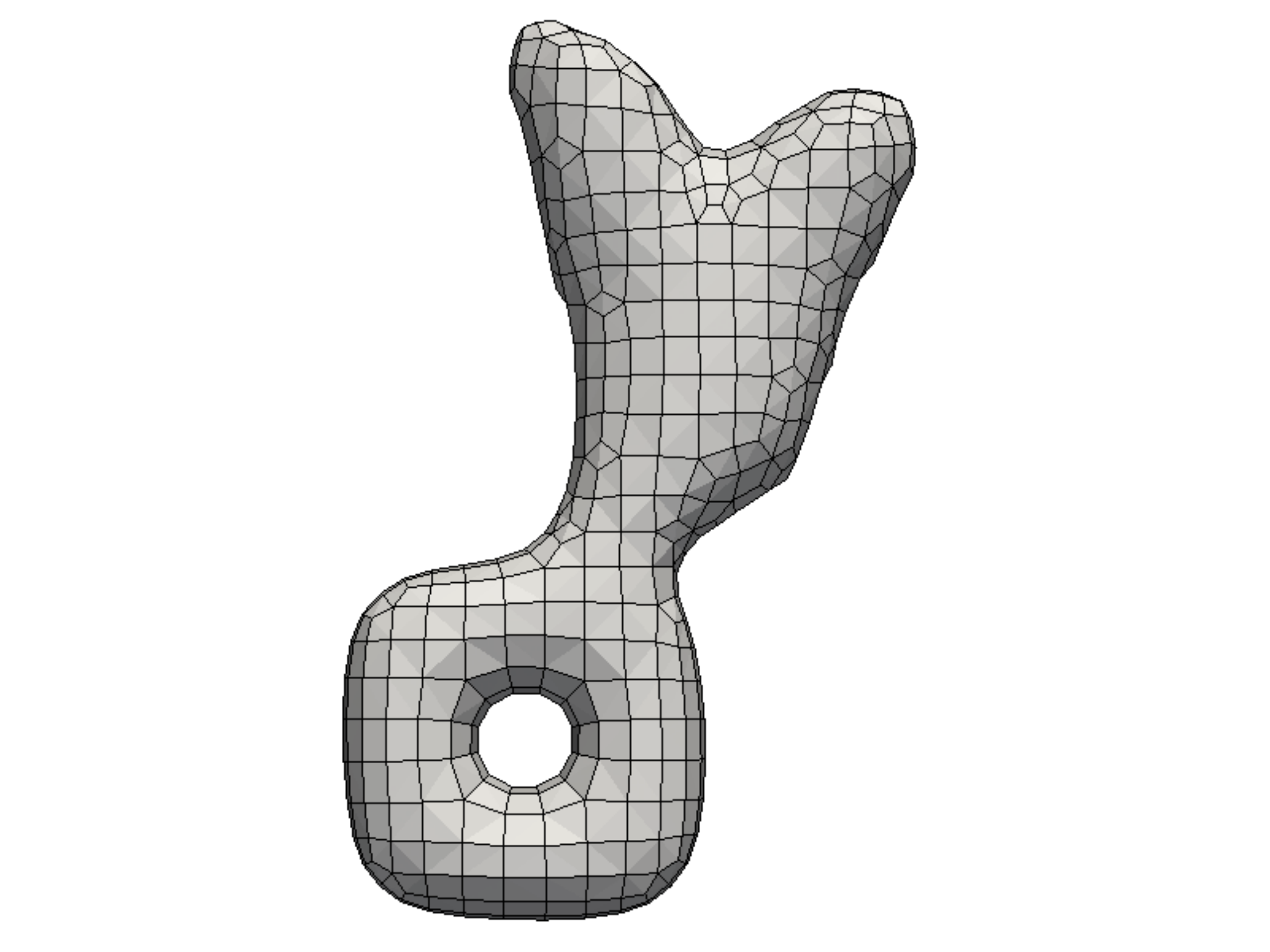}} \\\cline{2-7}
     & \checkmark & \parbox[m]{6em}{\includegraphics[trim={0cm 0cm 0cm 0cm},clip, width=0.12\textwidth]{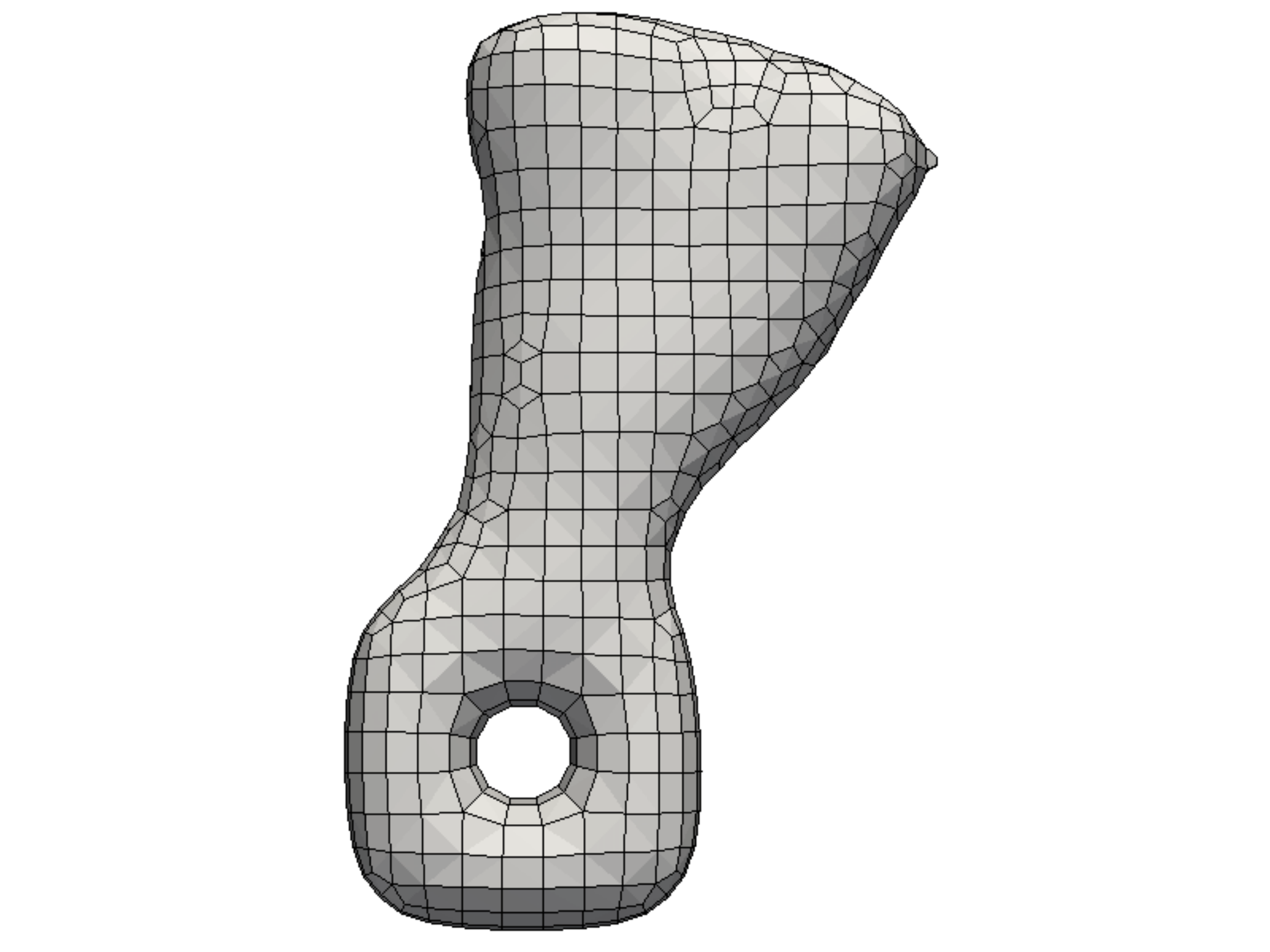}} & \parbox[m]{6em}{\includegraphics[trim={0cm 0cm 0cm 0cm},clip, width=0.12\textwidth]{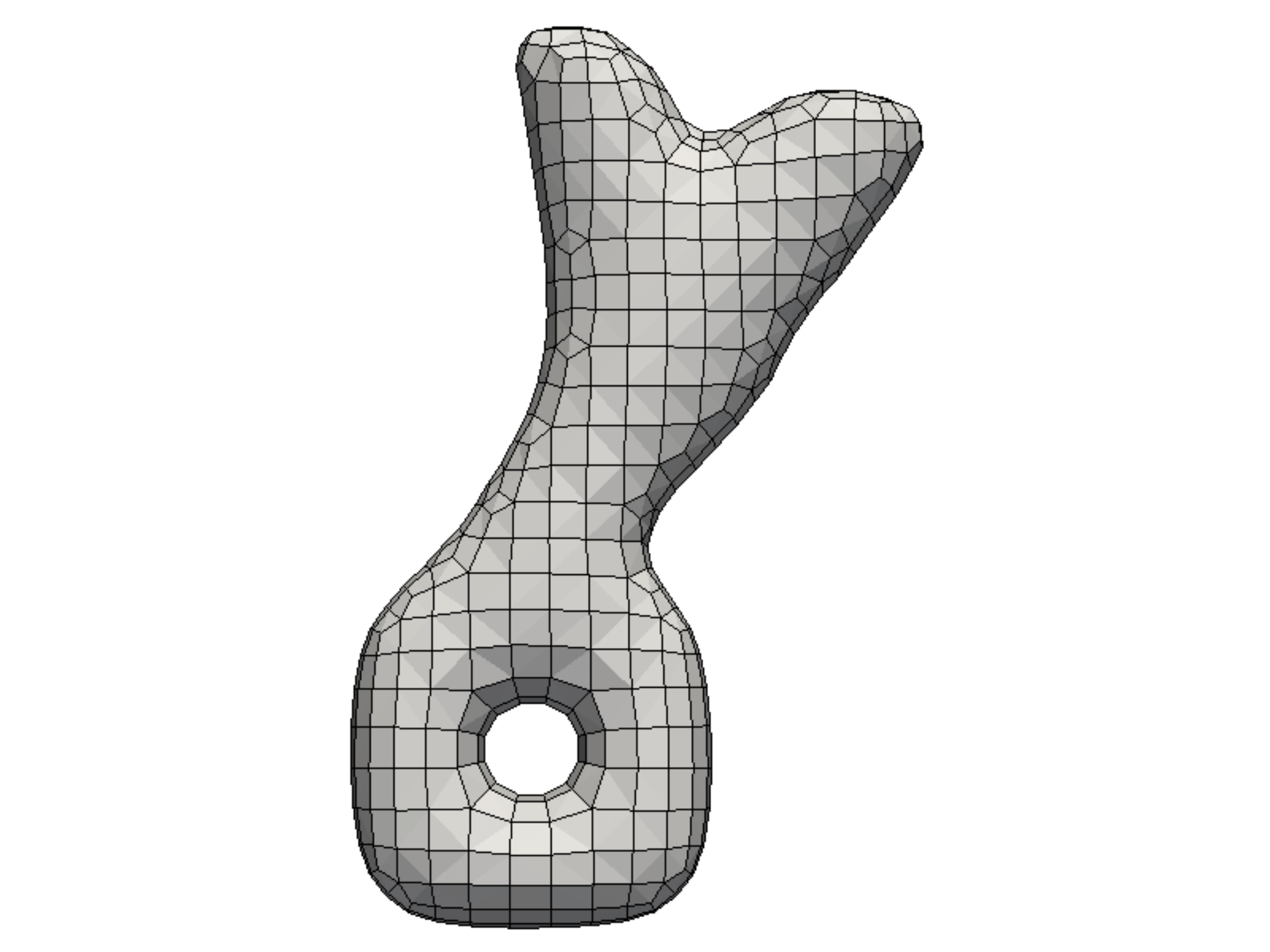}} & \parbox[m]{6em}{\includegraphics[trim={0cm 0cm 0cm 0cm},clip, width=0.12\textwidth]{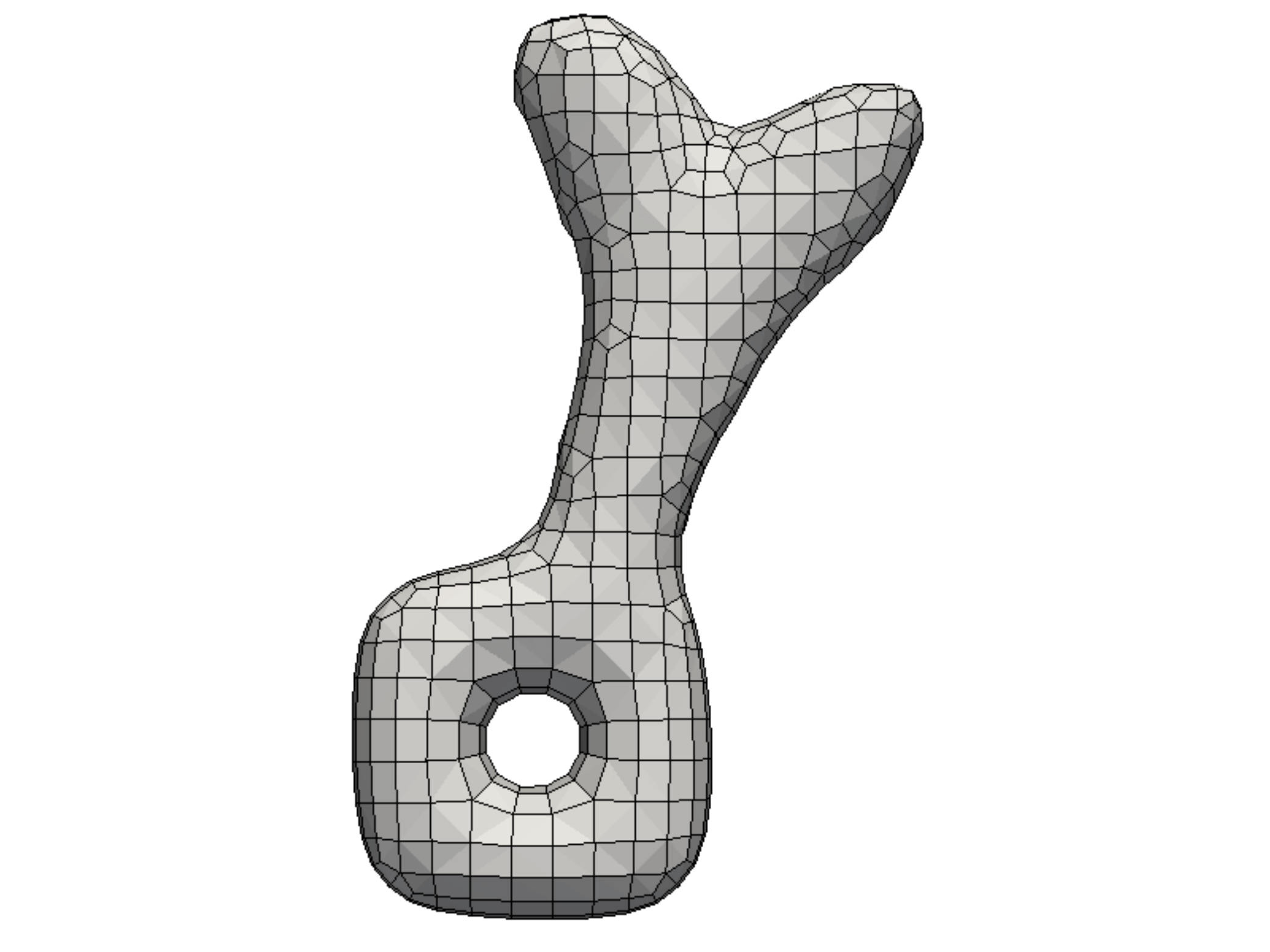}} & \parbox[m]{6em}{\includegraphics[trim={0cm 0cm 0cm 0cm},clip, width=0.12\textwidth]{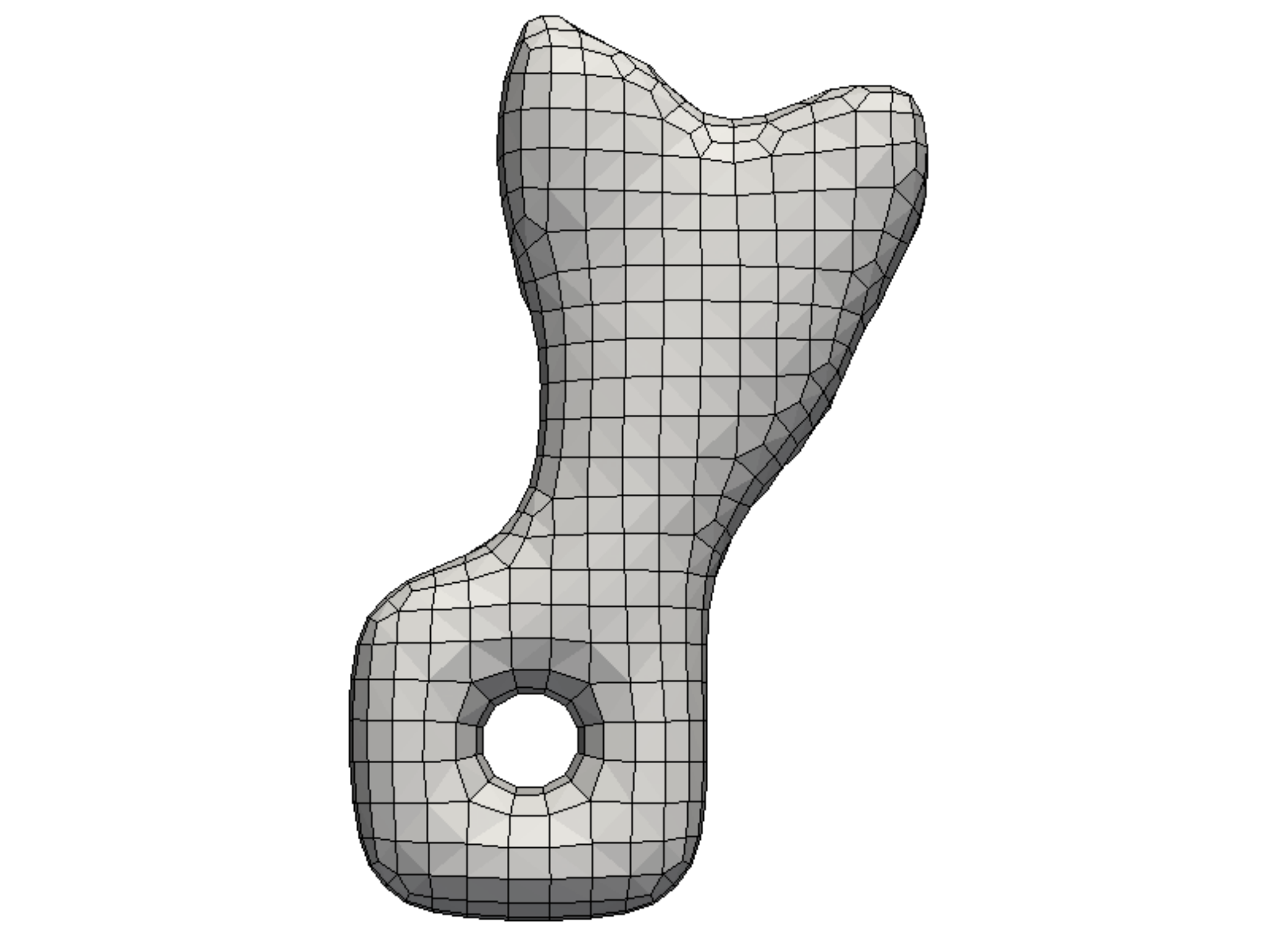}} & \parbox[m]{6em}{\includegraphics[trim={0cm 0cm 0cm 0cm},clip, width=0.12\textwidth]{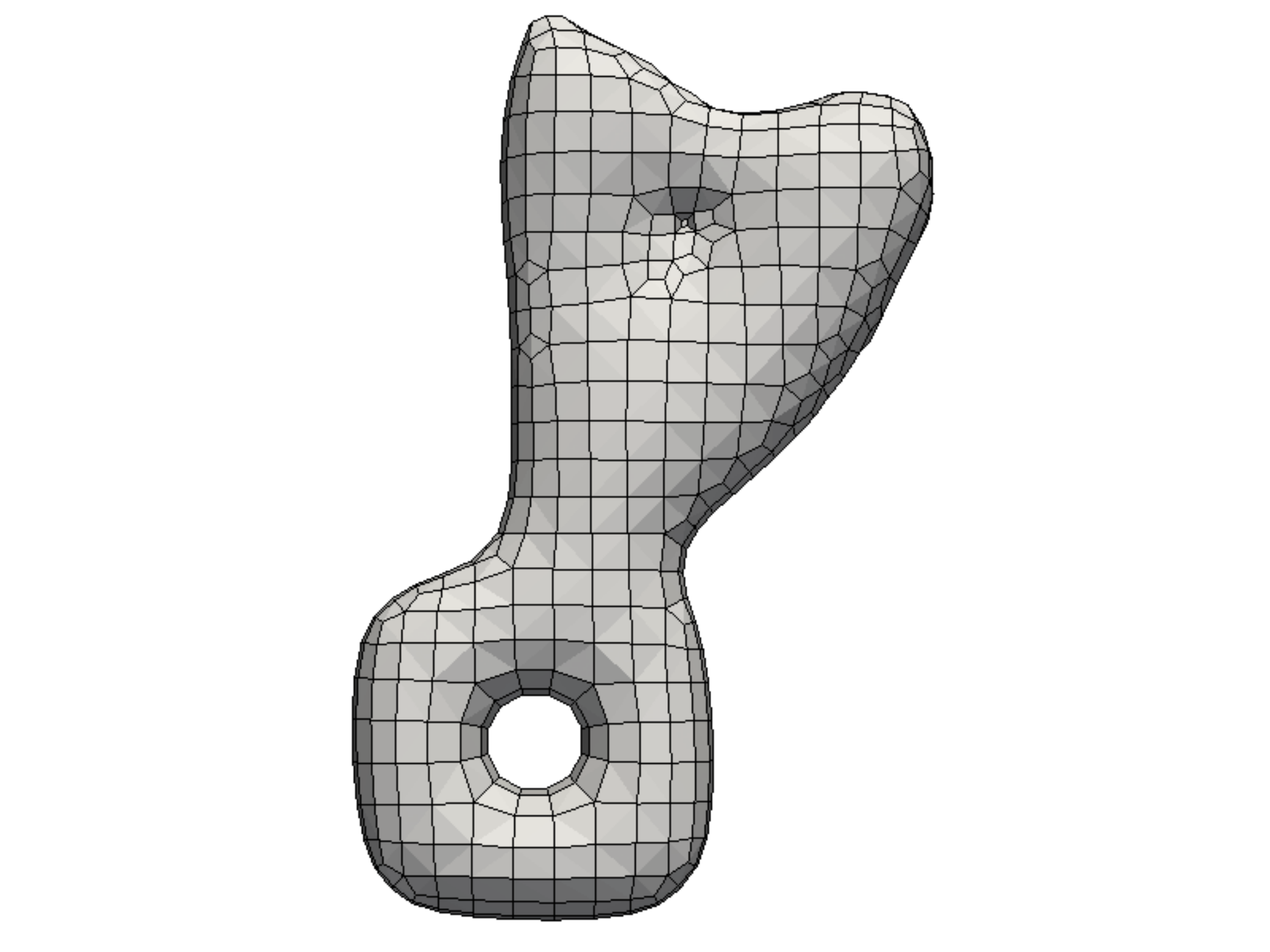}}\\
     \multicolumn{7}{c}{\fbox{\hspace{0.3cm}\parbox[m]{6em}{\includegraphics[trim={0cm 0cm 0cm 0cm},clip, width=0.12\textwidth]{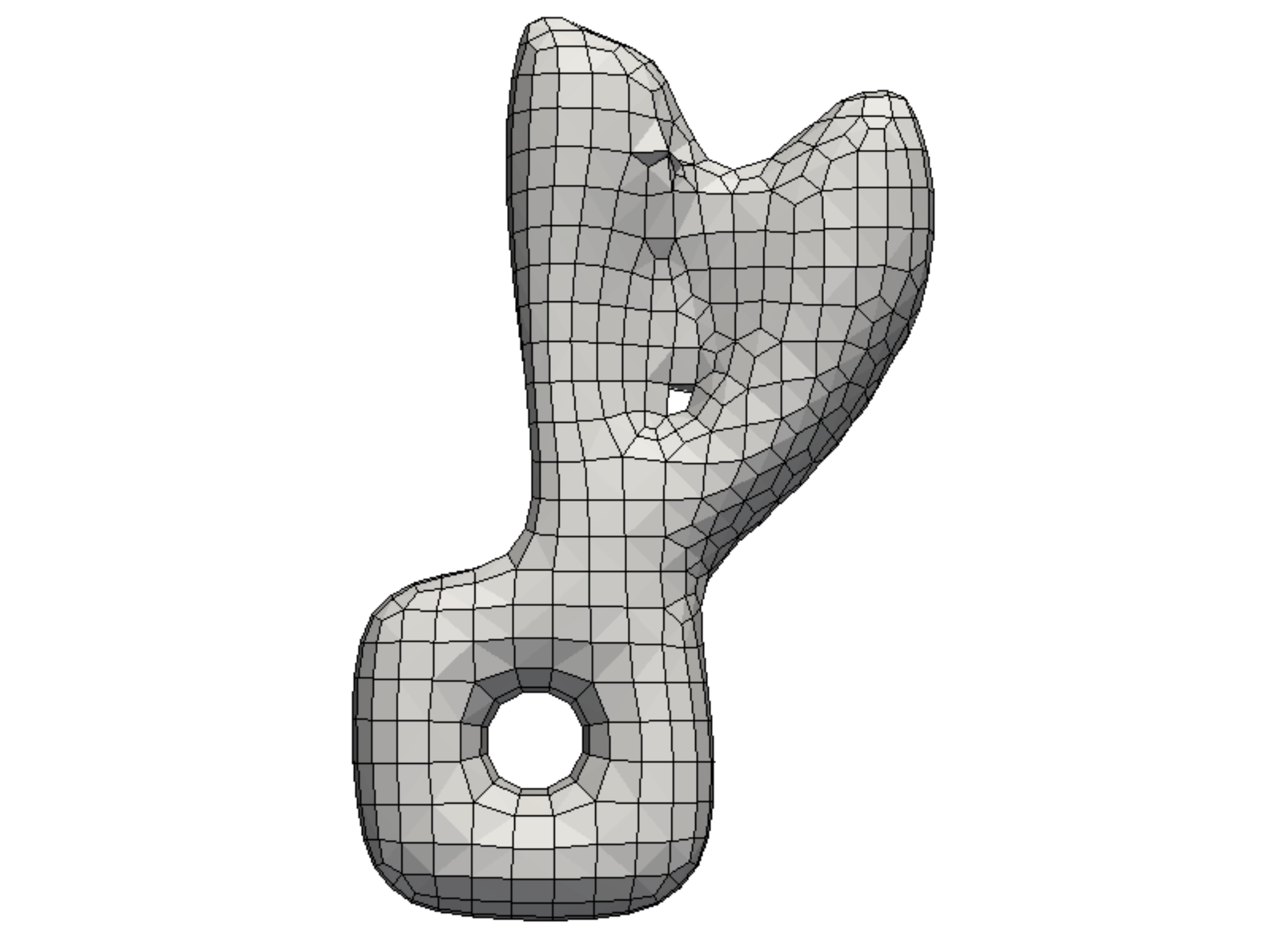}}}}
\end{tabular}
\end{subtable}
\begin{subtable}[h]{0.99\textwidth}
    \vspace{5mm}%
    \centering\setcellgapes{3pt}\makegapedcells
    \setlength\tabcolsep{3.5pt}
    \begin{tabular}{c|c||ScScScScSc}
    \multicolumn{2}{c||}{} & \multicolumn{5}{c}{training samples} \\\hline
     prepr. & equiv. & 10 & 50 & 100 & 500 & 1500 \\\hline
     \multirow{2}{*}{\rotatebox{90}{trivial}} & & \parbox[m]{6em}{\includegraphics[trim={0cm 0cm 0cm 0cm},clip, width=0.12\textwidth]{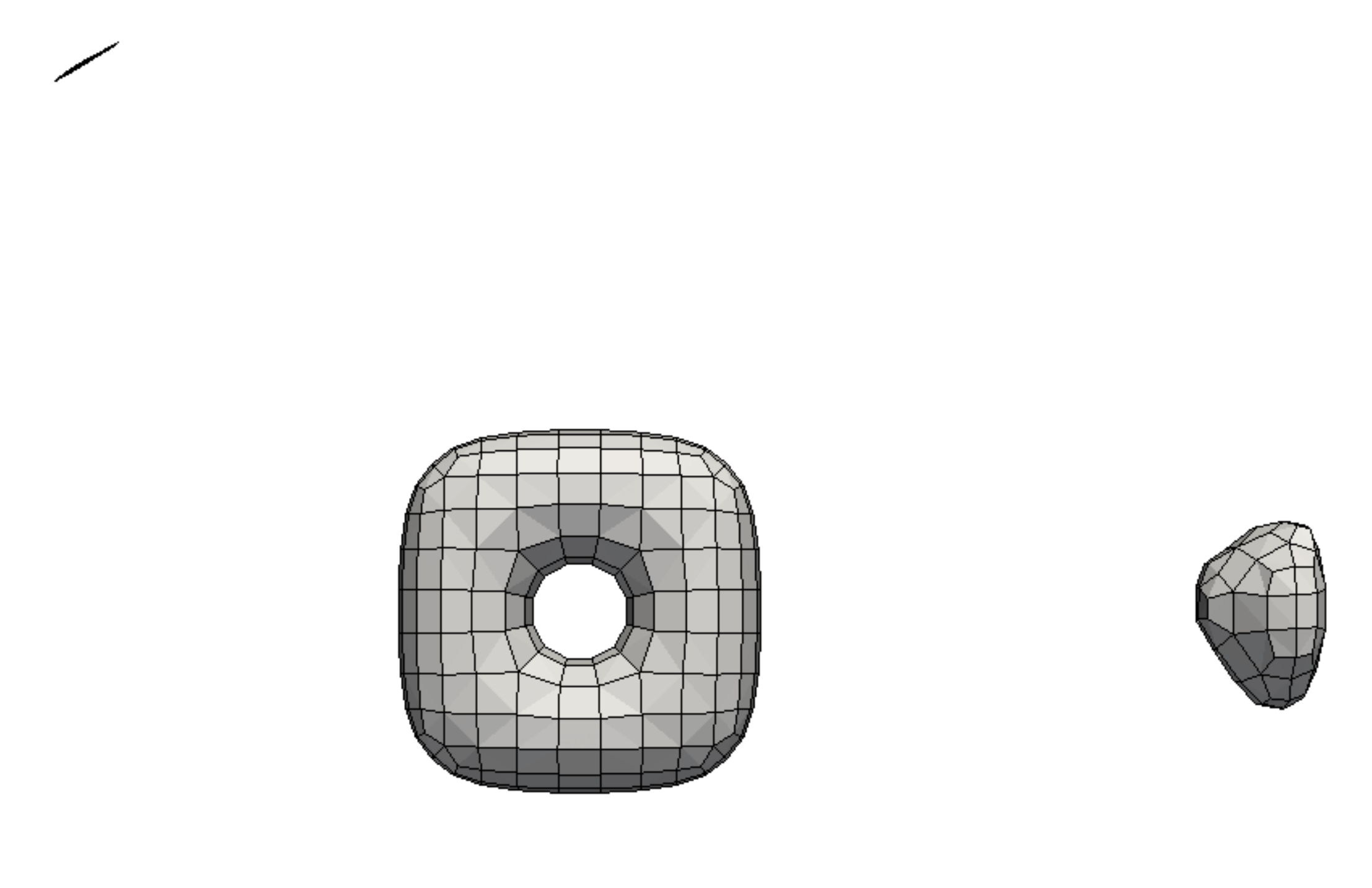}} & \parbox[m]{6em}{\includegraphics[trim={0cm 0cm 0cm 0cm},clip, width=0.12\textwidth]{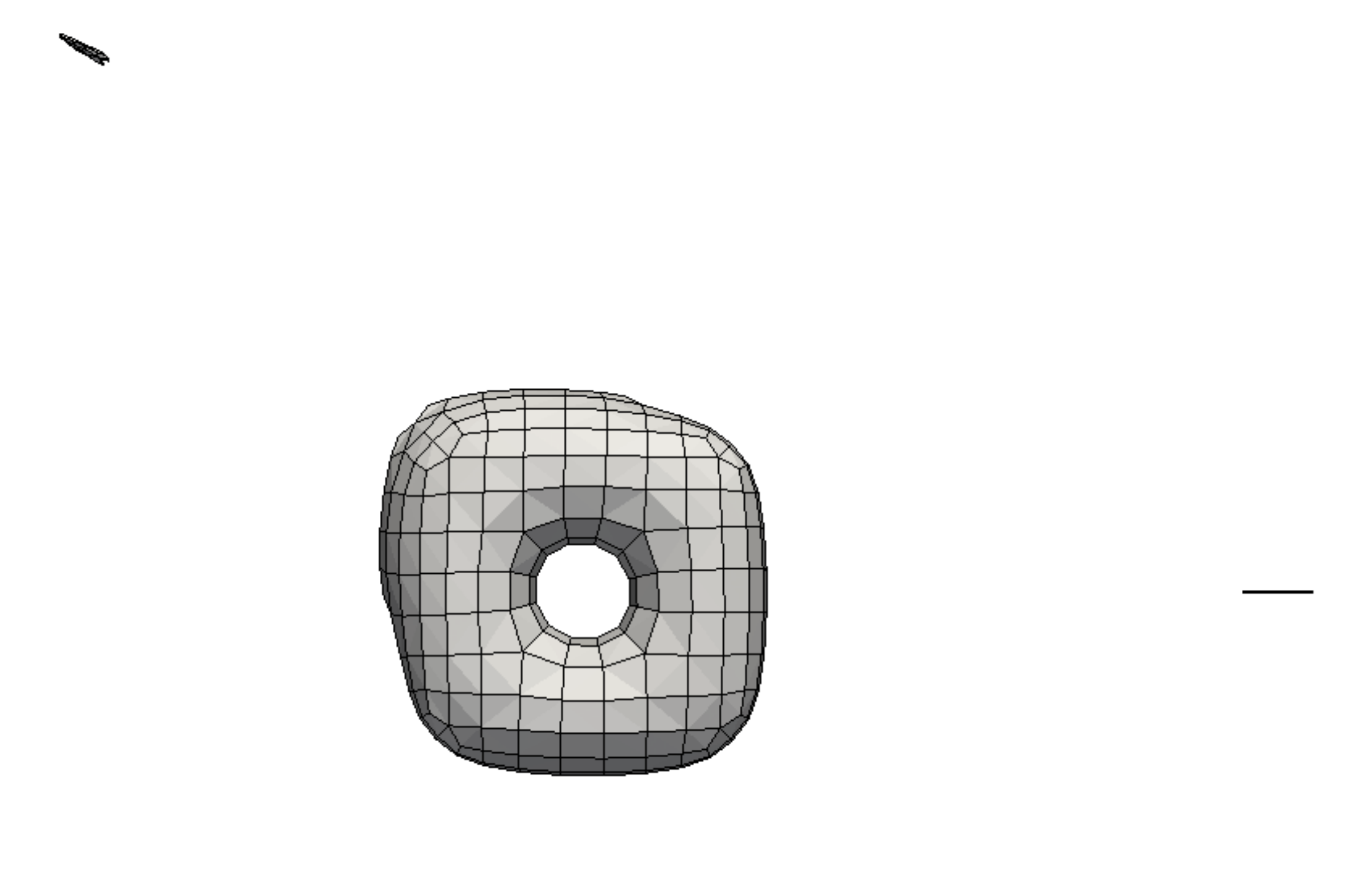}} & \parbox[m]{6em}{\includegraphics[trim={0cm 0cm 0cm 0cm},clip, width=0.12\textwidth]{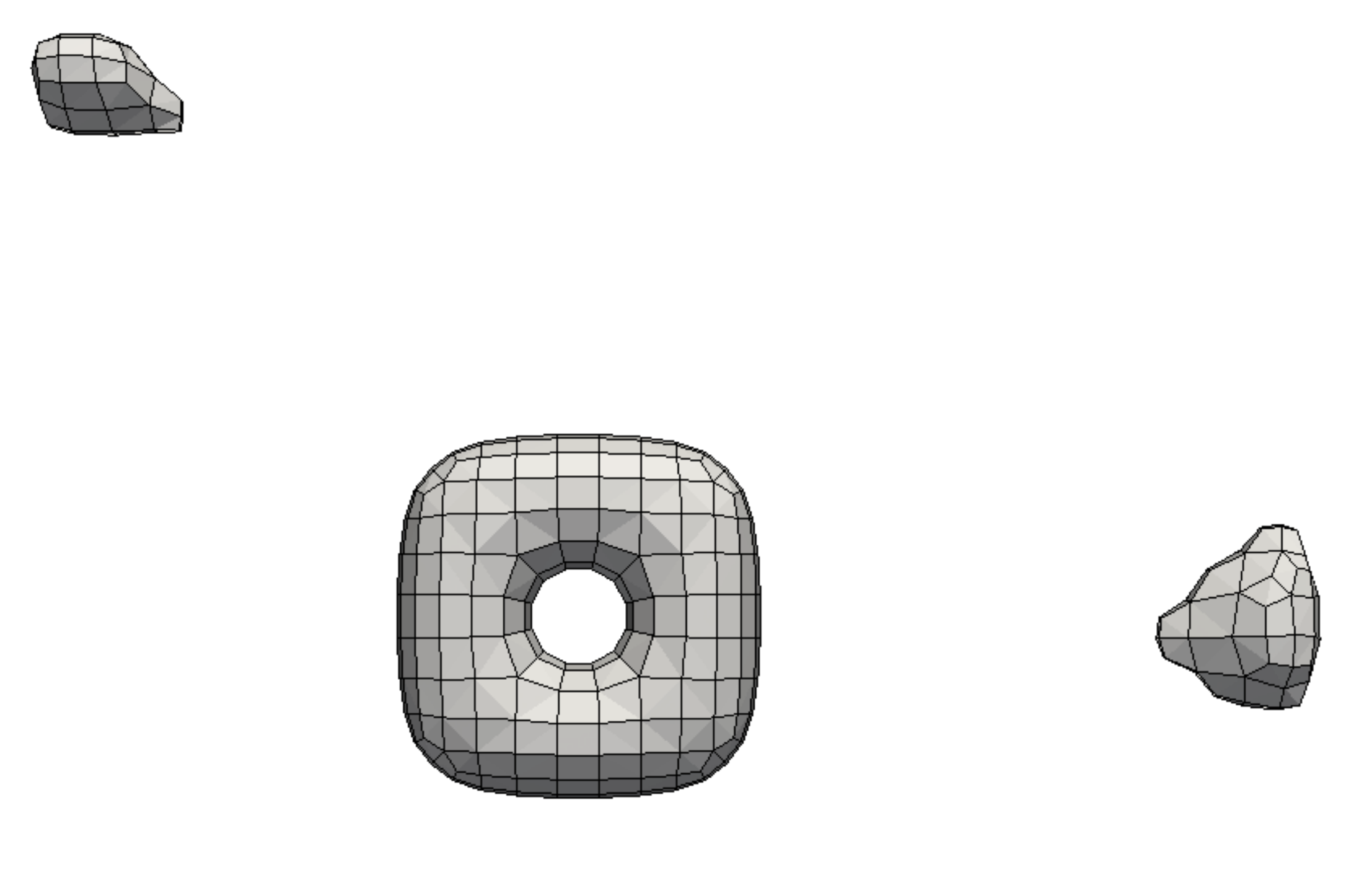}} & \parbox[m]{6em}{\includegraphics[trim={0cm 0cm 0cm 0cm},clip, width=0.12\textwidth]{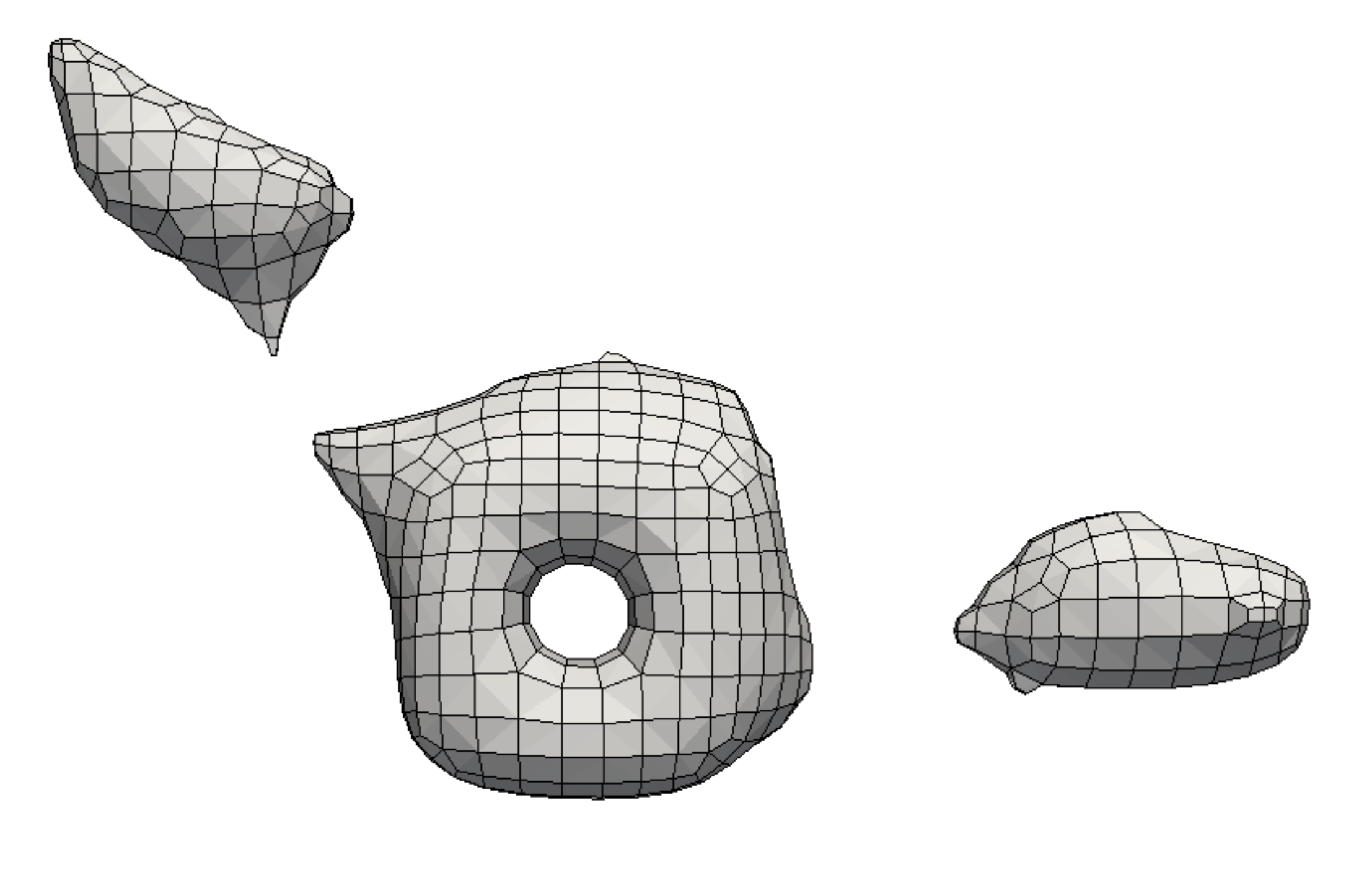}} & \parbox[m]{6em}{\includegraphics[trim={0cm 0cm 0cm 0cm},clip, width=0.12\textwidth]{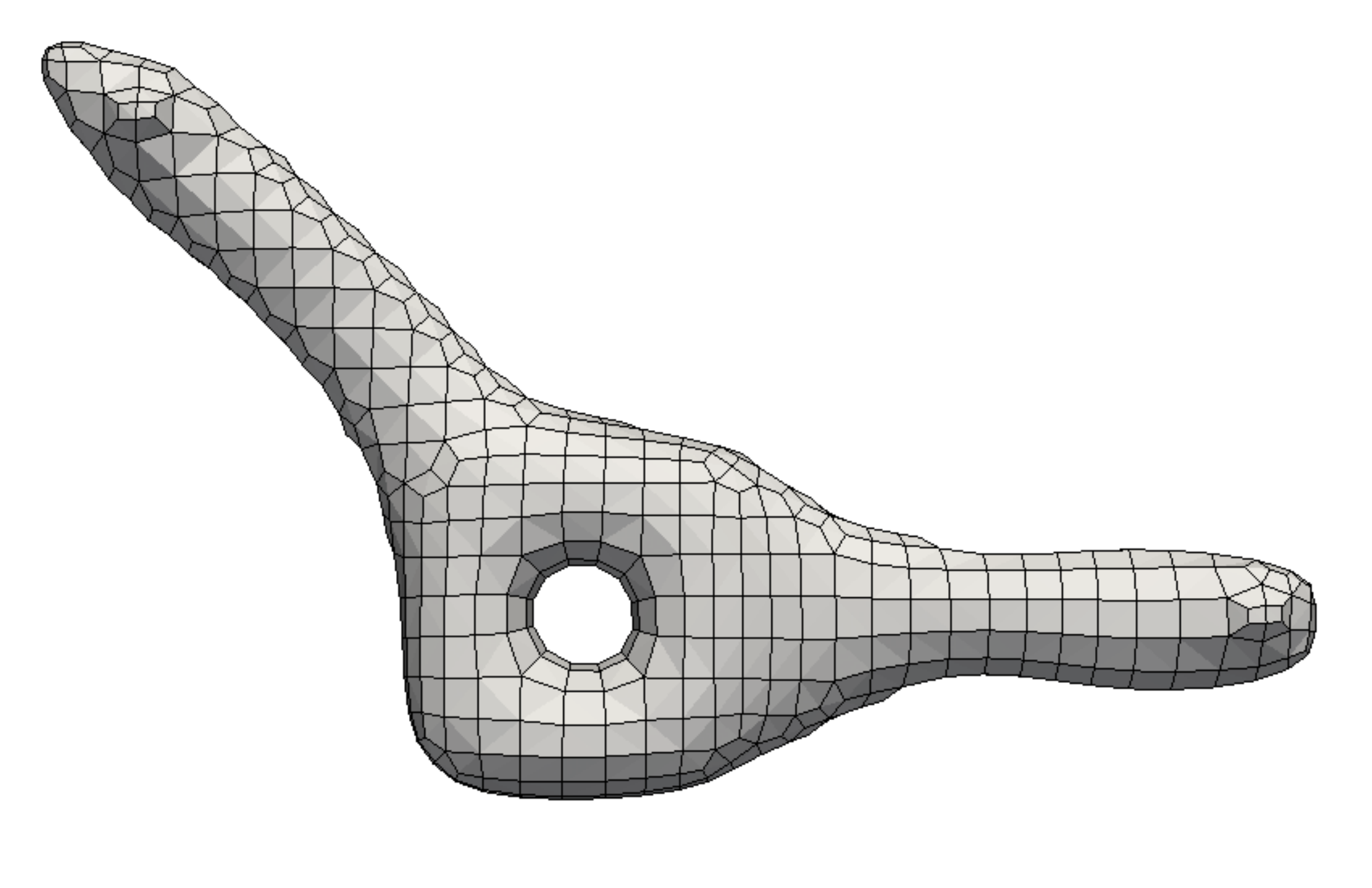}} \\\cline{2-7}
     & \checkmark
     & \parbox[m]{6em}{\includegraphics[trim={0cm 0cm 0cm 0cm},clip, width=0.12\textwidth]{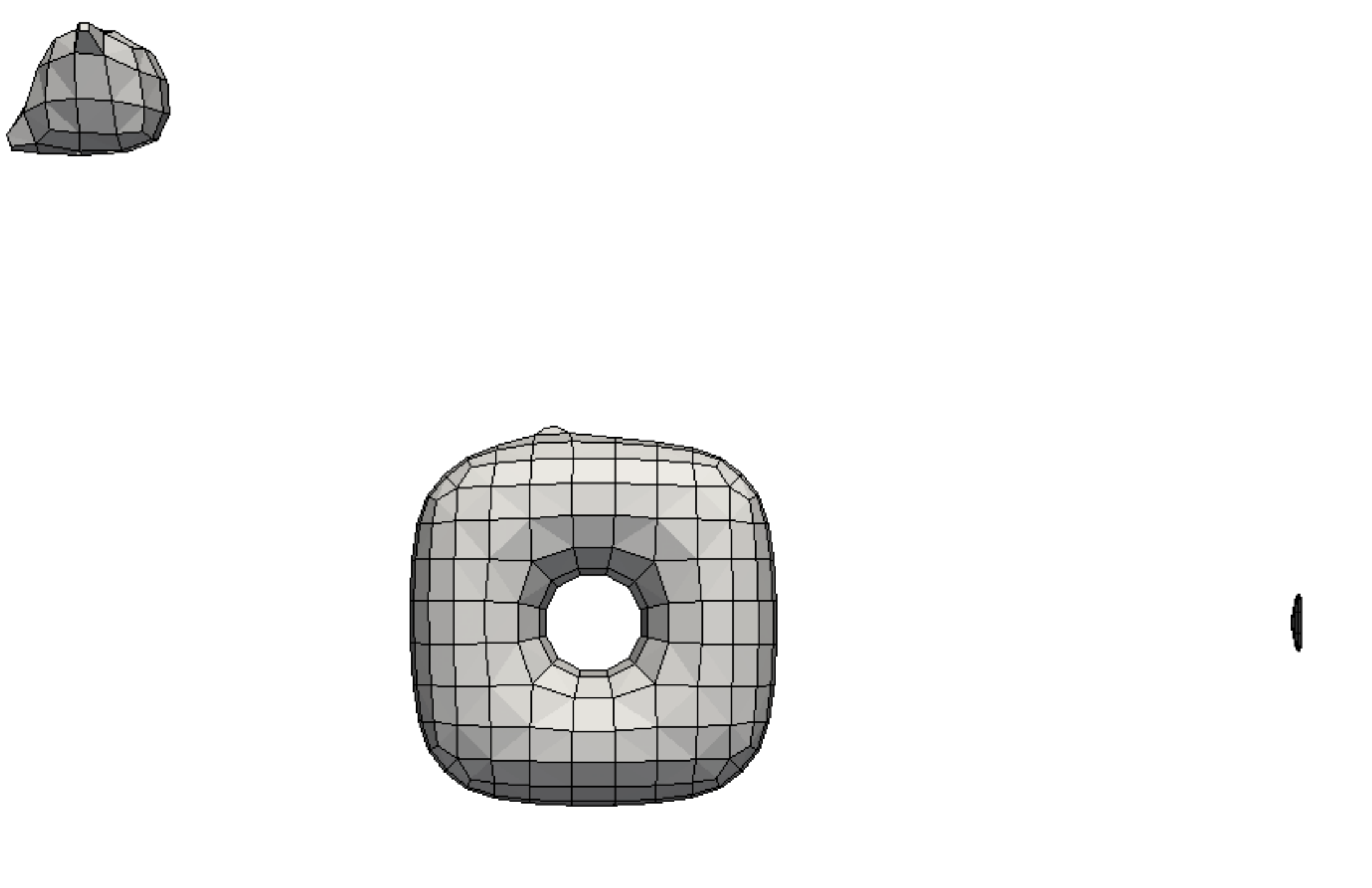}} & \parbox[m]{6em}{\includegraphics[trim={0cm 0cm 0cm 0cm},clip, width=0.12\textwidth]{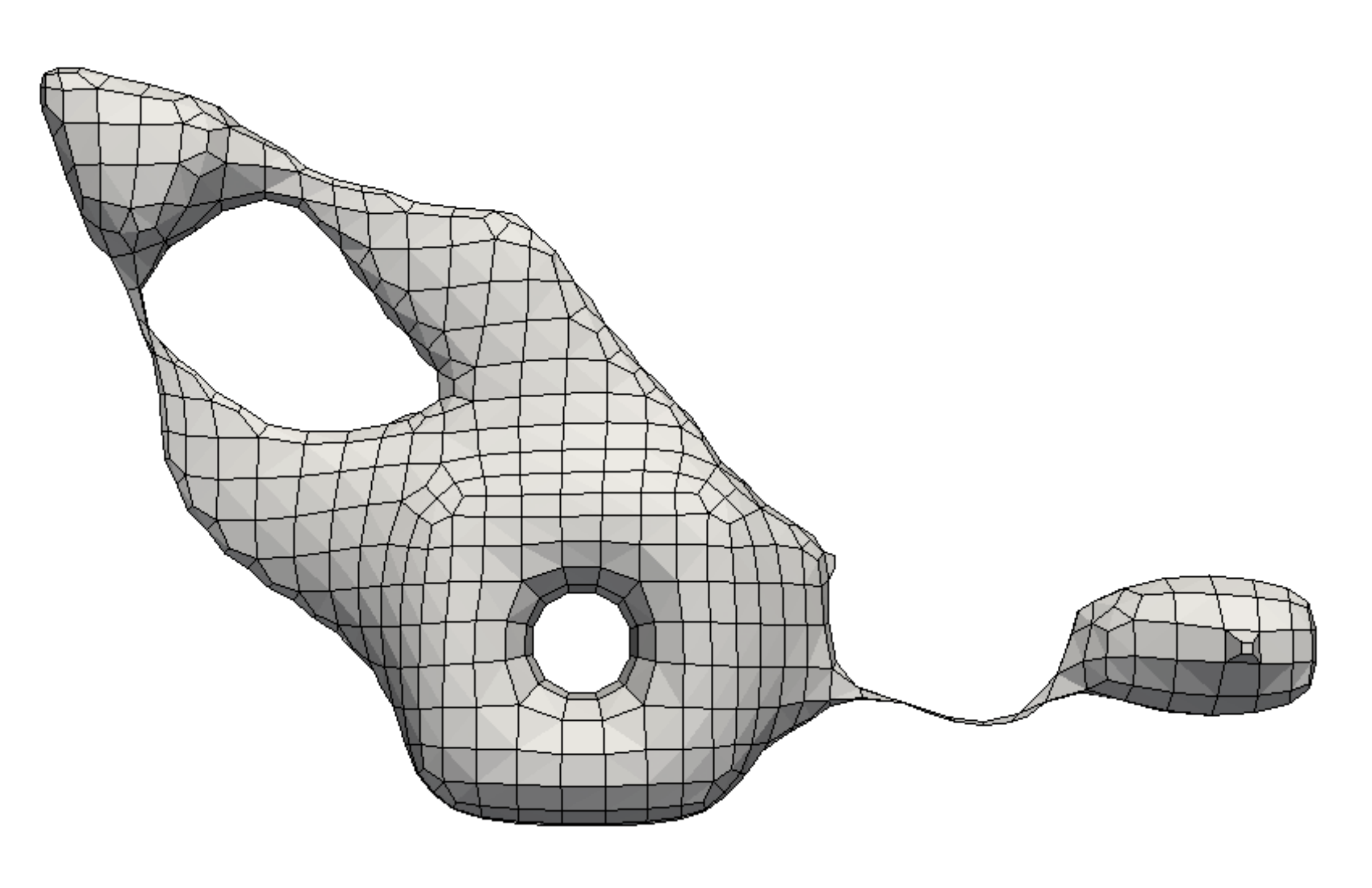}} & \parbox[m]{6em}{\includegraphics[trim={0cm 0cm 0cm 0cm},clip, width=0.12\textwidth]{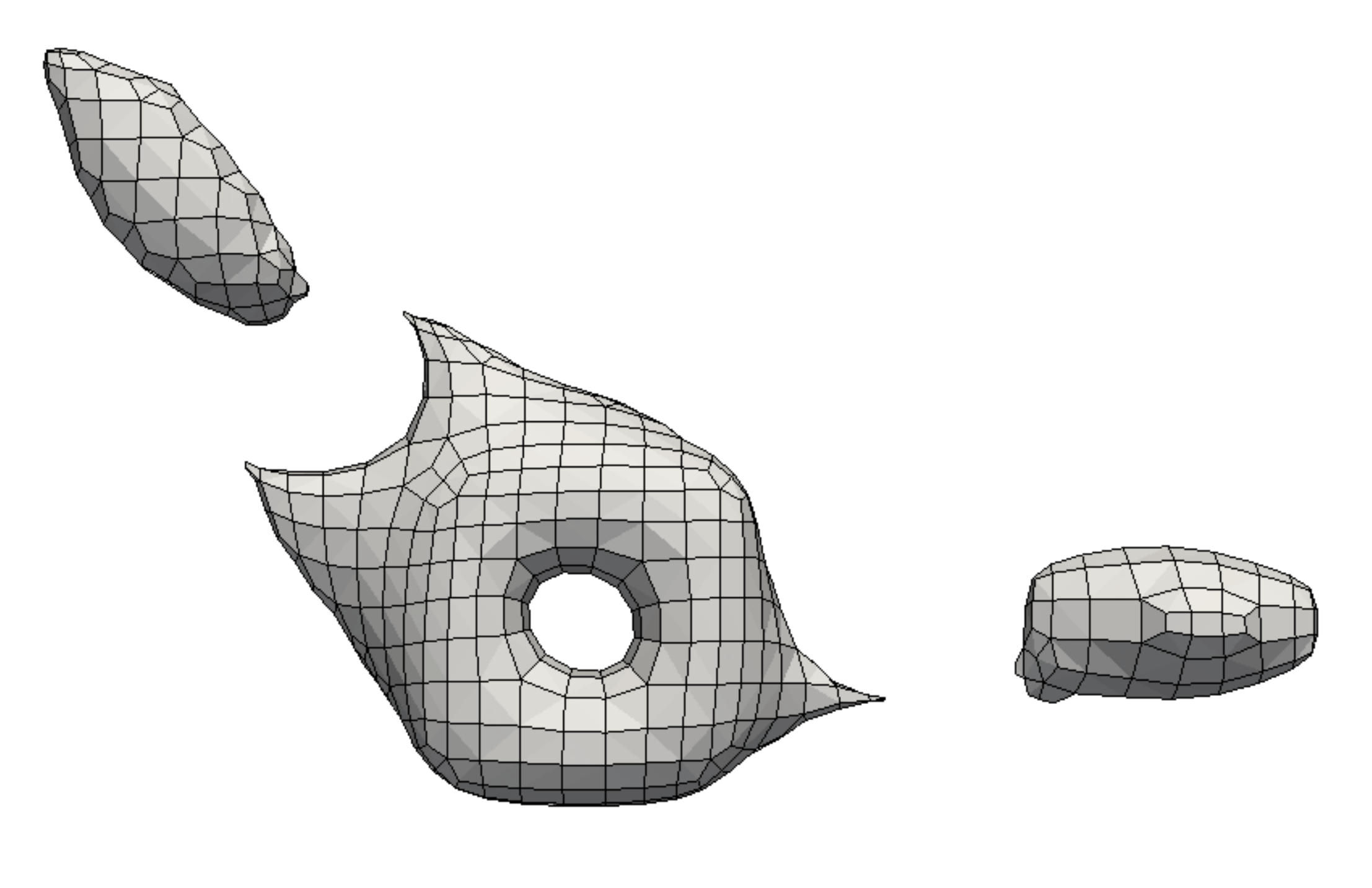}} & \parbox[m]{6em}{\includegraphics[trim={0cm 0cm 0cm 0cm},clip, width=0.12\textwidth]{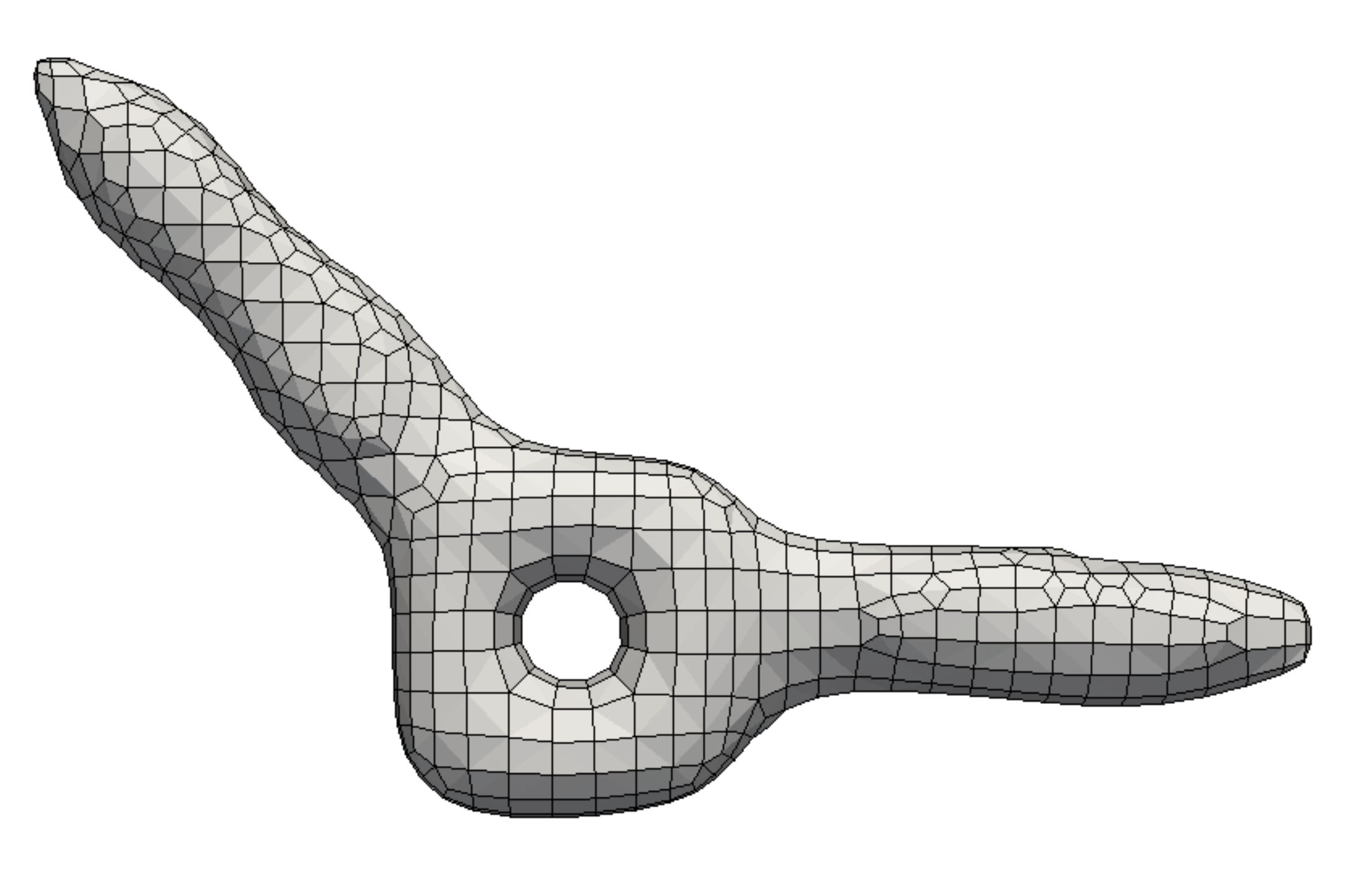}} & \parbox[m]{6em}{\includegraphics[trim={0cm 0cm 0cm 0cm},clip, width=0.12\textwidth]{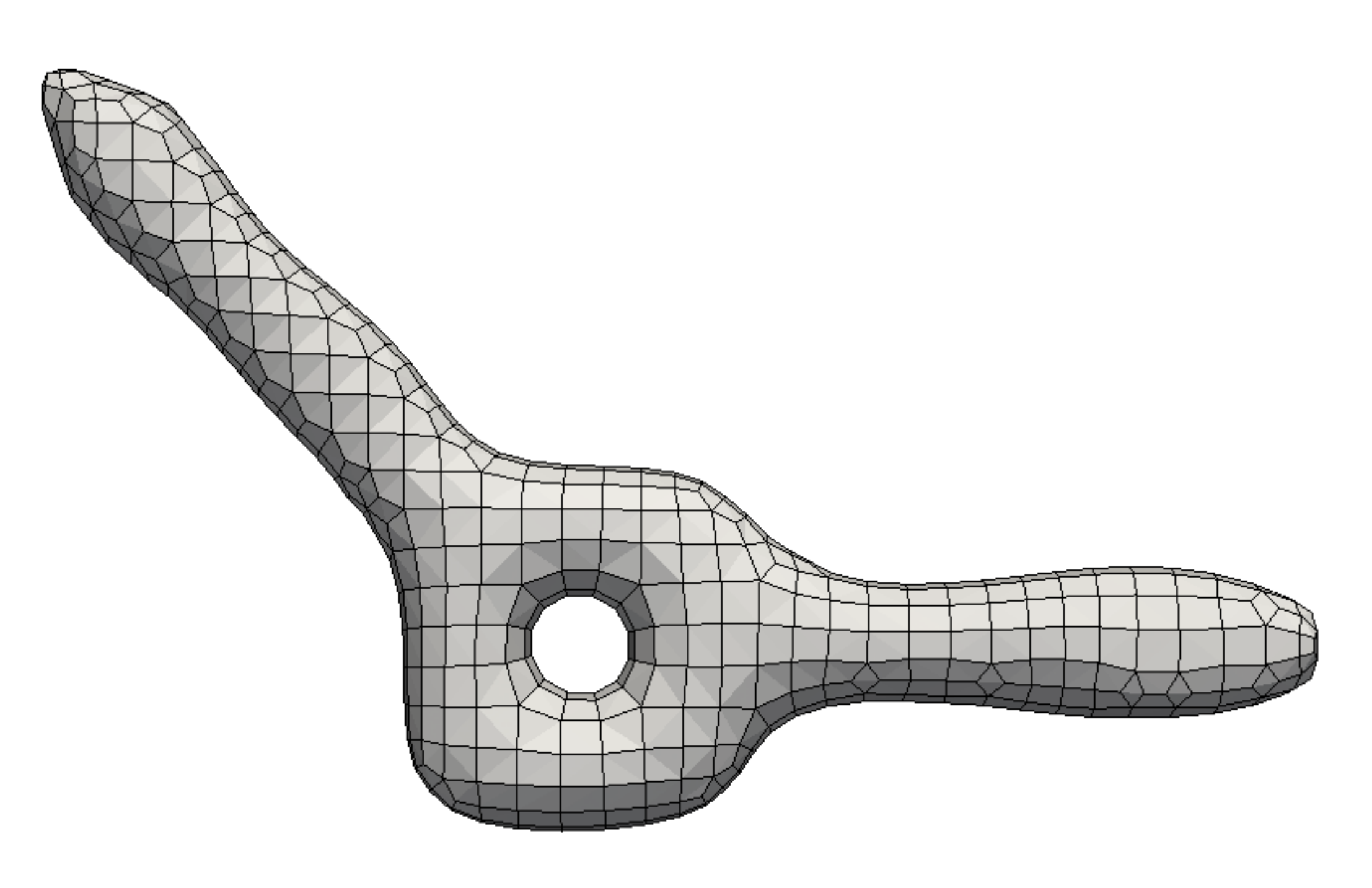}} \\\hline
     \multirow{2}{*}{\rotatebox{90}{trivial+PDE}} & & \parbox[m]{6em}{\includegraphics[trim={0cm 0cm 0cm 0cm},clip, width=0.12\textwidth]{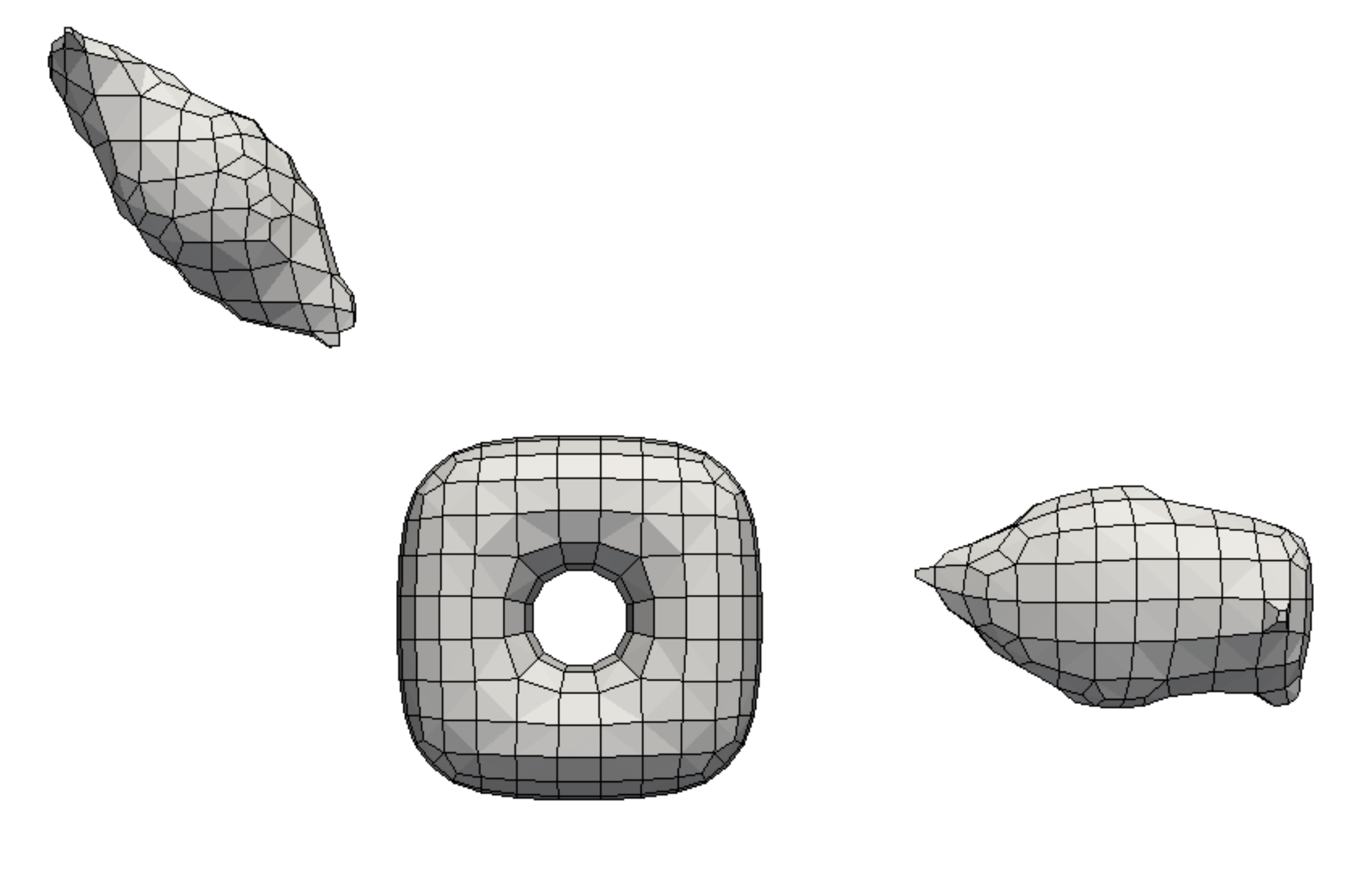}} & \parbox[m]{6em}{\includegraphics[trim={0cm 0cm 0cm 0cm},clip, width=0.12\textwidth]{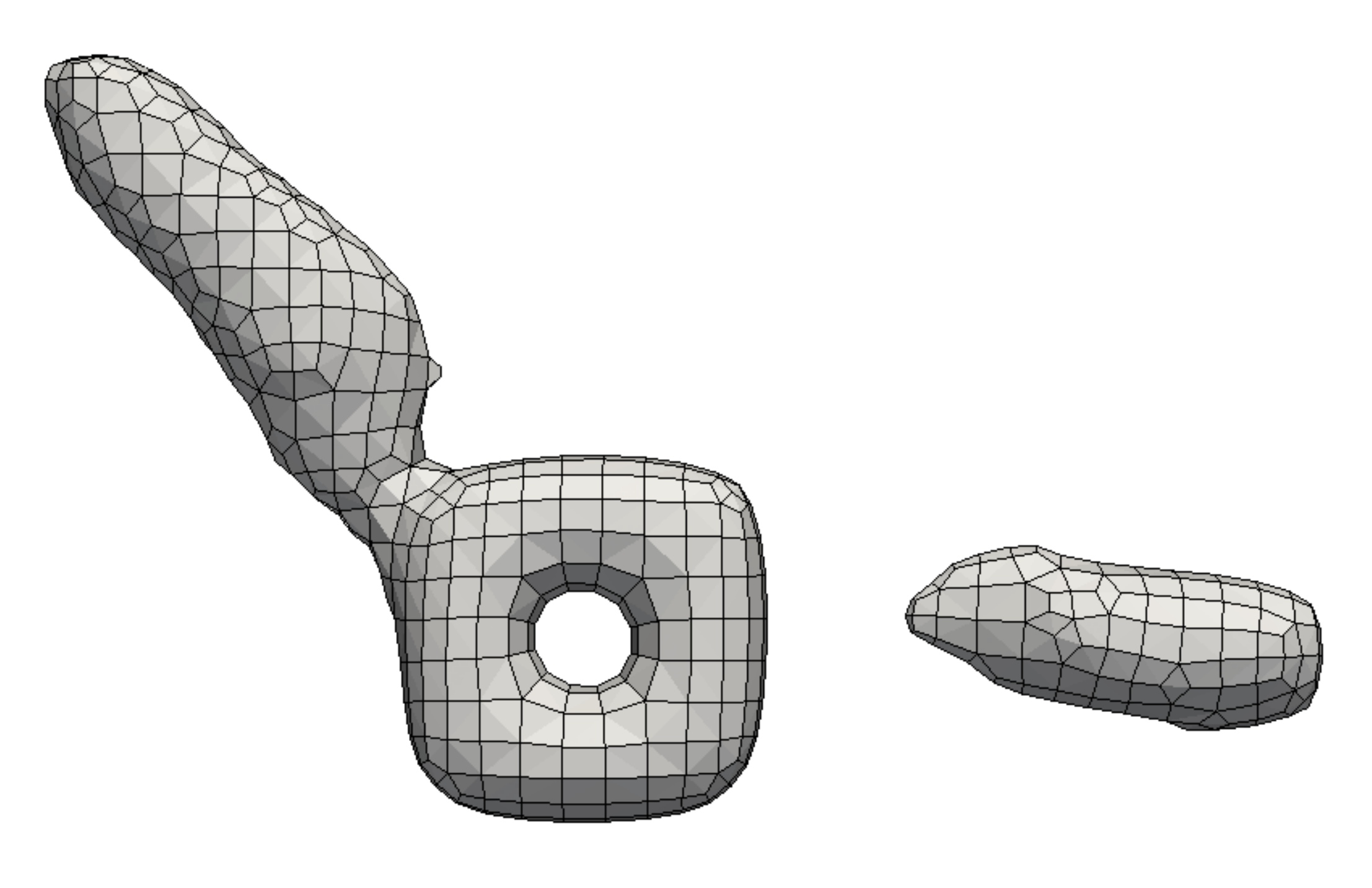}} & \parbox[m]{6em}{\includegraphics[trim={0cm 0cm 0cm 0cm},clip, width=0.12\textwidth]{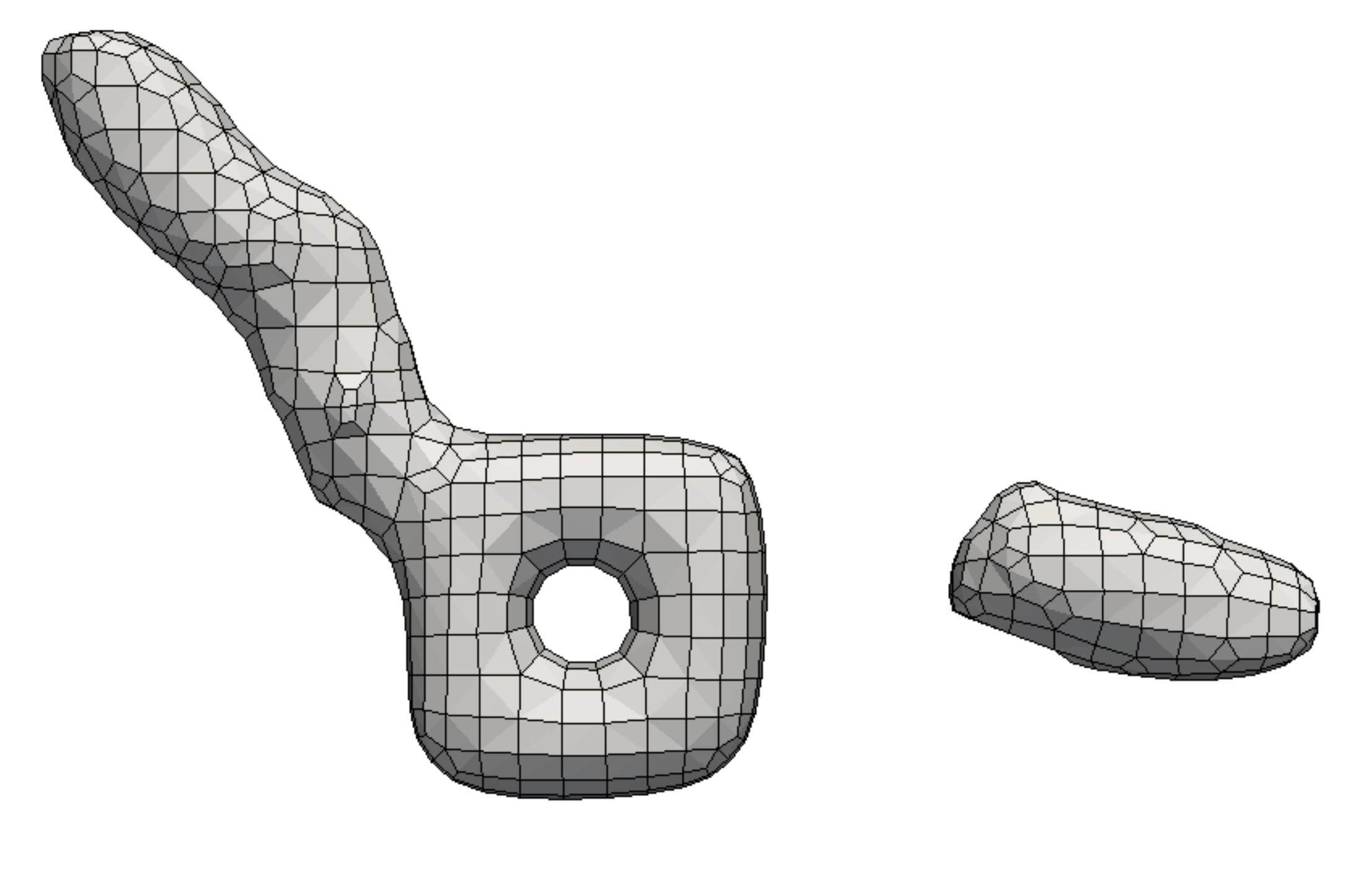}} & \parbox[m]{6em}{\includegraphics[trim={0cm 0cm 0cm 0cm},clip, width=0.12\textwidth]{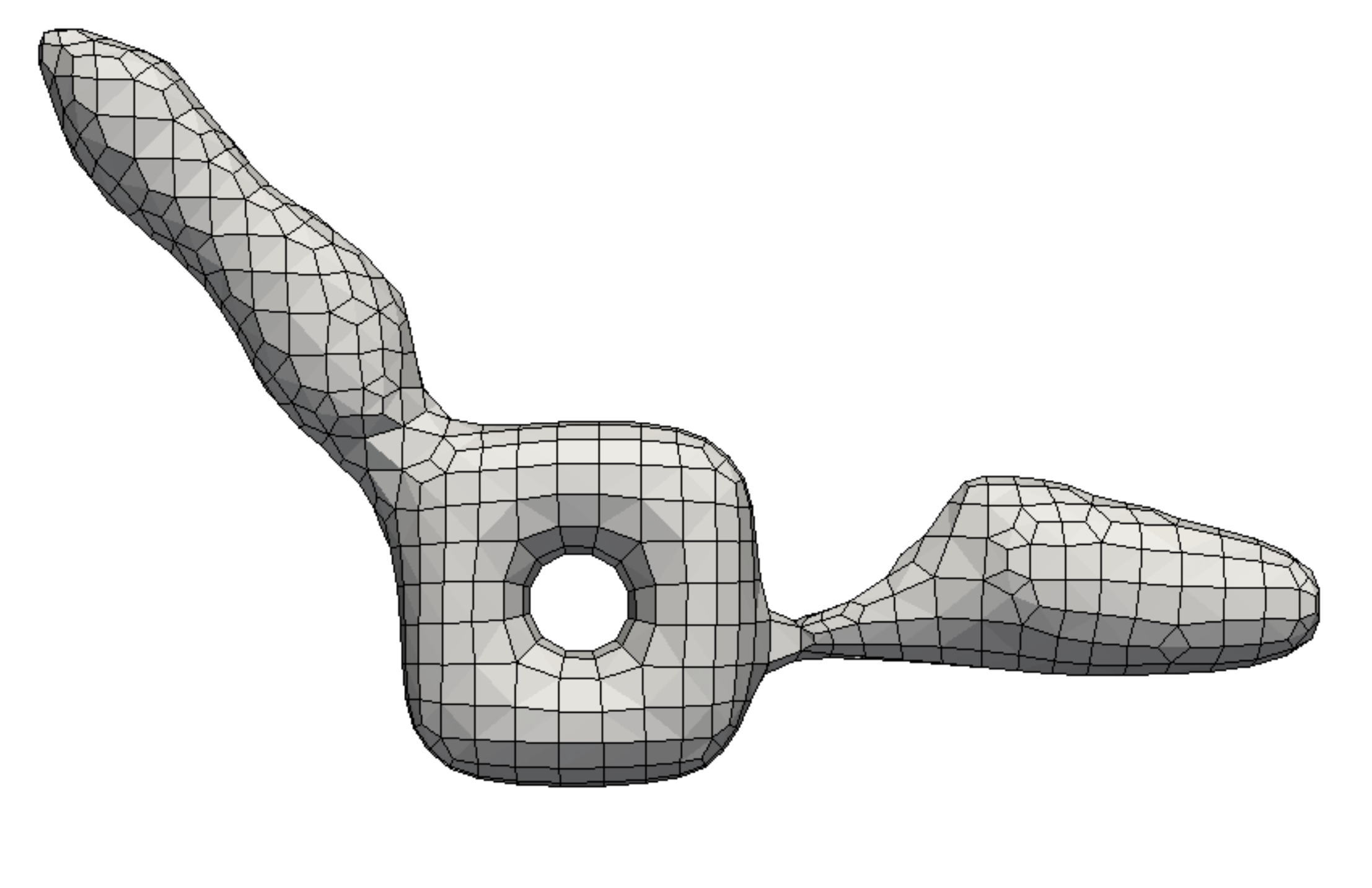}} & \parbox[m]{6em}{\includegraphics[trim={0cm 0cm 0cm 0cm},clip, width=0.12\textwidth]{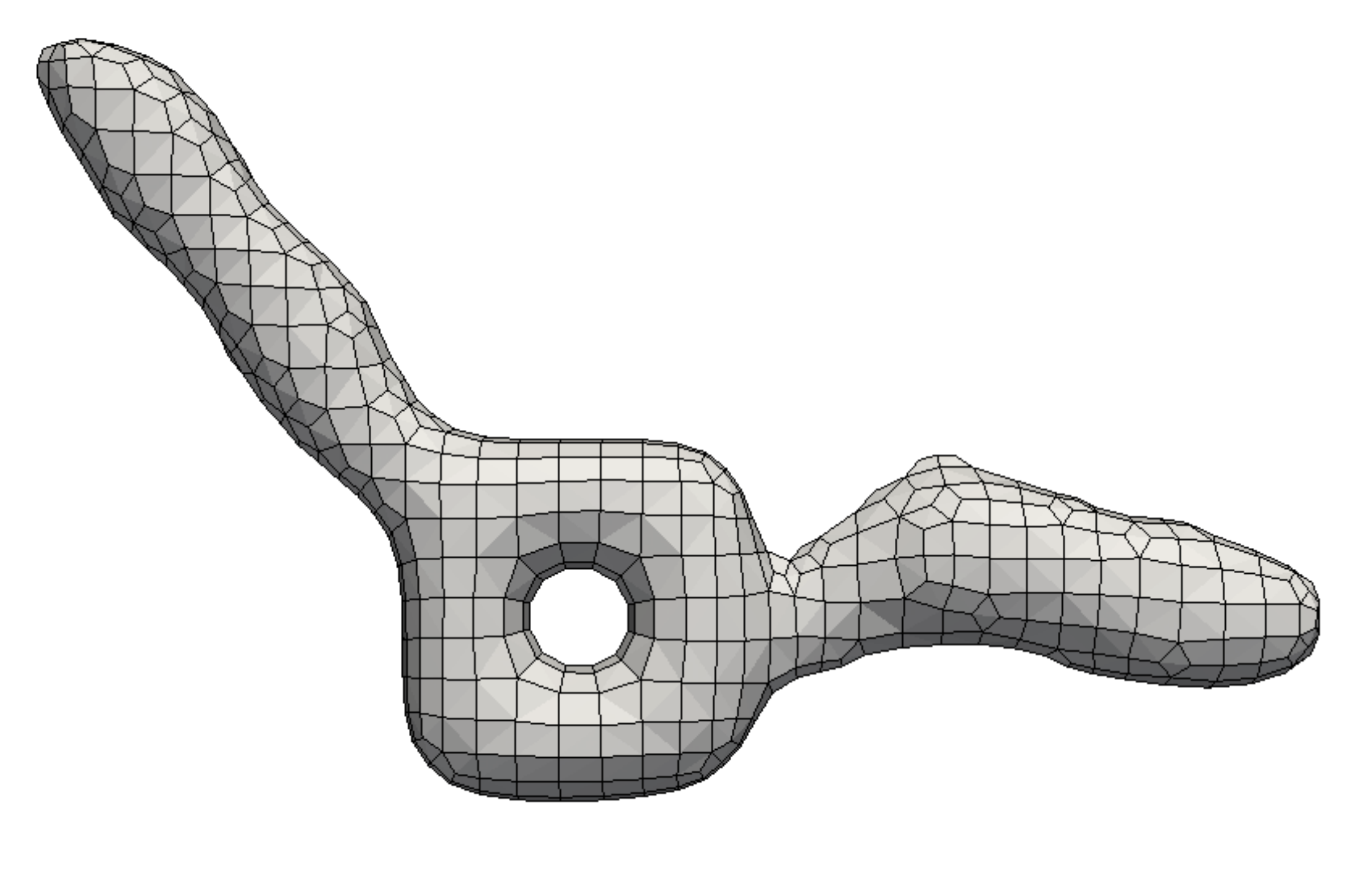}} \\\cline{2-7}
     & \checkmark & \parbox[m]{6em}{\includegraphics[trim={0cm 0cm 0cm 0cm},clip, width=0.12\textwidth]{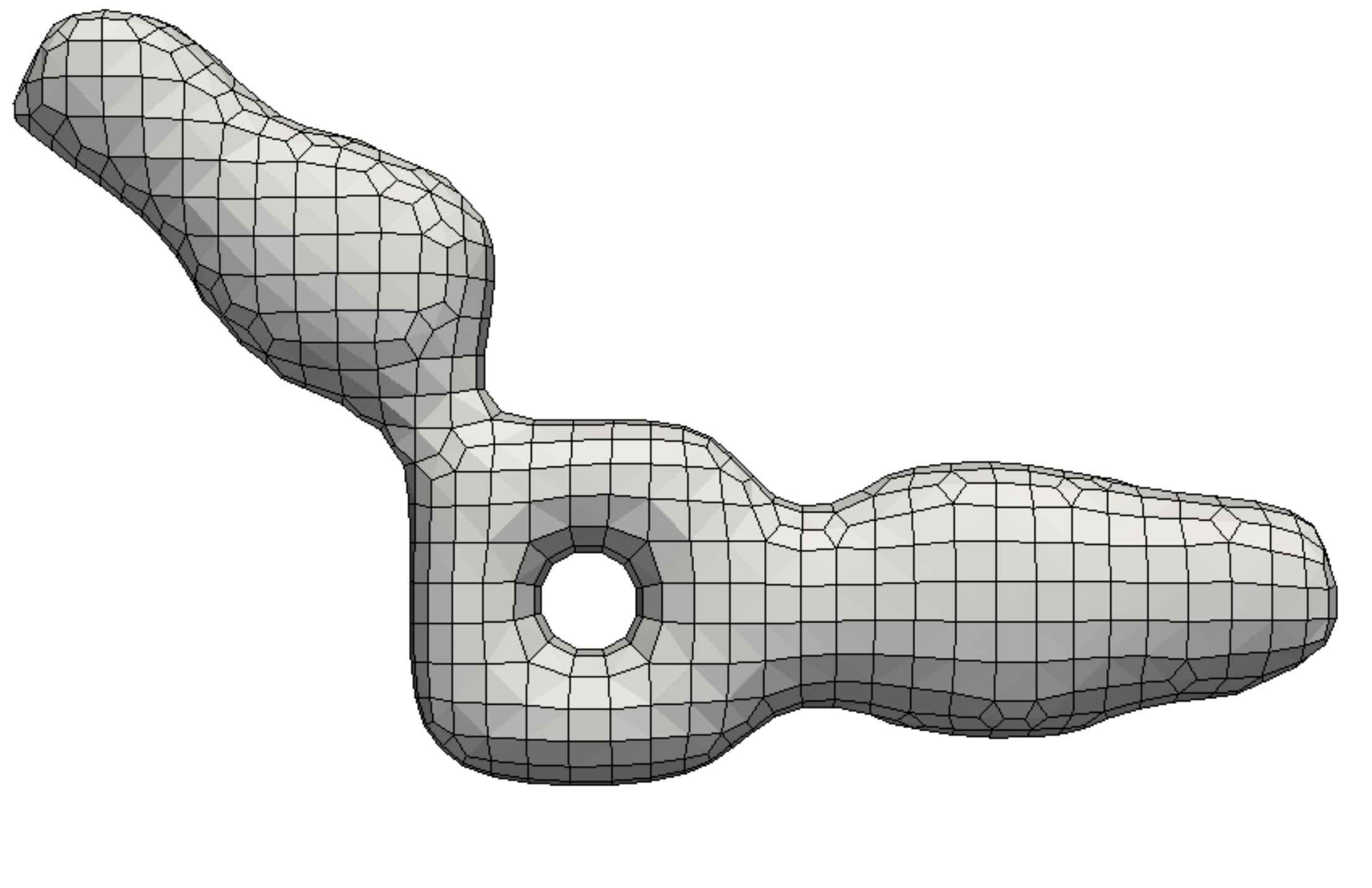}} & \parbox[m]{6em}{\includegraphics[trim={0cm 0cm 0cm 0cm},clip, width=0.12\textwidth]{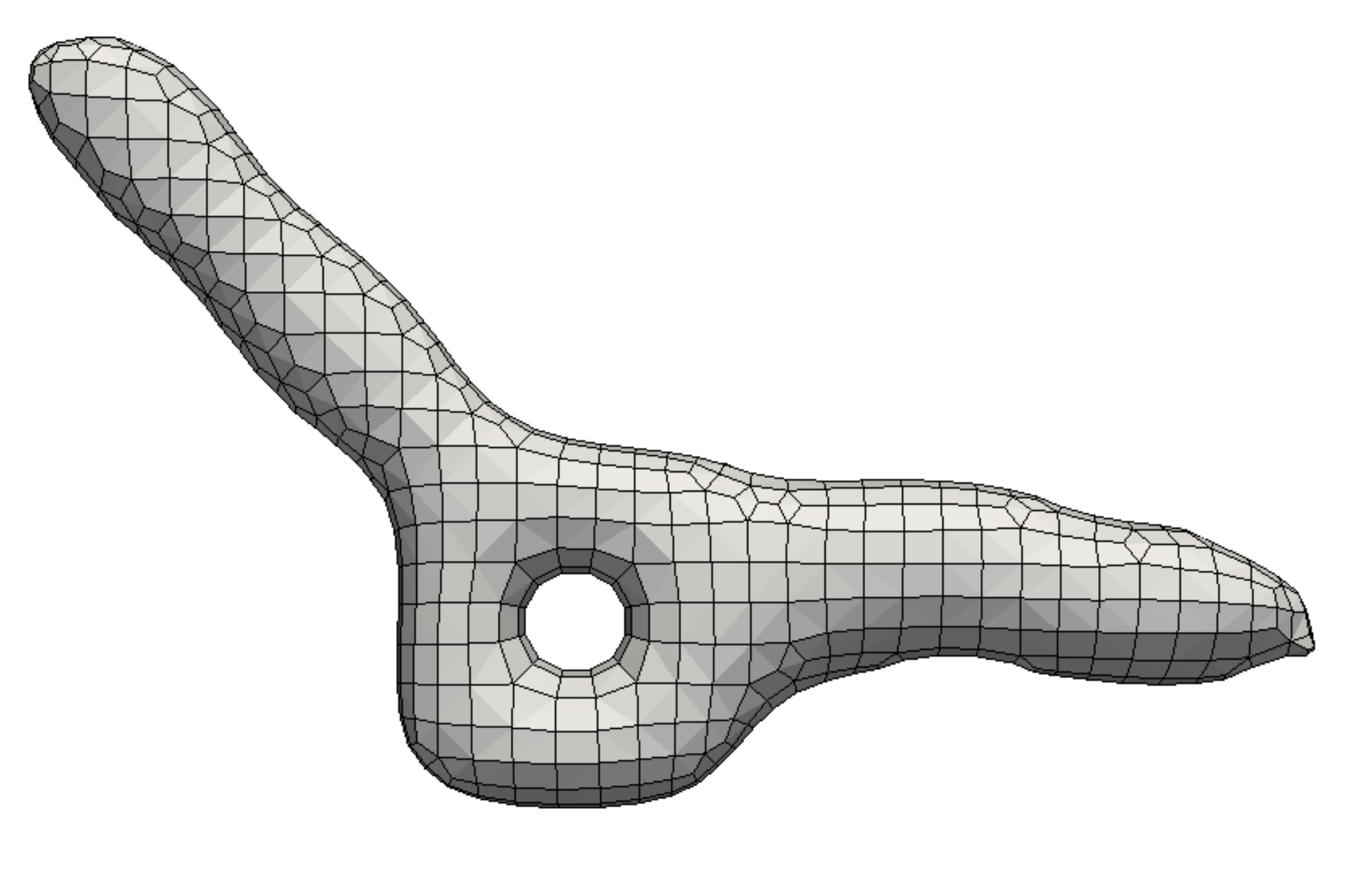}} & \parbox[m]{6em}{\includegraphics[trim={0cm 0cm 0cm 0cm},clip, width=0.12\textwidth]{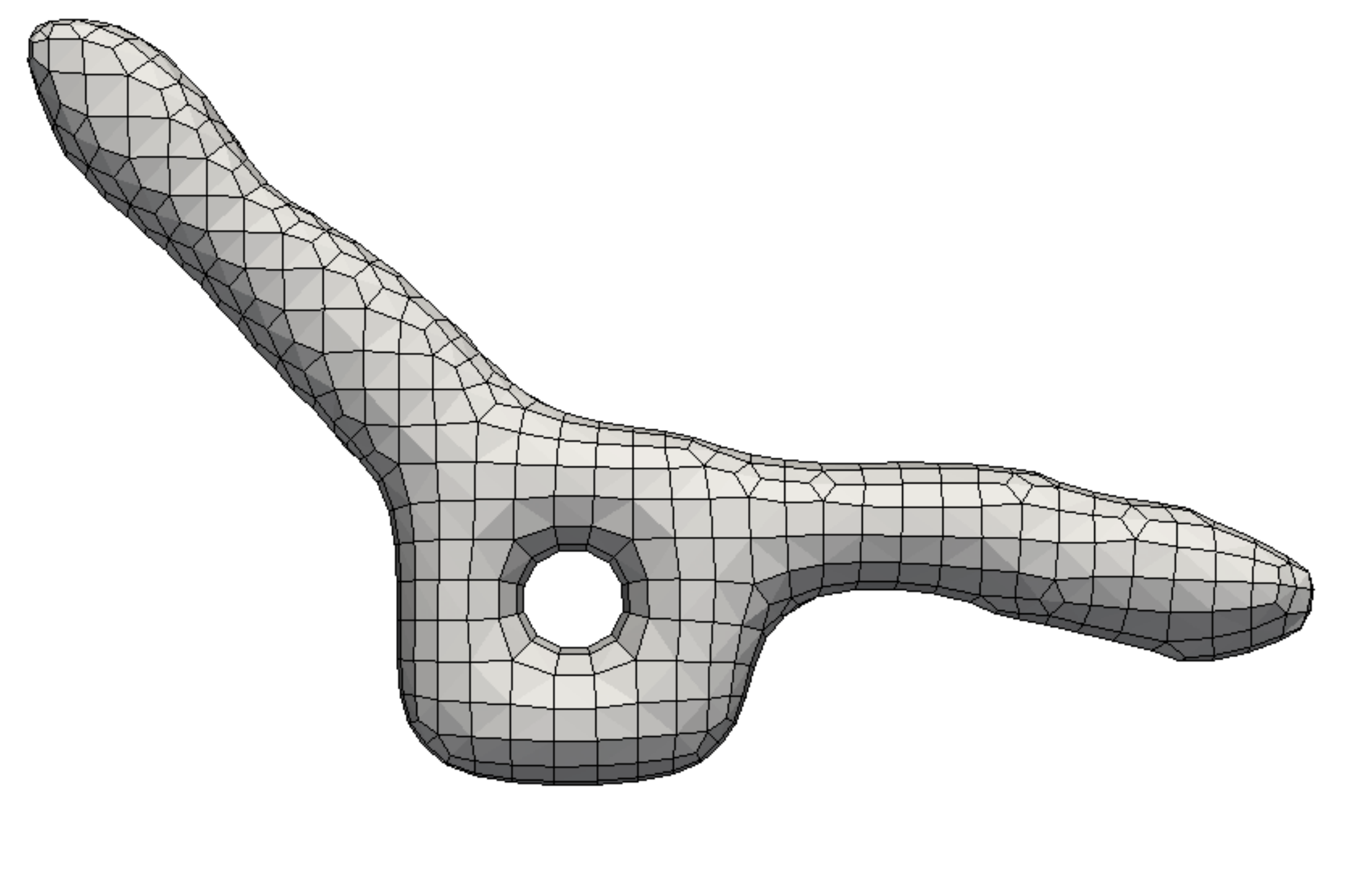}} & \parbox[m]{6em}{\includegraphics[trim={0cm 0cm 0cm 0cm},clip, width=0.12\textwidth]{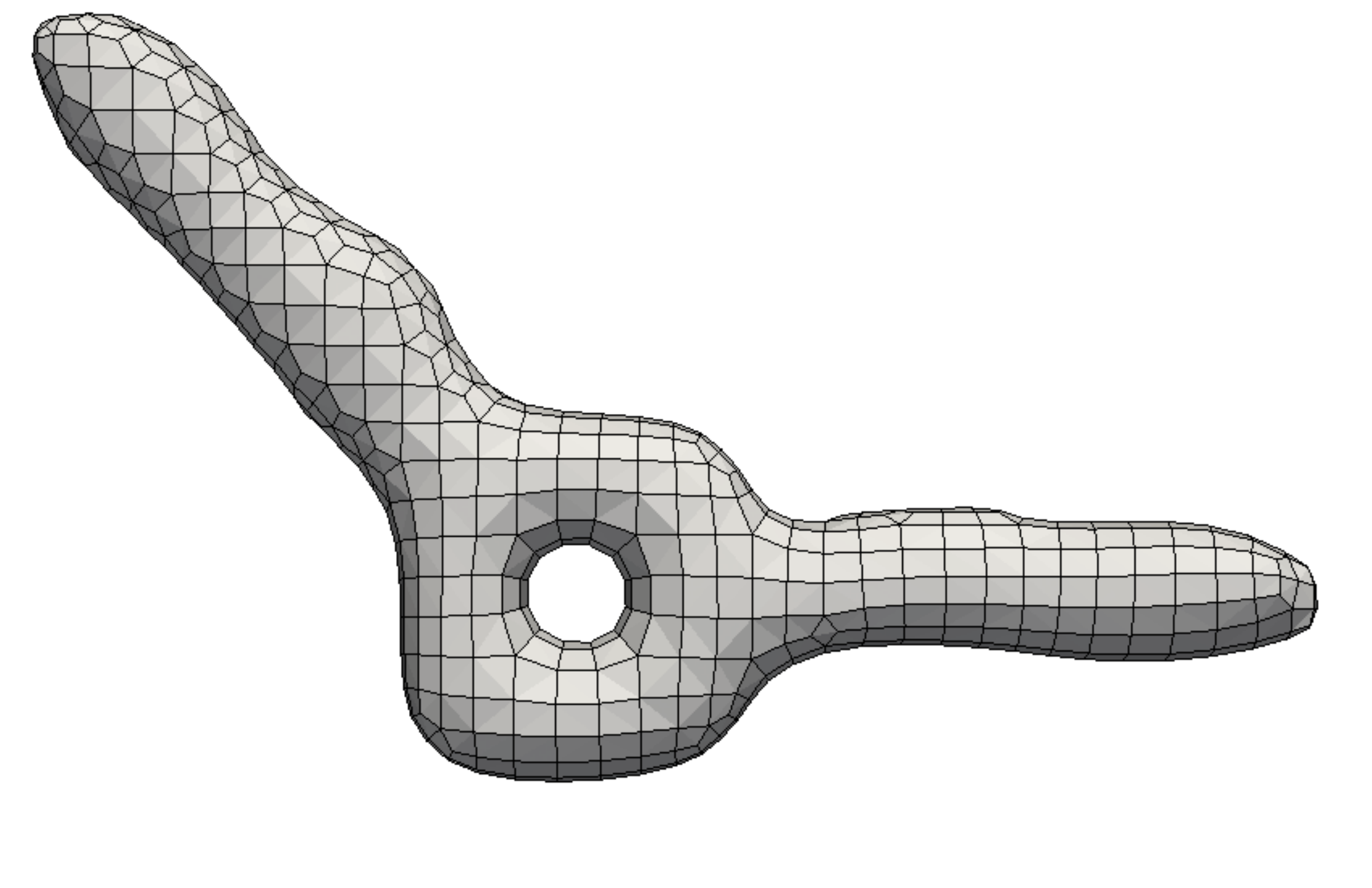}} & \parbox[m]{6em}{\includegraphics[trim={0cm 0cm 0cm 0cm},clip, width=0.12\textwidth]{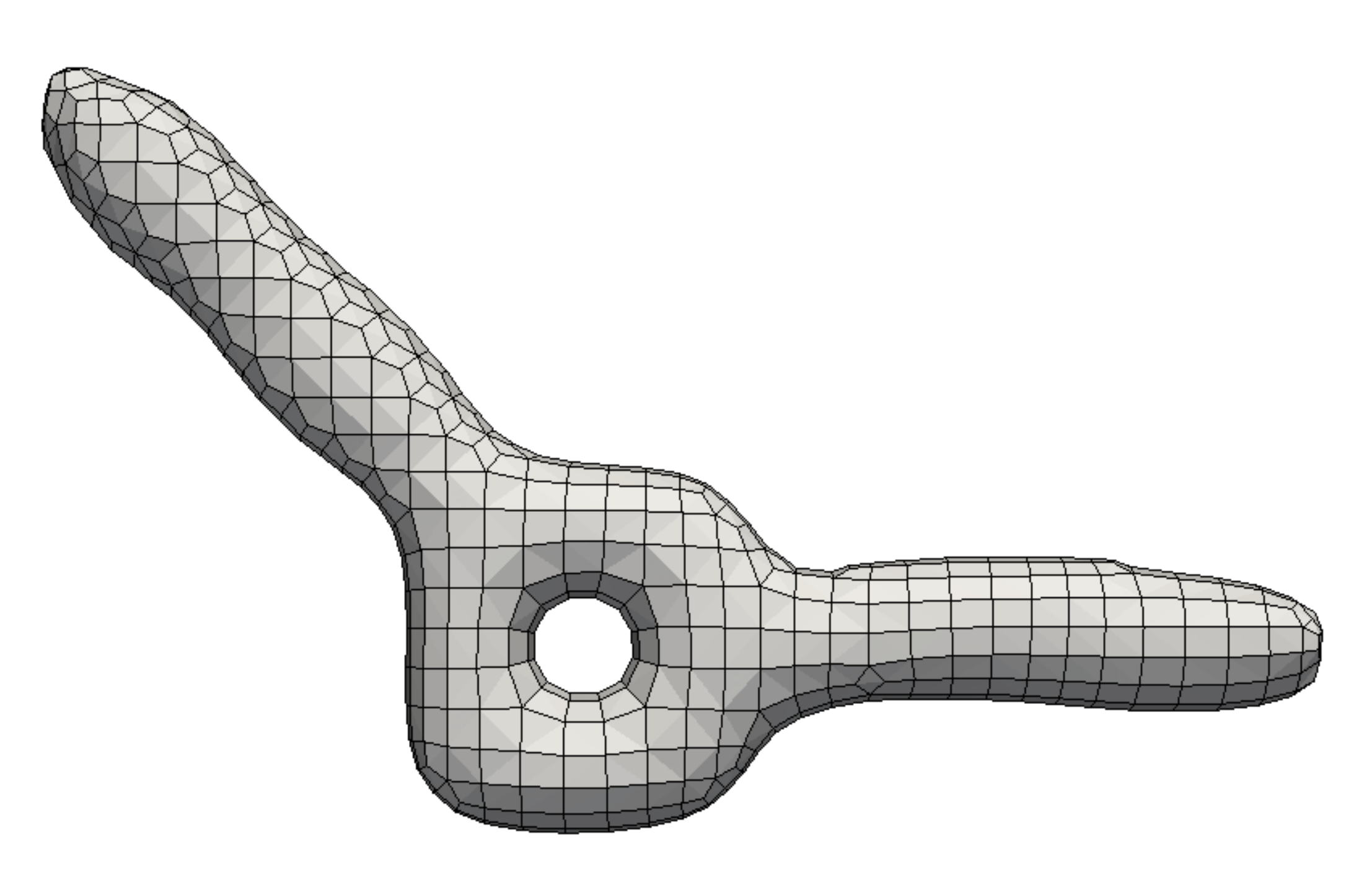}}\\
     \multicolumn{7}{c}{\fbox{\hspace{0.3cm}\parbox[m]{6em}{\includegraphics[trim={0cm 0cm 0cm 0cm},clip, width=0.12\textwidth]{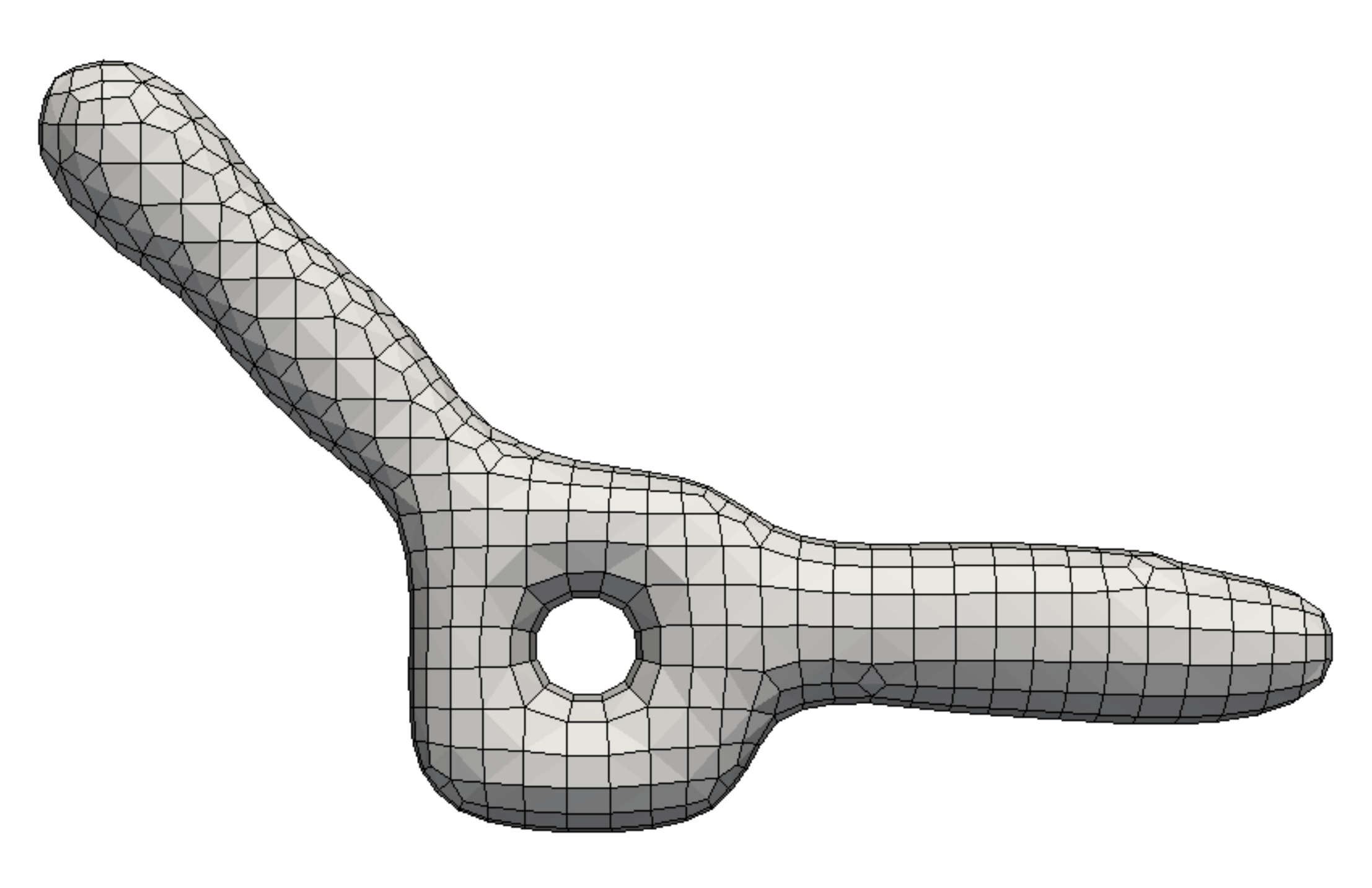}}}}
\end{tabular}
\end{subtable}
\end{table}

\begin{table}[]
\caption{Model predictions of two different problems from the \textbf{sphere simple} validation dataset, using the UNet with different preprocessings and equivariances. We train the models on subsets of the dataset and vary the training size along the columns of the table. At the boxes below the tables we show the corresponding ground truth density for each problem.}
\label{5_fig:sphere_simple_predictions}
\begin{subtable}[h]{0.99\textwidth}
    \centering\setcellgapes{3pt}\makegapedcells
    \setlength\tabcolsep{3.5pt}
    \begin{tabular}{c|c||ScScScScSc}
    \multicolumn{2}{c||}{} & \multicolumn{5}{c}{training samples} \\\hline
     prepr. & equiv. & 10 & 50 & 100 & 500 & 1500 \\\hline
     \multirow{2}{*}{\rotatebox{90}{trivial}} & & \parbox[m]{6em}{\includegraphics[trim={0cm 0cm 0cm 0cm},clip, width=0.12\textwidth]{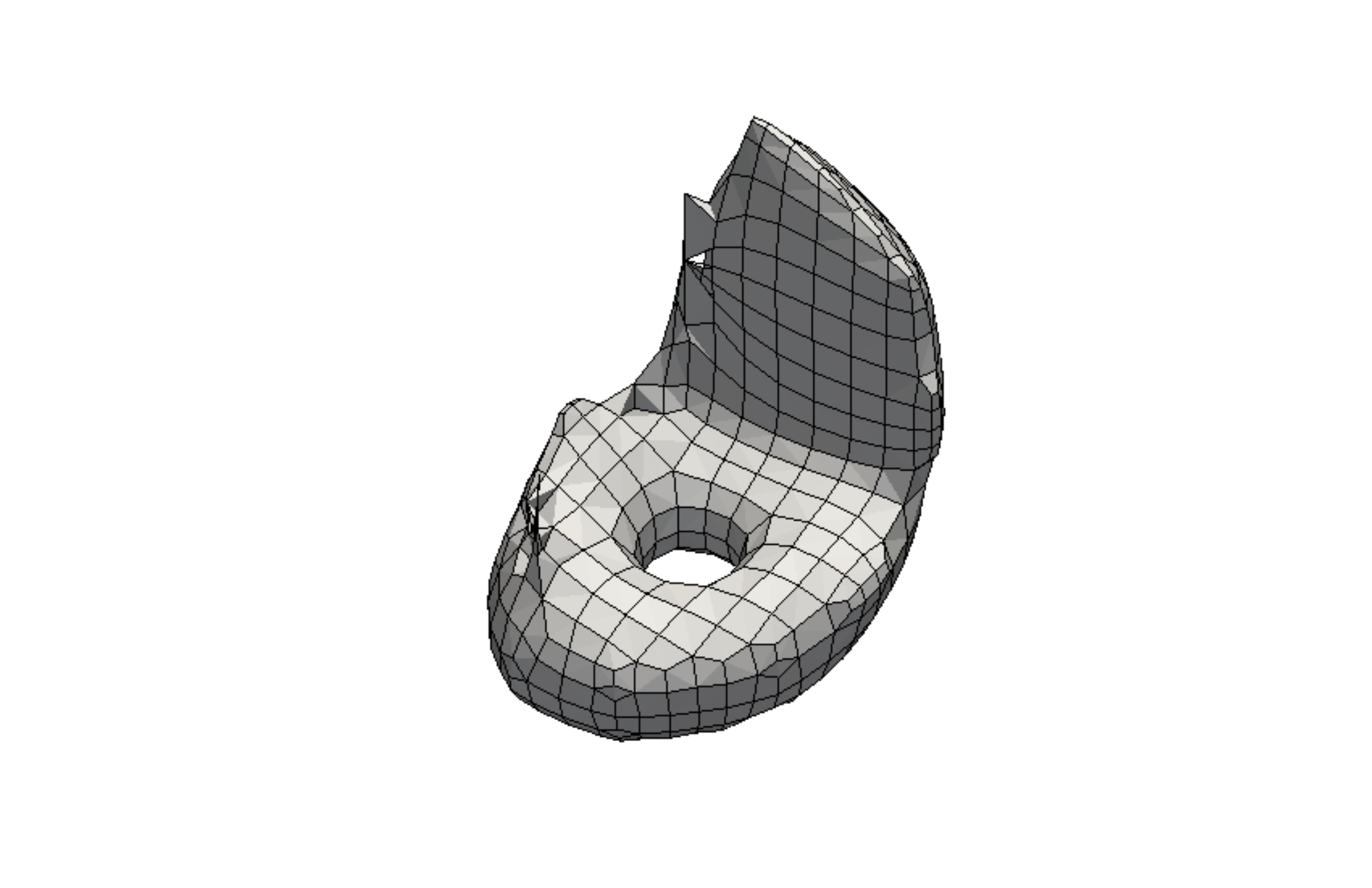}} & \parbox[m]{6em}{\includegraphics[trim={0cm 0cm 0cm 0cm},clip, width=0.12\textwidth]{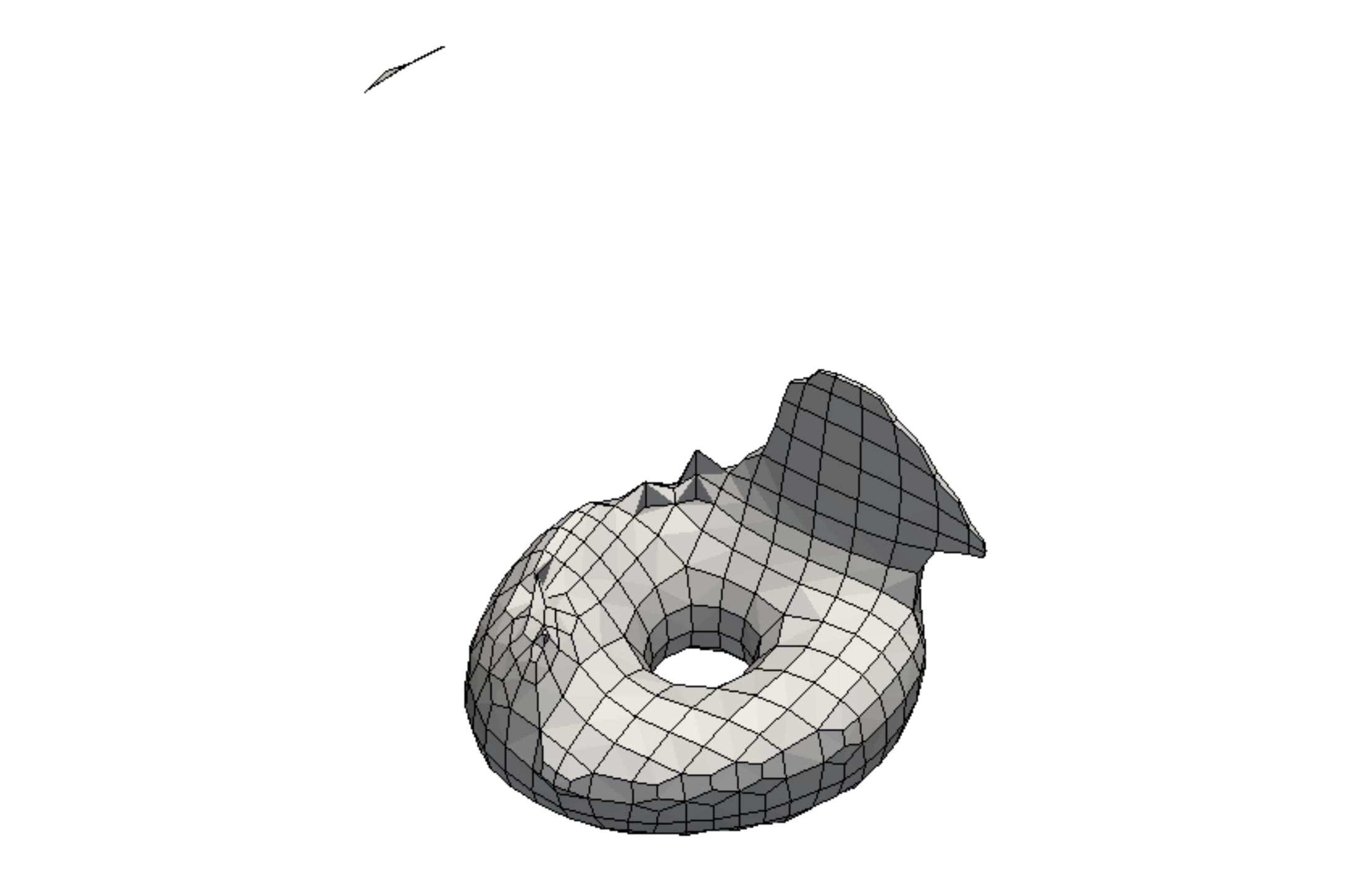}} & \parbox[m]{6em}{\includegraphics[trim={0cm 0cm 0cm 0cm},clip, width=0.12\textwidth]{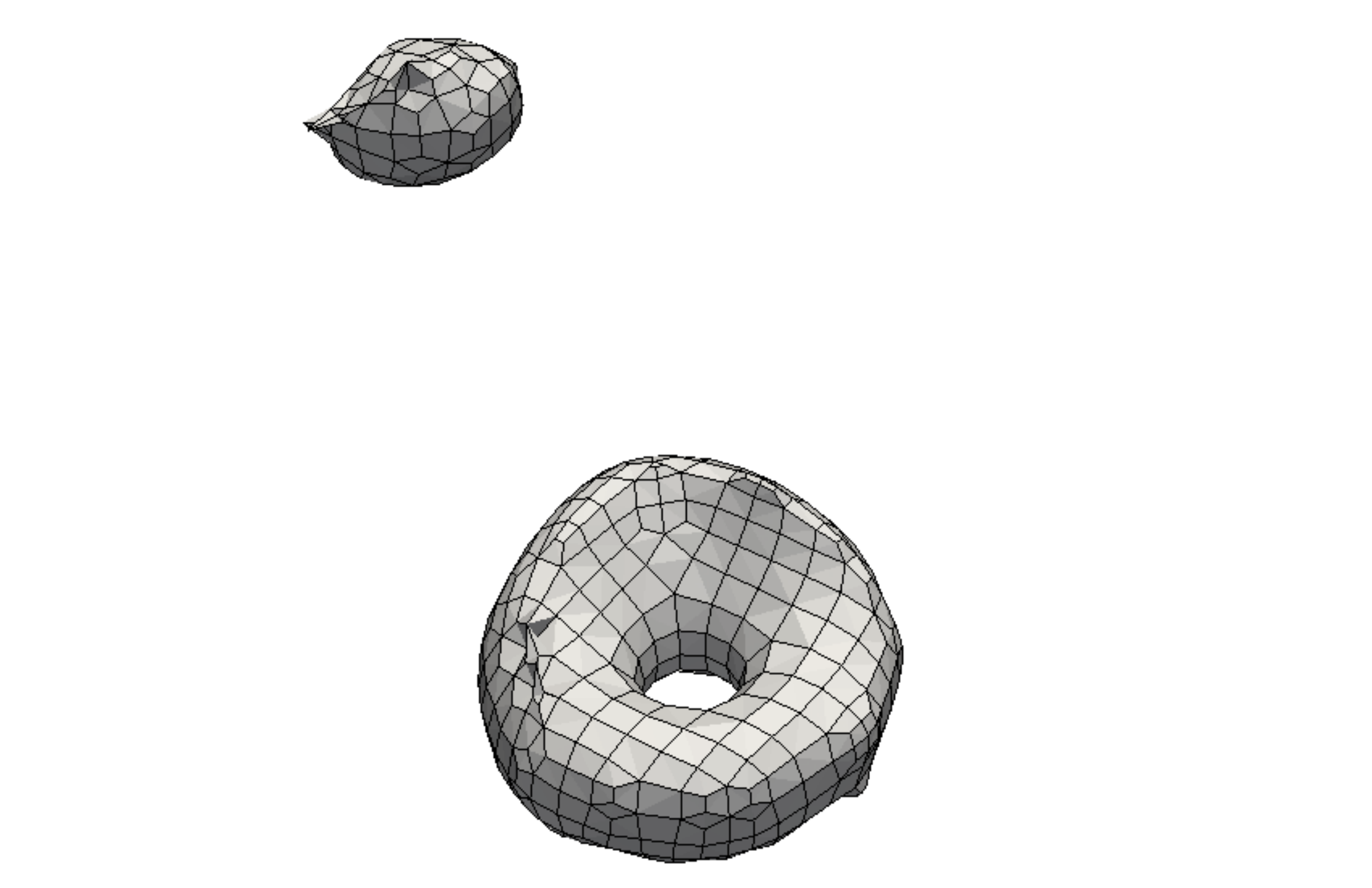}} & \parbox[m]{6em}{\includegraphics[trim={0cm 0cm 0cm 0cm},clip, width=0.12\textwidth]{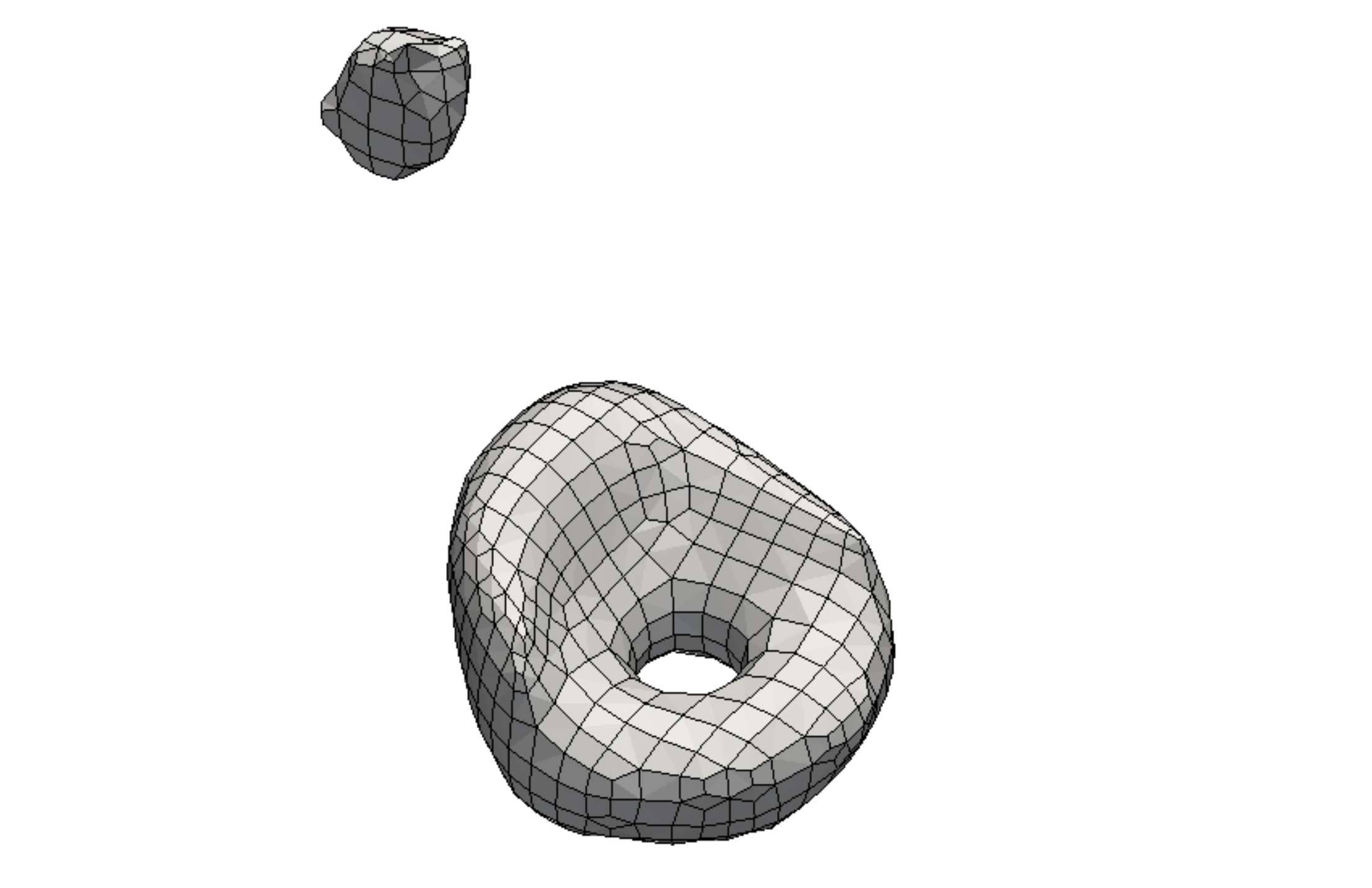}} & \parbox[m]{6em}{\includegraphics[trim={0cm 0cm 0cm 0cm},clip, width=0.12\textwidth]{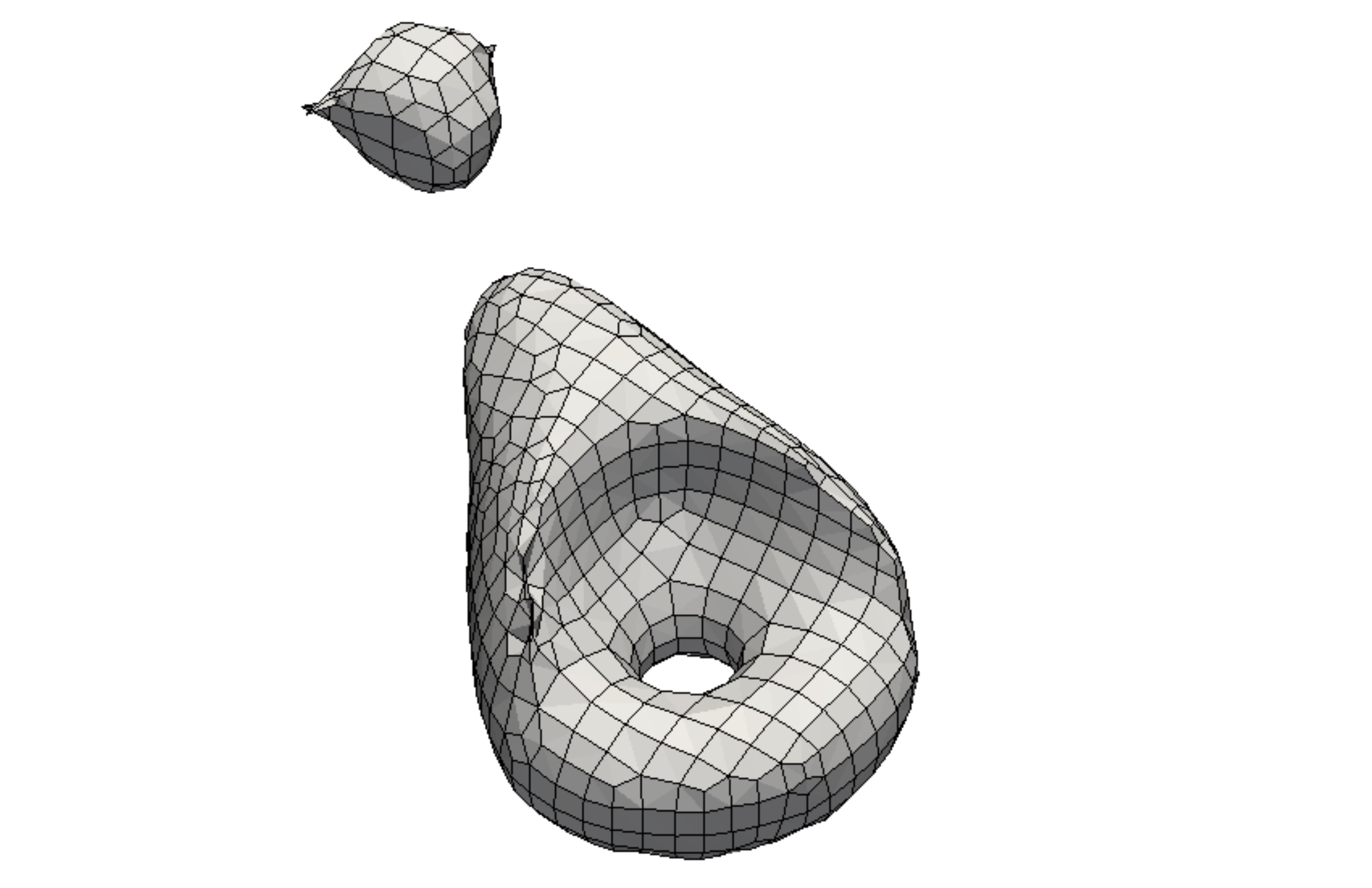}} \\\cline{2-7}
     & \checkmark
     & \parbox[m]{6em}{\includegraphics[trim={0cm 0cm 0cm 0cm},clip, width=0.12\textwidth]{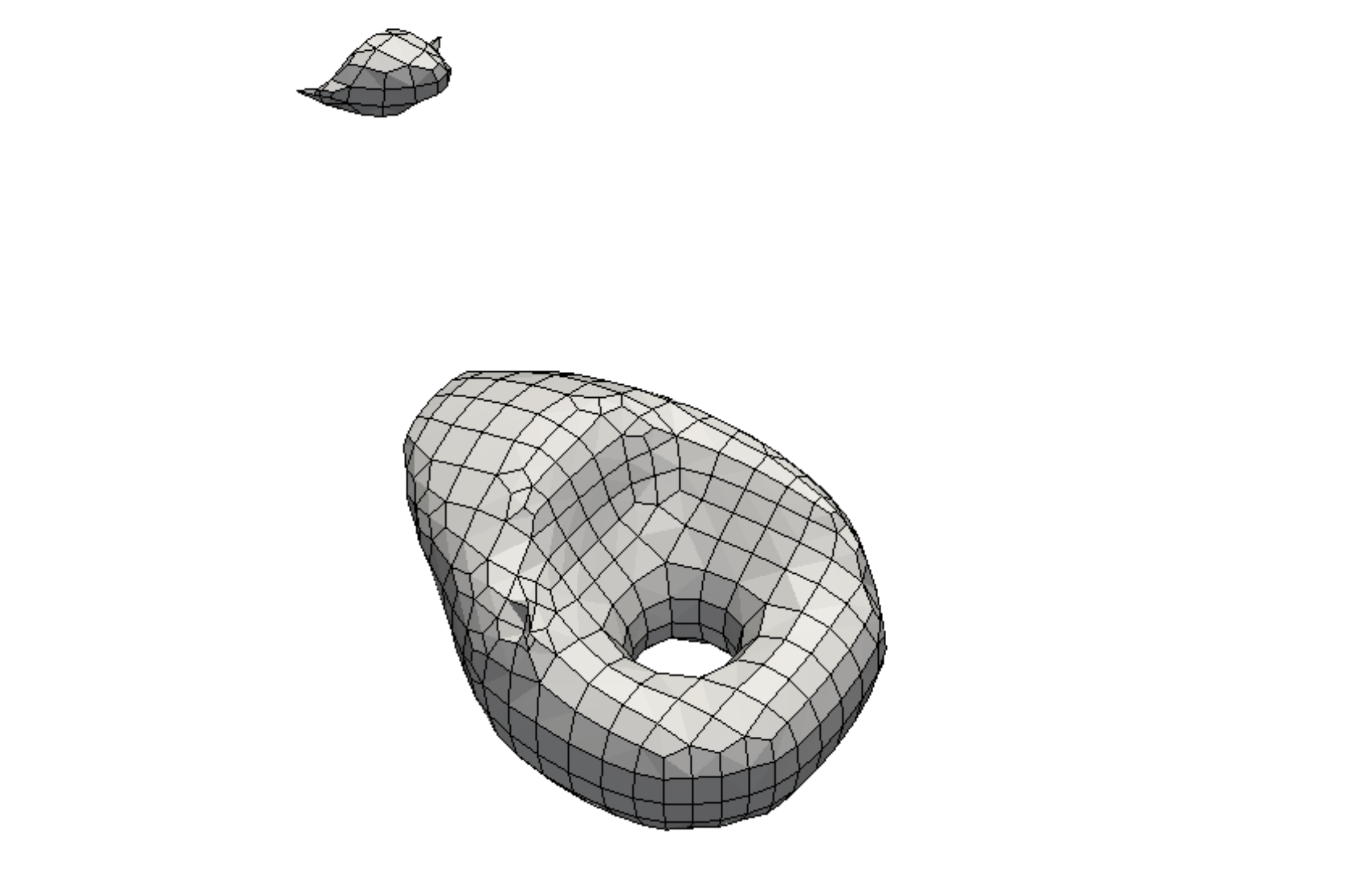}} & \parbox[m]{6em}{\includegraphics[trim={0cm 0cm 0cm 0cm},clip, width=0.12\textwidth]{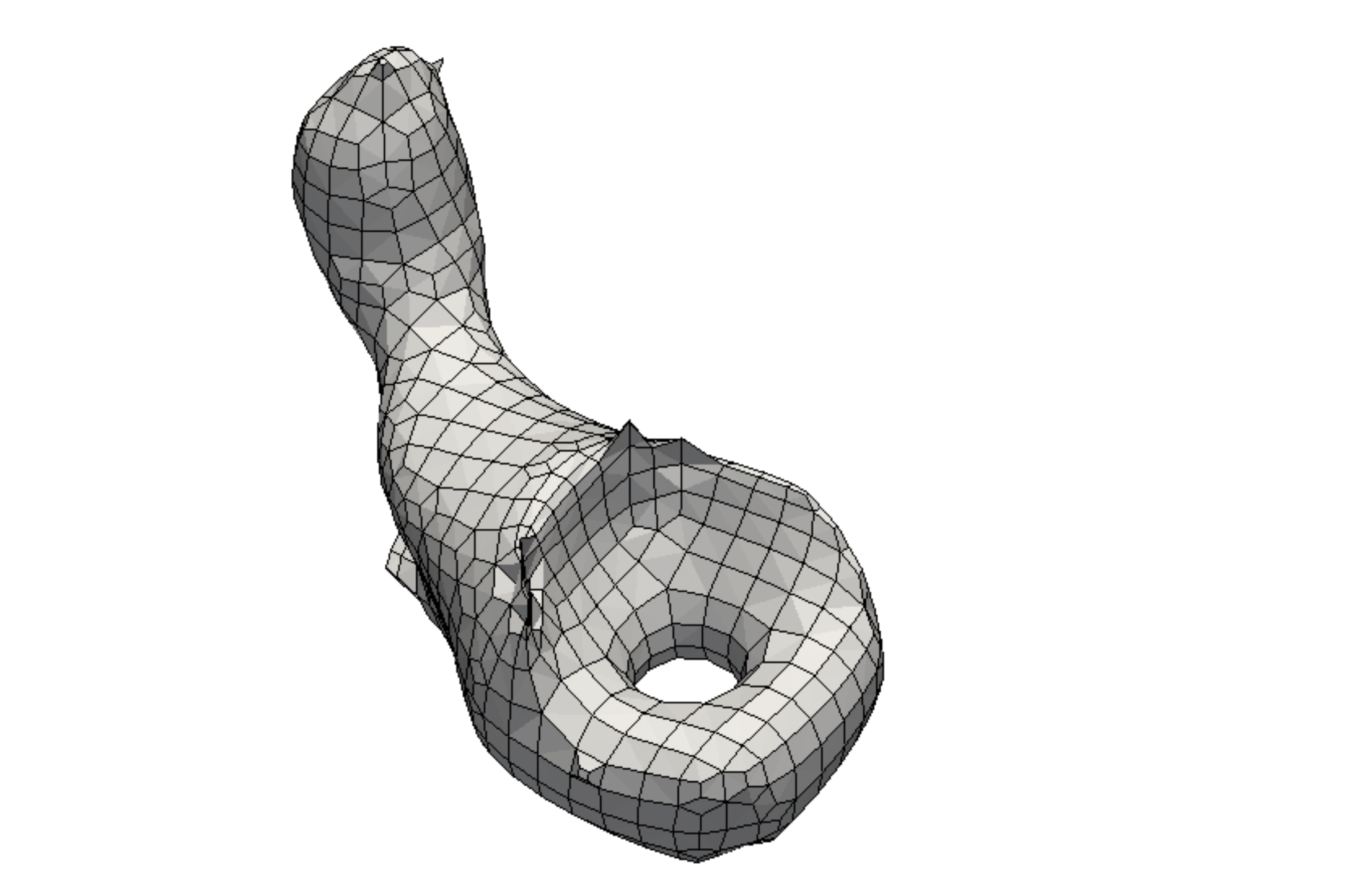}} & \parbox[m]{6em}{\includegraphics[trim={0cm 0cm 0cm 0cm},clip, width=0.12\textwidth]{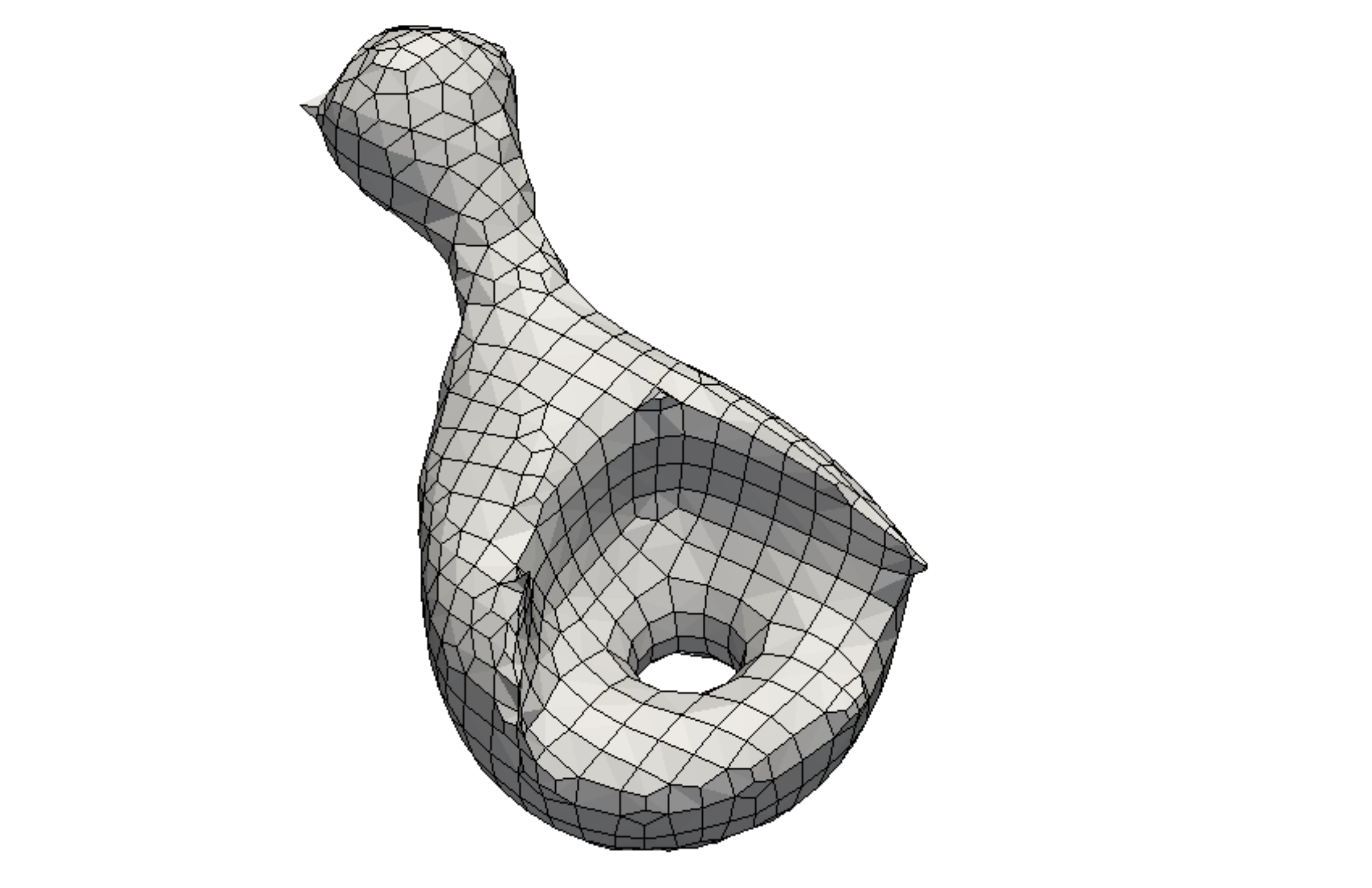}} & \parbox[m]{6em}{\includegraphics[trim={0cm 0cm 0cm 0cm},clip, width=0.12\textwidth]{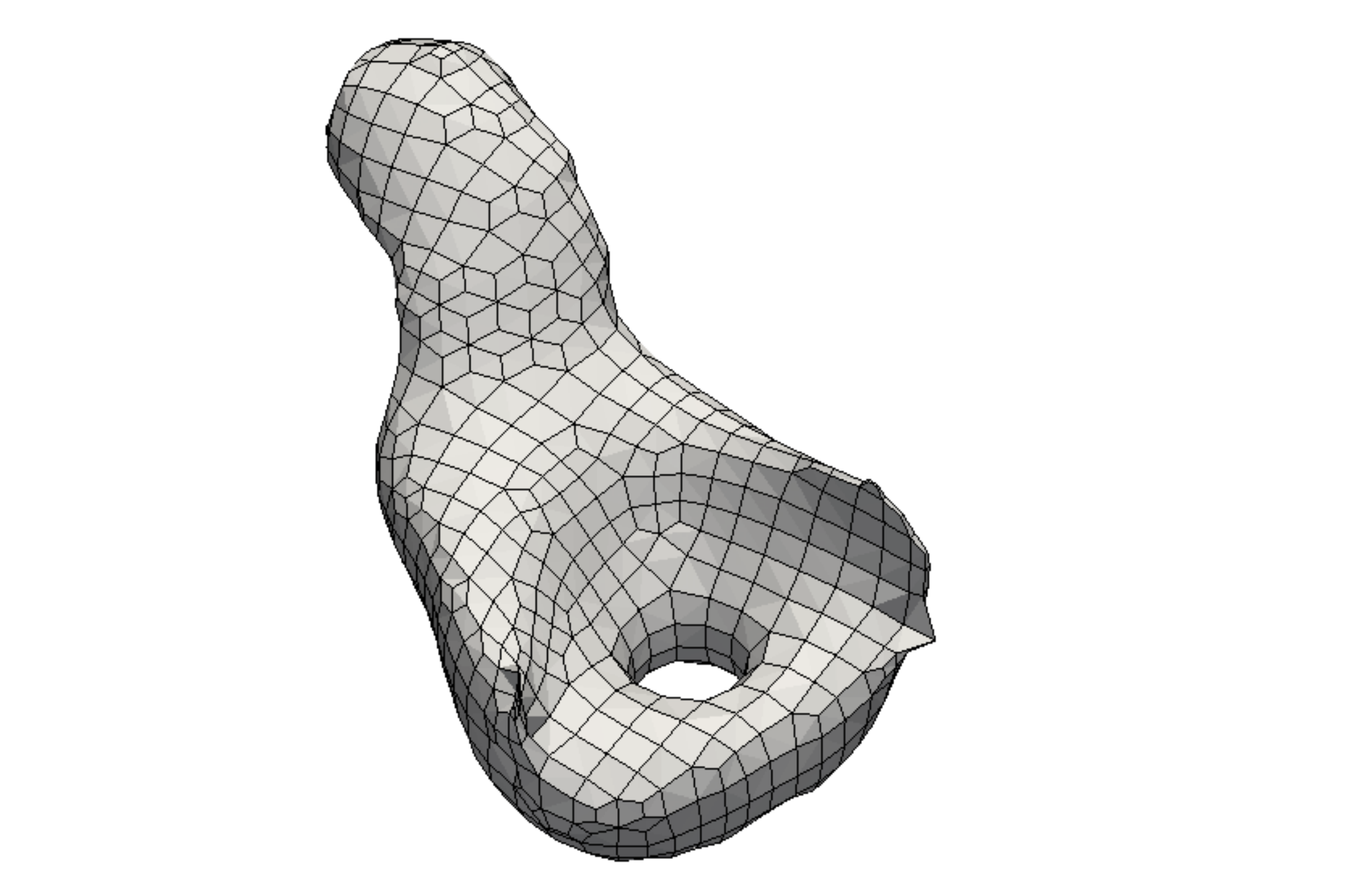}} & \parbox[m]{6em}{\includegraphics[trim={0cm 0cm 0cm 0cm},clip, width=0.12\textwidth]{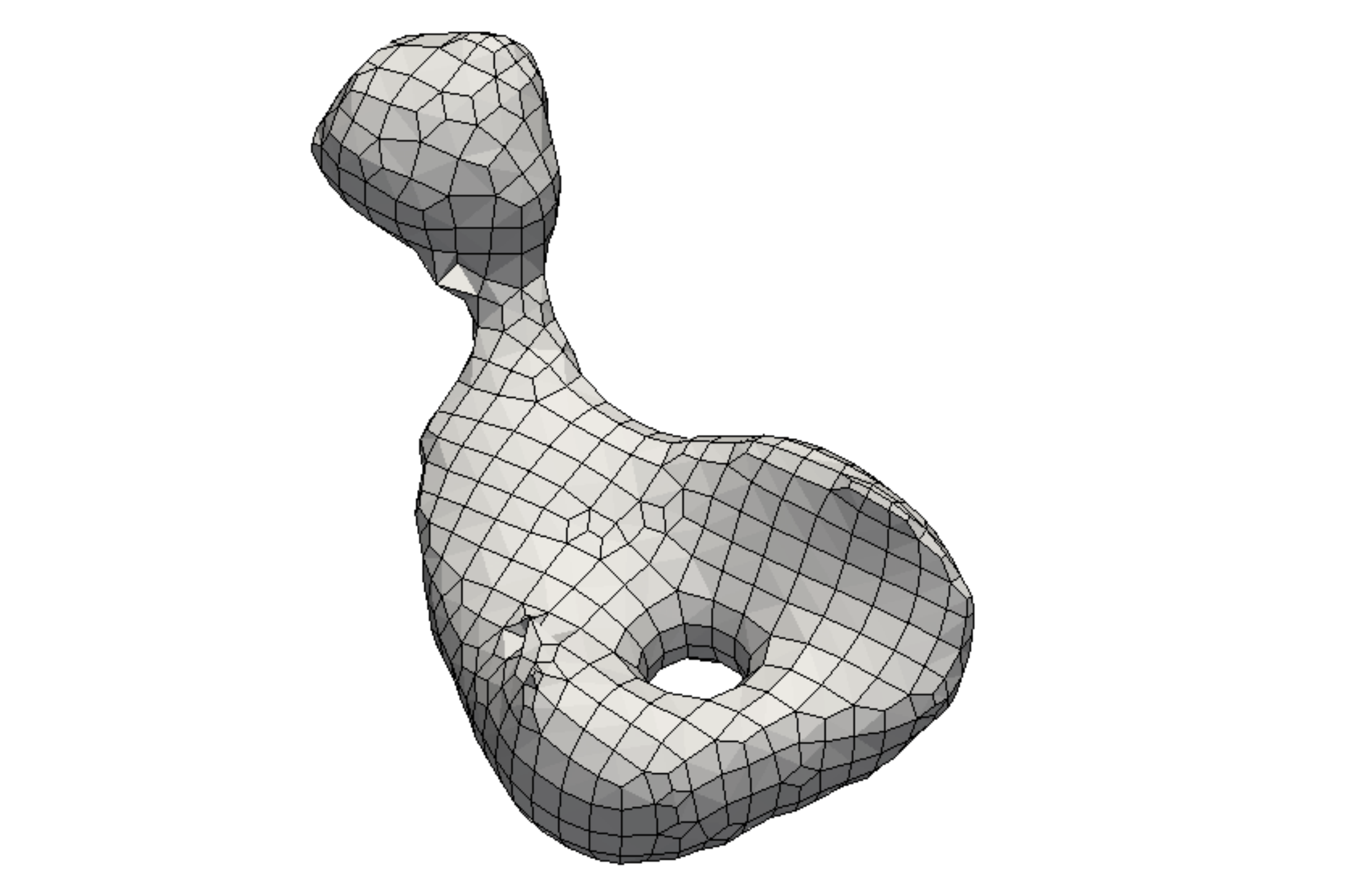}} \\\hline
     \multirow{2}{*}{\rotatebox{90}{trivial+PDE}} & & \parbox[m]{6em}{\includegraphics[trim={0cm 0cm 0cm 0cm},clip, width=0.12\textwidth]{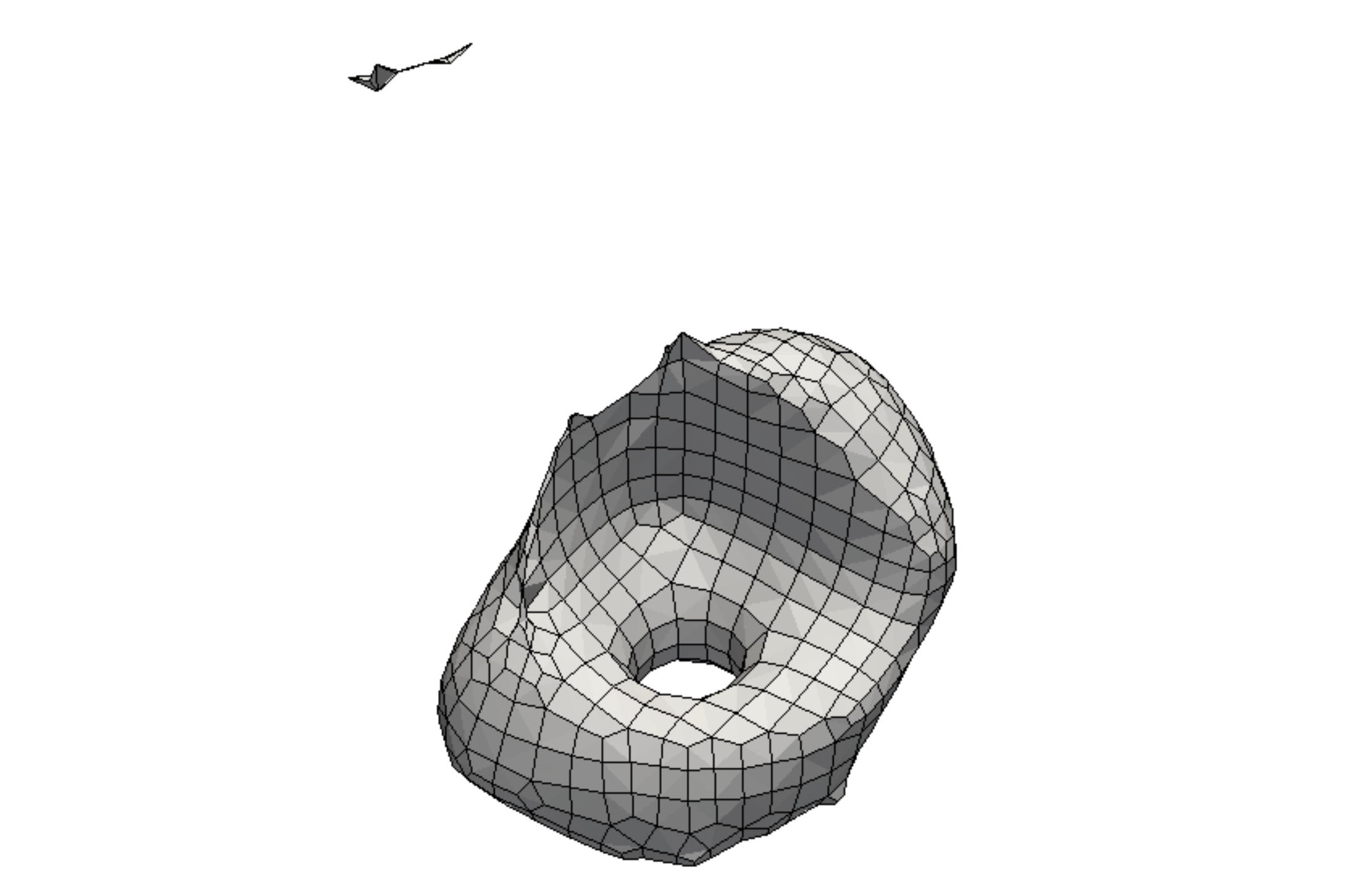}} & \parbox[m]{6em}{\includegraphics[trim={0cm 0cm 0cm 0cm},clip, width=0.12\textwidth]{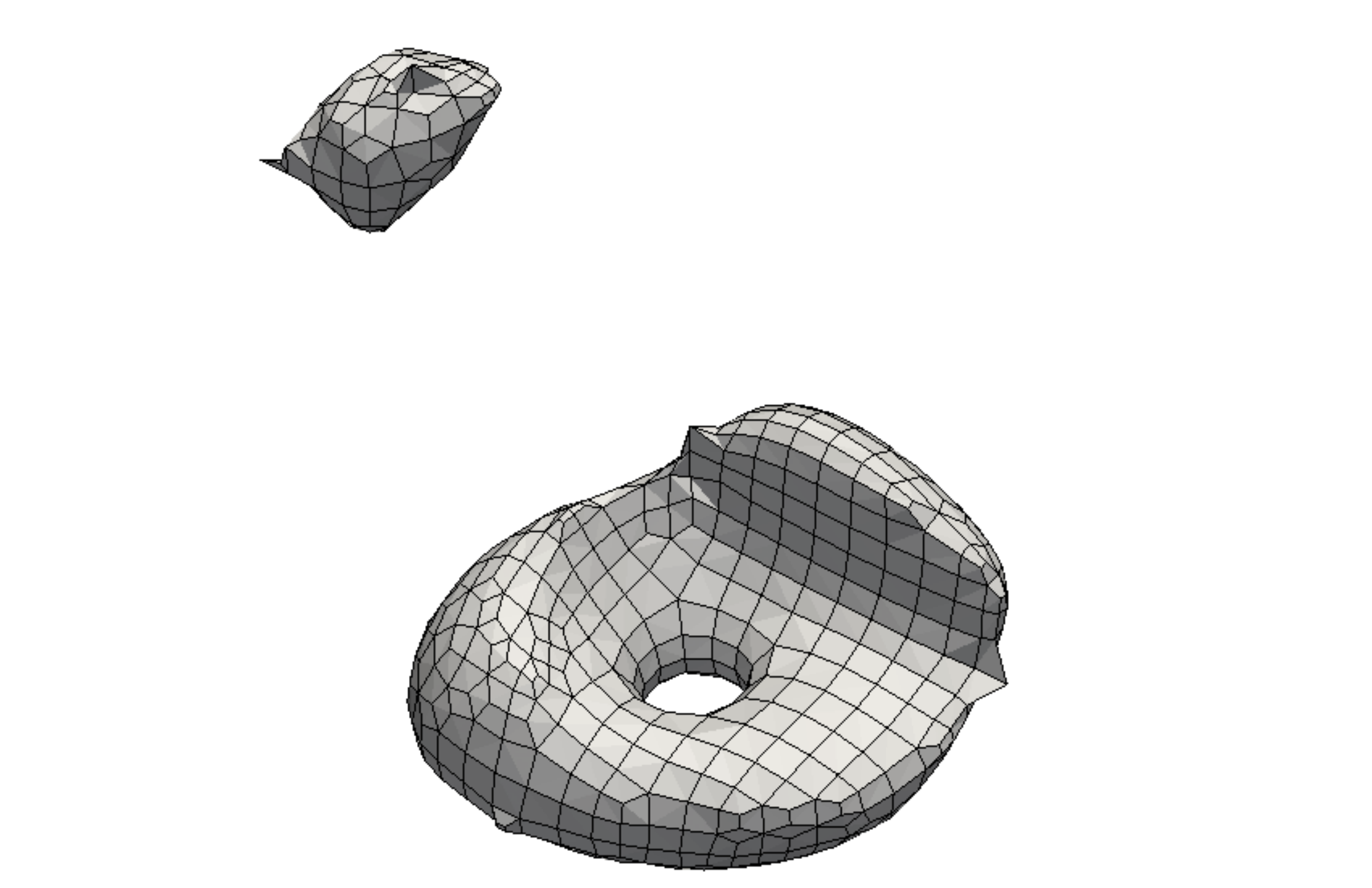}} & \parbox[m]{6em}{\includegraphics[trim={0cm 0cm 0cm 0cm},clip, width=0.12\textwidth]{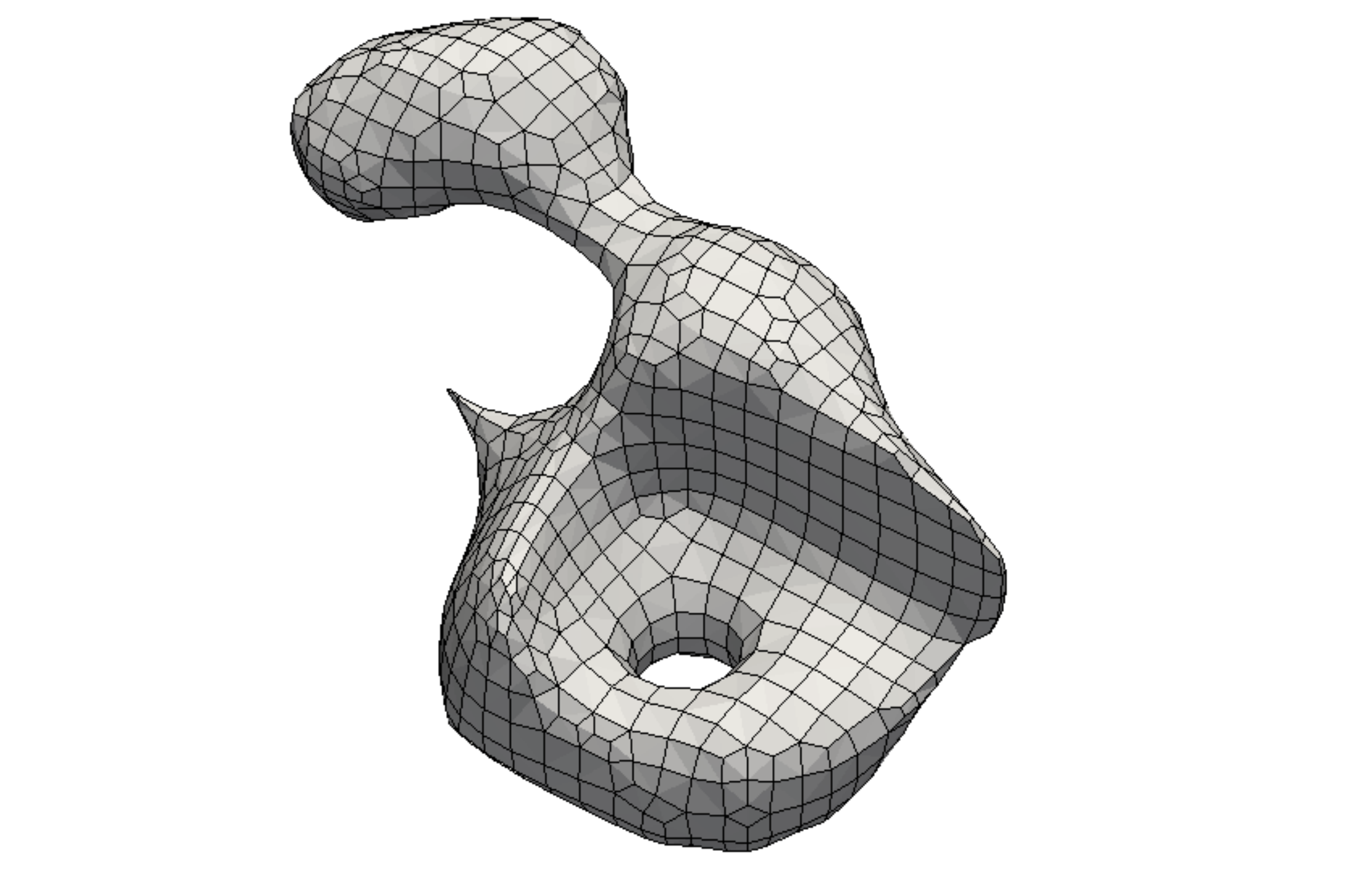}} & \parbox[m]{6em}{\includegraphics[trim={0cm 0cm 0cm 0cm},clip, width=0.12\textwidth]{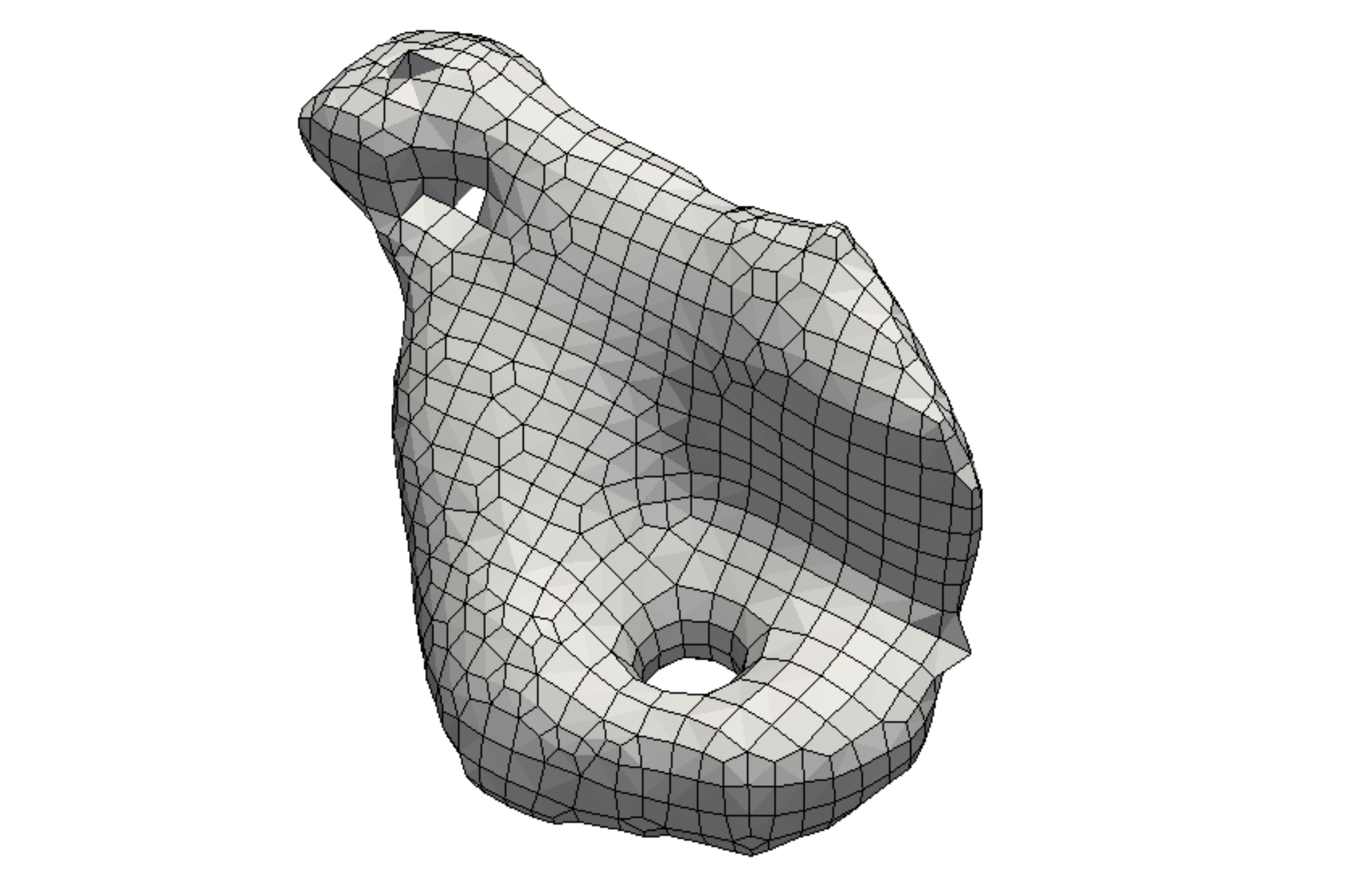}} & \parbox[m]{6em}{\includegraphics[trim={0cm 0cm 0cm 0cm},clip, width=0.12\textwidth]{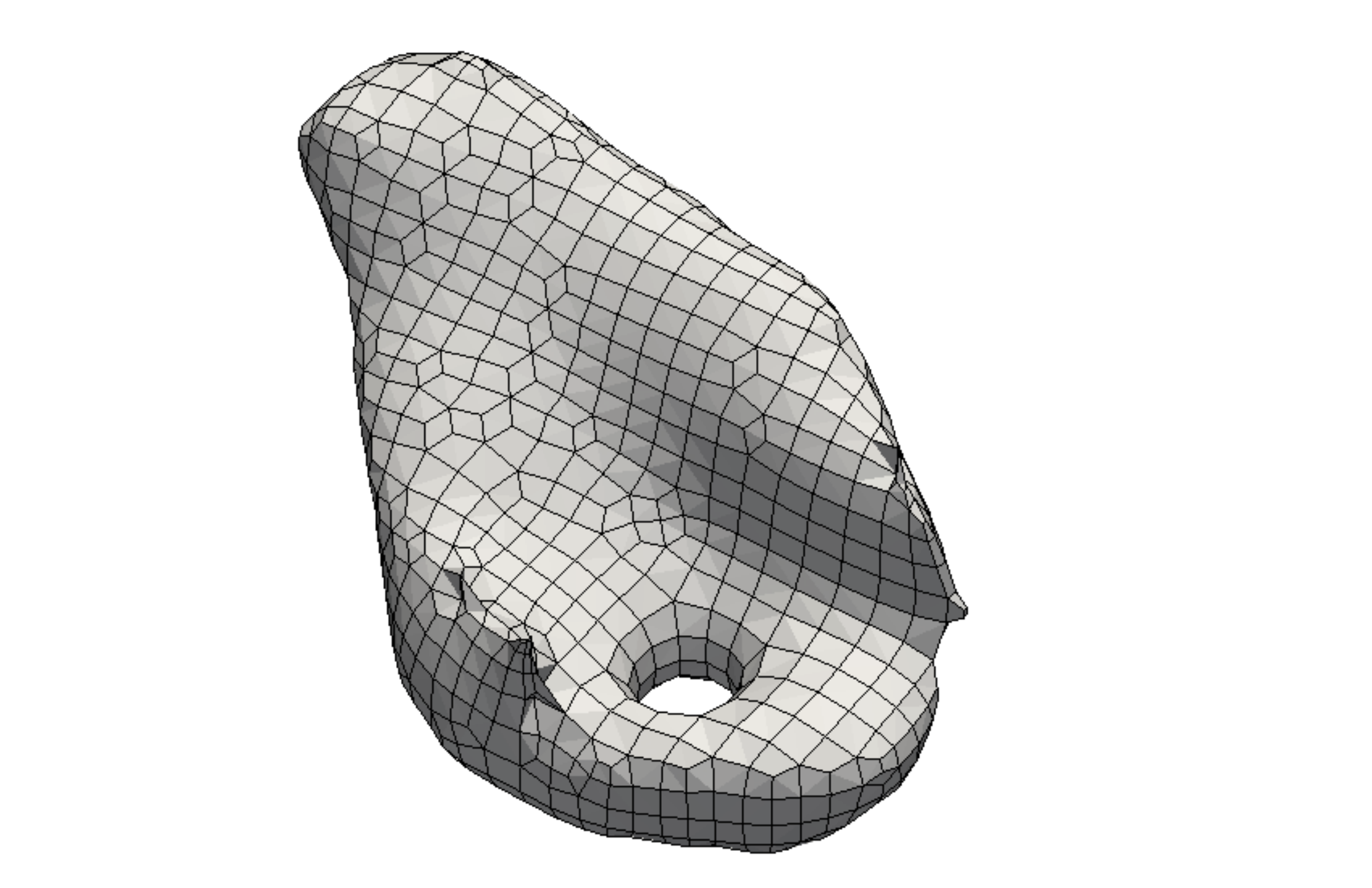}} \\\cline{2-7}
     & \checkmark & \parbox[m]{6em}{\includegraphics[trim={0cm 0cm 0cm 0cm},clip, width=0.12\textwidth]{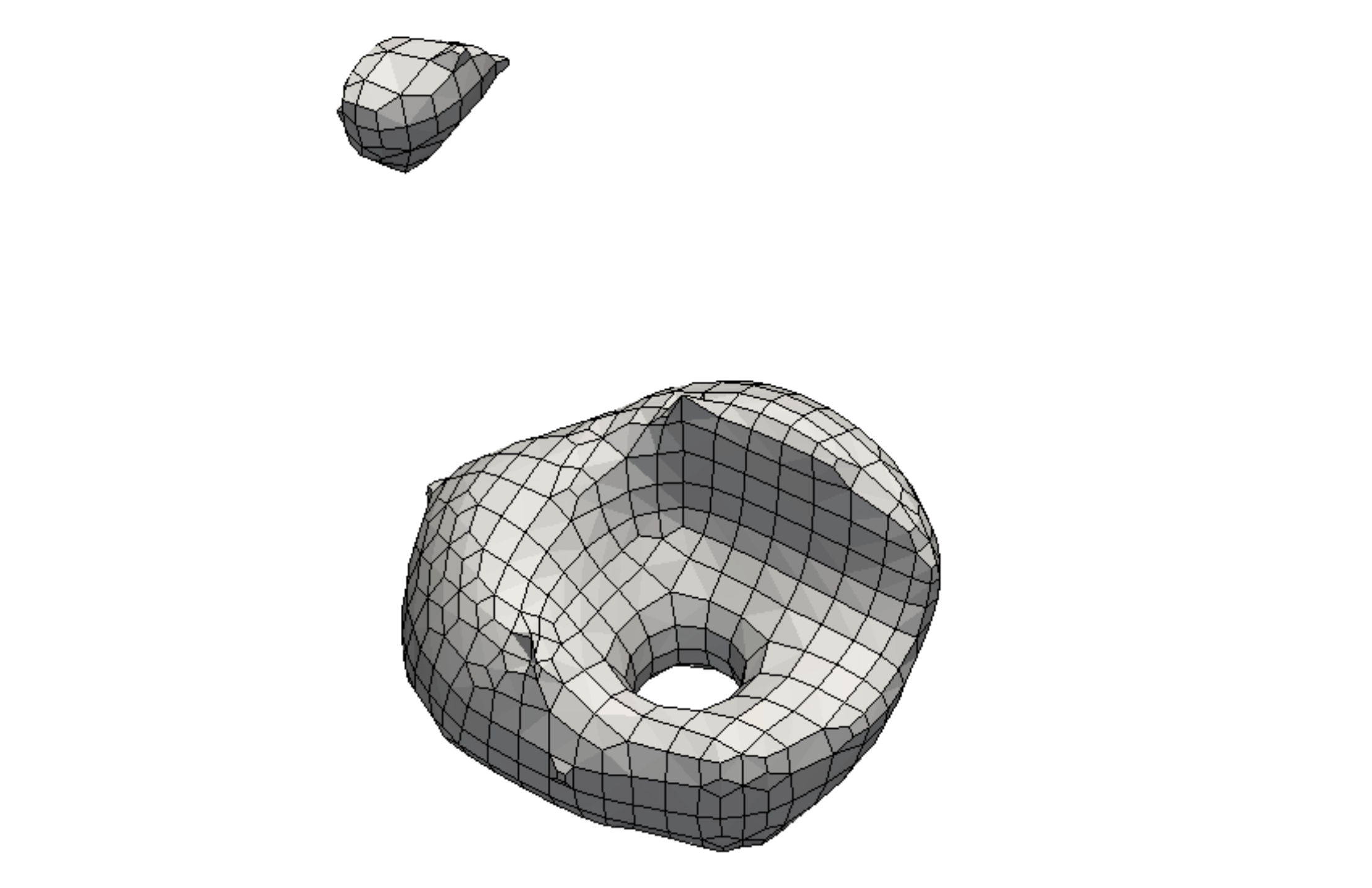}} & \parbox[m]{6em}{\includegraphics[trim={0cm 0cm 0cm 0cm},clip, width=0.12\textwidth]{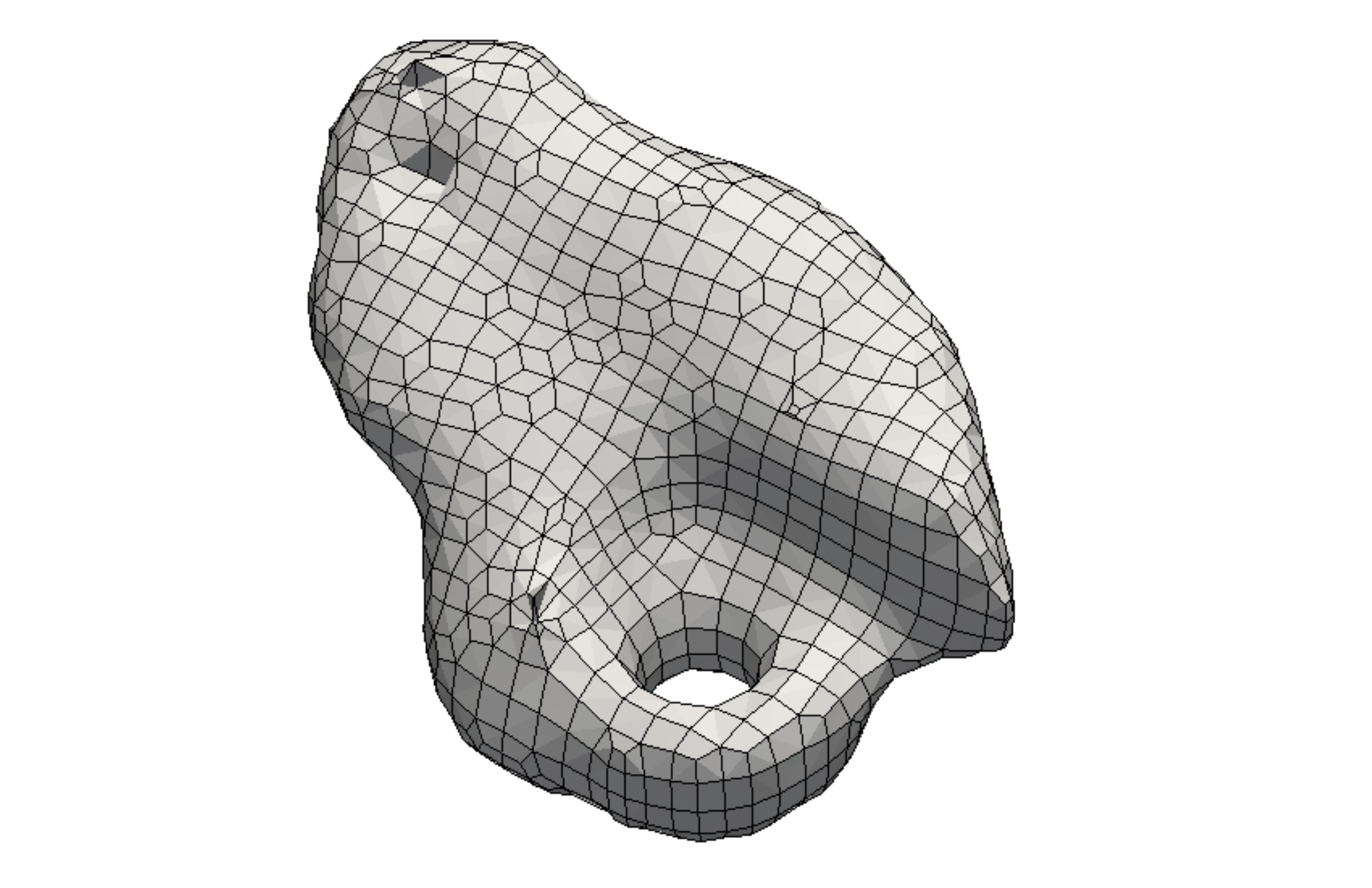}} & \parbox[m]{6em}{\includegraphics[trim={0cm 0cm 0cm 0cm},clip, width=0.12\textwidth]{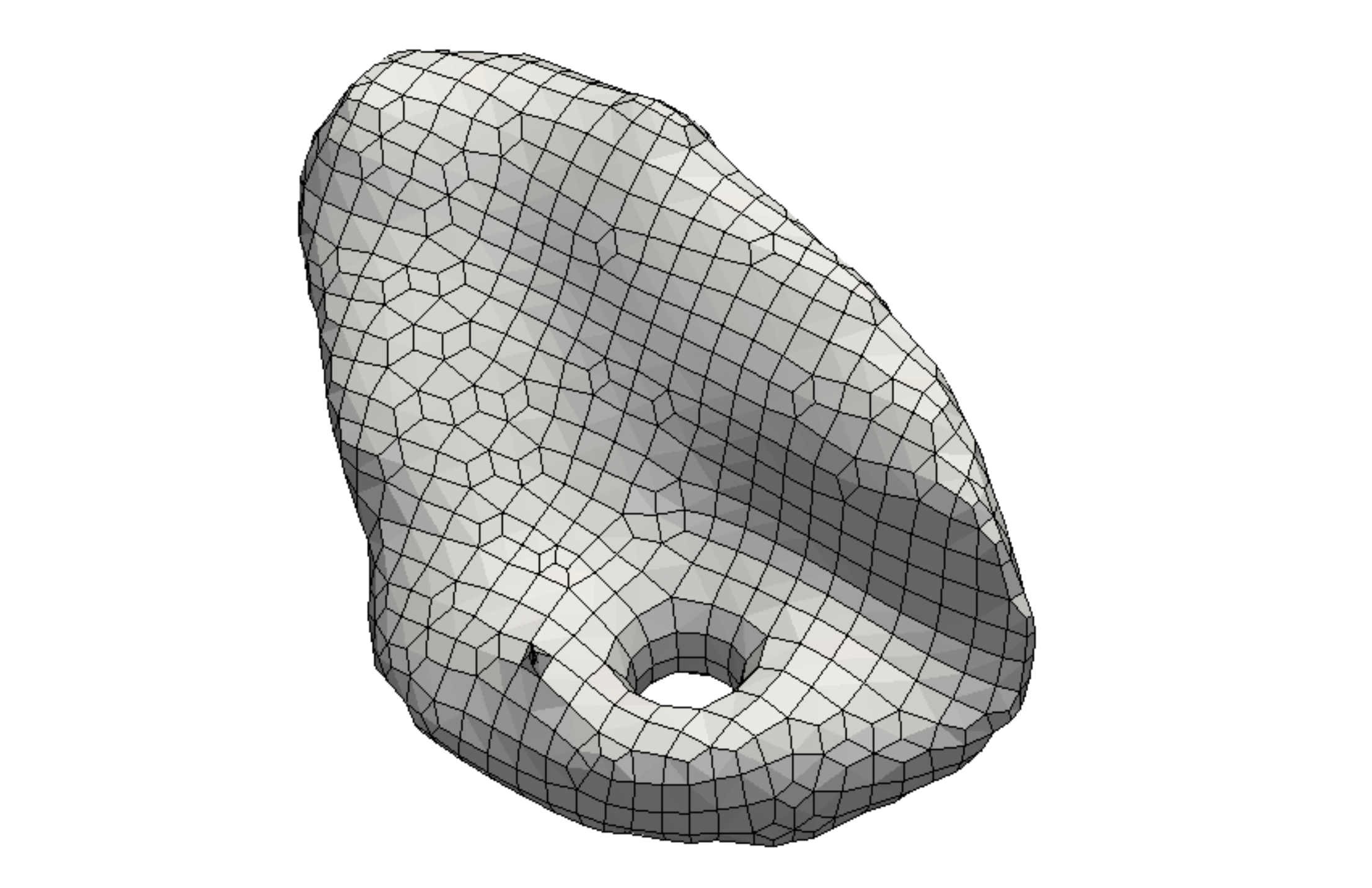}} & \parbox[m]{6em}{\includegraphics[trim={0cm 0cm 0cm 0cm},clip, width=0.12\textwidth]{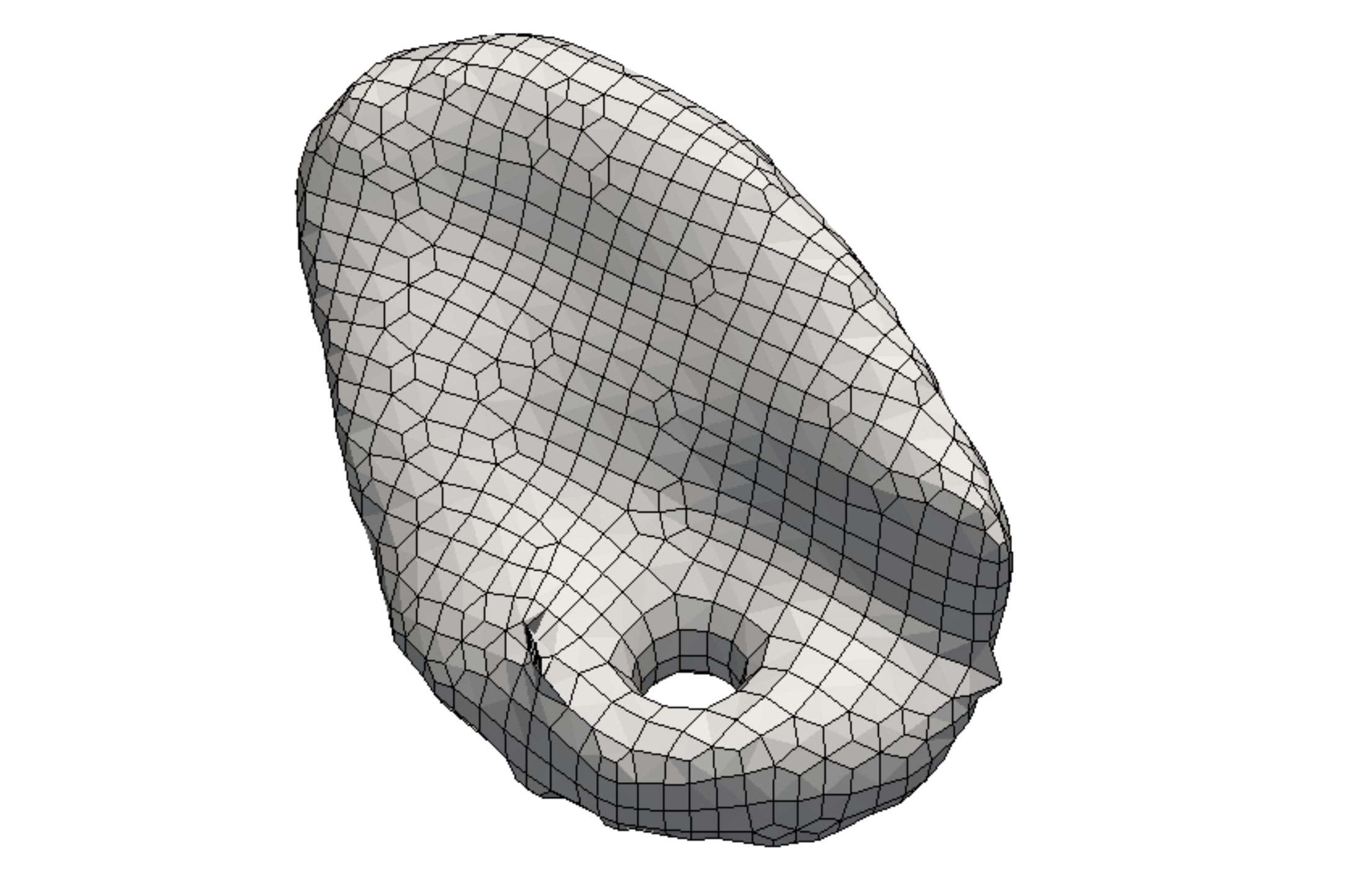}} & \parbox[m]{6em}{\includegraphics[trim={0cm 0cm 0cm 0cm},clip, width=0.12\textwidth]{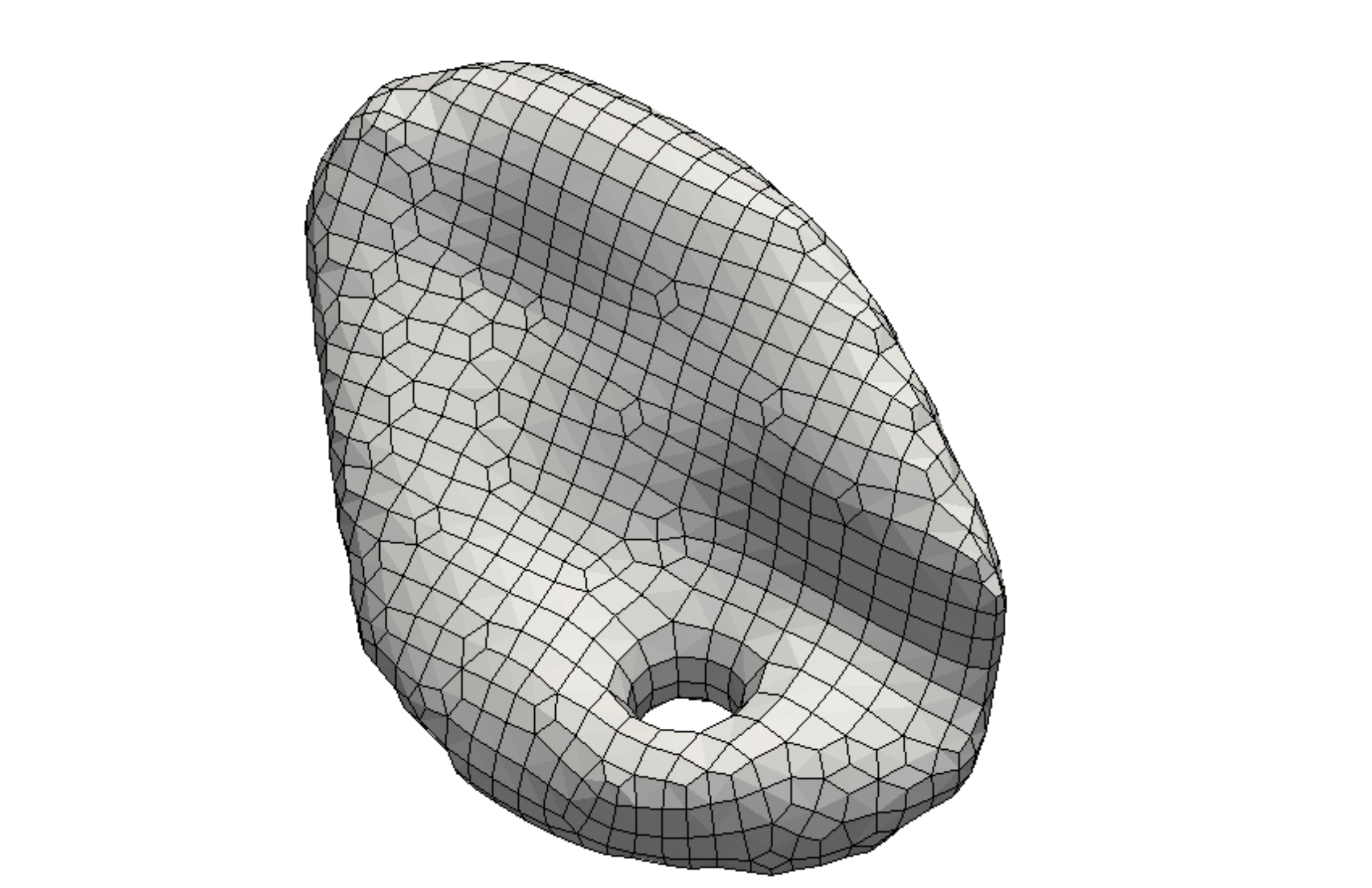}}\\
     \multicolumn{7}{c}{\fbox{\hspace{0.3cm}\parbox[m]{6em}{\includegraphics[trim={0cm 0cm 0cm 0cm},clip, width=0.12\textwidth]{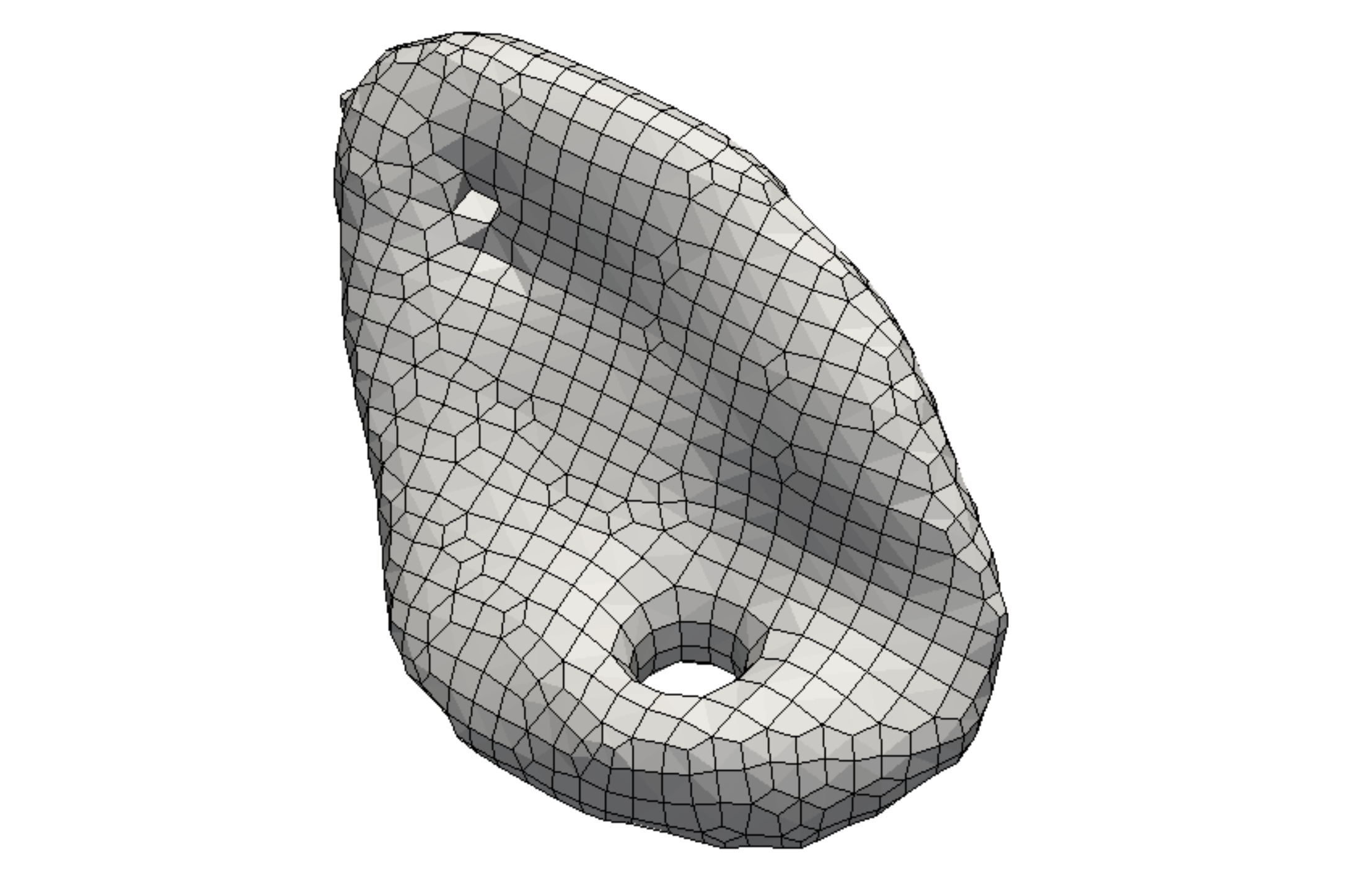}}}}
\end{tabular}
\end{subtable}
\begin{subtable}[h]{0.99\textwidth}
    \vspace{5mm}%
    \centering\setcellgapes{3pt}\makegapedcells
    \setlength\tabcolsep{3.5pt}
    \begin{tabular}{c|c||ScScScScSc}
    \multicolumn{2}{c||}{} & \multicolumn{5}{c}{training samples} \\\hline
     prepr. & equiv. & 10 & 50 & 100 & 500 & 1500 \\\hline
     \multirow{2}{*}{\rotatebox{90}{trivial}} & & \parbox[m]{6em}{\includegraphics[trim={0cm 0cm 0cm 0cm},clip, width=0.12\textwidth]{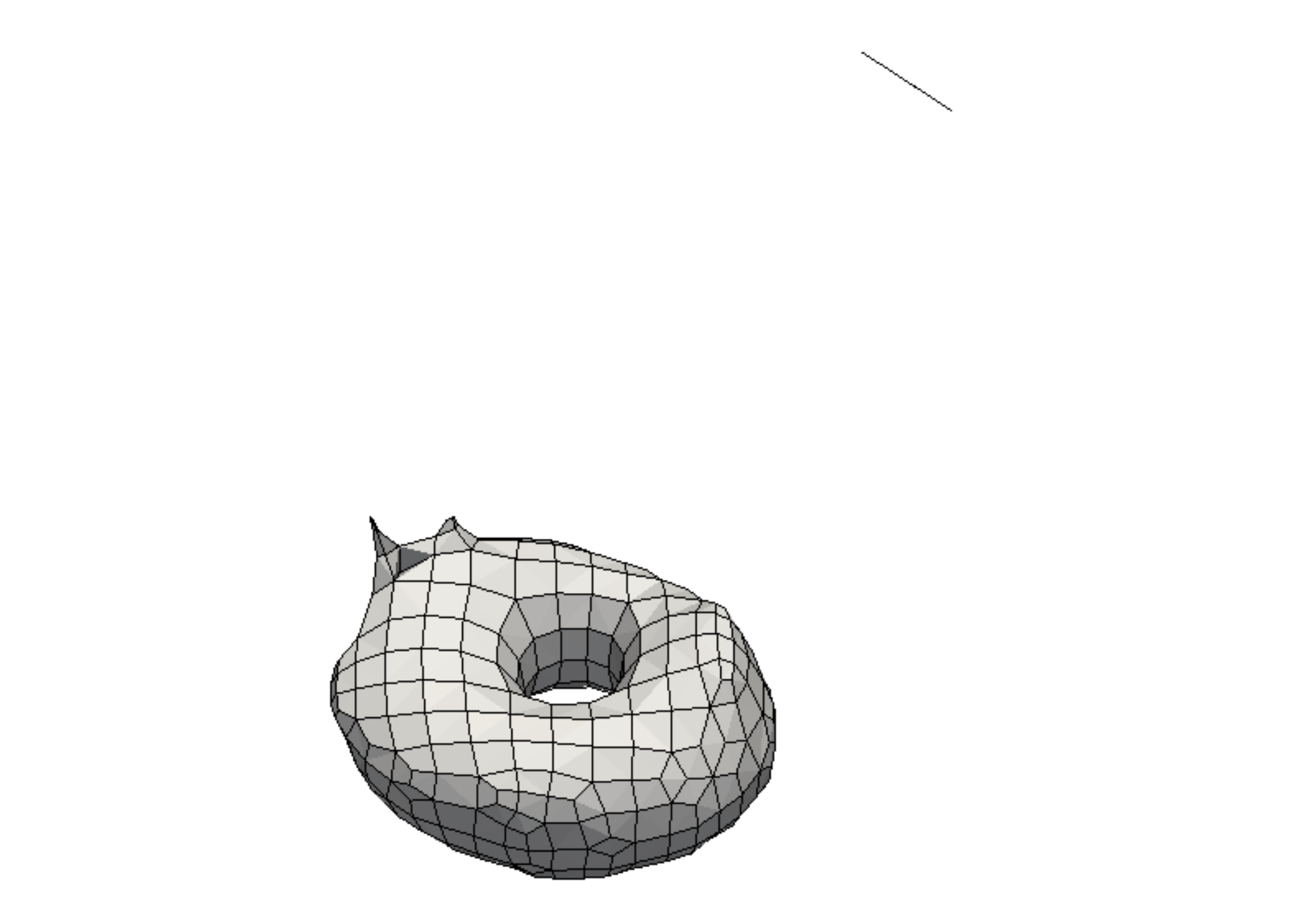}} & \parbox[m]{6em}{\includegraphics[trim={0cm 0cm 0cm 0cm},clip, width=0.12\textwidth]{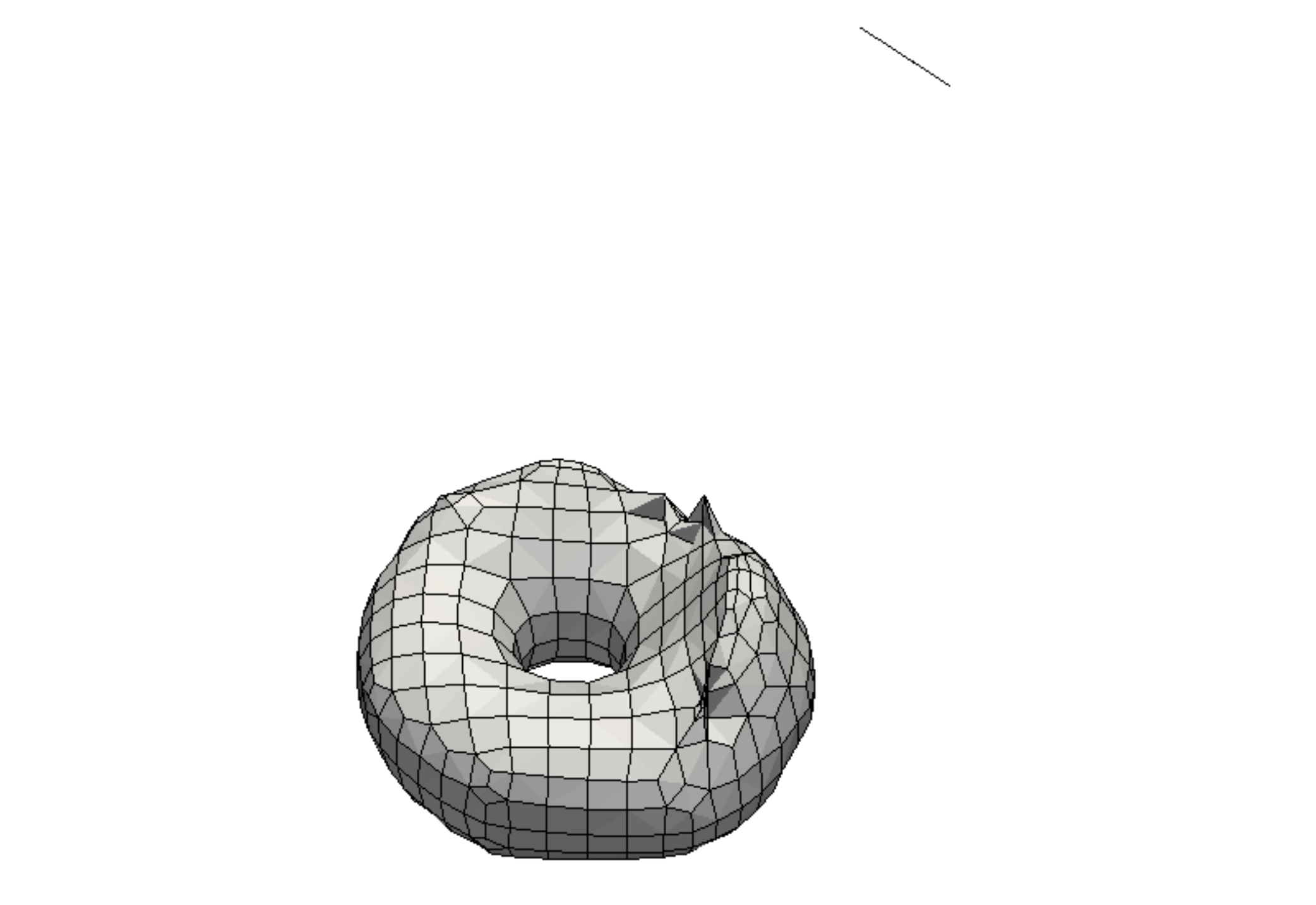}} & \parbox[m]{6em}{\includegraphics[trim={0cm 0cm 0cm 0cm},clip, width=0.12\textwidth]{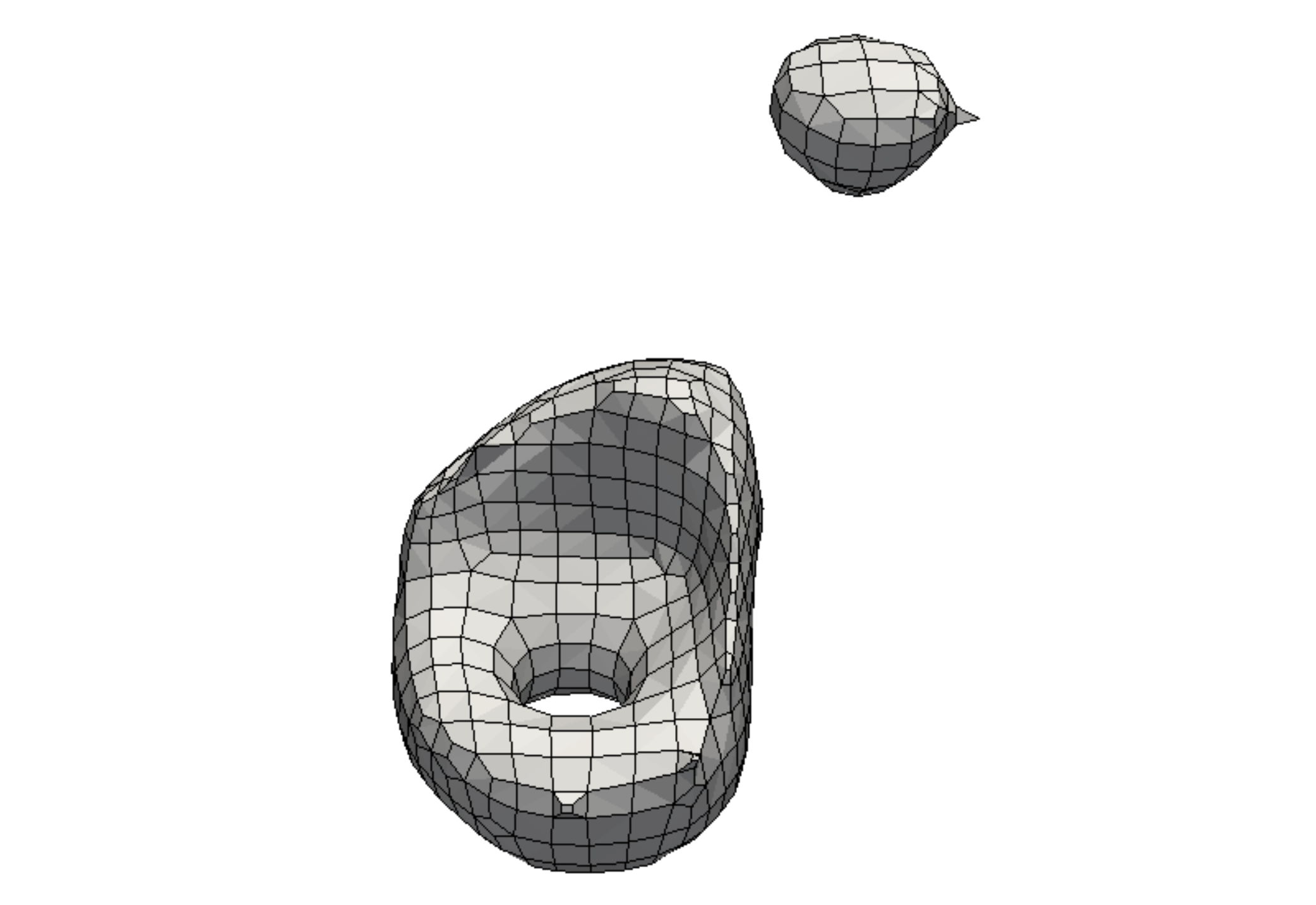}} & \parbox[m]{6em}{\includegraphics[trim={0cm 0cm 0cm 0cm},clip, width=0.12\textwidth]{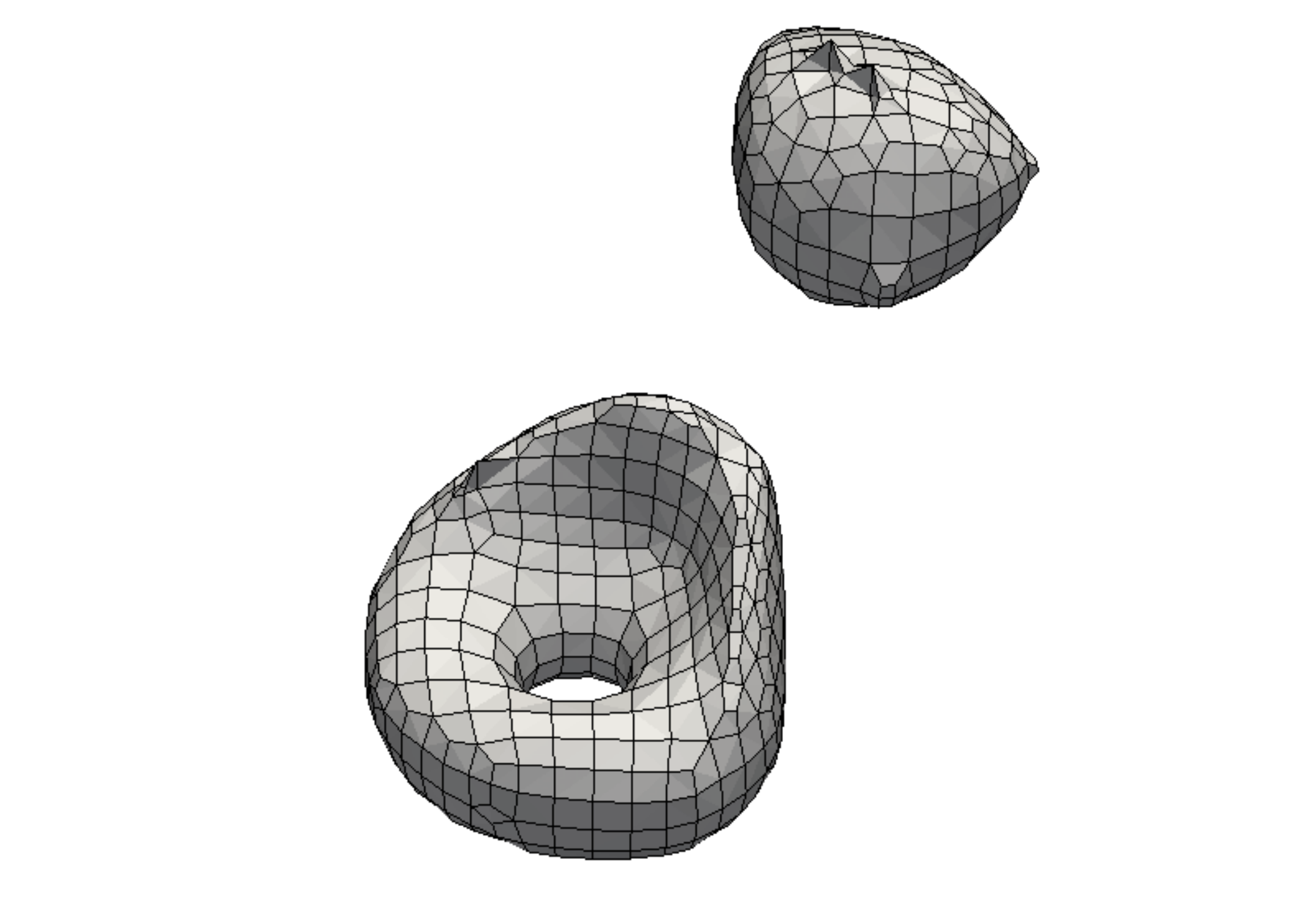}} & \parbox[m]{6em}{\includegraphics[trim={0cm 0cm 0cm 0cm},clip, width=0.12\textwidth]{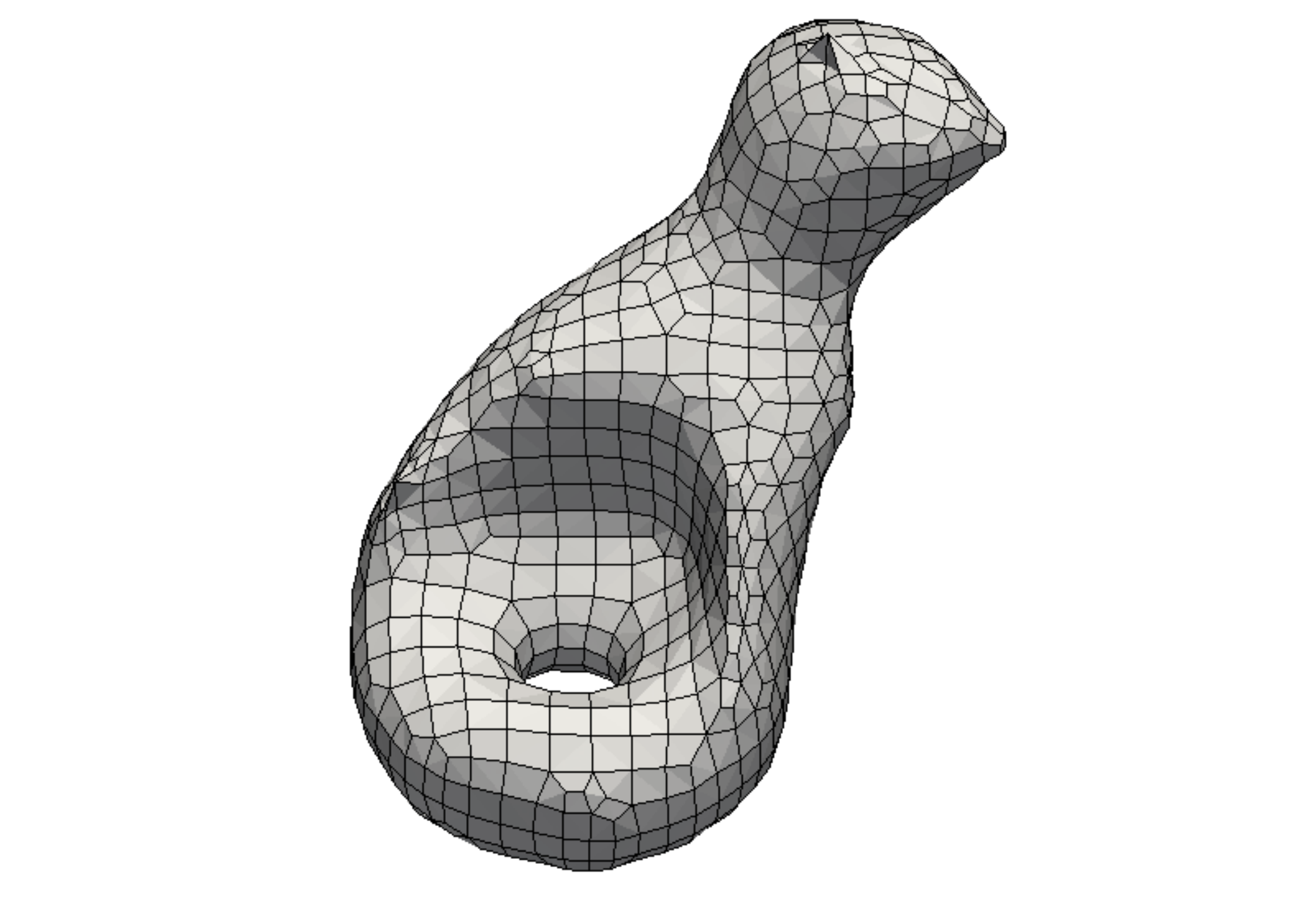}} \\\cline{2-7}
     & \checkmark
     & \parbox[m]{6em}{\includegraphics[trim={0cm 0cm 0cm 0cm},clip, width=0.12\textwidth]{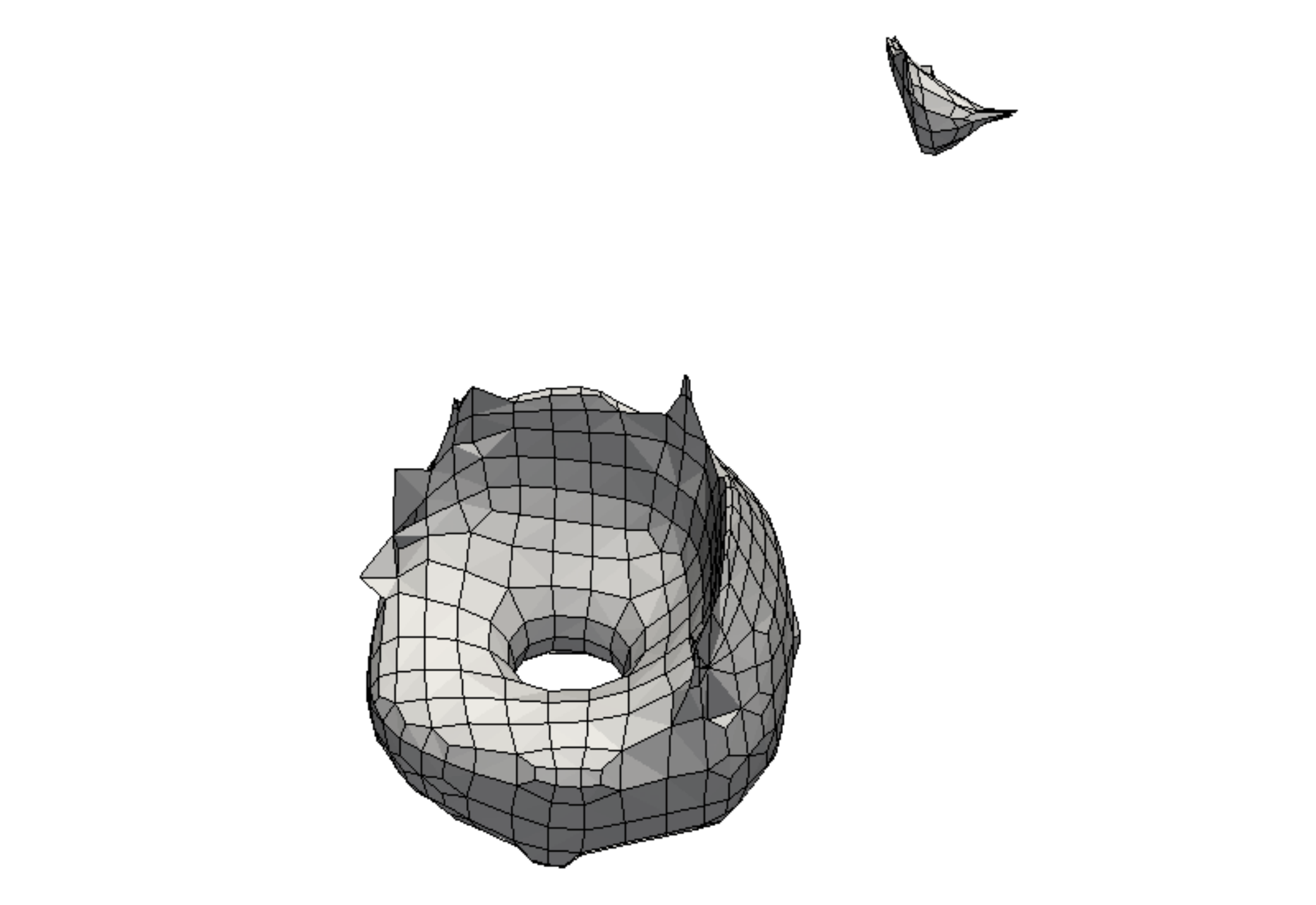}} & \parbox[m]{6em}{\includegraphics[trim={0cm 0cm 0cm 0cm},clip, width=0.12\textwidth]{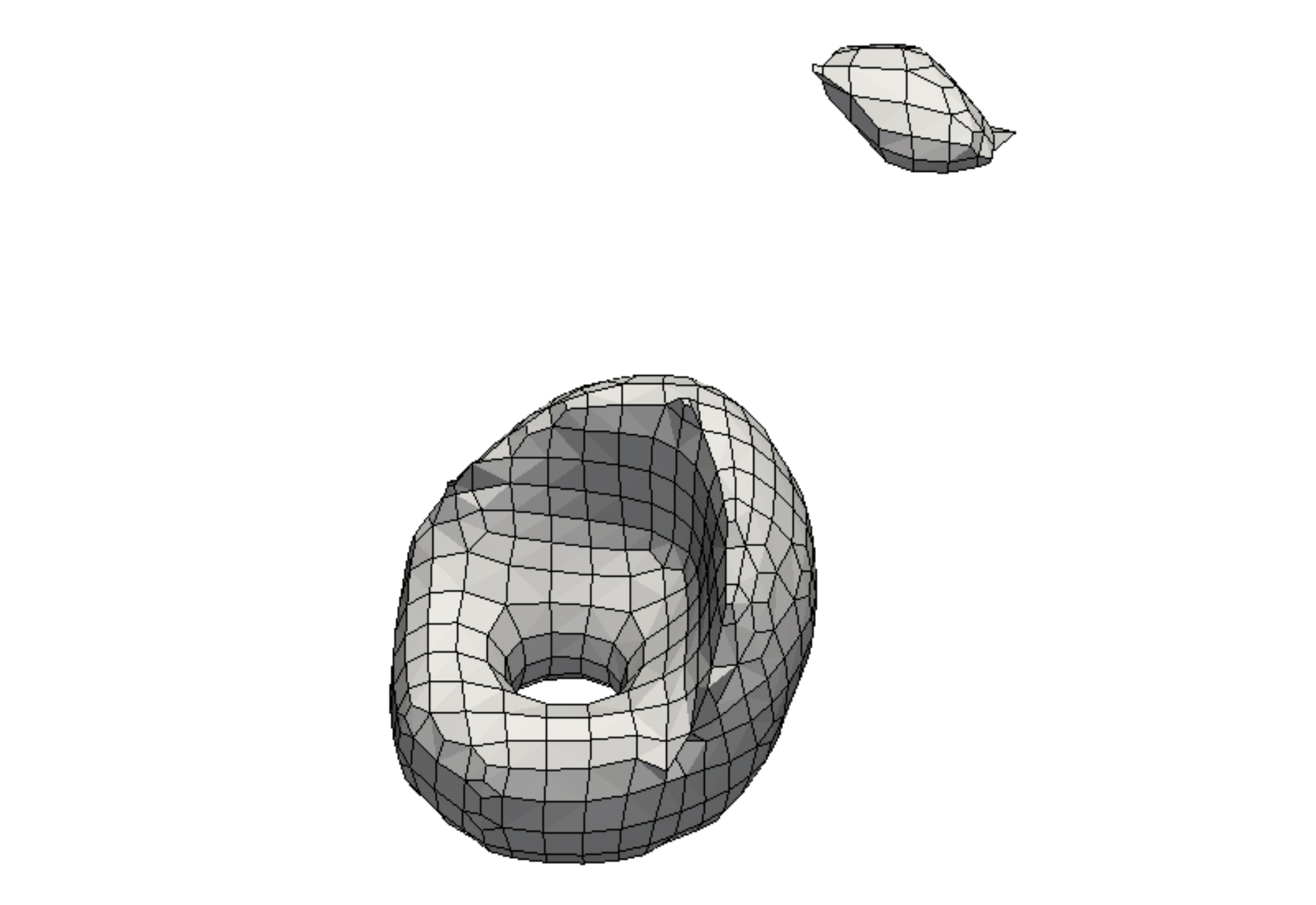}} & \parbox[m]{6em}{\includegraphics[trim={0cm 0cm 0cm 0cm},clip, width=0.12\textwidth]{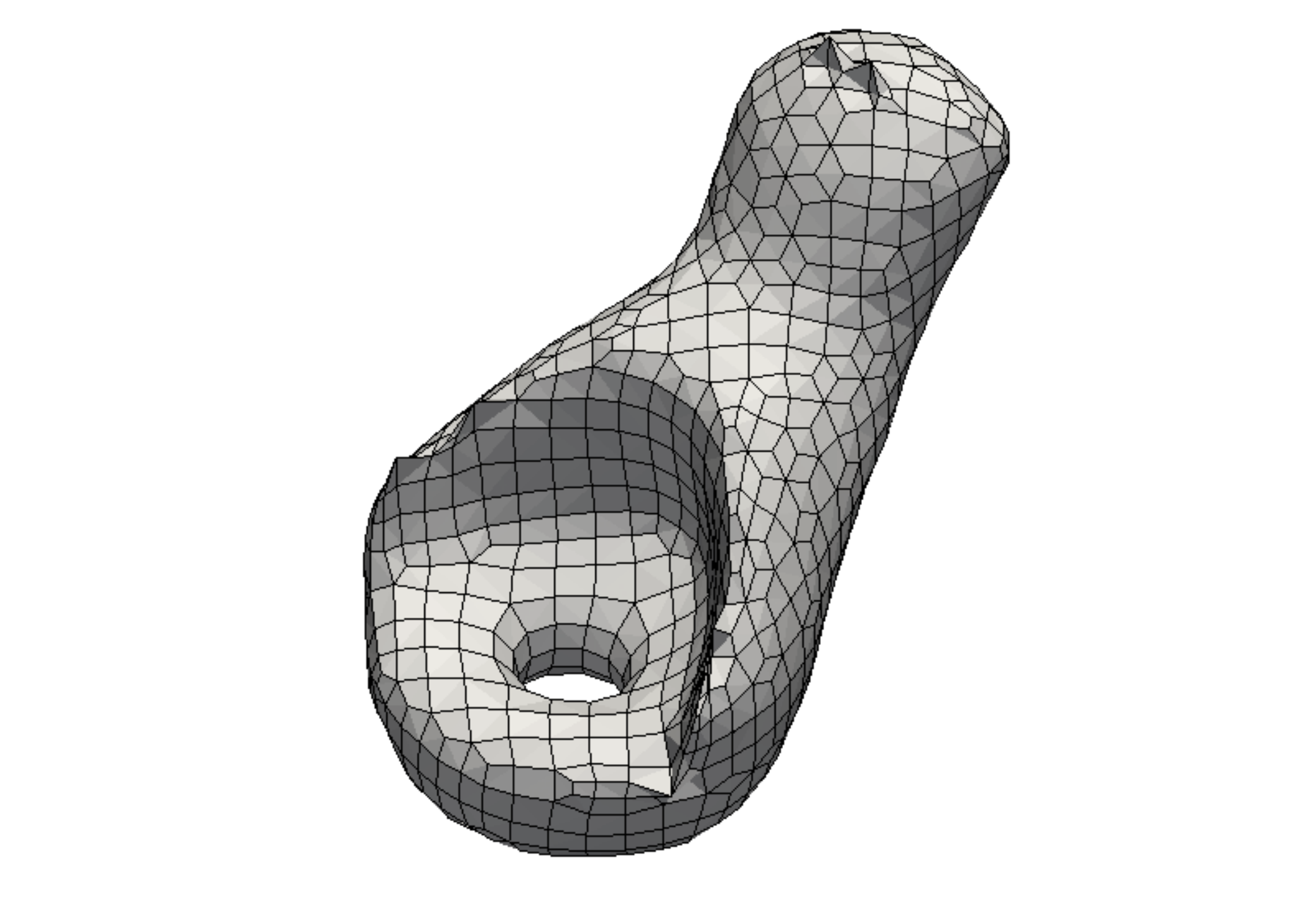}} & \parbox[m]{6em}{\includegraphics[trim={0cm 0cm 0cm 0cm},clip, width=0.12\textwidth]{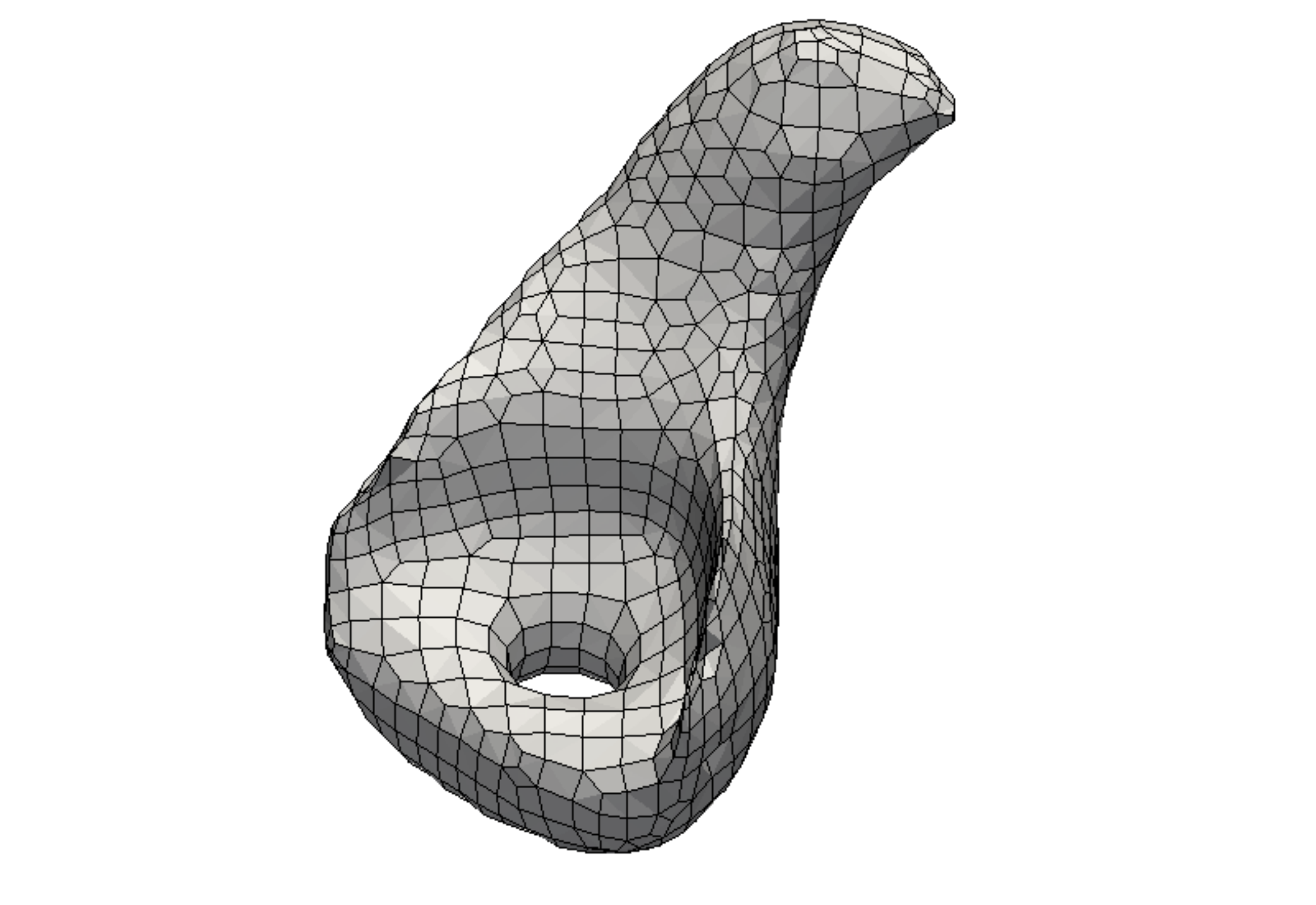}} & \parbox[m]{6em}{\includegraphics[trim={0cm 0cm 0cm 0cm},clip, width=0.12\textwidth]{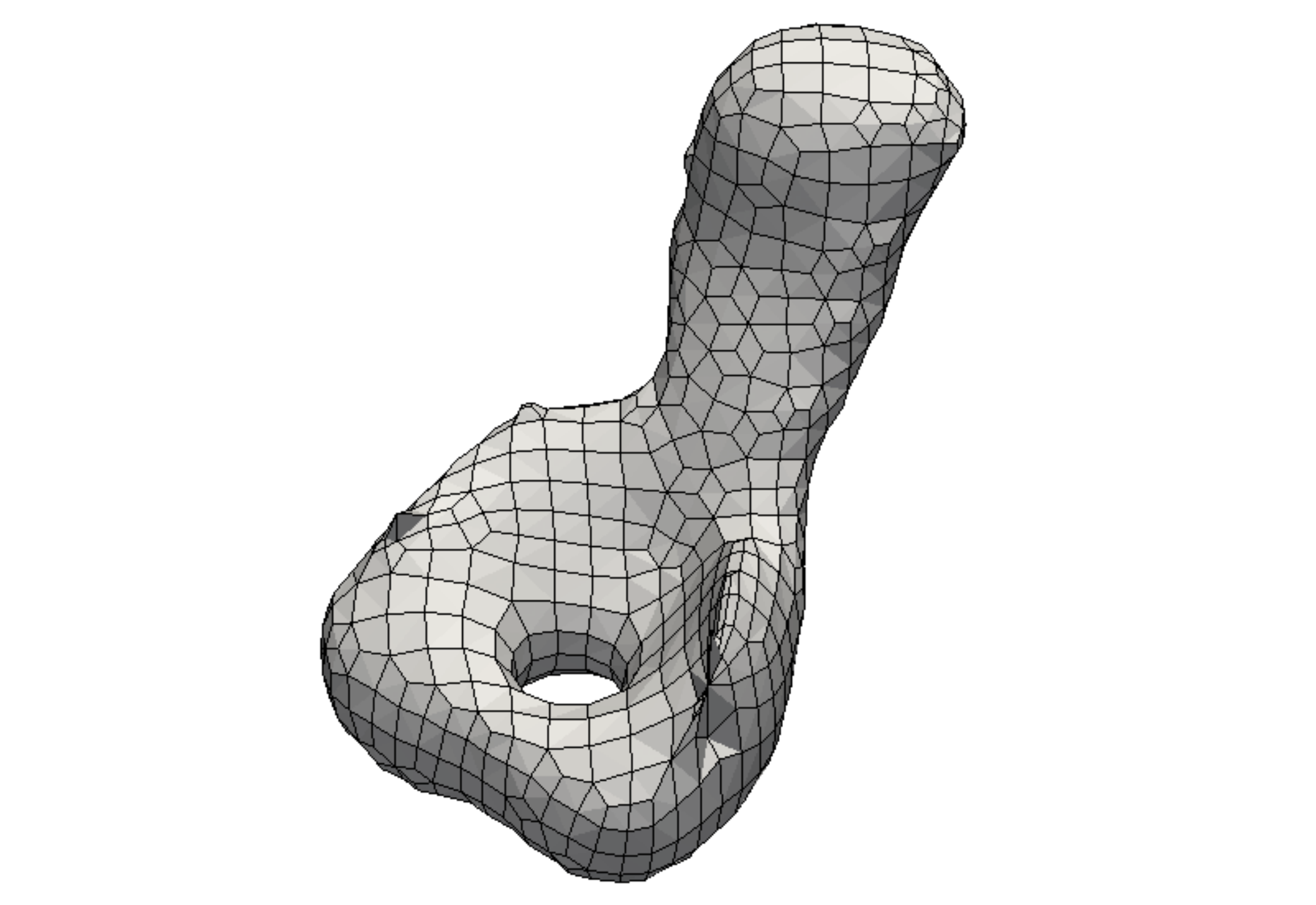}} \\\hline
     \multirow{2}{*}{\rotatebox{90}{trivial+PDE}} & & \parbox[m]{6em}{\includegraphics[trim={0cm 0cm 0cm 0cm},clip, width=0.12\textwidth]{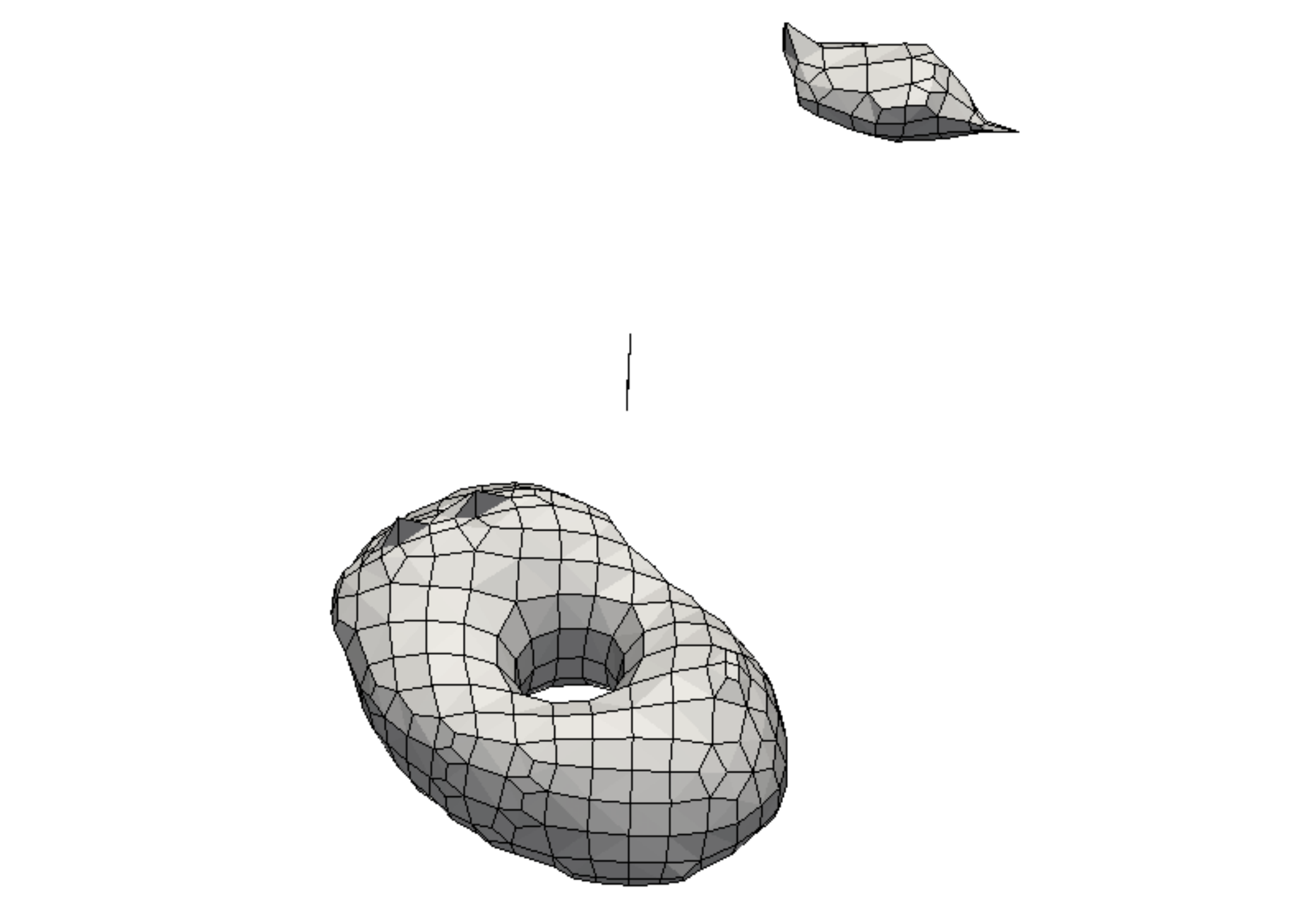}} & \parbox[m]{6em}{\includegraphics[trim={0cm 0cm 0cm 0cm},clip, width=0.12\textwidth]{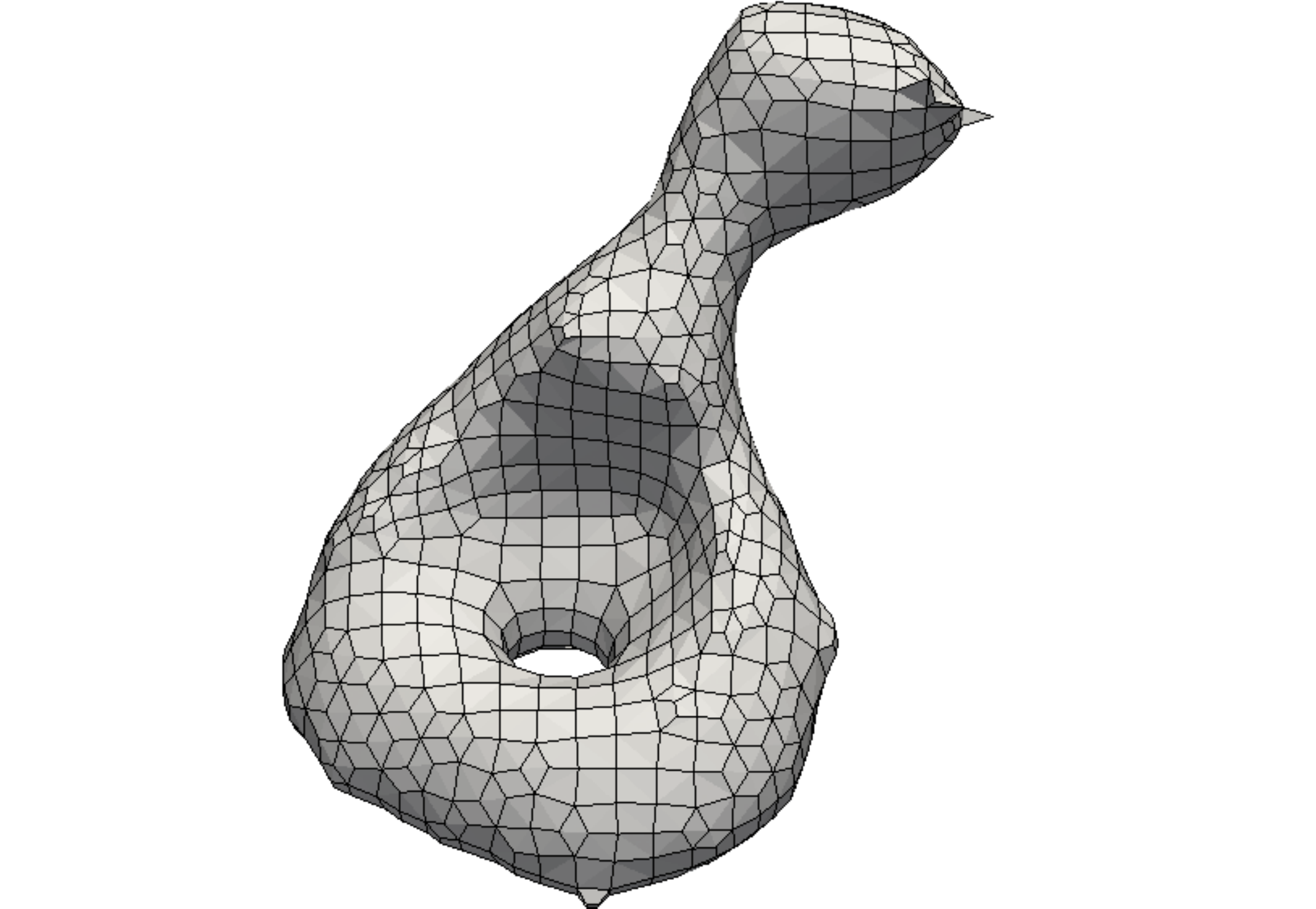}} & \parbox[m]{6em}{\includegraphics[trim={0cm 0cm 0cm 0cm},clip, width=0.12\textwidth]{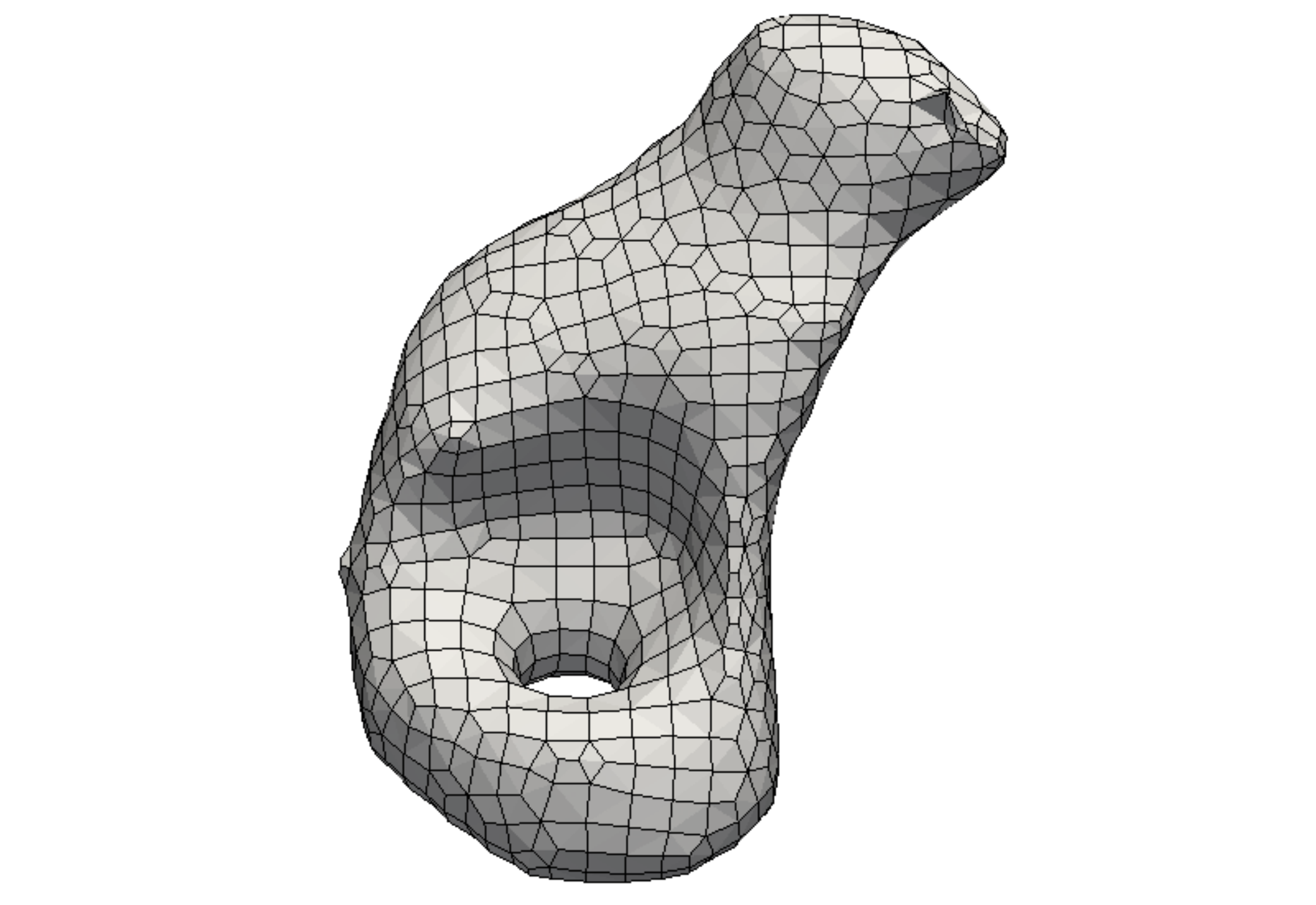}} & \parbox[m]{6em}{\includegraphics[trim={0cm 0cm 0cm 0cm},clip, width=0.12\textwidth]{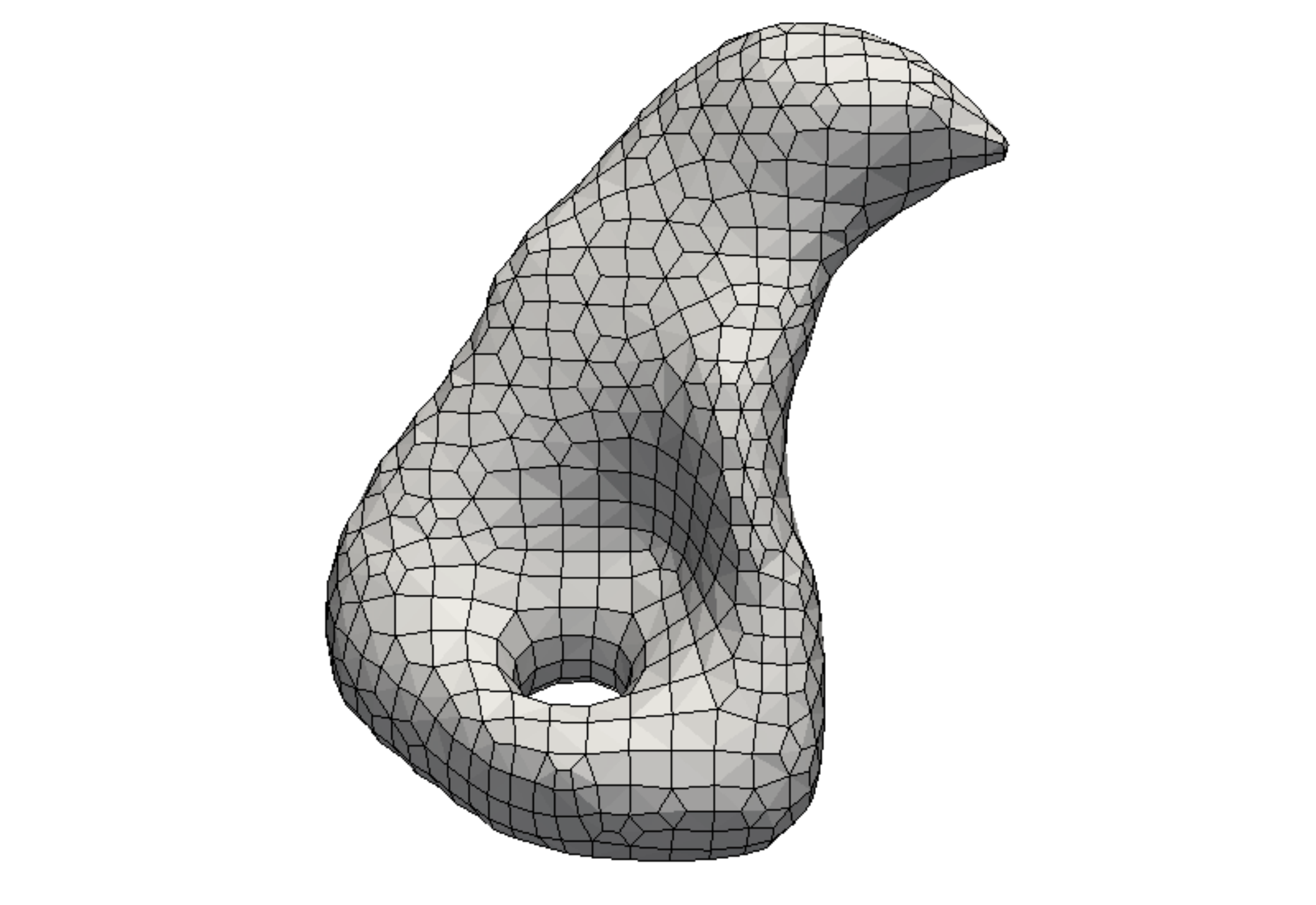}} & \parbox[m]{6em}{\includegraphics[trim={0cm 0cm 0cm 0cm},clip, width=0.12\textwidth]{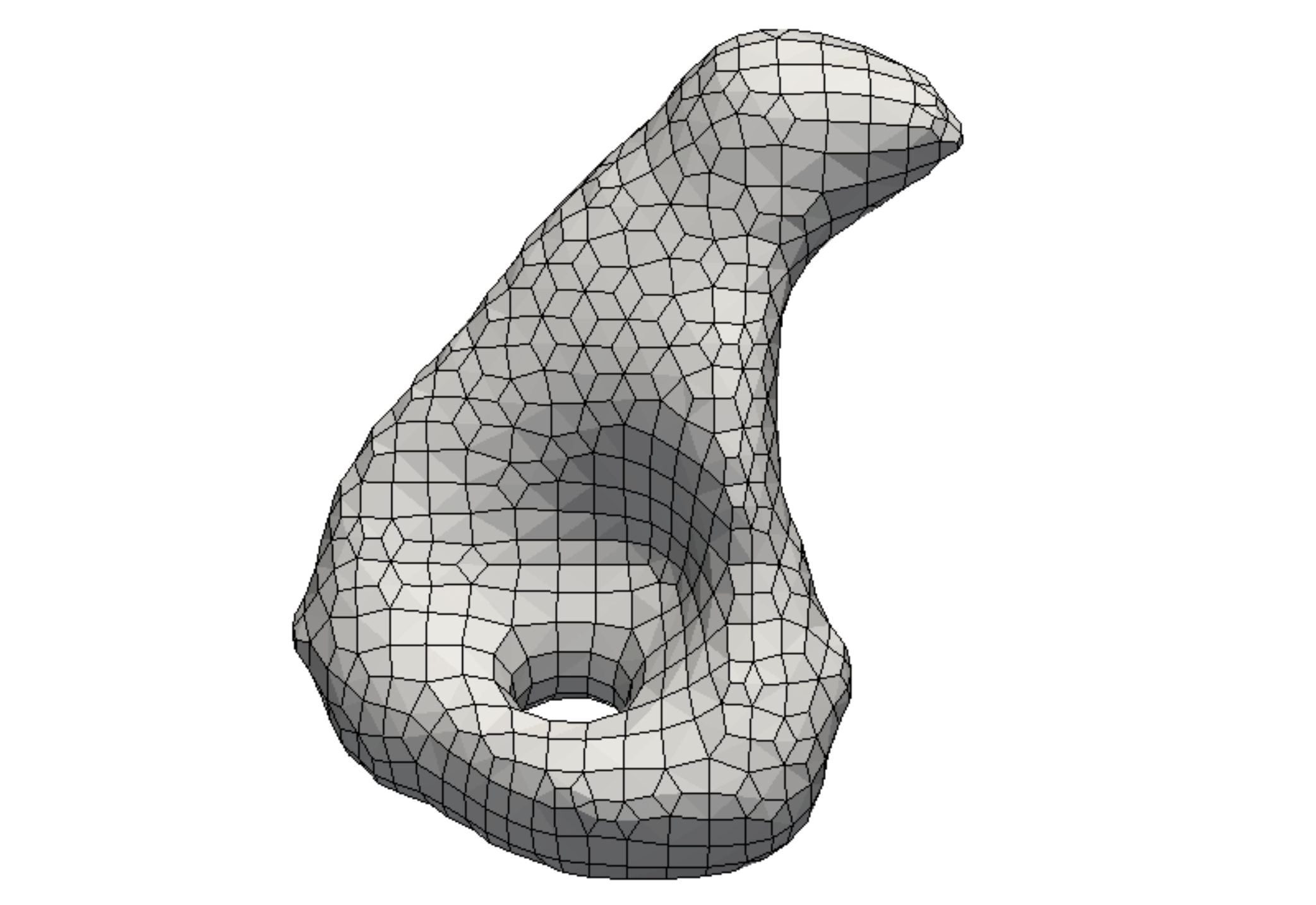}} \\\cline{2-7}
     & \checkmark & \parbox[m]{6em}{\includegraphics[trim={0cm 0cm 0cm 0cm},clip, width=0.12\textwidth]{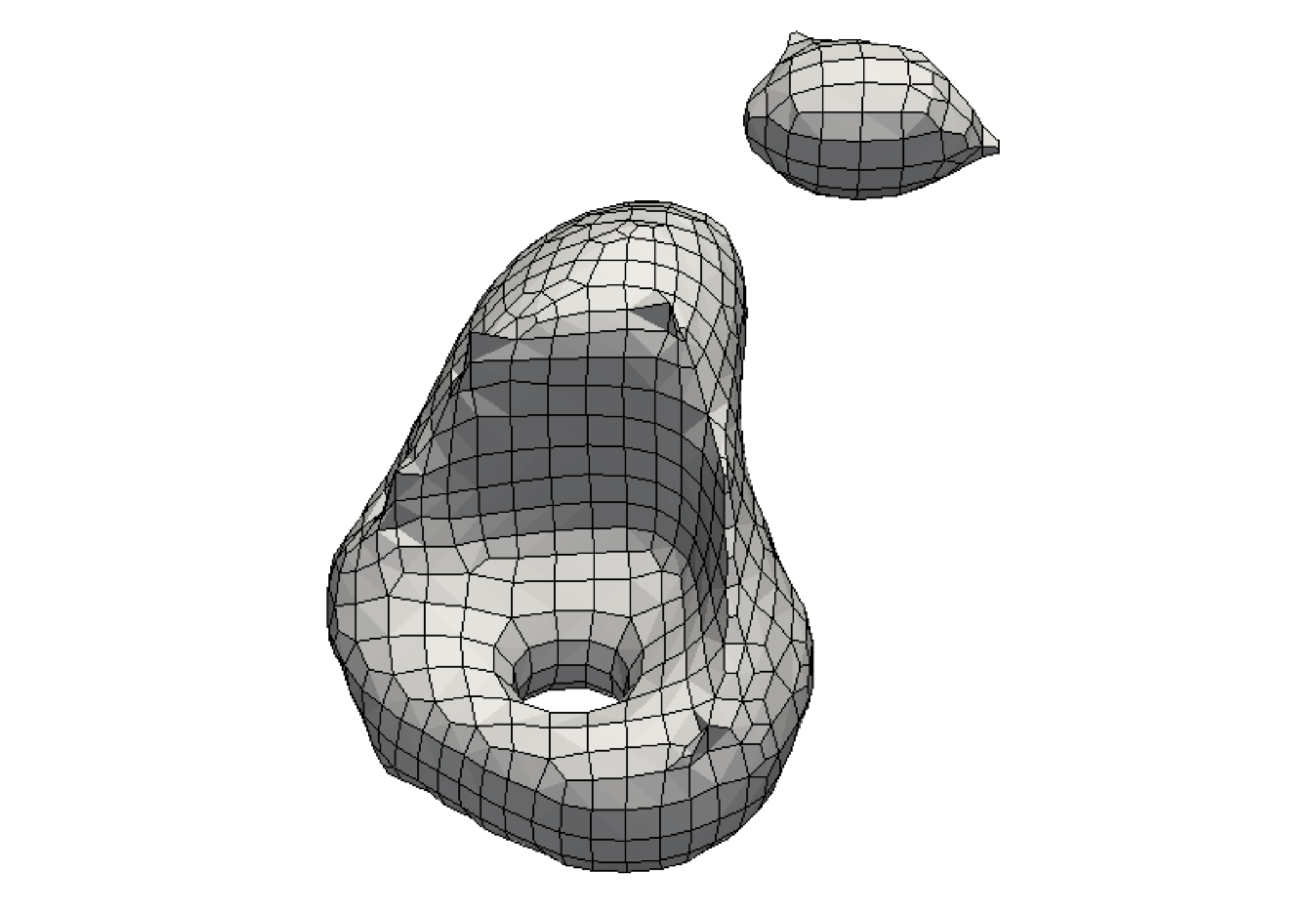}} & \parbox[m]{6em}{\includegraphics[trim={0cm 0cm 0cm 0cm},clip, width=0.12\textwidth]{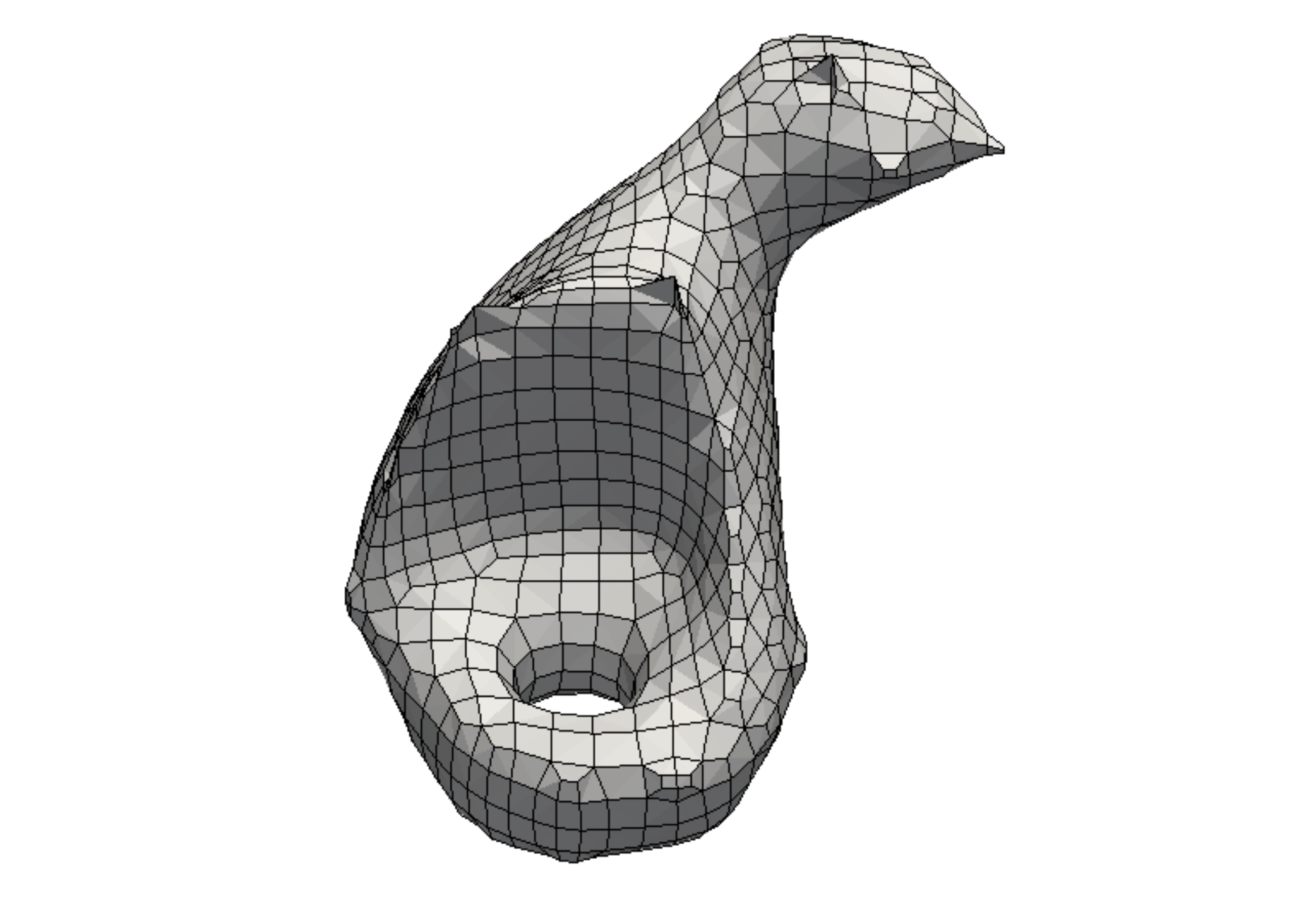}} & \parbox[m]{6em}{\includegraphics[trim={0cm 0cm 0cm 0cm},clip, width=0.12\textwidth]{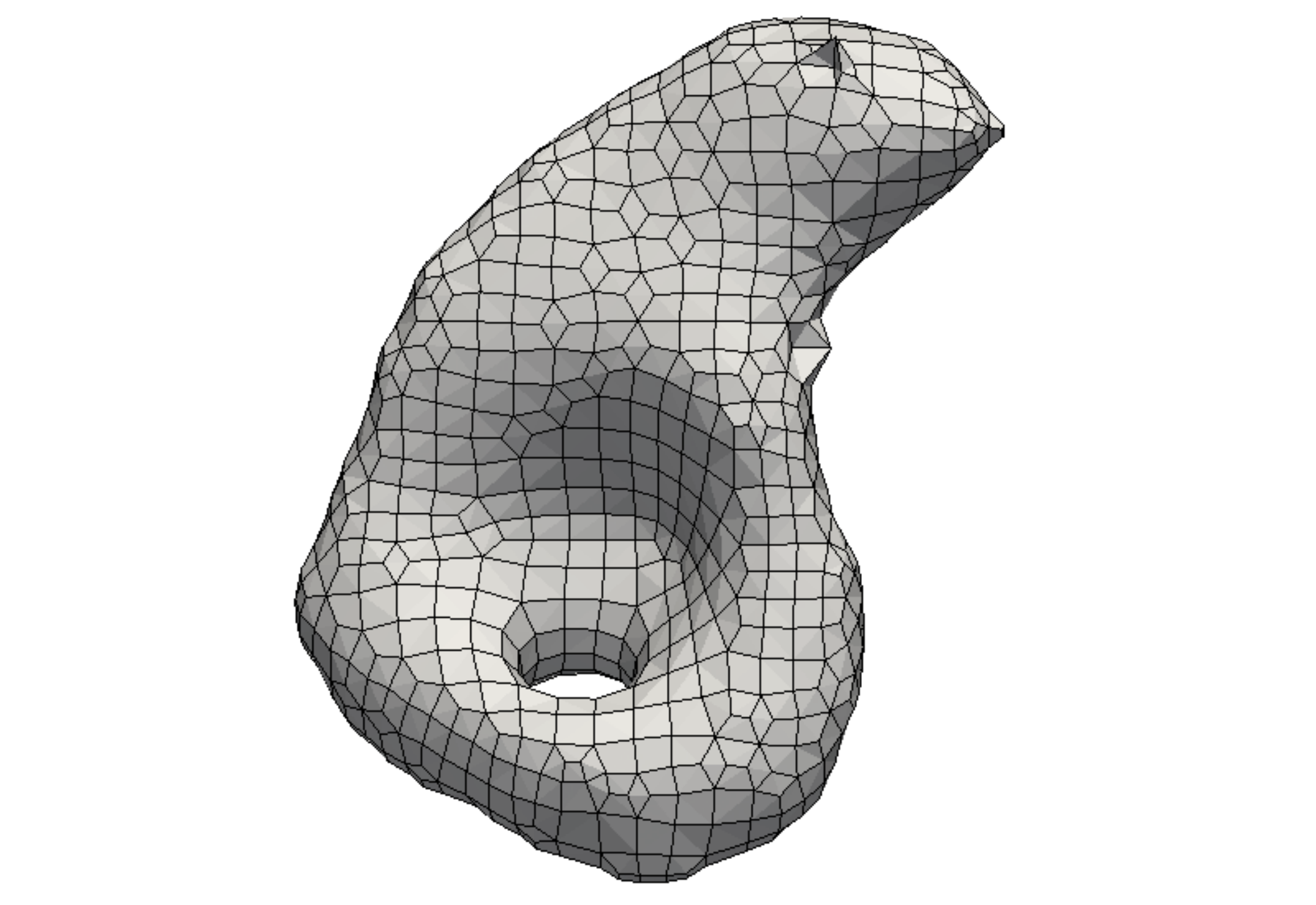}} & \parbox[m]{6em}{\includegraphics[trim={0cm 0cm 0cm 0cm},clip, width=0.12\textwidth]{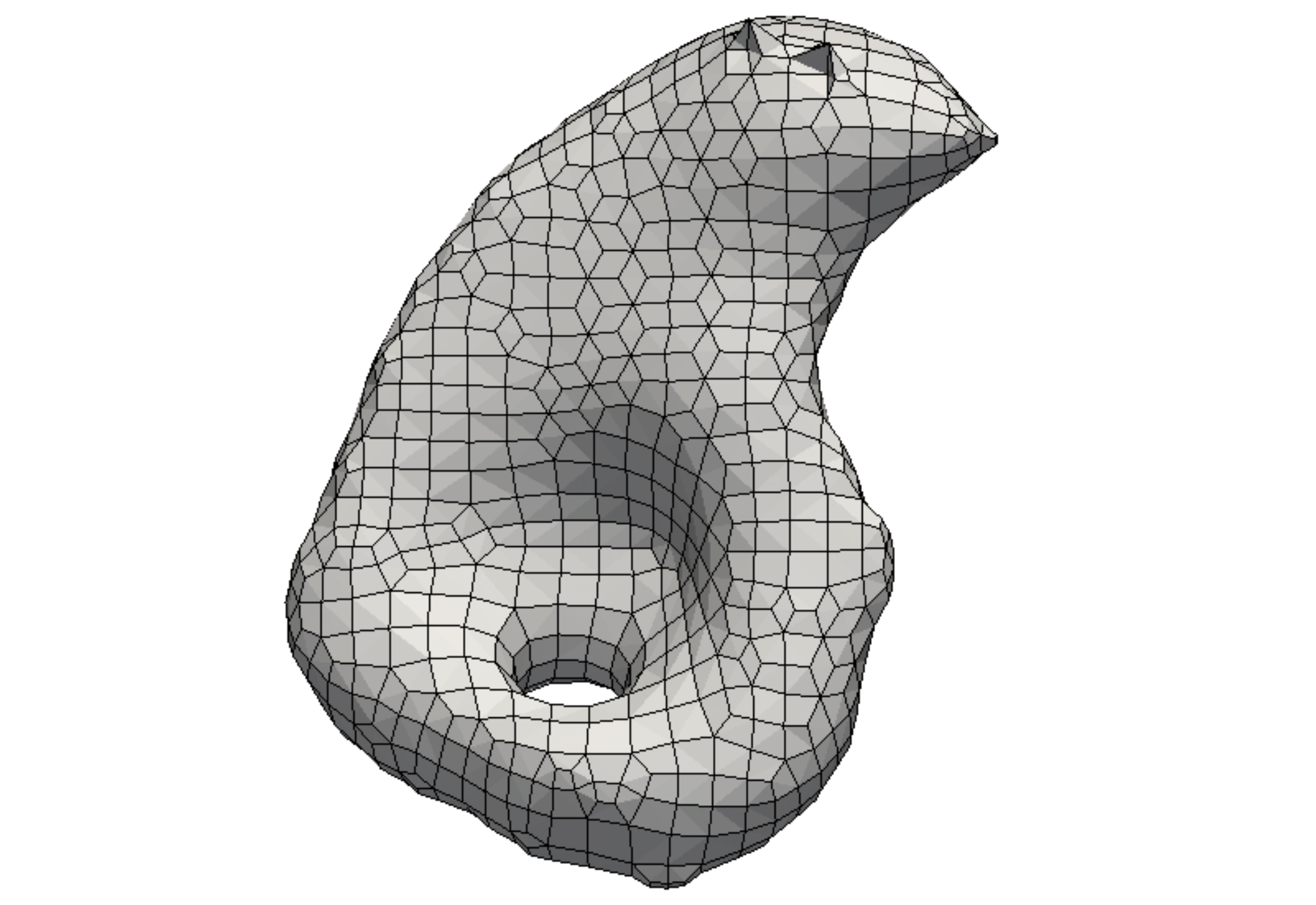}} & \parbox[m]{6em}{\includegraphics[trim={0cm 0cm 0cm 0cm},clip, width=0.12\textwidth]{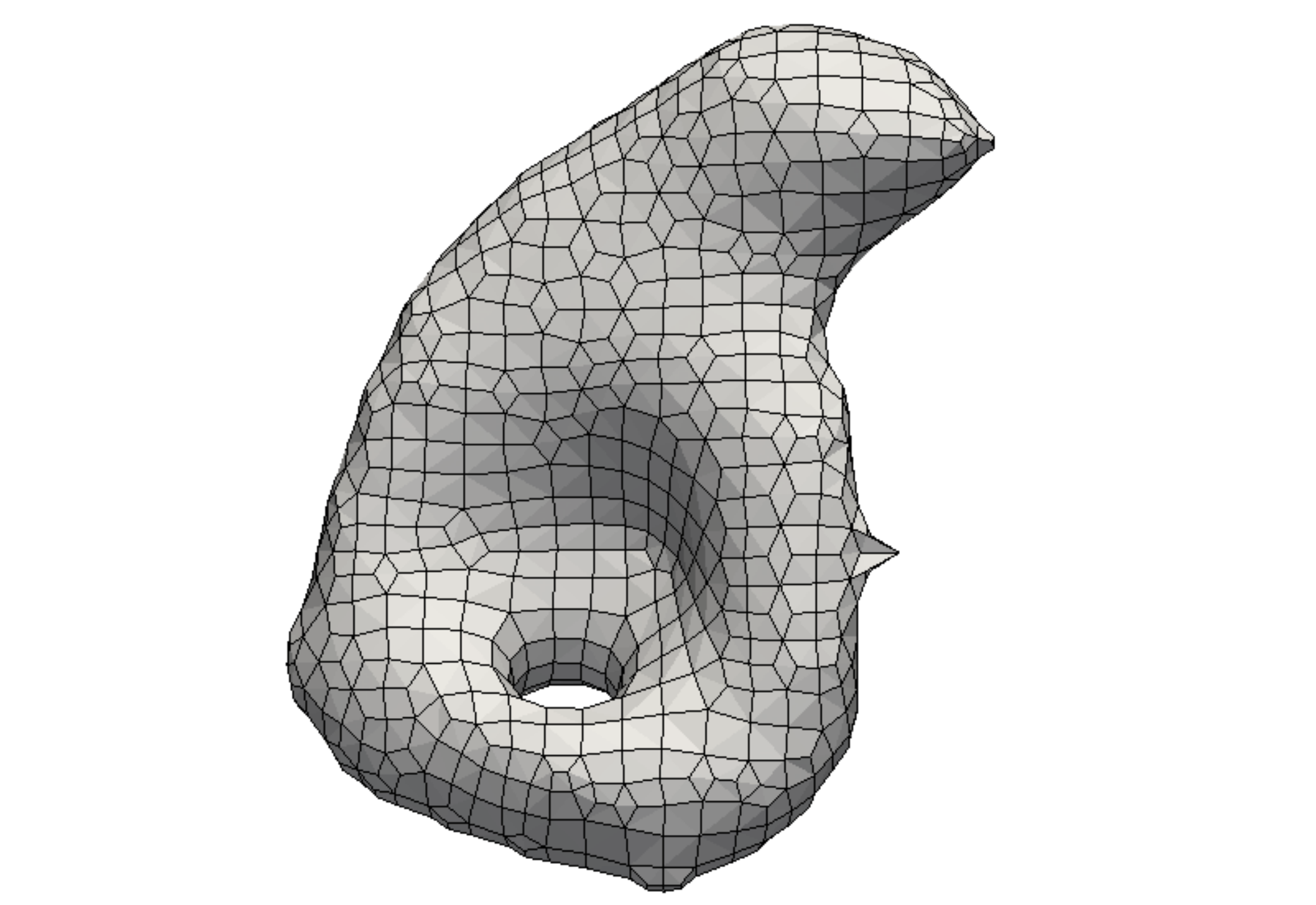}}\\
     \multicolumn{7}{c}{\fbox{\hspace{0.3cm}\parbox[m]{6em}{\includegraphics[trim={0cm 0cm 0cm 0cm},clip, width=0.12\textwidth]{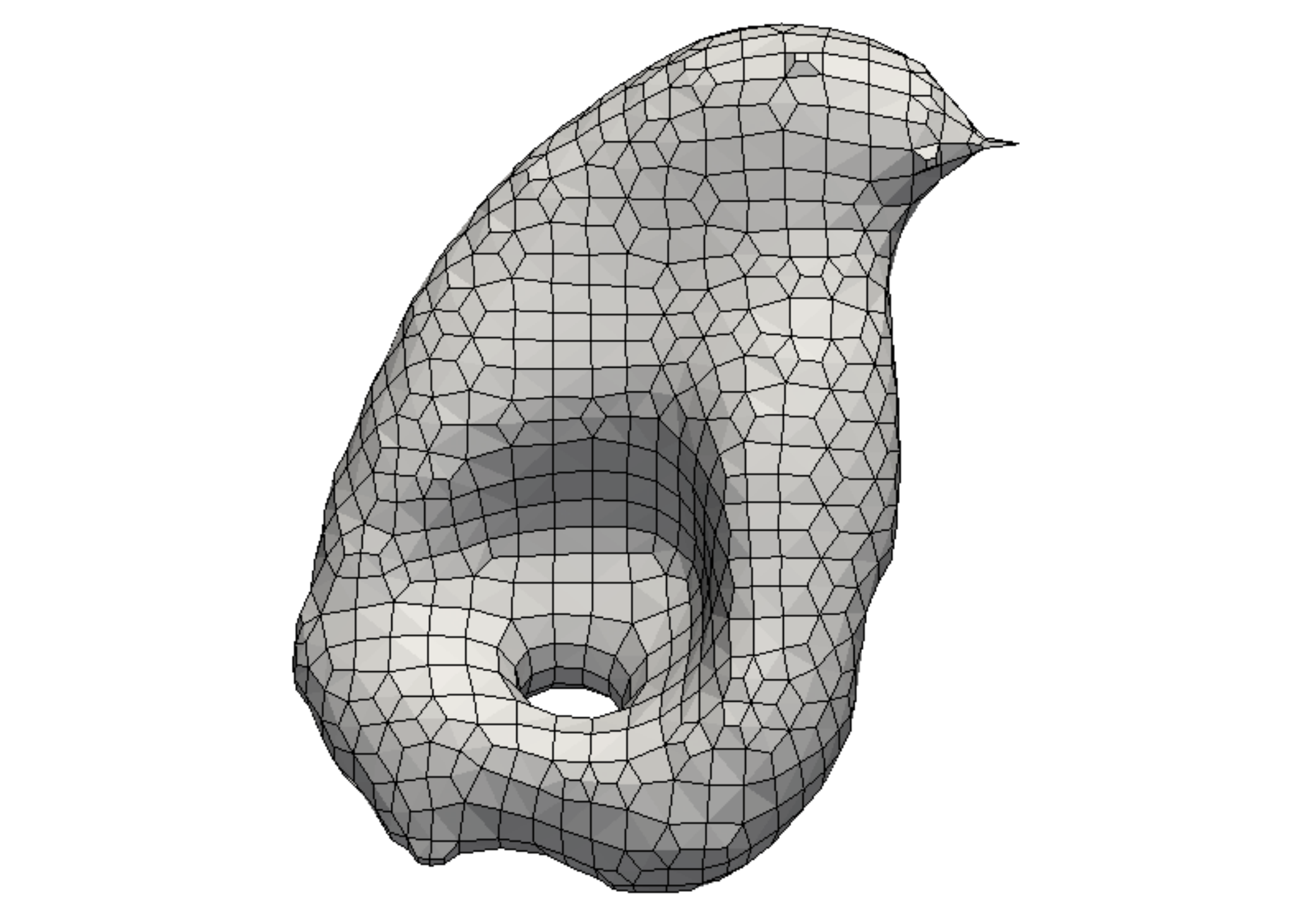}}}}
\end{tabular}
\end{subtable}
\end{table}

\begin{table}[]
\caption{Model predictions of two different problems from the \textbf{sphere complex} validation dataset, using the UNet with different preprocessings and equivariances. We train the models on subsets of the dataset and vary the training size along the columns of the table. At the boxes below the tables we show the corresponding ground truth density for each problem.}
\label{5_fig:sphere_complex_predictions}
\begin{subtable}[h]{0.99\textwidth}
    \centering\setcellgapes{3pt}\makegapedcells
    \setlength\tabcolsep{3.5pt}
    \begin{tabular}{c|c||ScScScScSc}
    \multicolumn{2}{c||}{} & \multicolumn{5}{c}{training samples} \\\hline
     prepr. & equiv. & 10 & 50 & 100 & 500 & 1500 \\\hline
     \multirow{2}{*}{\rotatebox{90}{trivial}} & & \parbox[m]{6em}{\includegraphics[trim={0cm 0cm 0cm 0cm},clip, width=0.12\textwidth]{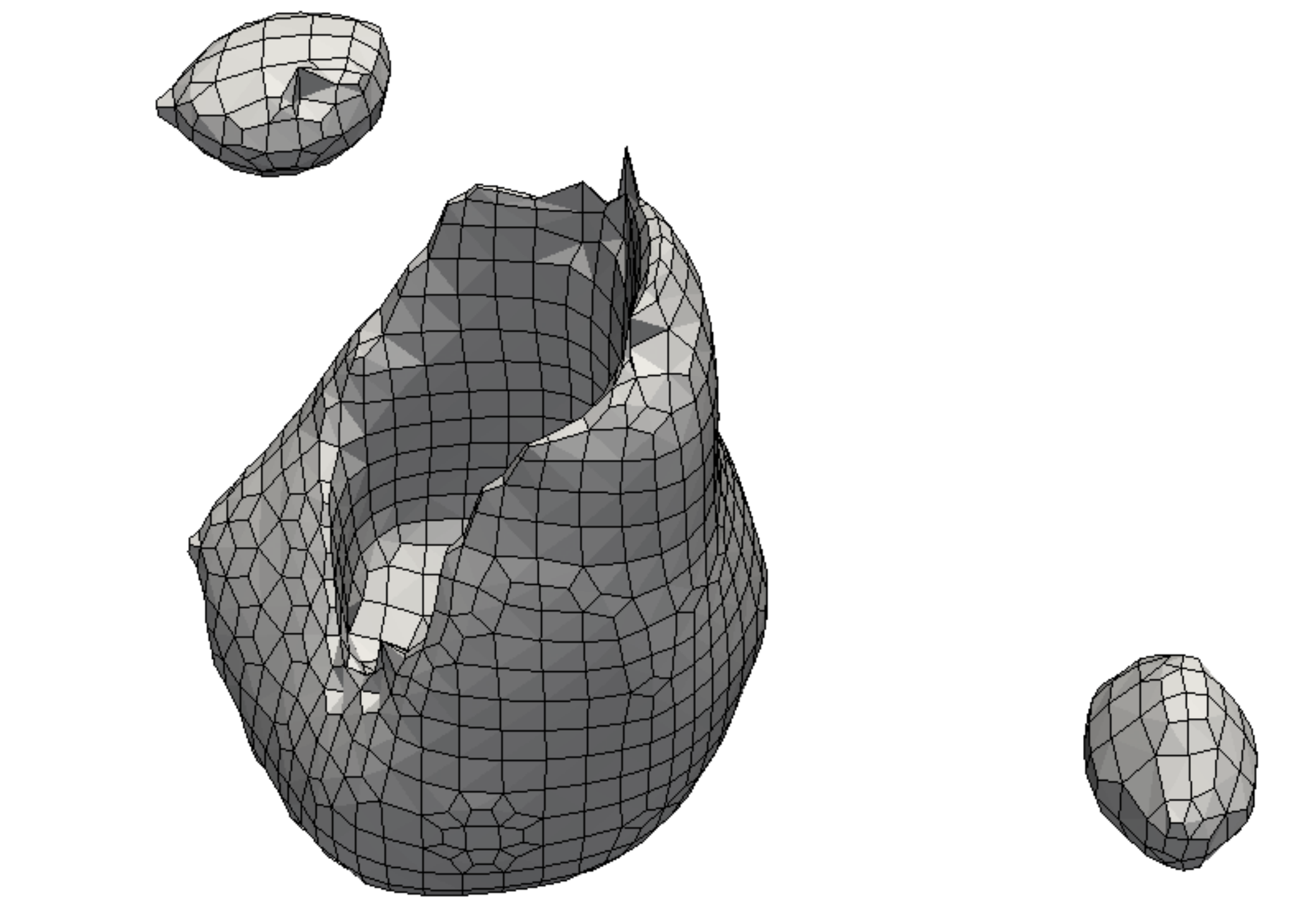}} & \parbox[m]{6em}{\includegraphics[trim={0cm 0cm 0cm 0cm},clip, width=0.12\textwidth]{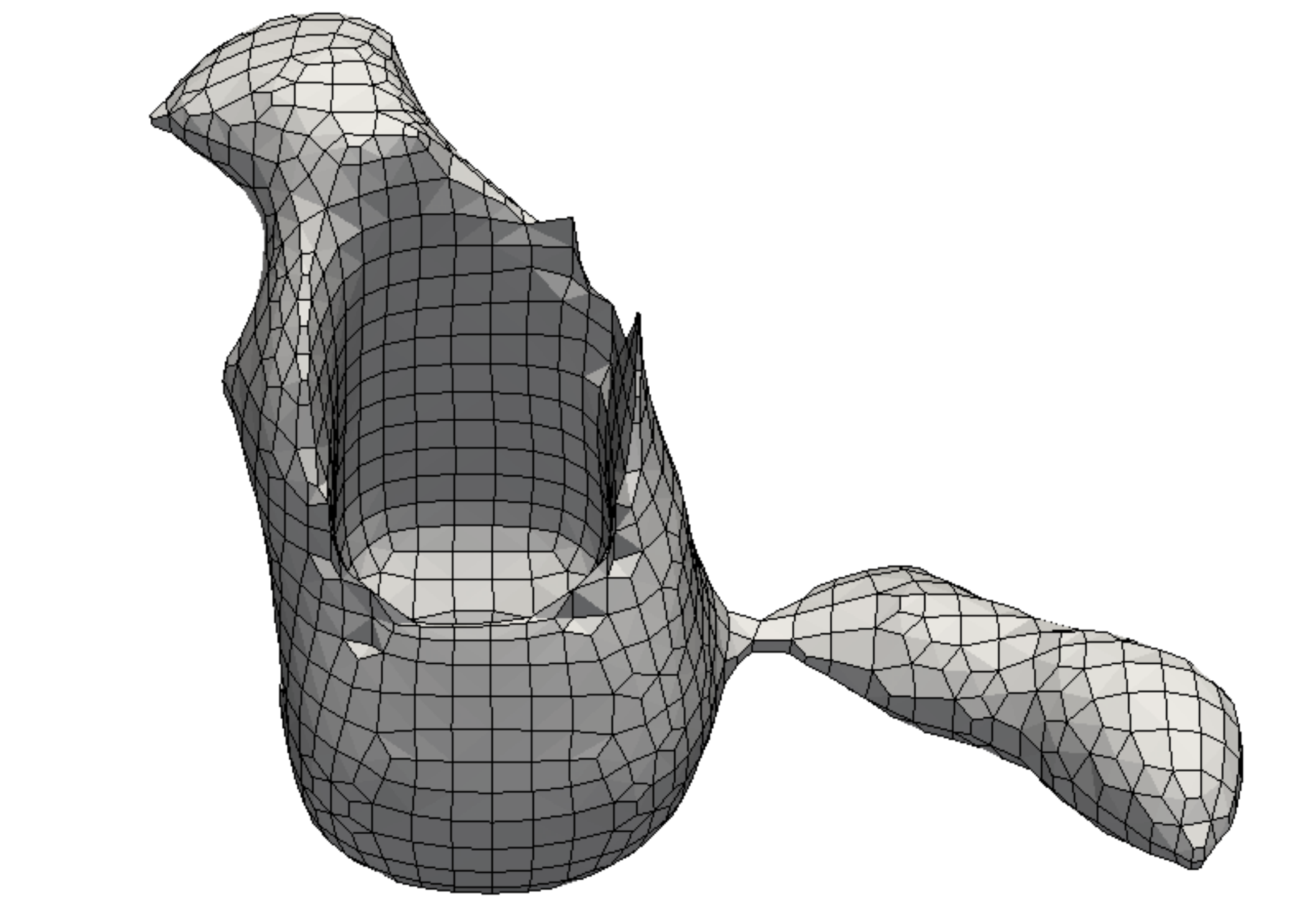}} & \parbox[m]{6em}{\includegraphics[trim={0cm 0cm 0cm 0cm},clip, width=0.12\textwidth]{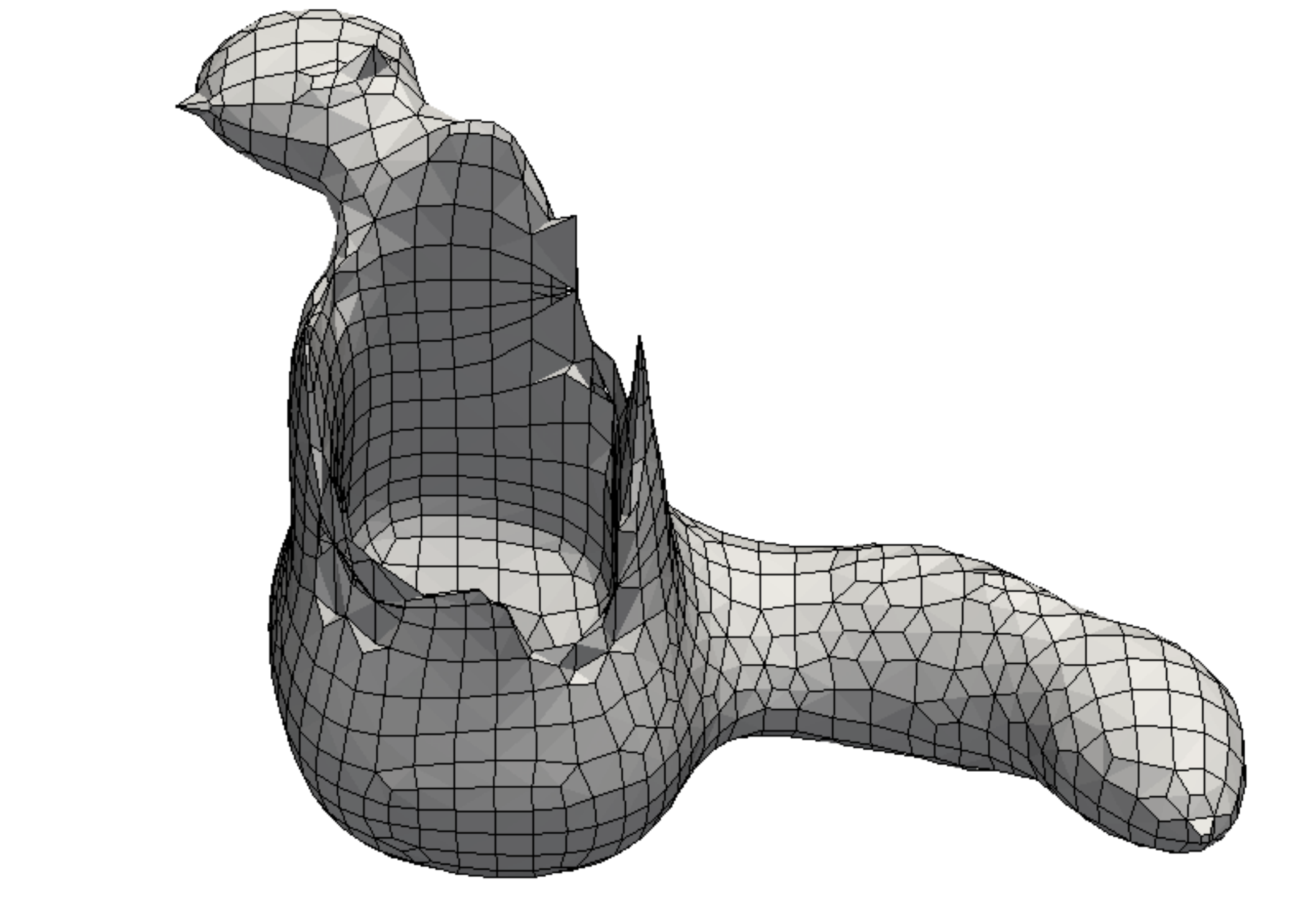}} & \parbox[m]{6em}{\includegraphics[trim={0cm 0cm 0cm 0cm},clip, width=0.12\textwidth]{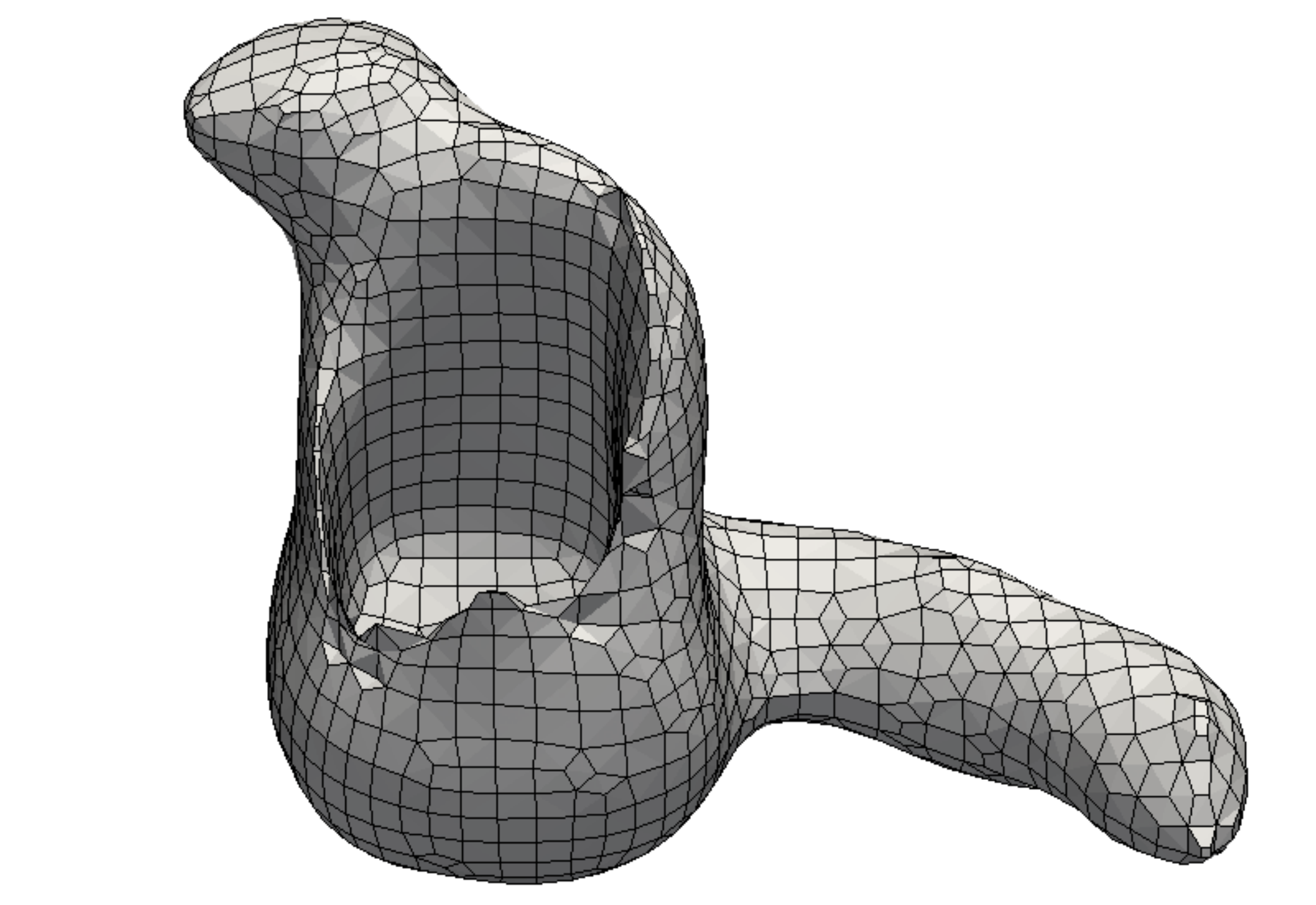}} & \parbox[m]{6em}{\includegraphics[trim={0cm 0cm 0cm 0cm},clip, width=0.12\textwidth]{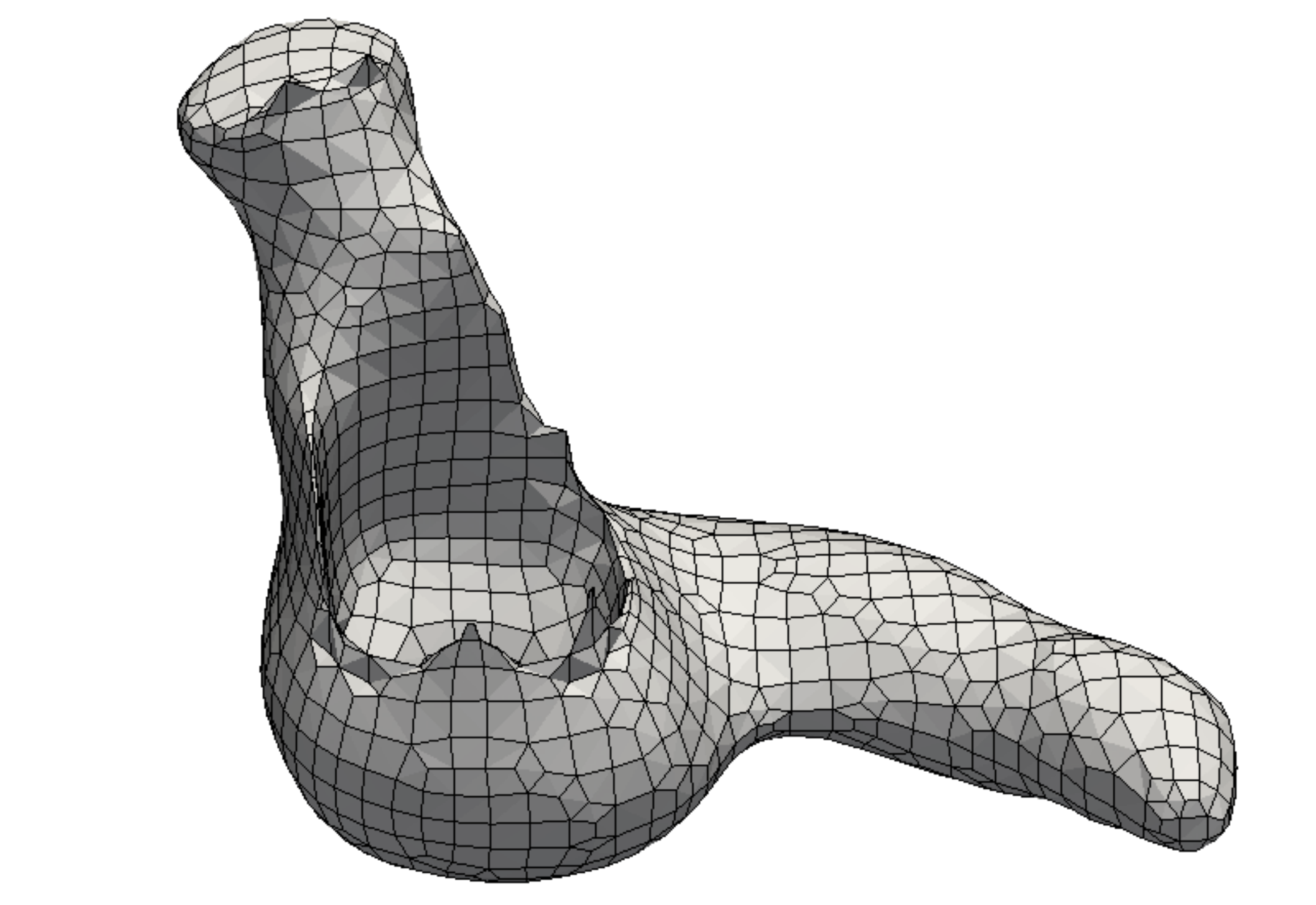}} \\\cline{2-7}
     & \checkmark
     & \parbox[m]{6em}{\includegraphics[trim={0cm 0cm 0cm 0cm},clip, width=0.12\textwidth]{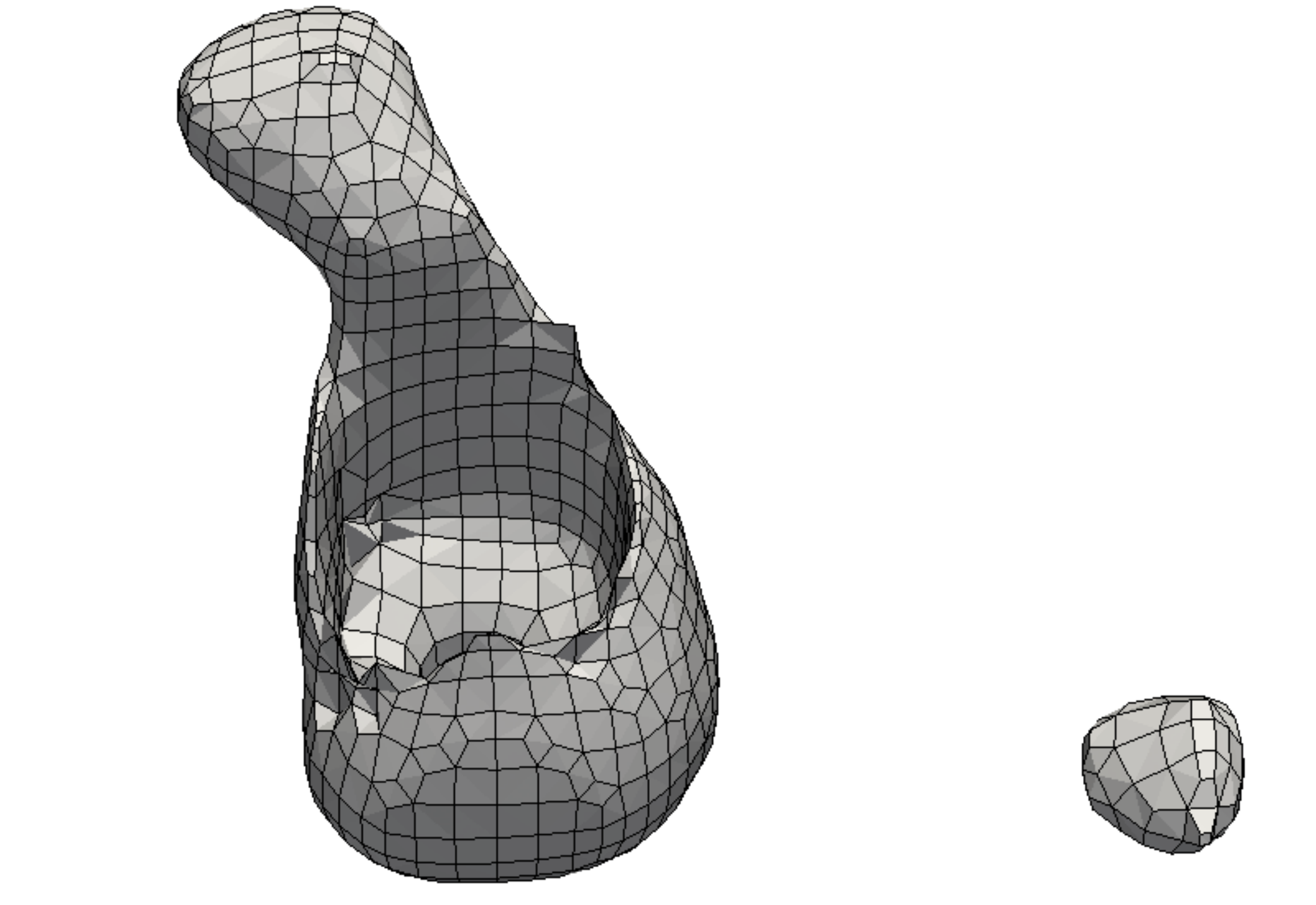}} & \parbox[m]{6em}{\includegraphics[trim={0cm 0cm 0cm 0cm},clip, width=0.12\textwidth]{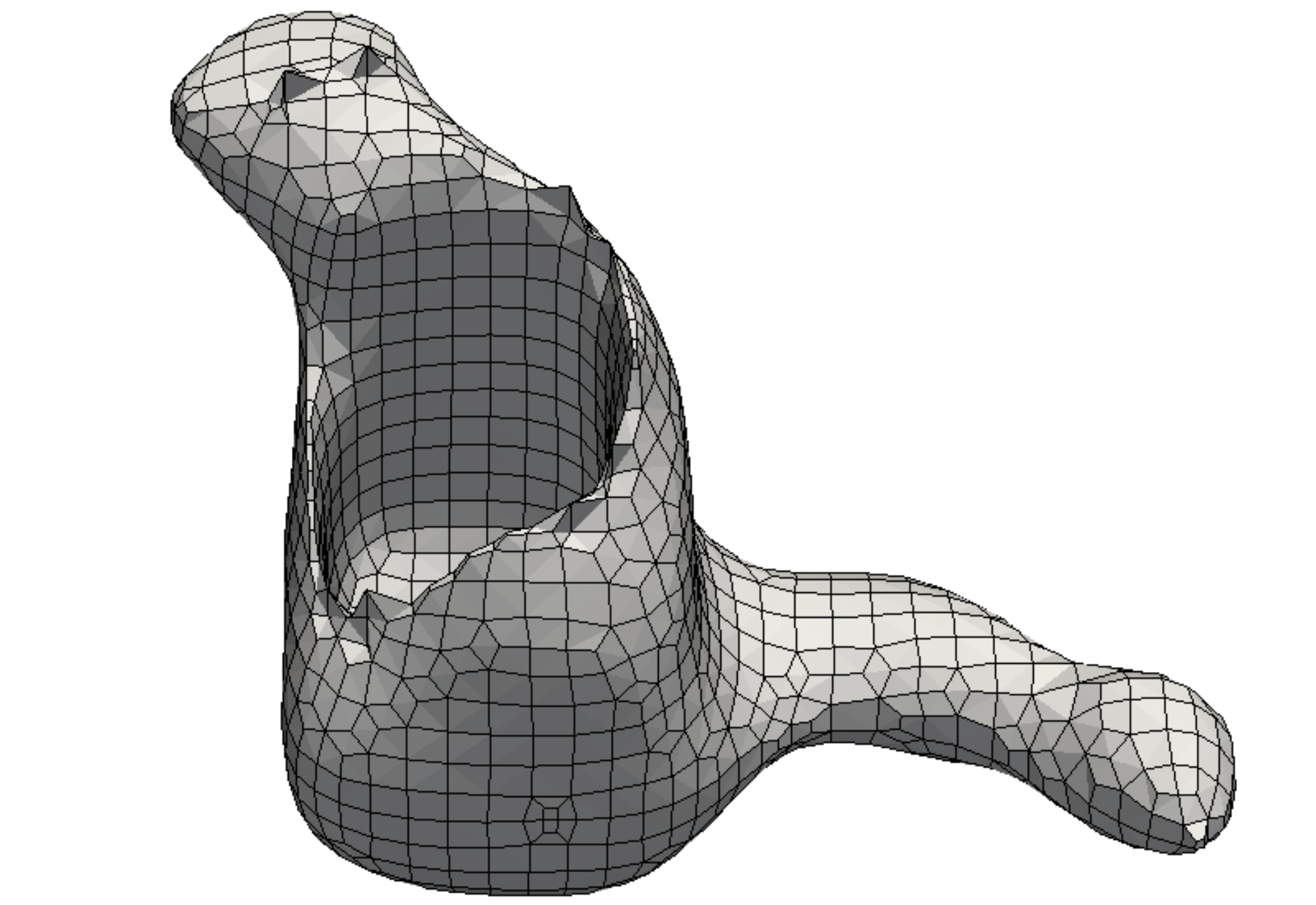}} & \parbox[m]{6em}{\includegraphics[trim={0cm 0cm 0cm 0cm},clip, width=0.12\textwidth]{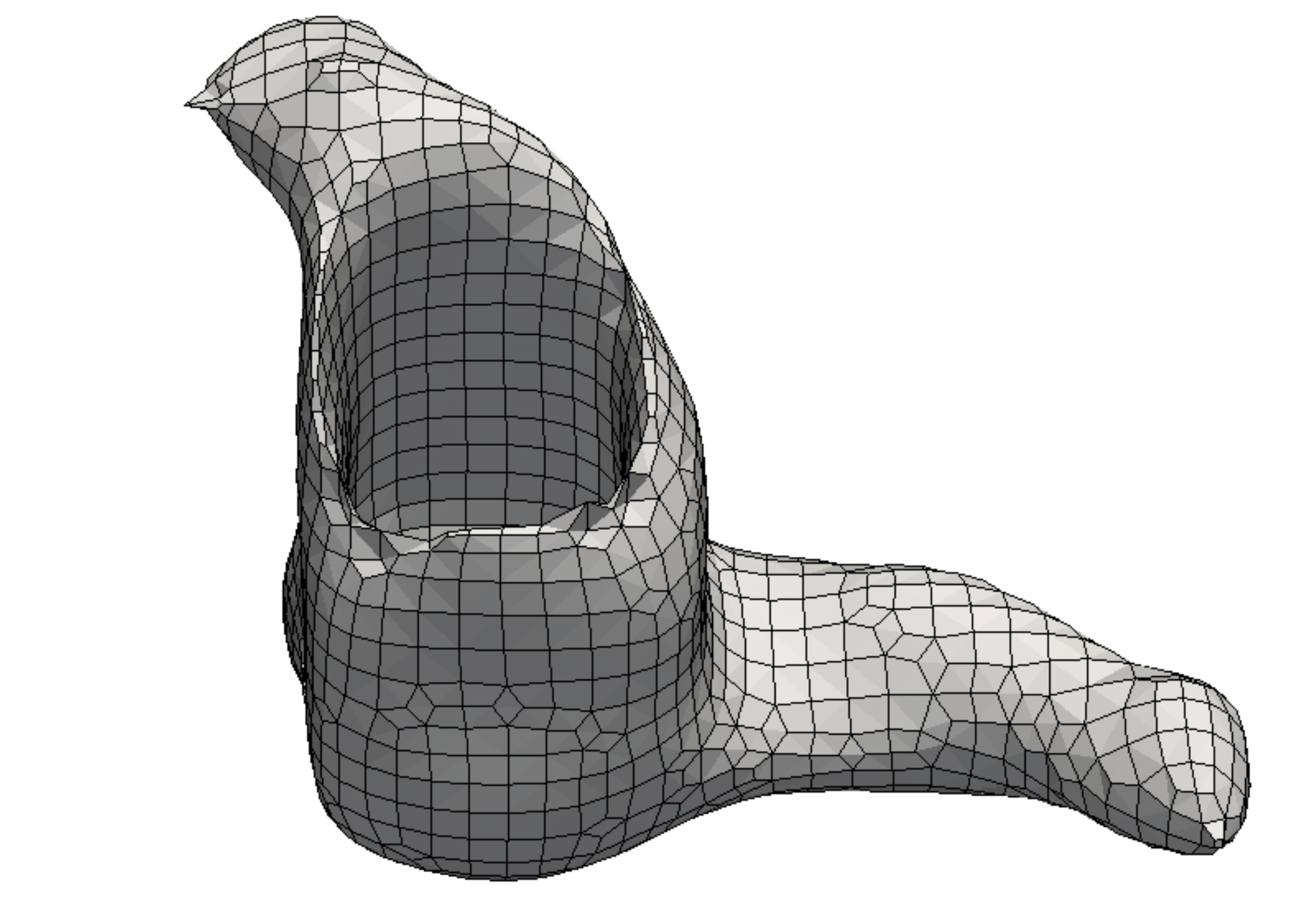}} & \parbox[m]{6em}{\includegraphics[trim={0cm 0cm 0cm 0cm},clip, width=0.12\textwidth]{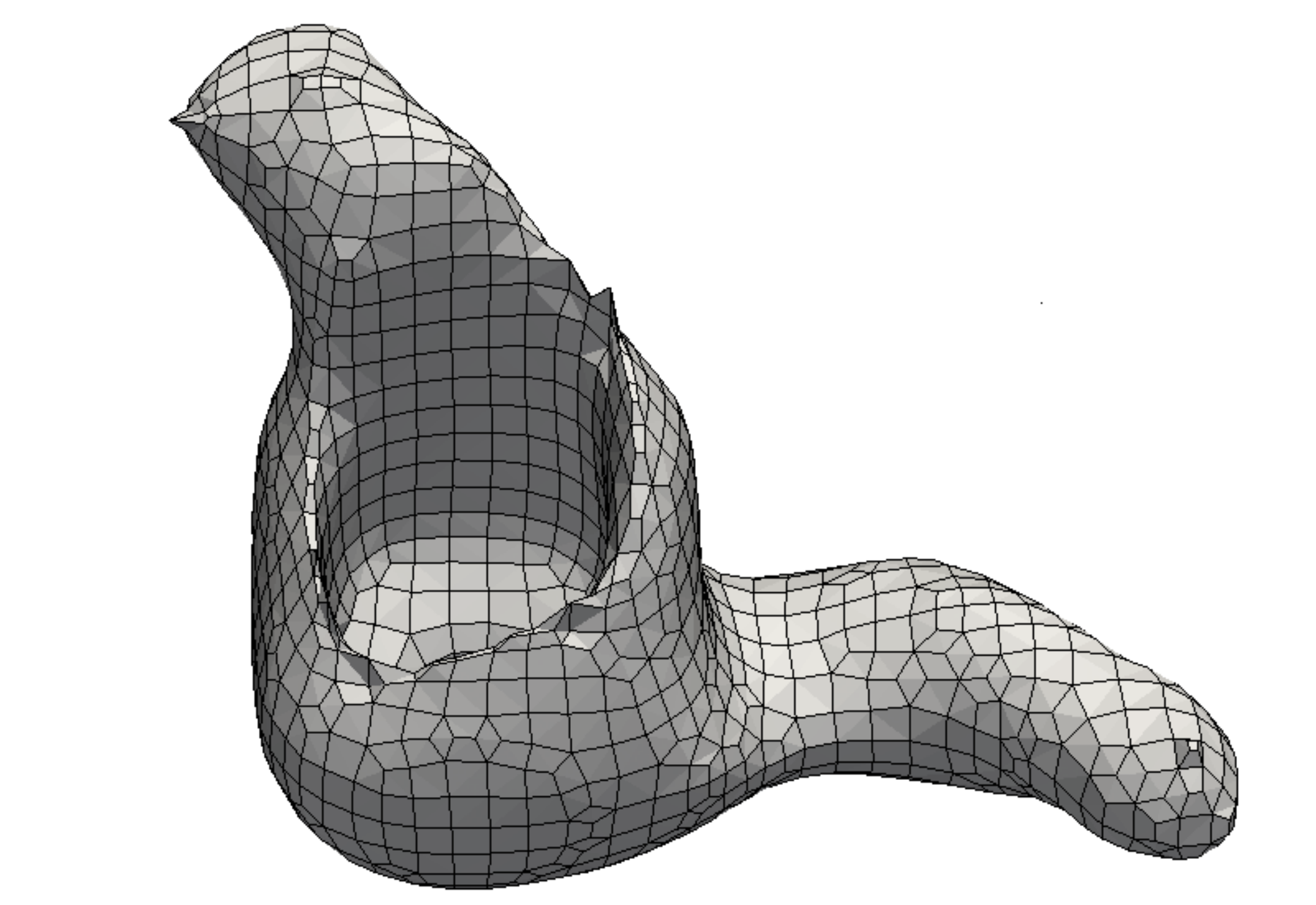}} & \parbox[m]{6em}{\includegraphics[trim={0cm 0cm 0cm 0cm},clip, width=0.12\textwidth]{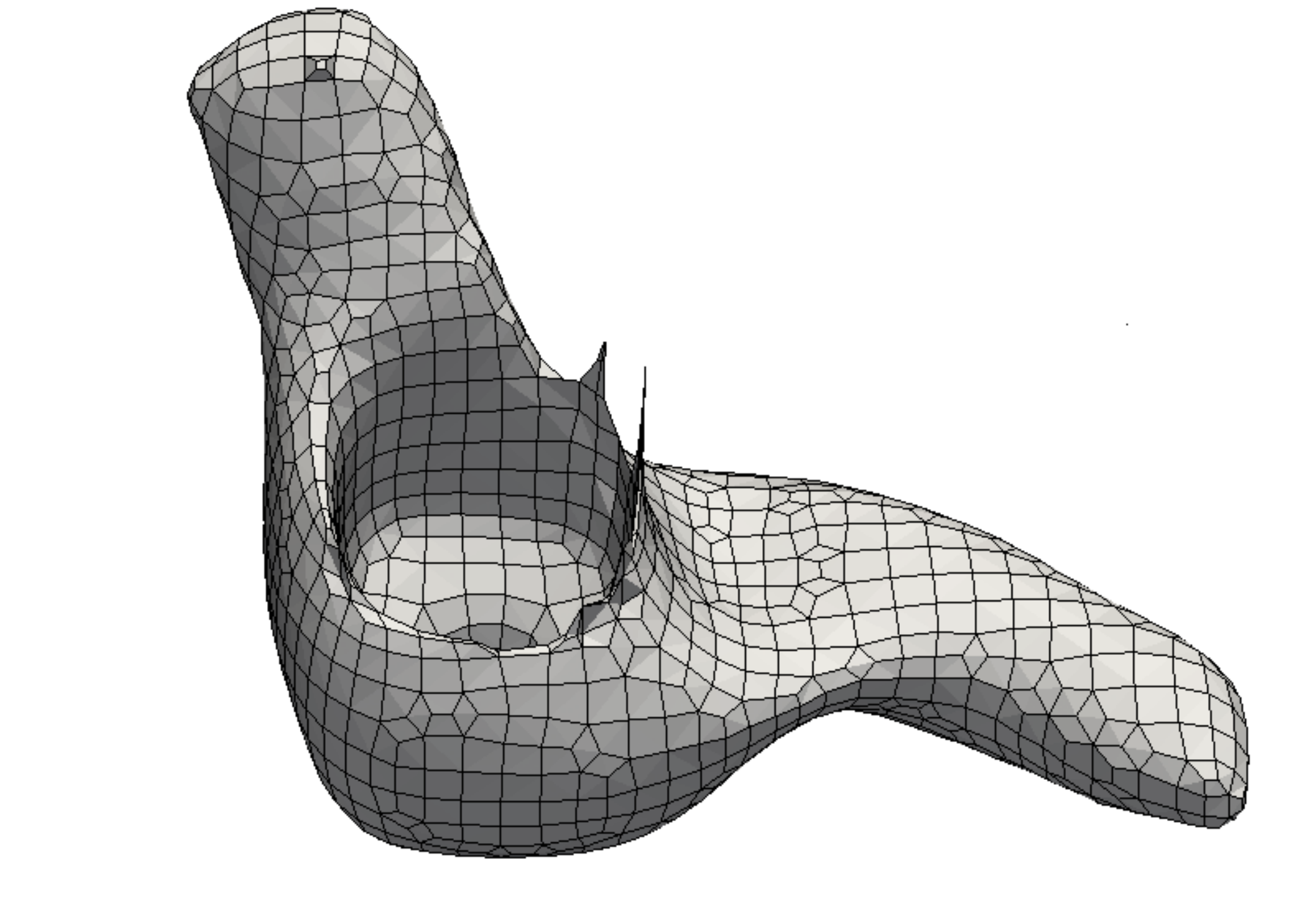}} \\\hline
     \multirow{2}{*}{\rotatebox{90}{trivial+PDE}} & & \parbox[m]{6em}{\includegraphics[trim={0cm 0cm 0cm 0cm},clip, width=0.12\textwidth]{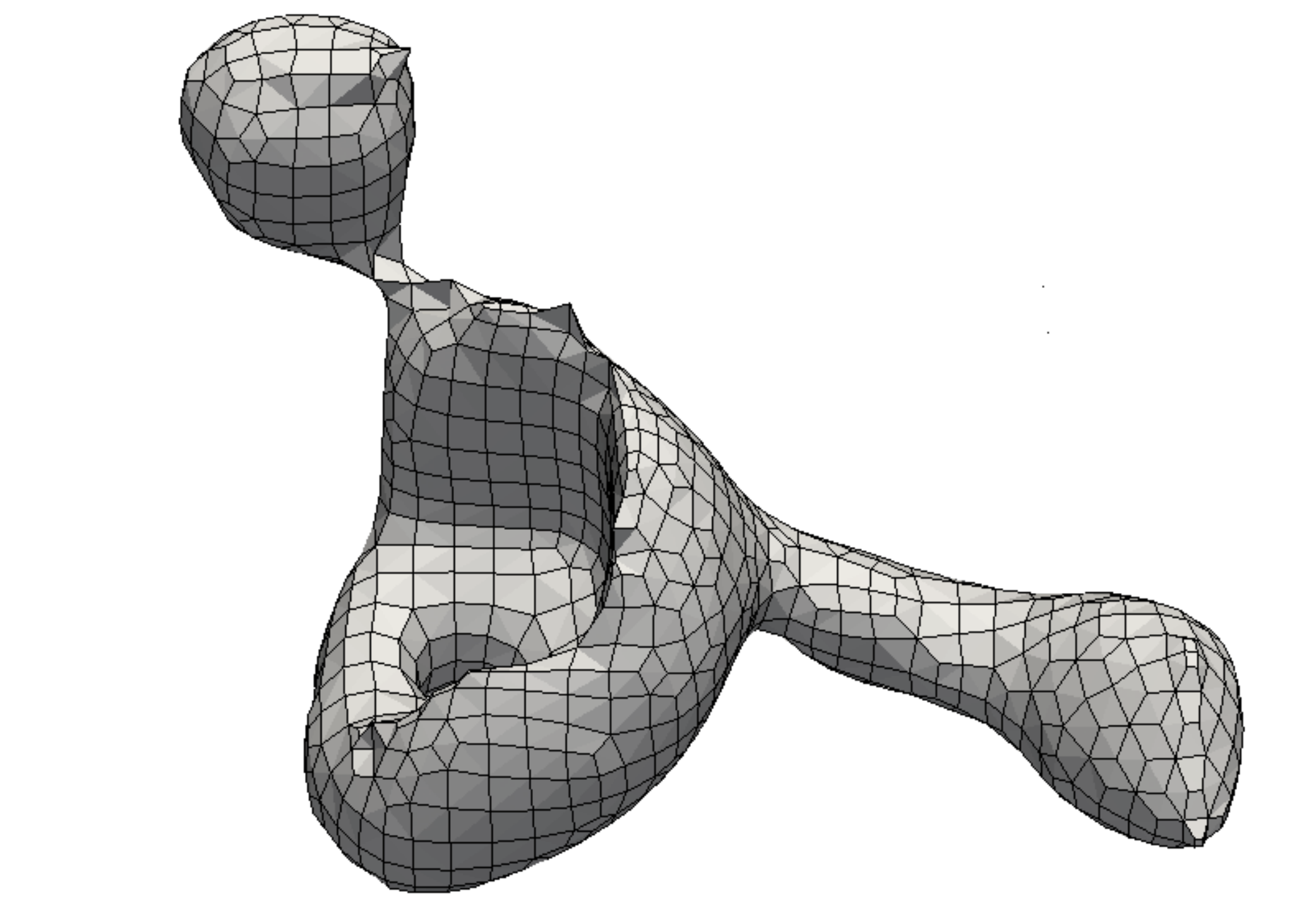}} & \parbox[m]{6em}{\includegraphics[trim={0cm 0cm 0cm 0cm},clip, width=0.12\textwidth]{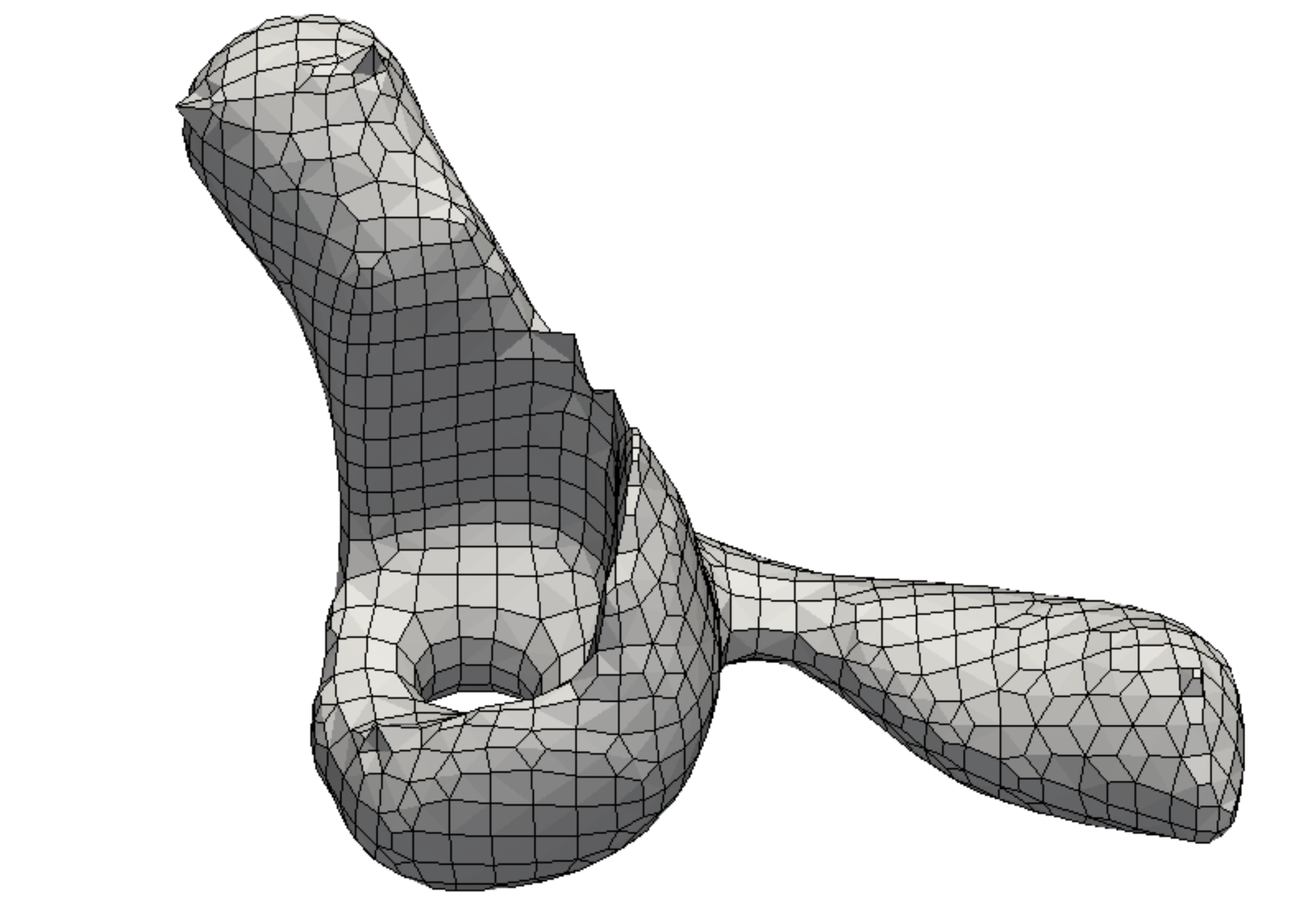}} & \parbox[m]{6em}{\includegraphics[trim={0cm 0cm 0cm 0cm},clip, width=0.12\textwidth]{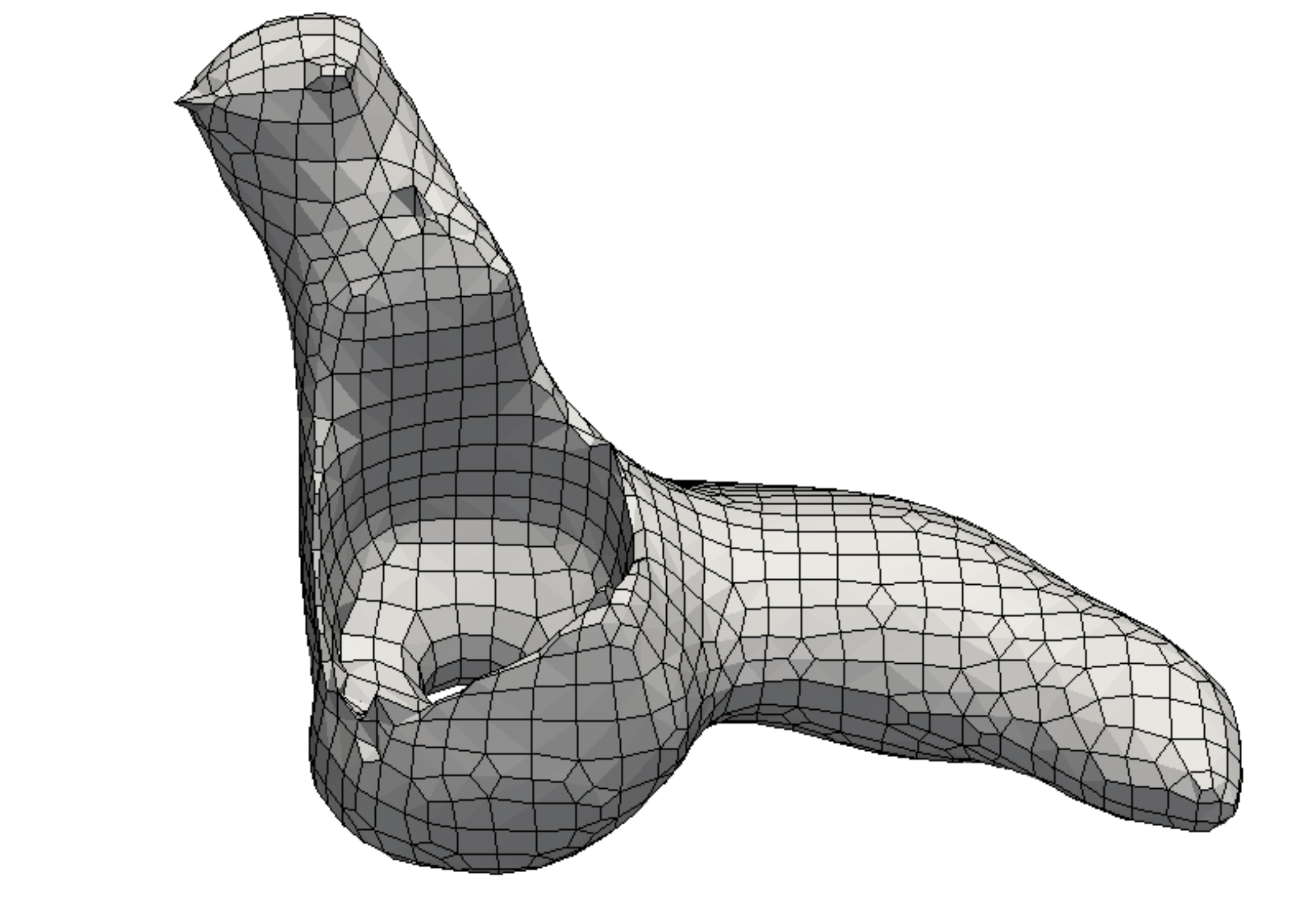}} & \parbox[m]{6em}{\includegraphics[trim={0cm 0cm 0cm 0cm},clip, width=0.12\textwidth]{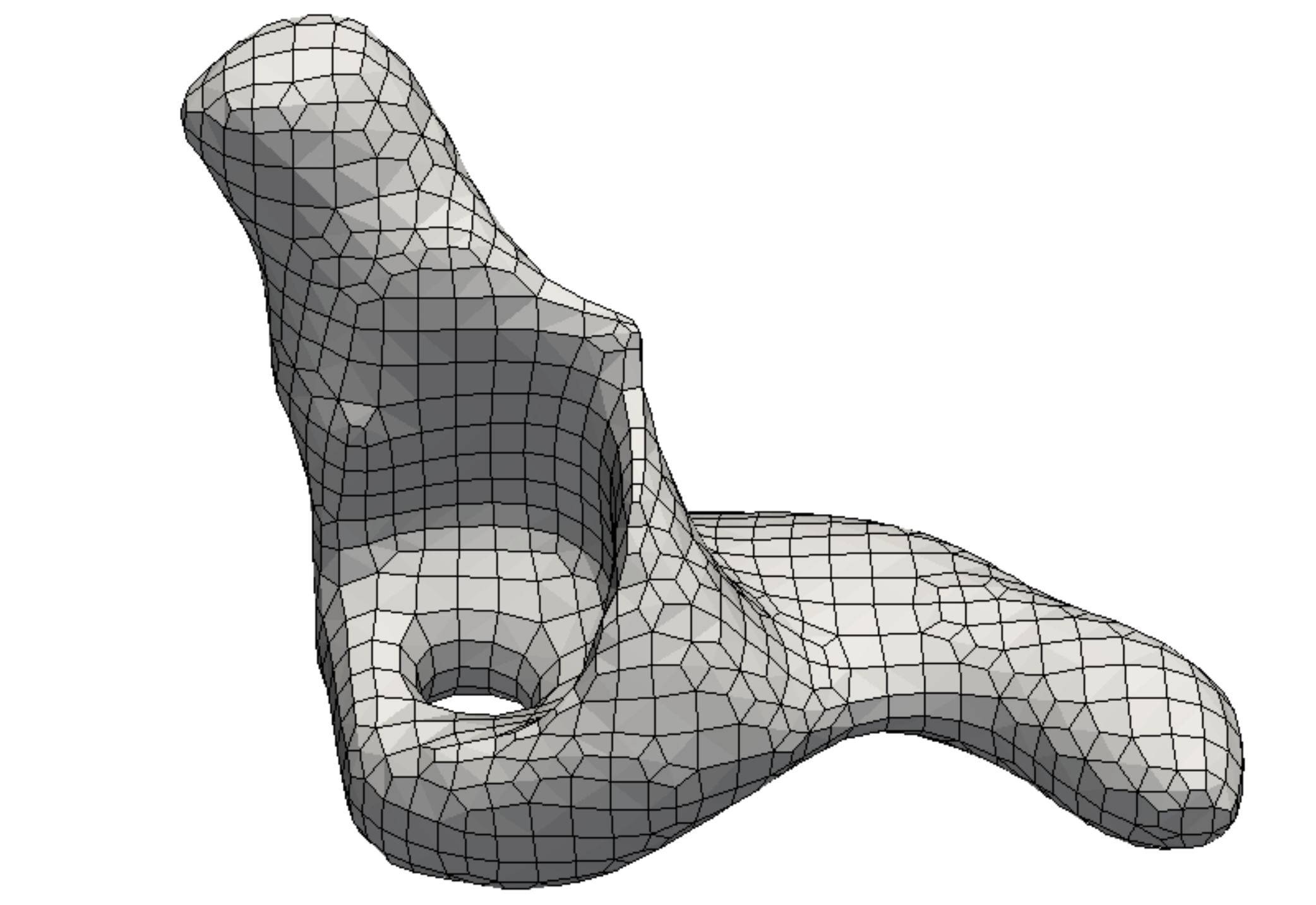}} & \parbox[m]{6em}{\includegraphics[trim={0cm 0cm 0cm 0cm},clip, width=0.12\textwidth]{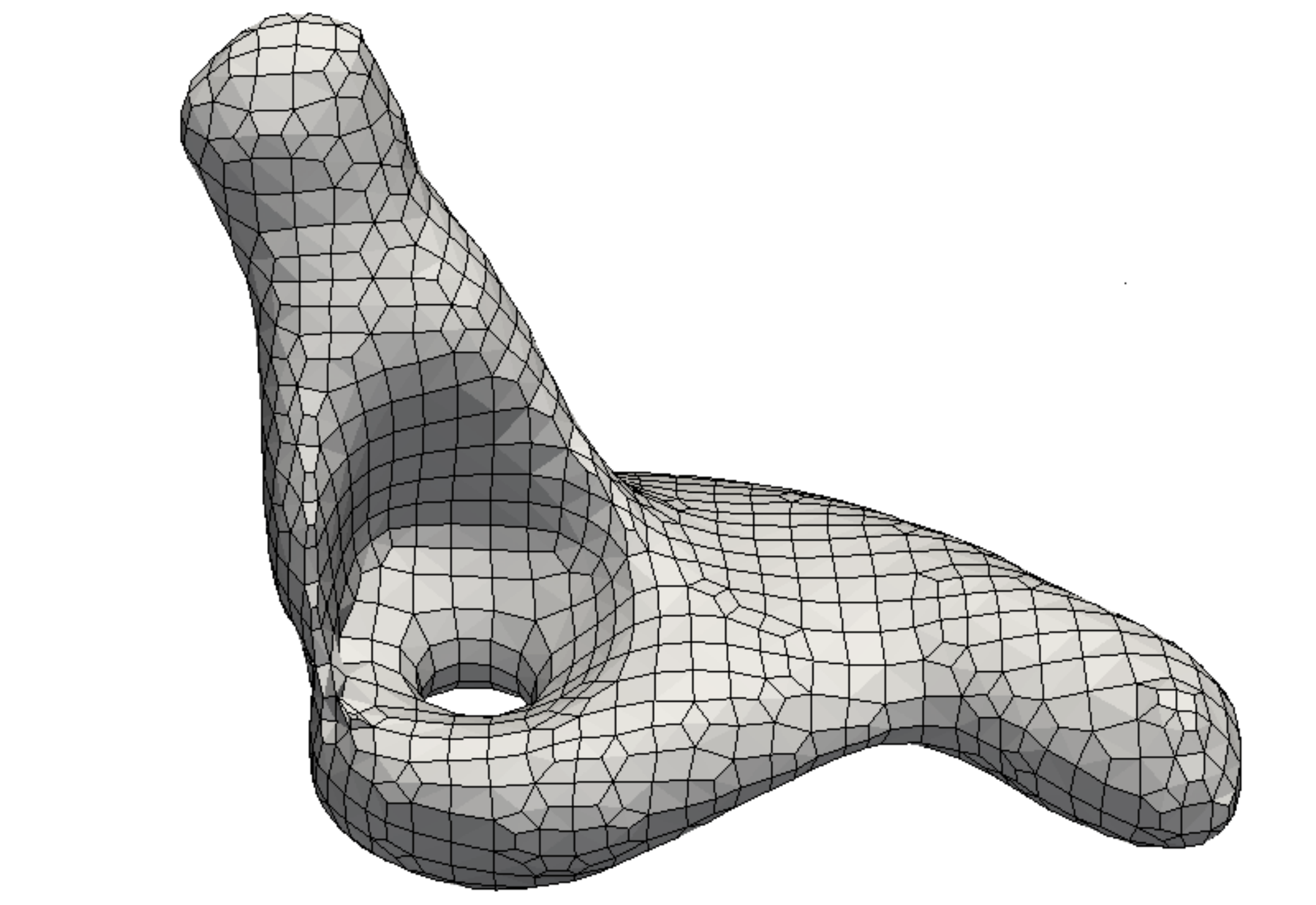}} \\\cline{2-7}
     & \checkmark & \parbox[m]{6em}{\includegraphics[trim={0cm 0cm 0cm 0cm},clip, width=0.12\textwidth]{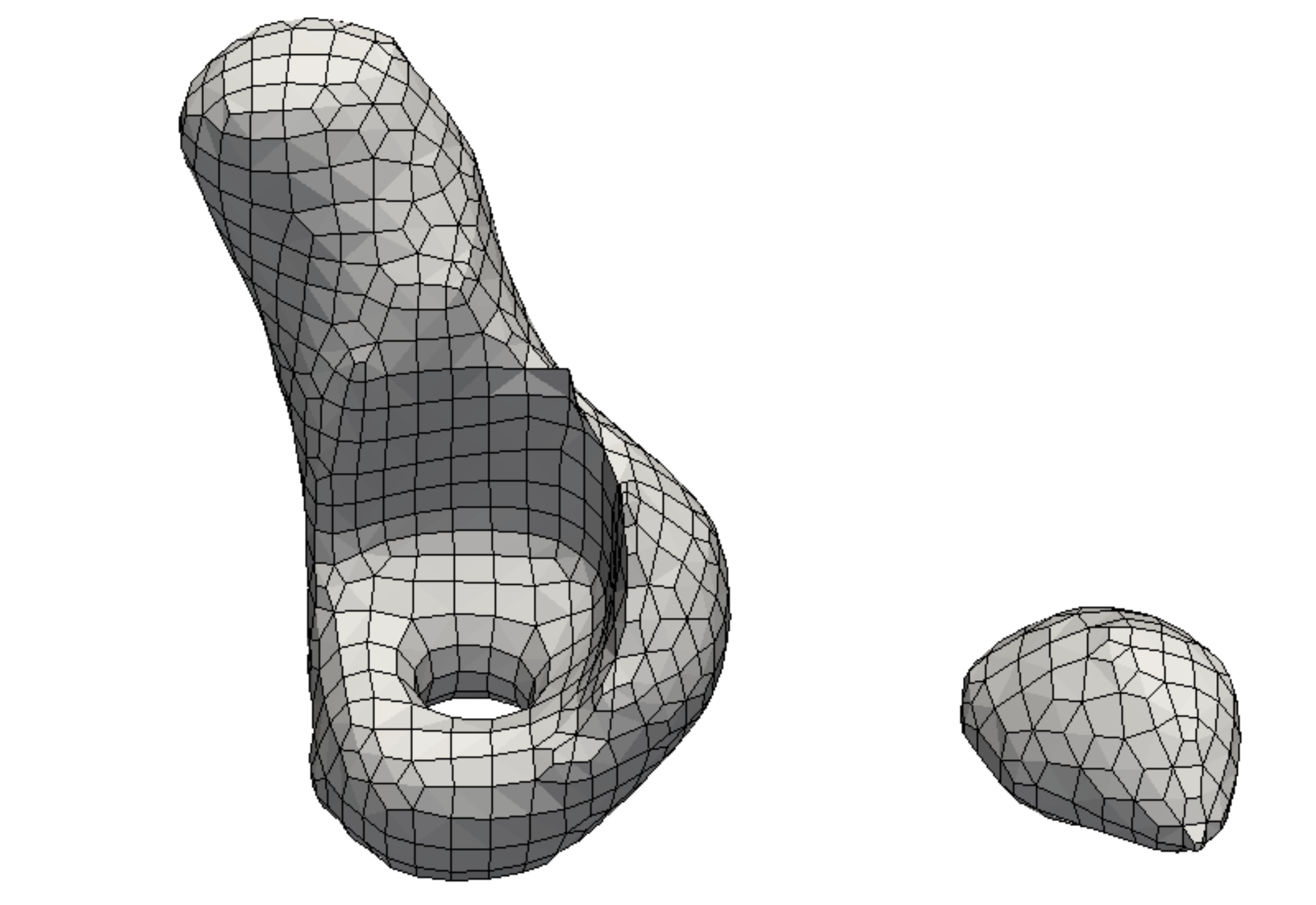}} & \parbox[m]{6em}{\includegraphics[trim={0cm 0cm 0cm 0cm},clip, width=0.12\textwidth]{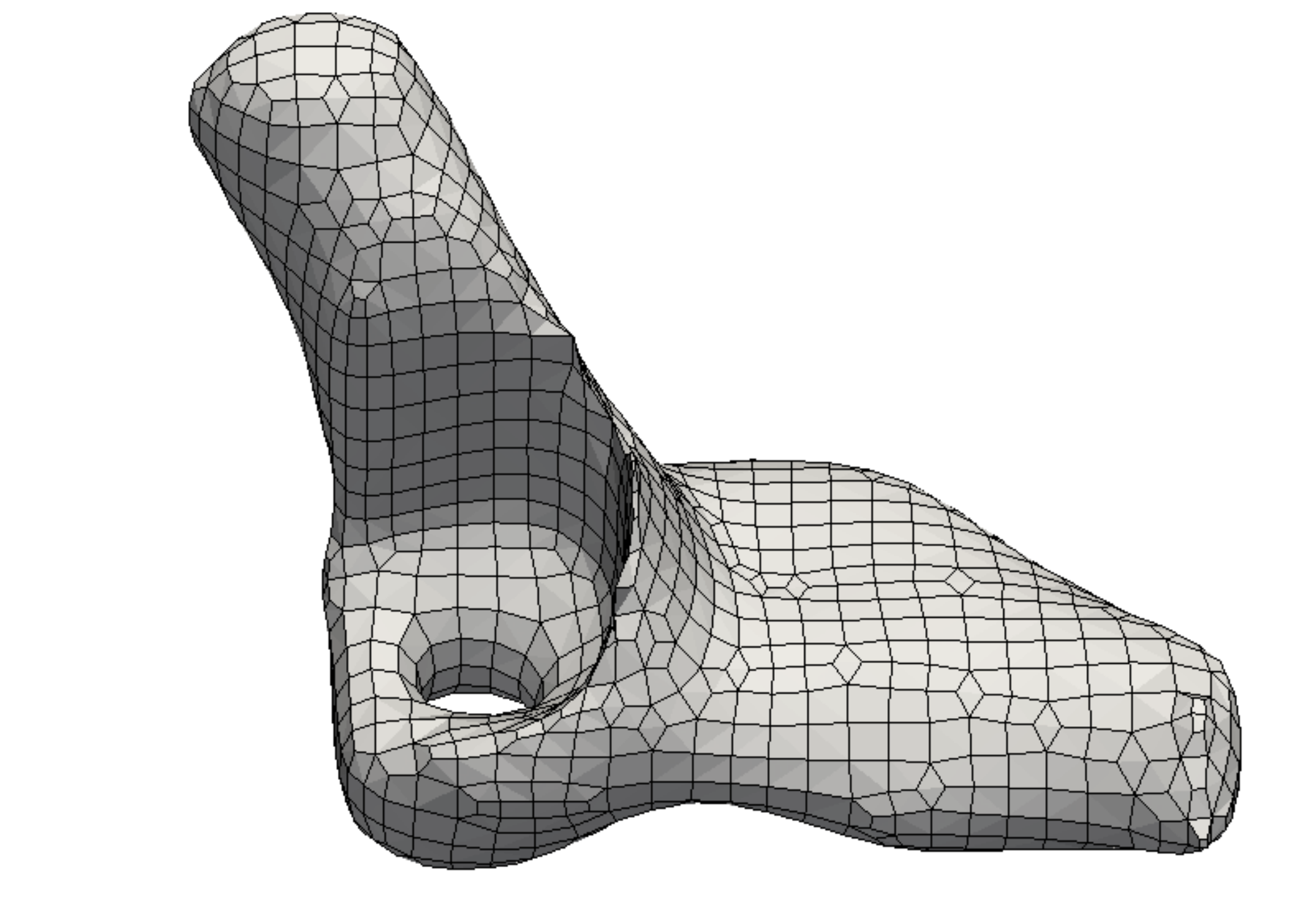}} & \parbox[m]{6em}{\includegraphics[trim={0cm 0cm 0cm 0cm},clip, width=0.12\textwidth]{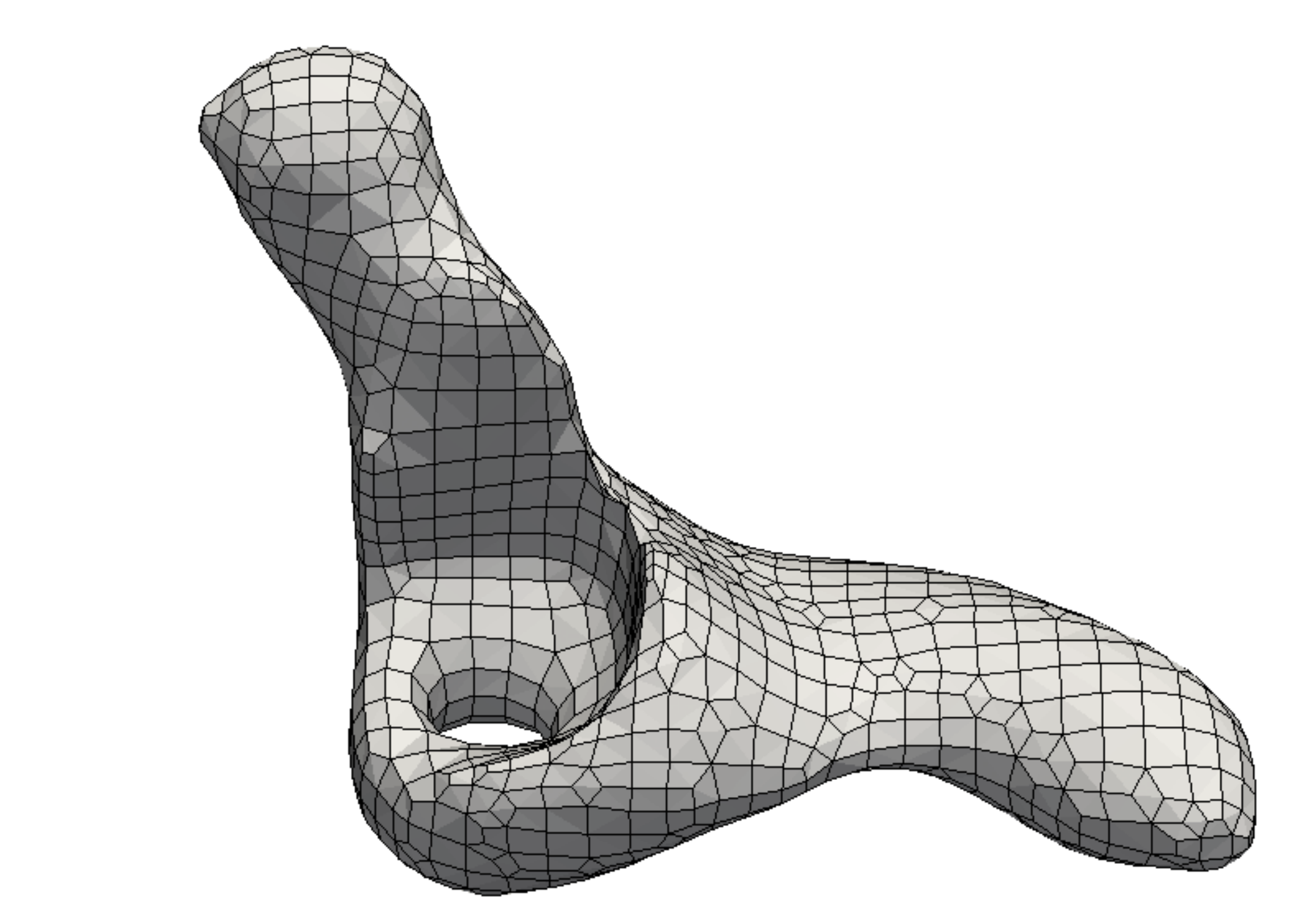}} & \parbox[m]{6em}{\includegraphics[trim={0cm 0cm 0cm 0cm},clip, width=0.12\textwidth]{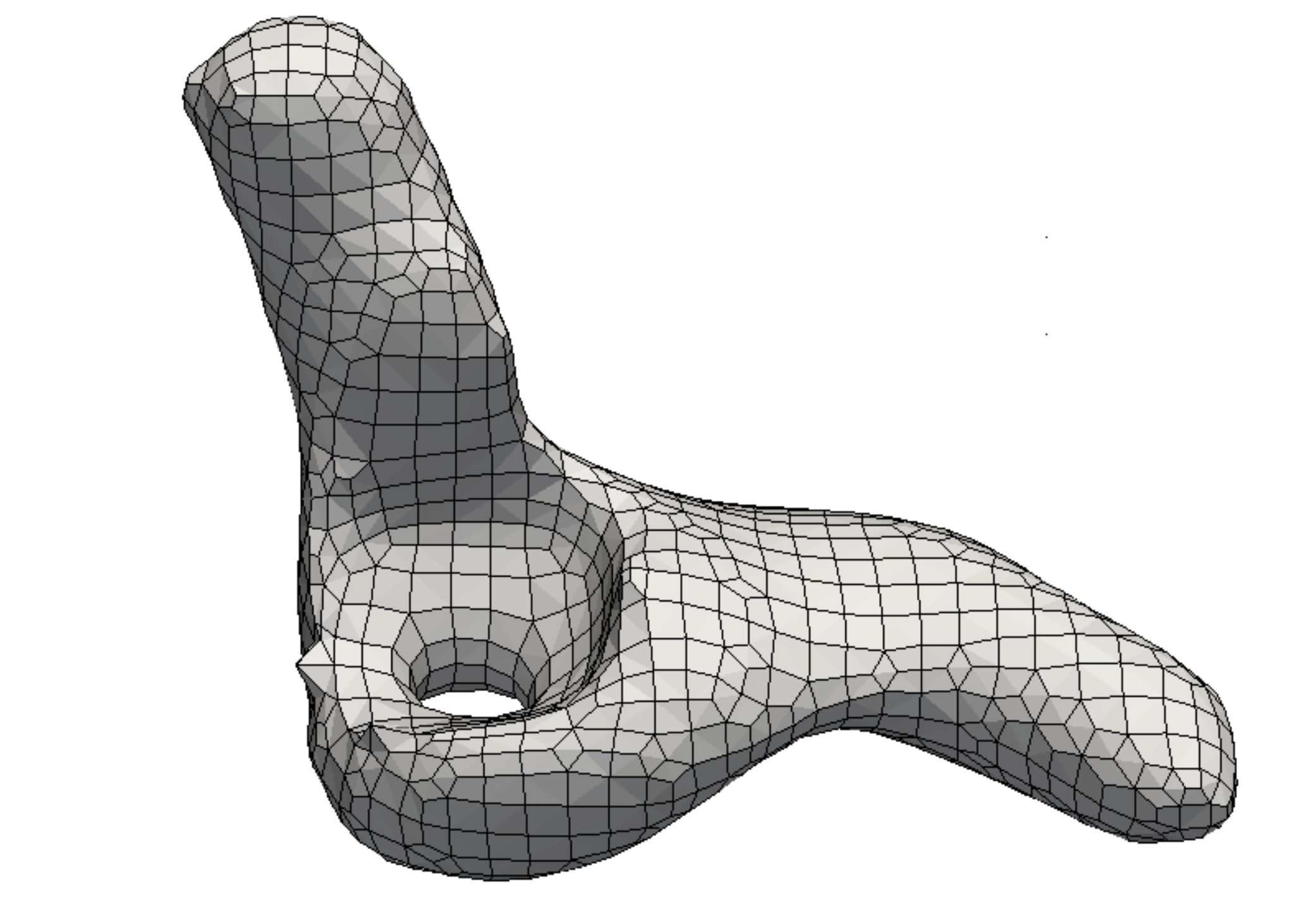}} & \parbox[m]{6em}{\includegraphics[trim={0cm 0cm 0cm 0cm},clip, width=0.12\textwidth]{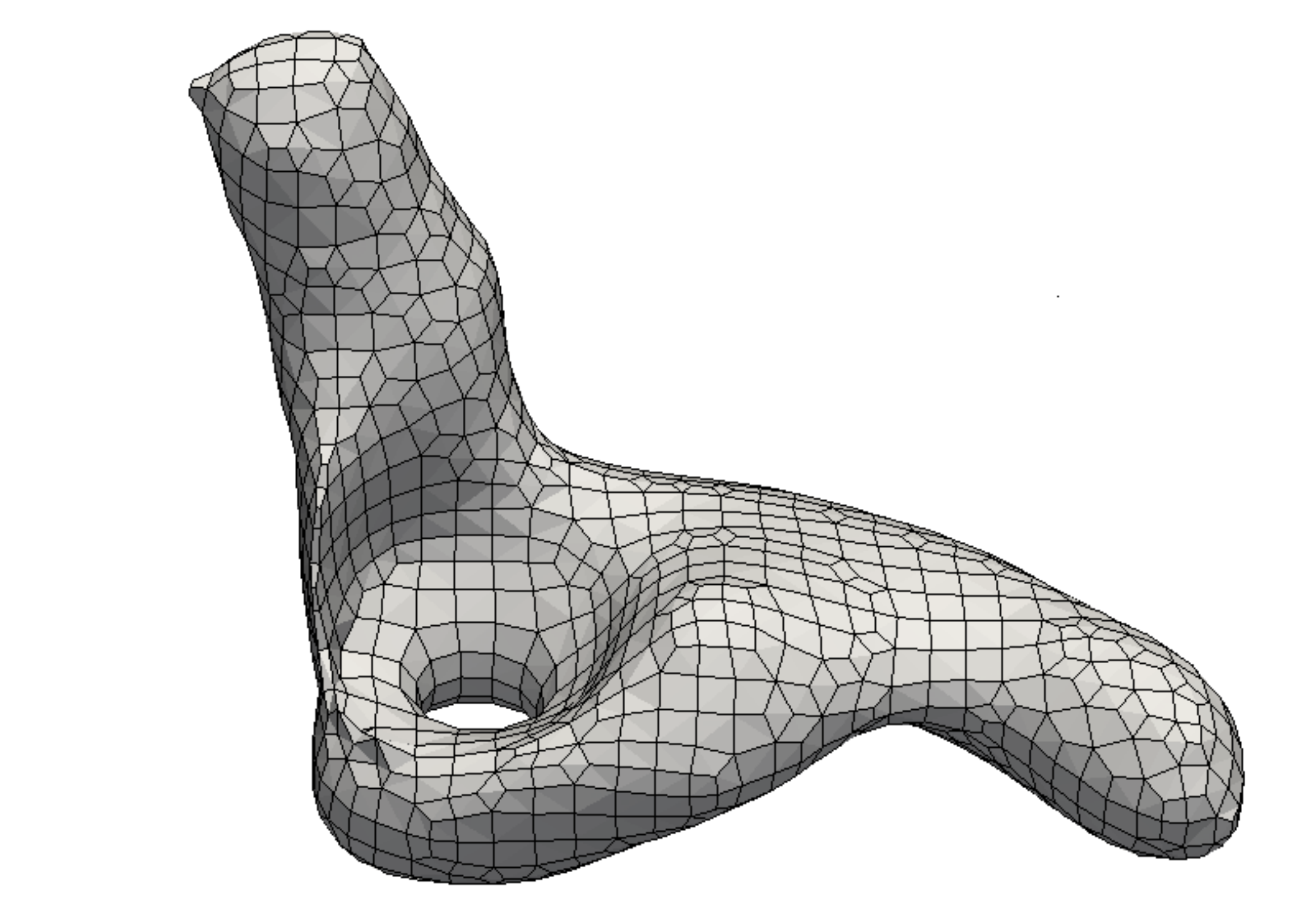}}\\
     \multicolumn{7}{c}{\fbox{\hspace{0.3cm}\parbox[m]{6em}{\includegraphics[trim={0cm 0cm 0cm 0cm},clip, width=0.12\textwidth]{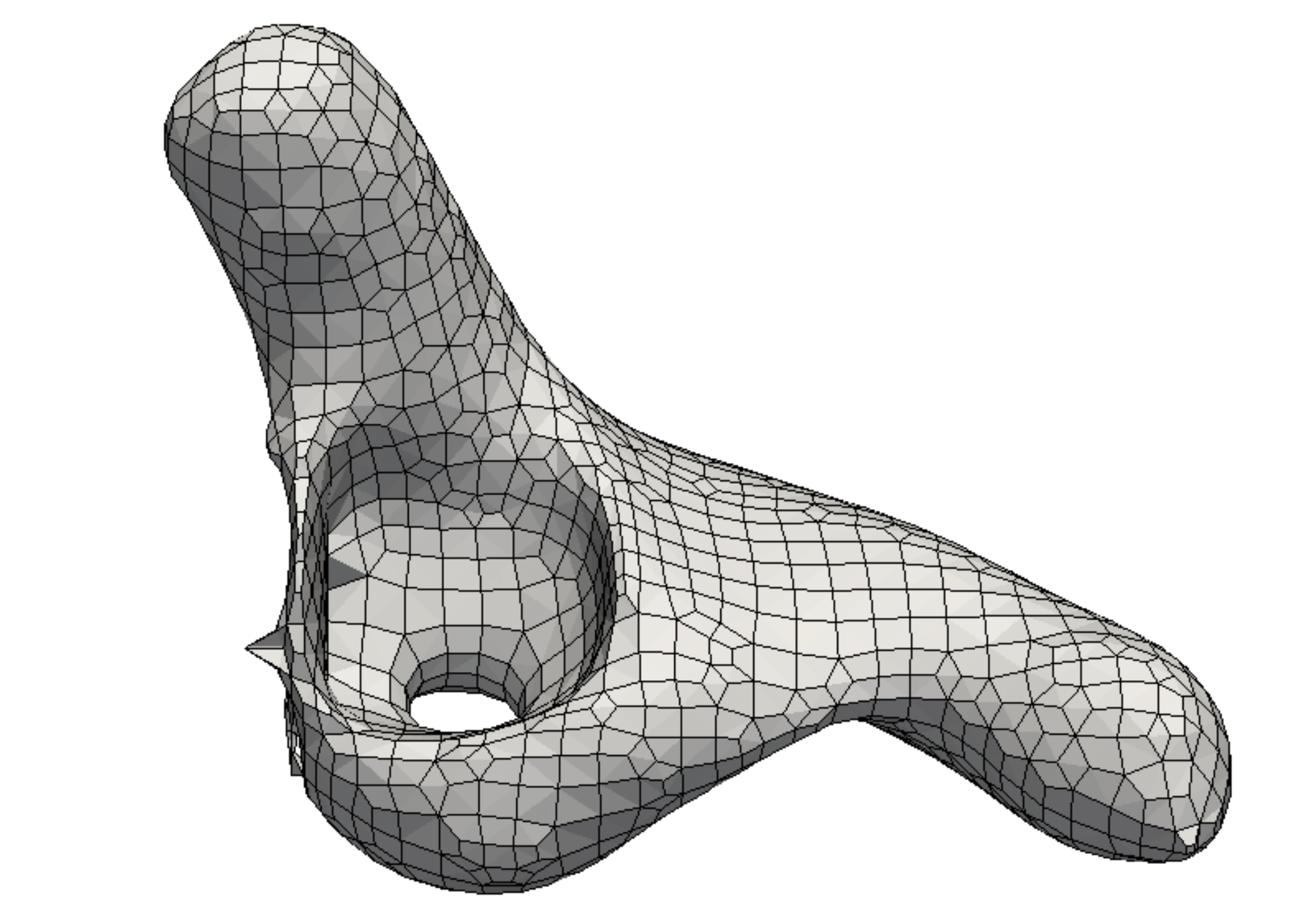}}}}
\end{tabular}
\end{subtable}
\begin{subtable}[h]{0.99\textwidth}
    \vspace{5mm}%
    \centering\setcellgapes{3pt}\makegapedcells
    \setlength\tabcolsep{3.5pt}
    \begin{tabular}{c|c||ScScScScSc}
    \multicolumn{2}{c||}{} & \multicolumn{5}{c}{training samples} \\\hline
     prepr. & equiv. & 10 & 50 & 100 & 500 & 1500 \\\hline
     \multirow{2}{*}{\rotatebox{90}{trivial}} & & \parbox[m]{6em}{\includegraphics[trim={0cm 0cm 0cm 0cm},clip, width=0.12\textwidth]{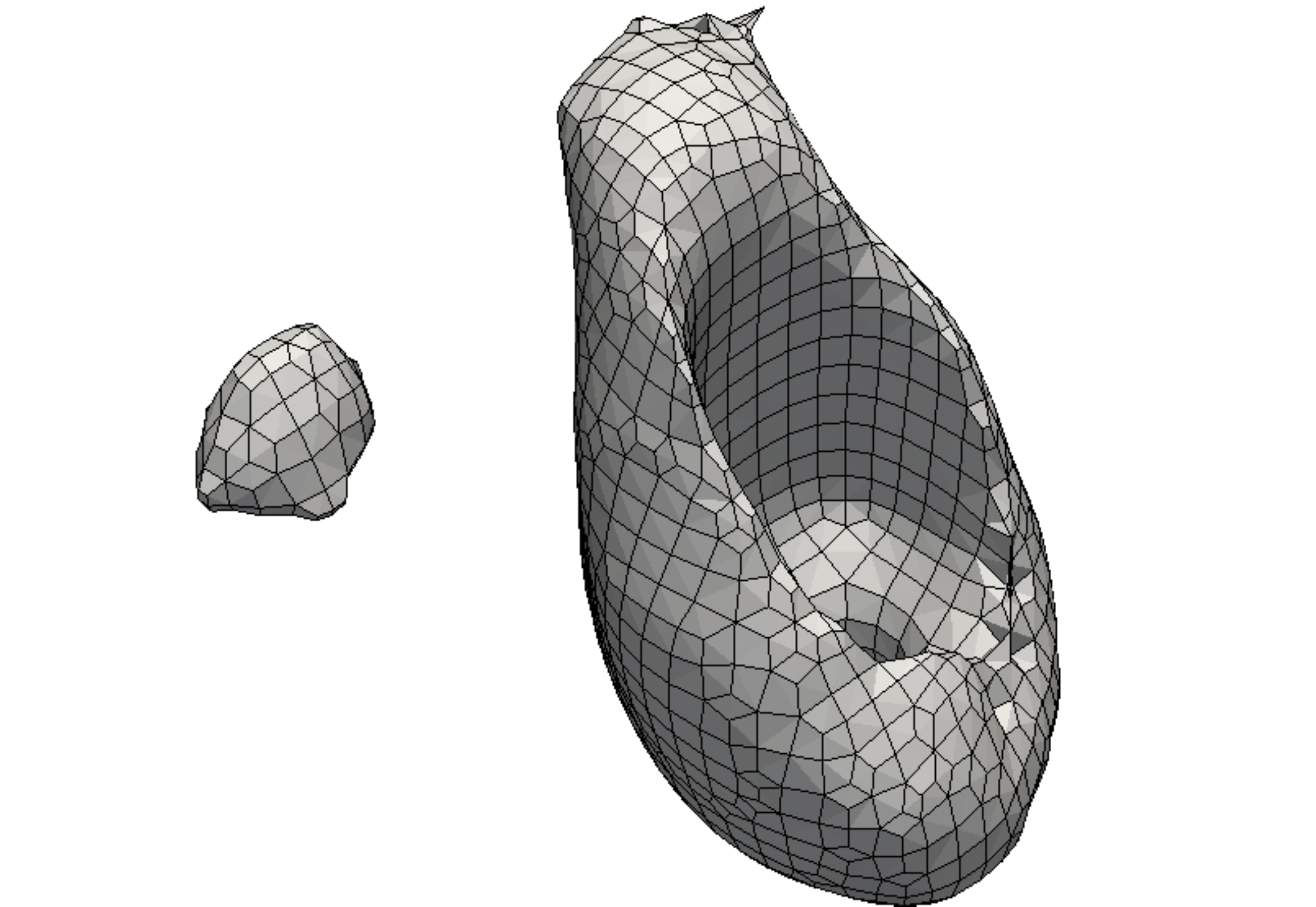}} & \parbox[m]{6em}{\includegraphics[trim={0cm 0cm 0cm 0cm},clip, width=0.12\textwidth]{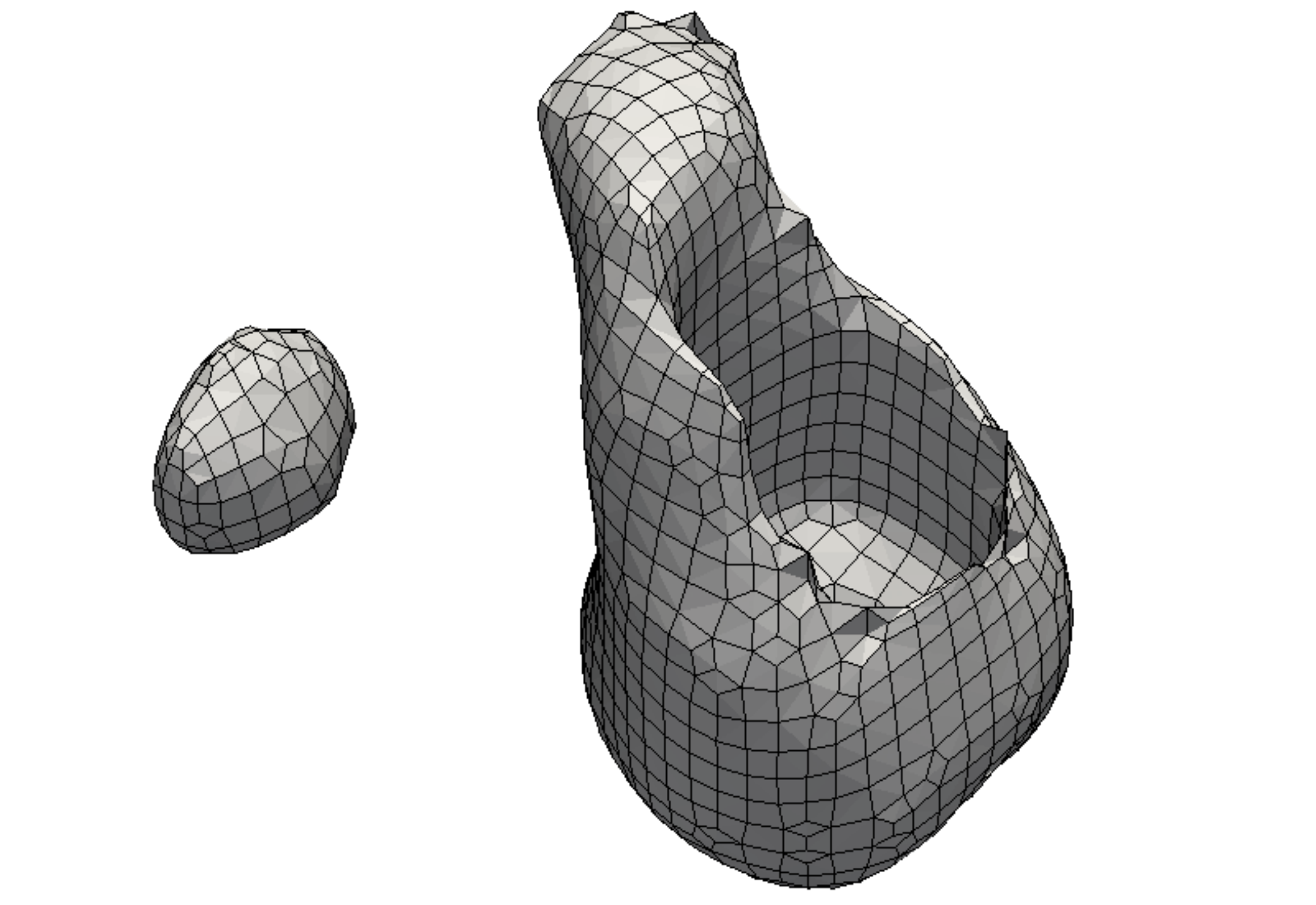}} & \parbox[m]{6em}{\includegraphics[trim={0cm 0cm 0cm 0cm},clip, width=0.12\textwidth]{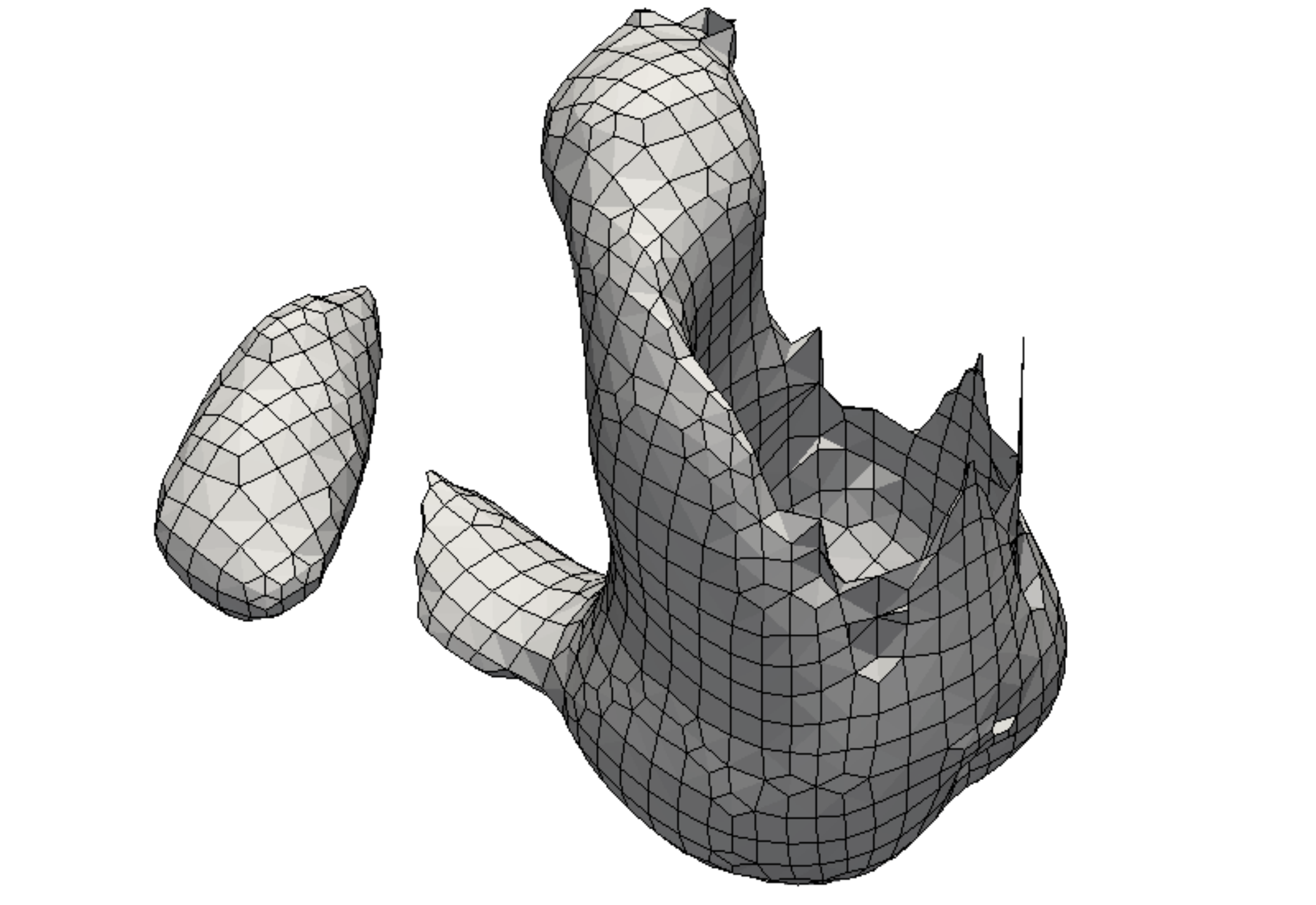}} & \parbox[m]{6em}{\includegraphics[trim={0cm 0cm 0cm 0cm},clip, width=0.12\textwidth]{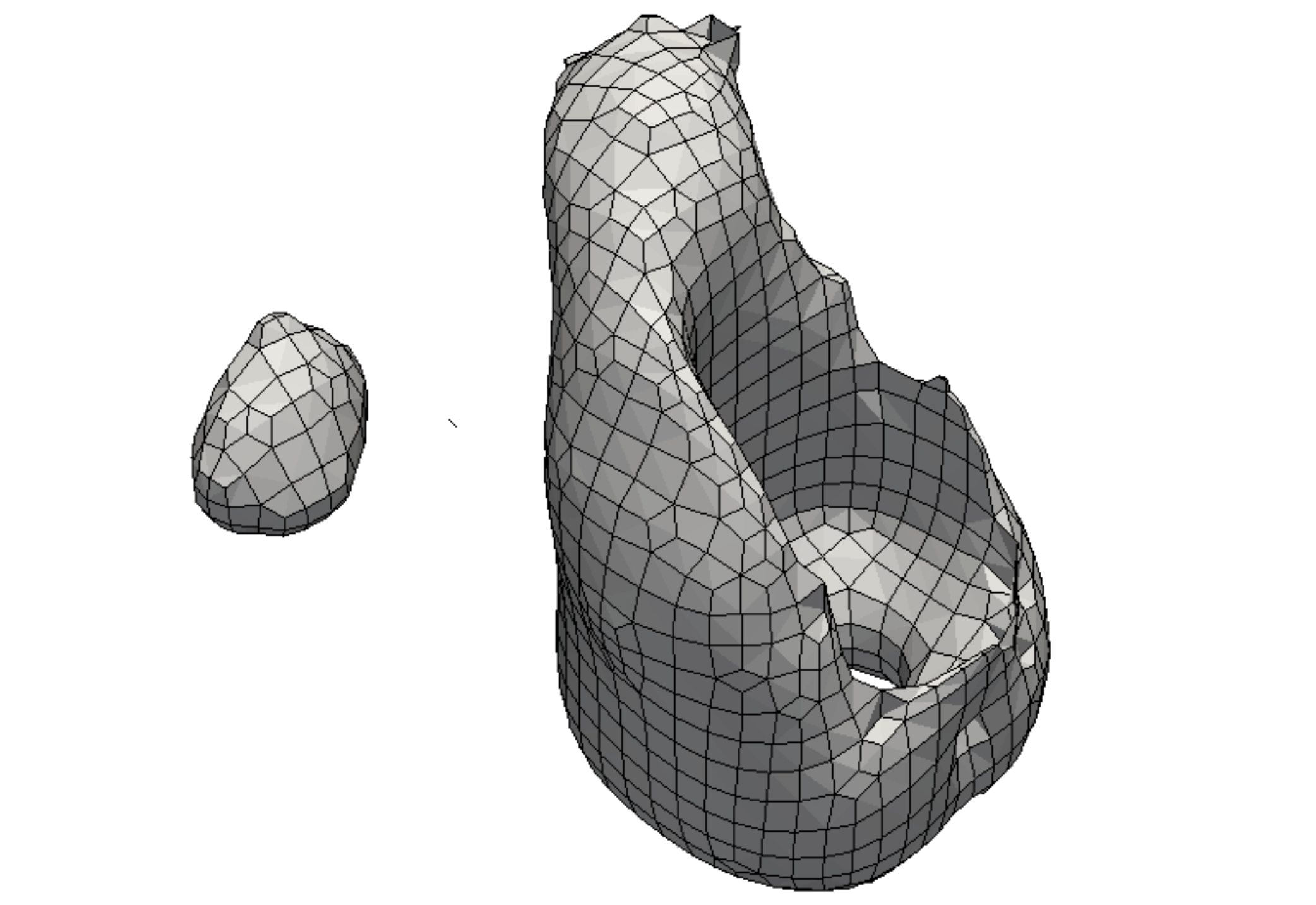}} & \parbox[m]{6em}{\includegraphics[trim={0cm 0cm 0cm 0cm},clip, width=0.12\textwidth]{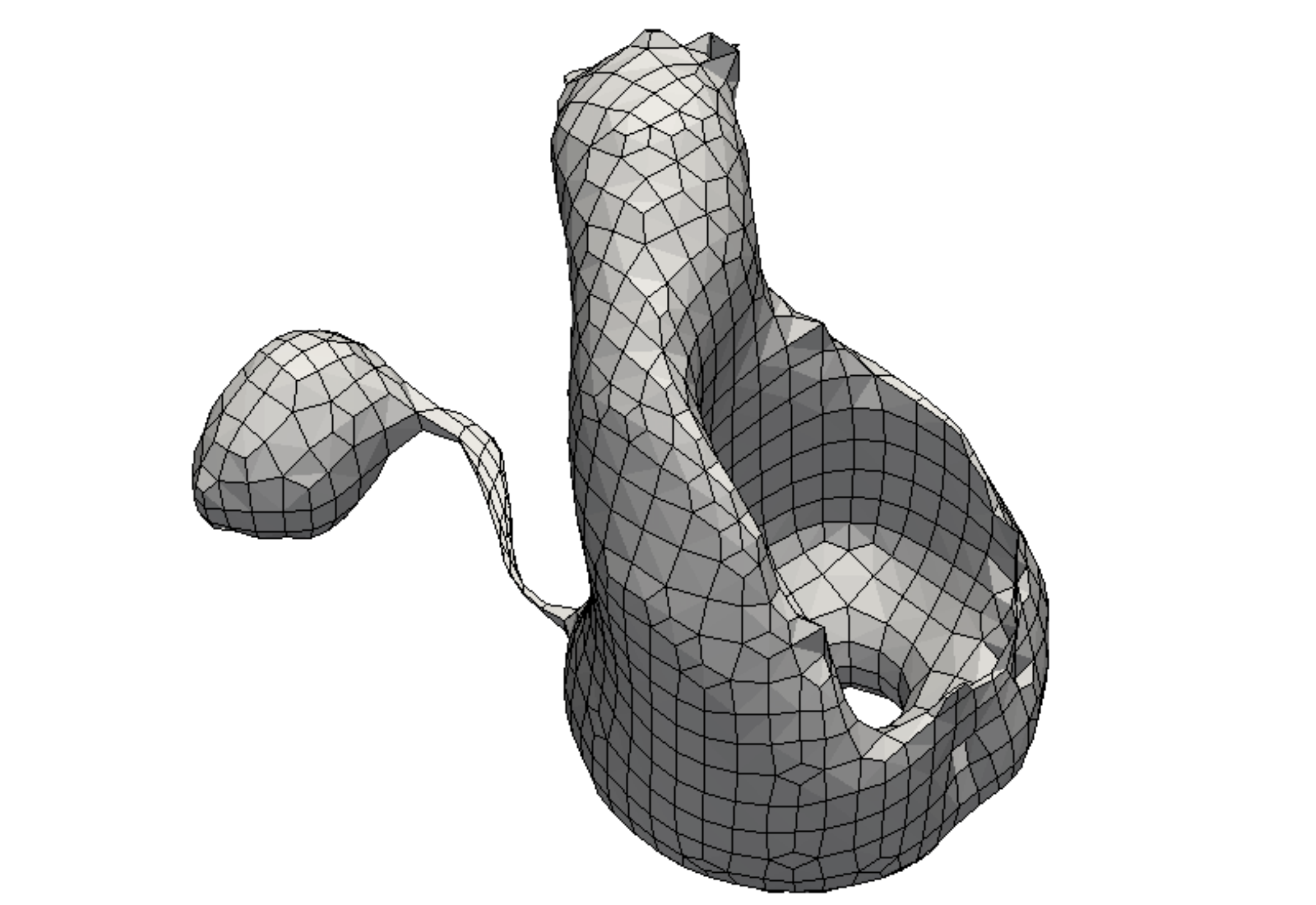}} \\\cline{2-7}
     & \checkmark
     & \parbox[m]{6em}{\includegraphics[trim={0cm 0cm 0cm 0cm},clip, width=0.12\textwidth]{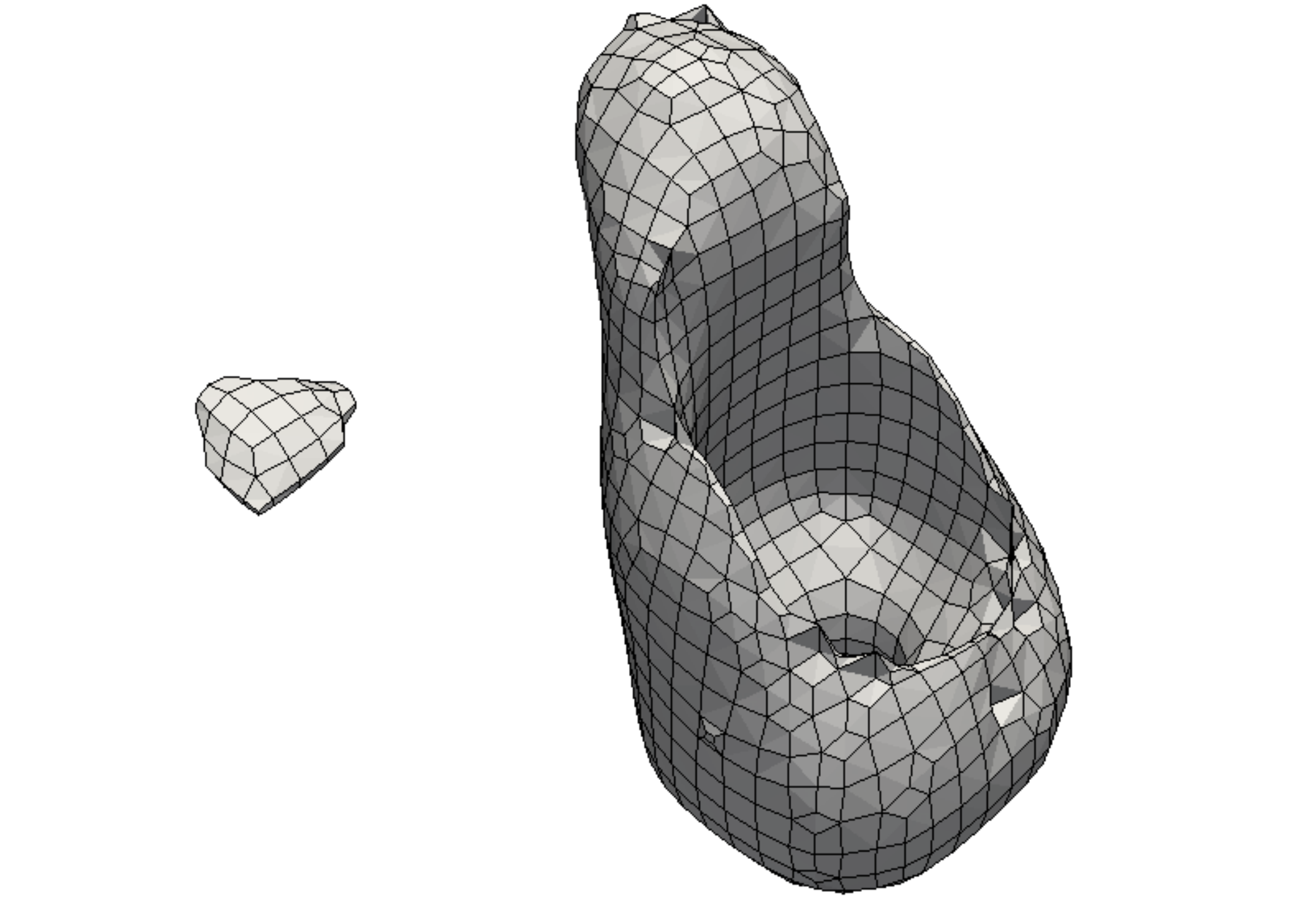}} & \parbox[m]{6em}{\includegraphics[trim={0cm 0cm 0cm 0cm},clip, width=0.12\textwidth]{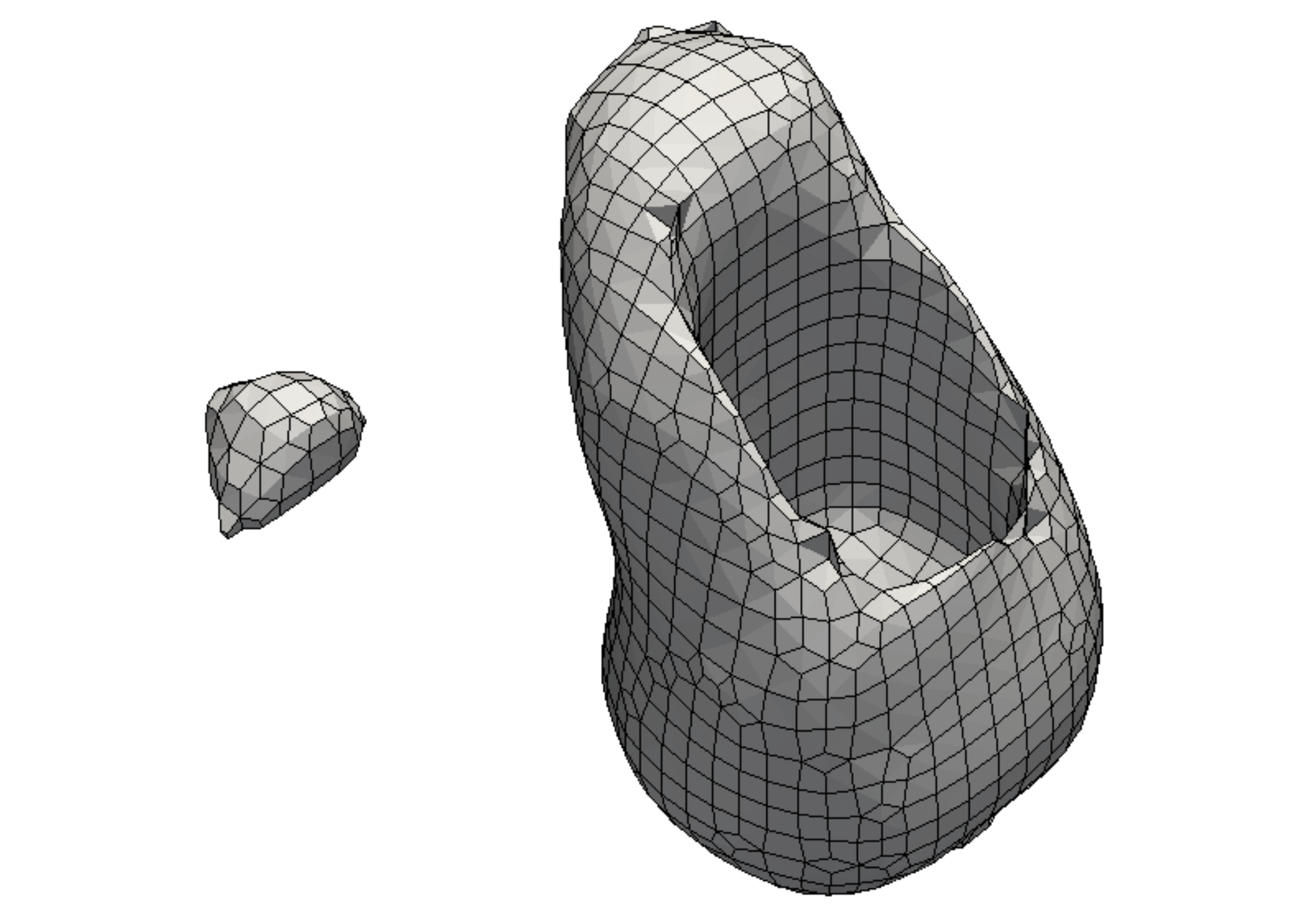}} & \parbox[m]{6em}{\includegraphics[trim={0cm 0cm 0cm 0cm},clip, width=0.12\textwidth]{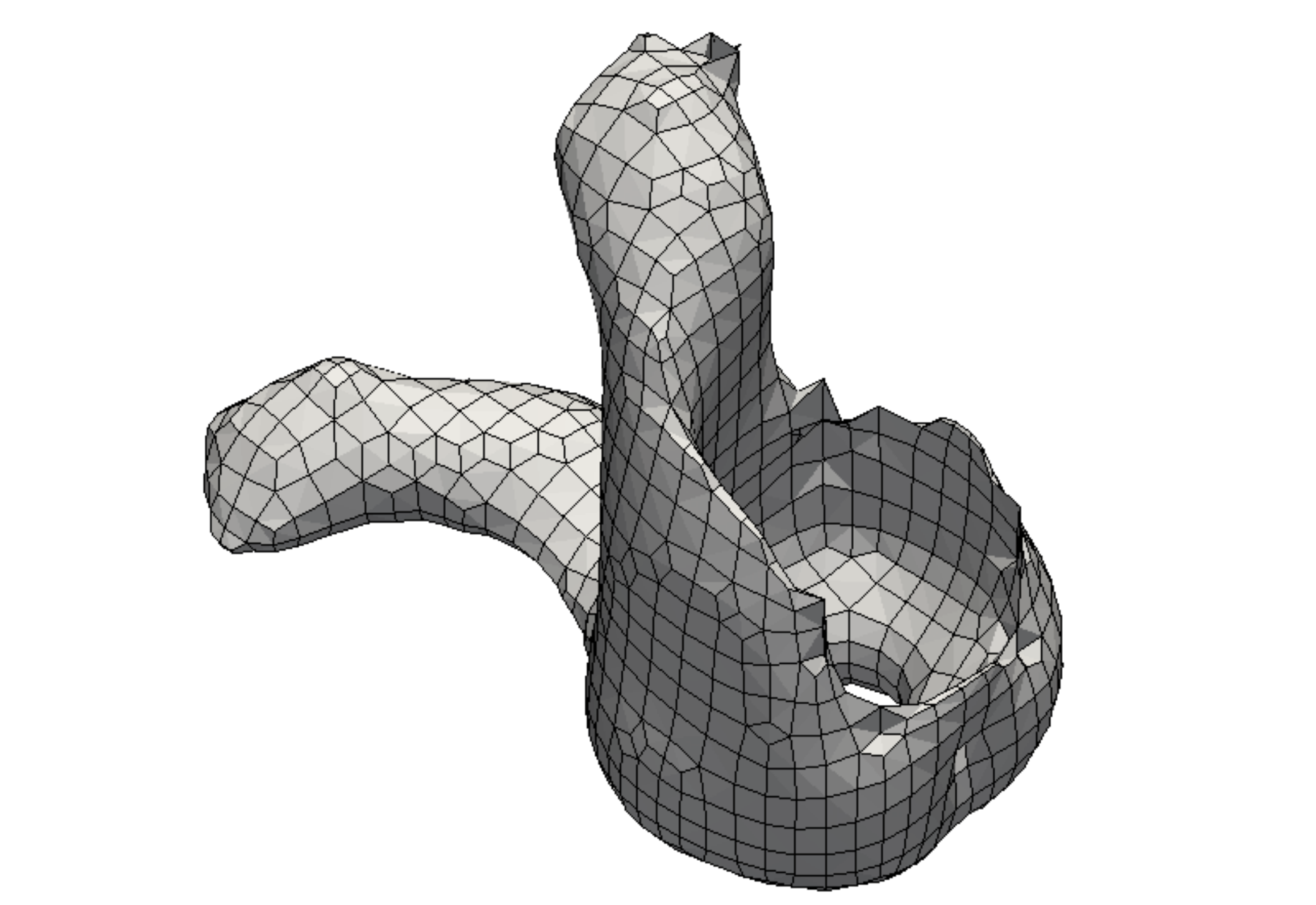}} & \parbox[m]{6em}{\includegraphics[trim={0cm 0cm 0cm 0cm},clip, width=0.12\textwidth]{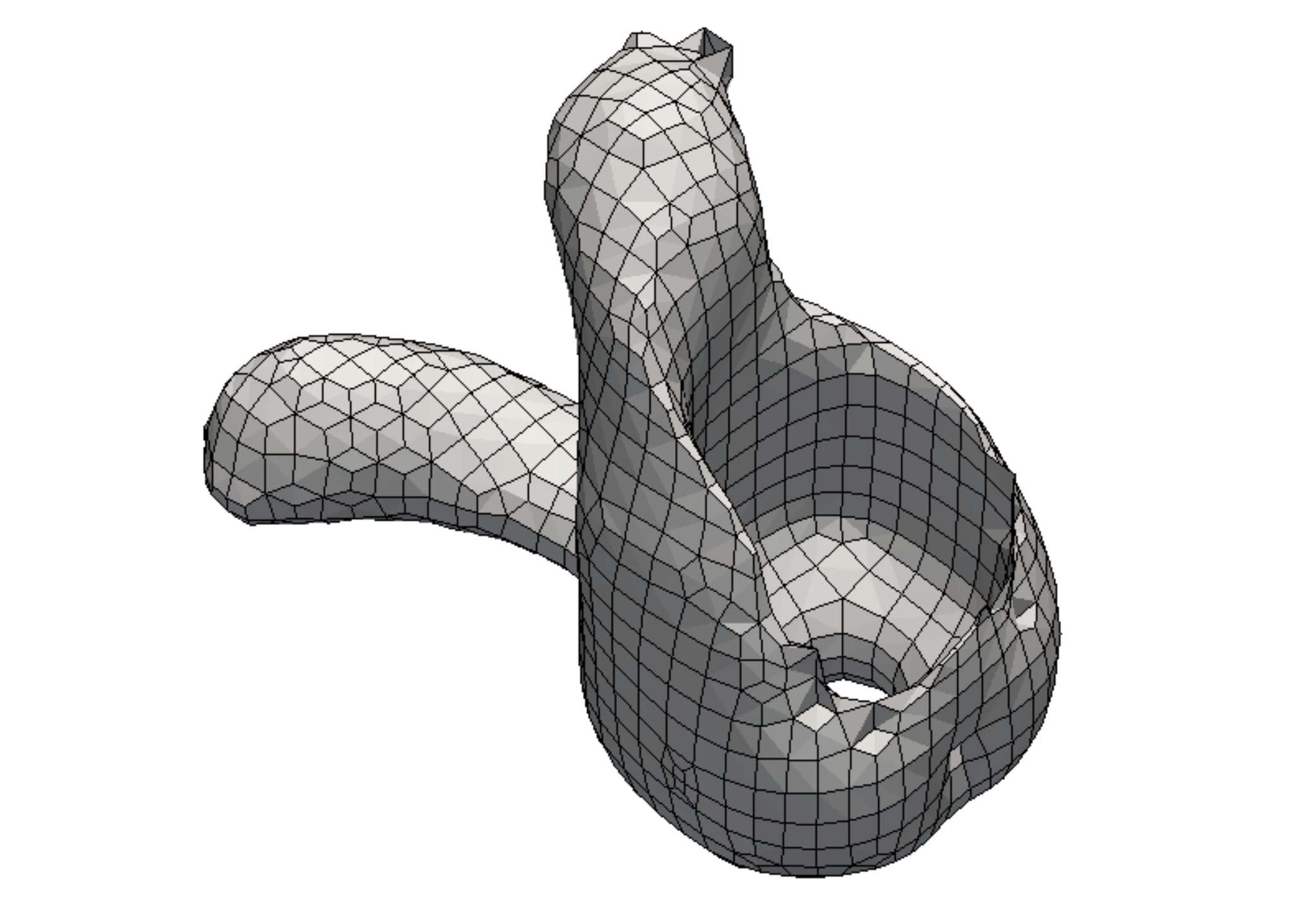}} & \parbox[m]{6em}{\includegraphics[trim={0cm 0cm 0cm 0cm},clip, width=0.12\textwidth]{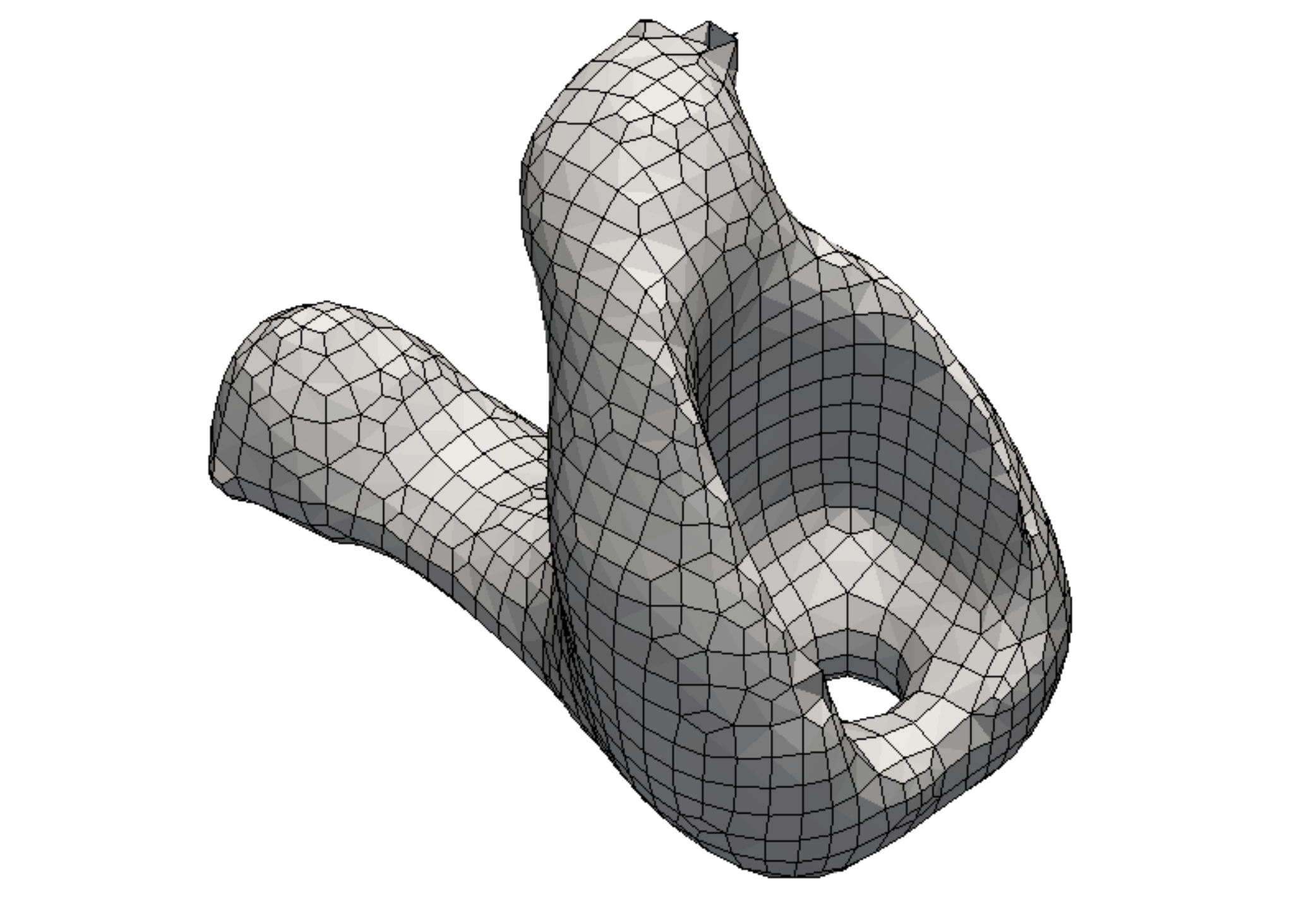}} \\\hline
     \multirow{2}{*}{\rotatebox{90}{trivial+PDE}} & & \parbox[m]{6em}{\includegraphics[trim={0cm 0cm 0cm 0cm},clip, width=0.12\textwidth]{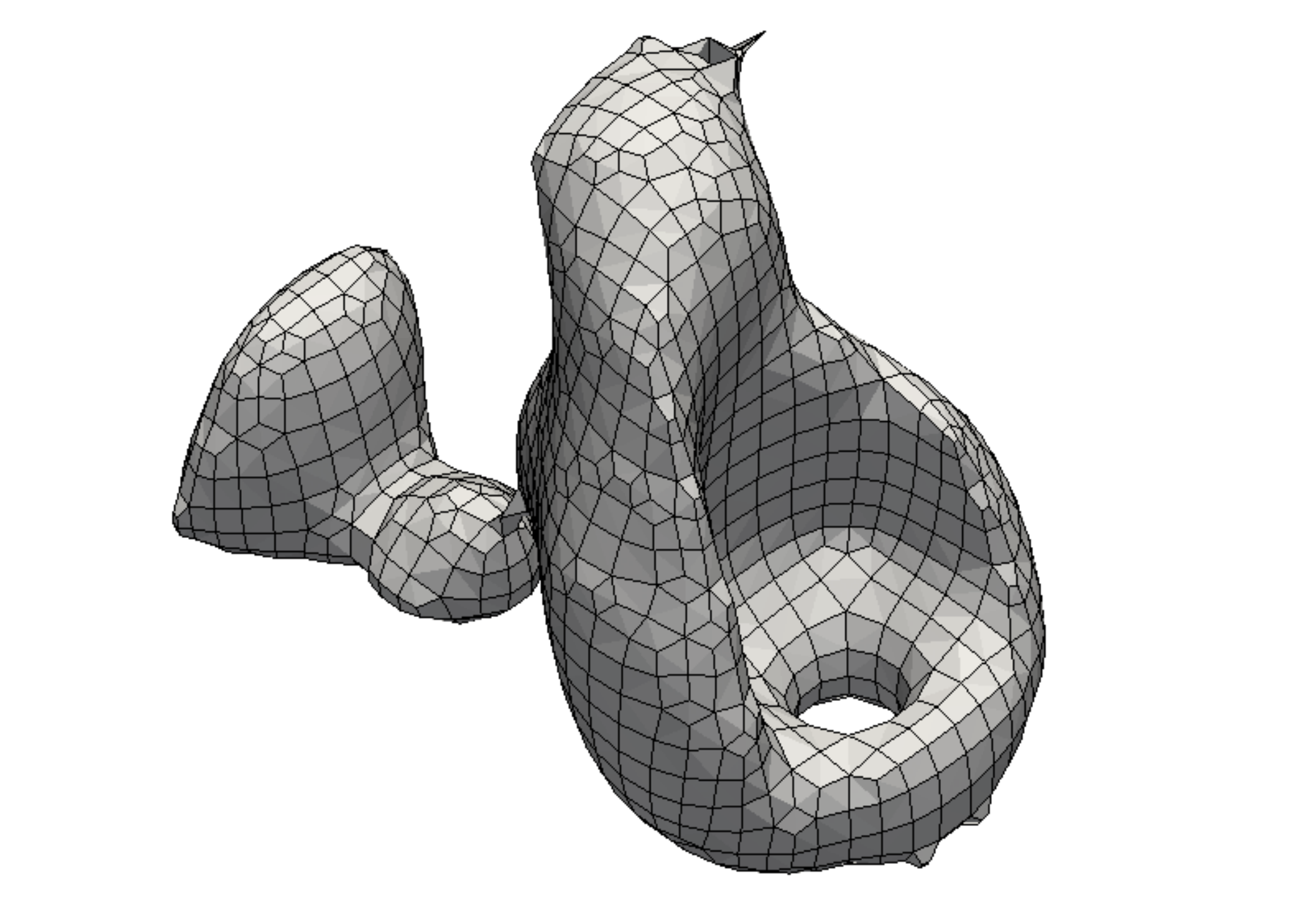}} & \parbox[m]{6em}{\includegraphics[trim={0cm 0cm 0cm 0cm},clip, width=0.12\textwidth]{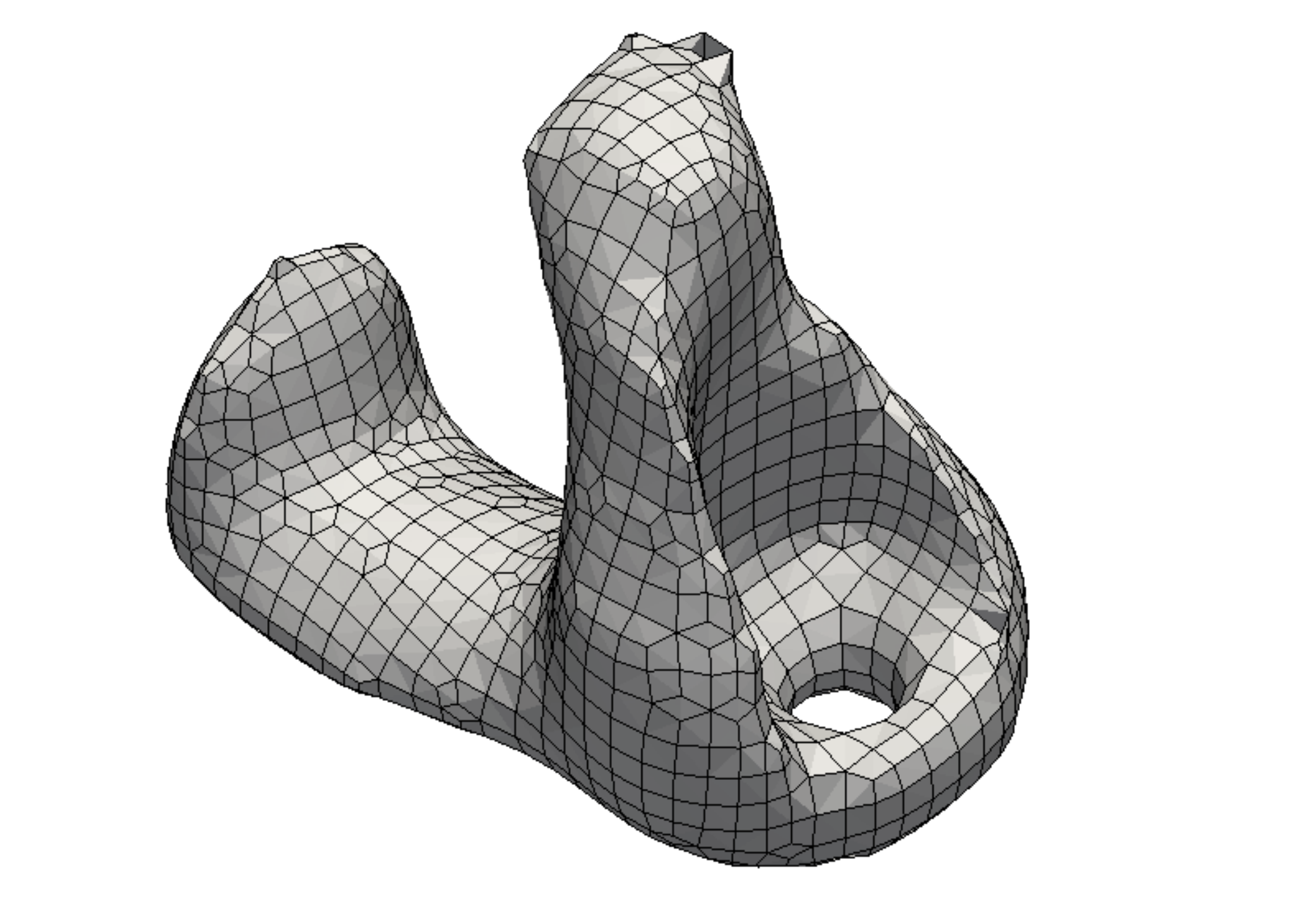}} & \parbox[m]{6em}{\includegraphics[trim={0cm 0cm 0cm 0cm},clip, width=0.12\textwidth]{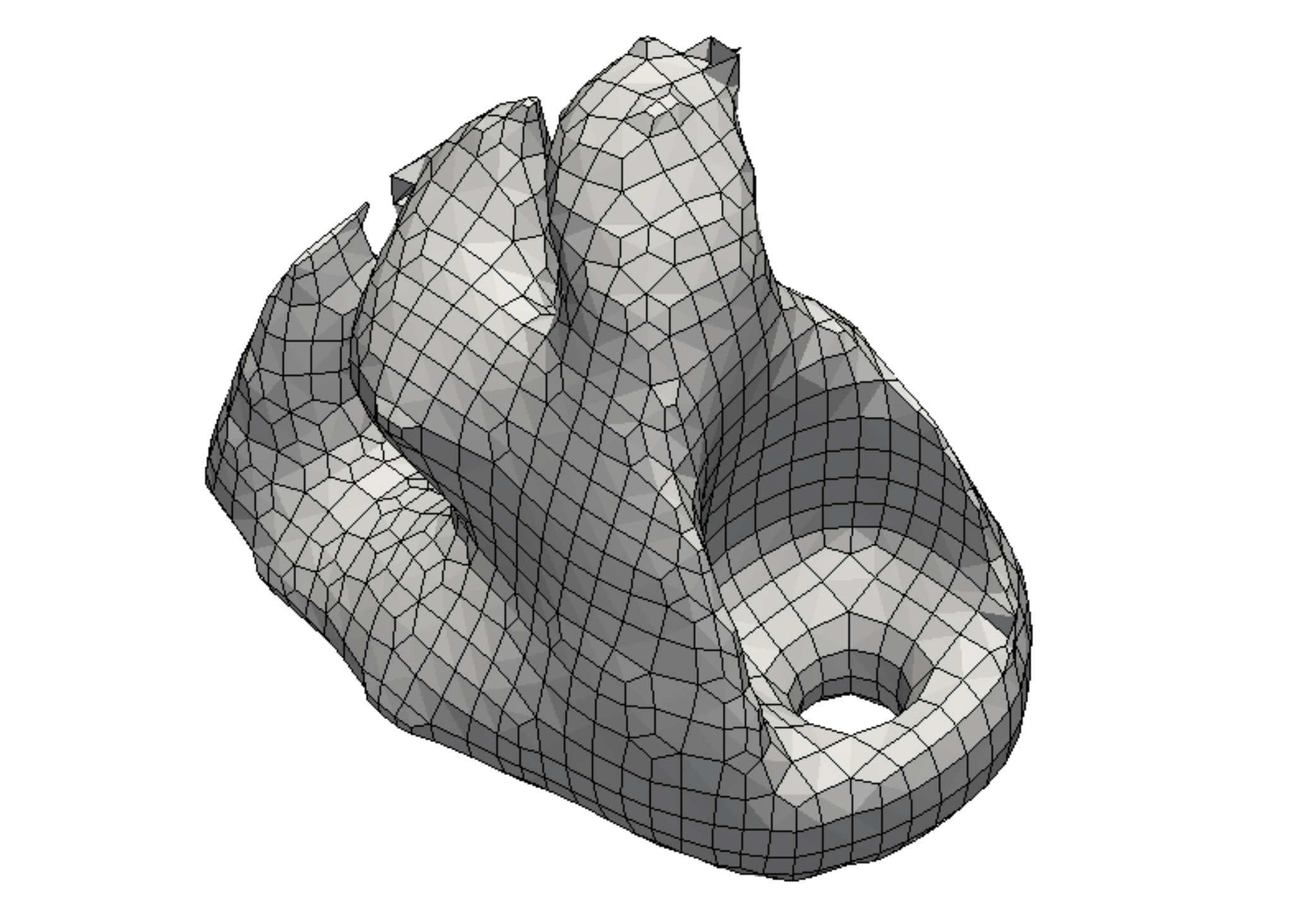}} & \parbox[m]{6em}{\includegraphics[trim={0cm 0cm 0cm 0cm},clip, width=0.12\textwidth]{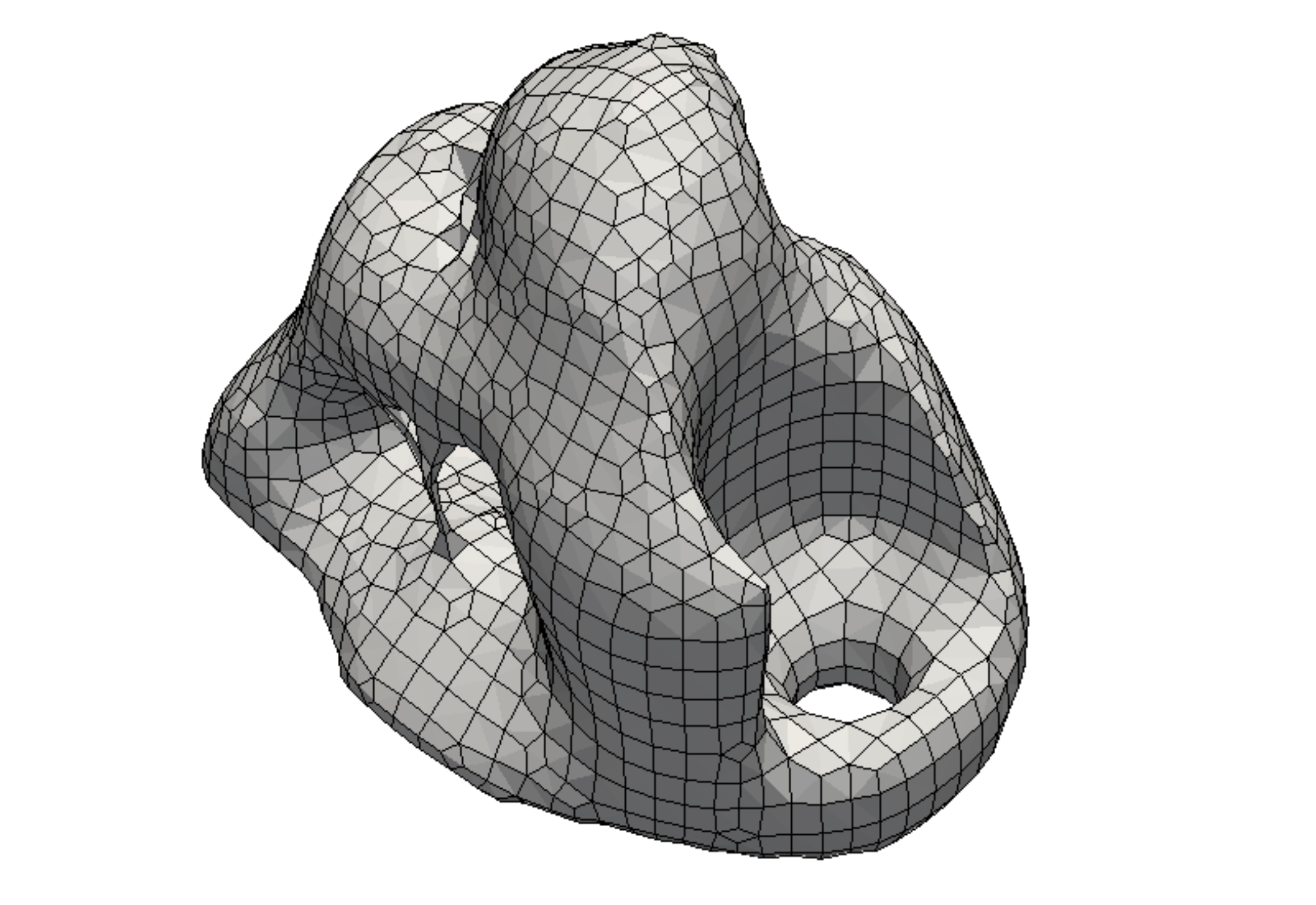}} & \parbox[m]{6em}{\includegraphics[trim={0cm 0cm 0cm 0cm},clip, width=0.12\textwidth]{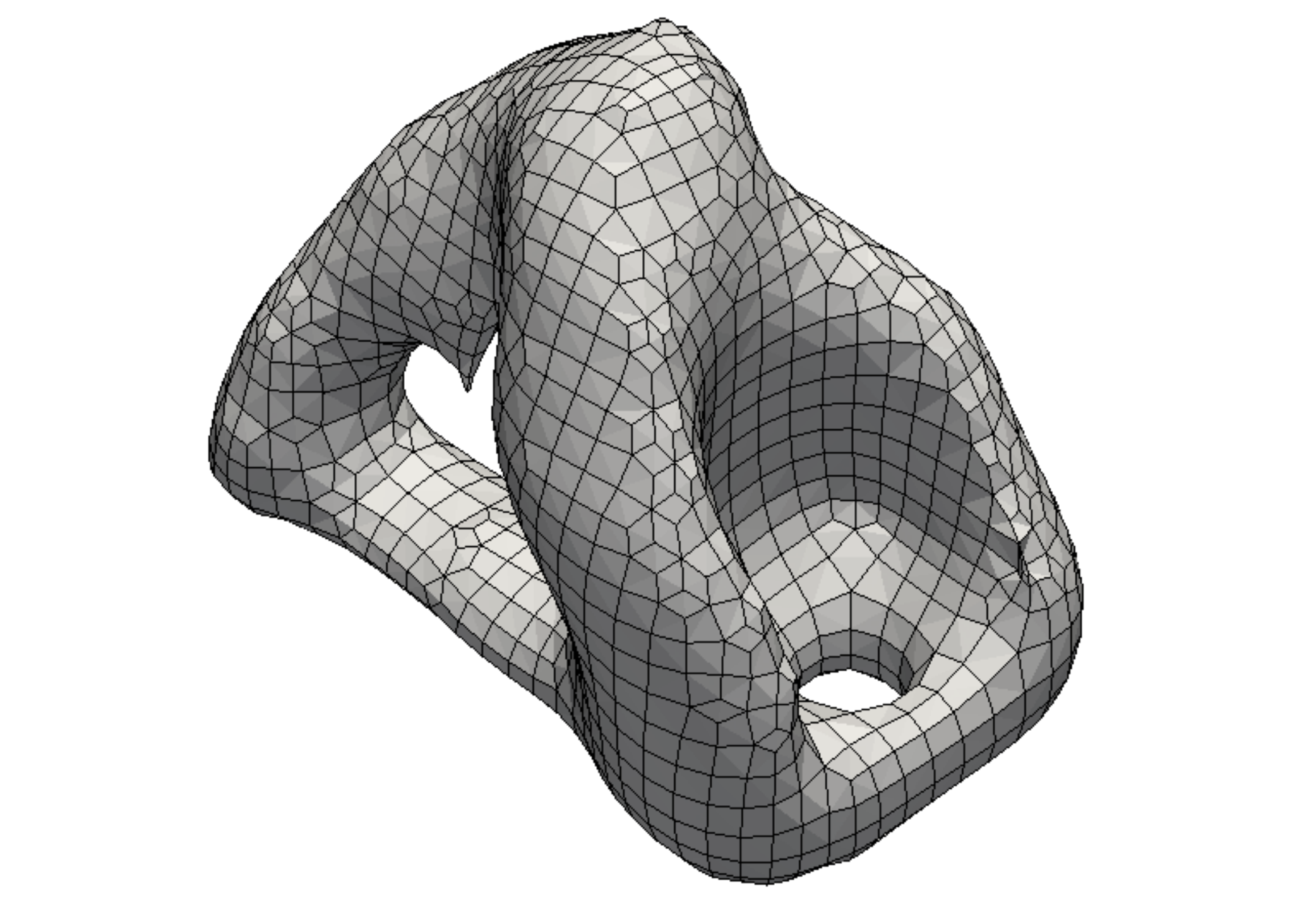}} \\\cline{2-7}
     & \checkmark & \parbox[m]{6em}{\includegraphics[trim={0cm 0cm 0cm 0cm},clip, width=0.12\textwidth]{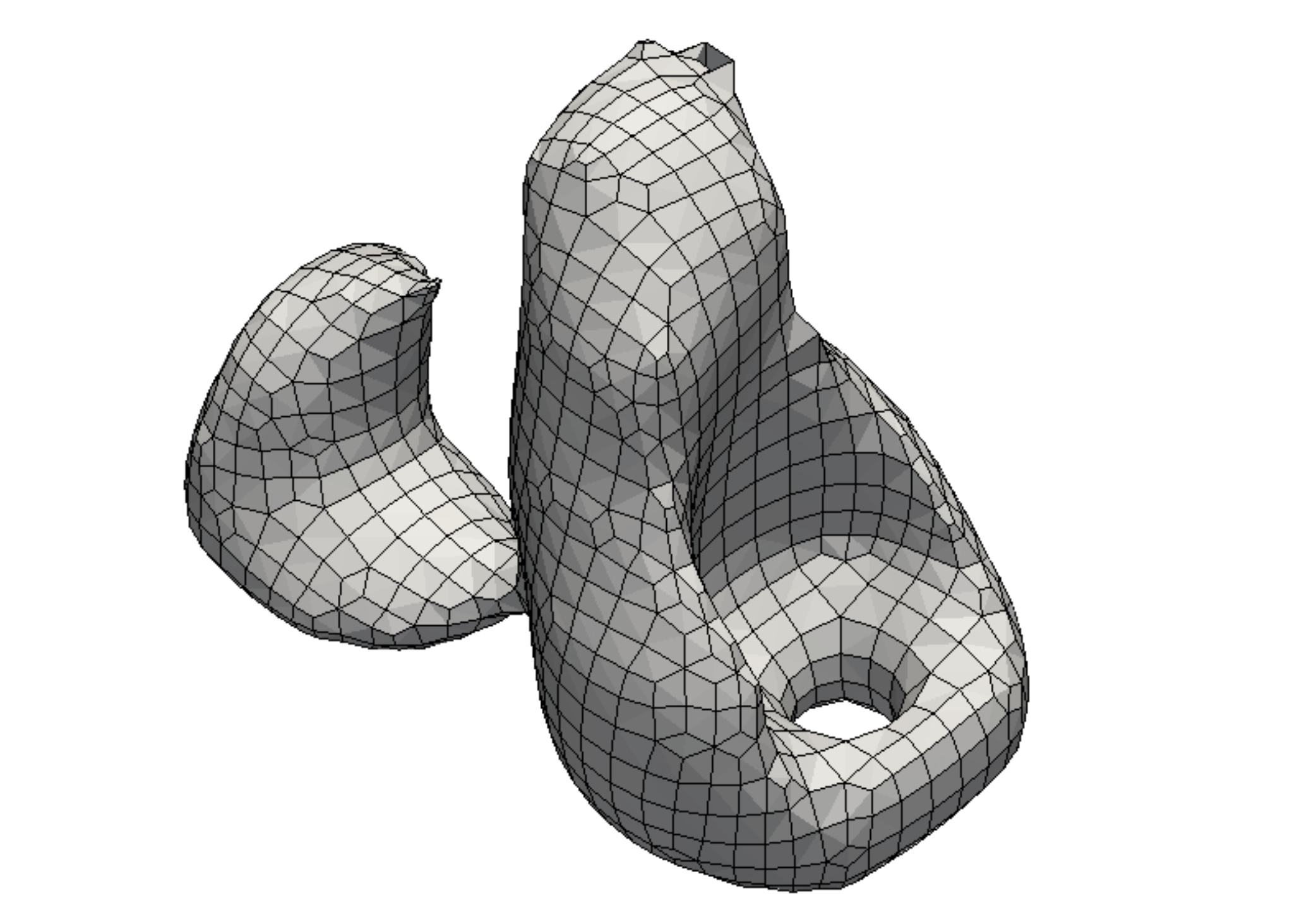}} & \parbox[m]{6em}{\includegraphics[trim={0cm 0cm 0cm 0cm},clip, width=0.12\textwidth]{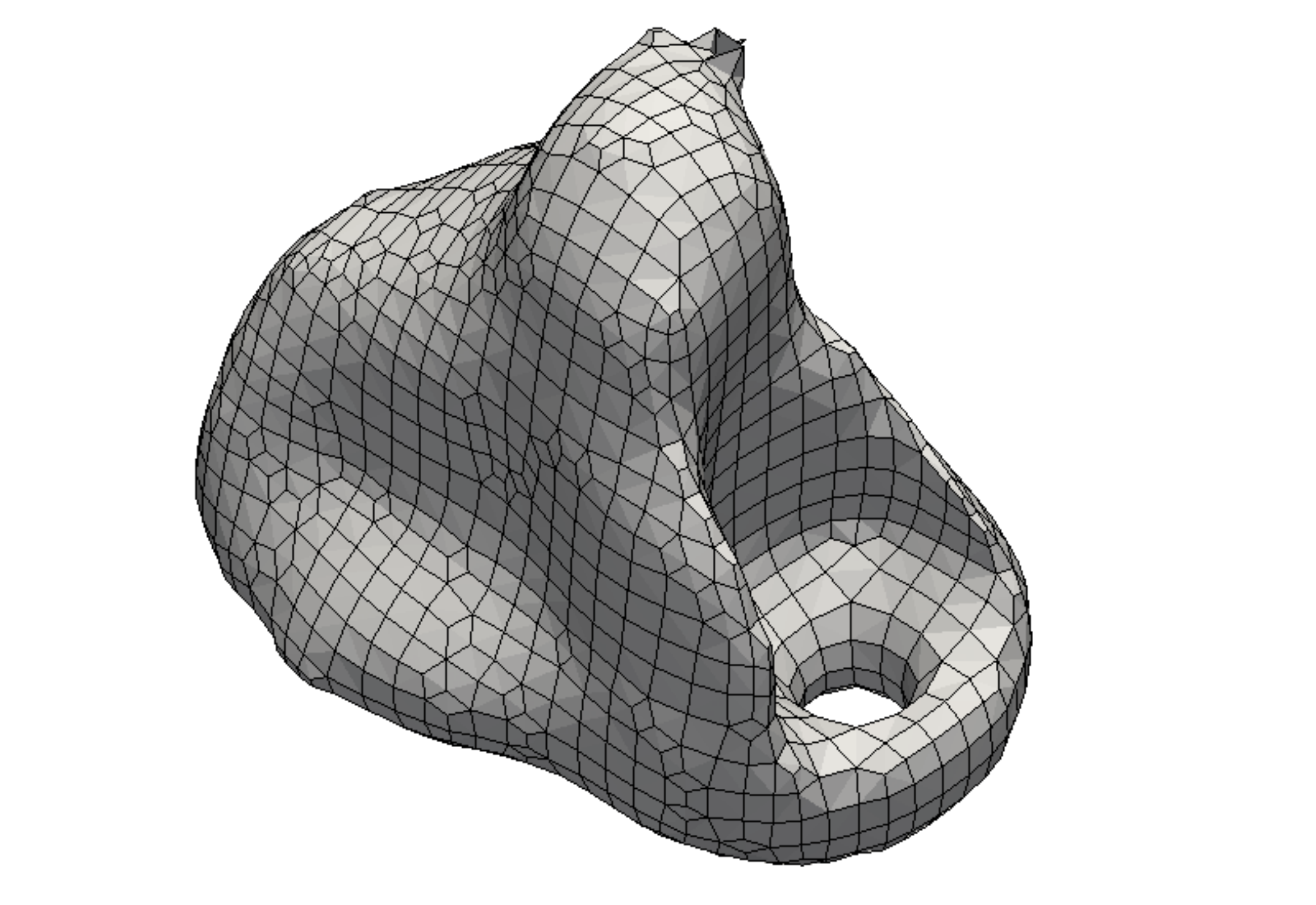}} & \parbox[m]{6em}{\includegraphics[trim={0cm 0cm 0cm 0cm},clip, width=0.12\textwidth]{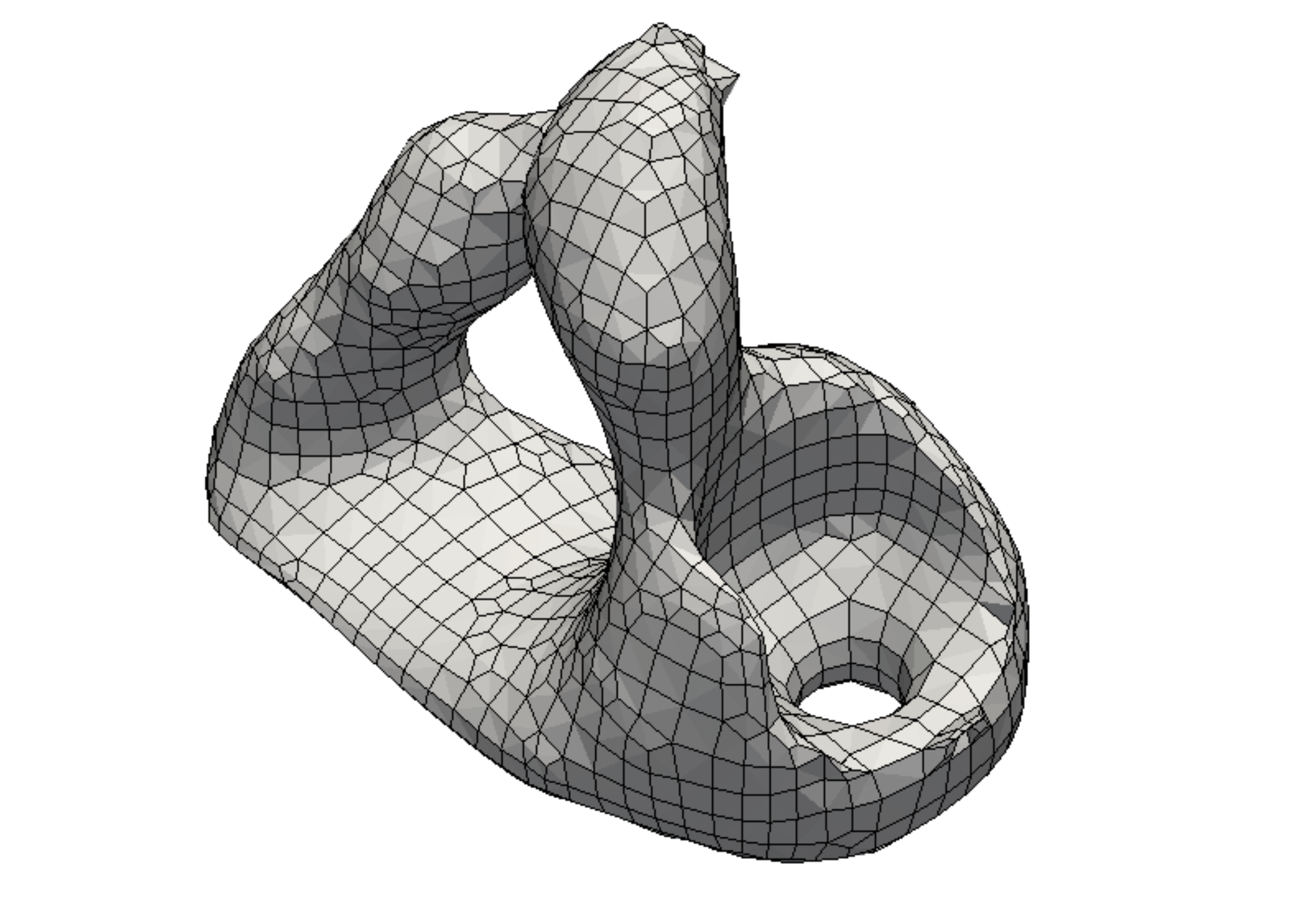}} & \parbox[m]{6em}{\includegraphics[trim={0cm 0cm 0cm 0cm},clip, width=0.12\textwidth]{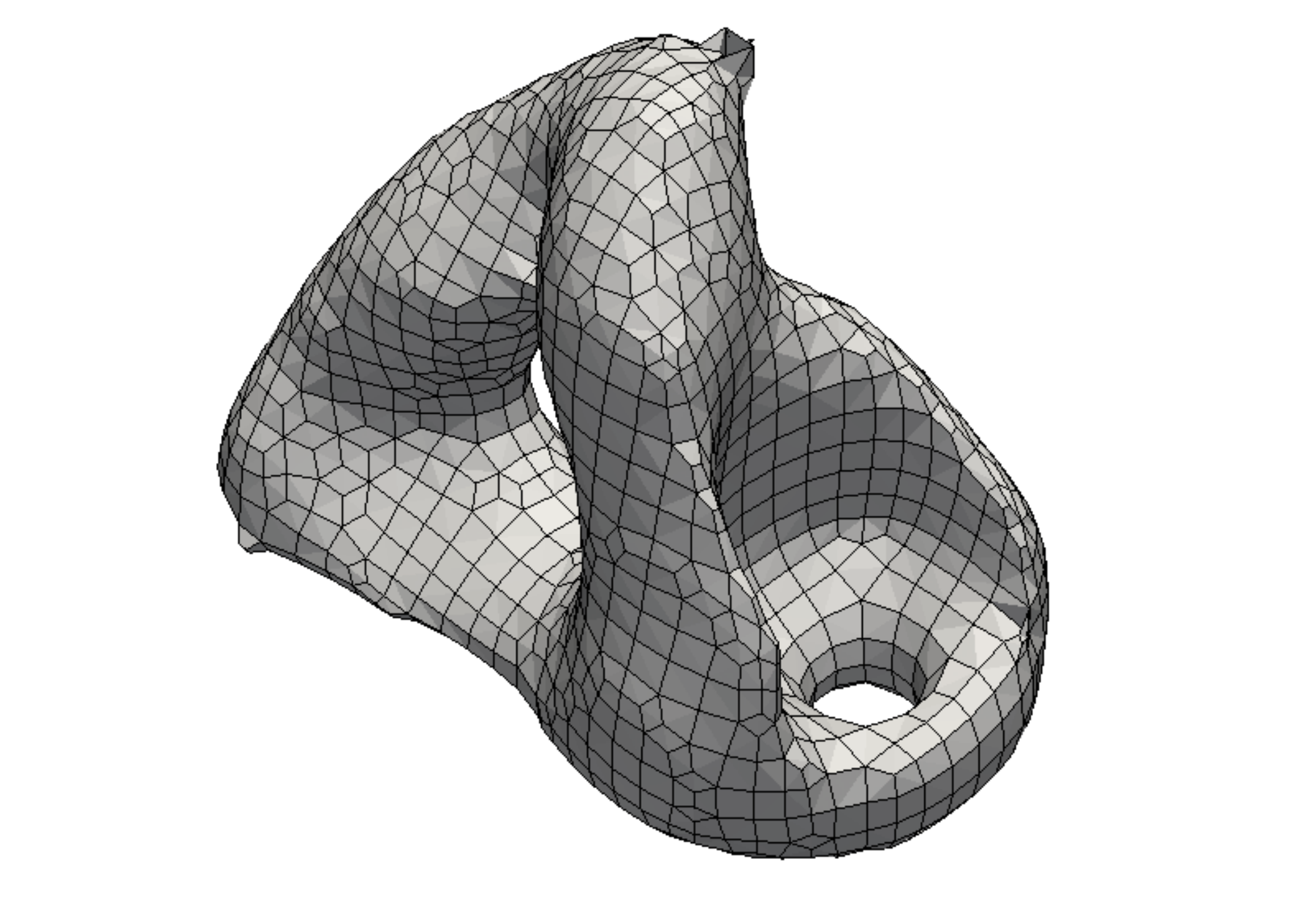}} & \parbox[m]{6em}{\includegraphics[trim={0cm 0cm 0cm 0cm},clip, width=0.12\textwidth]{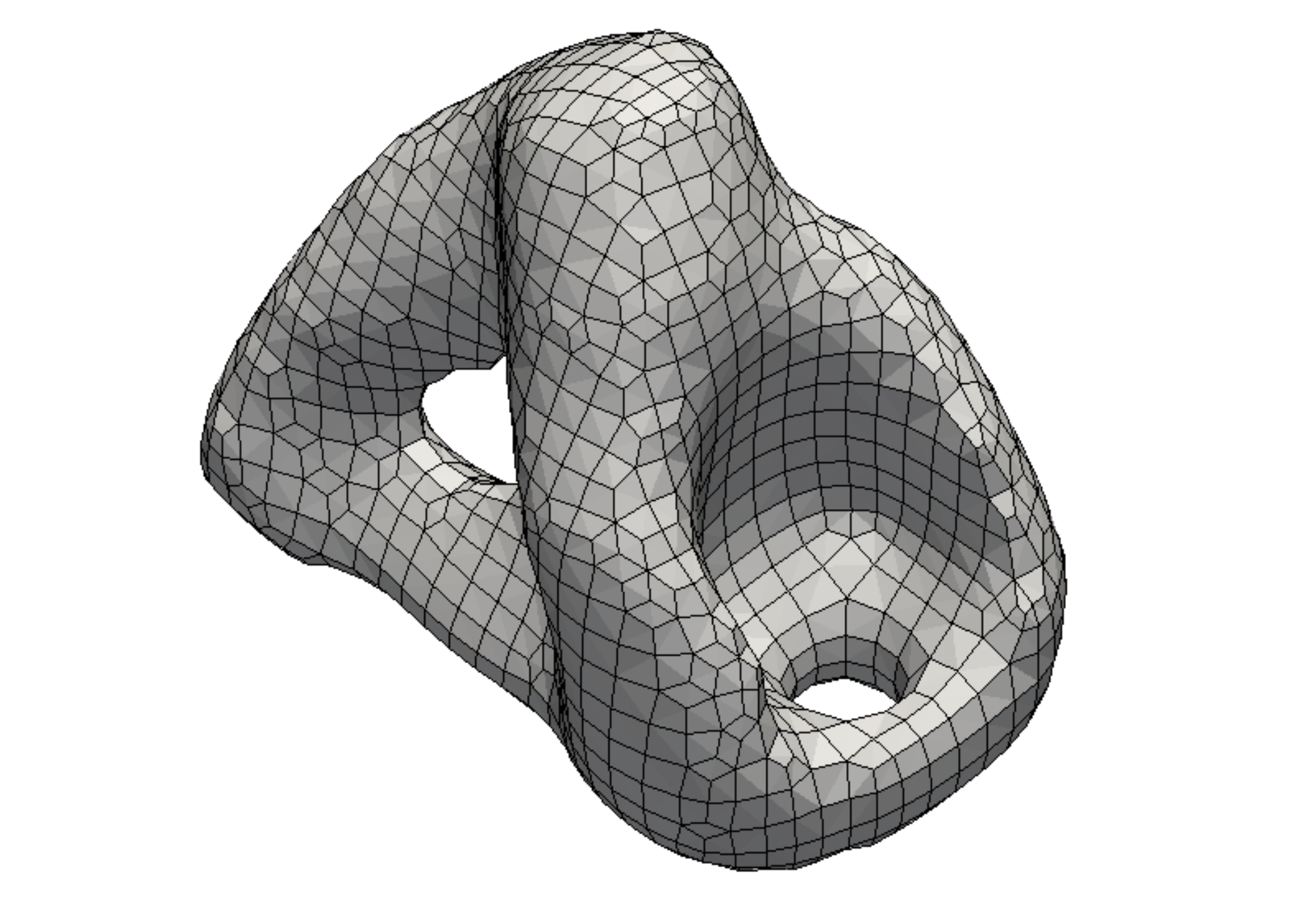}}\\
     \multicolumn{7}{c}{\fbox{\hspace{0.3cm}\parbox[m]{6em}{\includegraphics[trim={0cm 0cm 0cm 0cm},clip, width=0.12\textwidth]{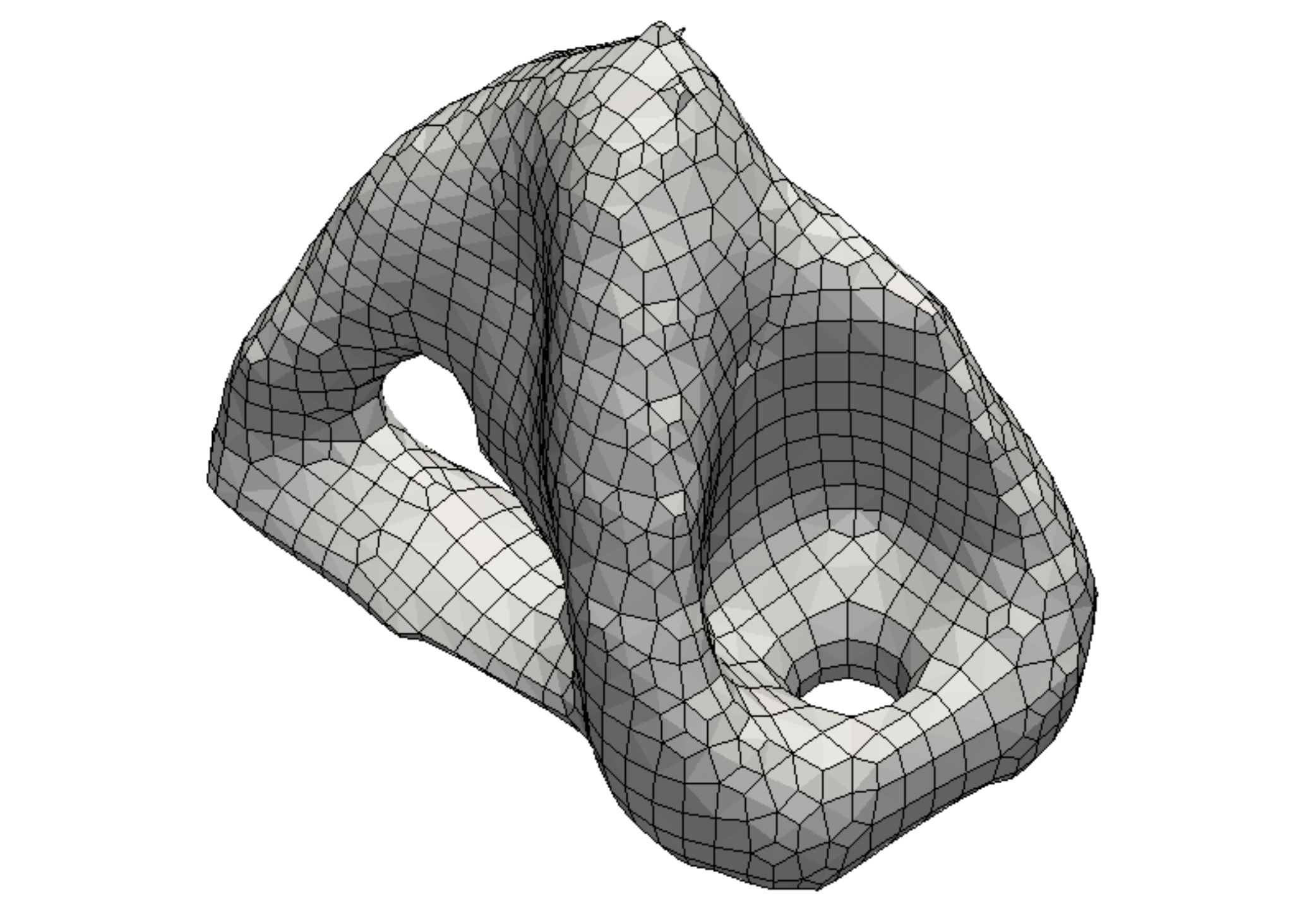}}}}
\end{tabular}
    \end{subtable}
\end{table}
\FloatBarrier
\FloatBarrier
\section*{Acknowledgments}
We thank Marco Gosch, Rielson Falck, Christian Knorr (ArianeGroup) and Daniel Siegel (Synera) for providing the datasets. We also thank Denis Regenbrecht and Susanne Heckrodt (DLR, German Aerospace Center) and Janek Gödeke (University of Bremen) for their support.

\bibliographystyle{siamplain}
\bibliography{ms}

\begin{thebibliography}{10}

\bibitem{aage2015topology}
{\sc N.~Aage, E.~Andreassen, and B.~S. Lazarov}, {\em Topology optimization
  using petsc: An easy-to-use, fully parallel, open source topology
  optimization framework}, Structural and Multidisciplinary Optimization, 51
  (2015), pp.~565--572.

\bibitem{abueidda2020topology}
{\sc D.~W. Abueidda, S.~Koric, and N.~A. Sobh}, {\em Topology optimization of
  2d structures with nonlinearities using deep learning}, Computers \&
  Structures, 237 (2020), p.~106283.

\bibitem{banga20183d}
{\sc S.~Banga, H.~Gehani, S.~Bhilare, S.~Patel, and L.~Kara}, {\em 3d topology
  optimization using convolutional neural networks}, arXiv preprint
  arXiv:1808.07440,  (2018).

\bibitem{bendsoe2003topology}
{\sc M.~P. Bendsoe and O.~Sigmund}, {\em Topology optimization: theory,
  methods, and applications}, Springer Science \& Business Media, 2003.

\bibitem{bengio2017deep}
{\sc Y.~Bengio, I.~Goodfellow, and A.~Courville}, {\em Deep learning}, vol.~1,
  MIT press Cambridge, MA, USA, 2017.

\bibitem{bogatskiy2020lorentz}
{\sc A.~Bogatskiy, B.~Anderson, J.~Offermann, M.~Roussi, D.~Miller, and
  R.~Kondor}, {\em Lorentz group equivariant neural network for particle
  physics}, in International Conference on Machine Learning, PMLR, 2020,
  pp.~992--1002.

\bibitem{borrvall2003topology}
{\sc T.~Borrvall and J.~Petersson}, {\em Topology optimization of fluids in
  stokes flow}, International journal for numerical methods in fluids, 41
  (2003), pp.~77--107.

\bibitem{cang2019one}
{\sc R.~Cang, H.~Yao, and Y.~Ren}, {\em One-shot generation of near-optimal
  topology through theory-driven machine learning}, Computer-Aided Design, 109
  (2019), pp.~12--21.

\bibitem{chandrasekhar2021tounn}
{\sc A.~Chandrasekhar and K.~Suresh}, {\em Tounn: topology optimization using
  neural networks}, Structural and Multidisciplinary Optimization, 63 (2021),
  pp.~1135--1149.

\bibitem{chen2021nerv}
{\sc H.~Chen, B.~He, H.~Wang, Y.~Ren, S.~N. Lim, and A.~Shrivastava}, {\em
  Nerv: Neural representations for videos}, Advances in Neural Information
  Processing Systems, 34 (2021), pp.~21557--21568.

\bibitem{chi2021universal}
{\sc H.~Chi, Y.~Zhang, T.~L.~E. Tang, L.~Mirabella, L.~Dalloro, L.~Song, and
  G.~H. Paulino}, {\em Universal machine learning for topology optimization},
  Computer Methods in Applied Mechanics and Engineering, 375 (2021), p.~112739.

\bibitem{cciccek20163d}
{\sc {\"O}.~{\c{C}}i{\c{c}}ek, A.~Abdulkadir, S.~S. Lienkamp, T.~Brox, and
  O.~Ronneberger}, {\em 3d u-net: learning dense volumetric segmentation from
  sparse annotation}, in International conference on medical image computing
  and computer-assisted intervention, Springer, 2016, pp.~424--432.

\bibitem{cohen2016group}
{\sc T.~Cohen and M.~Welling}, {\em Group equivariant convolutional networks},
  in International conference on machine learning, PMLR, 2016, pp.~2990--2999.

\bibitem{dede2009multiphysics}
{\sc E.~M. Dede}, {\em Multiphysics topology optimization of heat transfer and
  fluid flow systems}, 715 (2009).

\bibitem{deng2020topology}
{\sc H.~Deng and A.~C. To}, {\em Topology optimization based on deep
  representation learning (drl) for compliance and stress-constrained design},
  Computational Mechanics, 66 (2020), pp.~449--469.

\bibitem{selto_dataset}
{\sc S.~Dittmer, D.~Erzmann, H.~Harms, R.~Falck, and M.~Gosch}, {\em Selto
  dataset}.
\newblock \url{https://doi.org/10.5281/zenodo.7034898}, 2023.

\bibitem{duhring2008acoustic}
{\sc M.~B. D{\"u}hring, J.~S. Jensen, and O.~Sigmund}, {\em Acoustic design by
  topology optimization}, Journal of sound and vibration, 317 (2008),
  pp.~557--575.

\bibitem{dumont2018robustness}
{\sc B.~Dumont, S.~Maggio, and P.~Montalvo}, {\em Robustness of
  rotation-equivariant networks to adversarial perturbations}, arXiv preprint
  arXiv:1802.06627,  (2018).

\bibitem{eschenauer2001topology}
{\sc H.~A. Eschenauer and N.~Olhoff}, {\em Topology optimization of continuum
  structures: a review}, Appl. Mech. Rev., 54 (2001), pp.~331--390.

\bibitem{gerken2021geometric}
{\sc J.~E. Gerken, J.~Aronsson, O.~Carlsson, H.~Linander, F.~Ohlsson,
  C.~Petersson, and D.~Persson}, {\em Geometric deep learning and equivariant
  neural networks}, arXiv preprint arXiv:2105.13926,  (2021).

\bibitem{goodfellow2014generative}
{\sc I.~Goodfellow, J.~Pouget-Abadie, M.~Mirza, B.~Xu, D.~Warde-Farley,
  S.~Ozair, A.~Courville, and Y.~Bengio}, {\em Generative adversarial nets},
  Advances in neural information processing systems, 27 (2014).

\bibitem{goodhart1975problems}
{\sc C.~Goodhart}, {\em Problems of monetary management: the uk experience in
  papers in monetary economics}, Monetary Economics, 1 (1975).

\bibitem{hoyer2019neural}
{\sc S.~Hoyer, J.~Sohl-Dickstein, and S.~Greydanus}, {\em Neural
  reparameterization improves structural optimization}, arXiv preprint
  arXiv:1909.04240,  (2019).

\bibitem{ioffe2015batch}
{\sc S.~Ioffe and C.~Szegedy}, {\em Batch normalization: Accelerating deep
  network training by reducing internal covariate shift}, in International
  conference on machine learning, PMLR, 2015, pp.~448--456.

\bibitem{jensen2021topology}
{\sc P.~D.~L. Jensen, F.~Wang, I.~Dimino, and O.~Sigmund}, {\em Topology
  optimization of large-scale 3d morphing wing structures}, in Actuators,
  vol.~10, MDPI, 2021, p.~217.

\bibitem{kawaguchi2017generalization}
{\sc K.~Kawaguchi, L.~P. Kaelbling, and Y.~Bengio}, {\em Generalization in deep
  learning}, arXiv preprint arXiv:1710.05468,  (2017).

\bibitem{kingma2014adam}
{\sc D.~P. Kingma and J.~Ba}, {\em Adam: A method for stochastic optimization},
  arXiv preprint arXiv:1412.6980,  (2014).

\bibitem{lee2020cnn}
{\sc S.~Lee, H.~Kim, Q.~X. Lieu, and J.~Lee}, {\em Cnn-based image recognition
  for topology optimization}, Knowledge-Based Systems, 198 (2020), p.~105887.

\bibitem{ma2020segmentation}
{\sc J.~Ma}, {\em Segmentation loss odyssey}, arXiv preprint arXiv:2005.13449,
  (2020).

\bibitem{mrowca2018flexible}
{\sc D.~Mrowca, C.~Zhuang, E.~Wang, N.~Haber, L.~F. Fei-Fei, J.~Tenenbaum, and
  D.~L. Yamins}, {\em Flexible neural representation for physics prediction},
  Advances in neural information processing systems, 31 (2018).

\bibitem{murphy2018janossy}
{\sc R.~L. Murphy, B.~Srinivasan, V.~Rao, and B.~Ribeiro}, {\em Janossy
  pooling: Learning deep permutation-invariant functions for variable-size
  inputs}, arXiv preprint arXiv:1811.01900,  (2018).

\bibitem{nie2021topologygan}
{\sc Z.~Nie, T.~Lin, H.~Jiang, and L.~B. Kara}, {\em Topologygan: Topology
  optimization using generative adversarial networks based on physical fields
  over the initial domain}, Journal of Mechanical Design, 143 (2021).

\bibitem{oh2019deep}
{\sc S.~Oh, Y.~Jung, S.~Kim, I.~Lee, and N.~Kang}, {\em Deep generative design:
  Integration of topology optimization and generative models}, Journal of
  Mechanical Design, 141 (2019).

\bibitem{paszke2017automatic}
{\sc A.~Paszke, S.~Gross, S.~Chintala, G.~Chanan, E.~Yang, Z.~DeVito, Z.~Lin,
  A.~Desmaison, L.~Antiga, and A.~Lerer}, {\em Automatic differentiation in
  pytorch},  (2017).

\bibitem{puny2021frame}
{\sc O.~Puny, M.~Atzmon, H.~Ben-Hamu, E.~J. Smith, I.~Misra, A.~Grover, and
  Y.~Lipman}, {\em Frame averaging for invariant and equivariant network
  design}, arXiv preprint arXiv:2110.03336,  (2021).

\bibitem{perezgarcia2020}
{\sc F.~Pérez-García}, {\em Pytorch implementation of 2d and 3d u-net
  (v0.7.5)}.
\newblock \url{https://github.com/fepegar/unet}, 2020.

\bibitem{qian2021accelerating}
{\sc C.~Qian and W.~Ye}, {\em Accelerating gradient-based topology optimization
  design with dual-model artificial neural networks}, Structural and
  Multidisciplinary Optimization, 63 (2021), pp.~1687--1707.

\bibitem{rade2020physics}
{\sc J.~Rade, A.~Balu, E.~Herron, J.~Pathak, R.~Ranade, S.~Sarkar, and
  A.~Krishnamurthy}, {\em Physics-consistent deep learning for structural
  topology optimization}, arXiv preprint arXiv:2012.05359,  (2020).

\bibitem{raissi2019physics}
{\sc M.~Raissi, P.~Perdikaris, and G.~E. Karniadakis}, {\em Physics-informed
  neural networks: A deep learning framework for solving forward and inverse
  problems involving nonlinear partial differential equations}, Journal of
  Computational physics, 378 (2019), pp.~686--707.

\bibitem{rawat2019novel}
{\sc S.~Rawat and M.-H.~H. Shen}, {\em A novel topology optimization approach
  using conditional deep learning}, arXiv preprint arXiv:1901.04859,  (2019).

\bibitem{ronneberger2015u}
{\sc O.~Ronneberger, P.~Fischer, and T.~Brox}, {\em U-net: Convolutional
  networks for biomedical image segmentation}, in International Conference on
  Medical image computing and computer-assisted intervention, Springer, 2015,
  pp.~234--241.

\bibitem{shen2019new}
{\sc M.-H.~H. Shen and L.~Chen}, {\em A new cgan technique for constrained
  topology design optimization}, arXiv preprint arXiv:1901.07675,  (2019).

\bibitem{sosnovik2019neural}
{\sc I.~Sosnovik and I.~Oseledets}, {\em Neural networks for topology
  optimization}, Russian Journal of Numerical Analysis and Mathematical
  Modelling, 34 (2019), pp.~215--223.

\bibitem{taubin1995curve}
{\sc G.~Taubin}, {\em Curve and surface smoothing without shrinkage}, in
  Proceedings of IEEE international conference on computer vision, IEEE, 1995,
  pp.~852--857.

\bibitem{thomas2018tensor}
{\sc N.~Thomas, T.~Smidt, S.~Kearnes, L.~Yang, L.~Li, K.~Kohlhoff, and
  P.~Riley}, {\em Tensor field networks: Rotation-and translation-equivariant
  neural networks for 3d point clouds}, arXiv preprint arXiv:1802.08219,
  (2018).

\bibitem{wang2021deep}
{\sc C.~Wang, S.~Yao, Z.~Wang, and J.~Hu}, {\em Deep super-resolution neural
  network for structural topology optimization}, Engineering Optimization, 53
  (2021), pp.~2108--2121.

\bibitem{weiler20183d}
{\sc M.~Weiler, M.~Geiger, M.~Welling, W.~Boomsma, and T.~S. Cohen}, {\em 3d
  steerable cnns: Learning rotationally equivariant features in volumetric
  data}, Advances in Neural Information Processing Systems, 31 (2018).

\bibitem{weiler2018learning}
{\sc M.~Weiler, F.~A. Hamprecht, and M.~Storath}, {\em Learning steerable
  filters for rotation equivariant cnns},  (2018), pp.~849--858.

\bibitem{xue2021efficient}
{\sc L.~Xue, J.~Liu, G.~Wen, and H.~Wang}, {\em Efficient, high-resolution
  topology optimization method based on convolutional neural networks},
  Frontiers of Mechanical Engineering, 16 (2021), pp.~80--96.

\bibitem{yoon2007topology}
{\sc G.~H. Yoon, J.~S. Jensen, and O.~Sigmund}, {\em Topology optimization of
  acoustic--structure interaction problems using a mixed finite element
  formulation}, International journal for numerical methods in engineering, 70
  (2007), pp.~1049--1075.

\bibitem{yu2019deep}
{\sc Y.~Yu, T.~Hur, J.~Jung, and I.~G. Jang}, {\em Deep learning for
  determining a near-optimal topological design without any iteration},
  Structural and Multidisciplinary Optimization, 59 (2019), pp.~787--799.

\bibitem{zehnder2021ntopo}
{\sc J.~Zehnder, Y.~Li, S.~Coros, and B.~Thomaszewski}, {\em Ntopo: Mesh-free
  topology optimization using implicit neural representations}, Advances in
  Neural Information Processing Systems, 34 (2021), pp.~10368--10381.

\bibitem{zhang2019deep}
{\sc Y.~Zhang, B.~Peng, X.~Zhou, C.~Xiang, and D.~Wang}, {\em A deep
  convolutional neural network for topology optimization with strong
  generalization ability}, arXiv preprint arXiv:1901.07761,  (2019).

\bibitem{zhang2021tonr}
{\sc Z.~Zhang, Y.~Li, W.~Zhou, X.~Chen, W.~Yao, and Y.~Zhao}, {\em Tonr: An
  exploration for a novel way combining neural network with topology
  optimization}, Computer Methods in Applied Mechanics and Engineering, 386
  (2021), p.~114083.

\bibitem{zhang2018road}
{\sc Z.~Zhang, Q.~Liu, and Y.~Wang}, {\em Road extraction by deep residual
  u-net}, IEEE Geoscience and Remote Sensing Letters, 15 (2018), pp.~749--753.

\end{thebibliography}

\newpage
\FloatBarrier
\appendix
\section{Network architecture and training details}
\label{appendix_a_network}

Our UNet architecture is inspired by {\c{C}}i{\c{c}}ek et al.\cite{cciccek20163d} and based on the implementation from Pérez-García\cite{perezgarcia2020}. Like in {\c{C}}i{\c{c}}ek et al.\cite{cciccek20163d}, each layer of our encoder contains two $3\times 3\times 3$ convolutions, each followed by batch normalization~\cite{ioffe2015batch} and a rectified linear unit (ReLU) activation function, and then a max pooling operation. For the sphere dataset, we choose a pooling kernel size of $2\times 2\times 2$; for the disc dataset, we choose $2\times 2\times 1$ to account for the lower number of voxels in the $z$-direction. In the decoding path, each layer consists of a nearest neighbor upsampling step, followed by two $3 \times 3 \times 3$ convolutions with batch normalization and ReLU activation functions. In the last layer, a $1\times 1\times 1$ convolution with a sigmoid activation function reduces the number of output channels to $1$ and acts as a binary voxel-wise classifier. We use skip connections to pass information from layers of equal resolution in the encoding path to the decoder and apply a dropout rate of $0.3$ to reduce overfitting. We use five encoding and four decoding blocks. Each encoding block doubles the number of channels, starting with $10$ output channels for the first block. We choose the padding, so the image resolution stays unaffected by the convolutions. Based on memory availability, we set the batch size to $128$ when training on the disc dataset and $28$ when training on the sphere dataset. All models have been implemented in PyTorch~\cite{paszke2017automatic} with a total parameter count of approximately $1.6$M. For training we used an Nvidia GeForce GTX 1080 Ti GPU with 11GB VRAM. For the number of training epochs and the approximate training times for different models trained on the largest training subset, see \cref{7_tab:nr_parameters}.

\begin{table}[h]
    \caption{The number of training epochs and the approximate training time for UNets with different preprocessings and equivariances, trained on the largest training subsets of the disc and sphere combined datasets.}
    \label{7_tab:nr_parameters}
    \centering
    \begin{subtable}{\textwidth}
    \centering
    \begin{tabular}{|c|c||c|c|}
    \hline
        preprocessing & equivariance & training epochs & training time (in minutes)\\ \hline
        trivial & & $210$ & $12$\\
        trivial & \checkmark & $120$ & $34$\\
        trivial+PDE & & $230$ & $14$ \\
        trivial+PDE & \checkmark & $110$ & $32$
        \\\hline
    \end{tabular}
    \subcaption{disc combined}
    \end{subtable}
    \begin{subtable}{\textwidth}
    \centering
    \begin{tabular}{|c|c||c|c|}
        \hline
        preprocessing & equivariance & training epochs & training time (in minutes) \\ \hline
        trivial & & $100$ & $5$\\
        trivial & \checkmark & $80$ & $17$\\
        trivial+PDE & & $170$ & $9$\\
        trivial+PDE & \checkmark & $140$ & $31$ \\\hline
    \end{tabular}
    \subcaption{sphere combined}
    \end{subtable}
\end{table}

\section{Ablations and negative results}
\label{appendix_ablations}
This section discusses several model alterations that did not lead to noticeable improvements in our numerical experiments. We think these results are still valuable to the community. 

Above, we exclusively used the dihedral symmetry group, $D_4$, in the equivariance wrapper. For the sphere dataset, we also examined the performance of the octahedral symmetry group $O_h$. While still constituting a significant benefit over not utilizing equivariance, this method was inferior to the use of the $D_4$ symmetry group. We suspect this is due to the fixed location of the sphere dataset's Dirichlet condition, negating the benefit of generalizing to more diverse locations of boundary conditions.

In addition to the preprocessings we examined in \Cref{sec_main_results}, we experimented with various other preprocessing combinations. For instance, for an alternative preprocessing strategy we construct a binary mask to extract all voxels with either Dirichlet boundary or load cases. The \textit{convex hull} of this mask is then given as the set of voxels included in the smallest convex polygon that surrounds all these voxels. In case of one load point and one Dirichlet voxel this results in the simplest possible density distribution that leads to a connected mechanical structure. This convex hull density distribution is then used as a $1$-channel input to the neural network. We found that convex hull preprocessing did not lead to any measurable improvements on the validation dataset, neither on its own nor in combination with other preprocessings. We speculate that this is due to the convex hull being invariant to the direction of the applied forces. We also found that applying PDE preprocessing on its own produced similar but somewhat inferior results compared to the trivial+PDE preprocessing. Also, adding the full initial stress tensor and initial displacements to the PDE preprocessing did not improve over the presented PDE preprocessing.

We also experimented with different network architectures. In addition to the UNet, we considered a ResNet~\cite{zhang2018road} of depth $10$ with $5$-layer CNN blocks. However, despite the UNet requiring less VRAM and computation time, it reliably outperformed the ResNet.

\newpage
\section{Examples of model predictions}
\label{appendix_b_samples}

\begin{table}[h!!!]
    \caption{Random samples from the disc dataset. The first columns displays prediction from the UNet with trivial preprocessing and without equivariance. The second columns shows predictions from the UNet with trivial+PDE preprocessing and equivariance, which is our best model pipeline. All models have been trained on $1500$ samples. In the third columns we show the ground truths densities corresponding to each sample.}
    \label{7_tab:random_samples_disc}
    \begin{subtable}{0.44\textwidth}
        \centering
        \setlength\tabcolsep{6pt}
        \resizebox{0.77\textwidth}{!}{\begin{tabular}{|c|c||c|}
\hline
\parbox{4em}{\centering UNet} & \parbox{4em}{\centering UNet\\+physics} & \xrowht{20pt}\parbox{4em}{\centering ground\\truth} \\
\hline\rule{0pt}{2cm}

\xrowht{20pt} 
\includegraphics[width=0.25\textwidth]{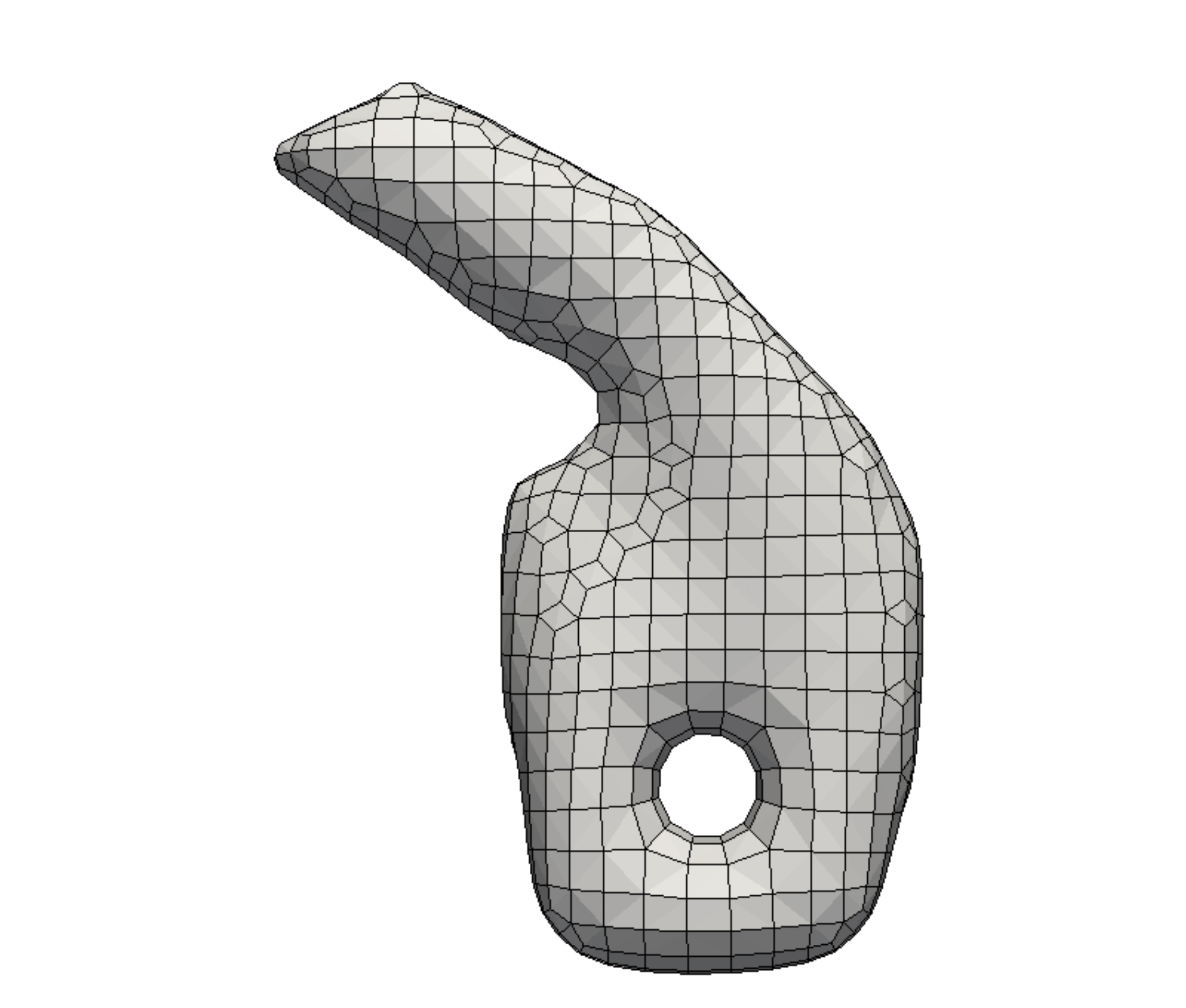} & \includegraphics[width=0.25\textwidth]{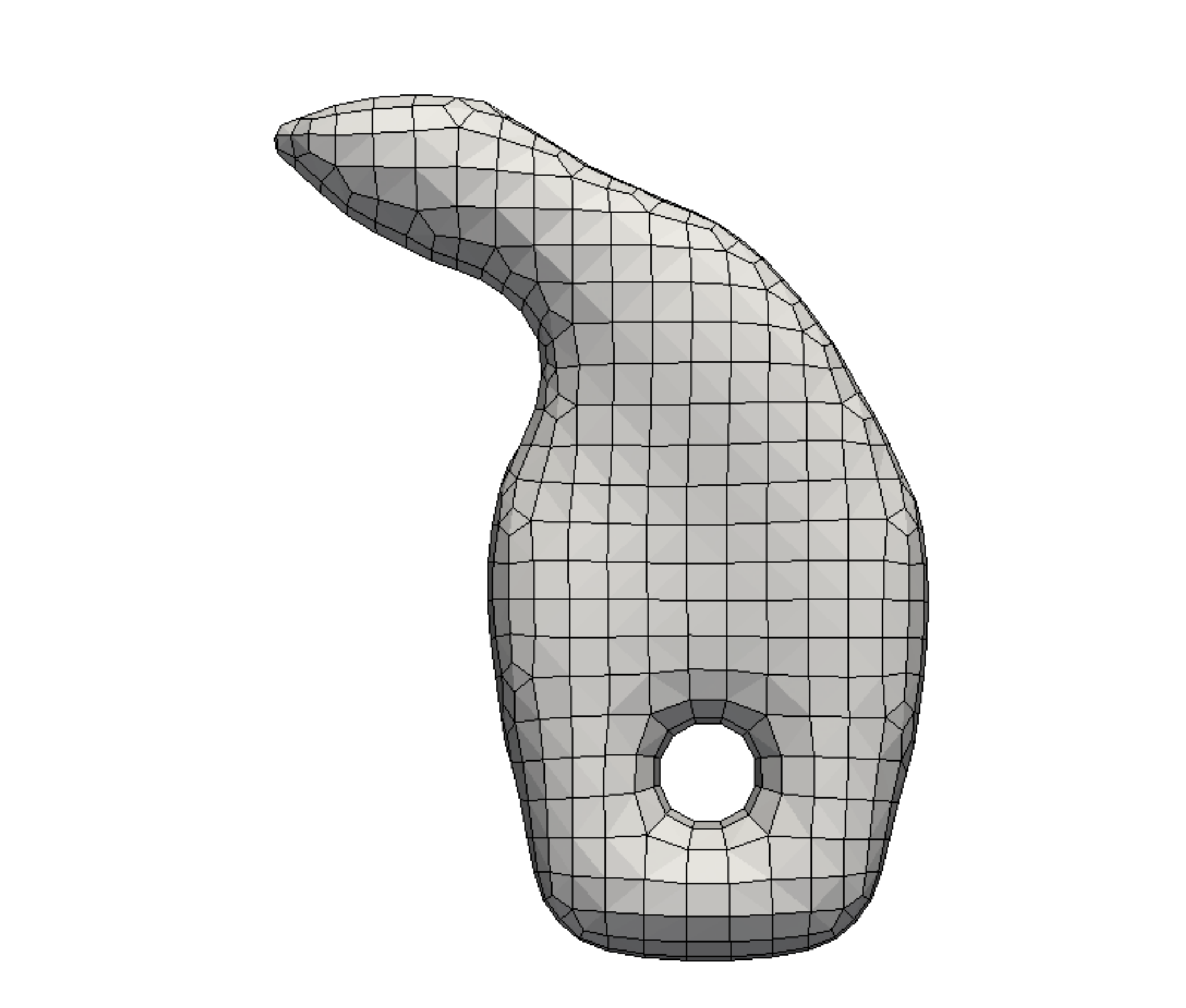} & \includegraphics[width=0.25\textwidth]{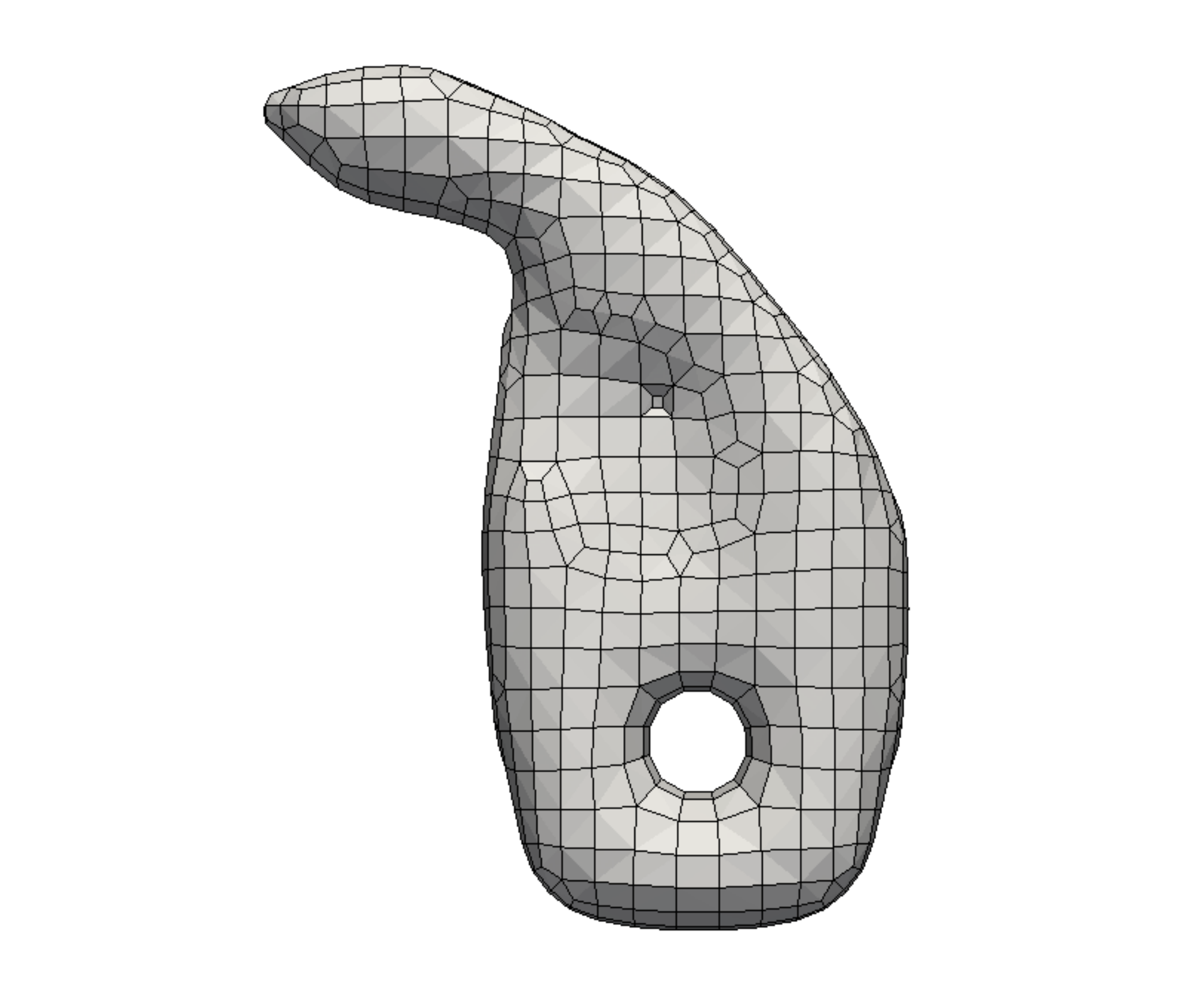}\\

\xrowht{20pt} 
\includegraphics[width=0.25\textwidth]{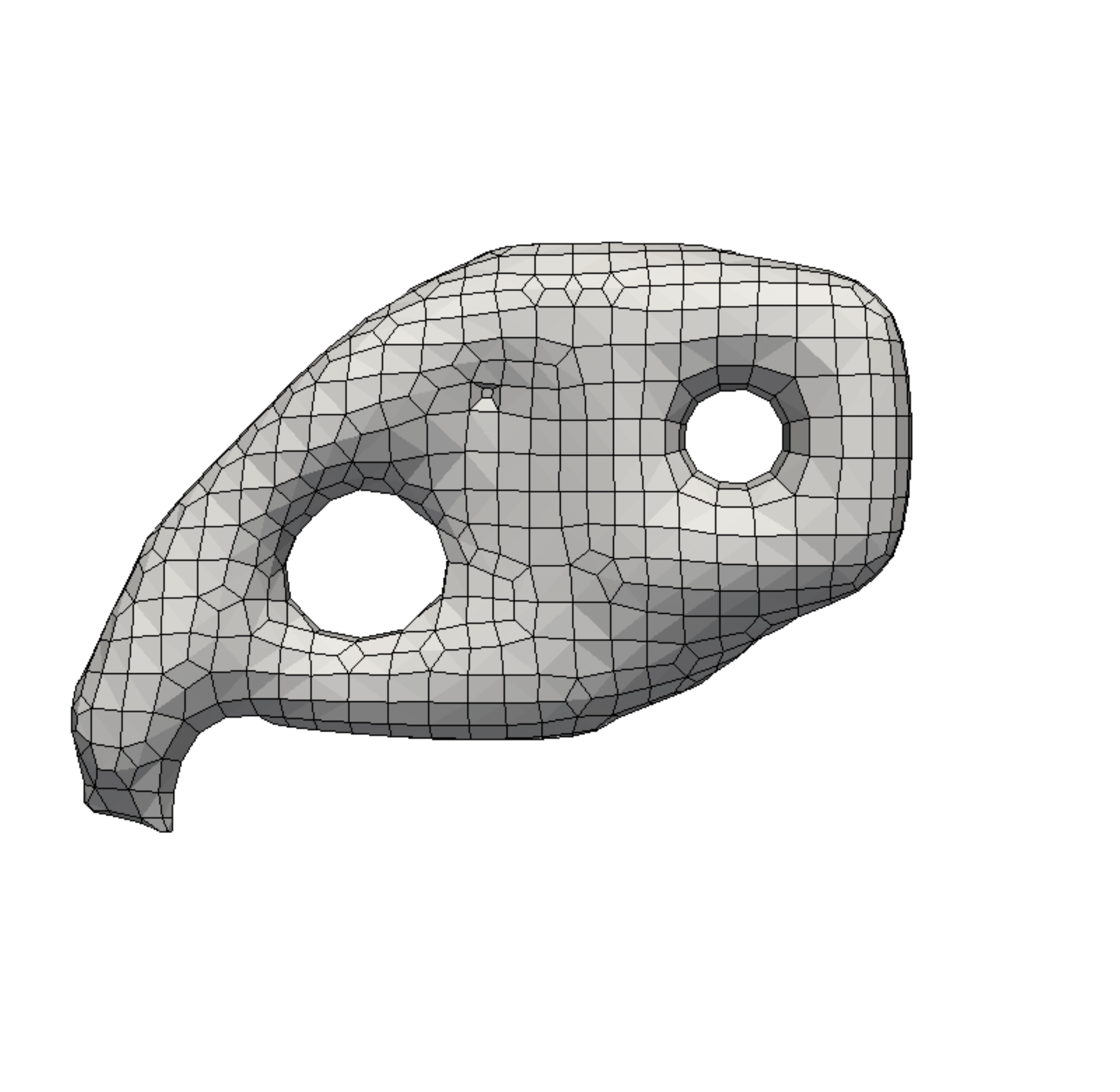} & \includegraphics[width=0.25\textwidth]{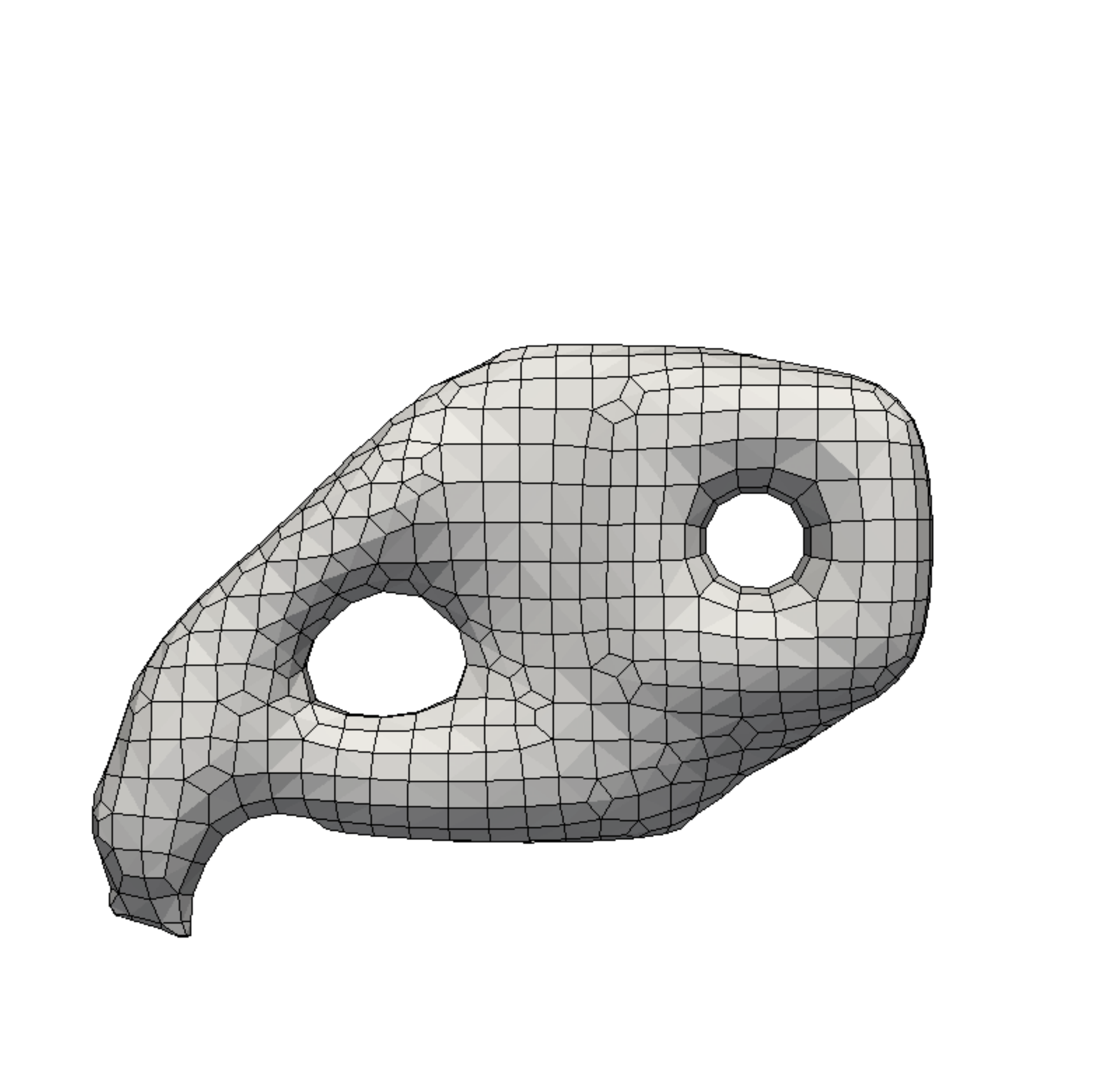} & \includegraphics[width=0.25\textwidth]{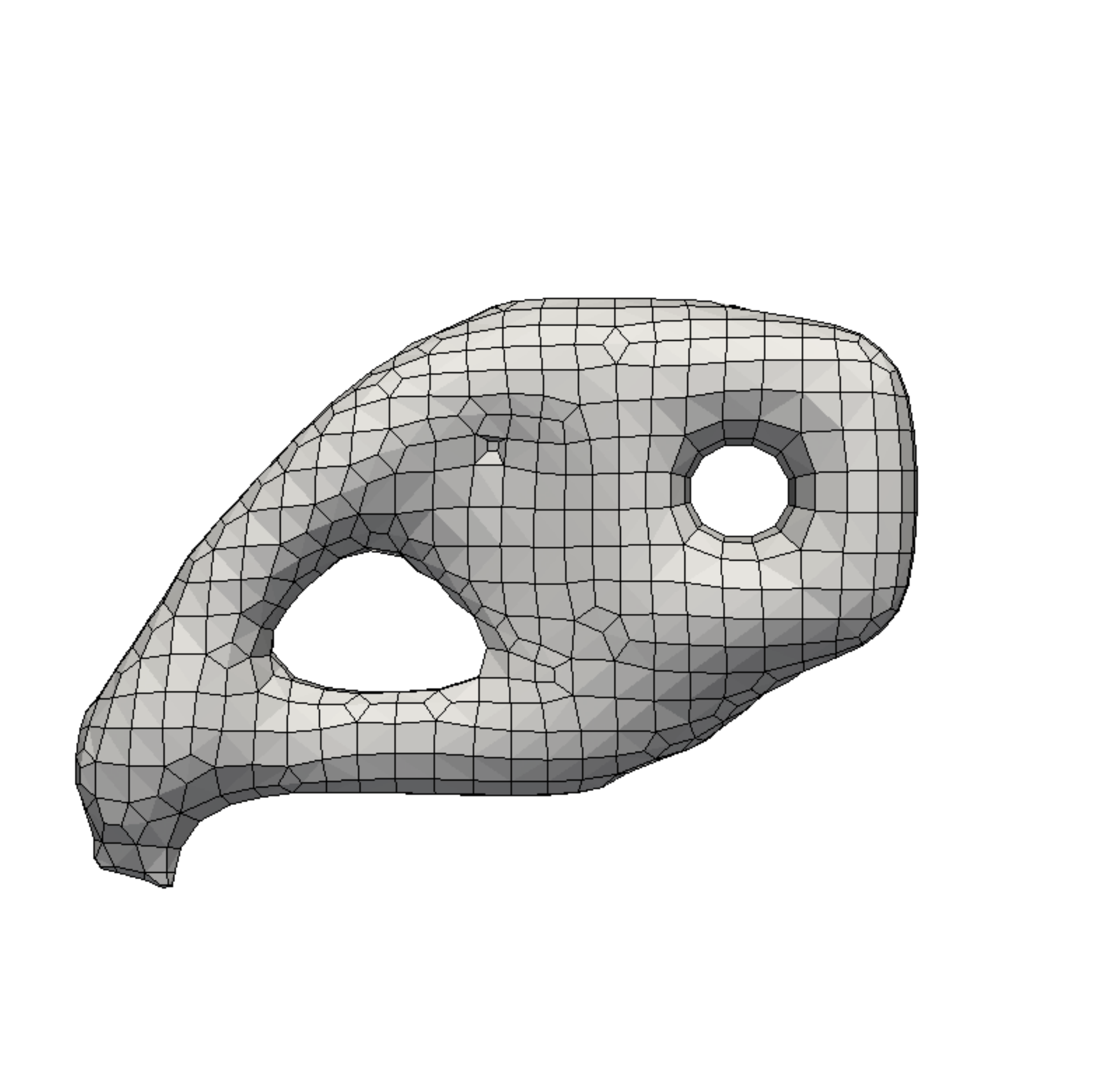}\\

\xrowht{20pt} 
\includegraphics[width=0.25\textwidth]{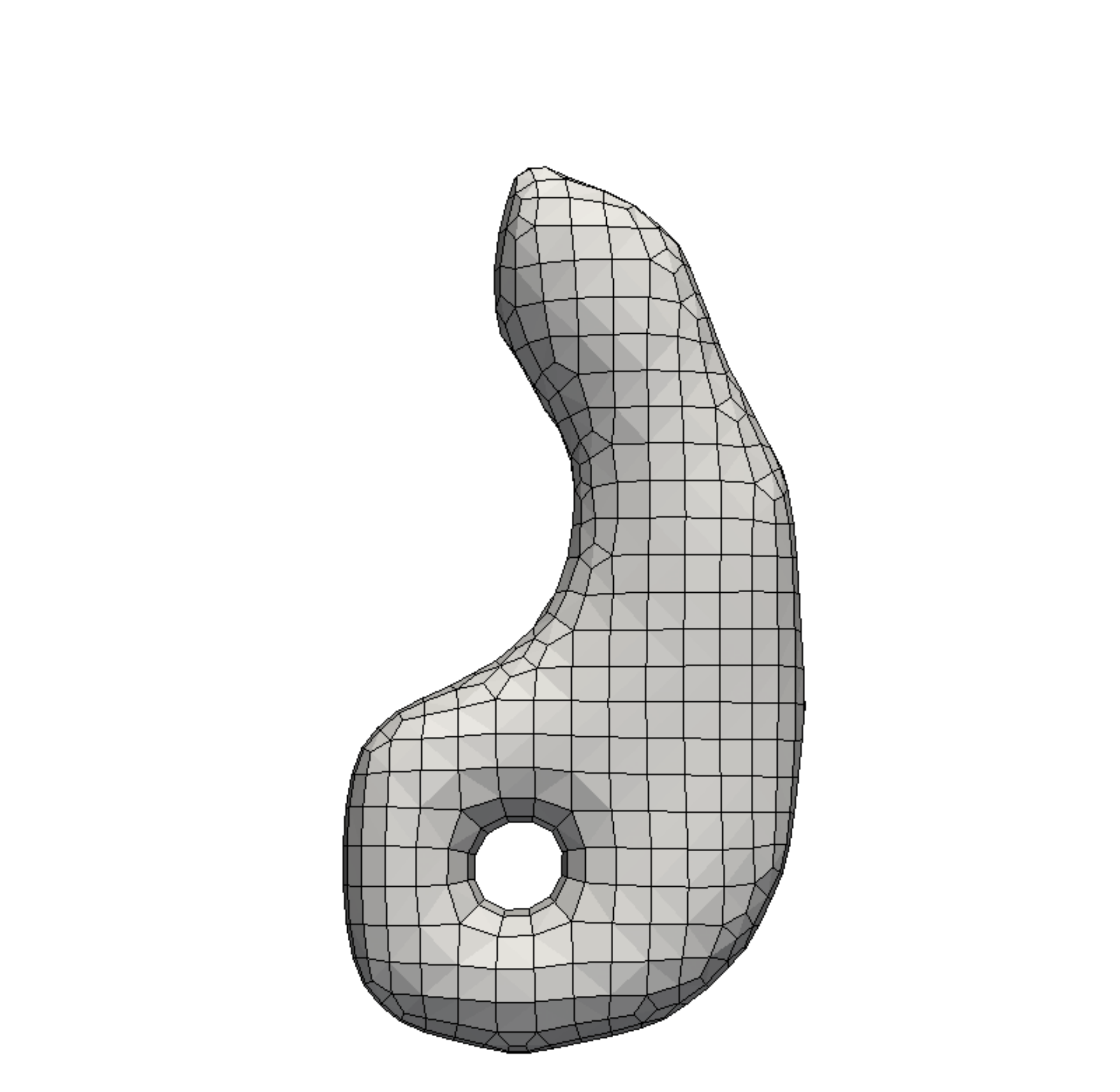} & \includegraphics[width=0.25\textwidth]{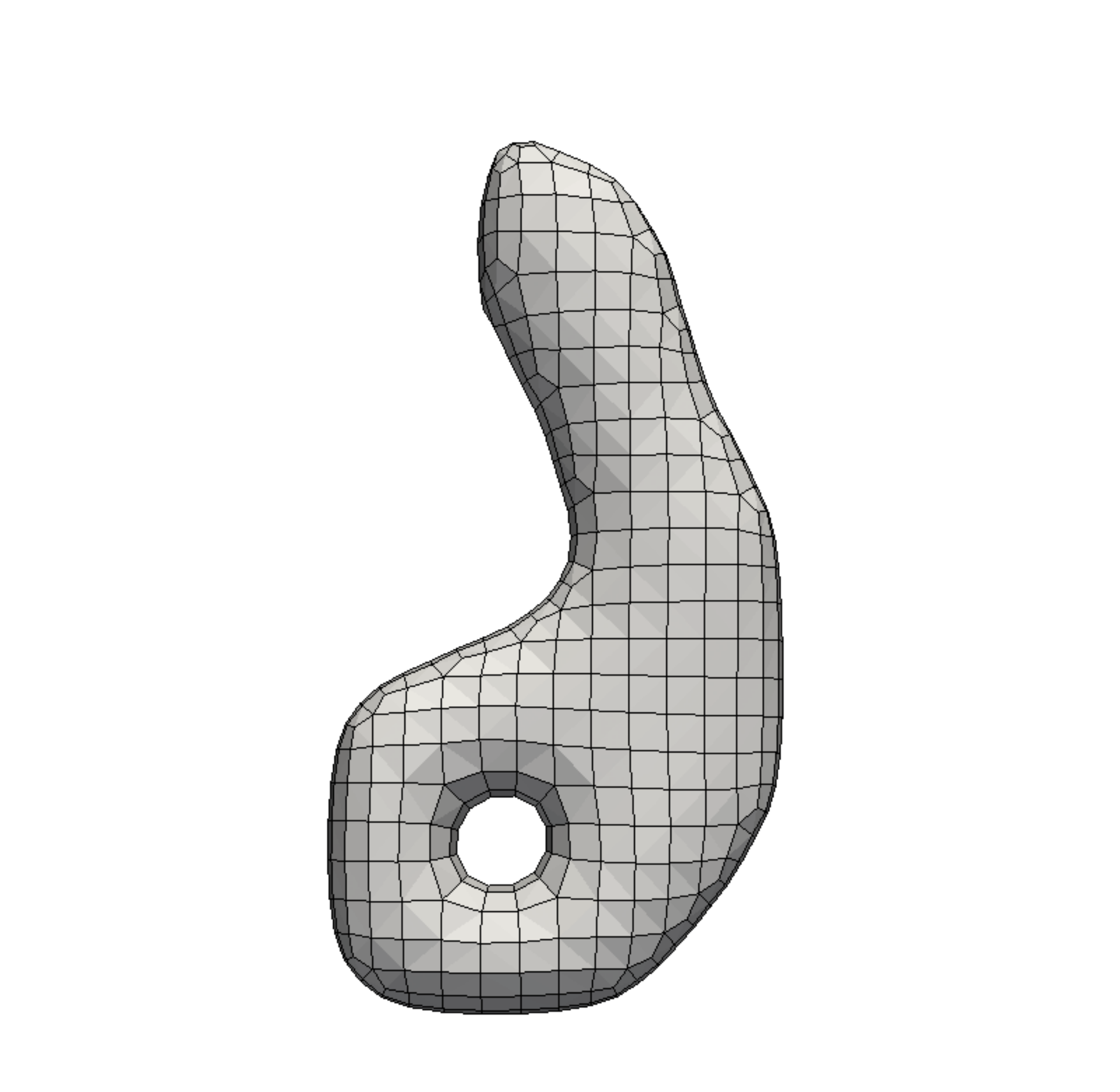} & \includegraphics[width=0.25\textwidth]{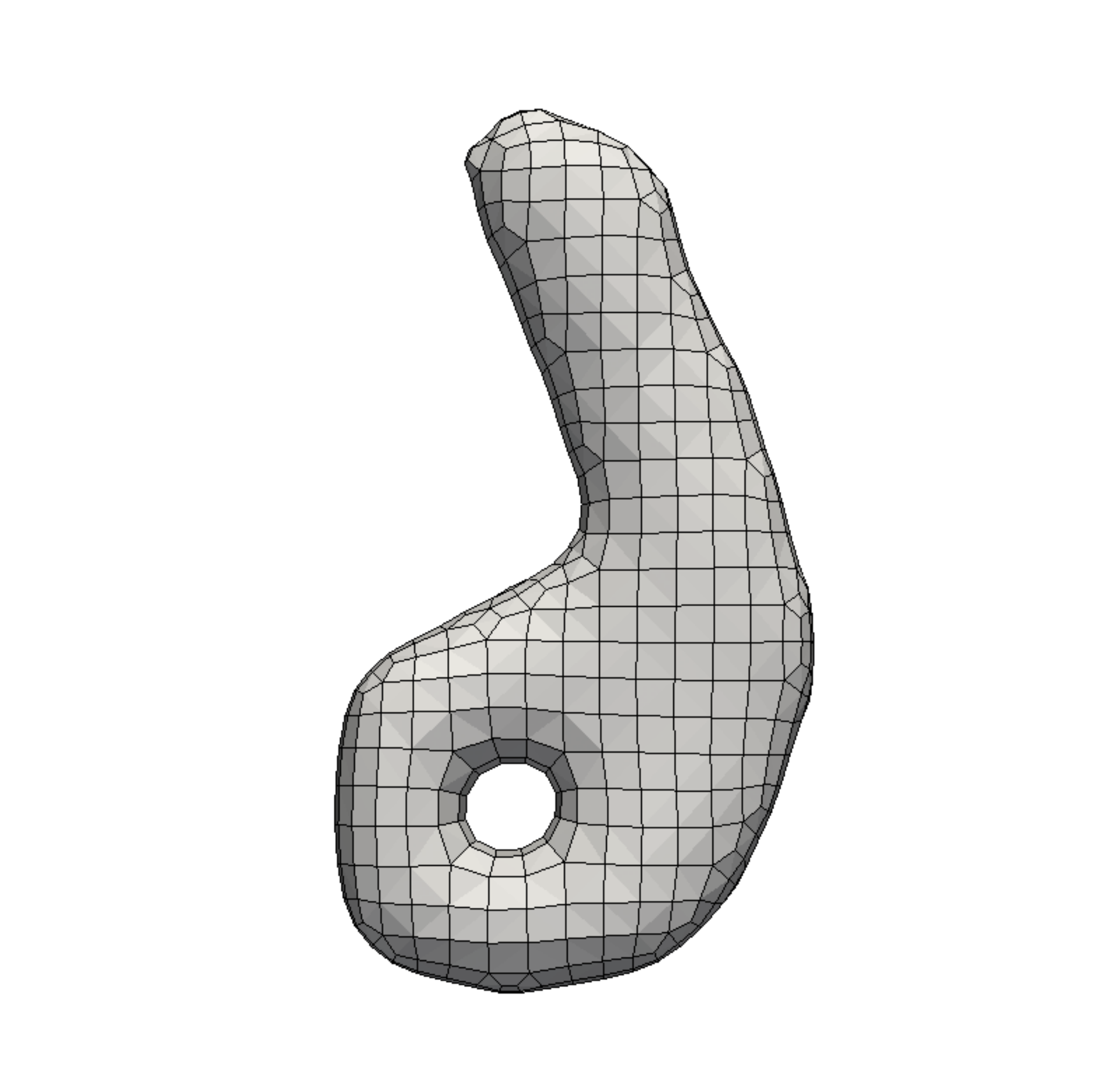}\\

\xrowht{20pt} 
\includegraphics[width=0.25\textwidth]{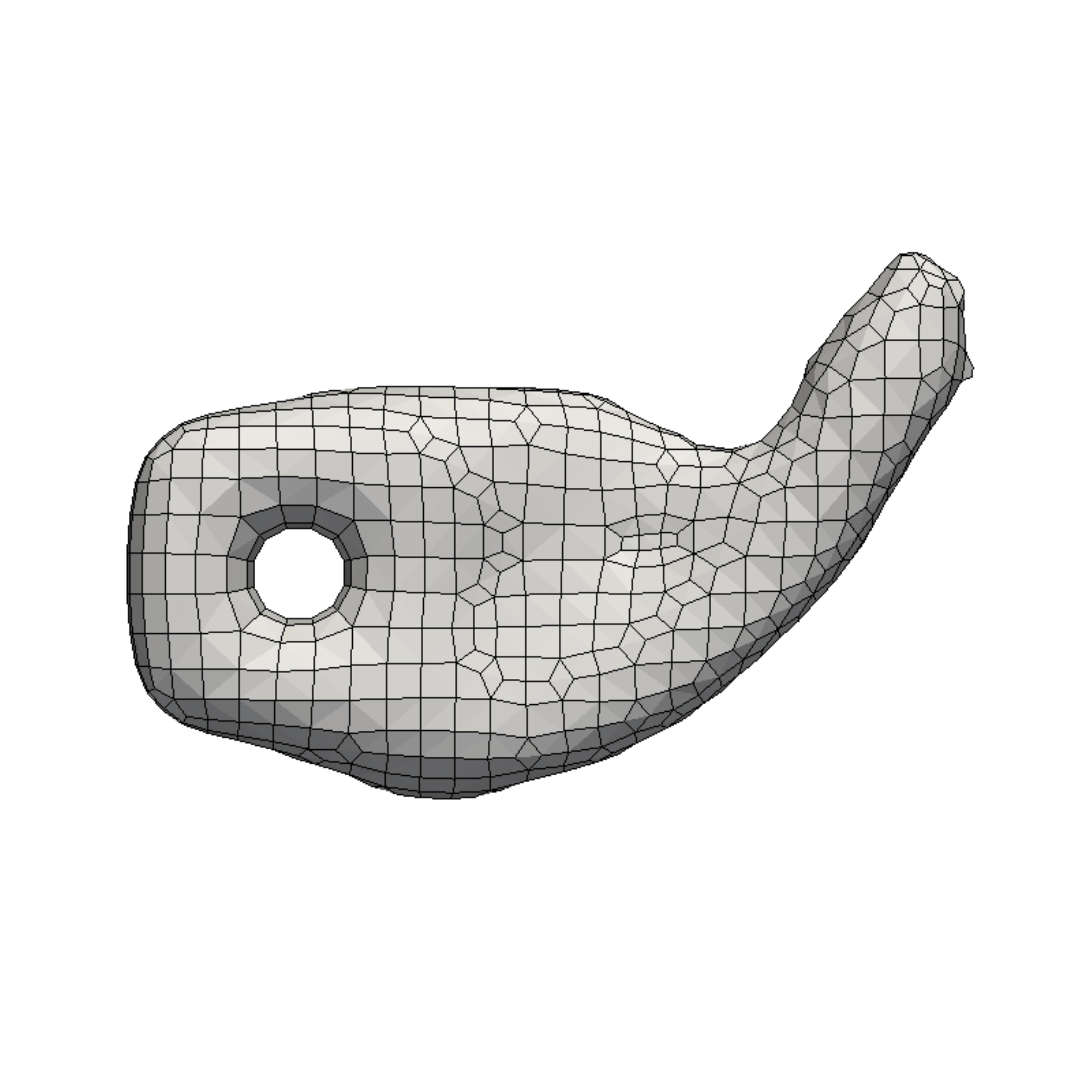} & \includegraphics[width=0.25\textwidth]{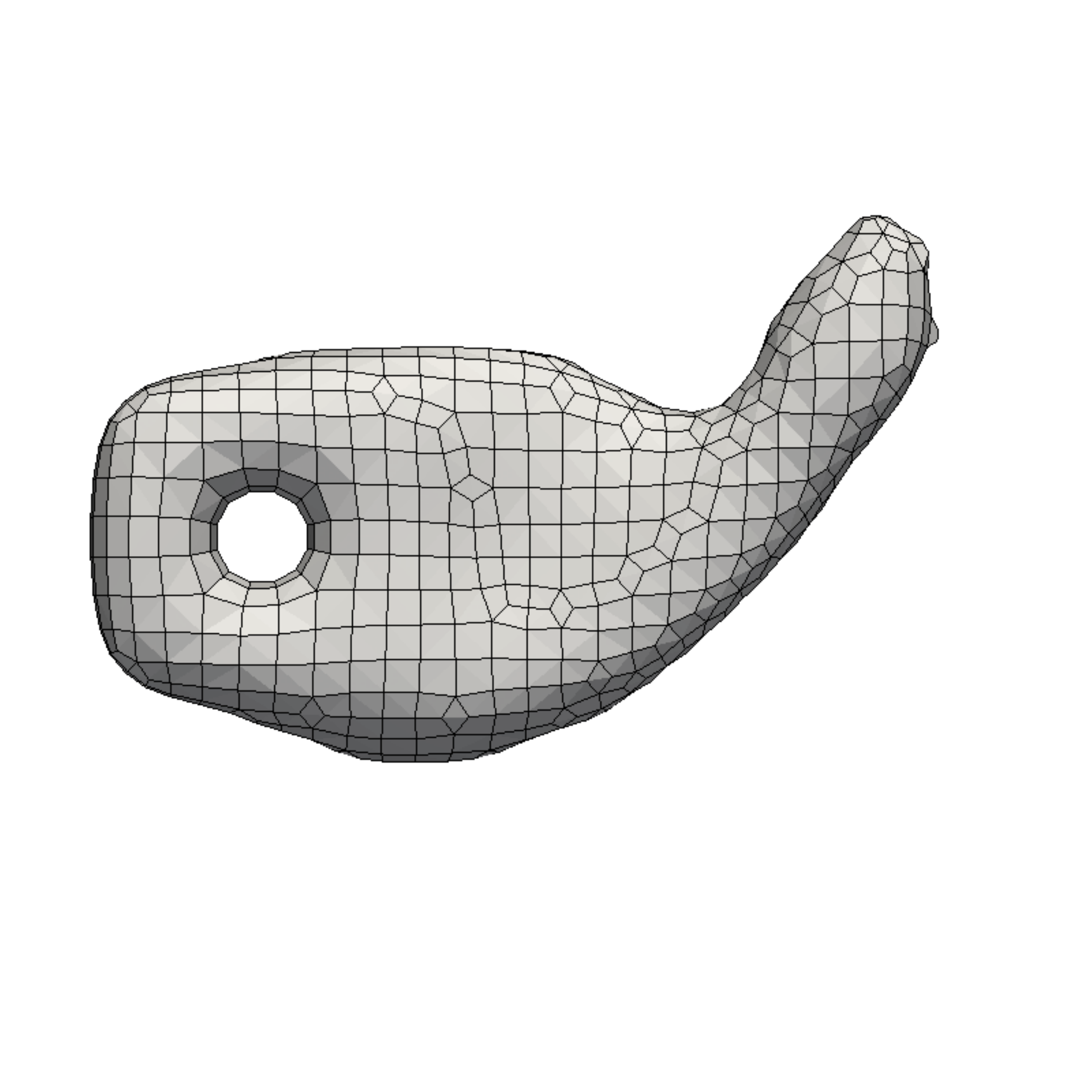} & \includegraphics[width=0.25\textwidth]{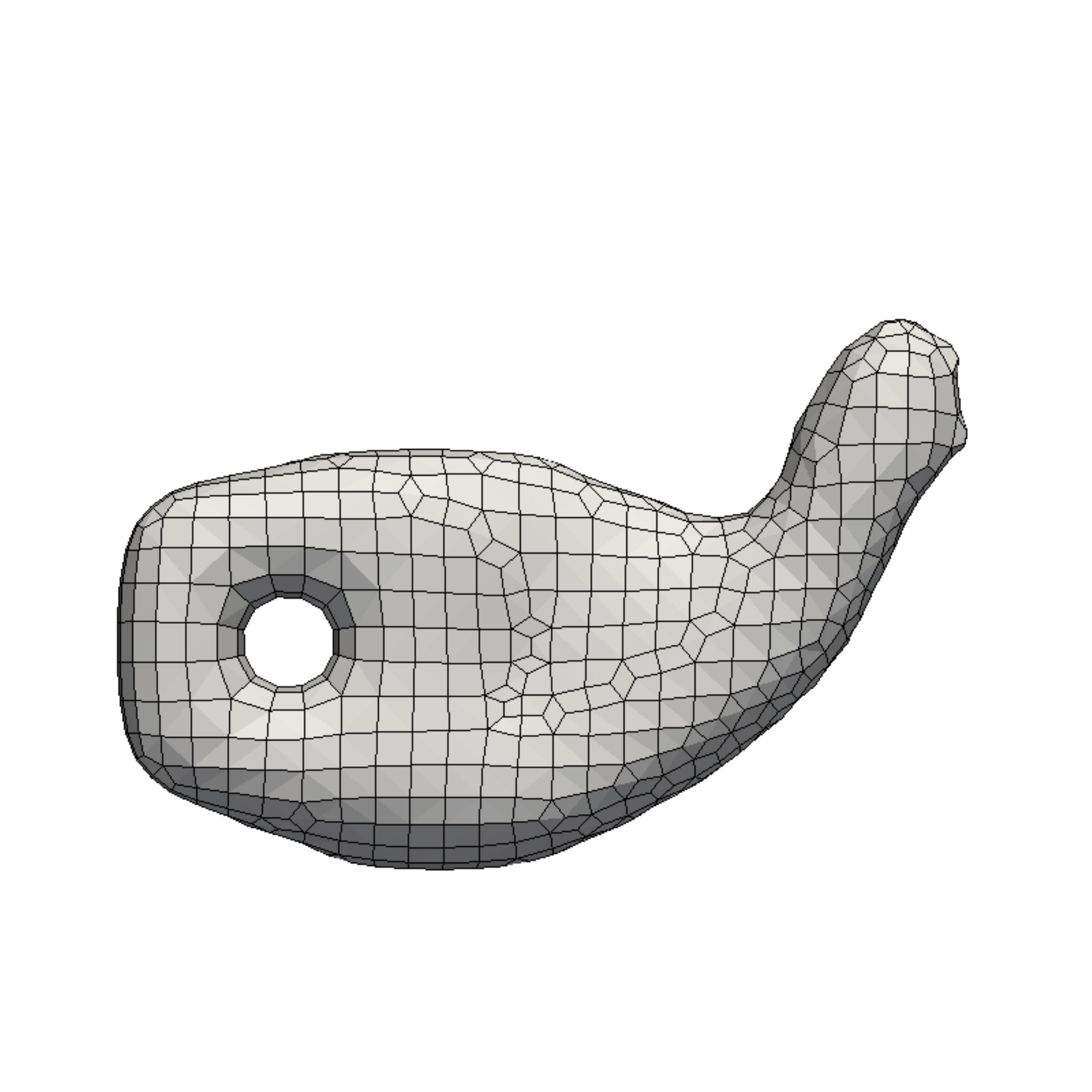}\\

\xrowht{20pt} 
\includegraphics[width=0.25\textwidth]{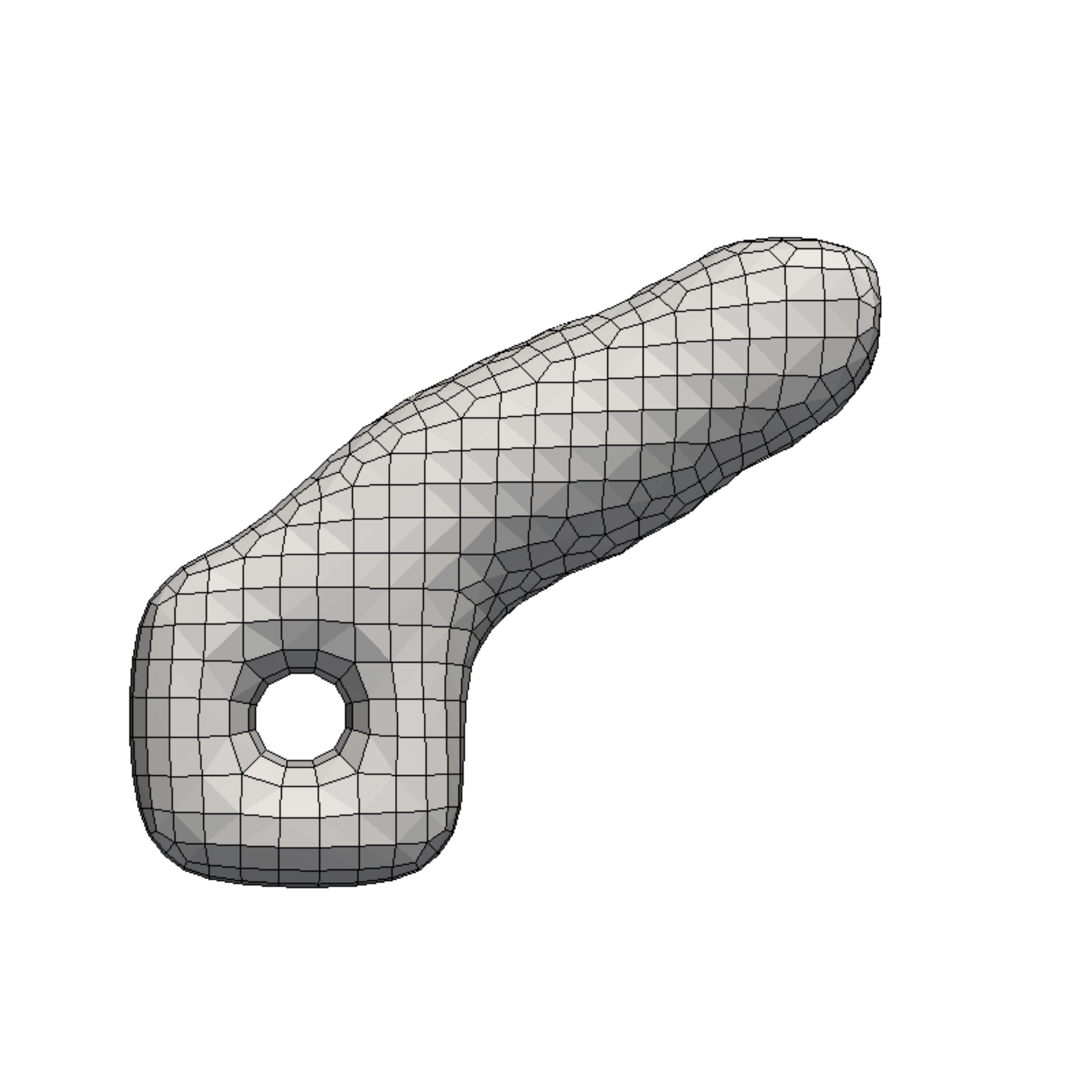} & \includegraphics[width=0.25\textwidth]{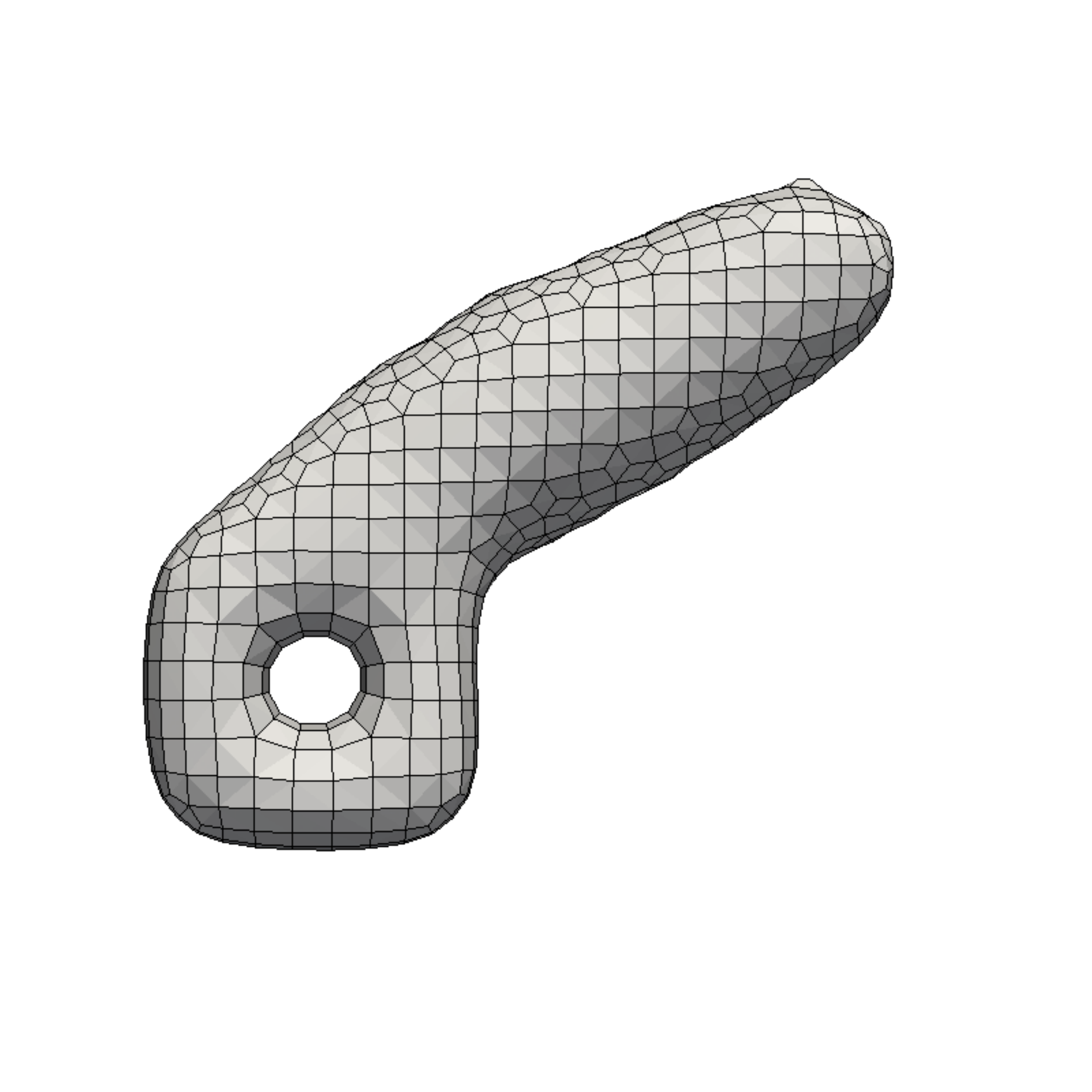} & \includegraphics[width=0.25\textwidth]{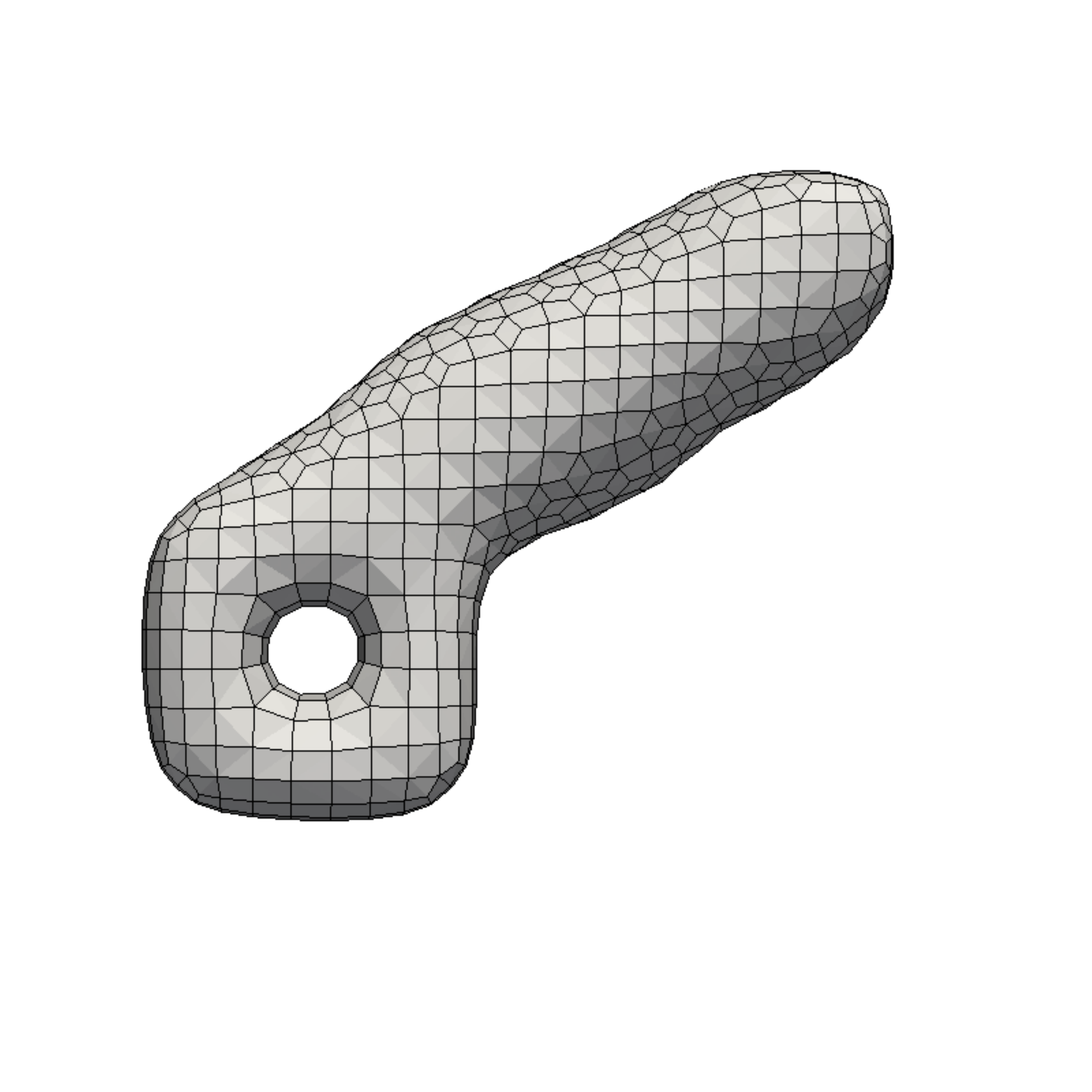}\\

\xrowht{20pt} 
\includegraphics[width=0.25\textwidth]{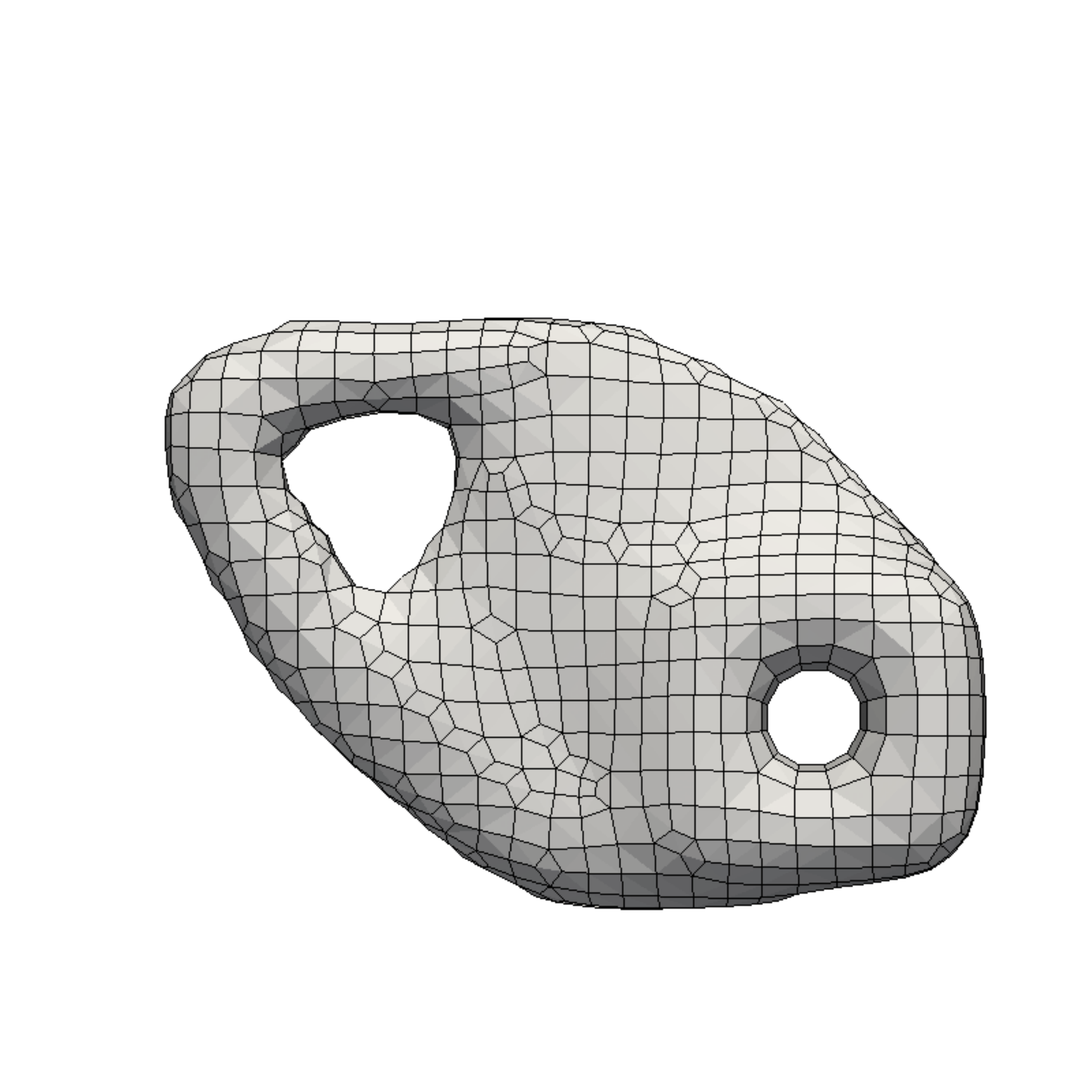} & \includegraphics[width=0.25\textwidth]{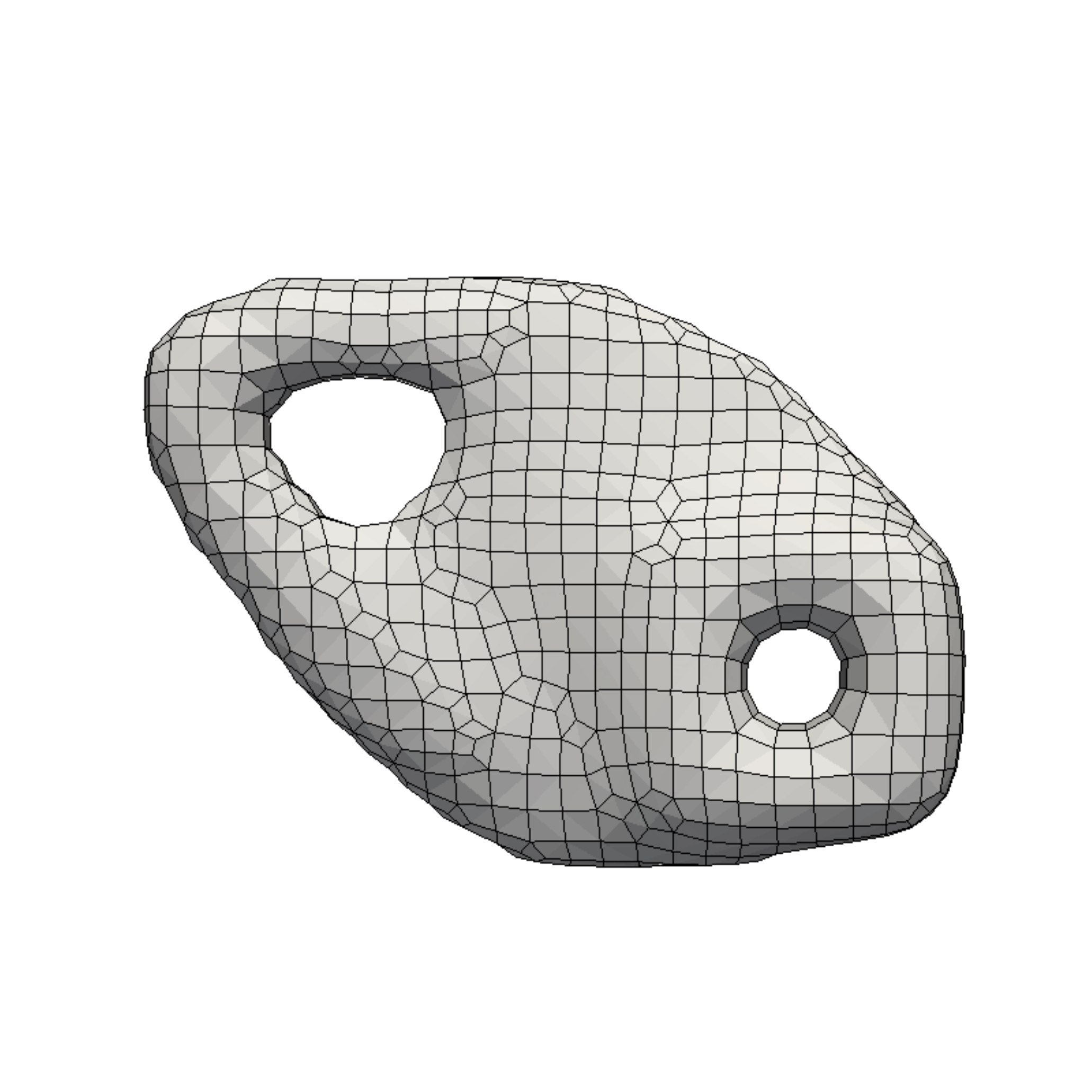} & \includegraphics[width=0.25\textwidth]{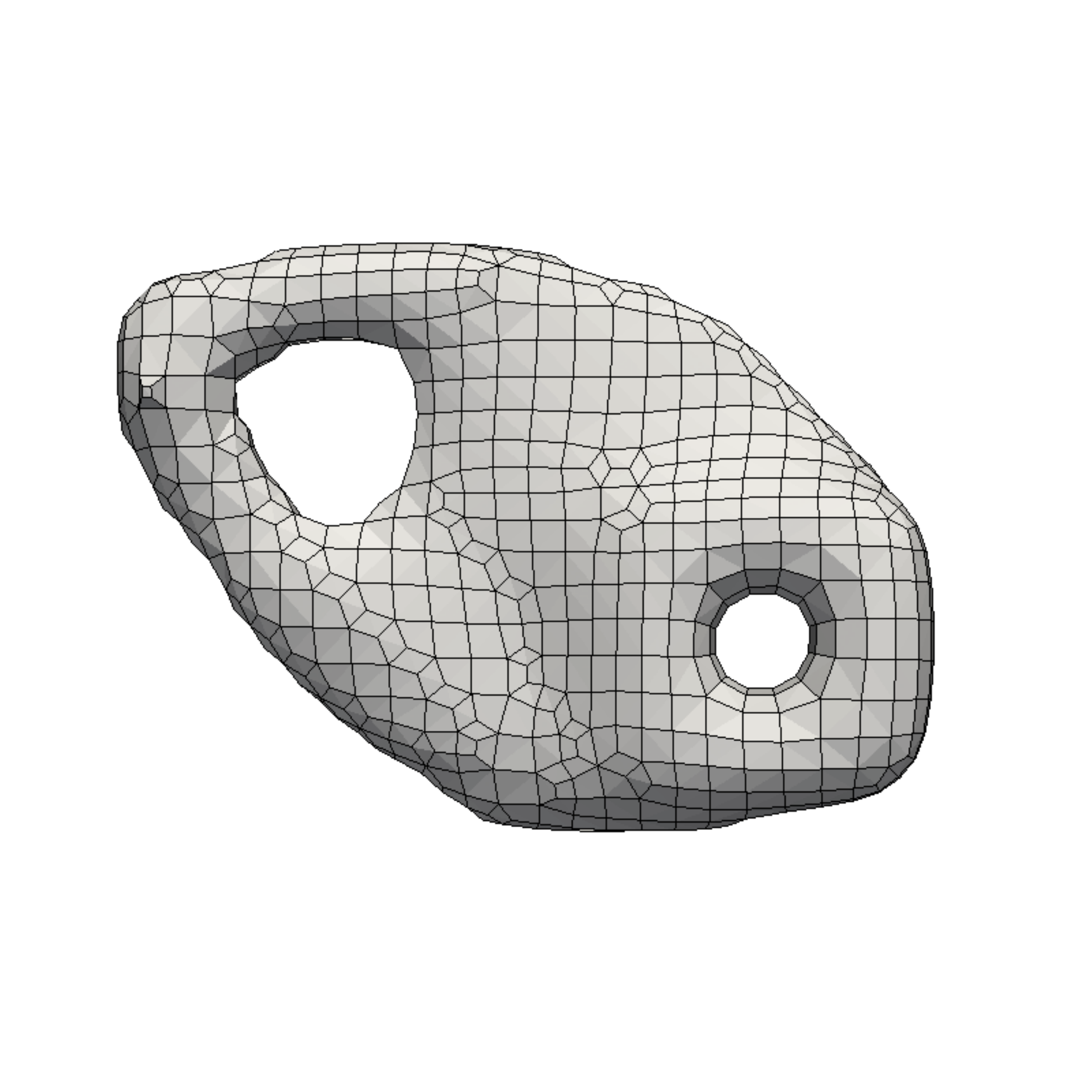}\\

\xrowht{20pt} 
\includegraphics[width=0.25\textwidth]{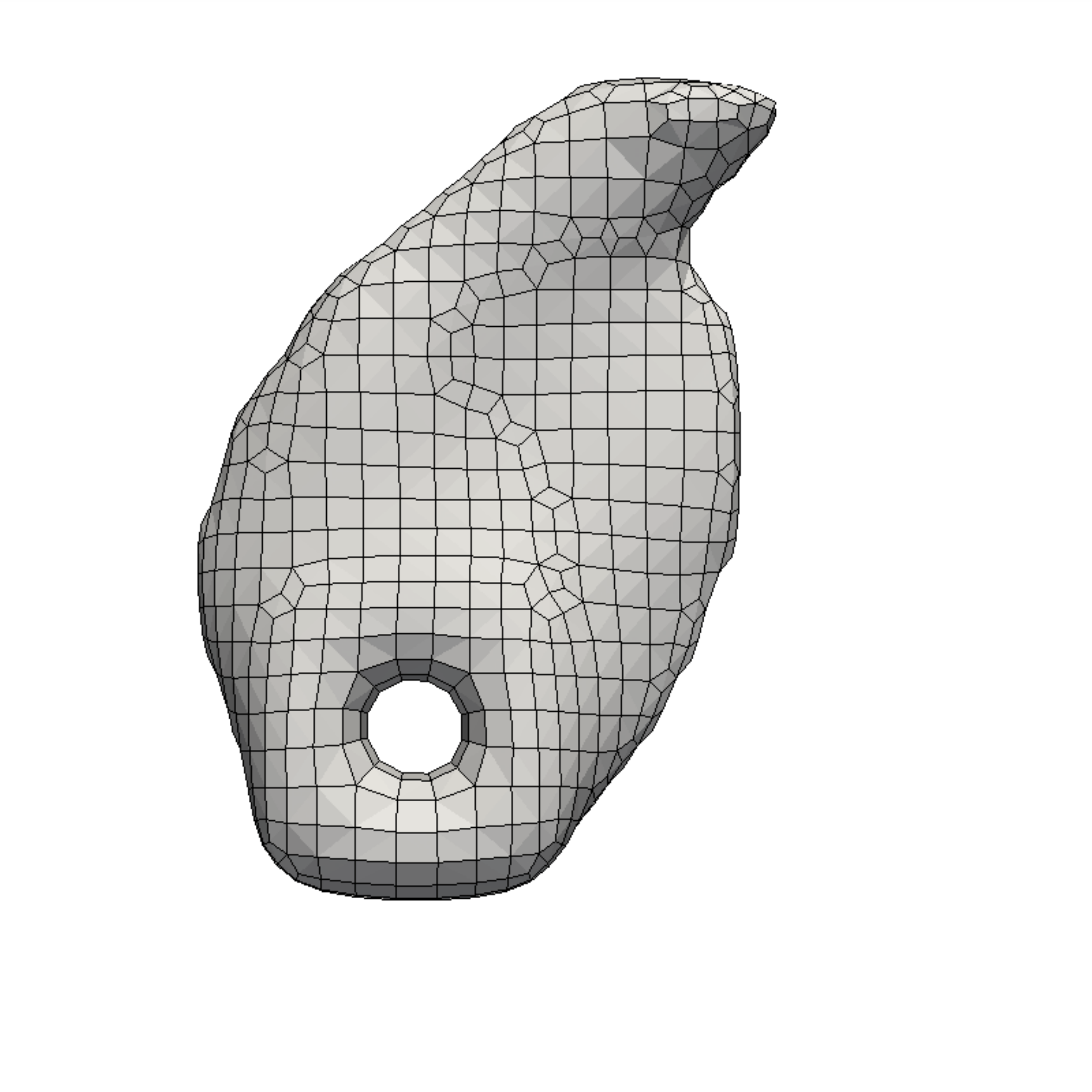} & \includegraphics[width=0.25\textwidth]{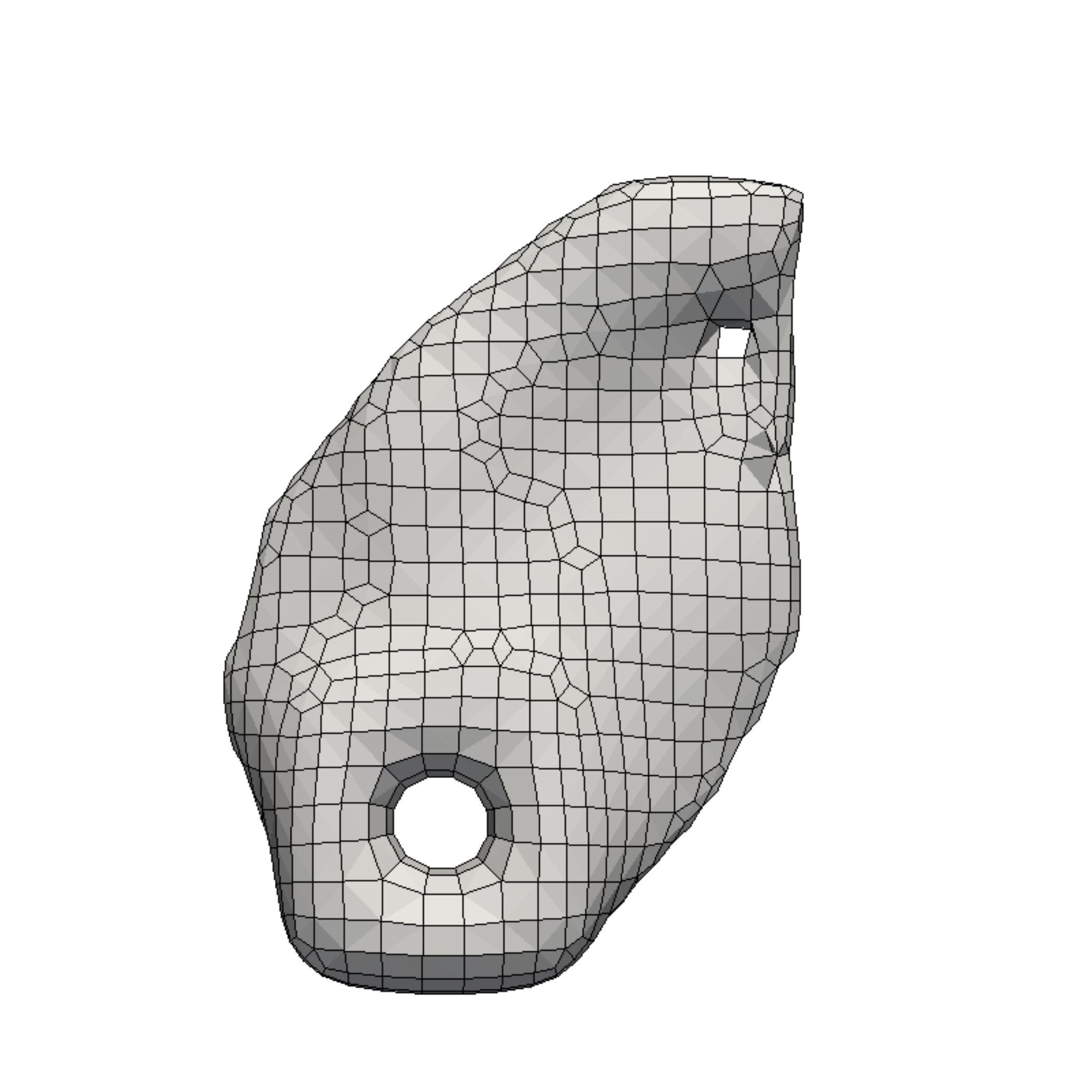} & \includegraphics[width=0.25\textwidth]{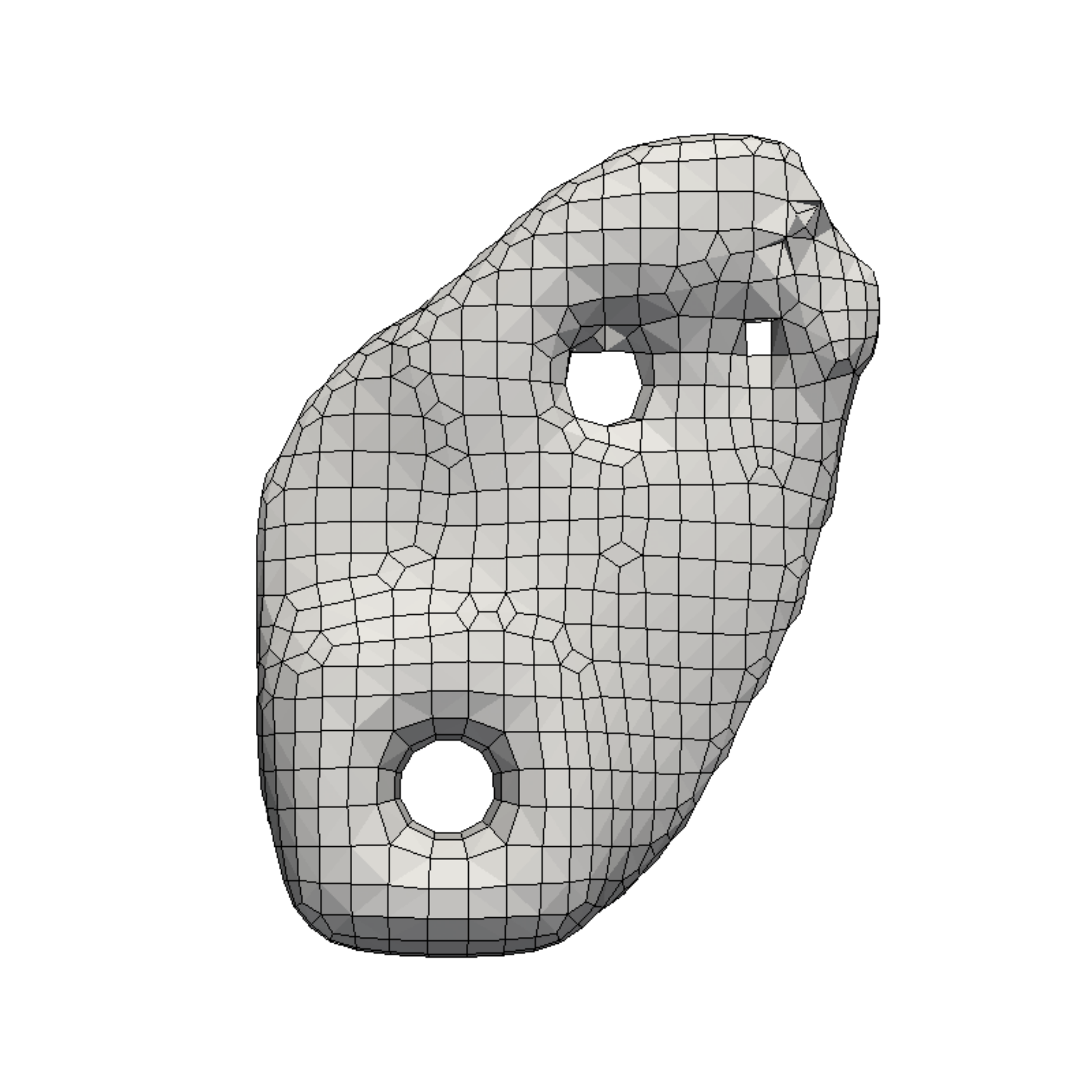}\\

\xrowht{20pt} 
\includegraphics[width=0.25\textwidth]{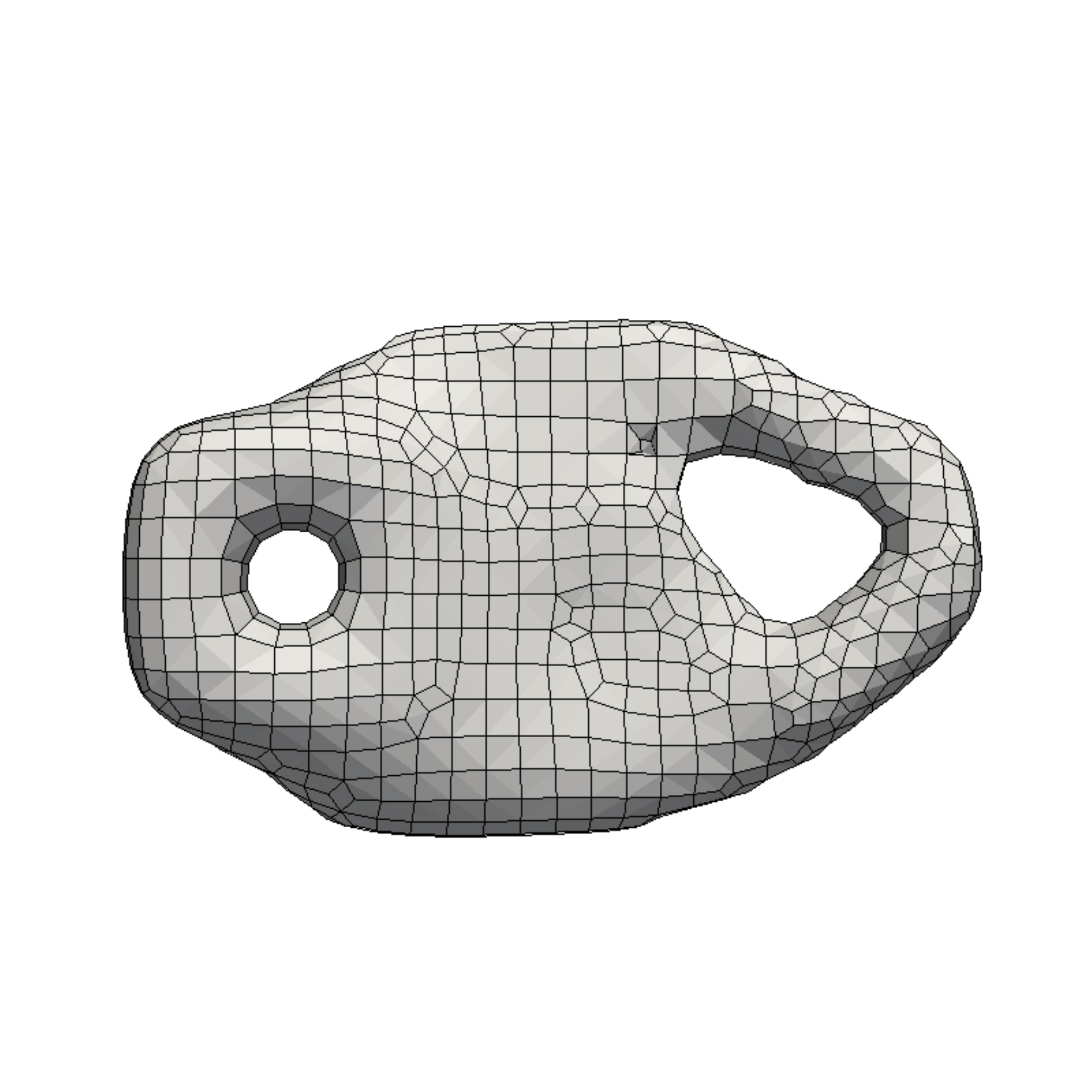} & \includegraphics[width=0.25\textwidth]{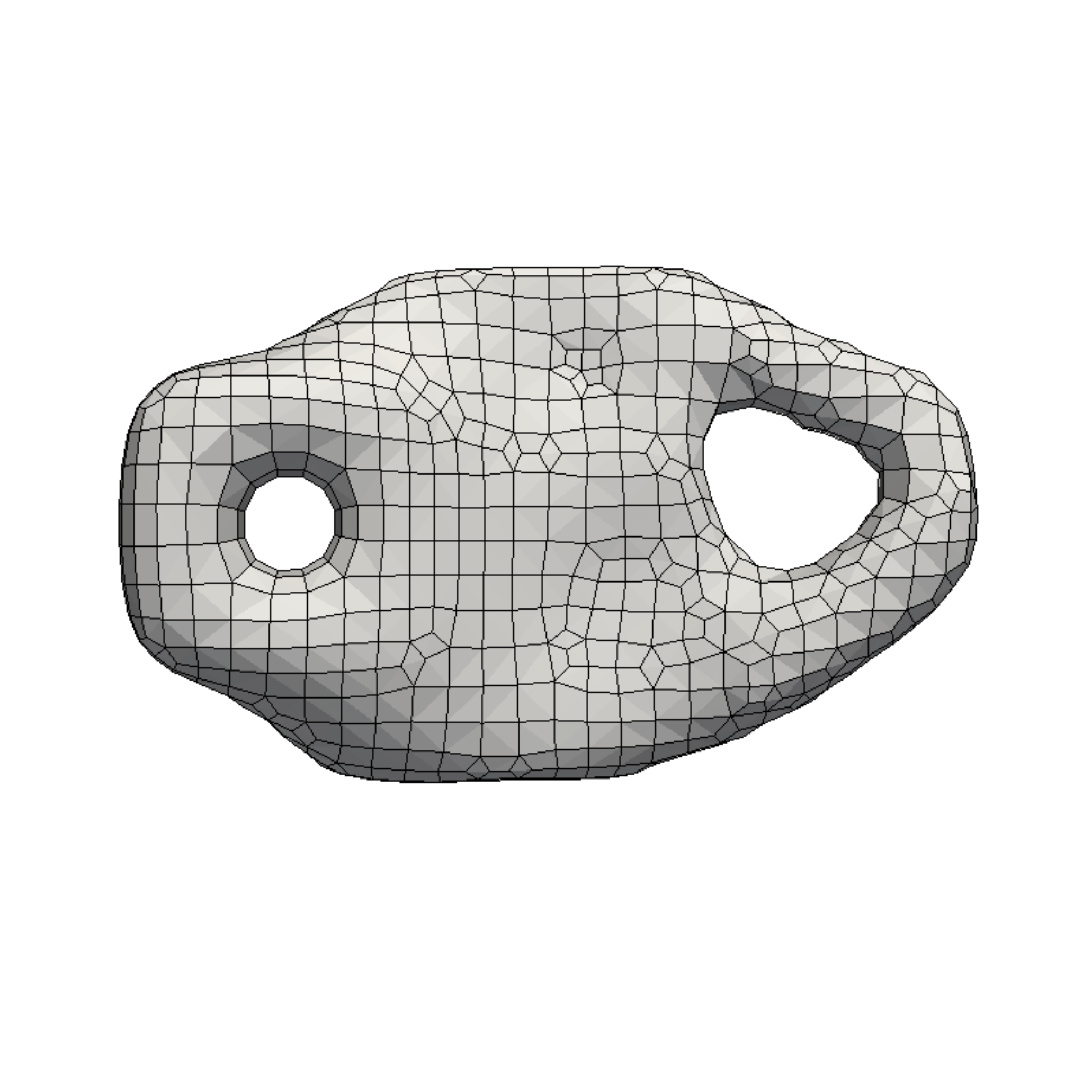} & \includegraphics[width=0.25\textwidth]{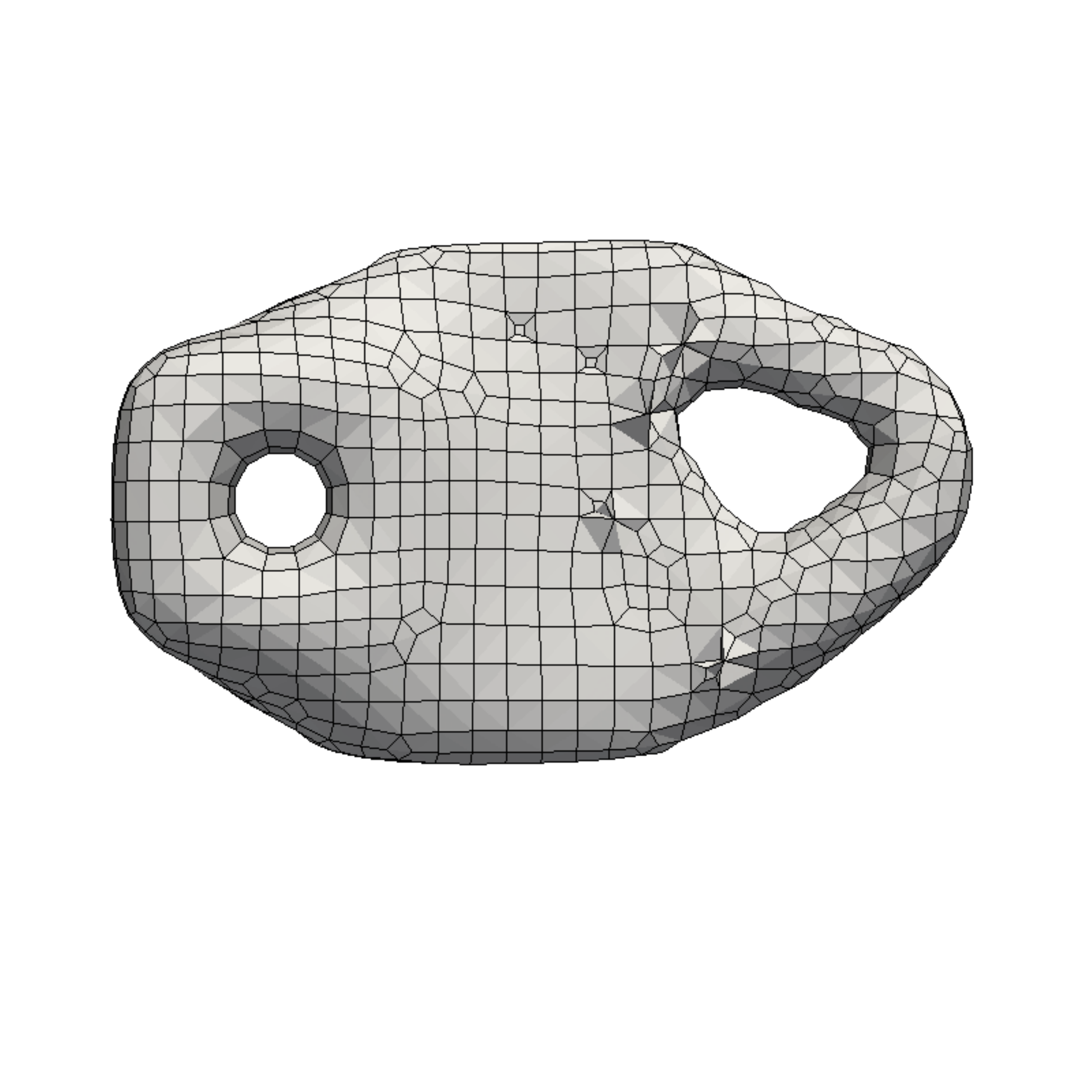}\\

\xrowht{20pt} 
\includegraphics[width=0.25\textwidth]{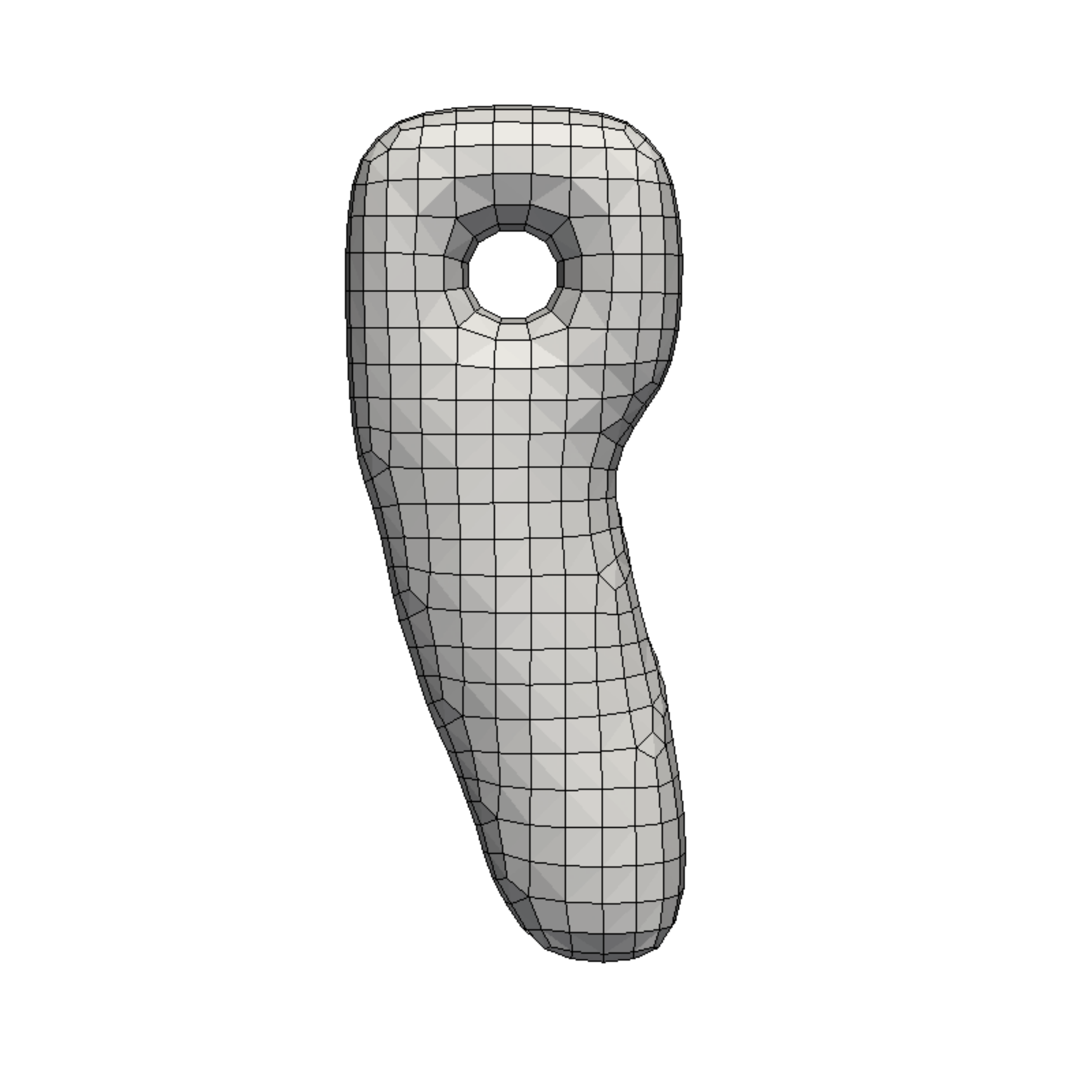} & \includegraphics[width=0.25\textwidth]{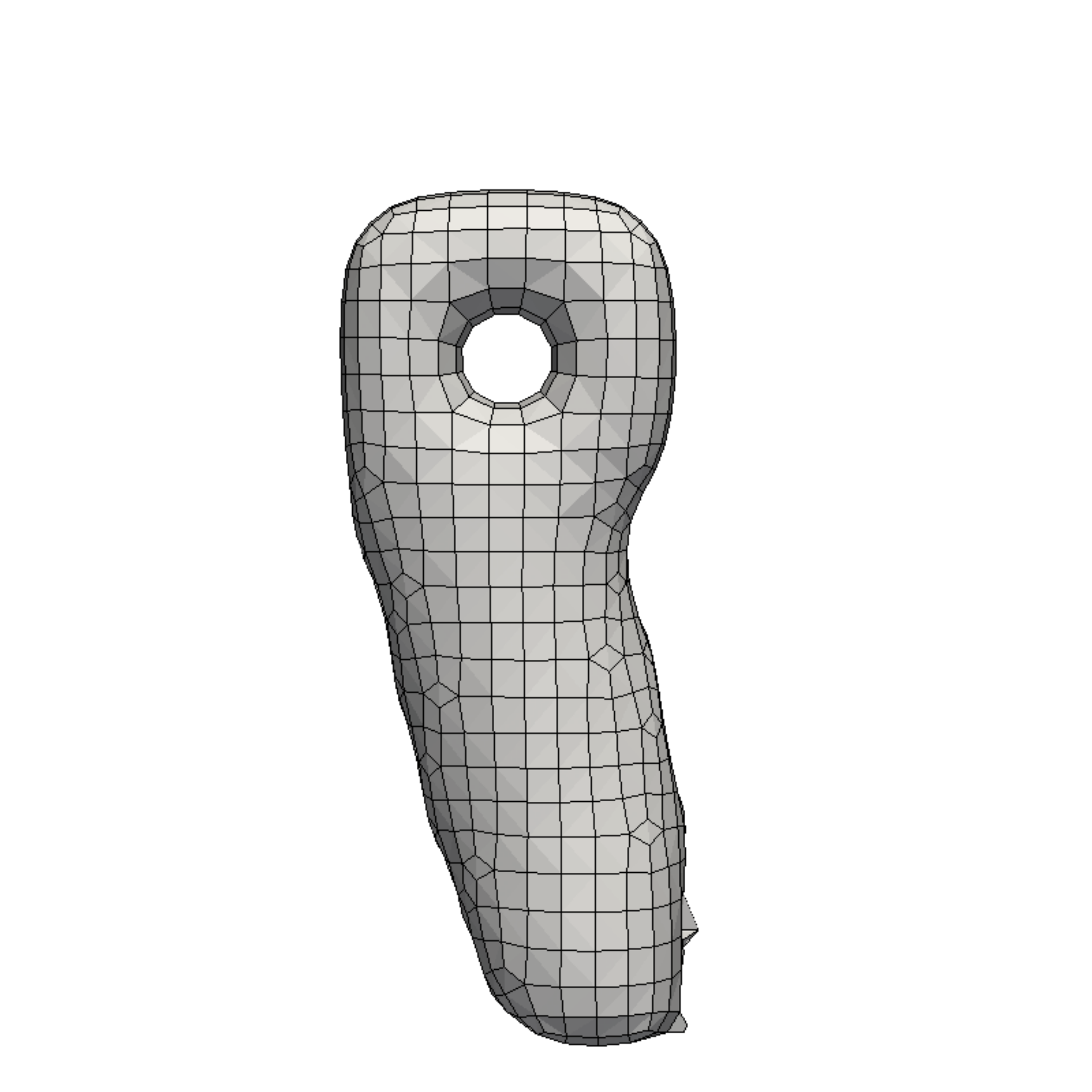} & \includegraphics[width=0.25\textwidth]{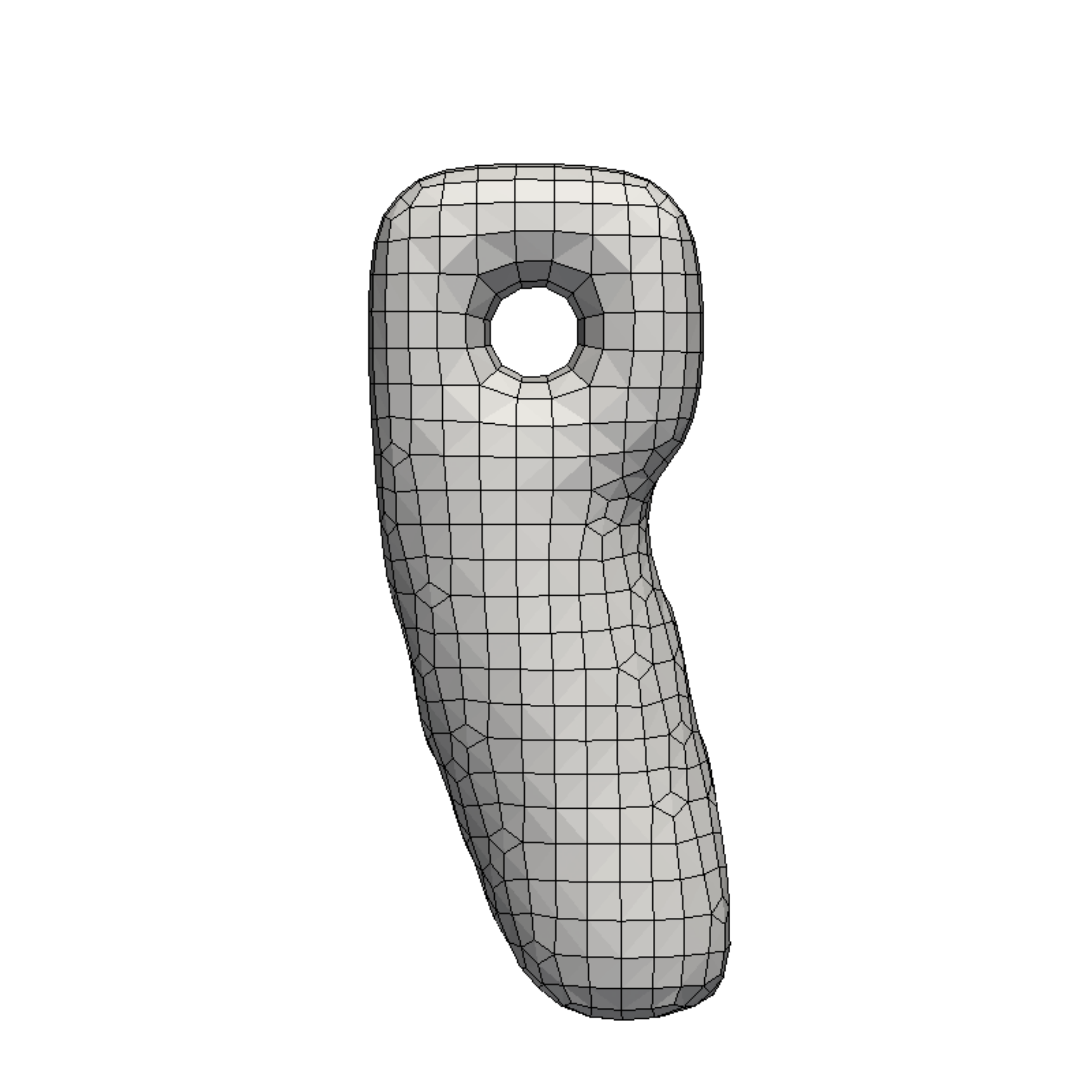}\\

\xrowht{20pt} 
\includegraphics[width=0.25\textwidth]{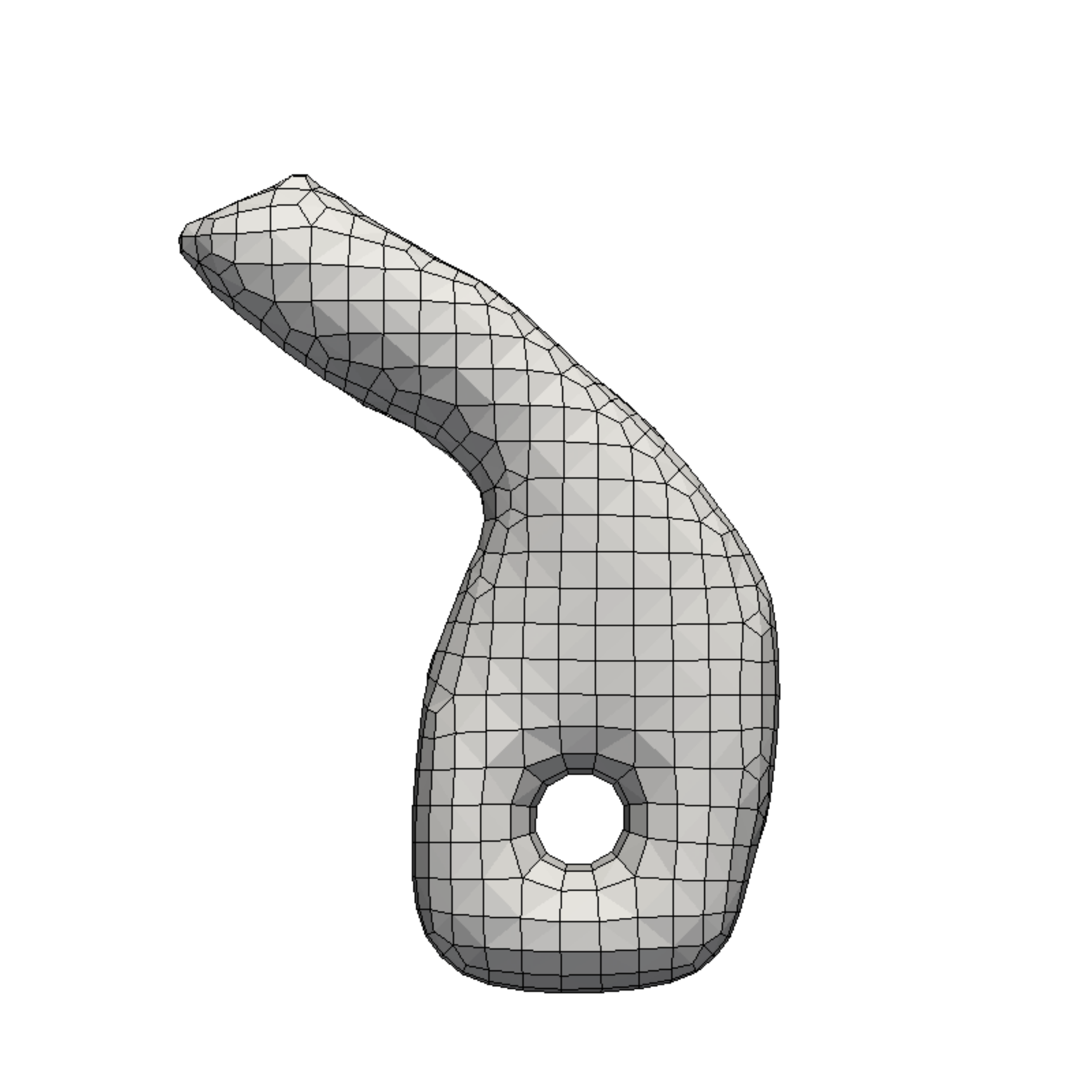} & \includegraphics[width=0.25\textwidth]{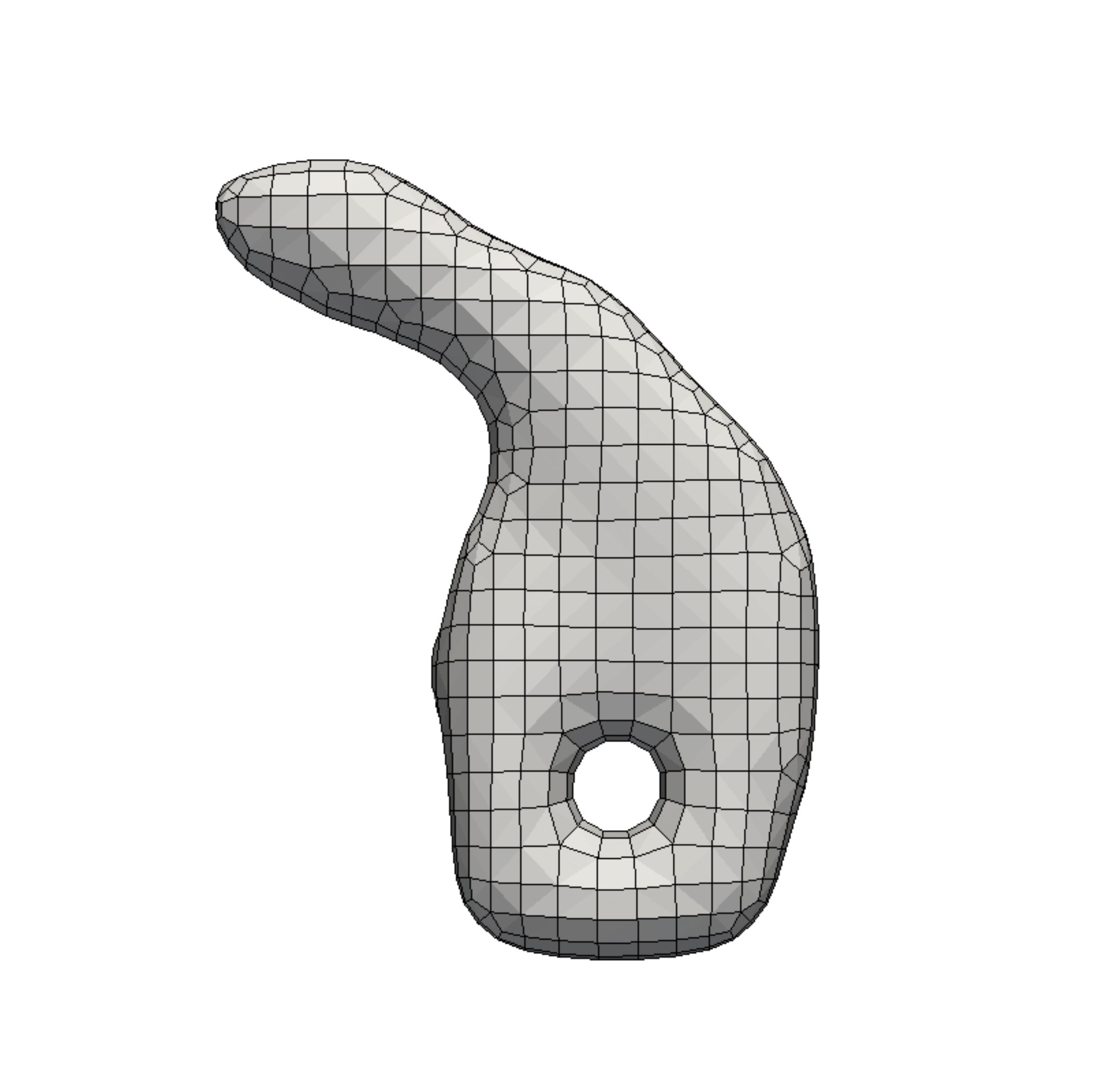} & \includegraphics[width=0.25\textwidth]{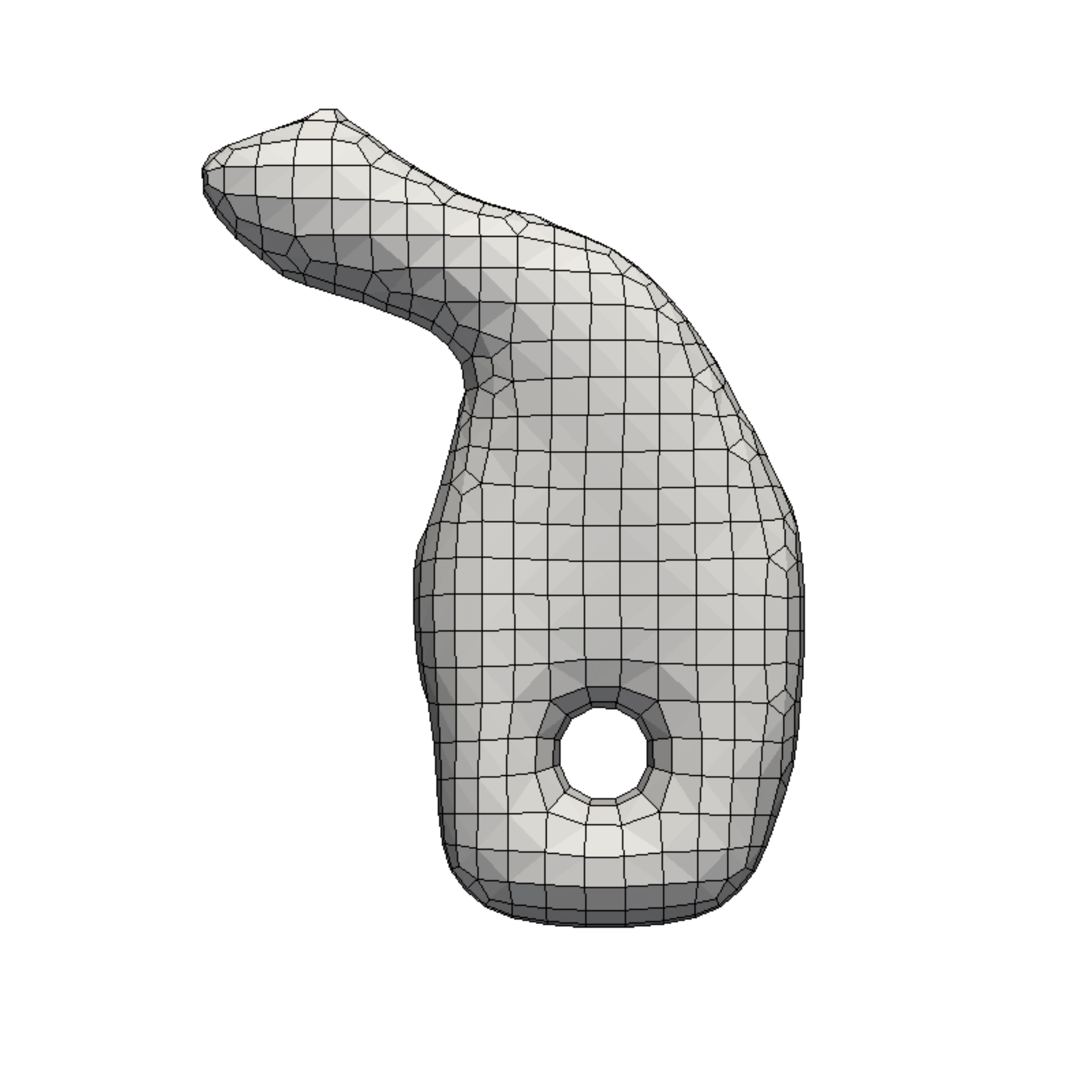}\\

\hline
\end{tabular}

}
        \subcaption{disc simple}
    \end{subtable}
    \hfill
    \begin{subtable}{0.44\textwidth}
        \centering
        \setlength\tabcolsep{6pt}
        \resizebox{0.77\textwidth}{!}{\begin{tabular}{|c|c||c|}
\hline
\parbox{4em}{\centering UNet} & \parbox{4em}{\centering UNet\\+physics} & \xrowht{20pt}\parbox{4em}{\centering ground\\truth} \\
\hline\rule{0pt}{2cm}

\xrowht{20pt} 
\includegraphics[width=0.25\textwidth]{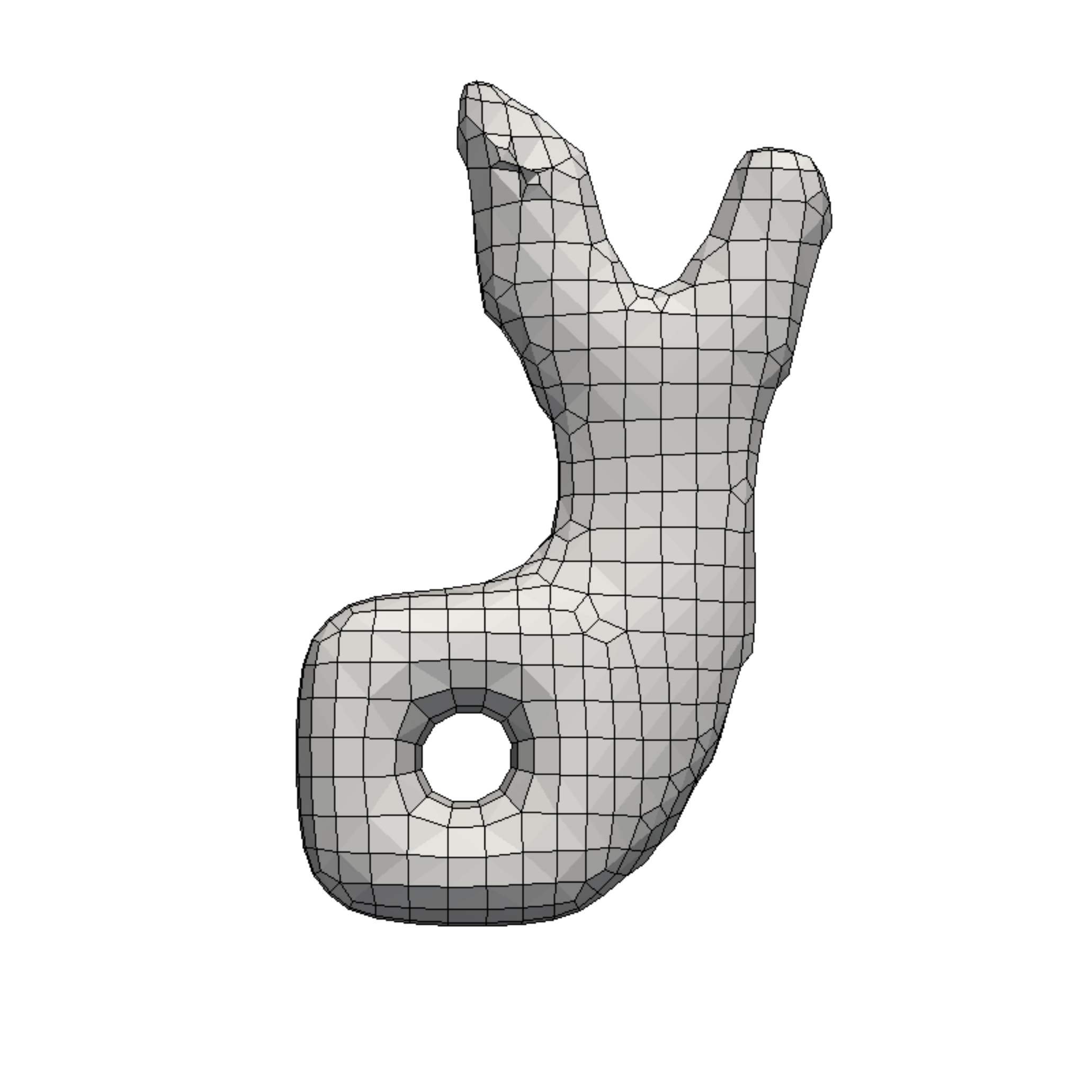} & \includegraphics[width=0.25\textwidth]{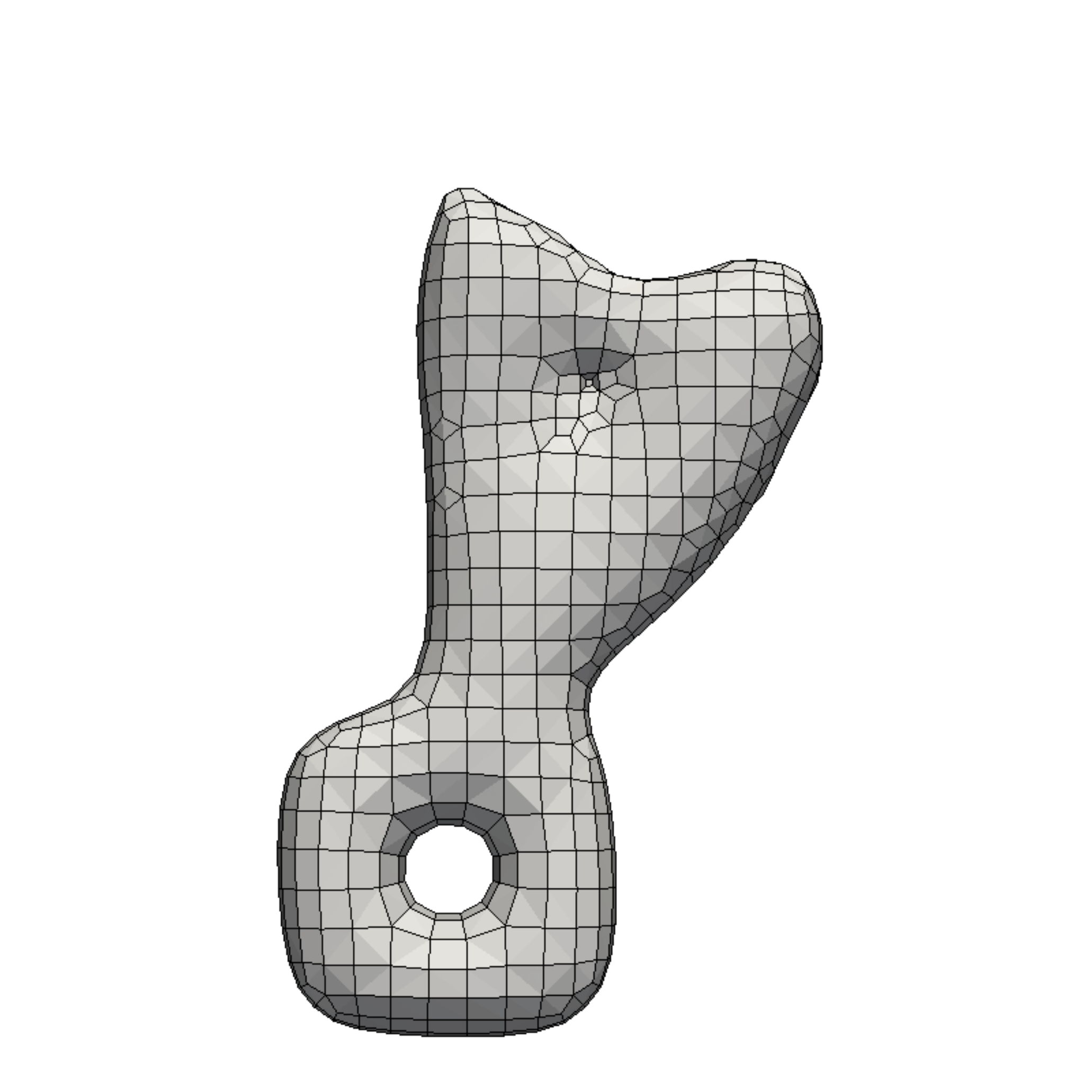} & \includegraphics[width=0.25\textwidth]{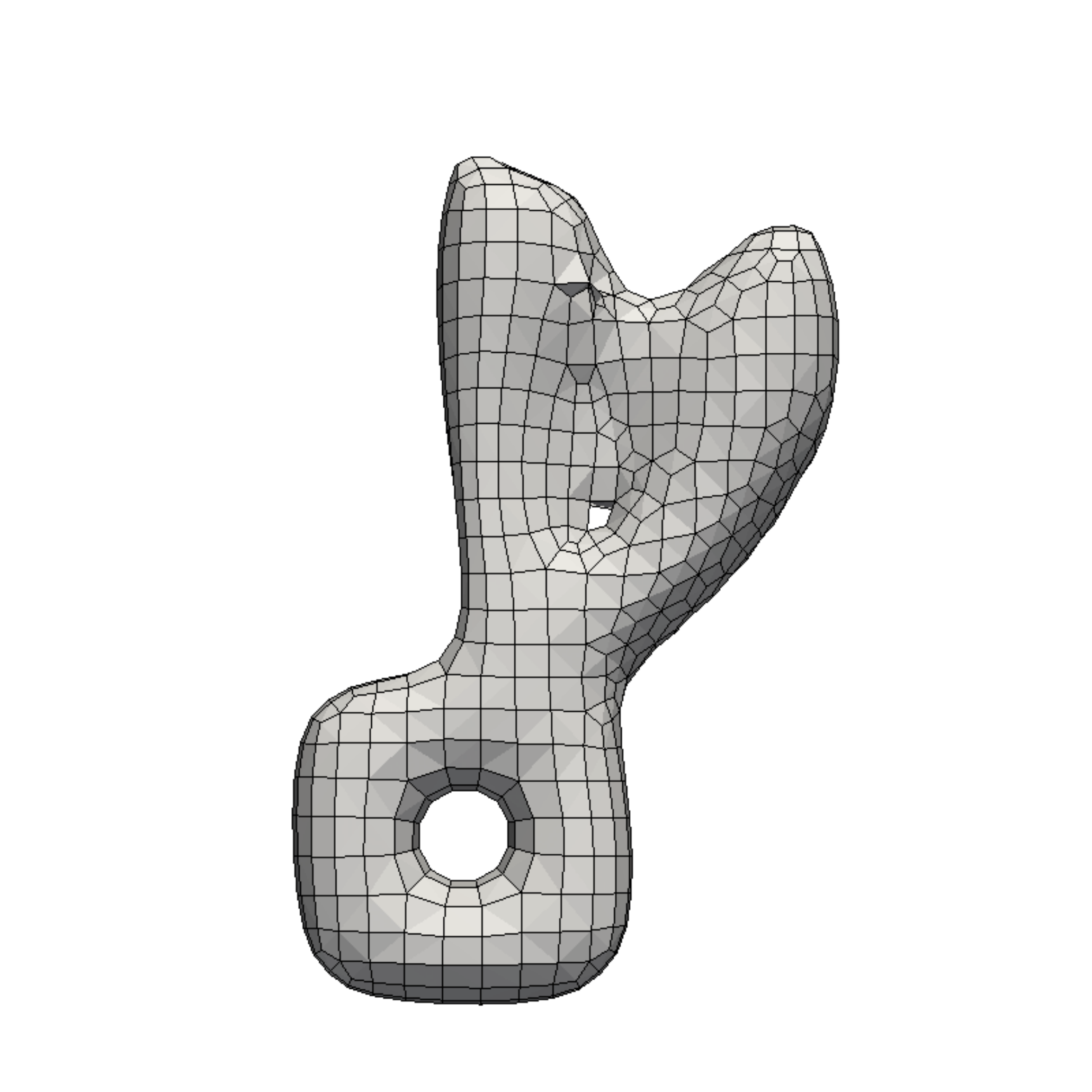}\\

\xrowht{20pt} 
\includegraphics[width=0.25\textwidth]{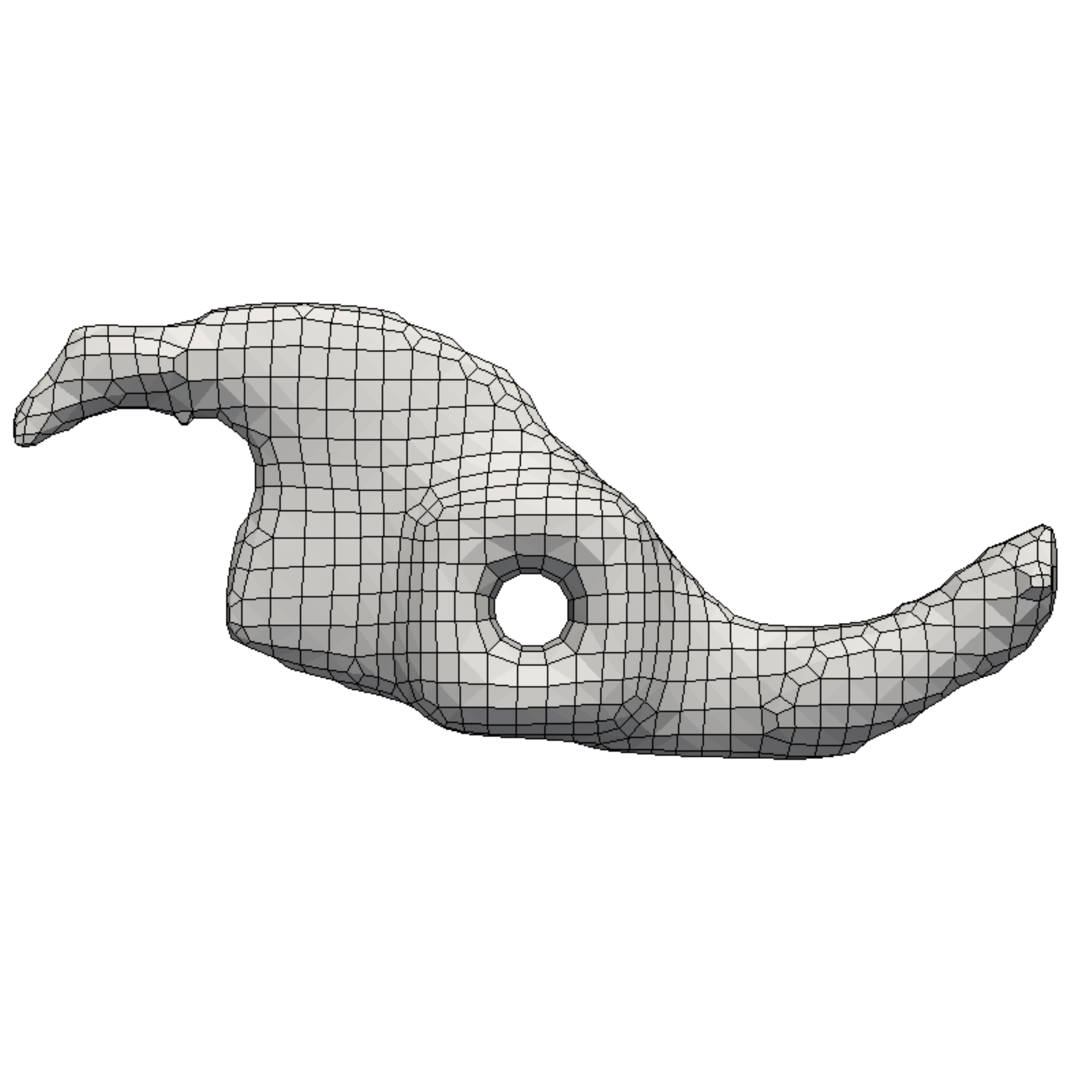} & \includegraphics[width=0.25\textwidth]{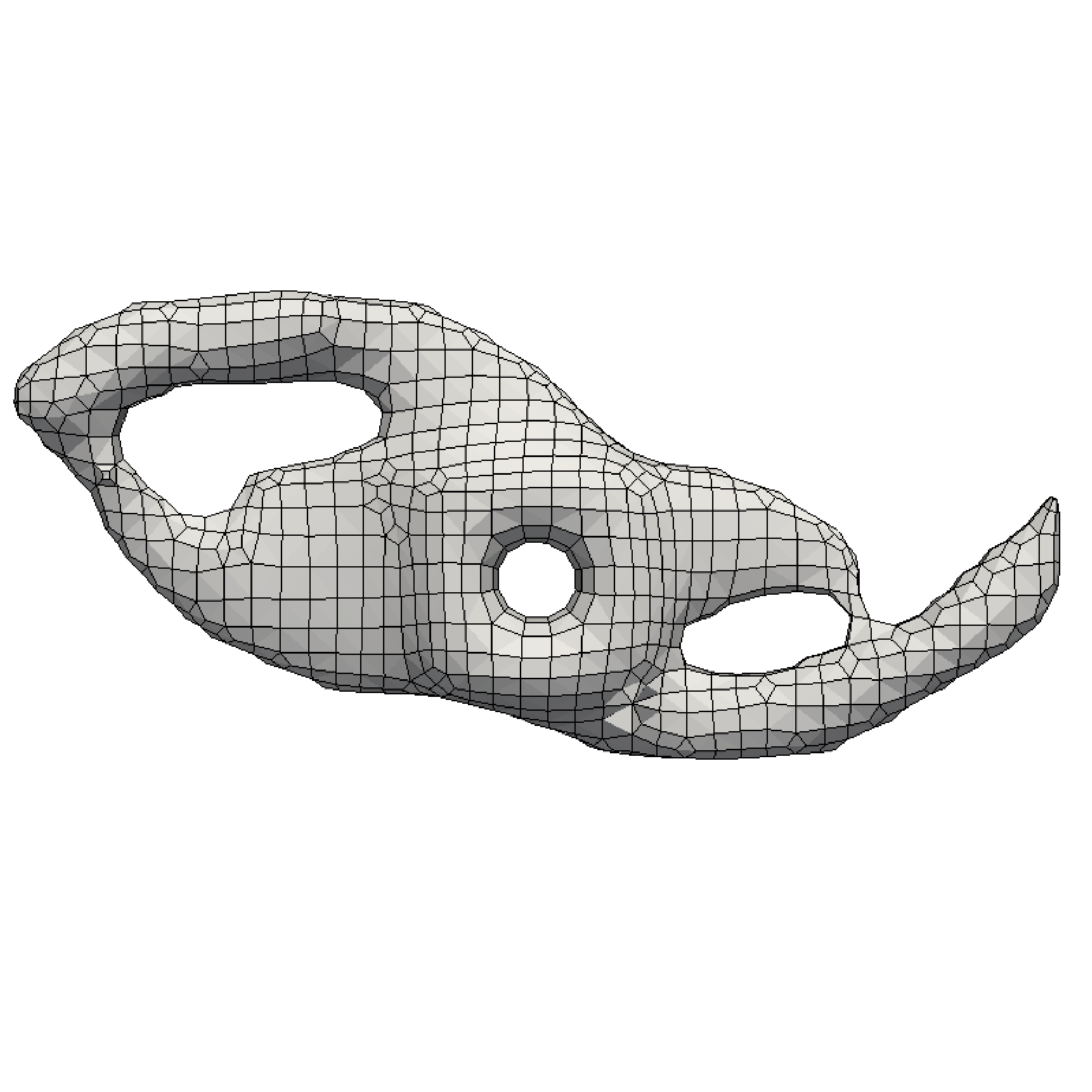} & \includegraphics[width=0.25\textwidth]{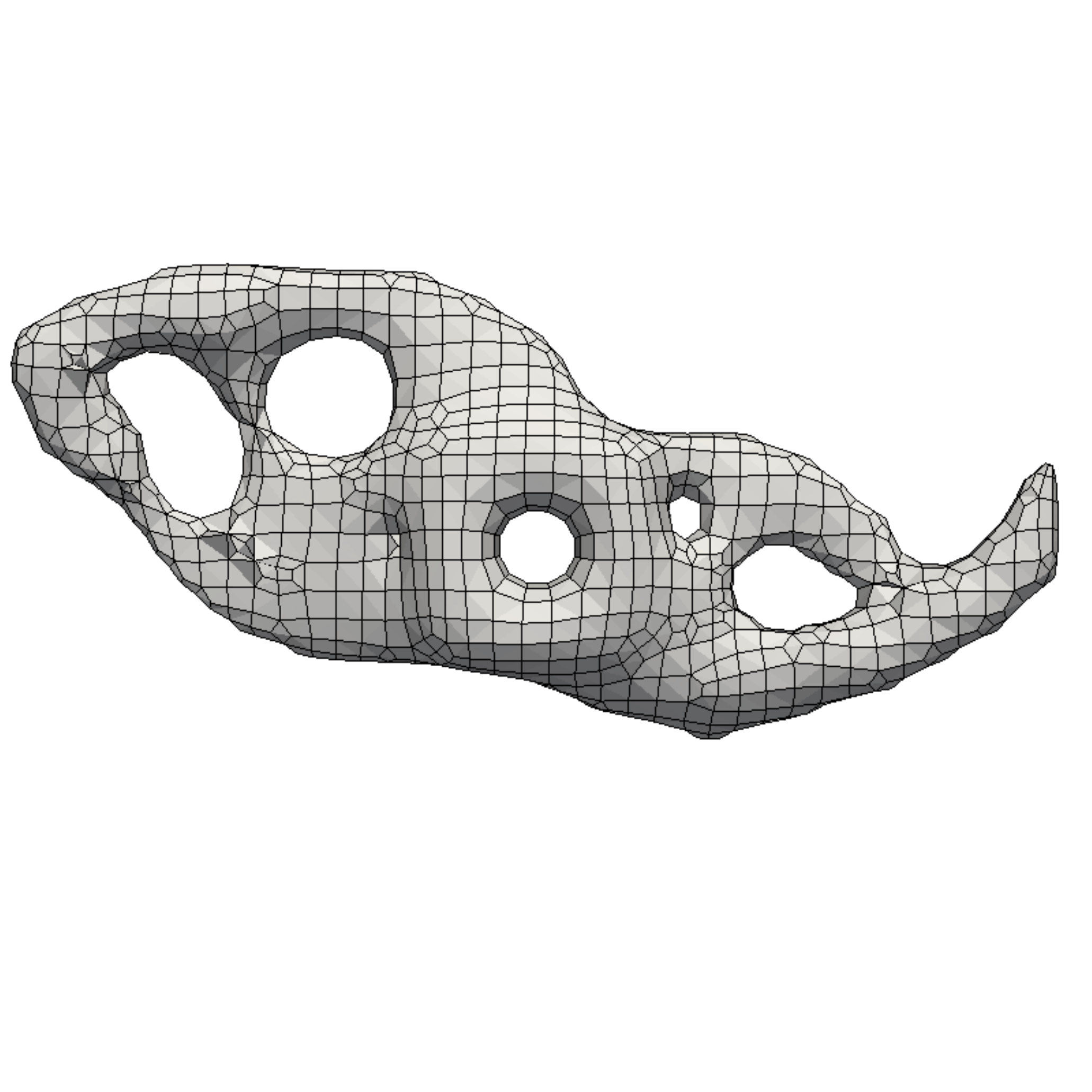}\\

\xrowht{20pt} 
\includegraphics[width=0.25\textwidth]{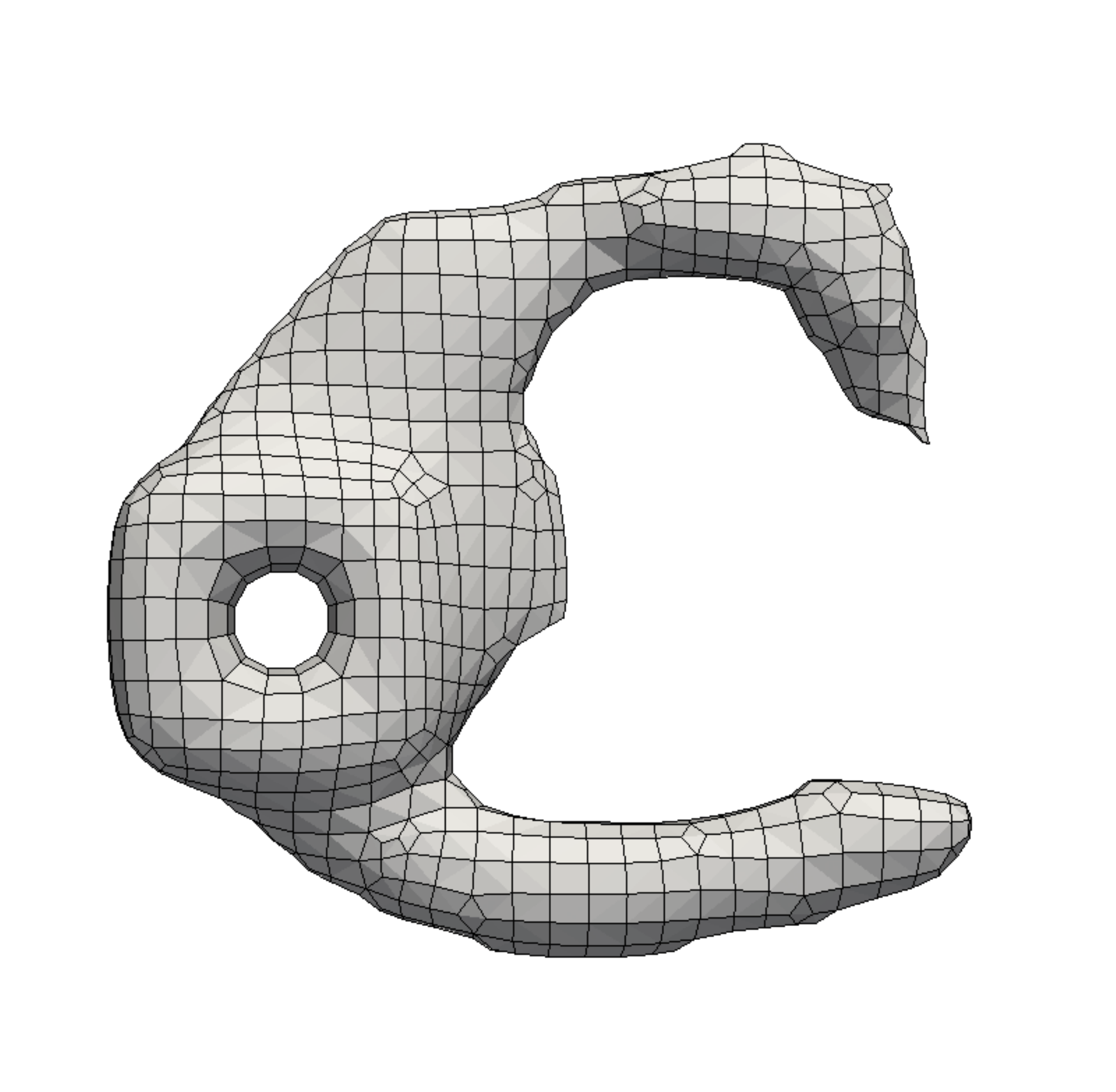} & \includegraphics[width=0.25\textwidth]{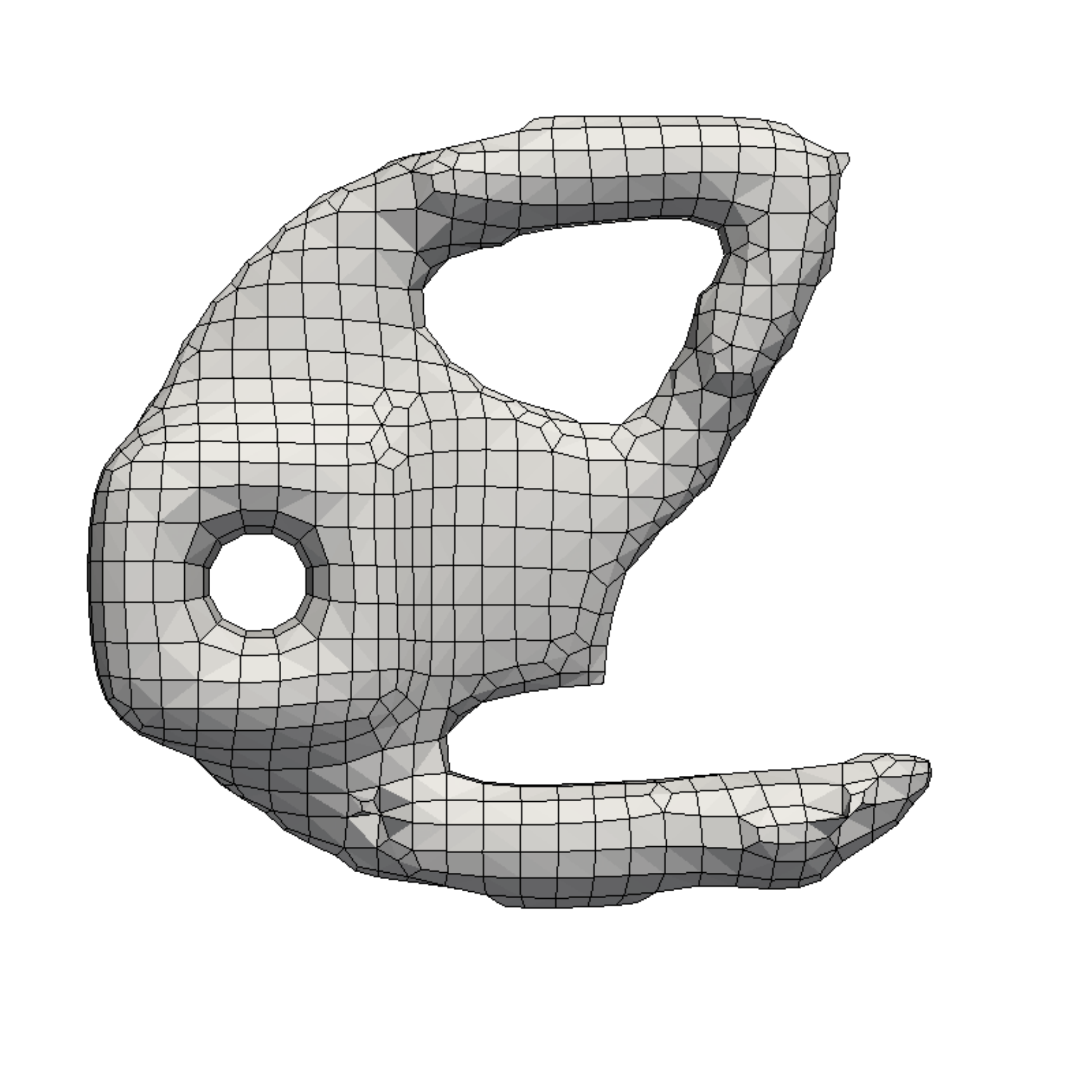} & \includegraphics[width=0.25\textwidth]{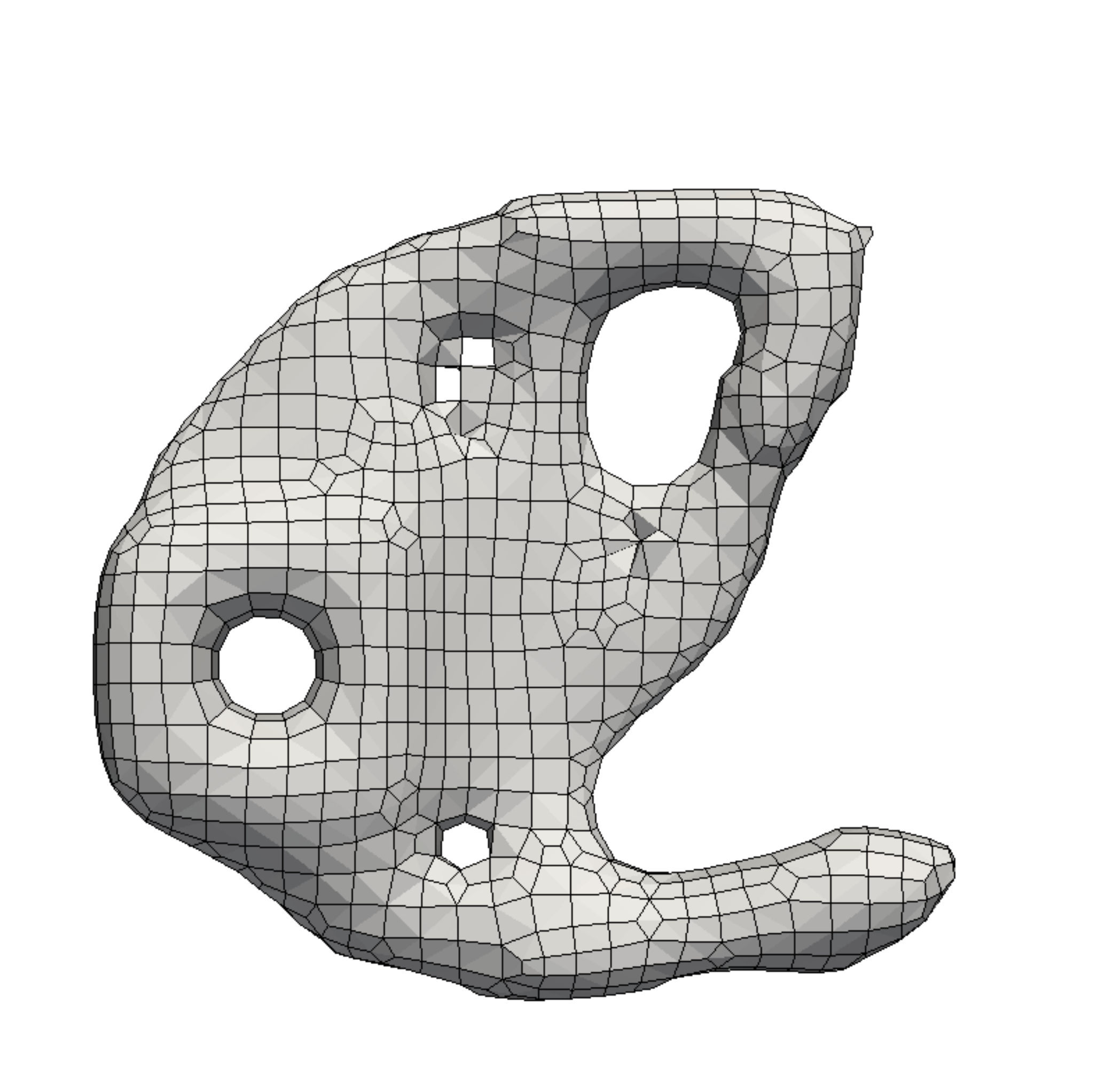}\\

\xrowht{20pt} 
\includegraphics[width=0.25\textwidth]{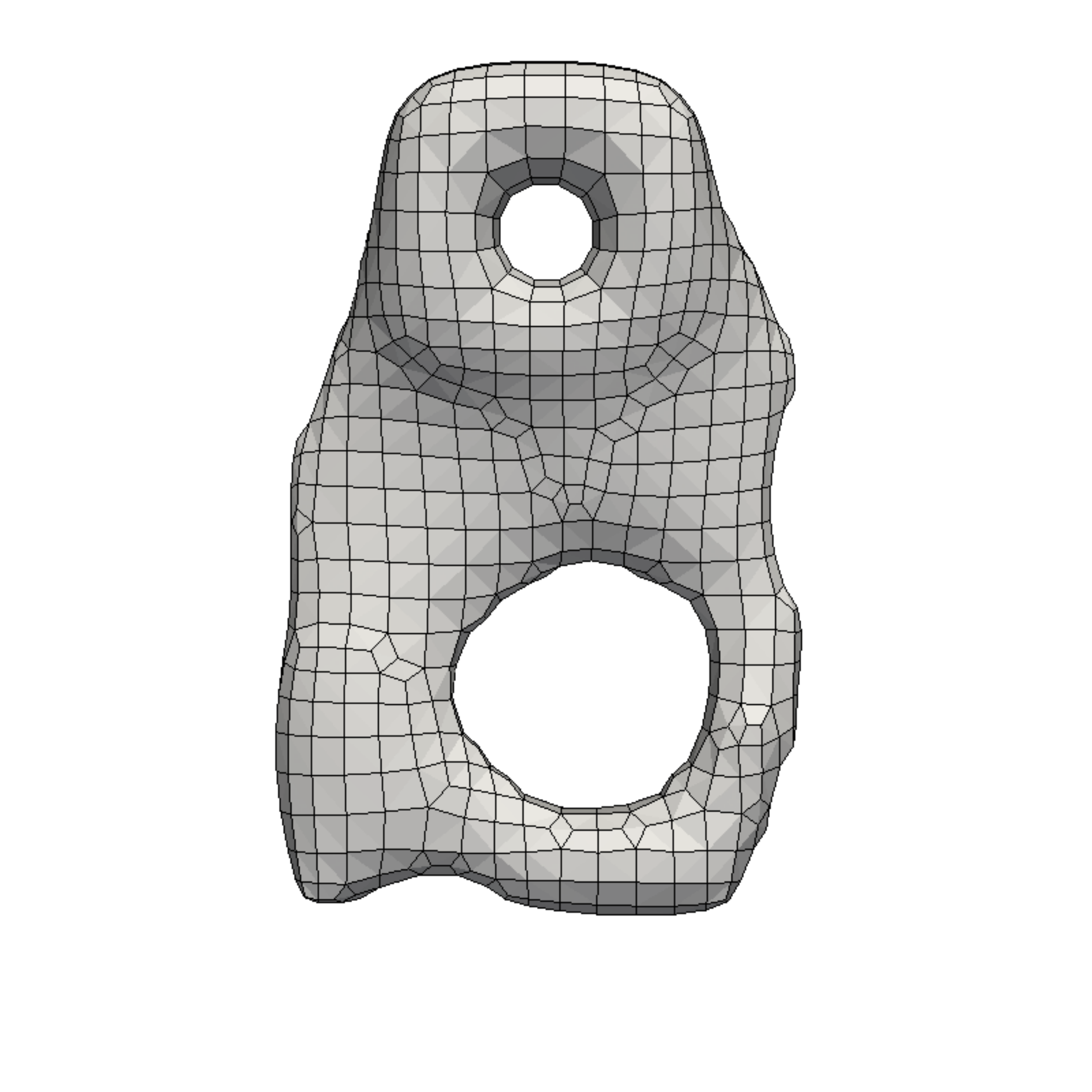} & \includegraphics[width=0.25\textwidth]{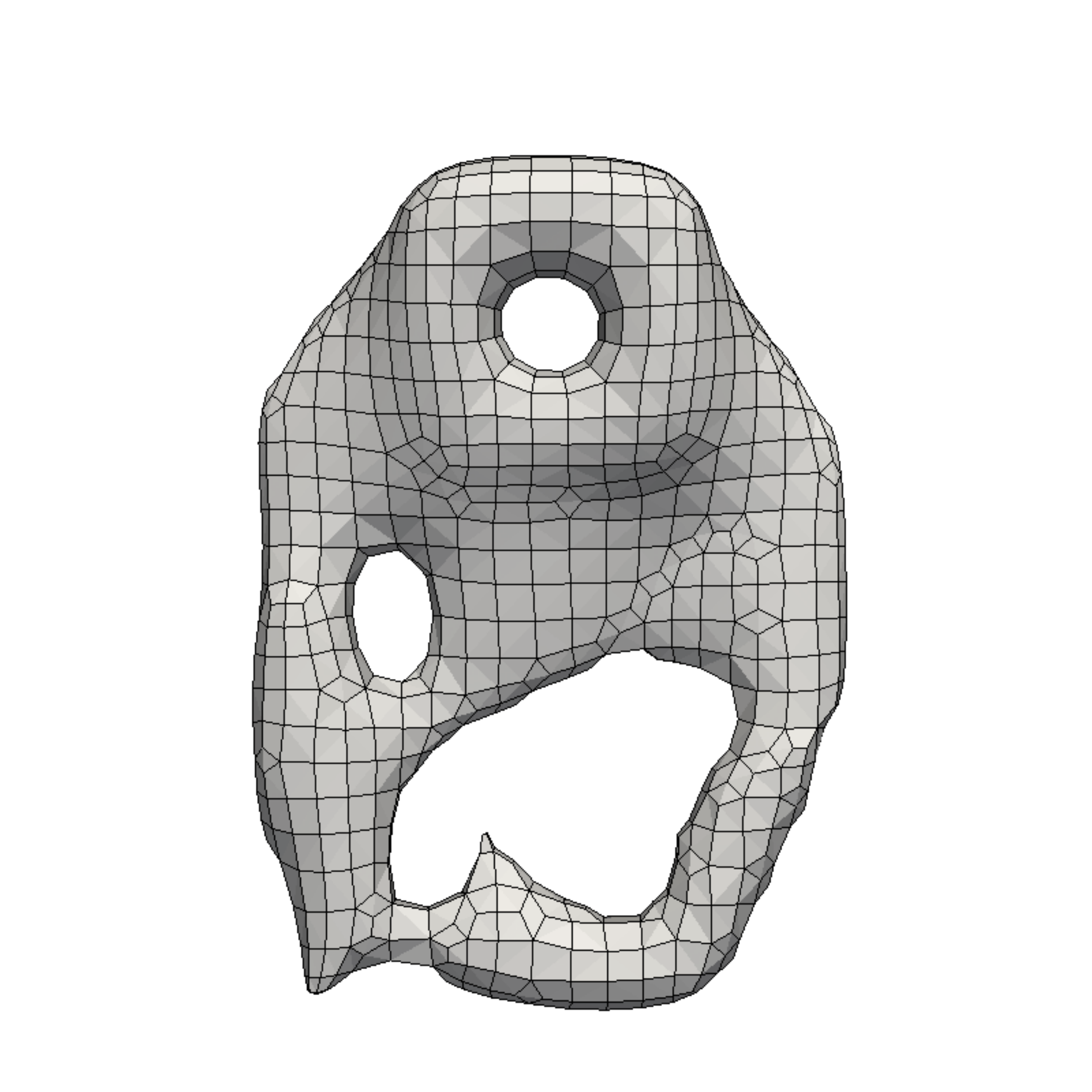} & \includegraphics[width=0.25\textwidth]{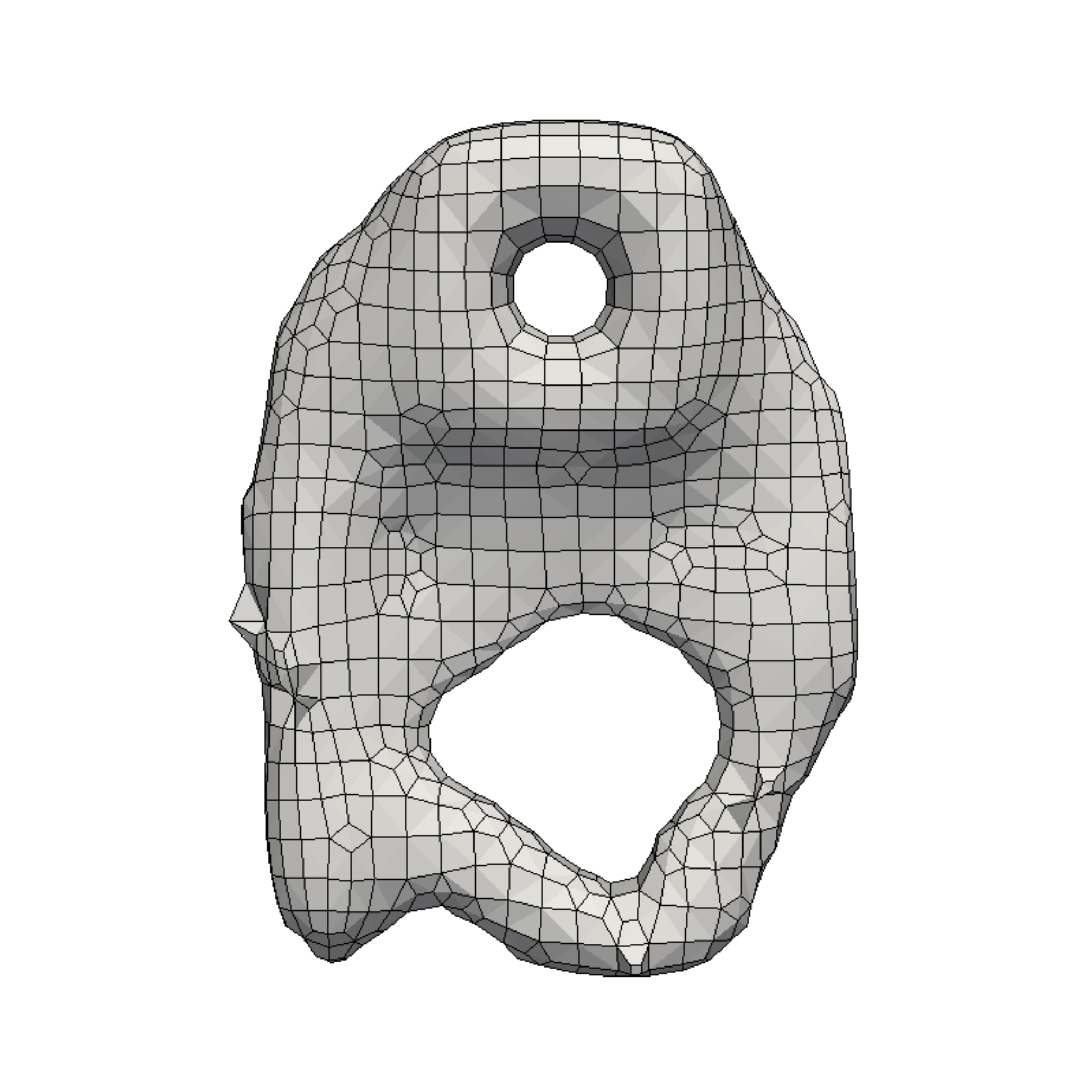}\\

\xrowht{20pt} 
\includegraphics[width=0.25\textwidth]{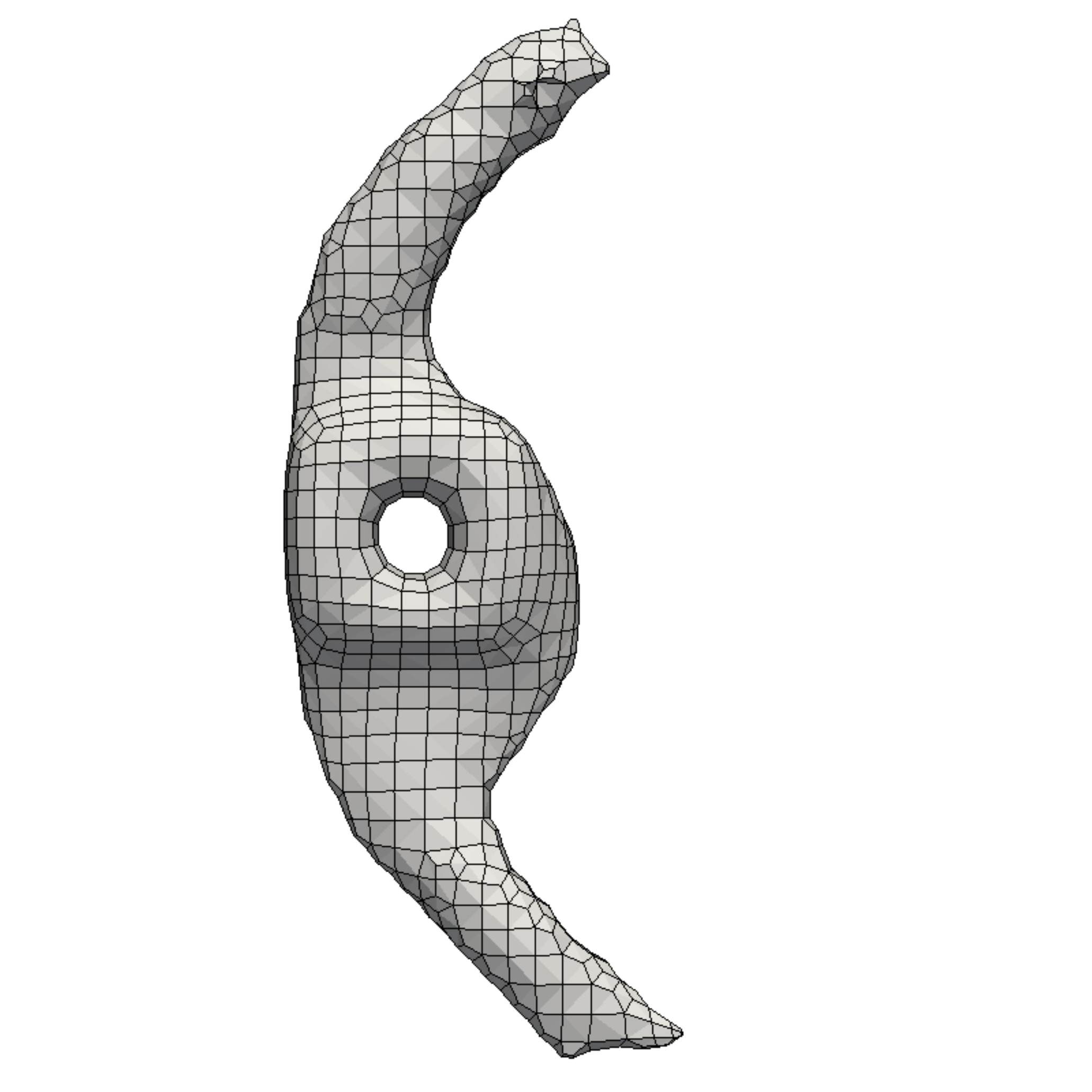} & \includegraphics[width=0.25\textwidth]{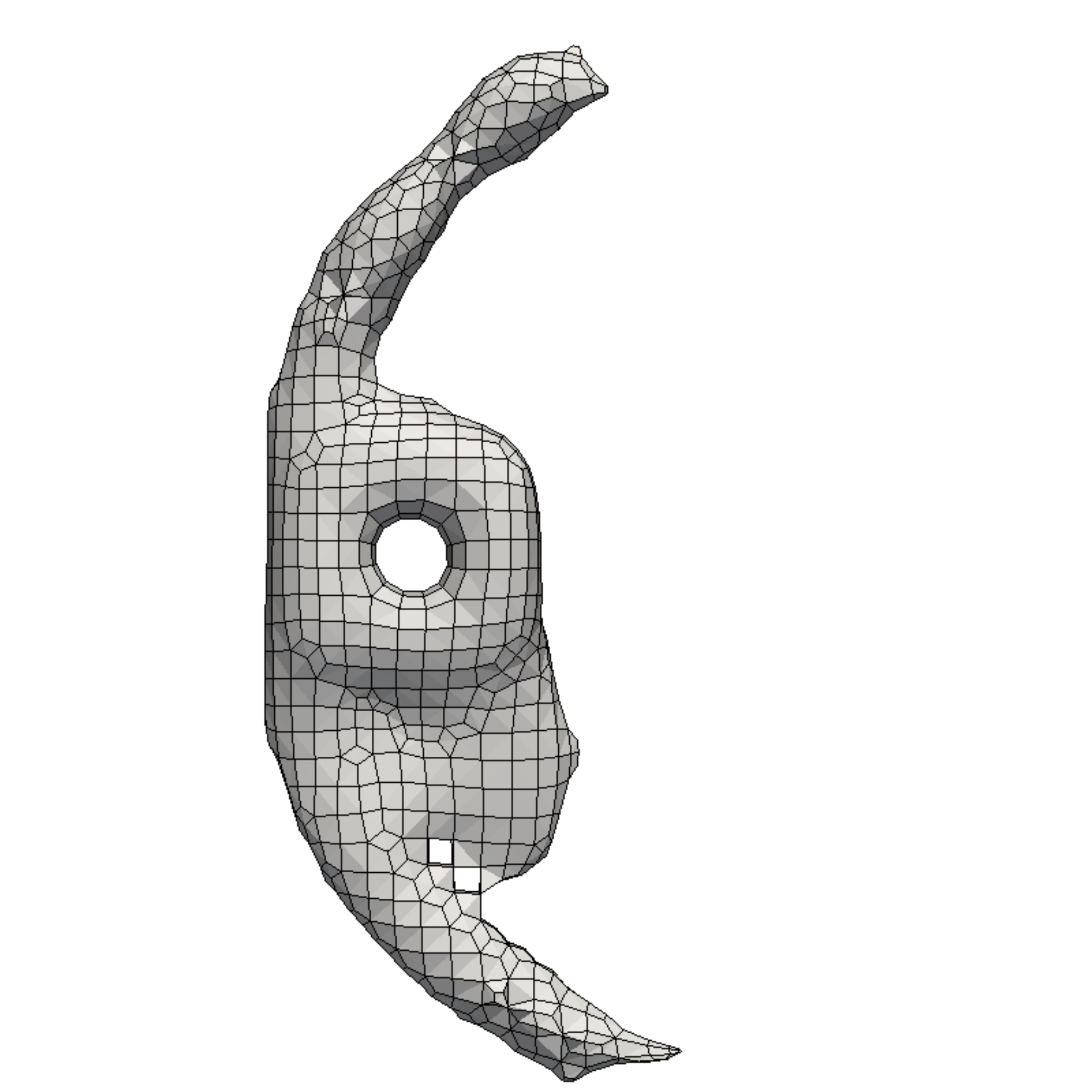} & \includegraphics[width=0.25\textwidth]{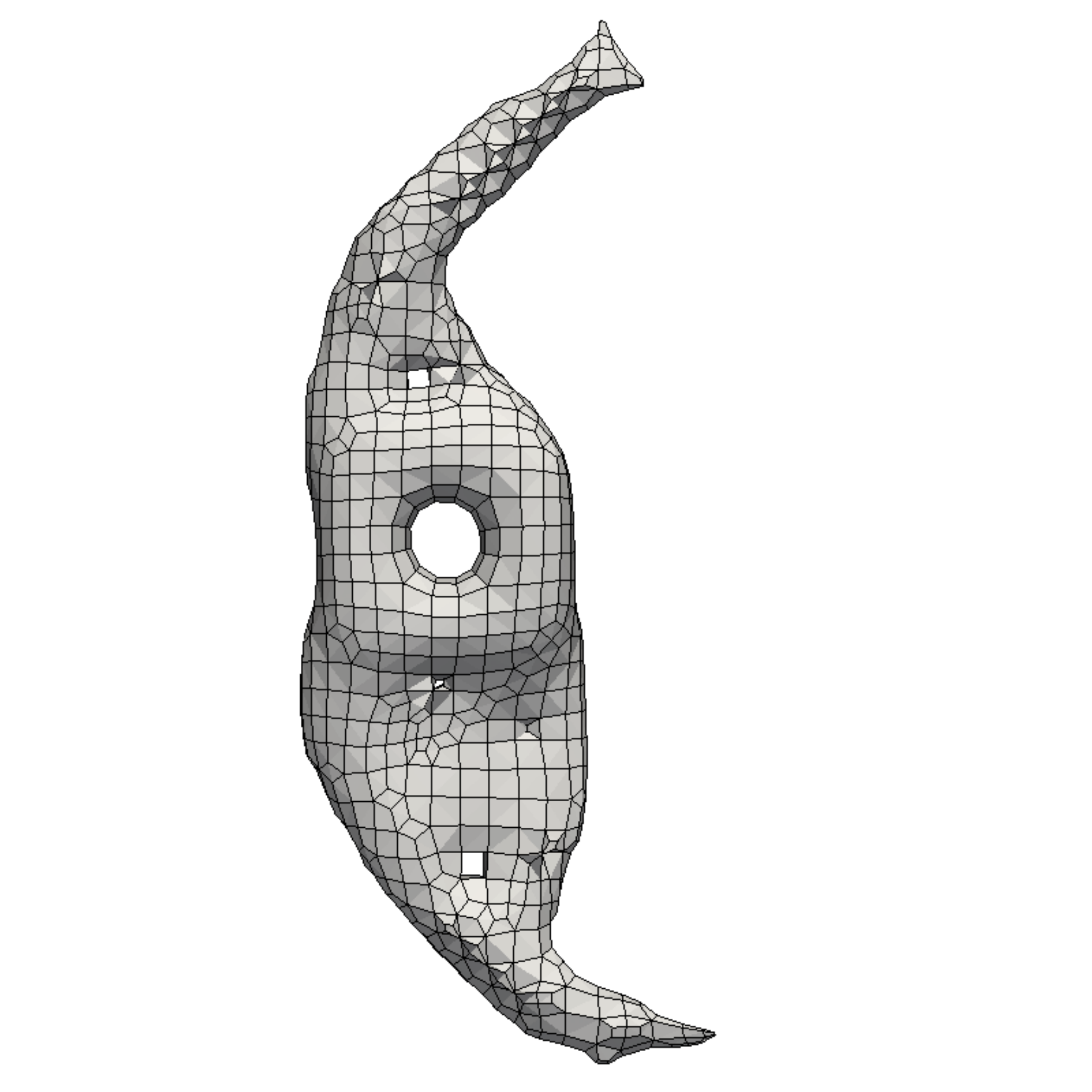}\\

\xrowht{20pt} 
\includegraphics[width=0.25\textwidth]{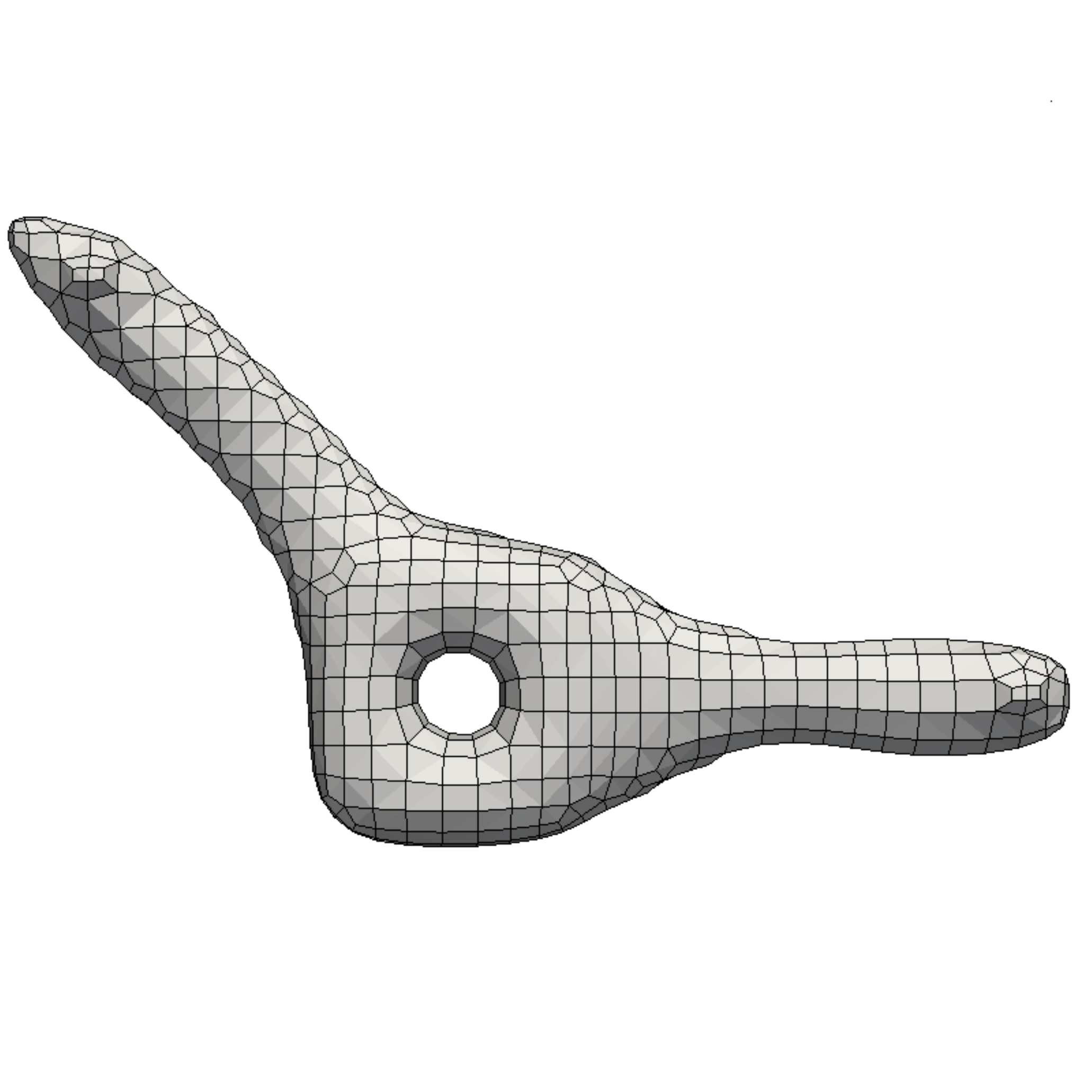} & \includegraphics[width=0.25\textwidth]{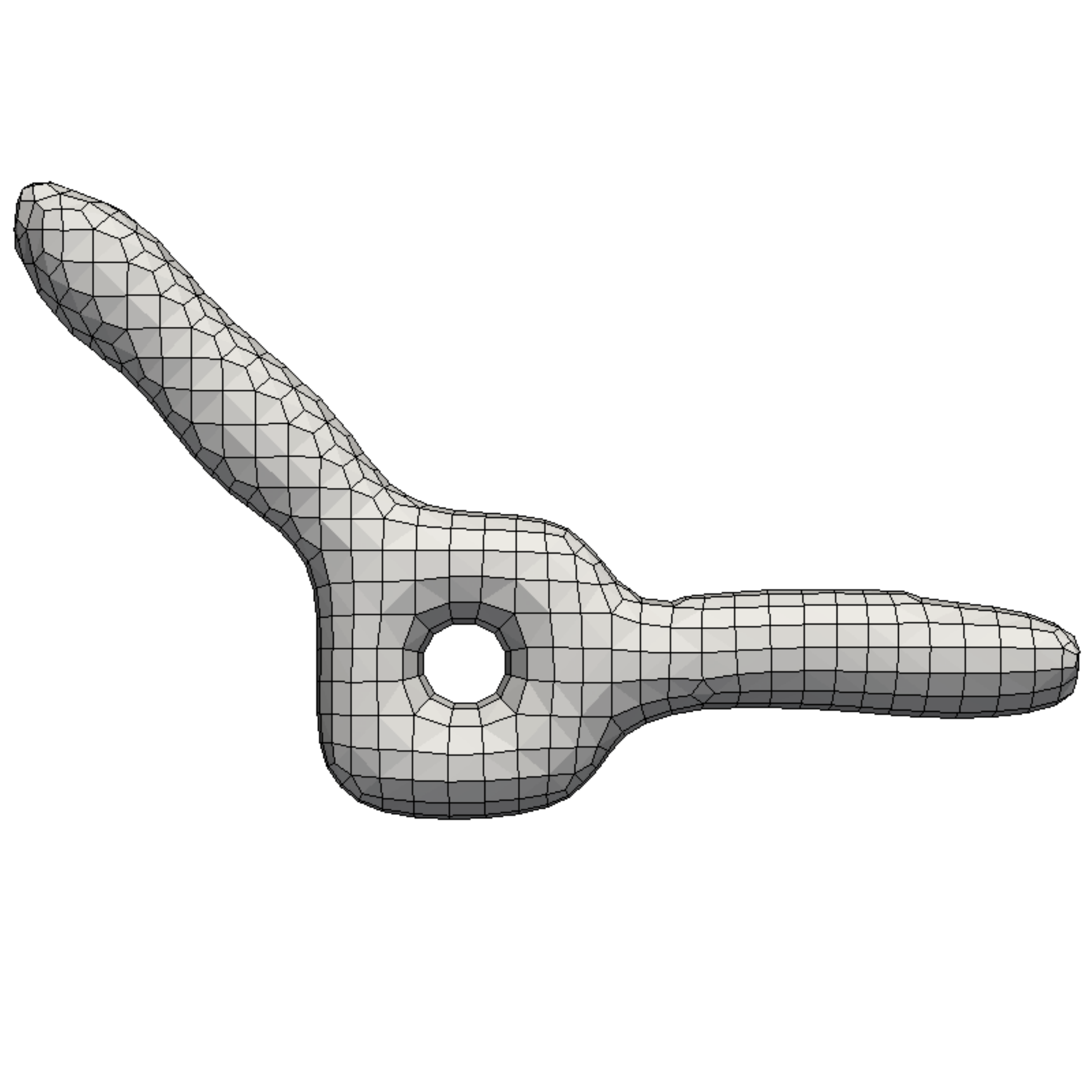} & \includegraphics[width=0.25\textwidth]{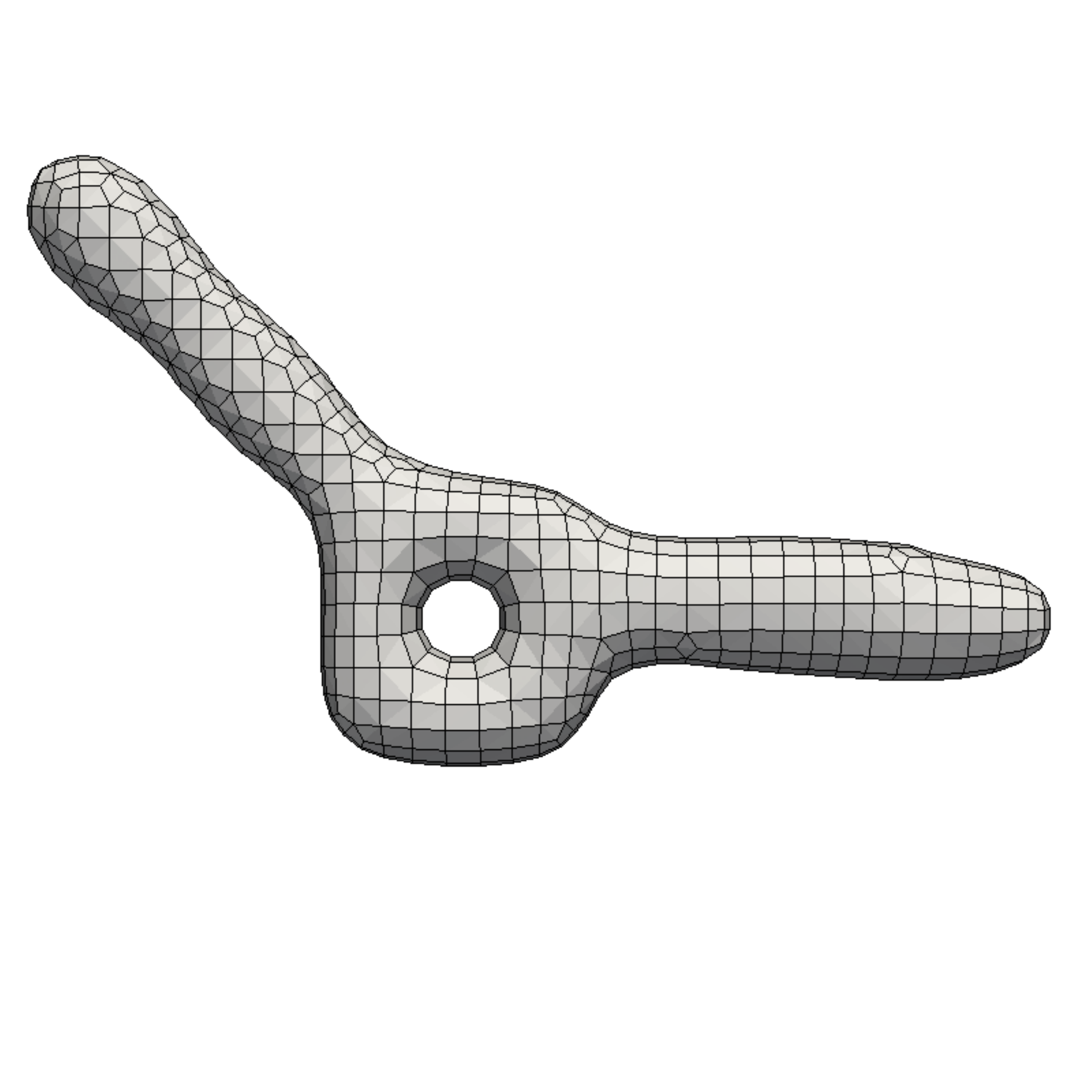}\\

\xrowht{20pt} 
\includegraphics[width=0.25\textwidth]{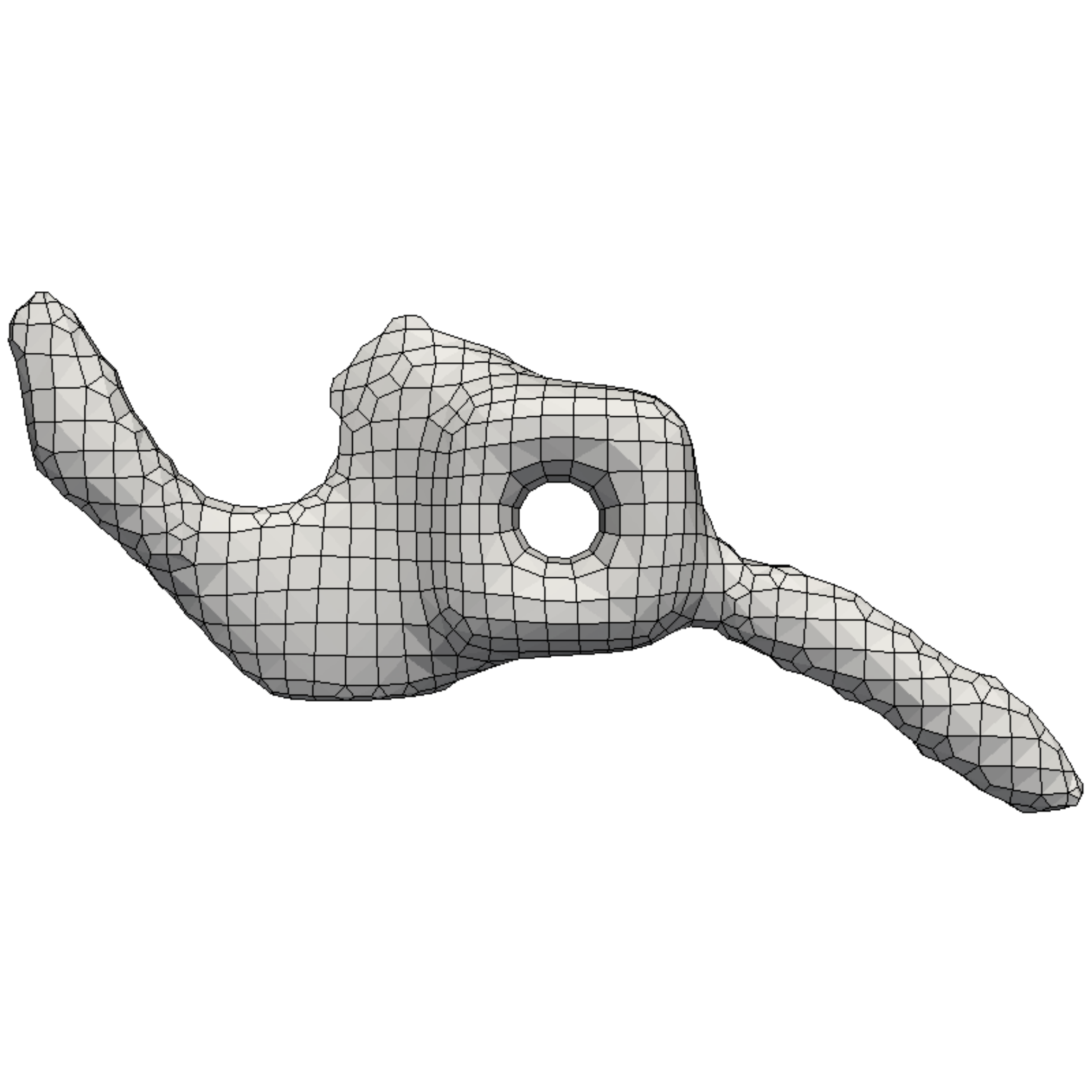} & \includegraphics[width=0.25\textwidth]{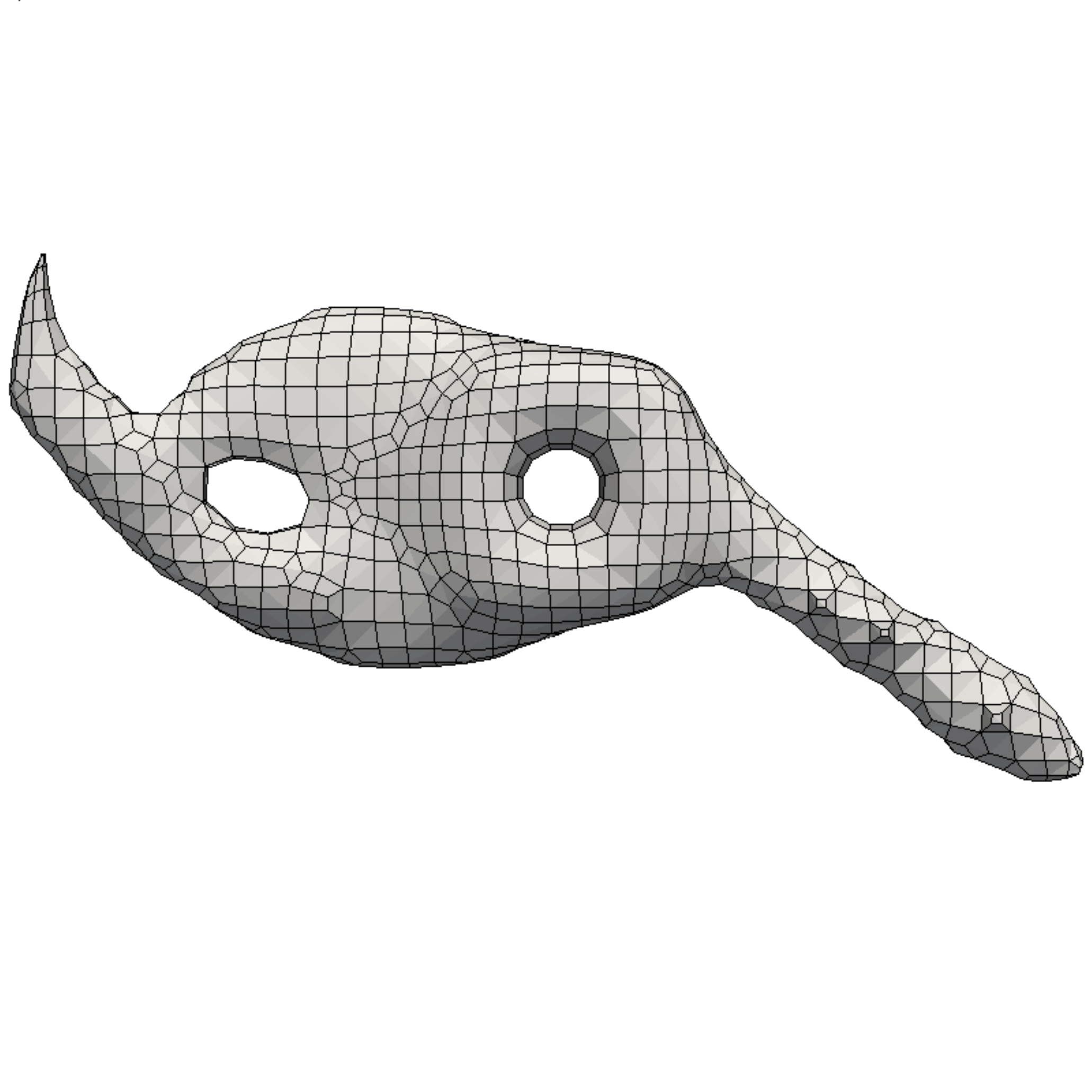} & \includegraphics[width=0.25\textwidth]{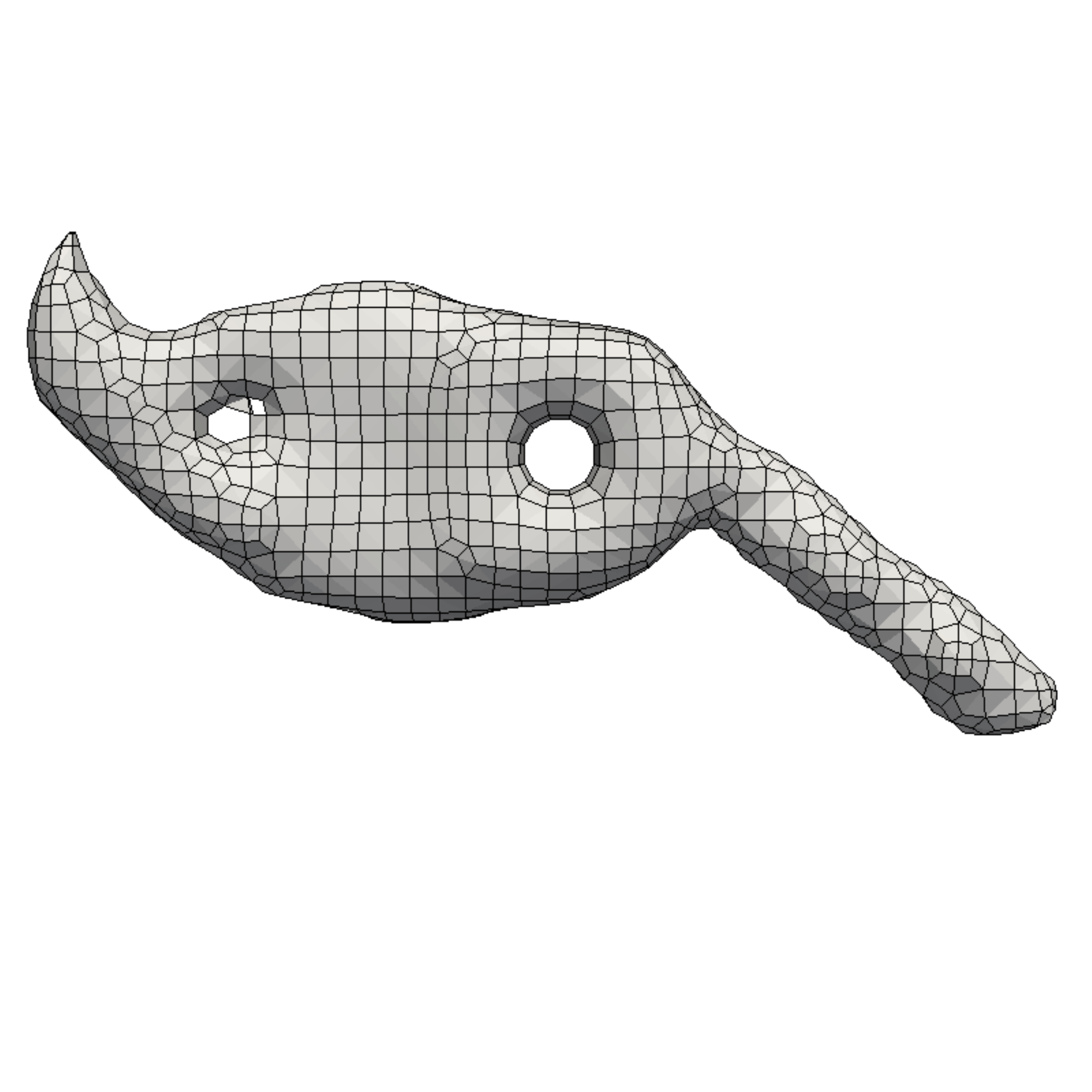}\\

\xrowht{20pt} 
\includegraphics[width=0.25\textwidth]{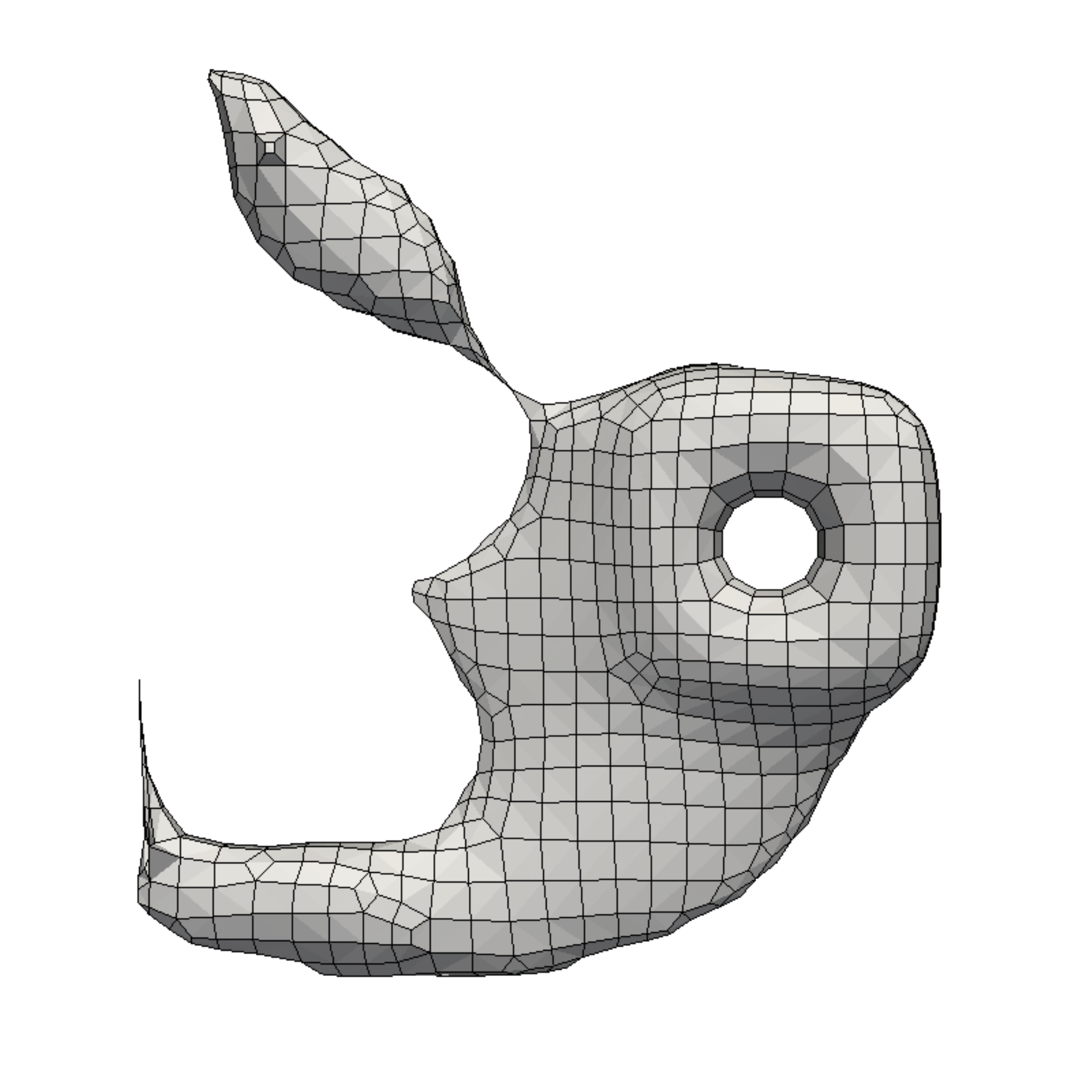} & \includegraphics[width=0.25\textwidth]{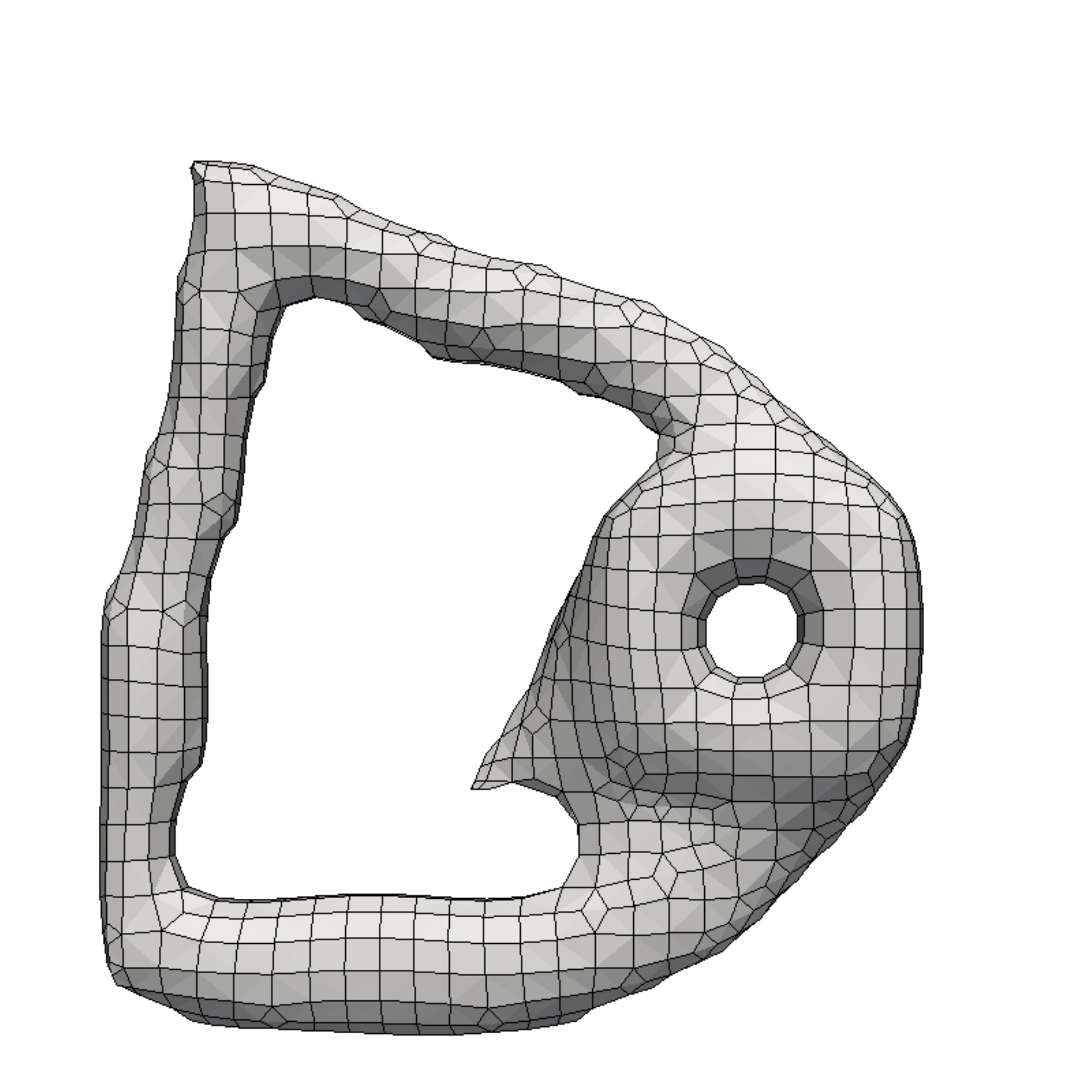} & \includegraphics[width=0.25\textwidth]{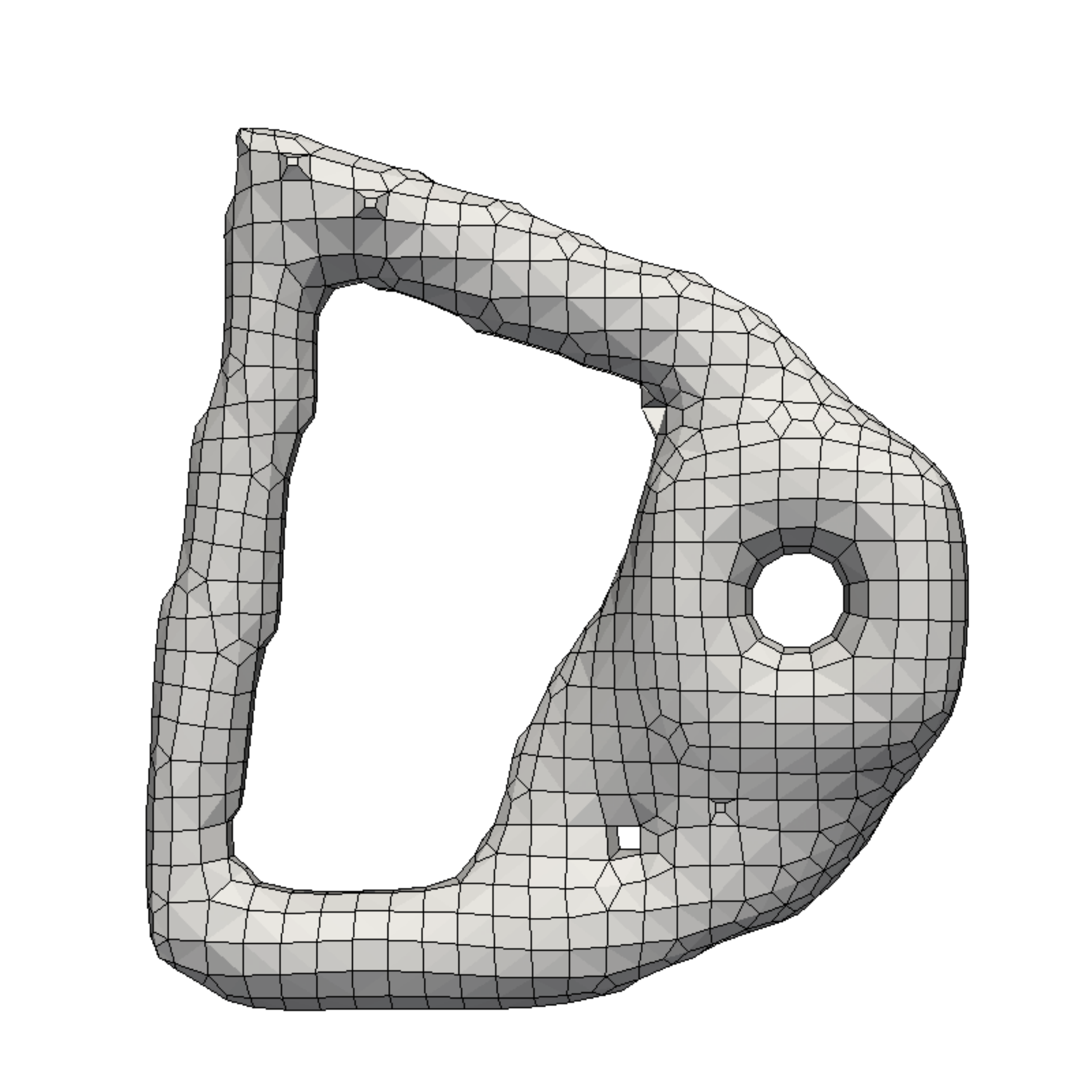}\\

\xrowht{20pt} 
\includegraphics[width=0.25\textwidth]{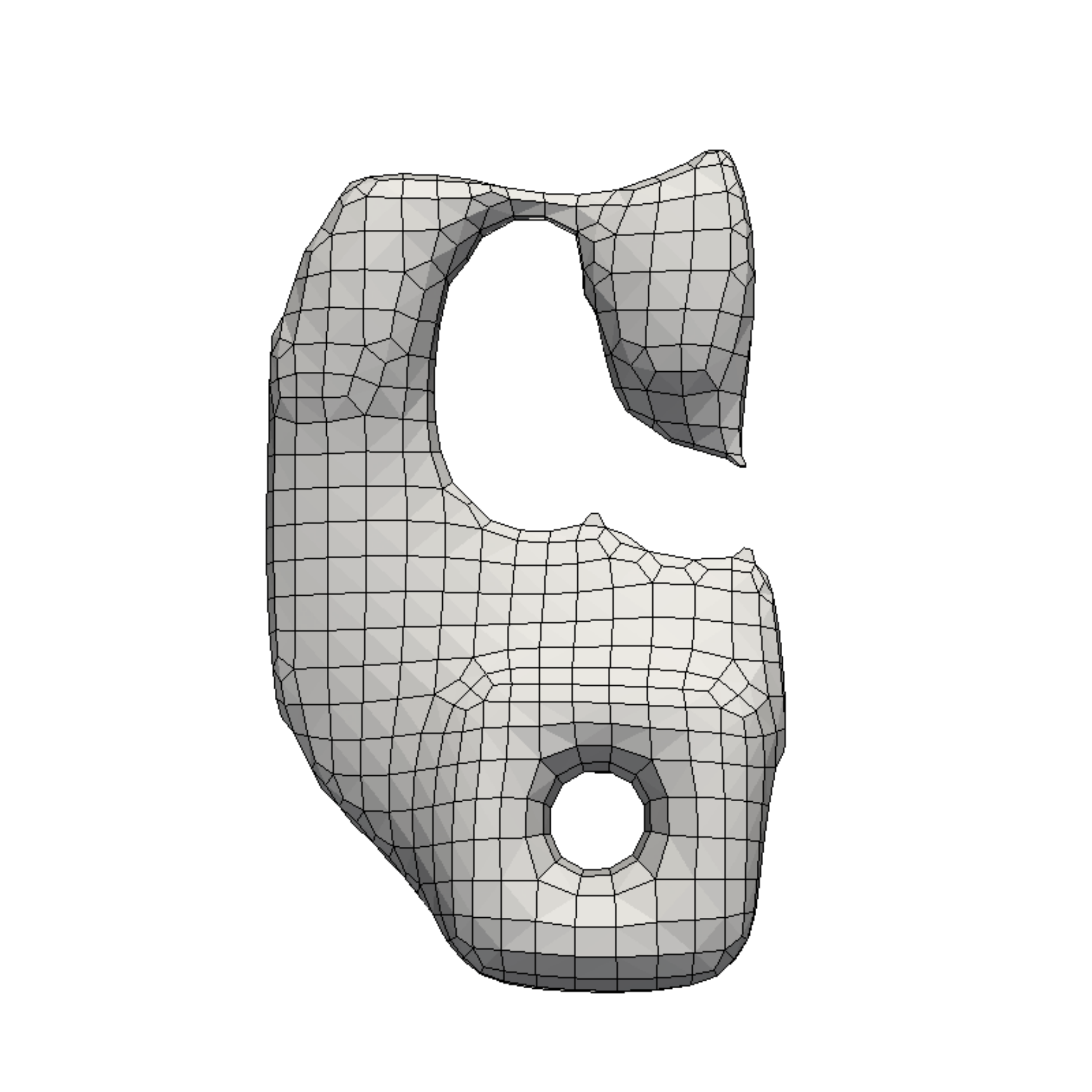} & \includegraphics[width=0.25\textwidth]{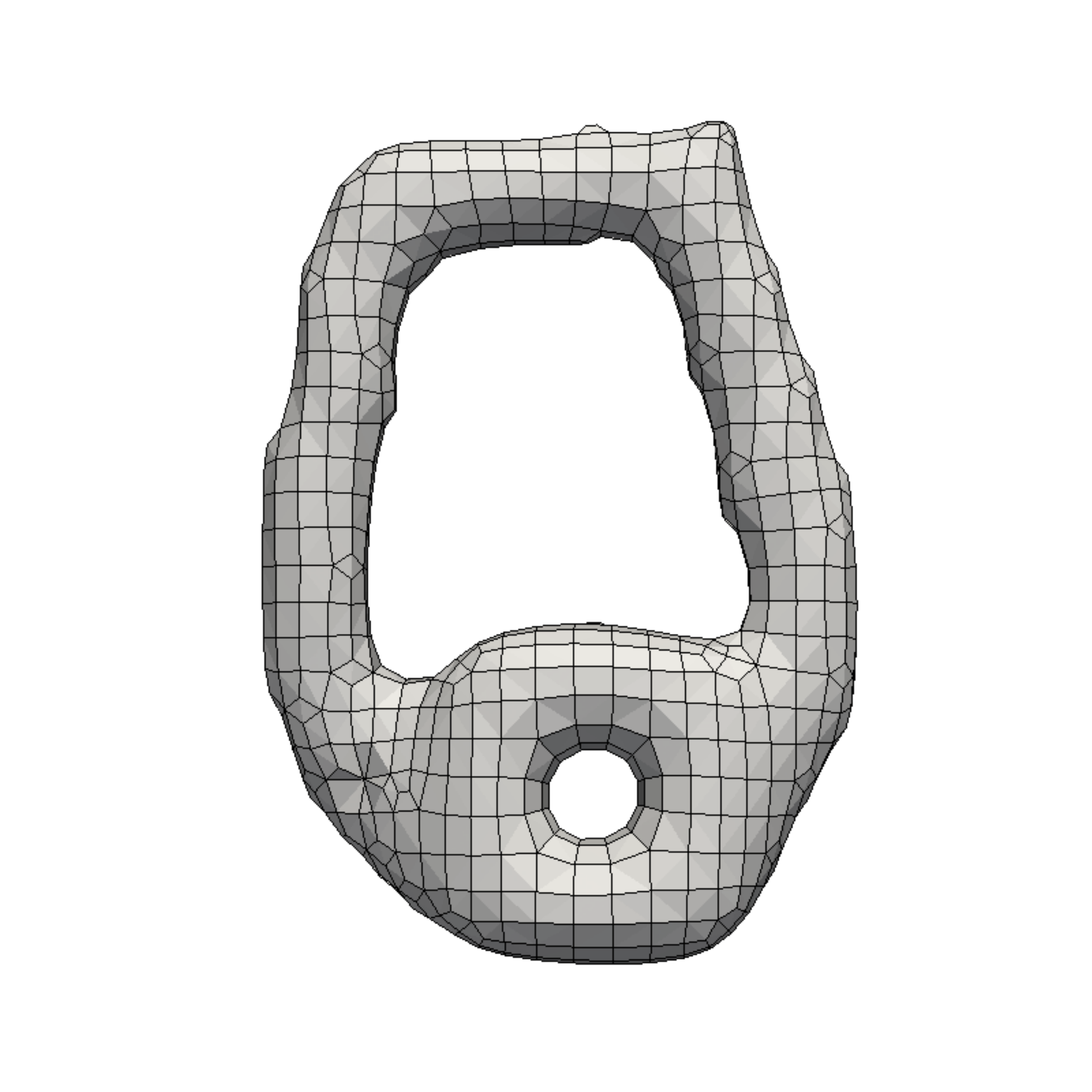} & \includegraphics[width=0.25\textwidth]{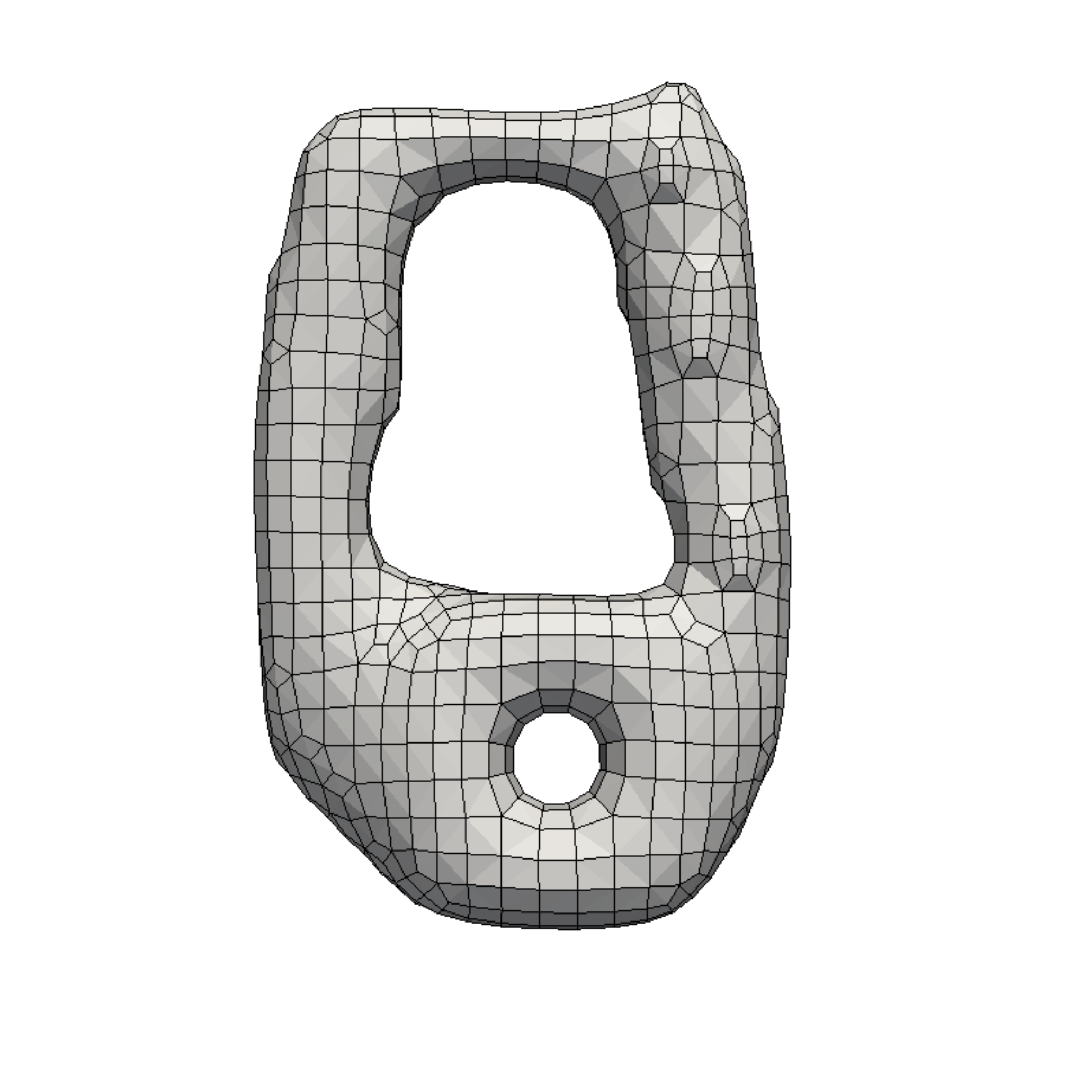}\\

\xrowht{20pt} 
\includegraphics[width=0.25\textwidth]{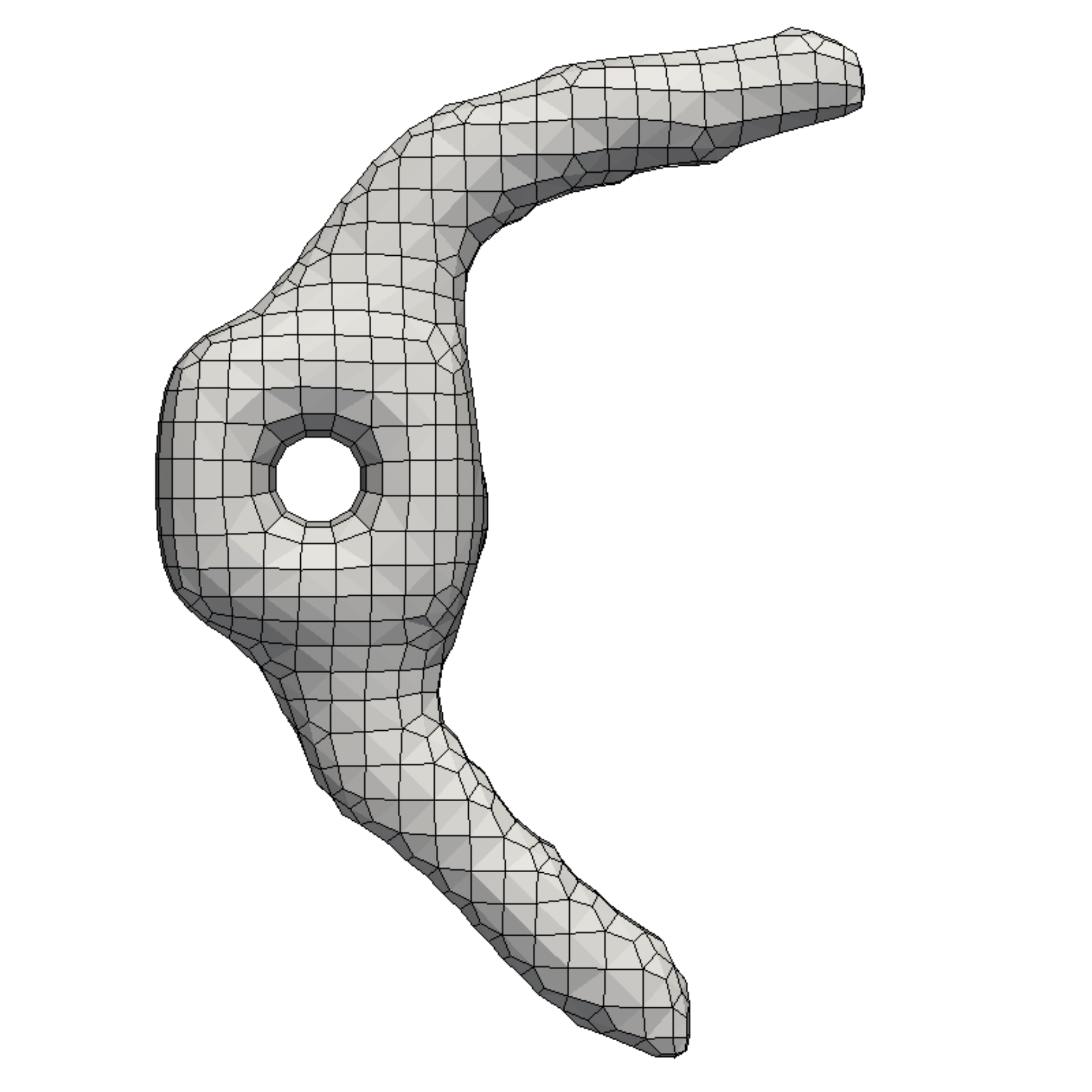} & \includegraphics[width=0.25\textwidth]{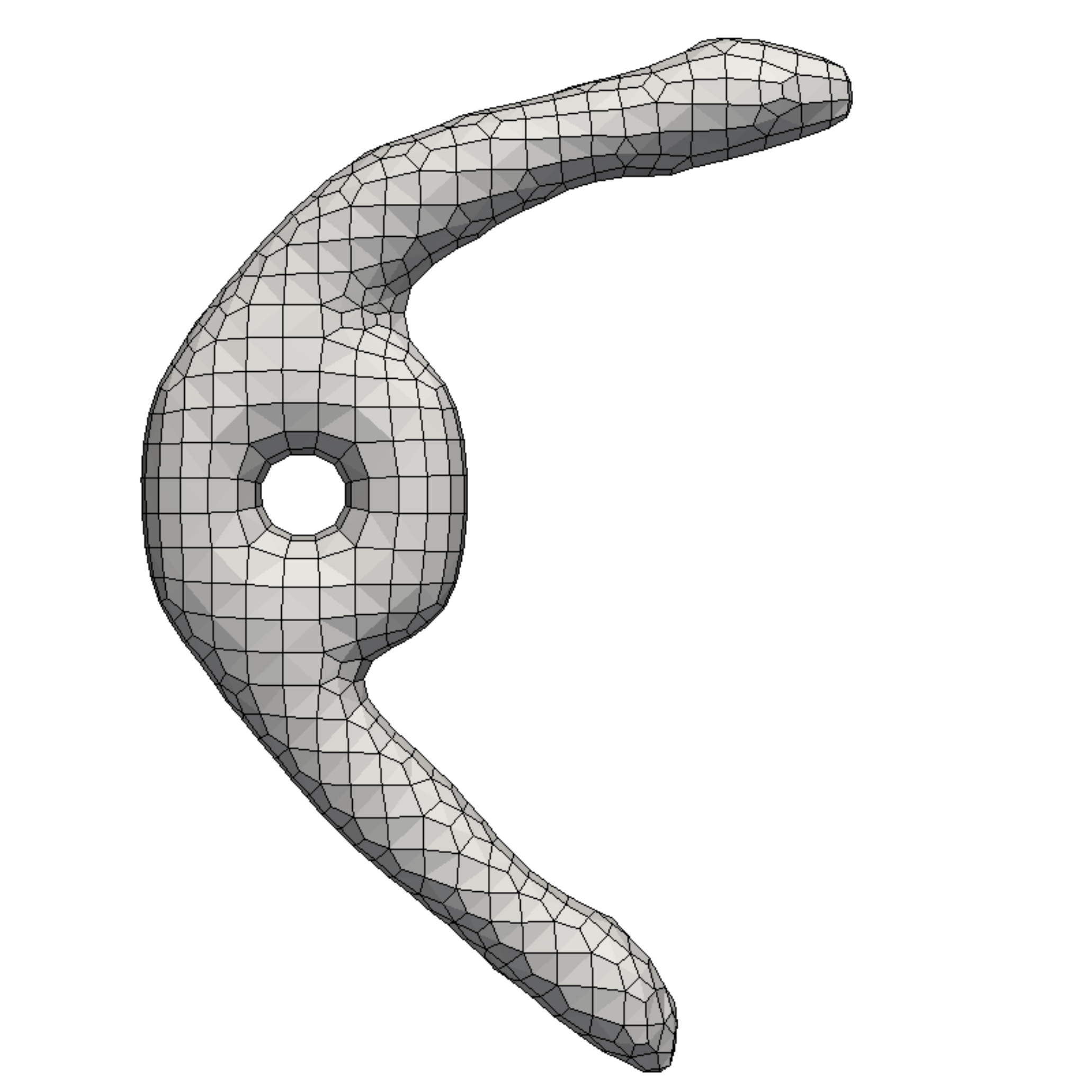} & \includegraphics[width=0.25\textwidth]{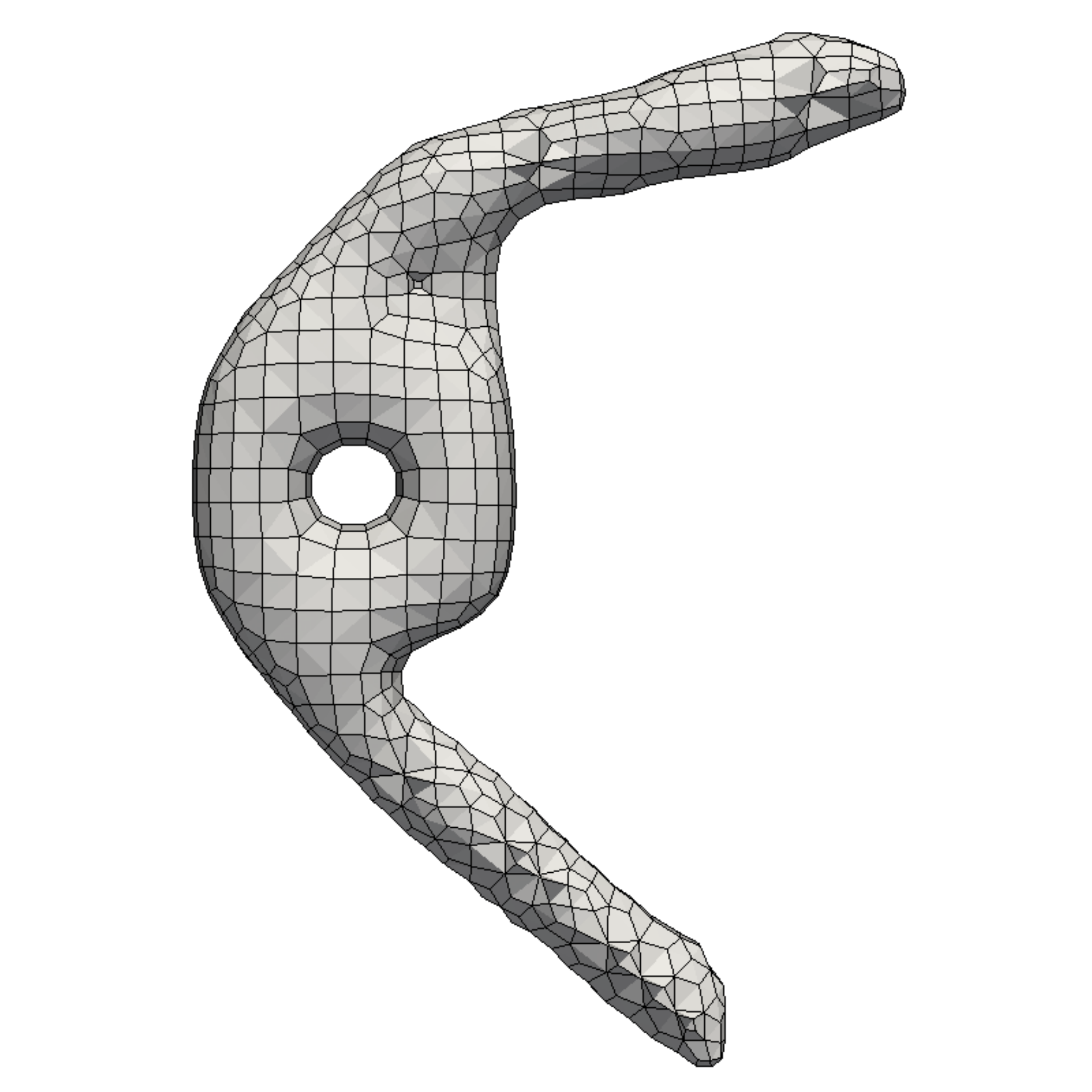}\\

\hline
\end{tabular}

}
        \subcaption{disc complex}
    \end{subtable}
\end{table}
\clearpage

\begin{table}[]
    \caption{Random samples from the sphere dataset. The first columns displays prediction from the UNet with trivial preprocessing and without equivariance. The second columns shows predictions from the UNet with trivial+PDE preprocessing and equivariance, which is our best model pipeline. All models have been trained on $150$ samples. In the third columns we show the ground truths densities corresponding to each sample.}
    \label{7_tab:random_samples_sphere}
    \begin{subtable}[t]{0.44\textwidth}
        \centering
        \setlength\tabcolsep{6pt}
        \resizebox{0.85\textwidth}{!}{\begin{tabular}{|c|c||c|}
\hline
\parbox{4em}{\centering UNet} & \parbox{4em}{\centering UNet\\+physics} & \xrowht{20pt}\parbox{4em}{\centering ground\\truth} \\
\hline\rule{0pt}{2cm}

\xrowht{20pt} 
\includegraphics[width=0.25\textwidth]{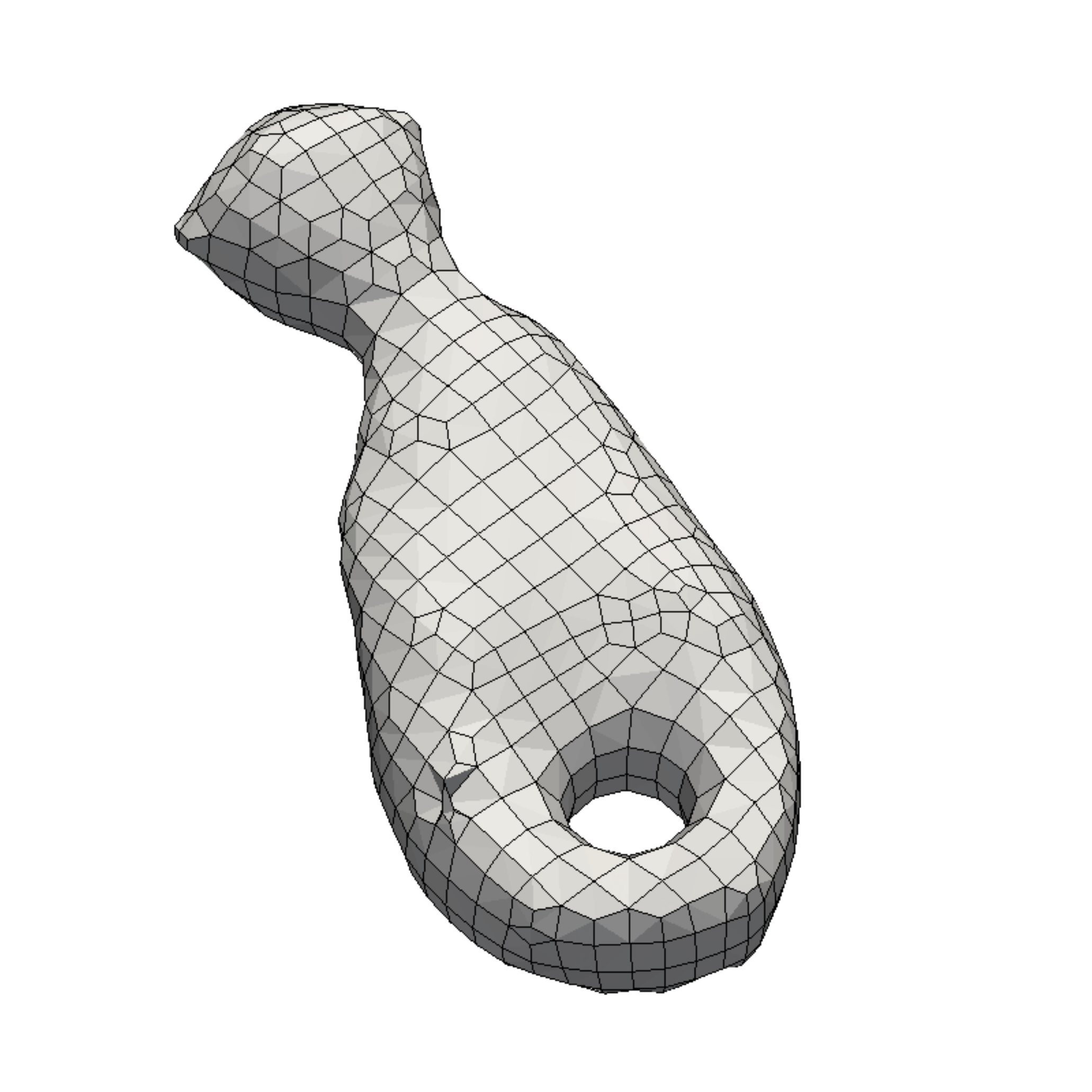} & \includegraphics[width=0.25\textwidth]{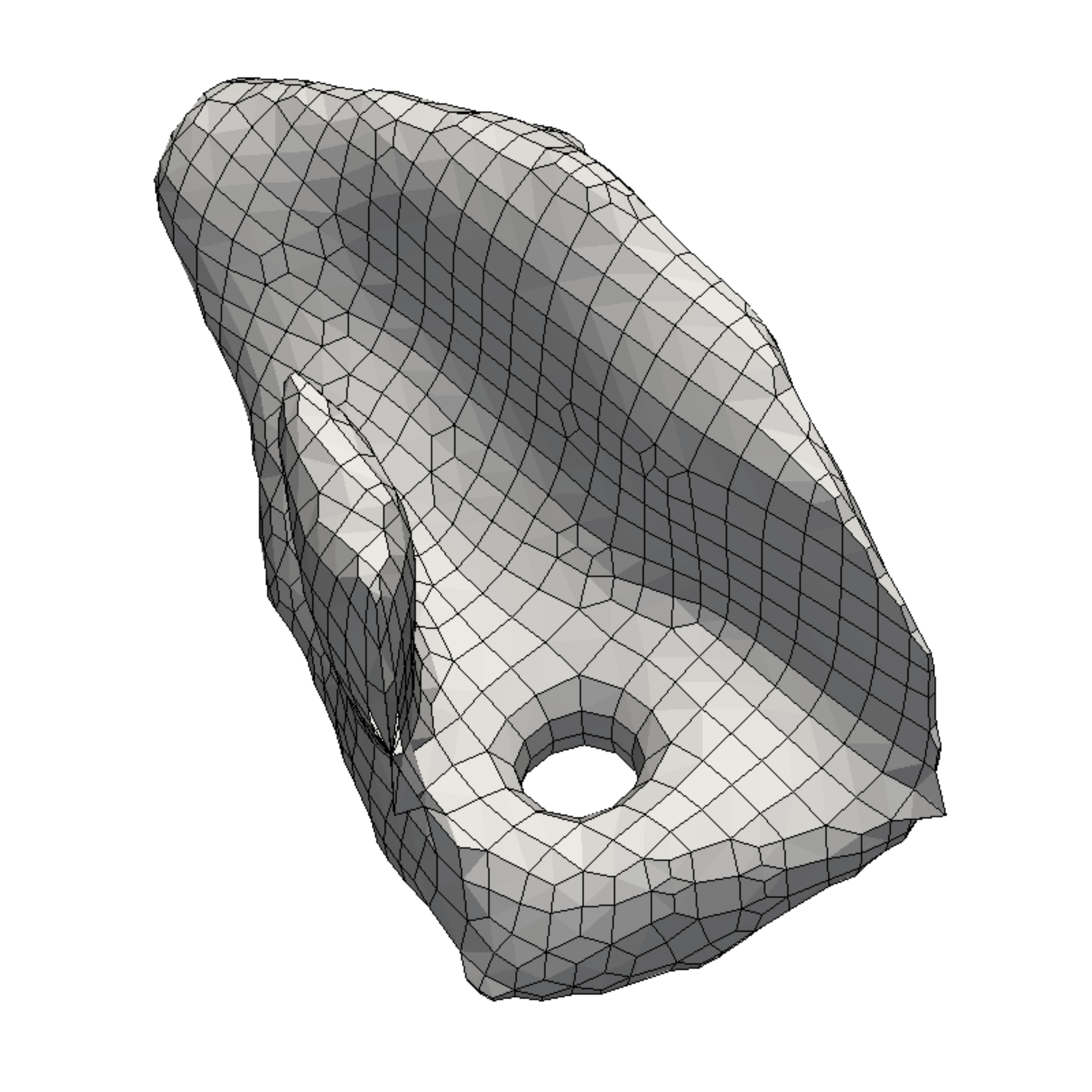} & \includegraphics[width=0.25\textwidth]{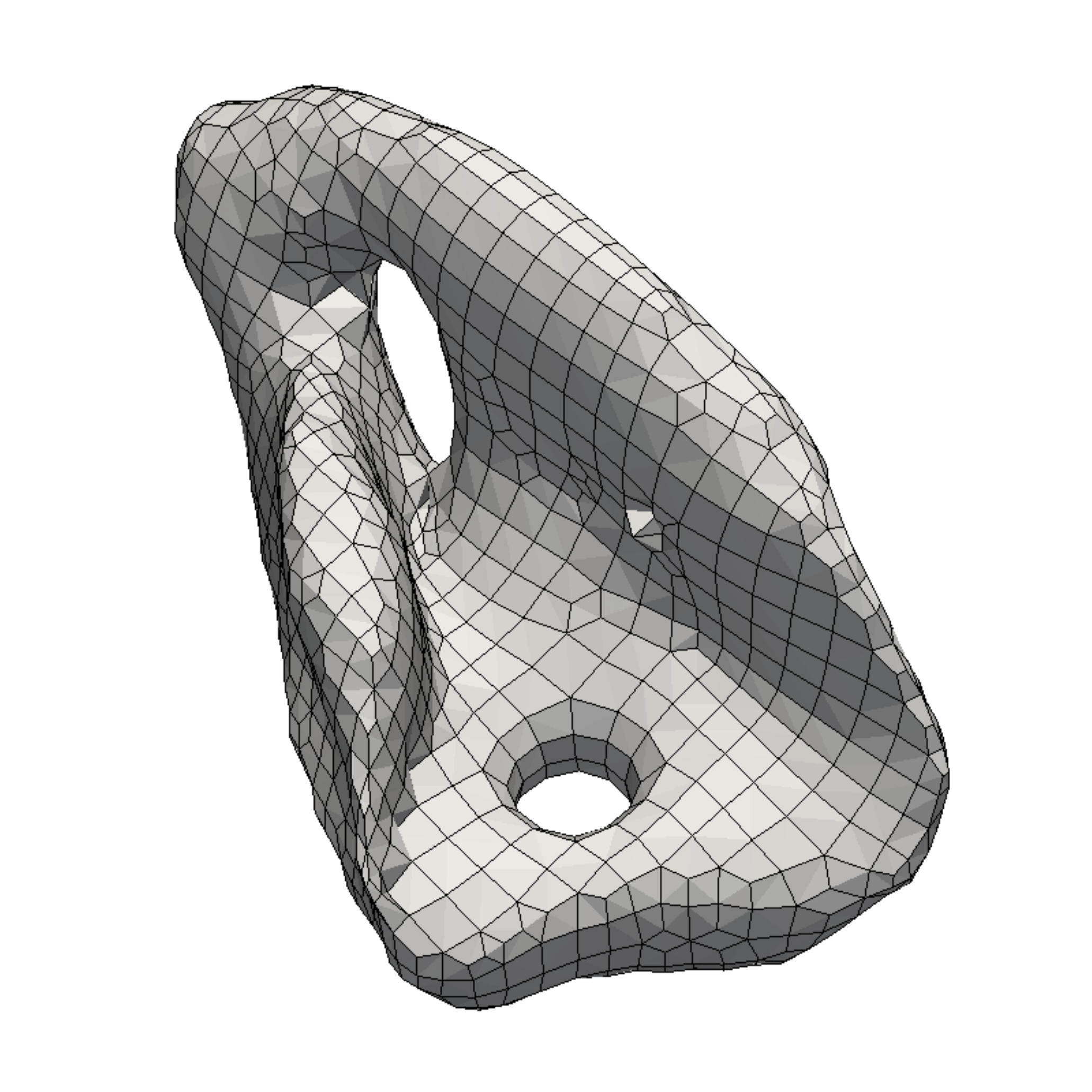}\\

\xrowht{20pt} 
\includegraphics[width=0.25\textwidth]{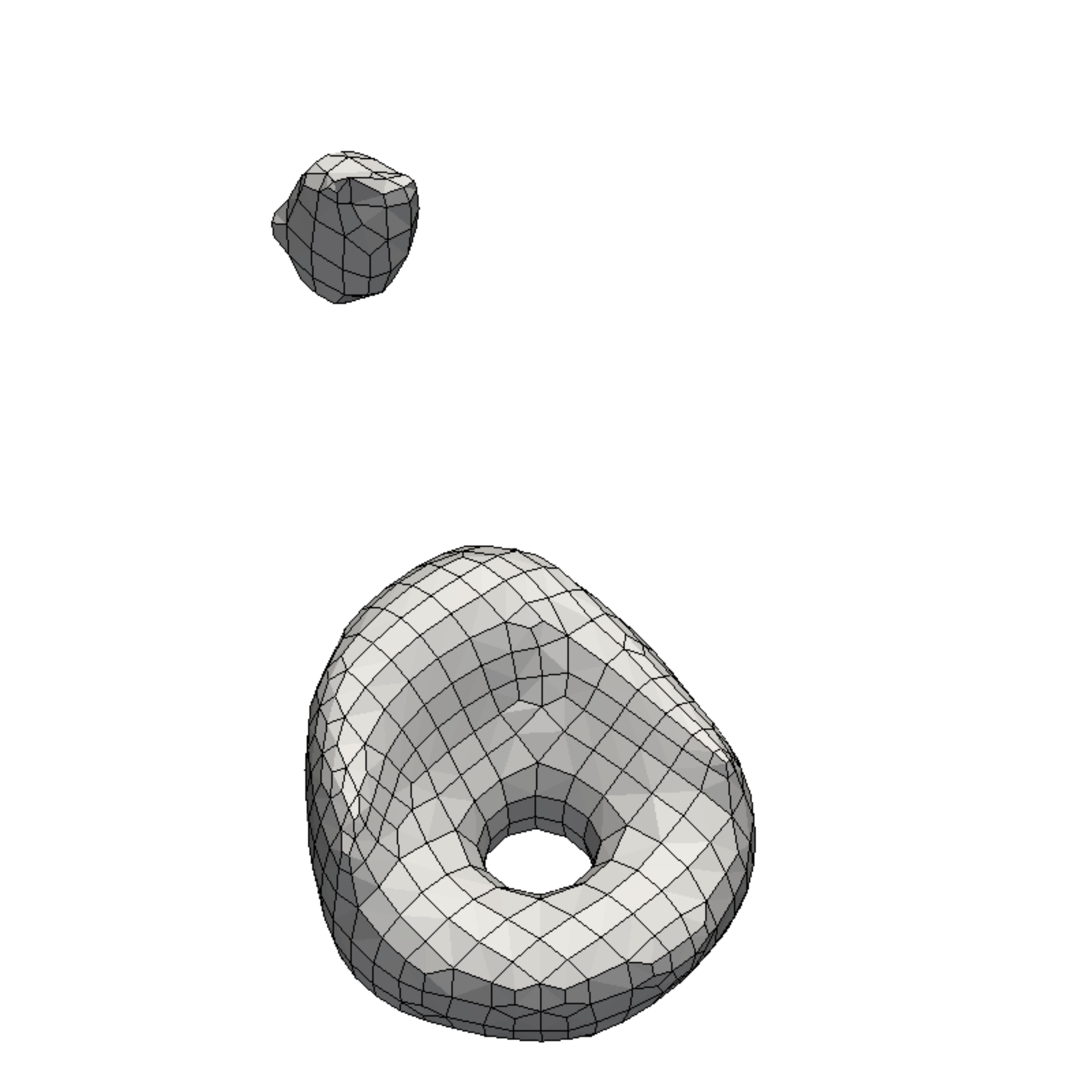} & \includegraphics[width=0.25\textwidth]{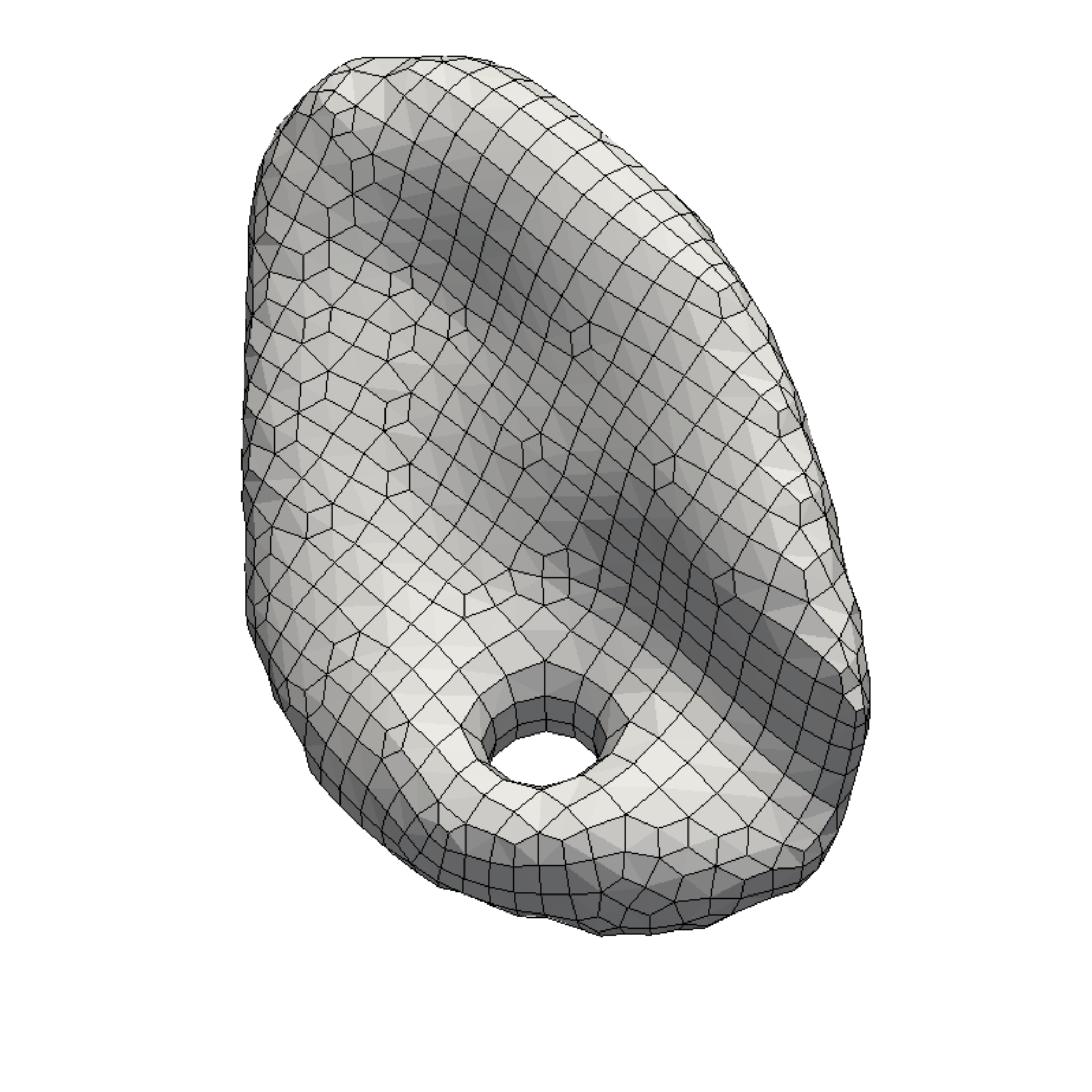} & \includegraphics[width=0.25\textwidth]{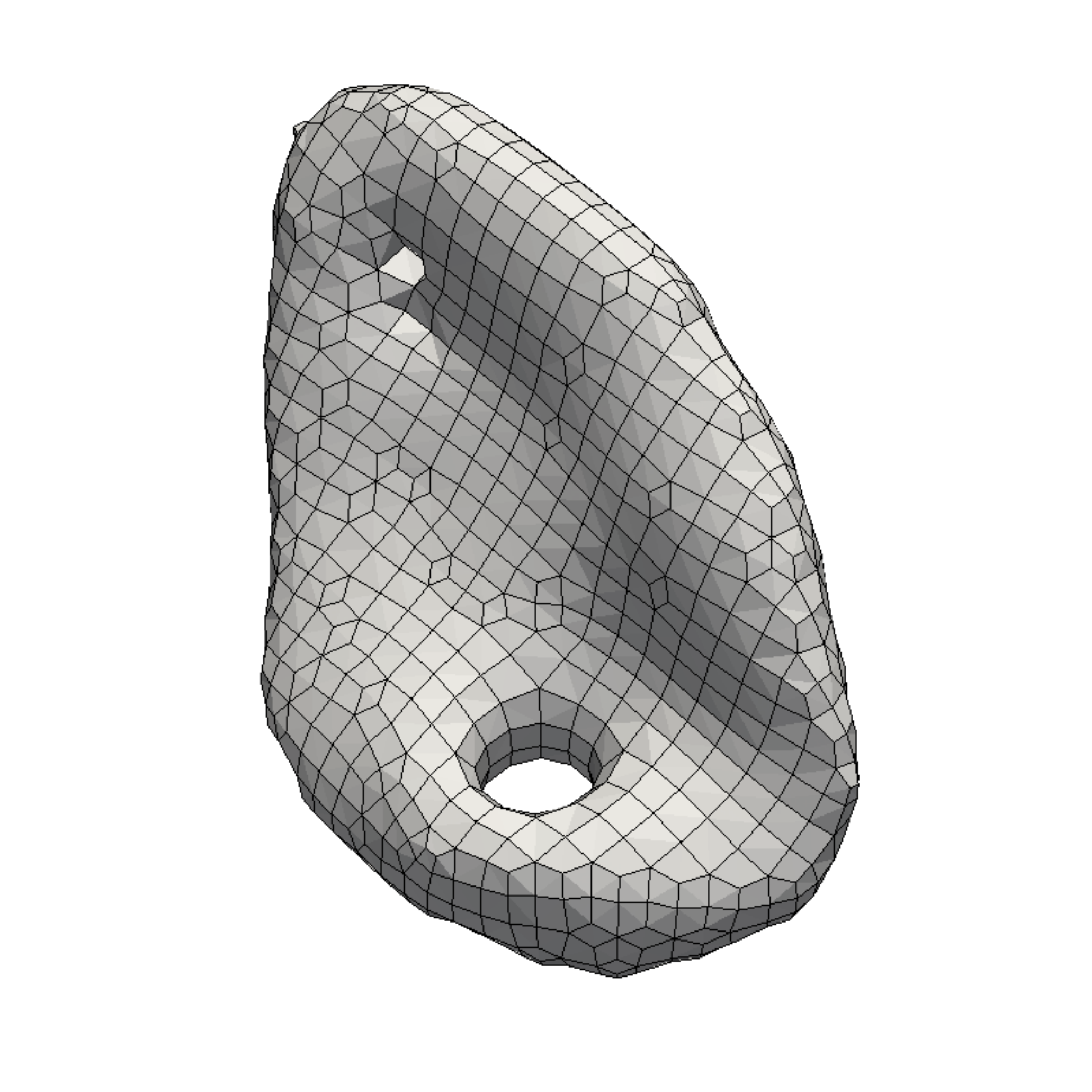}\\

\xrowht{20pt} 
\includegraphics[width=0.25\textwidth]{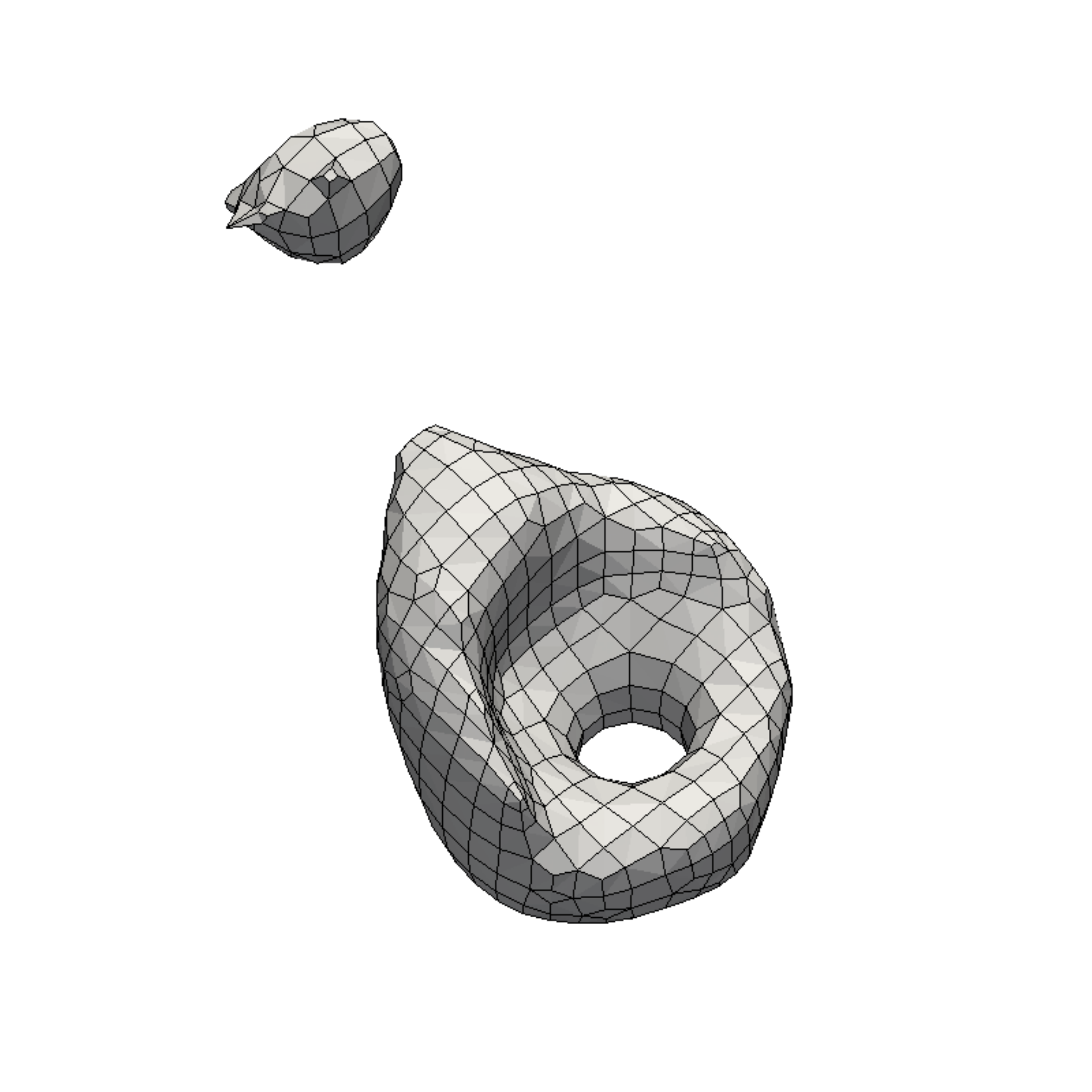} & \includegraphics[width=0.25\textwidth]{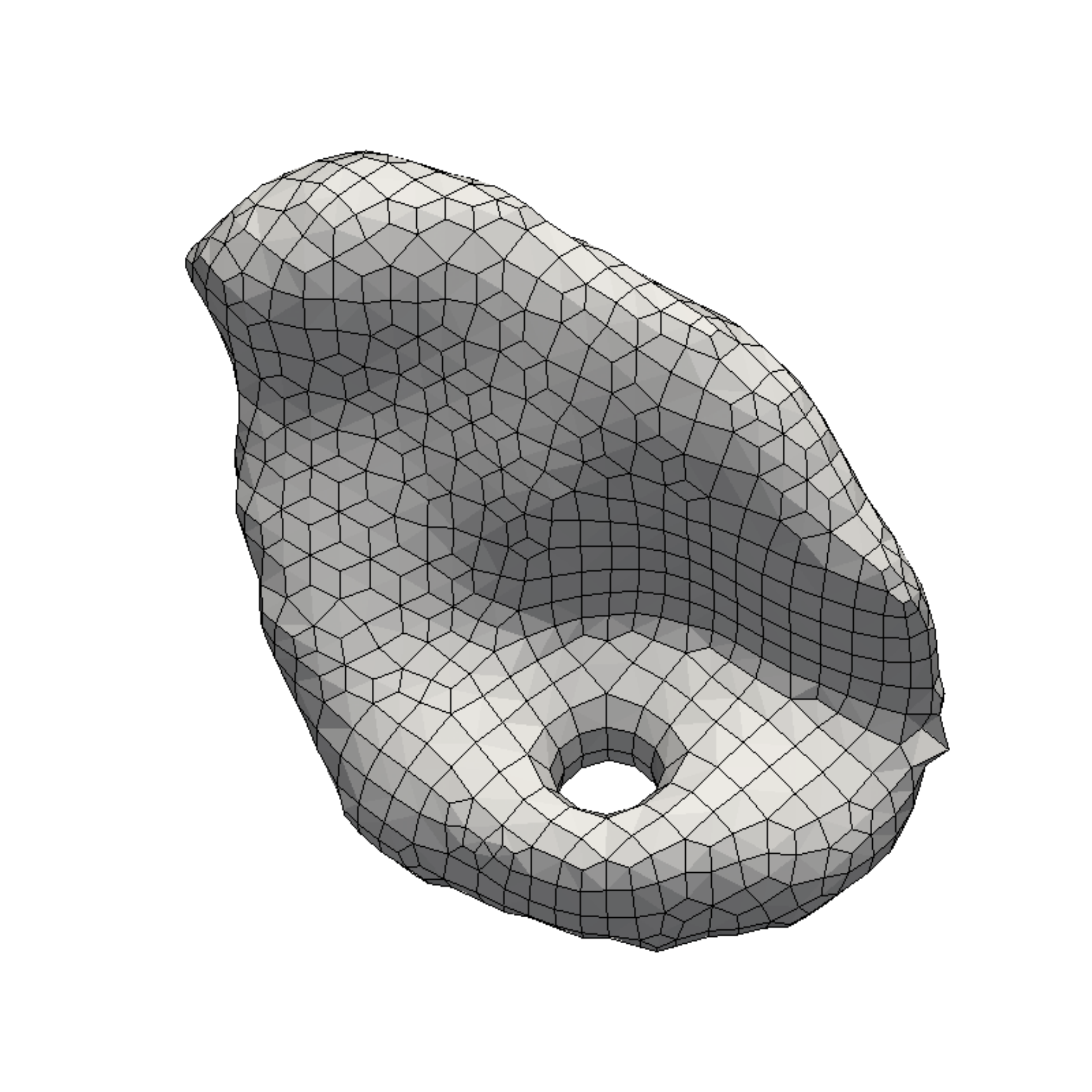} & \includegraphics[width=0.25\textwidth]{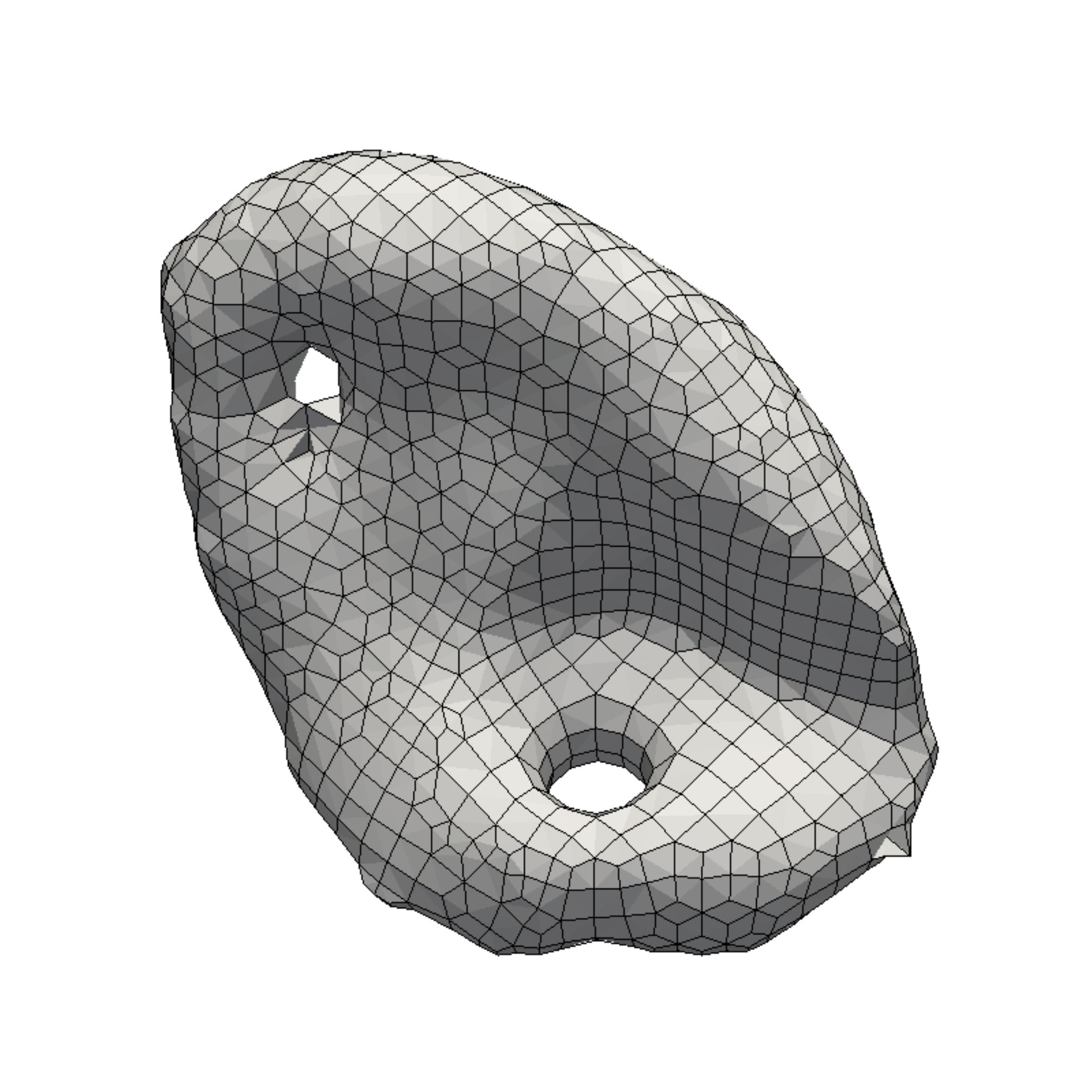}\\

\xrowht{20pt} 
\includegraphics[width=0.25\textwidth]{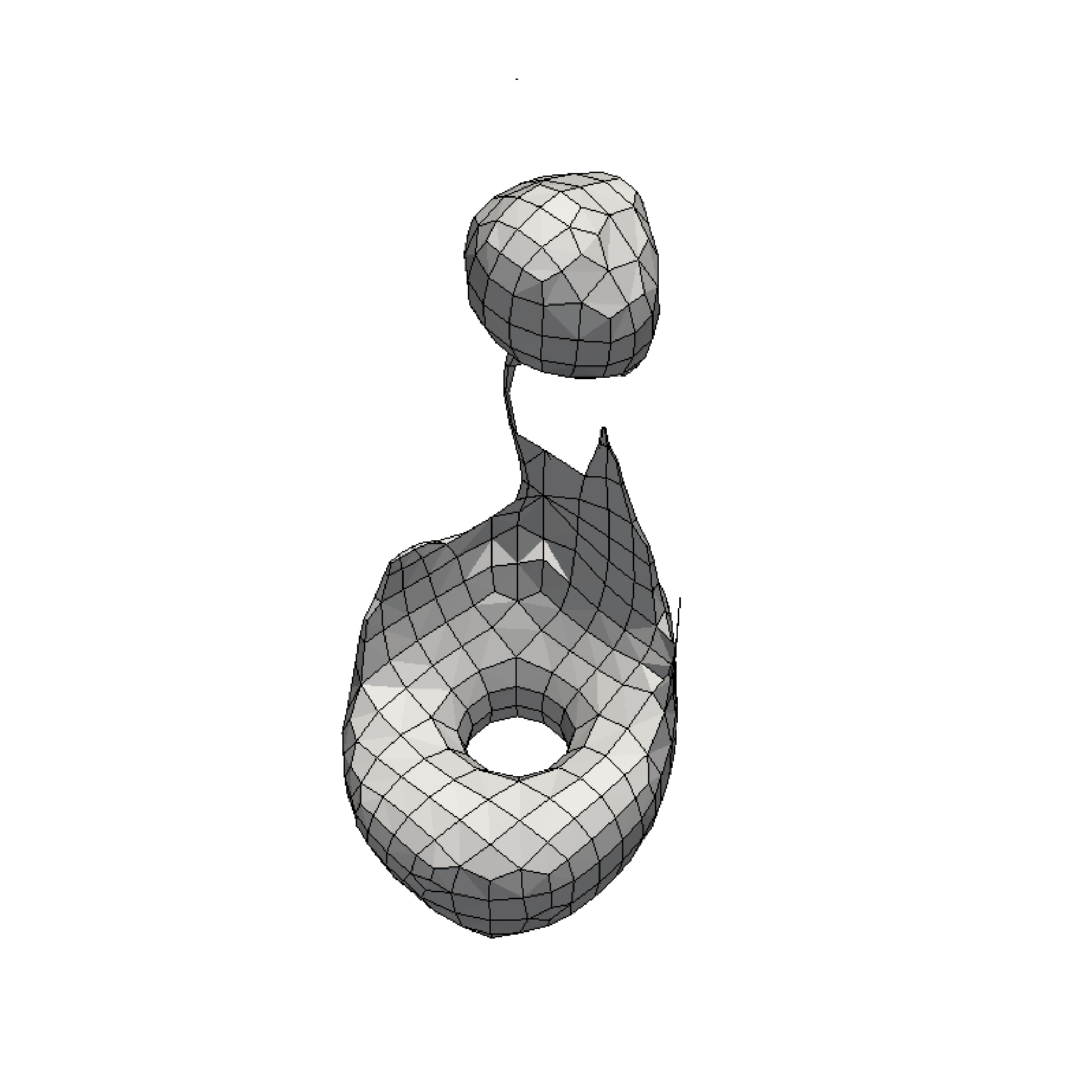} & \includegraphics[width=0.25\textwidth]{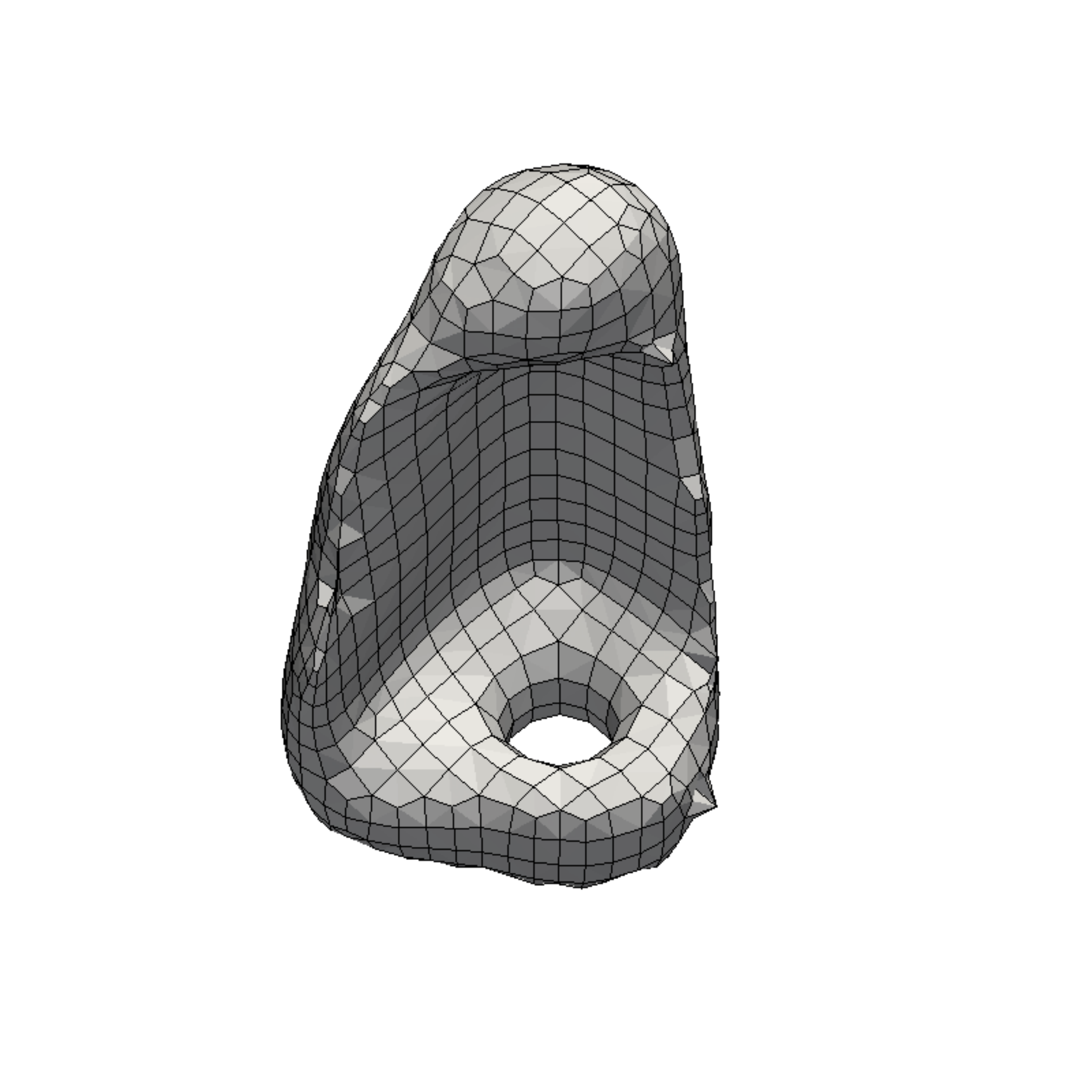} & \includegraphics[width=0.25\textwidth]{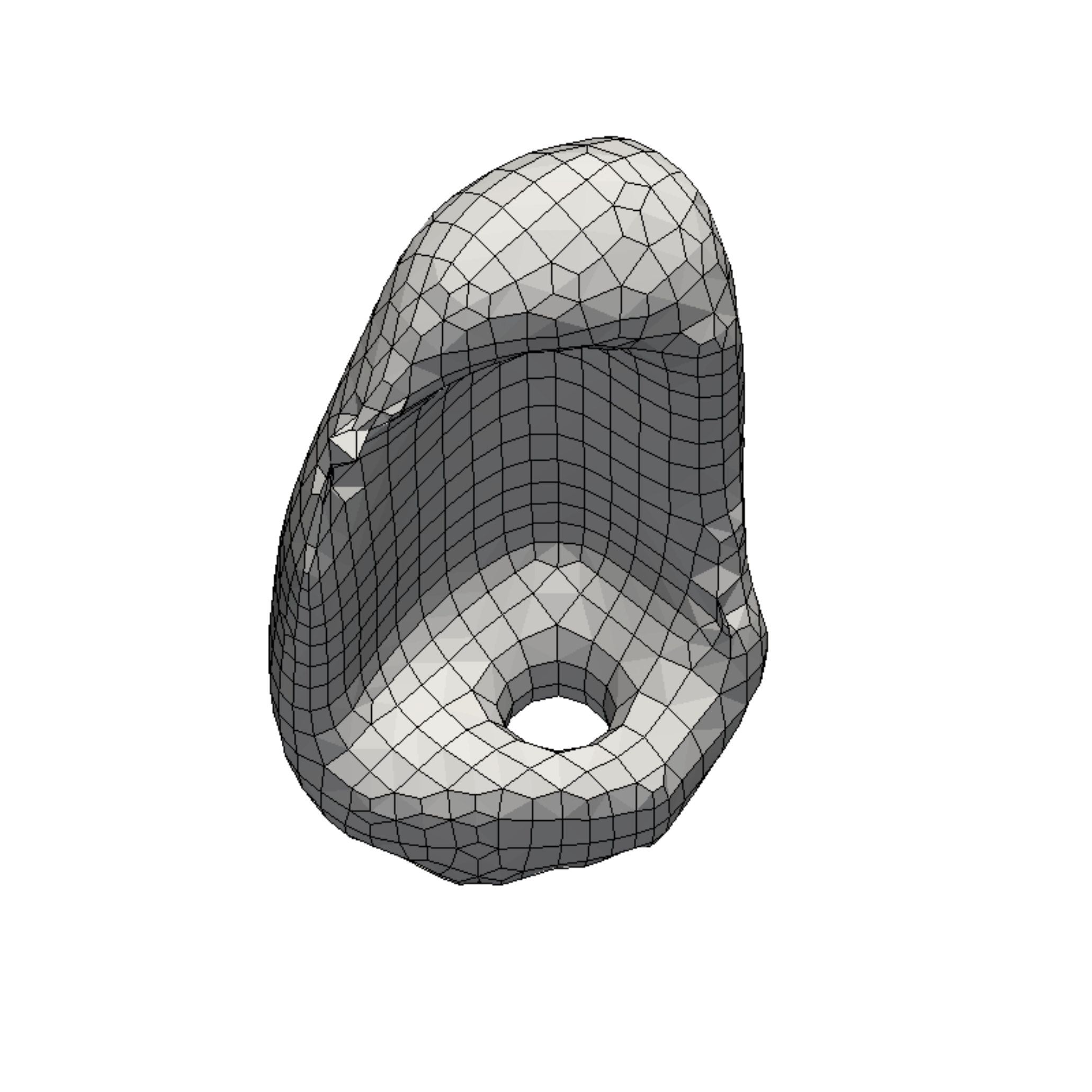}\\

\xrowht{20pt} 
\includegraphics[width=0.25\textwidth]{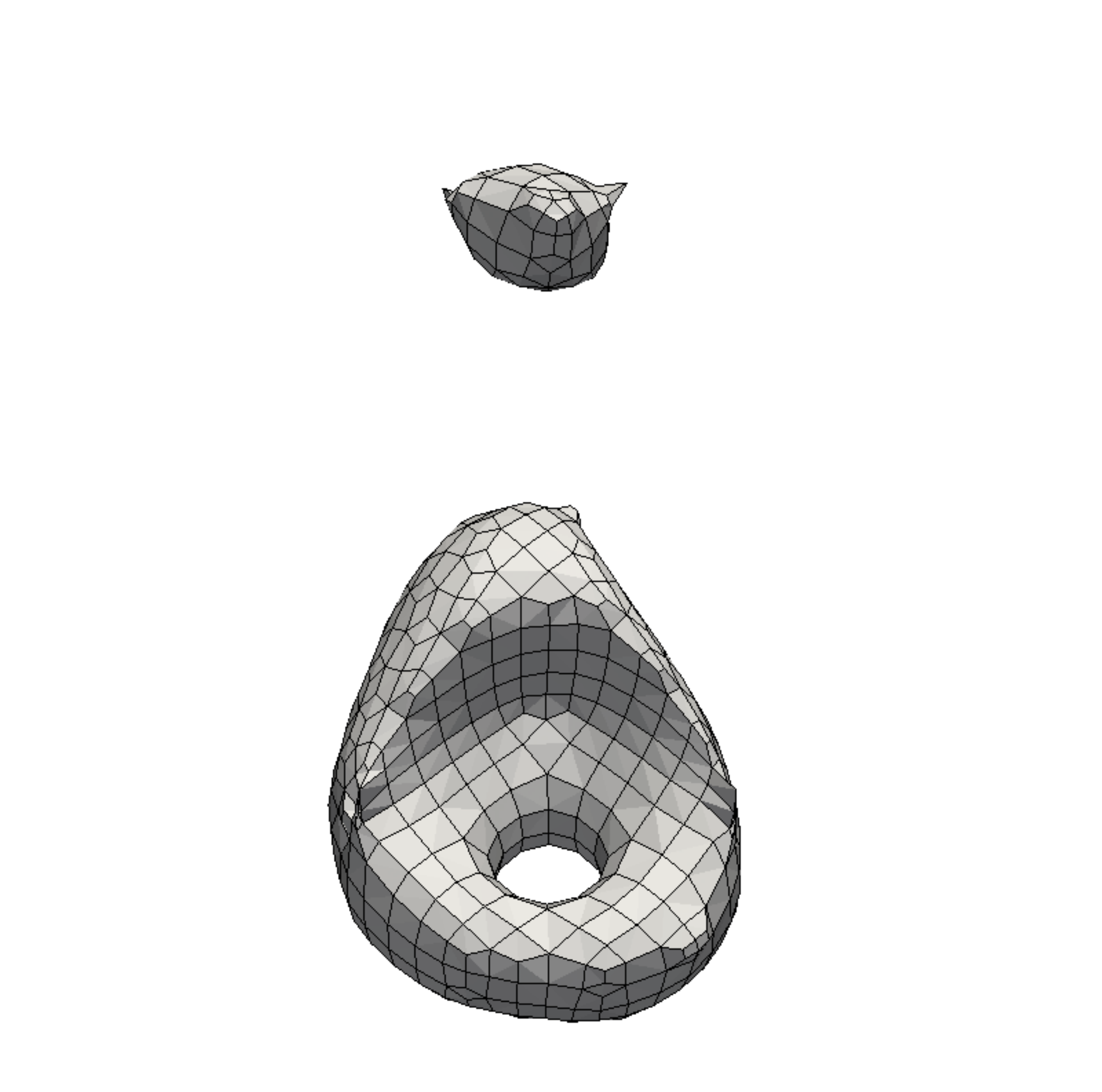} & \includegraphics[width=0.25\textwidth]{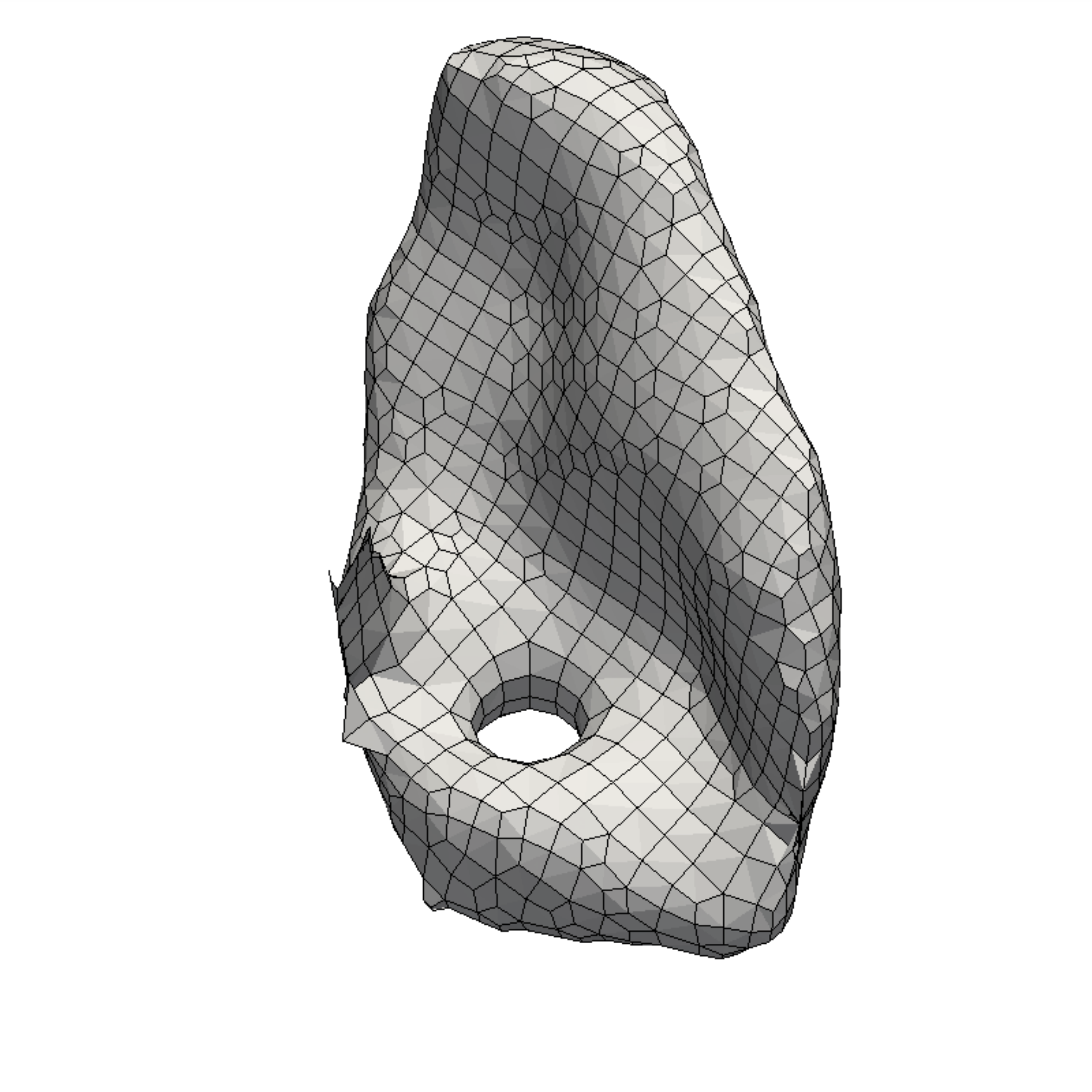} & \includegraphics[width=0.25\textwidth]{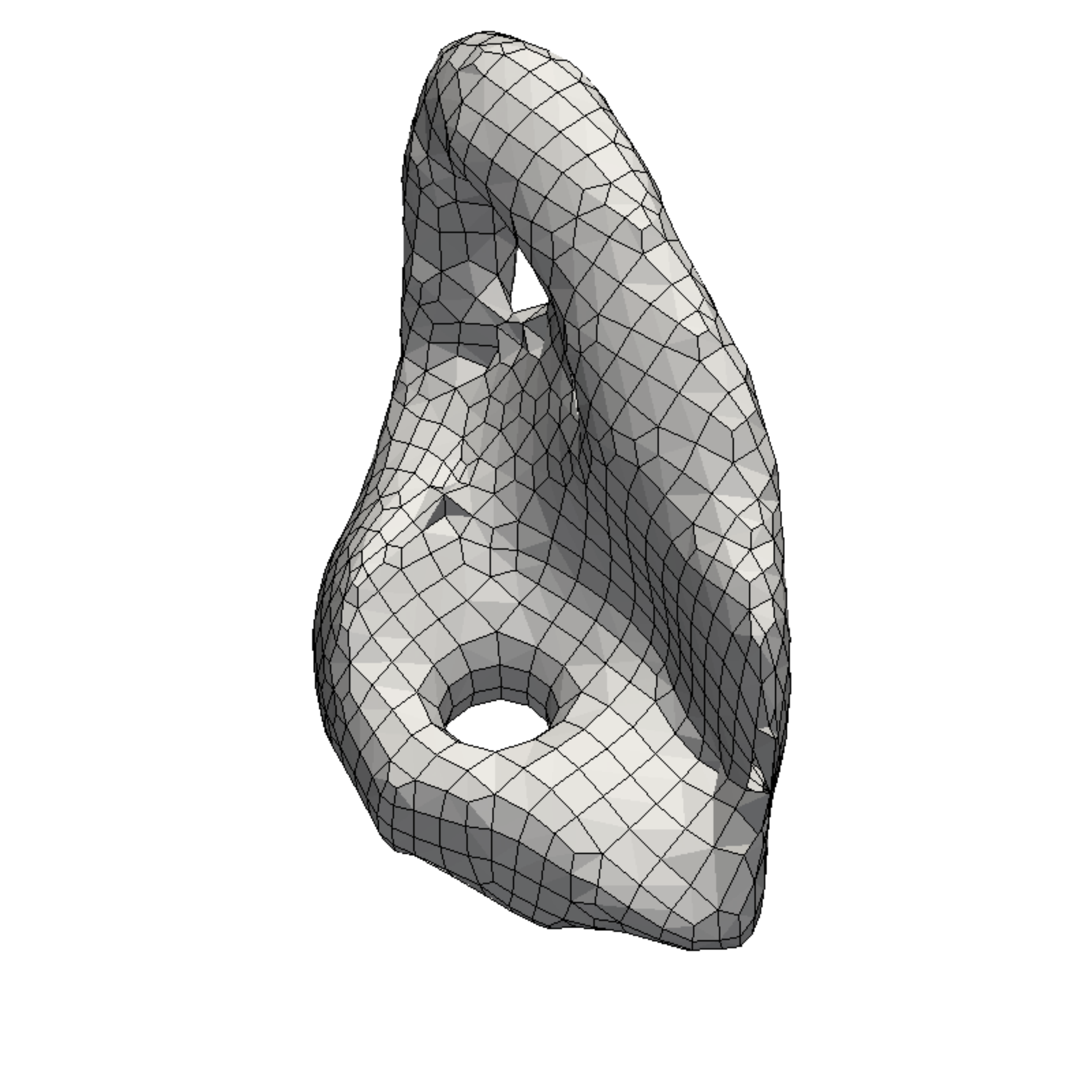}\\

\xrowht{20pt} 
\includegraphics[width=0.25\textwidth]{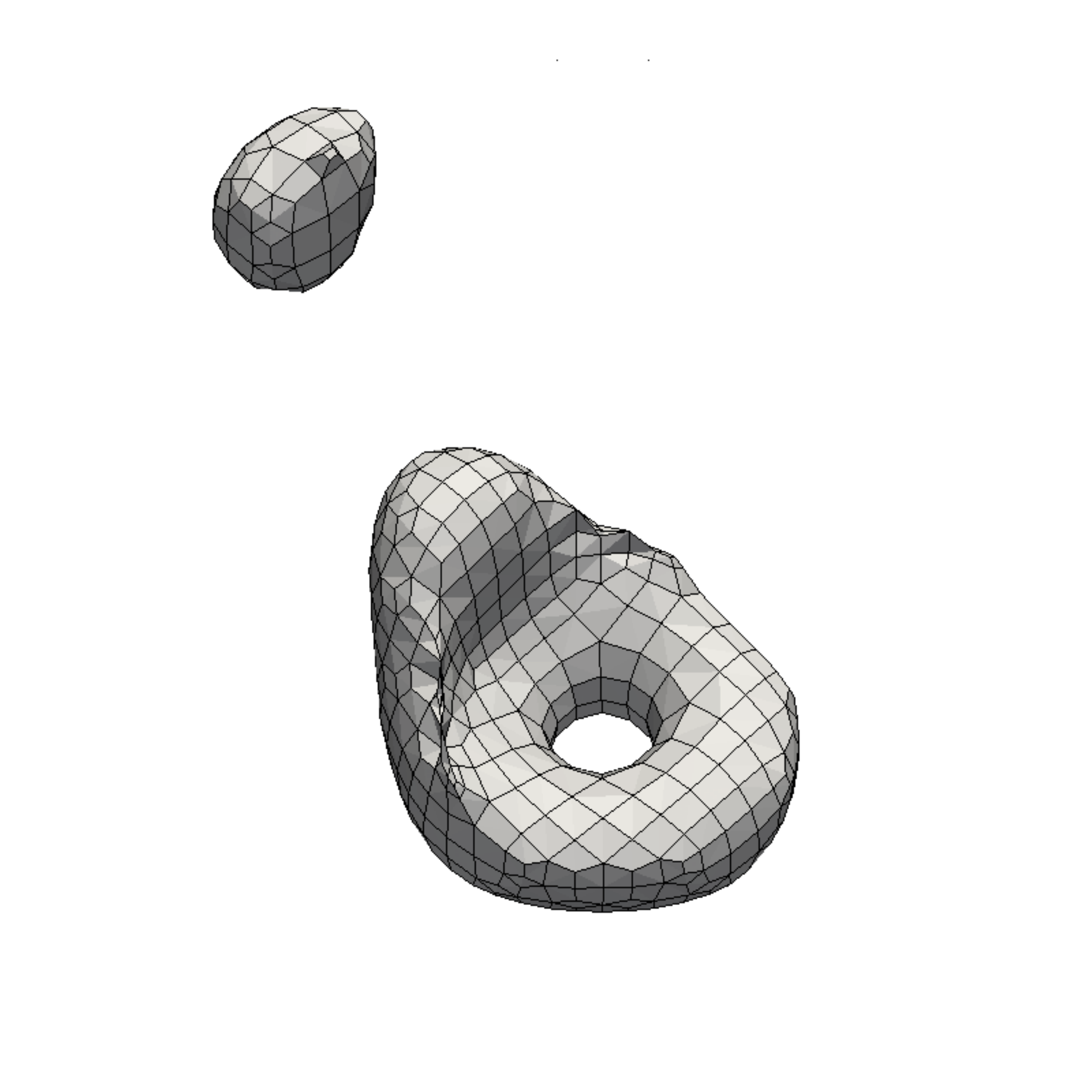} & \includegraphics[width=0.25\textwidth]{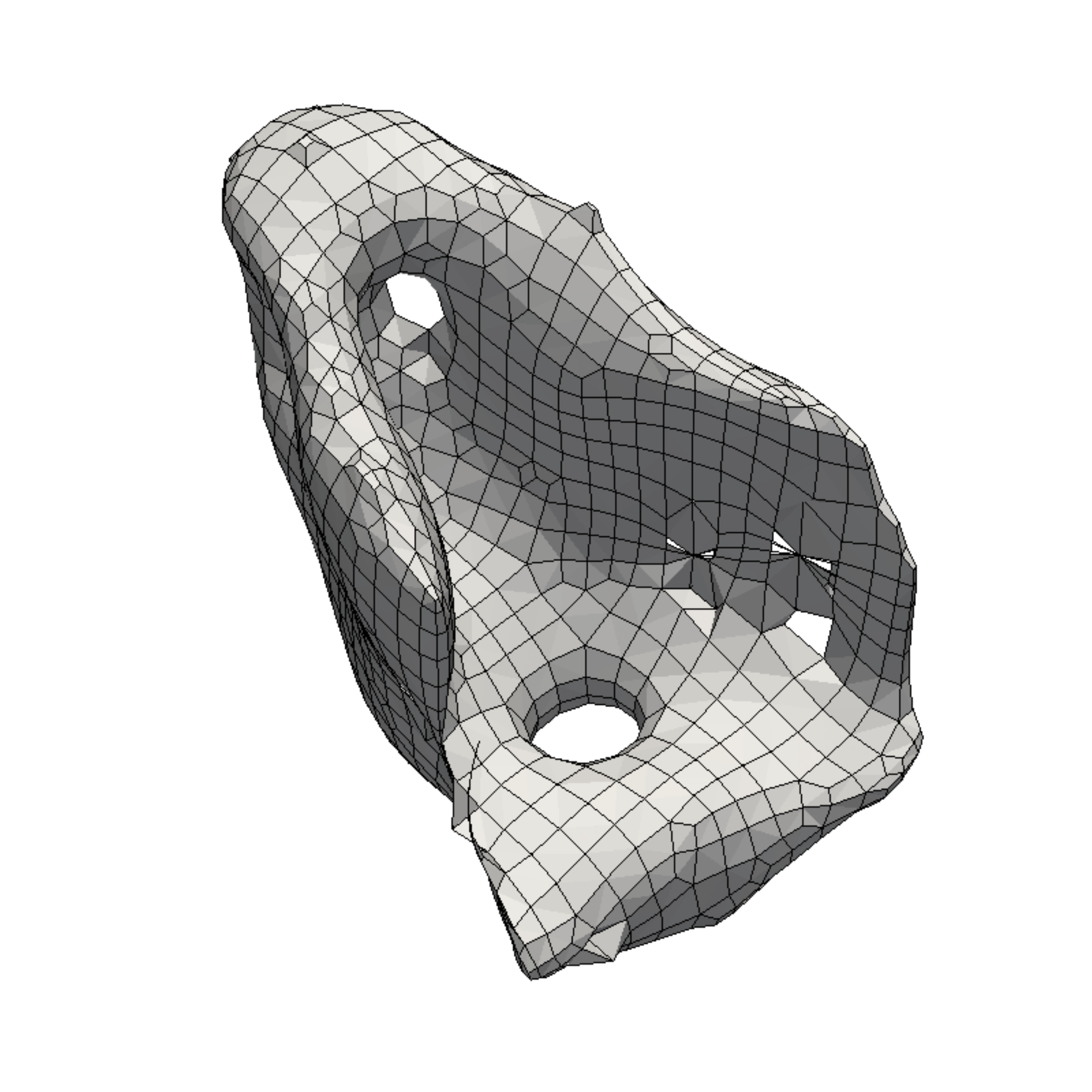} & \includegraphics[width=0.25\textwidth]{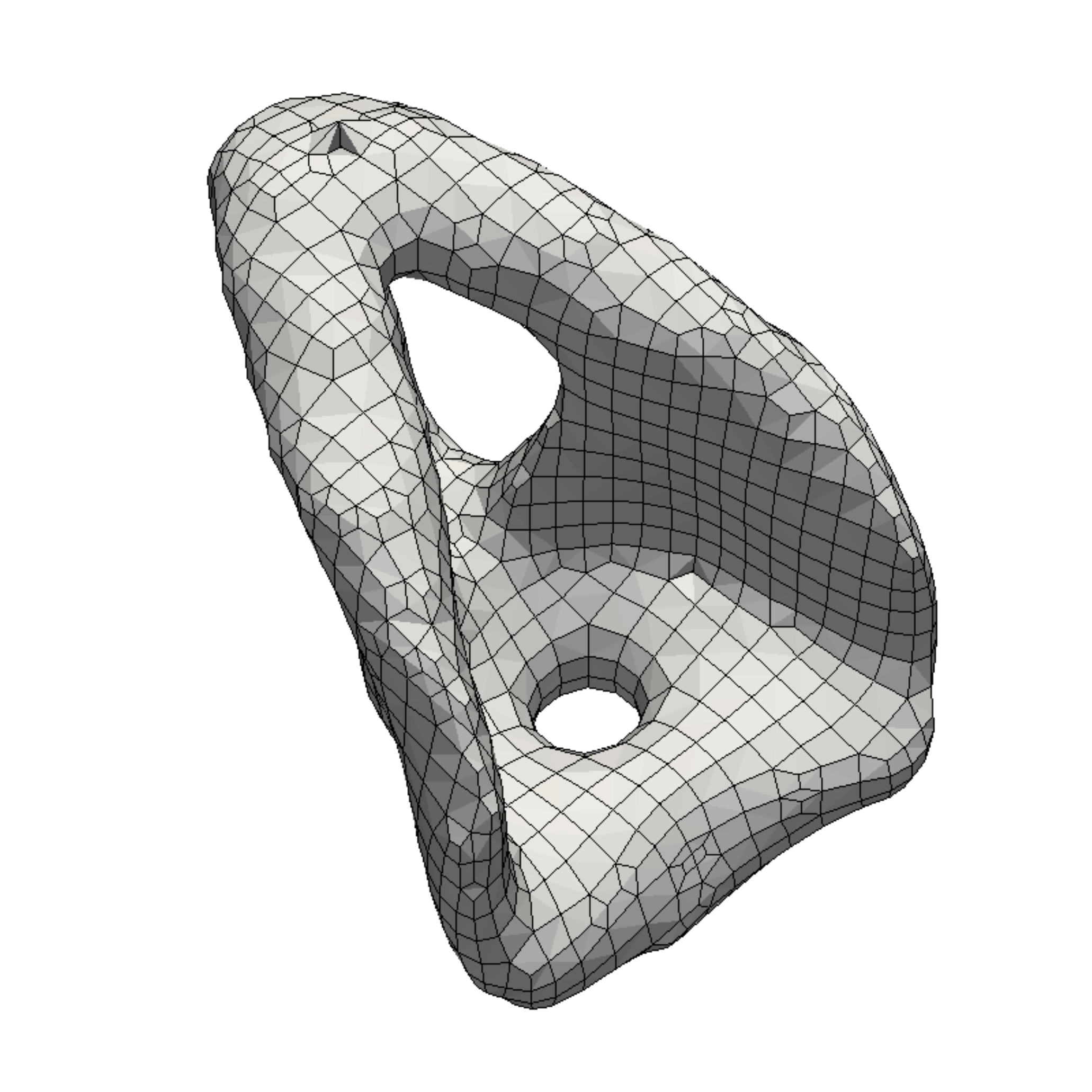}\\

\xrowht{20pt} 
\includegraphics[width=0.25\textwidth]{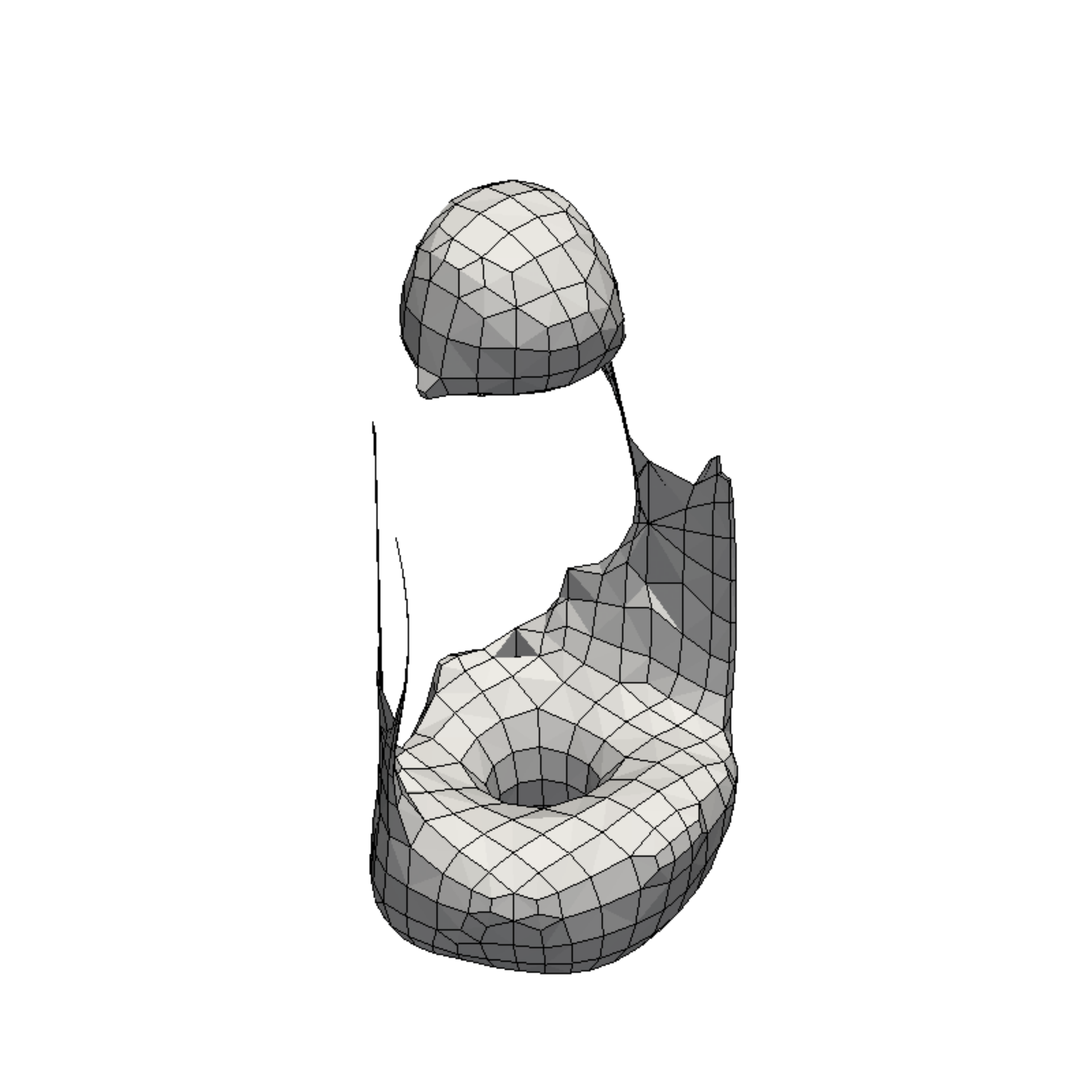} & \includegraphics[width=0.25\textwidth]{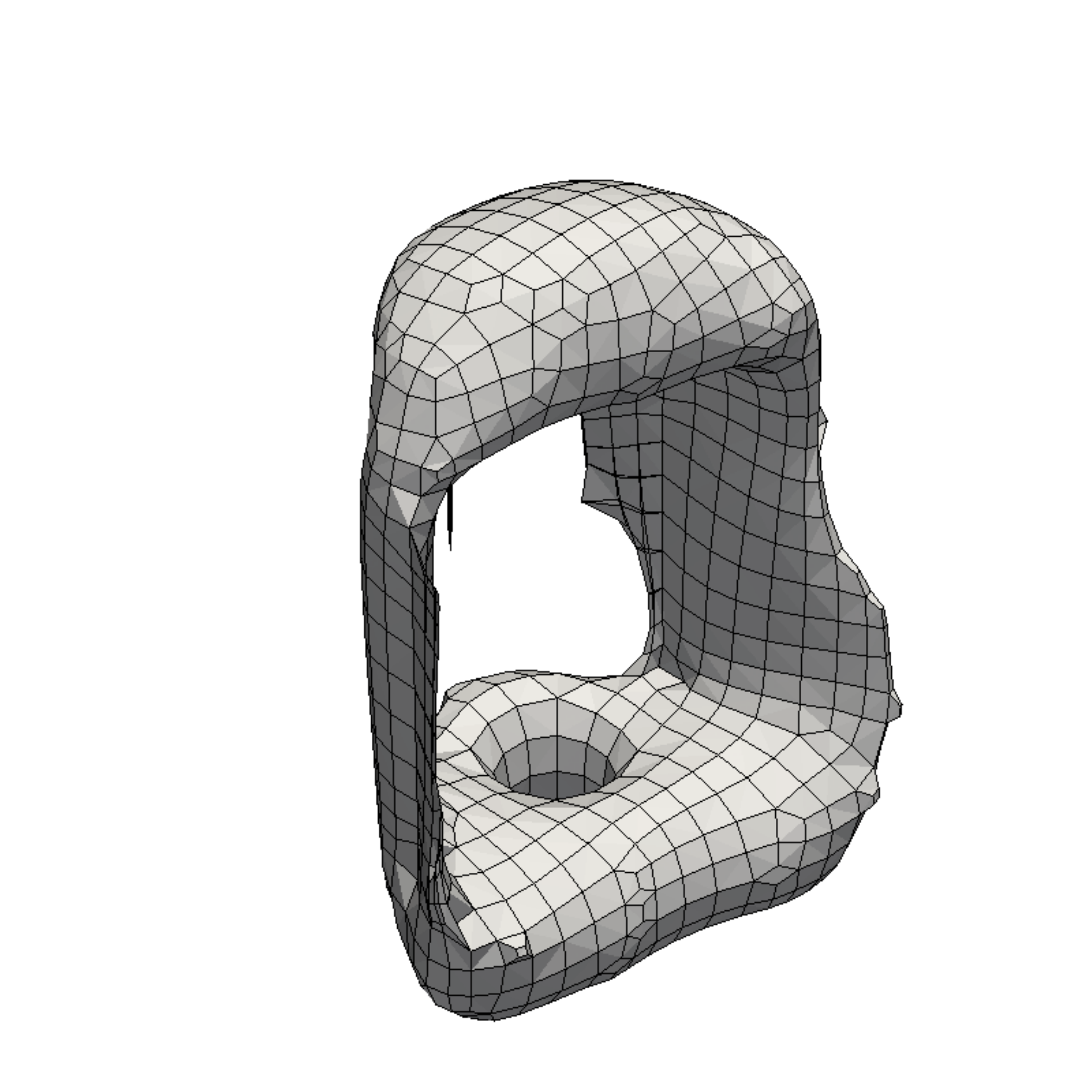} & \includegraphics[width=0.25\textwidth]{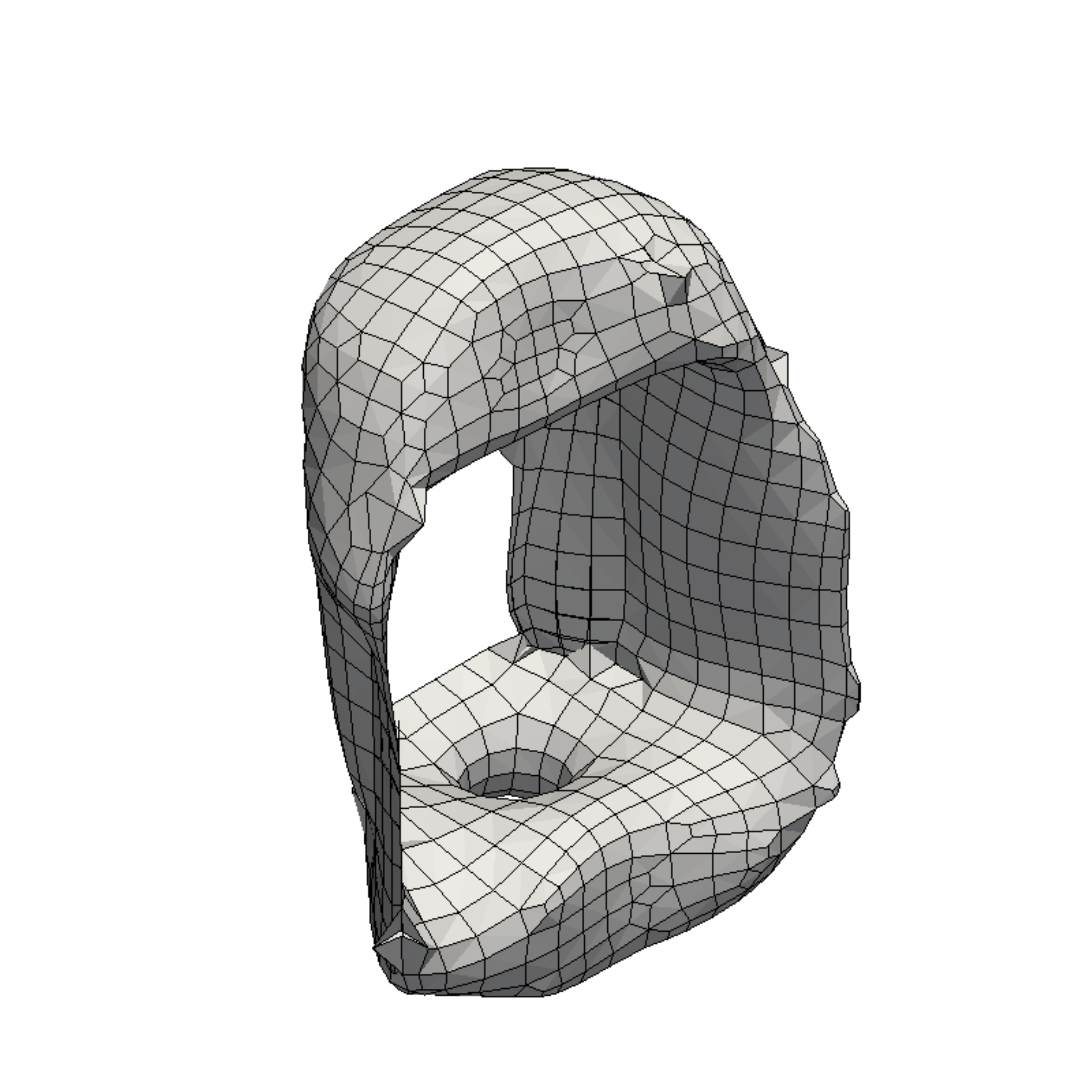}\\

\xrowht{20pt} 
\includegraphics[width=0.25\textwidth]{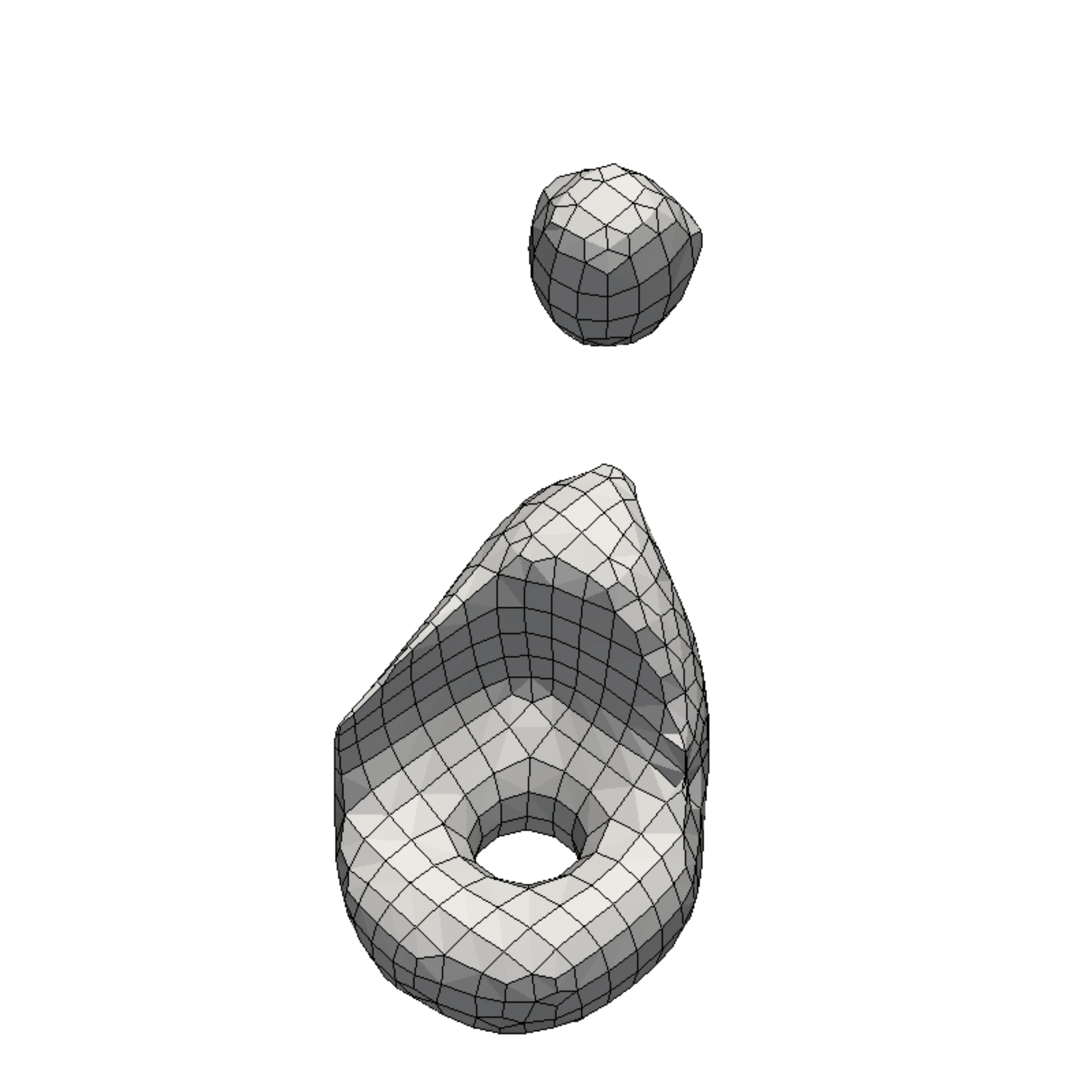} & \includegraphics[width=0.25\textwidth]{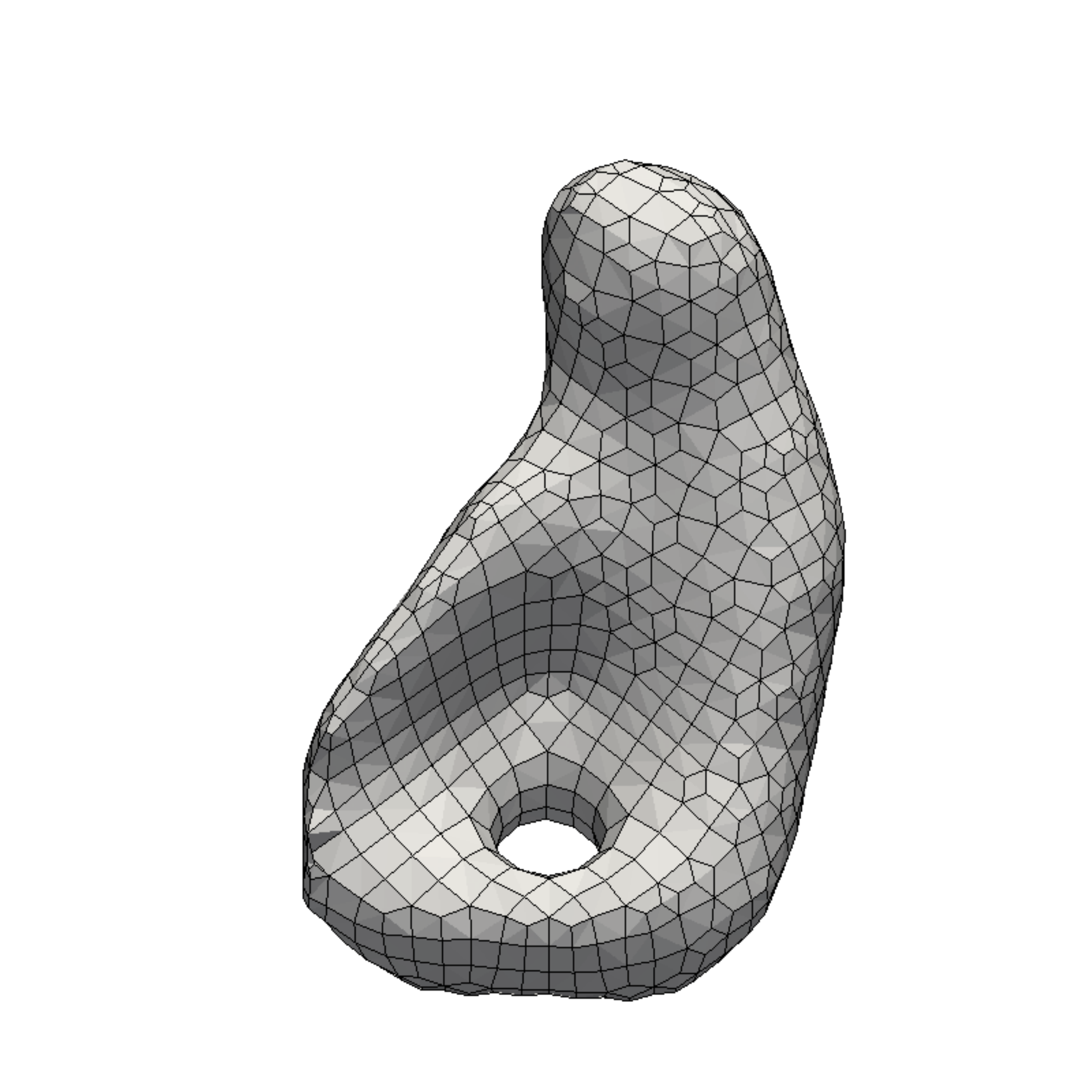} & \includegraphics[width=0.25\textwidth]{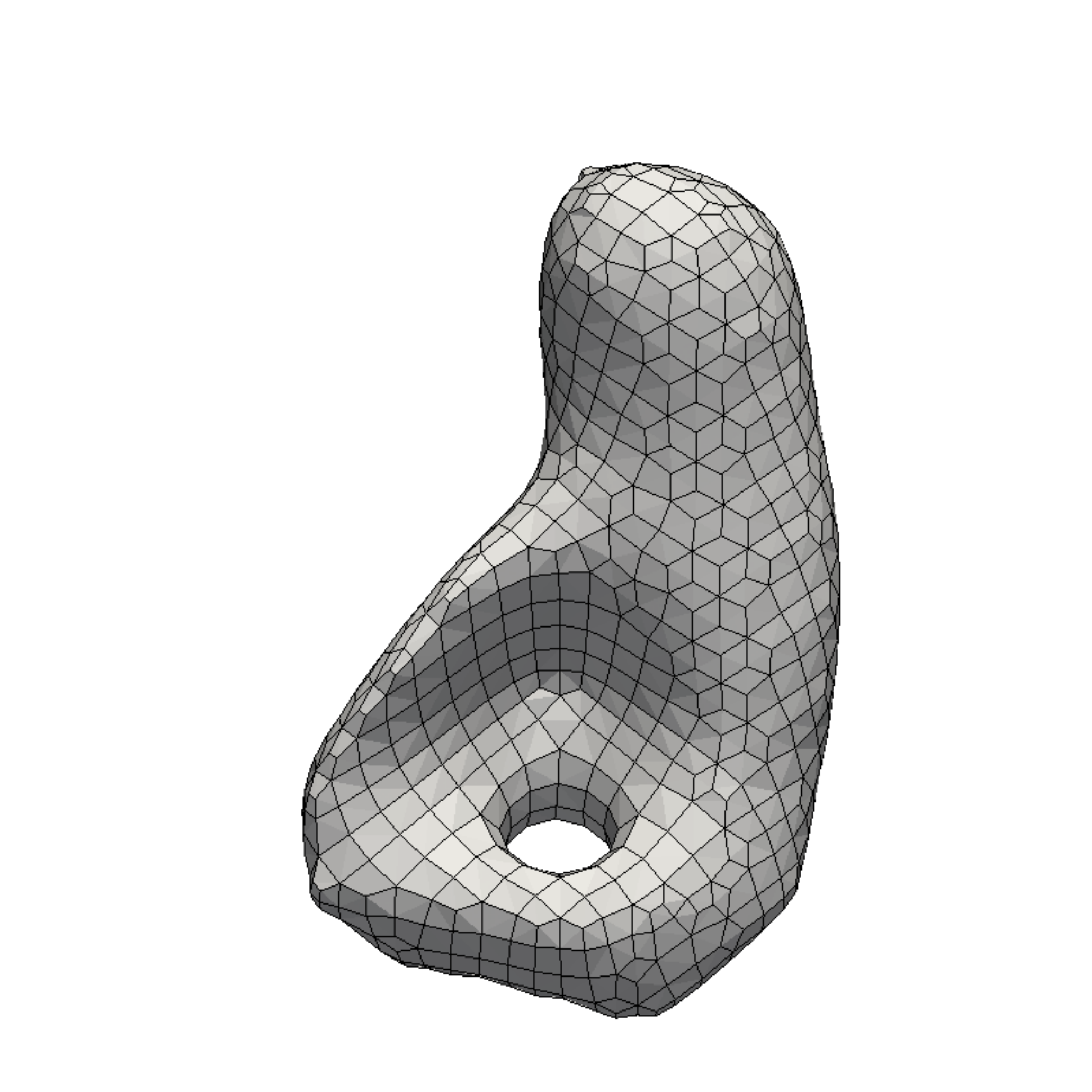}\\

\xrowht{20pt} 
\includegraphics[width=0.25\textwidth]{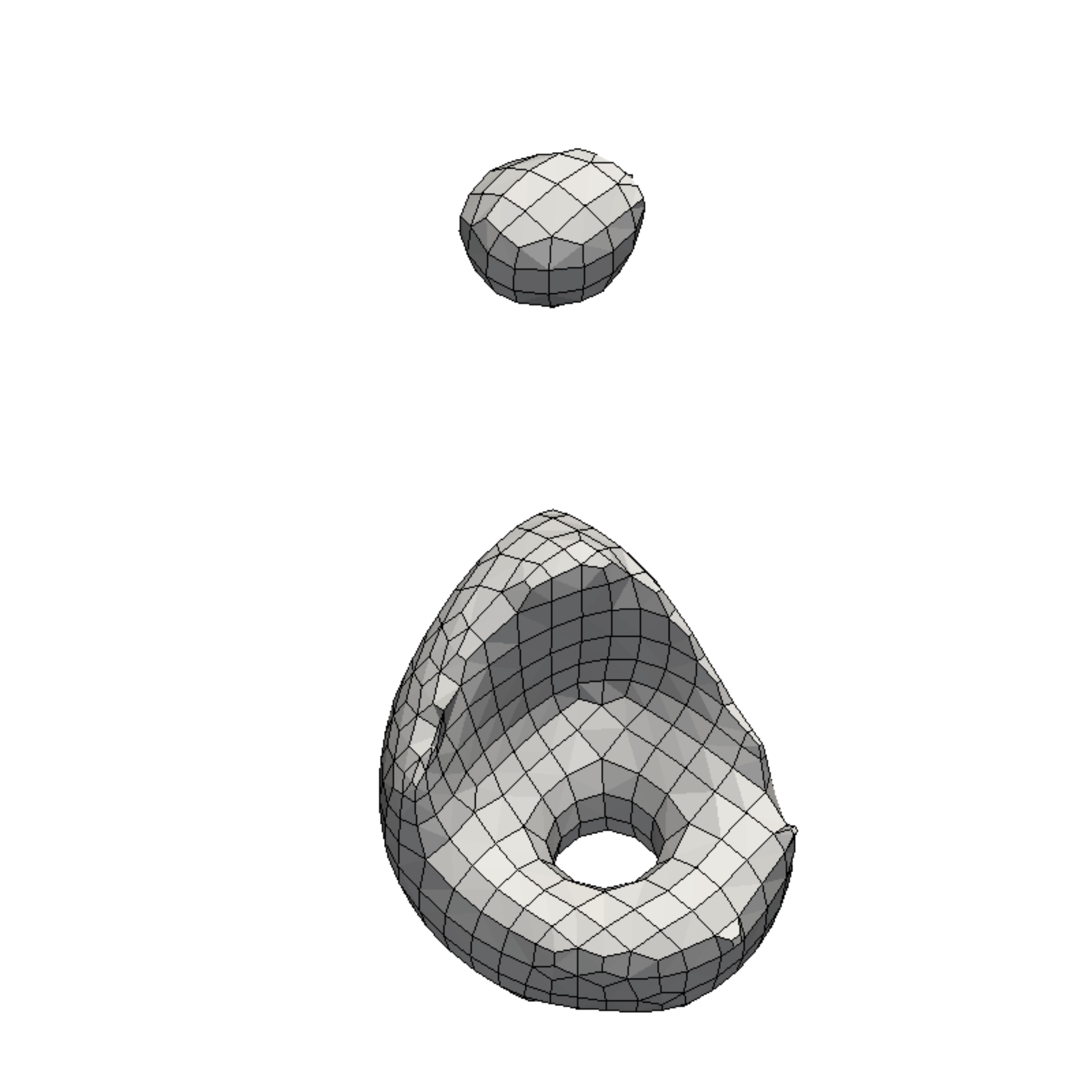} & \includegraphics[width=0.25\textwidth]{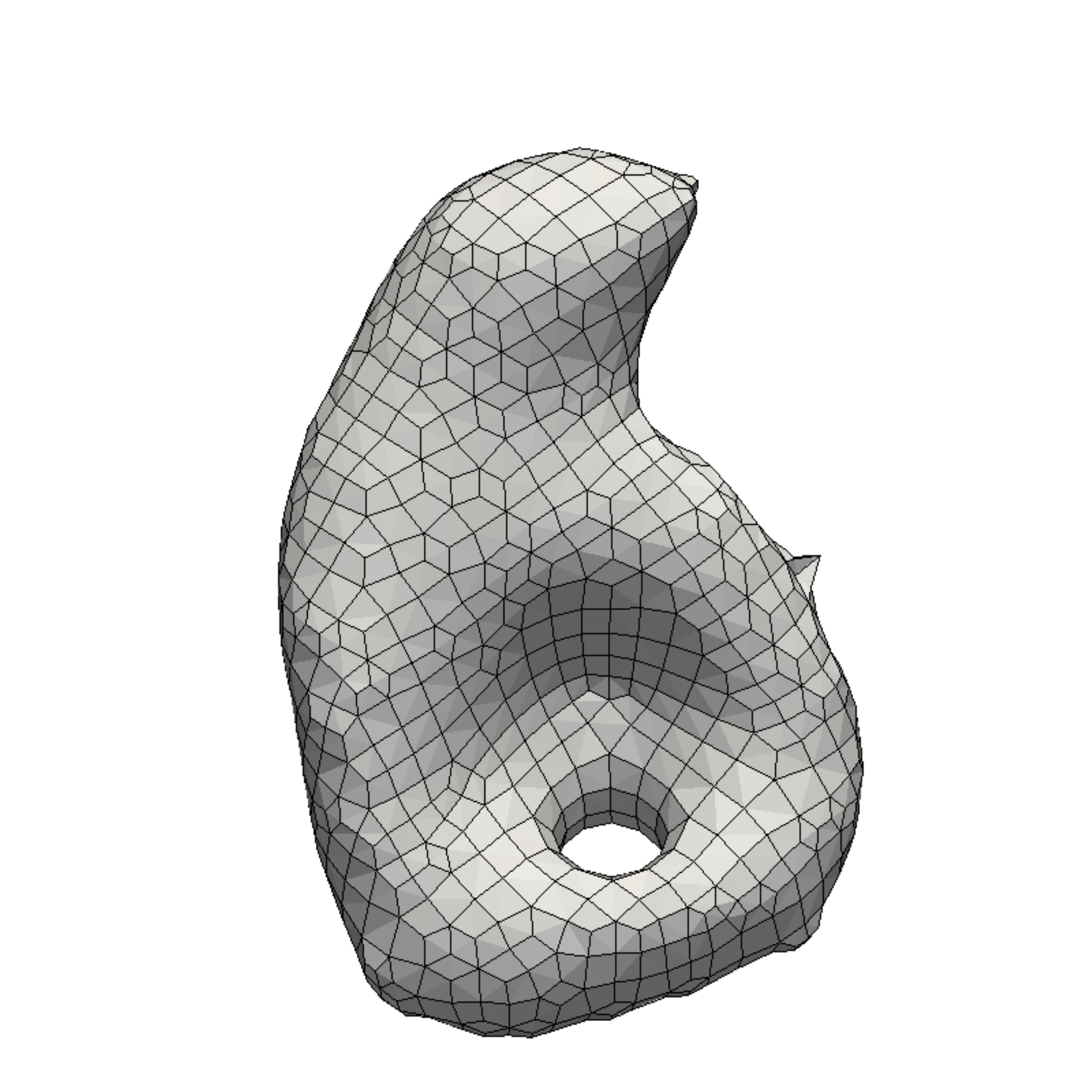} & \includegraphics[width=0.25\textwidth]{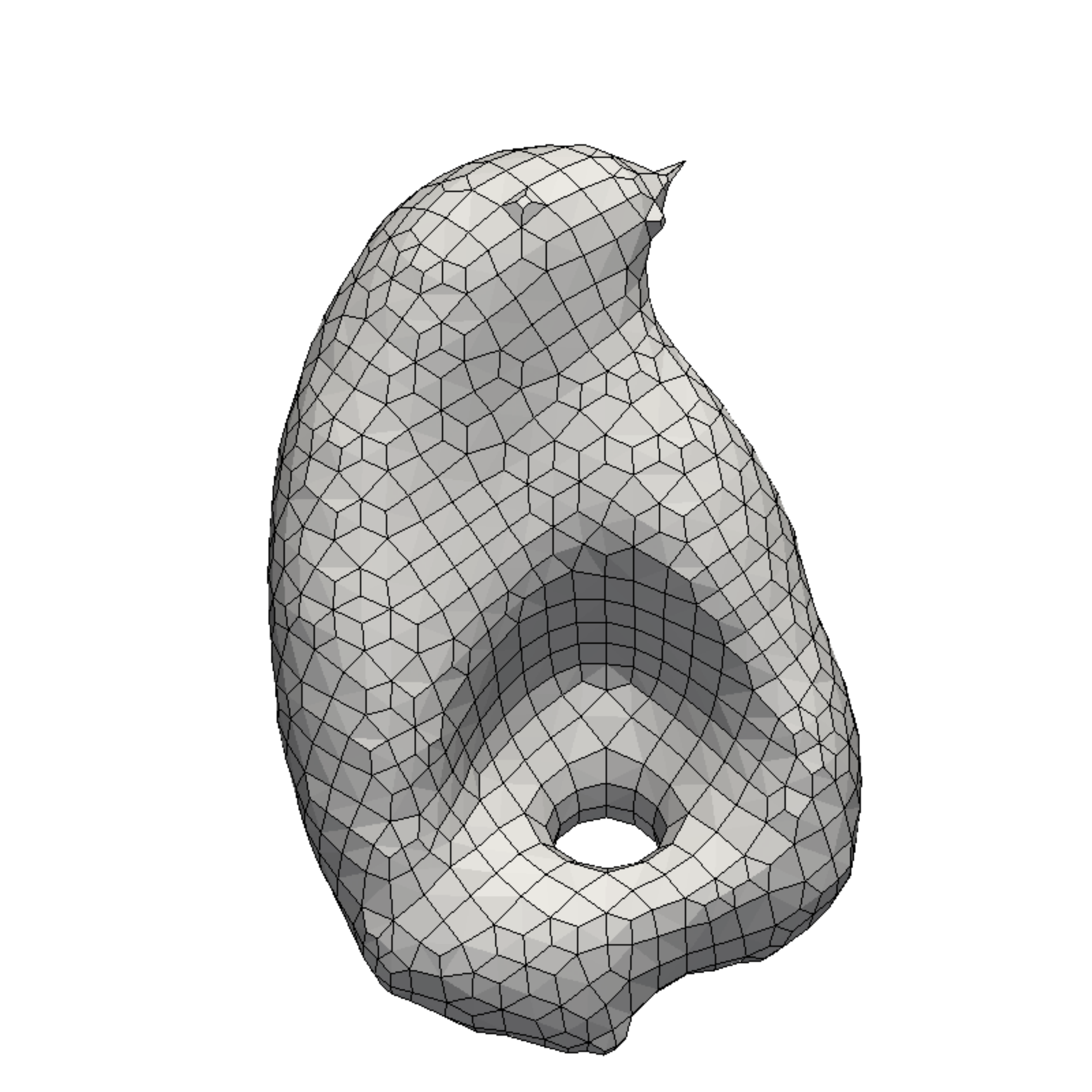}\\

\xrowht{20pt} 
\includegraphics[width=0.25\textwidth]{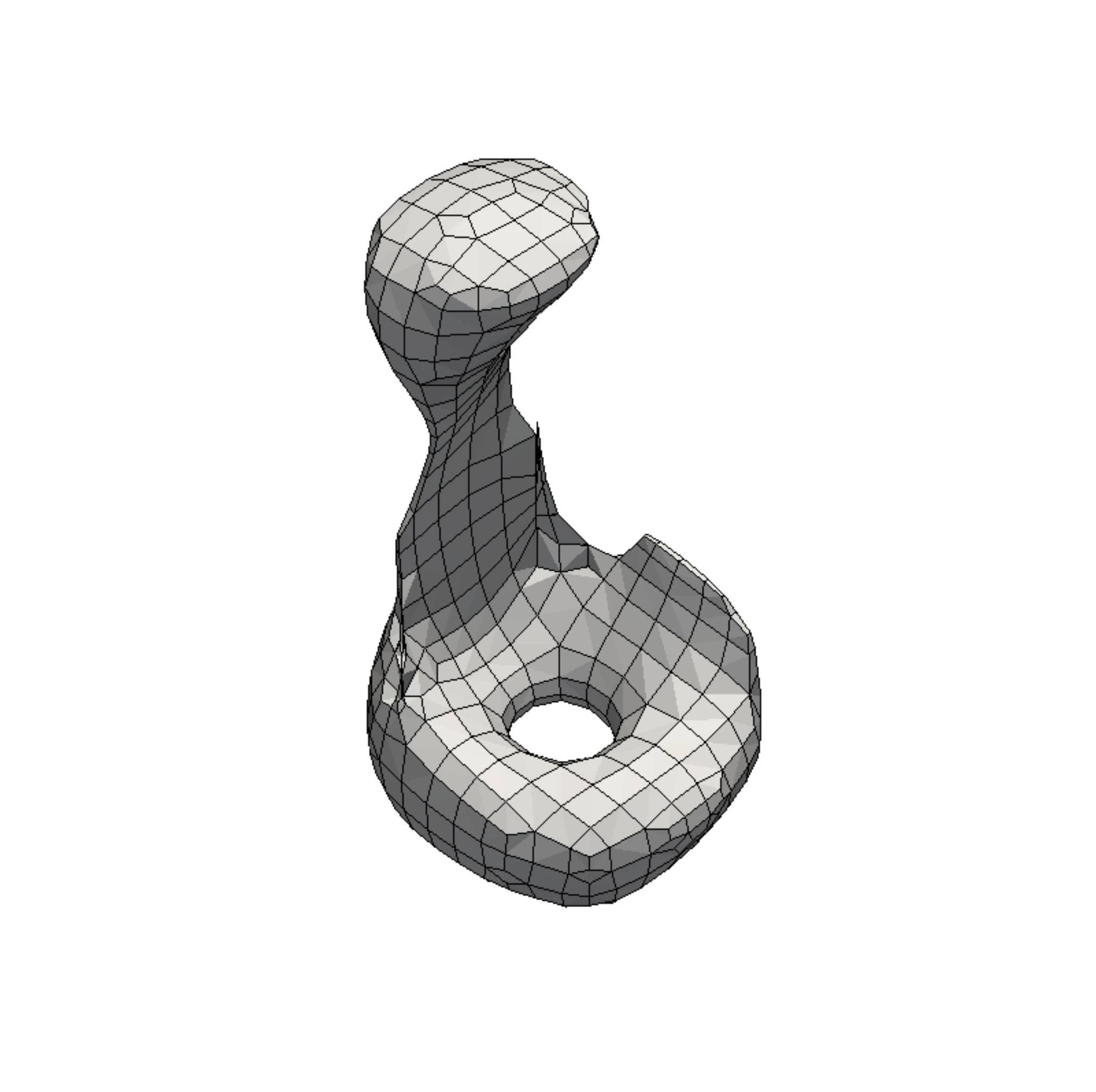} & \includegraphics[width=0.25\textwidth]{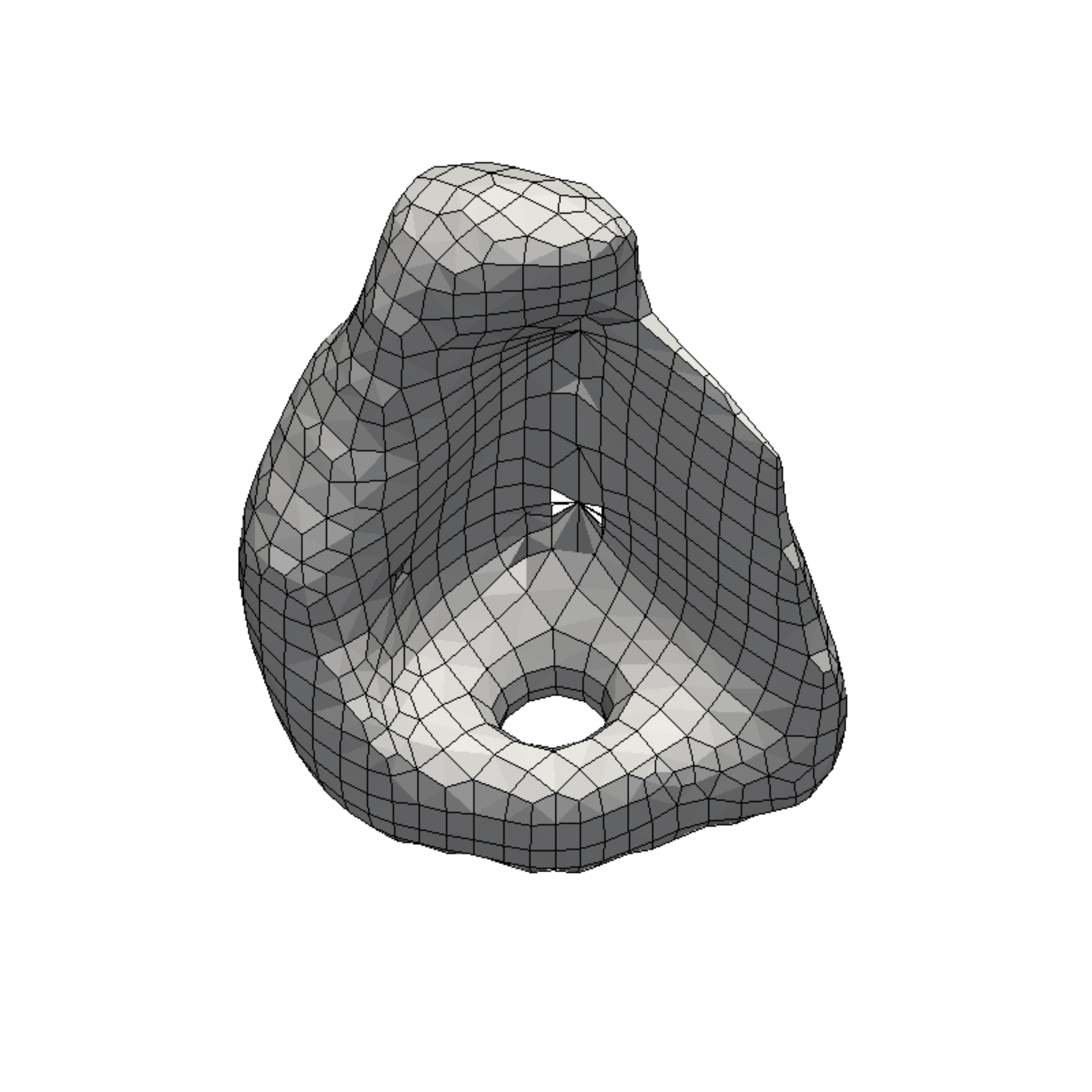} & \includegraphics[width=0.25\textwidth]{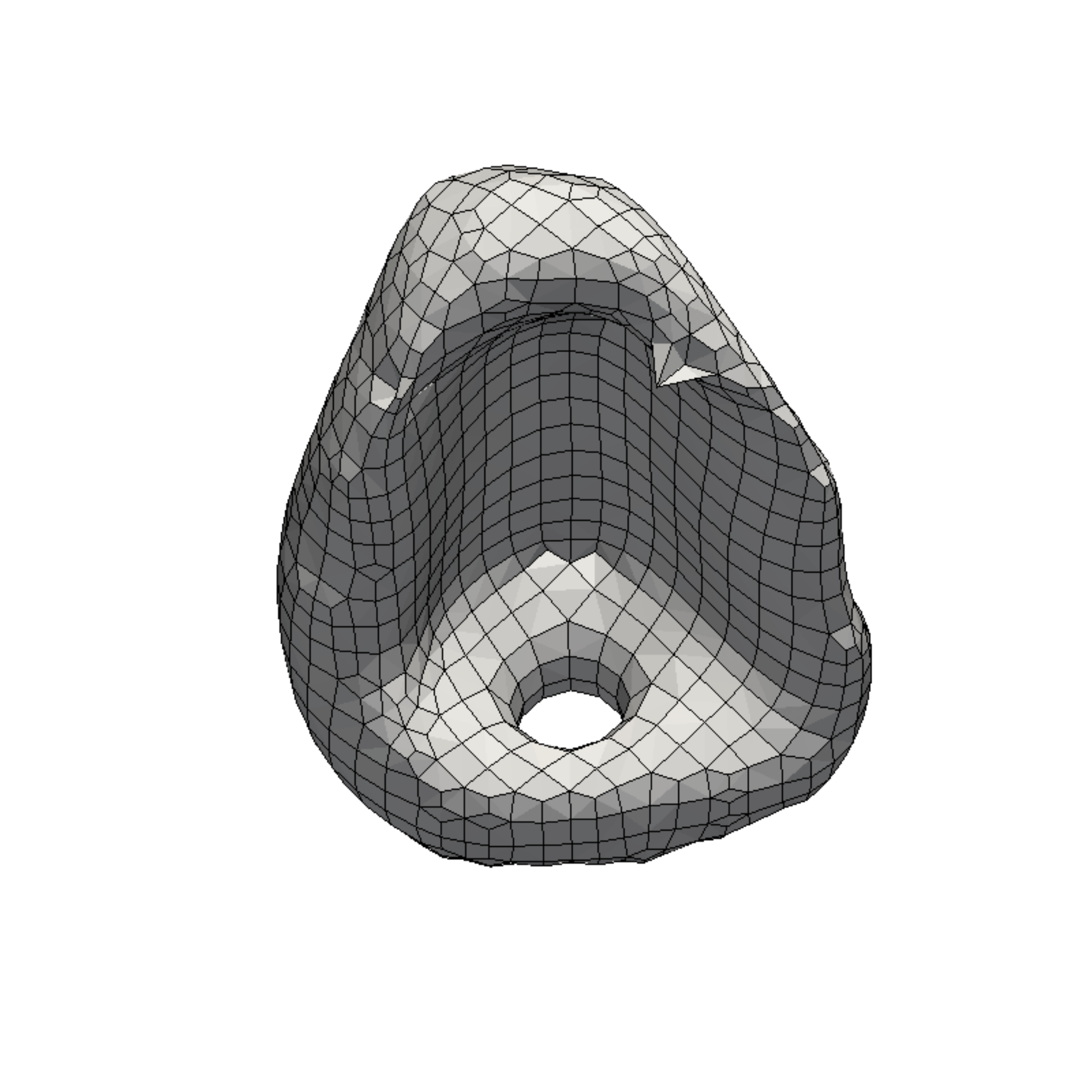}\\

\hline
\end{tabular}

}
        \subcaption{sphere simple}
    \end{subtable}
    \hfill
    \begin{subtable}[t]{0.44\textwidth}
        \centering
        \setlength\tabcolsep{6pt}
        \resizebox{0.85\textwidth}{!}{\begin{tabular}{|c|c||c|}
\hline
\parbox{4em}{\centering UNet} & \parbox{4em}{\centering UNet\\+physics} & \xrowht{20pt}\parbox{4em}{\centering ground\\truth} \\
\hline\rule{0pt}{2cm}

\xrowht{20pt} 
\includegraphics[width=0.25\textwidth]{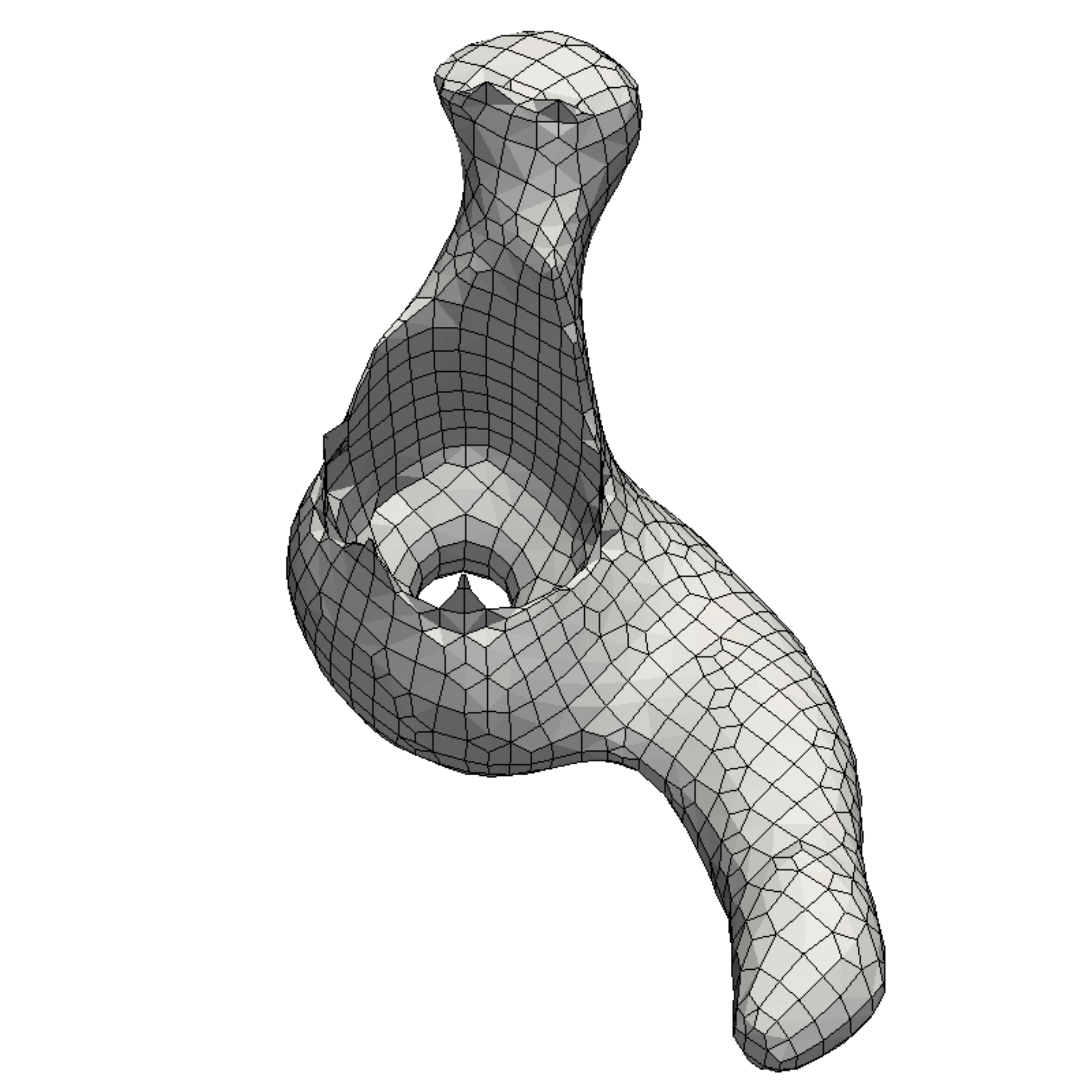} & \includegraphics[width=0.25\textwidth]{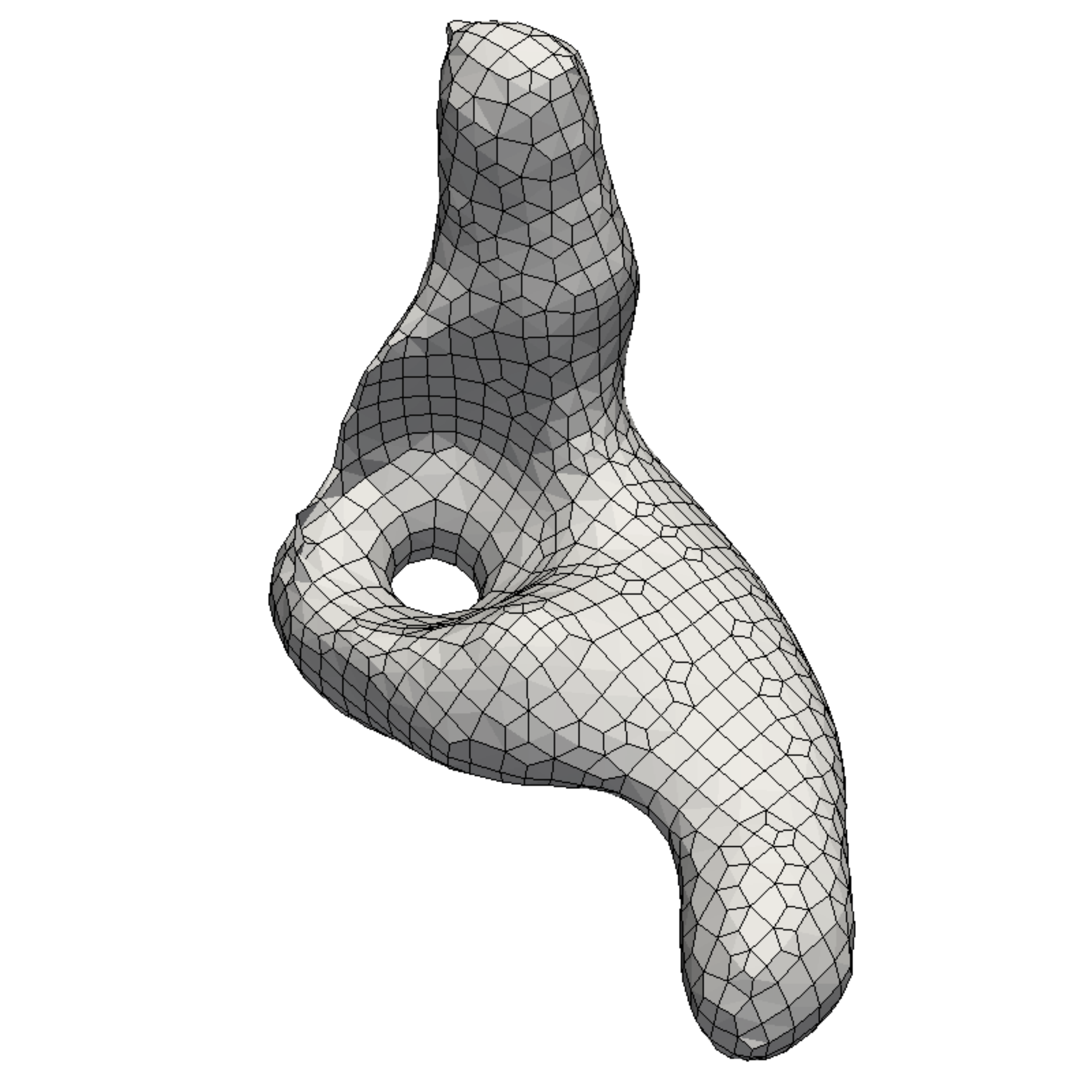} & \includegraphics[width=0.25\textwidth]{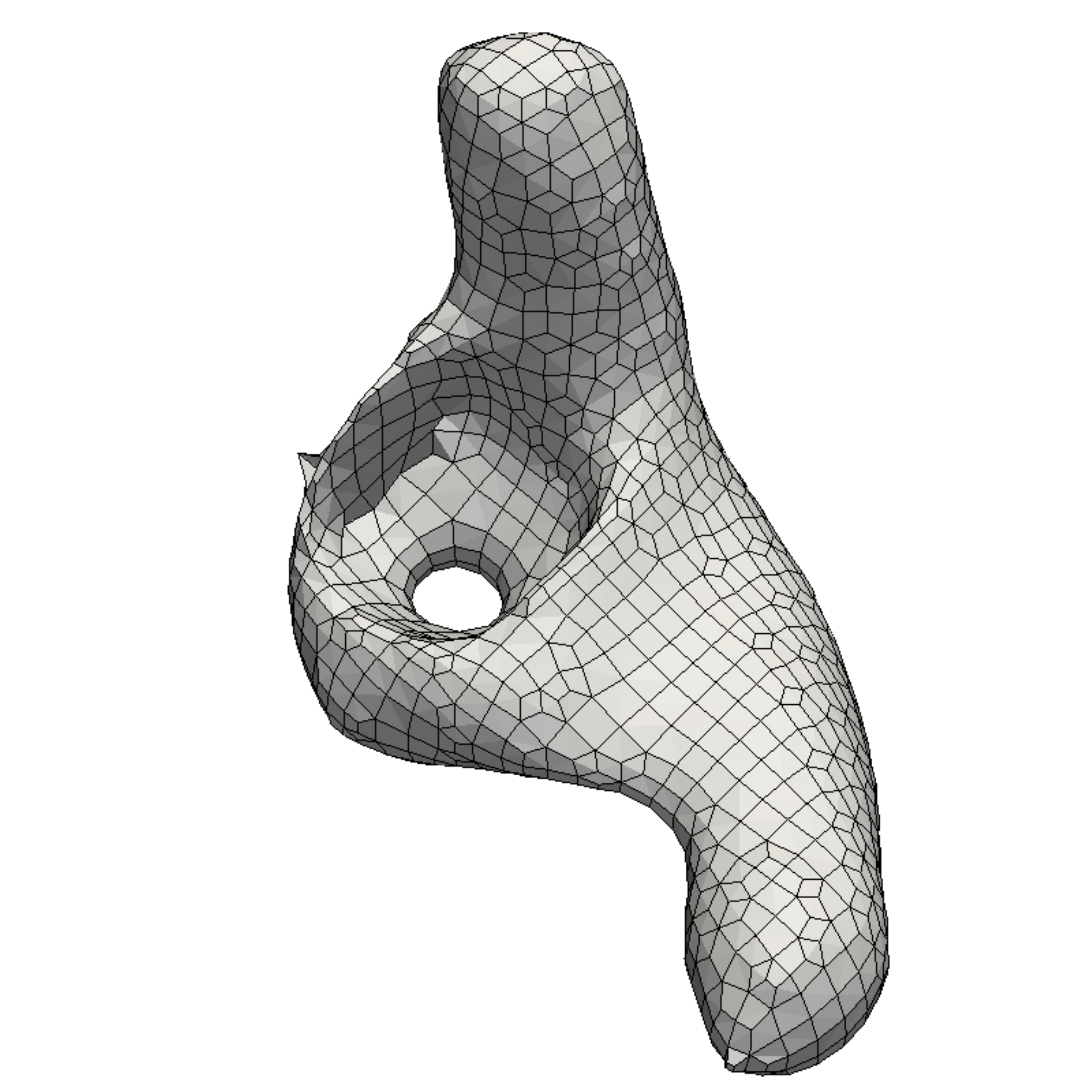}\\

\xrowht{20pt} 
\includegraphics[width=0.25\textwidth]{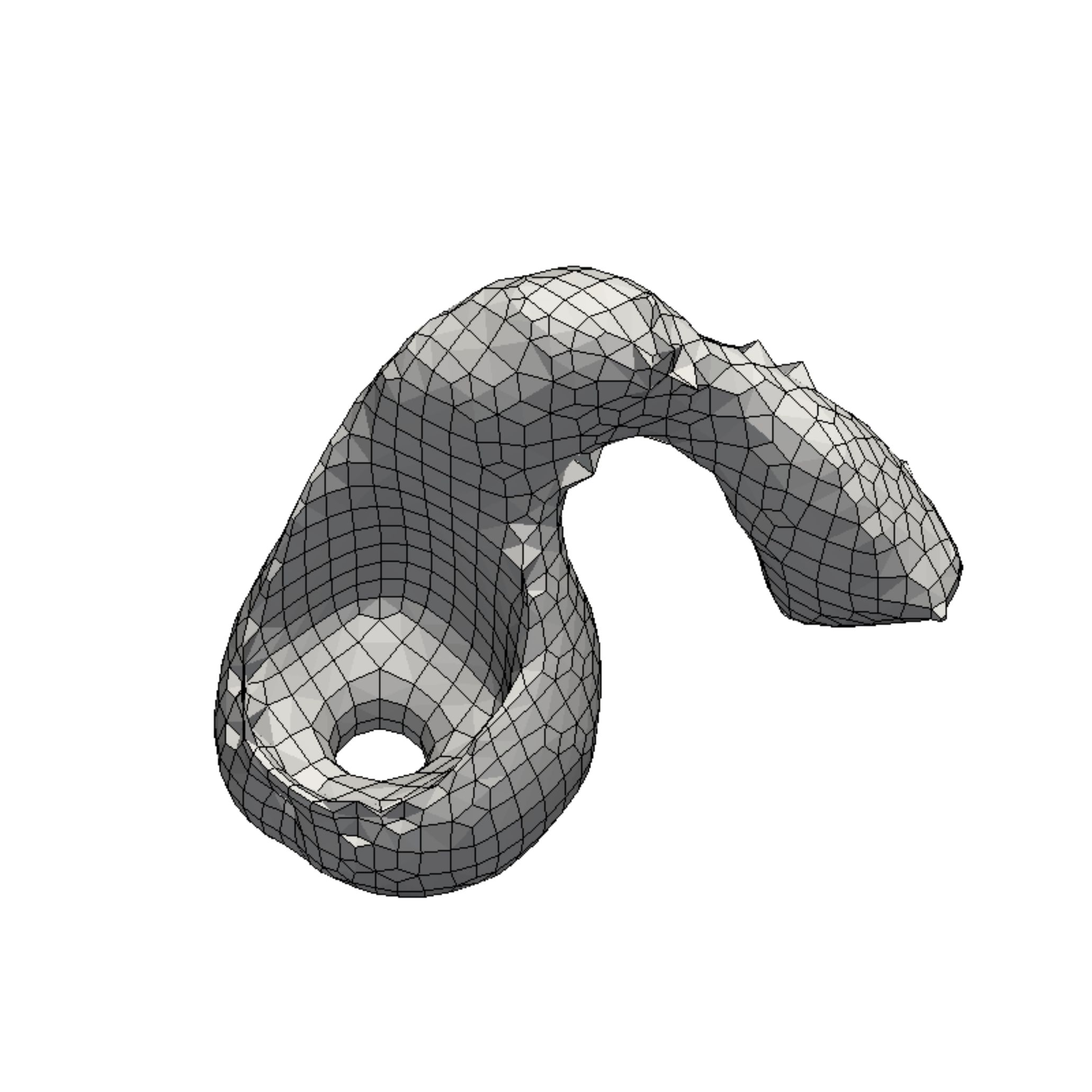} & \includegraphics[width=0.25\textwidth]{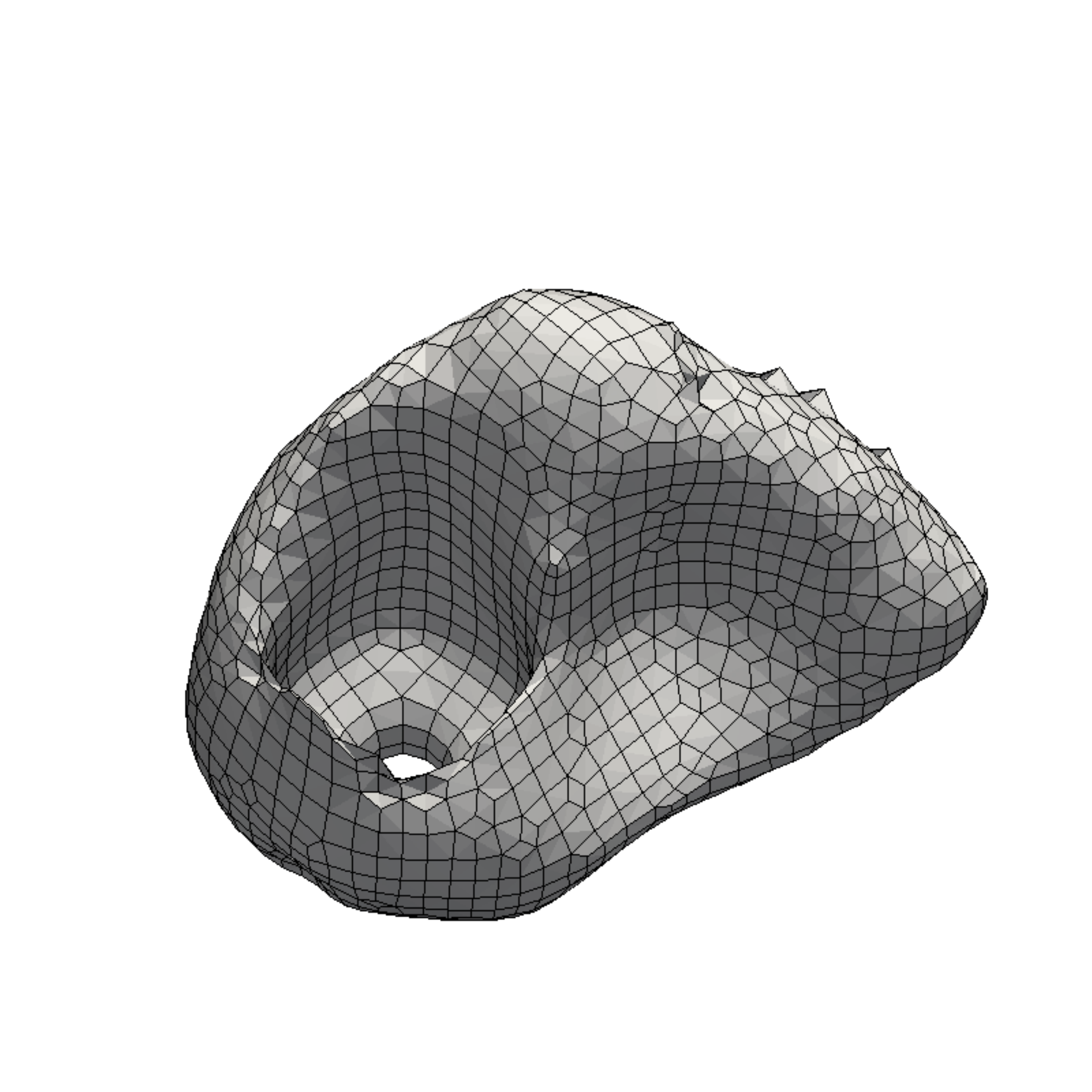} & \includegraphics[width=0.25\textwidth]{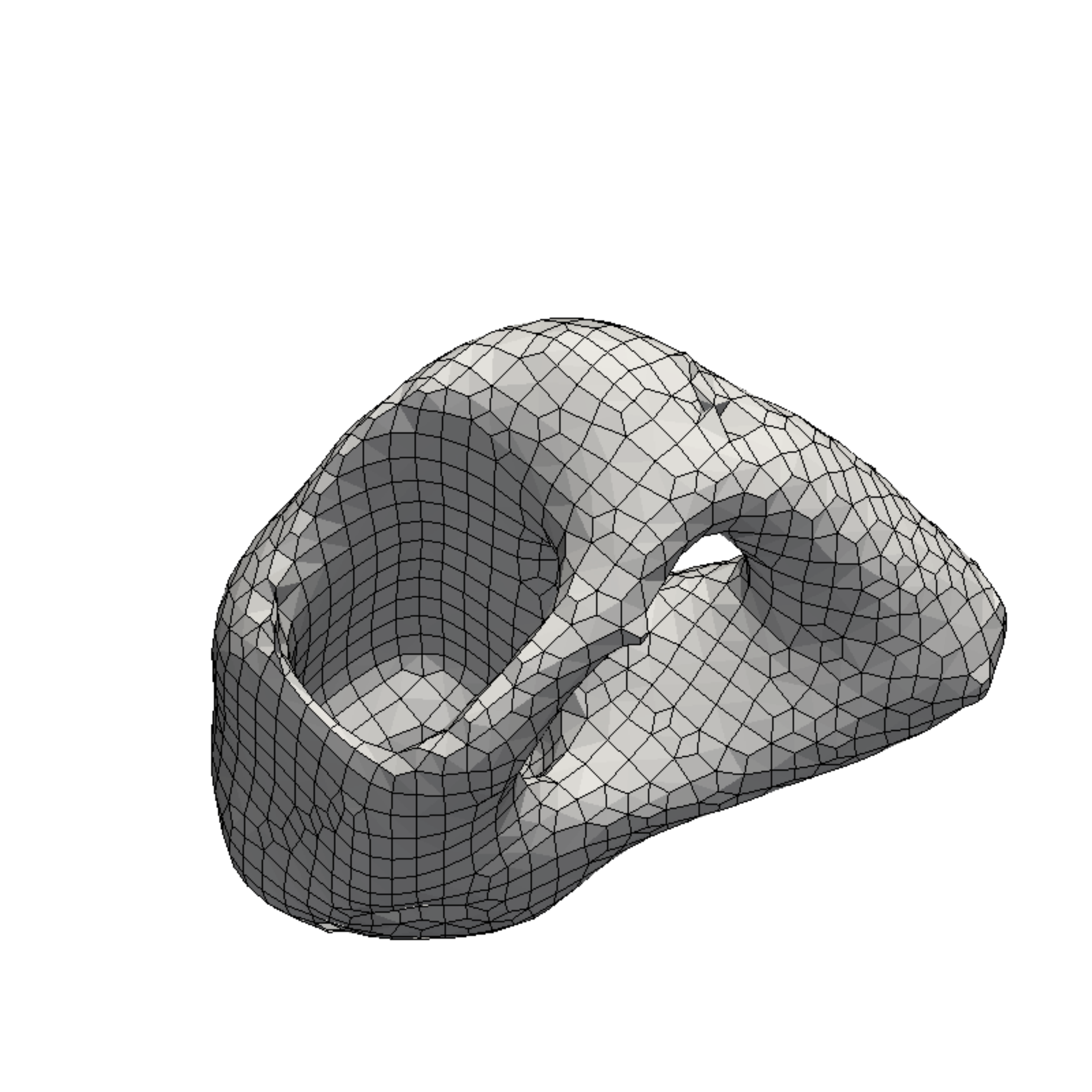}\\

\xrowht{20pt} 
\includegraphics[width=0.25\textwidth]{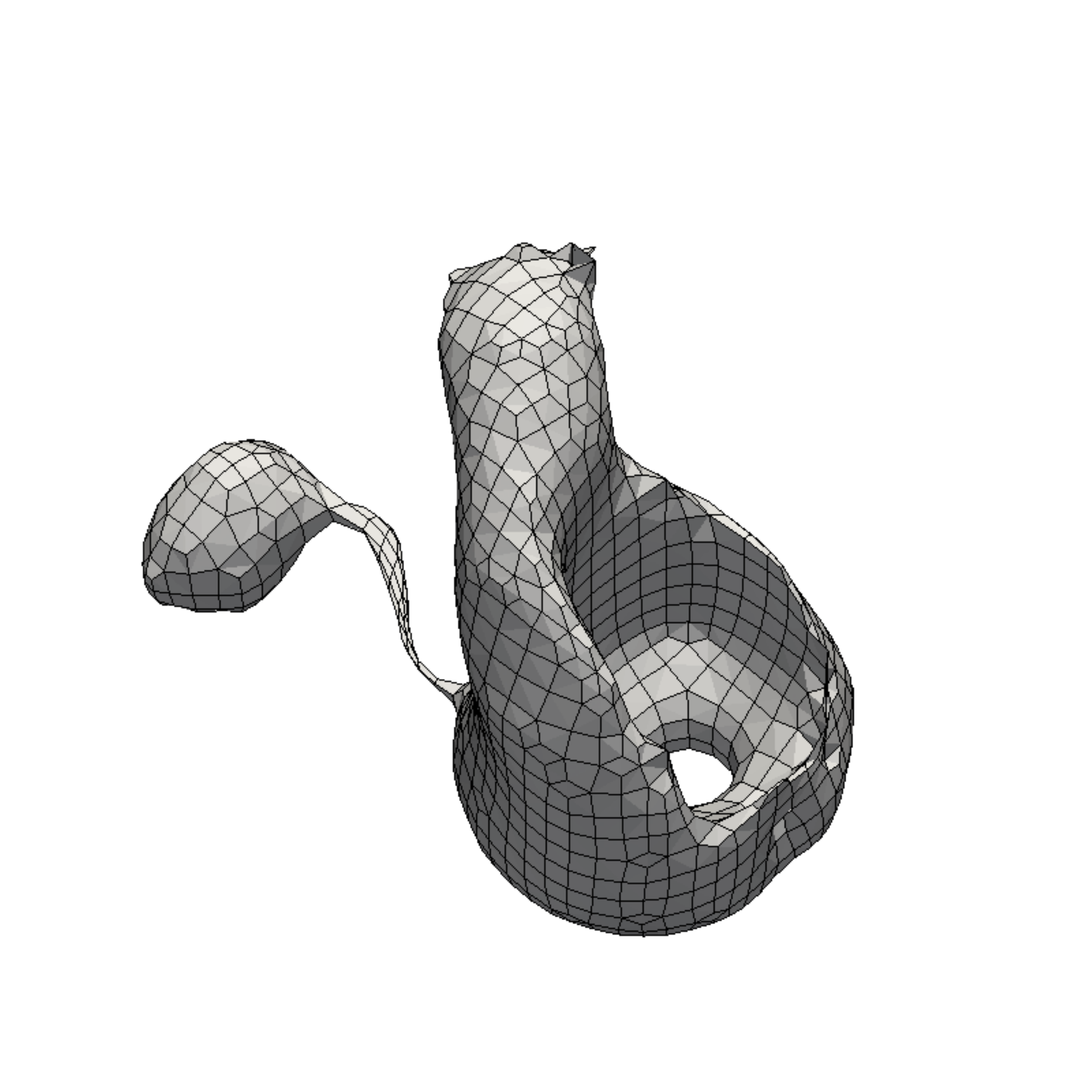} & \includegraphics[width=0.25\textwidth]{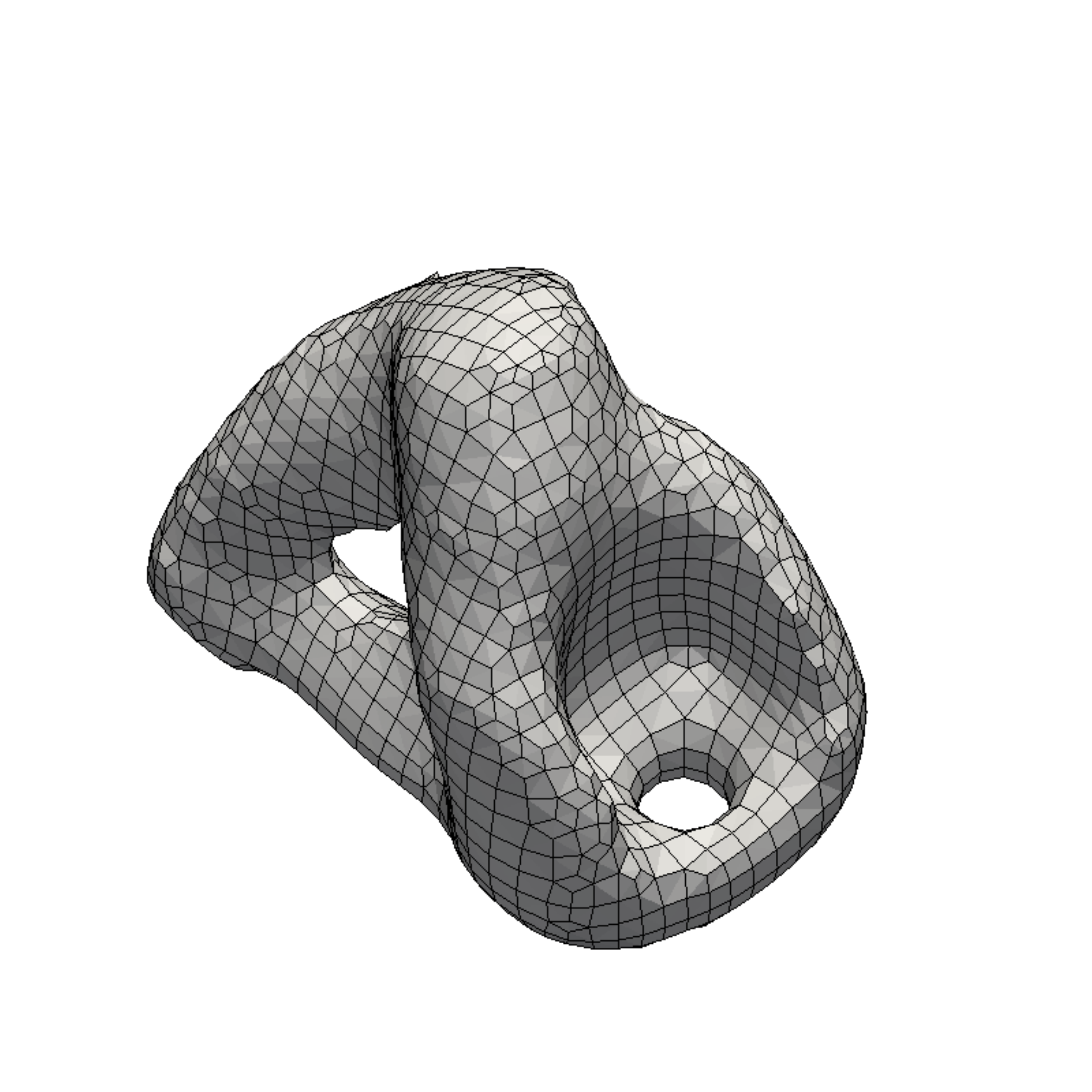} & \includegraphics[width=0.25\textwidth]{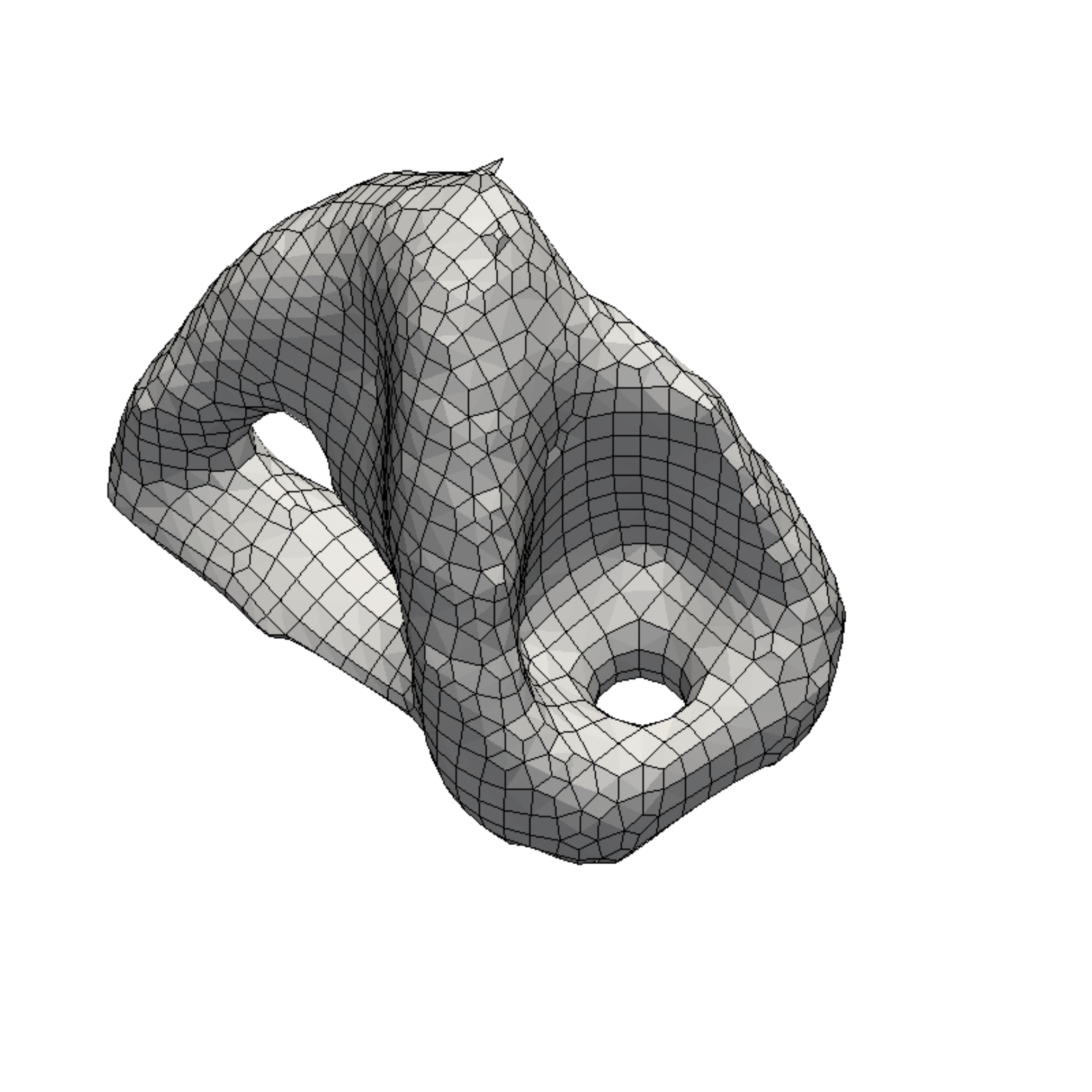}\\

\xrowht{20pt} 
\includegraphics[width=0.25\textwidth]{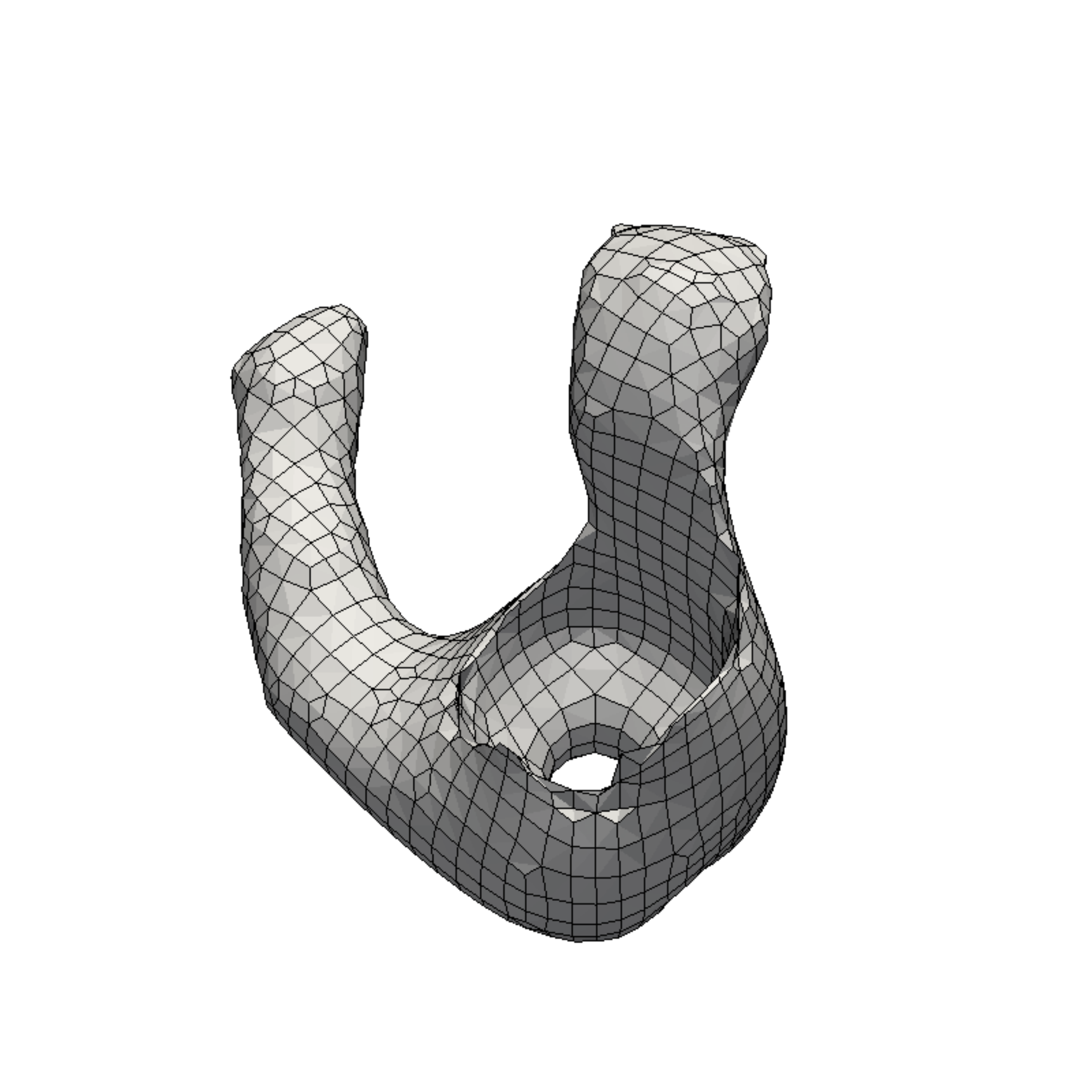} & \includegraphics[width=0.25\textwidth]{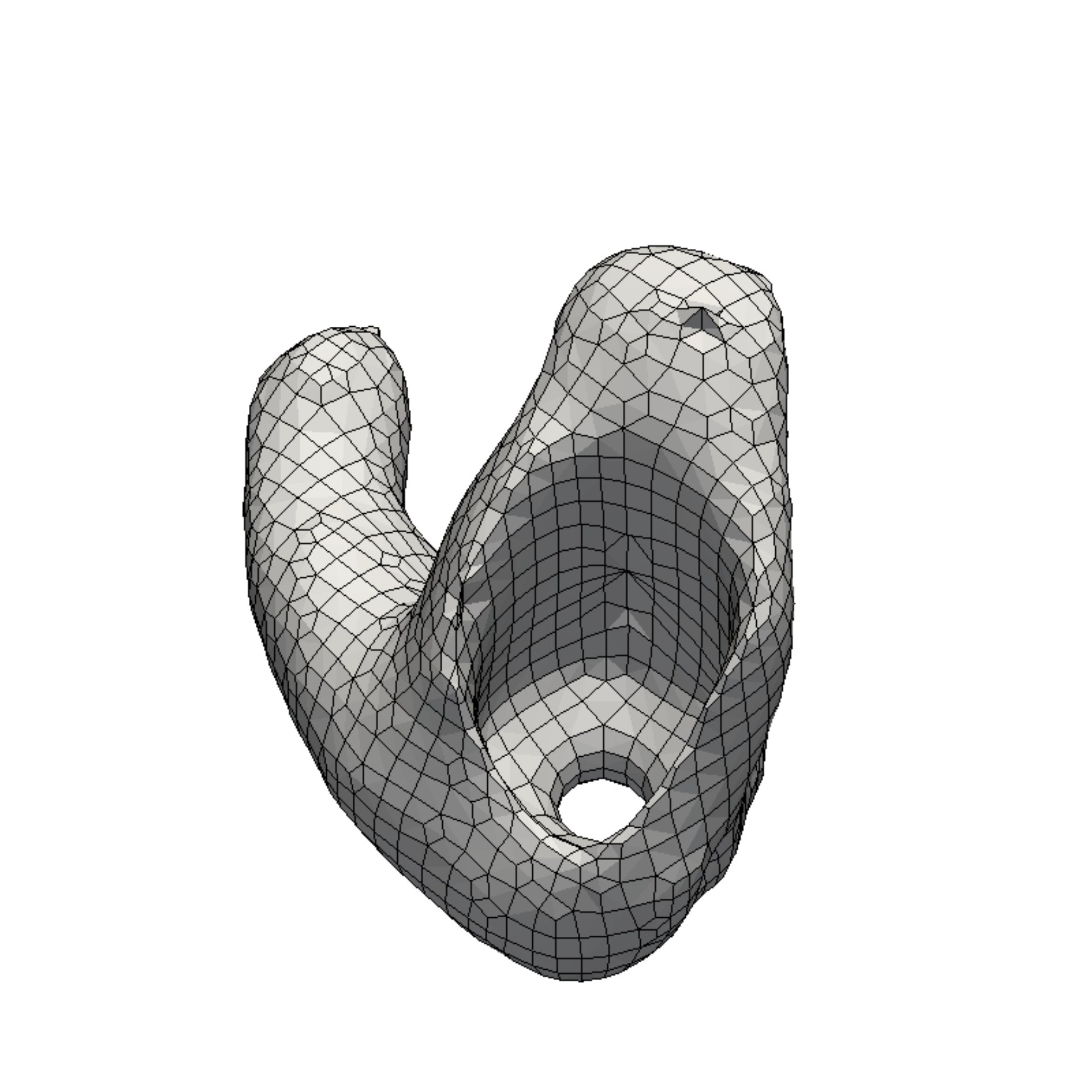} & \includegraphics[width=0.25\textwidth]{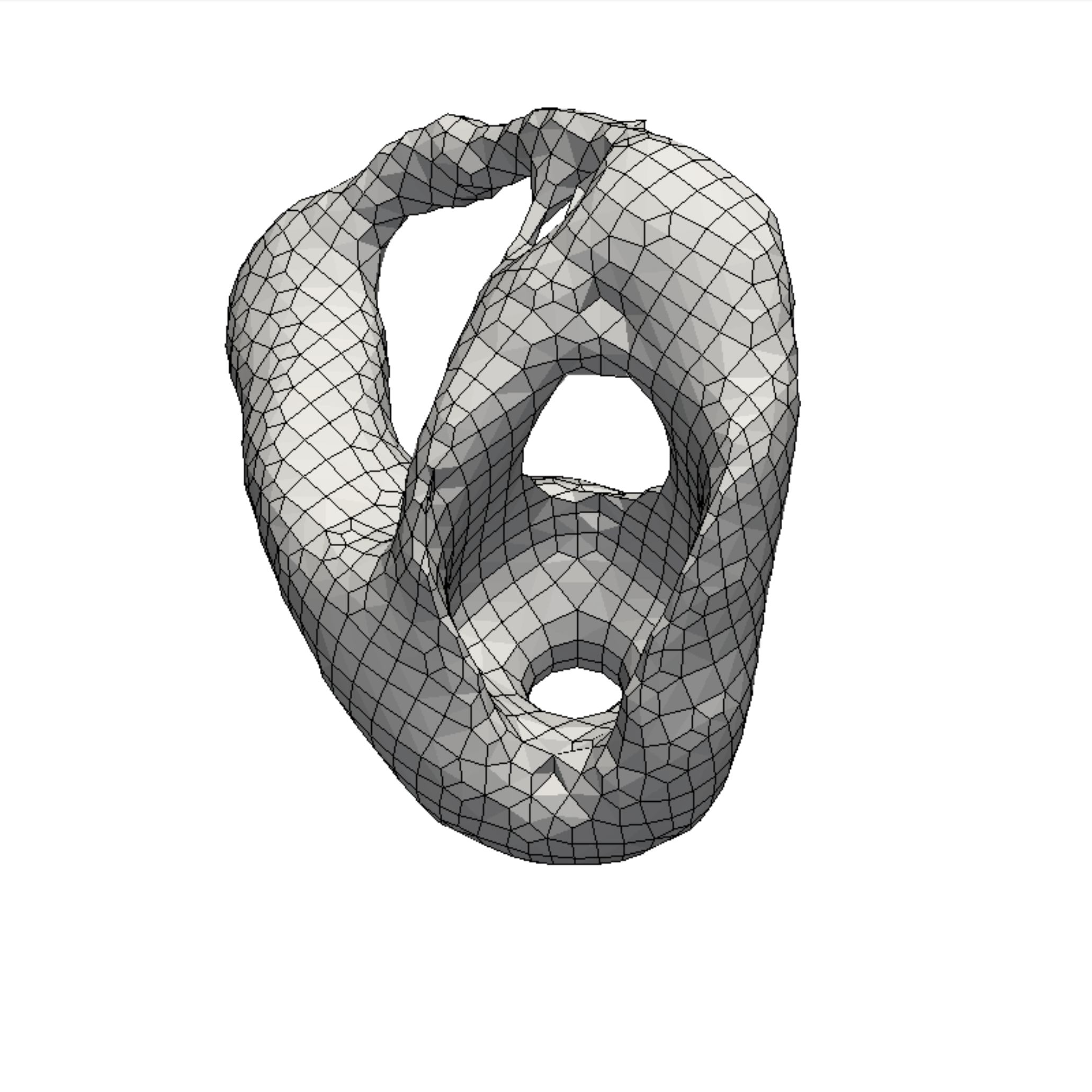}\\

\xrowht{20pt} 
\includegraphics[width=0.25\textwidth]{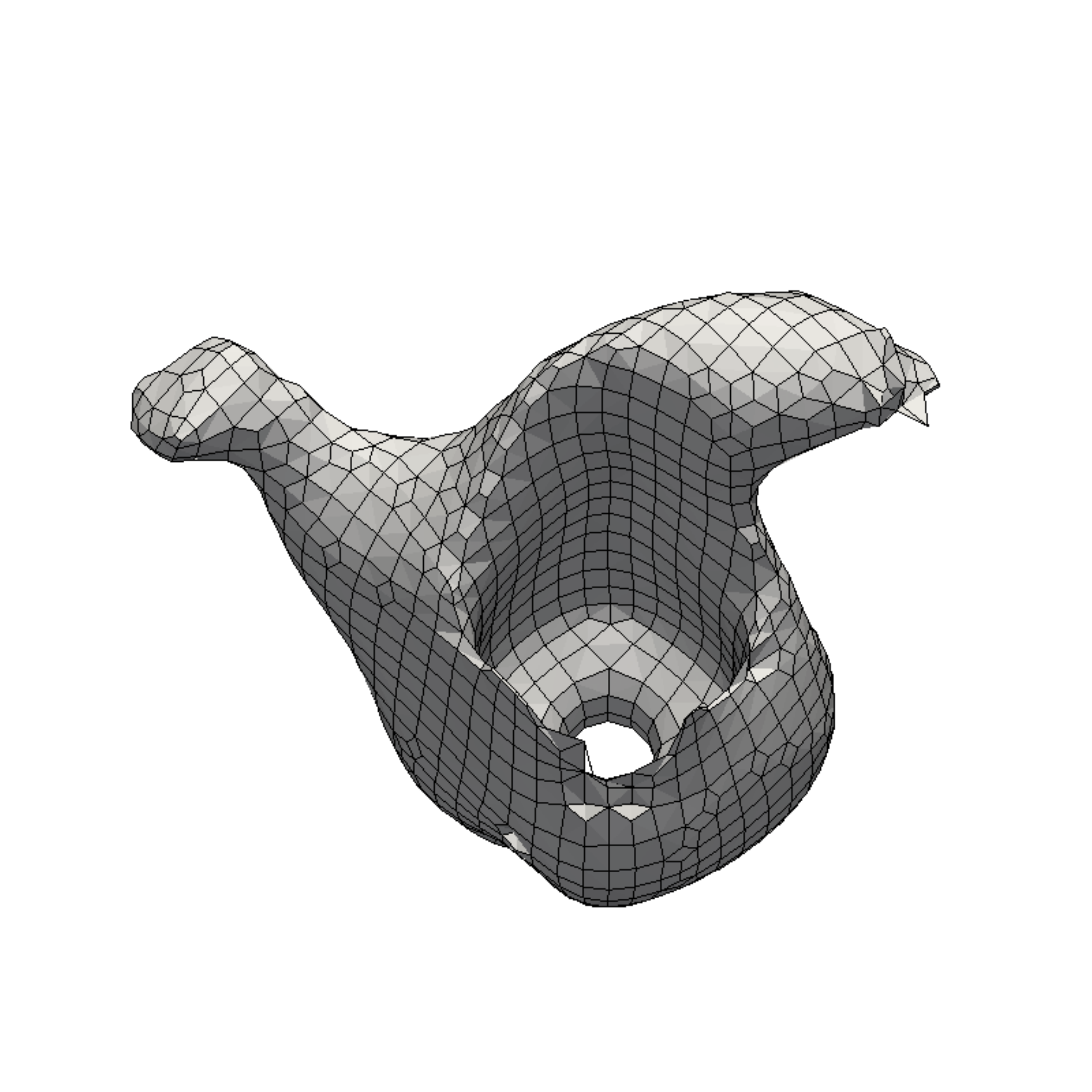} & \includegraphics[width=0.25\textwidth]{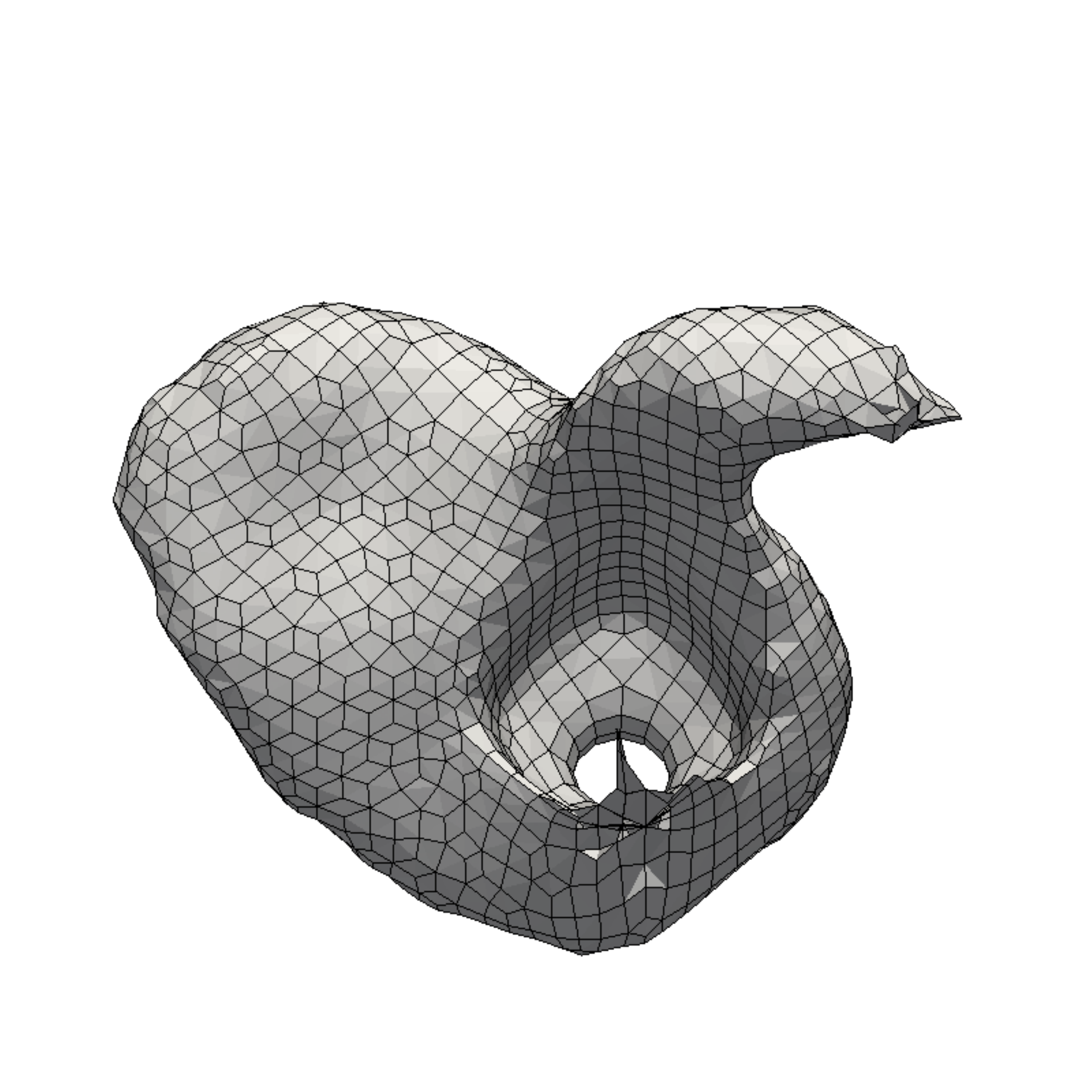} & \includegraphics[width=0.25\textwidth]{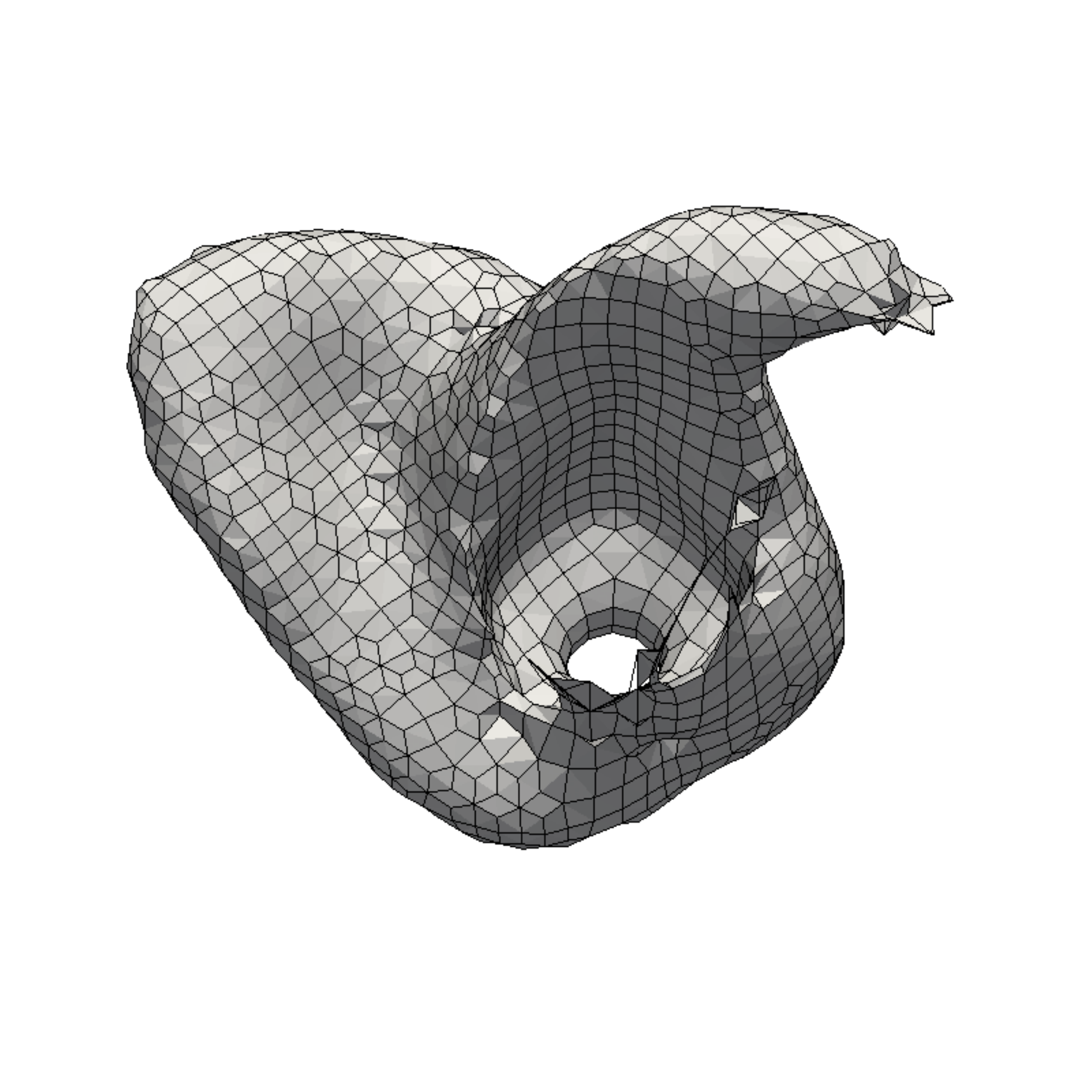}\\

\xrowht{20pt} 
\includegraphics[width=0.25\textwidth]{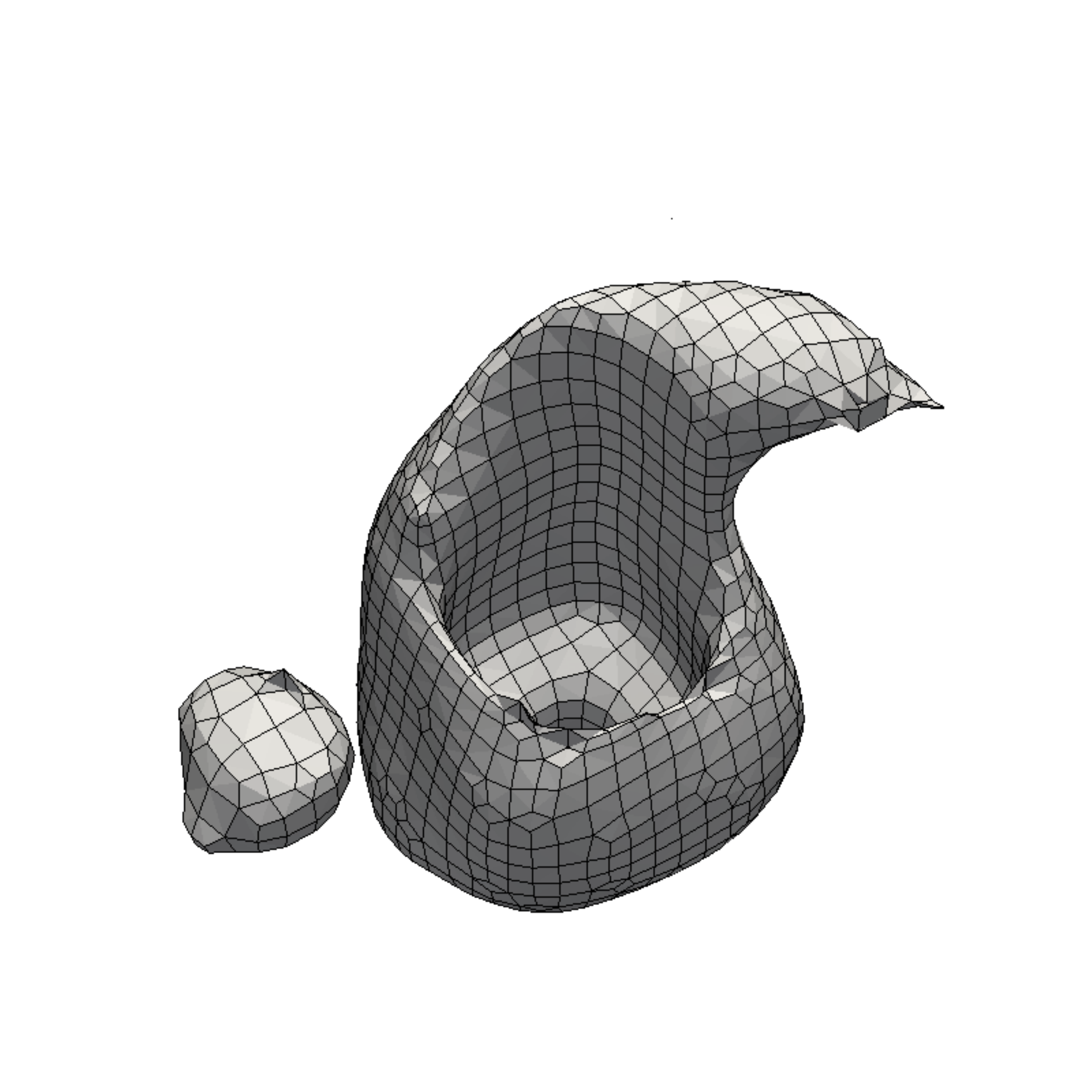} & \includegraphics[width=0.25\textwidth]{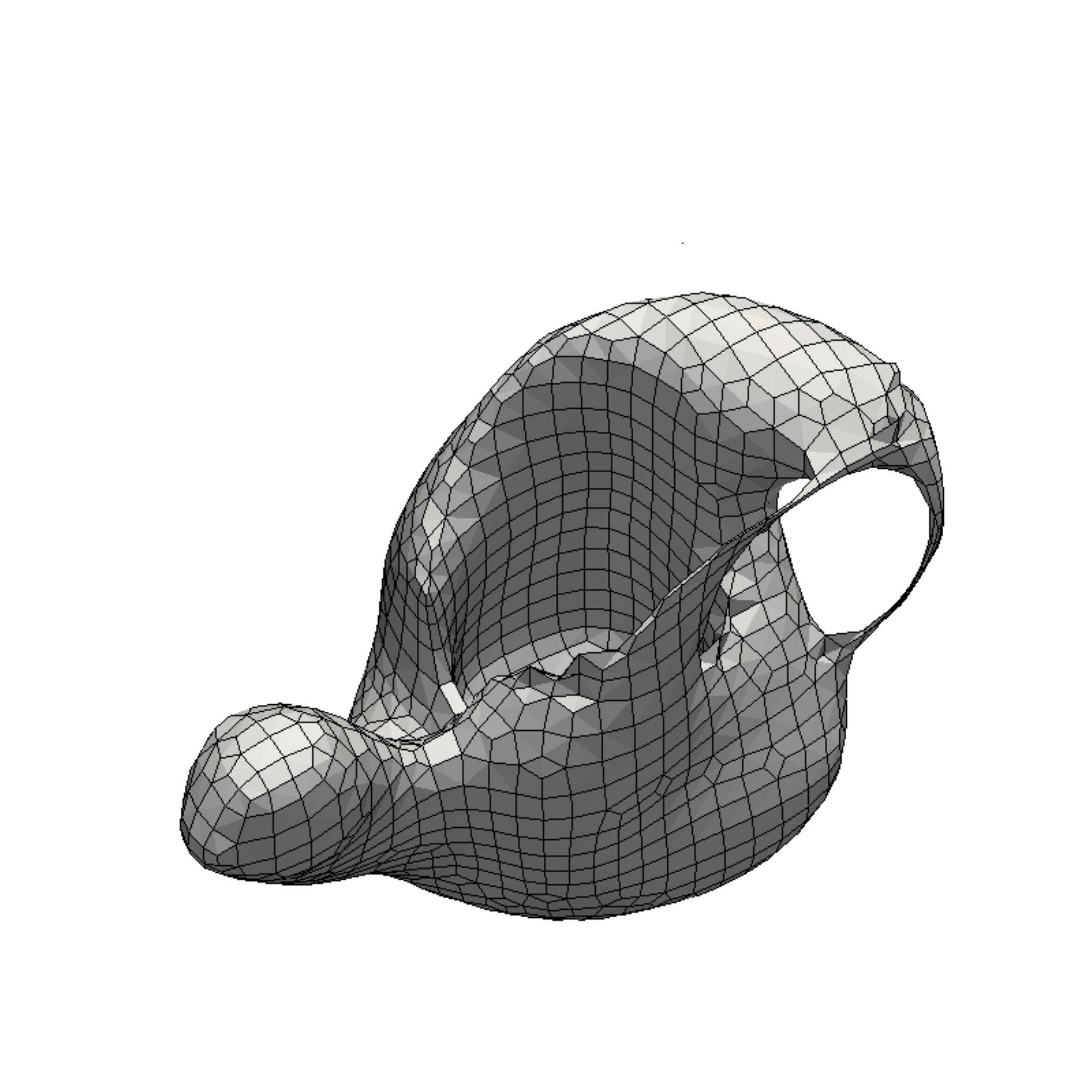} & \includegraphics[width=0.25\textwidth]{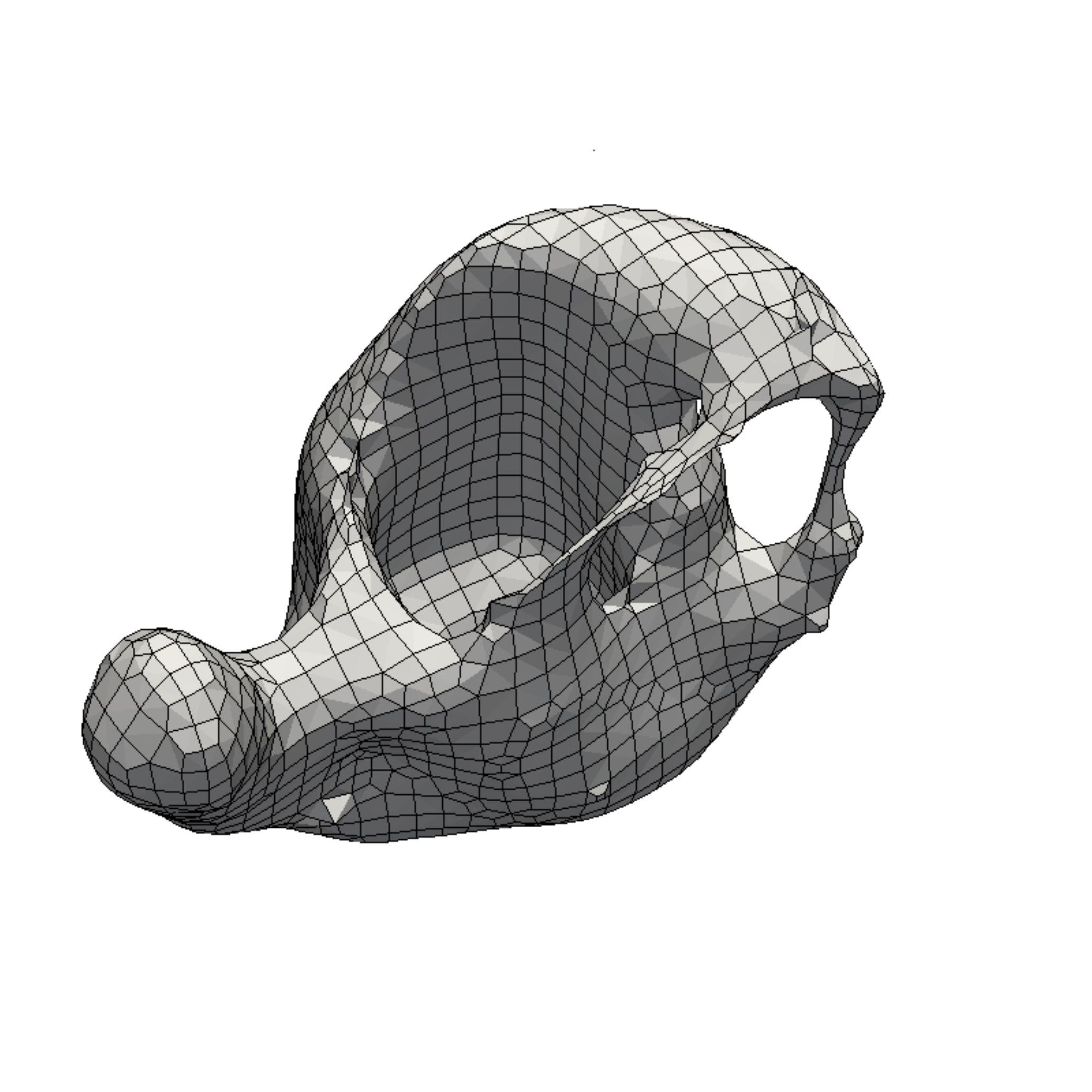}\\

\xrowht{20pt} 
\includegraphics[width=0.25\textwidth]{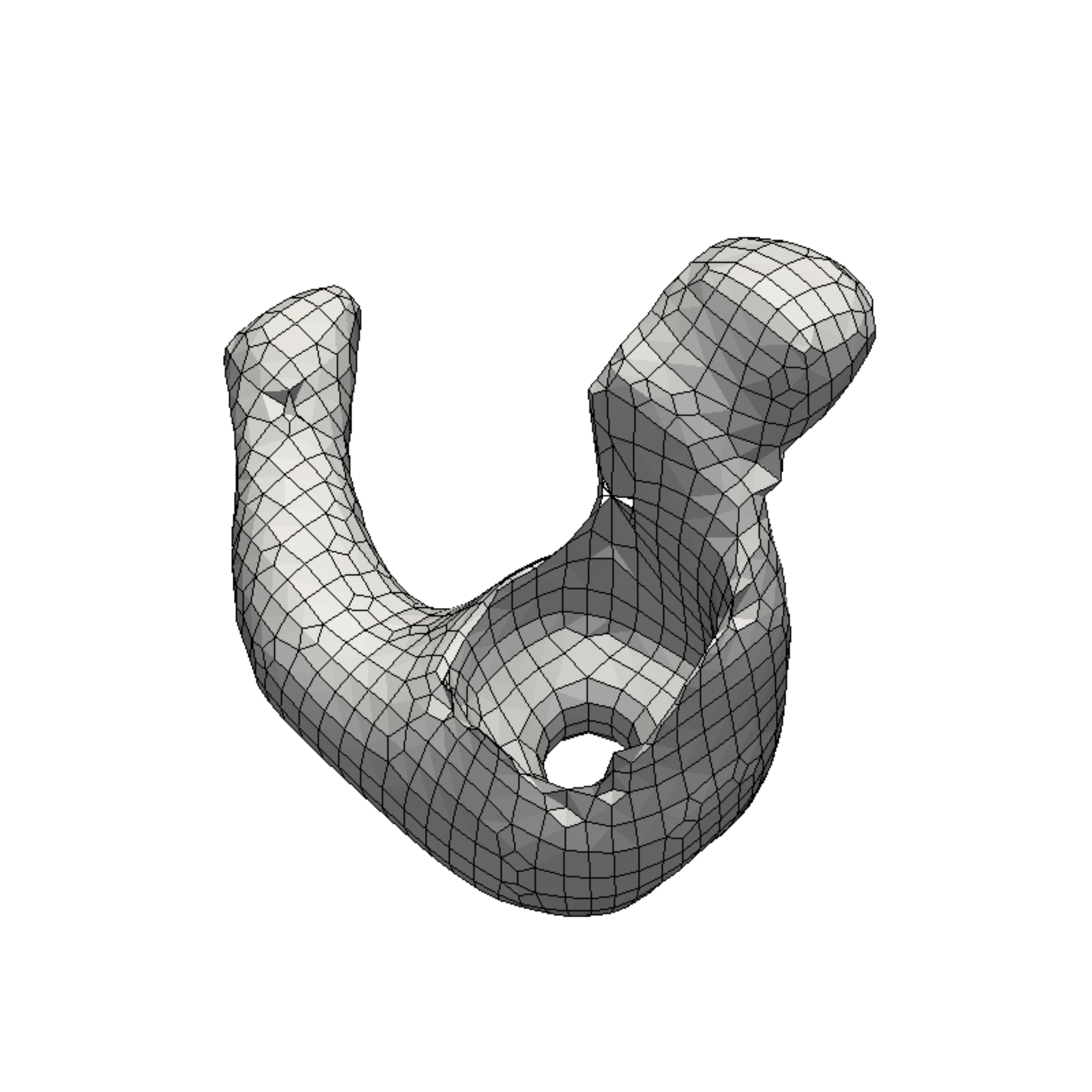} & \includegraphics[width=0.25\textwidth]{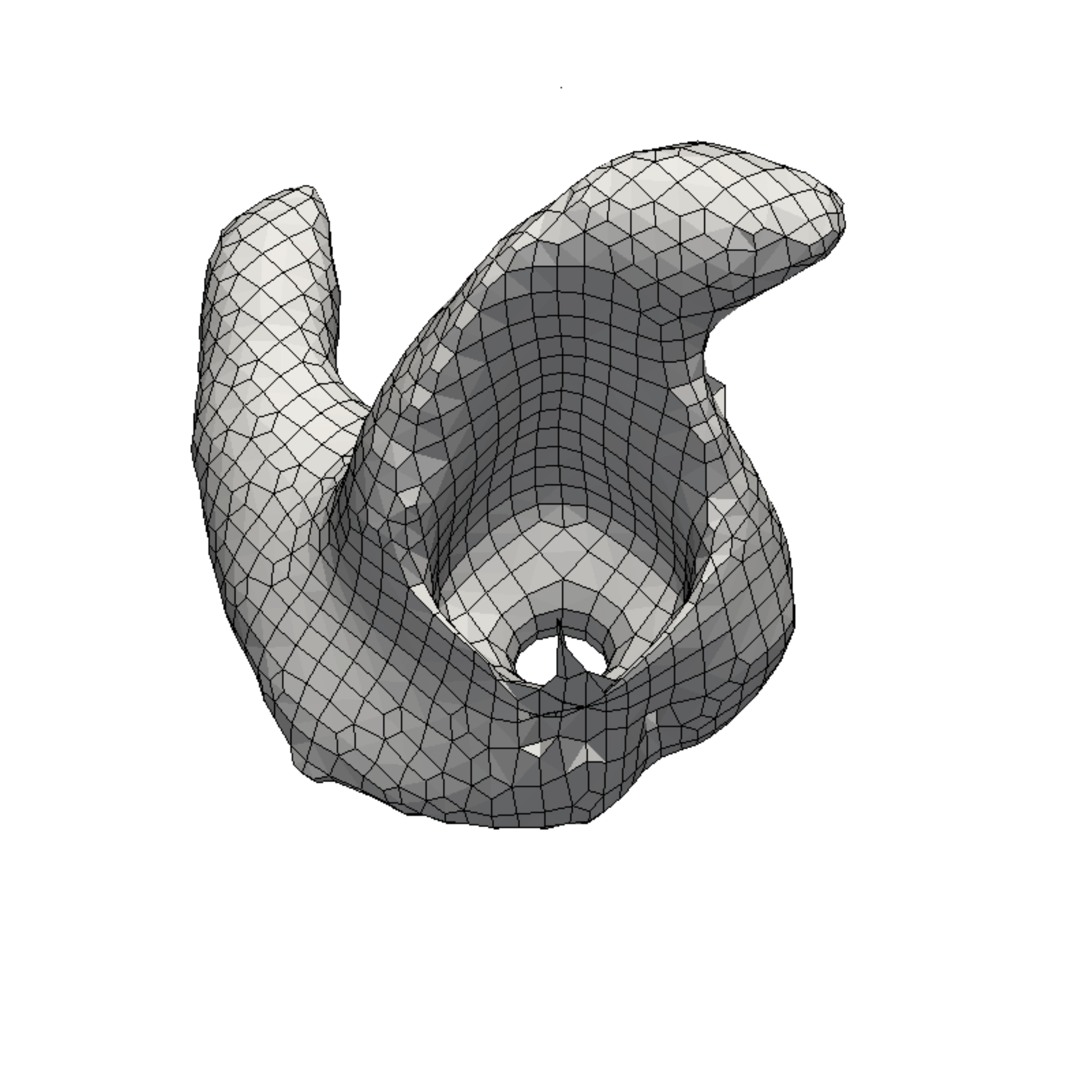} & \includegraphics[width=0.25\textwidth]{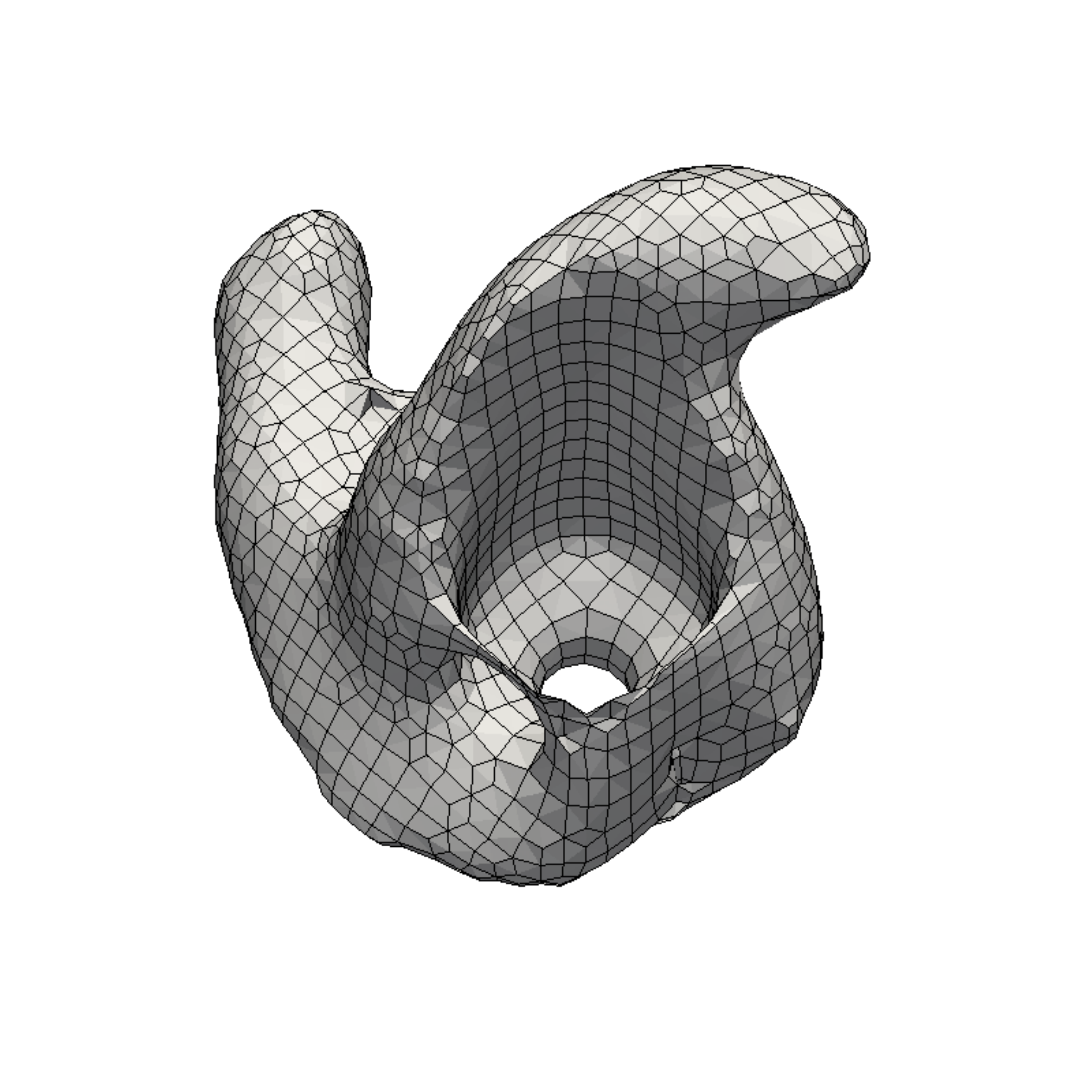}\\

\xrowht{20pt} 
\includegraphics[width=0.25\textwidth]{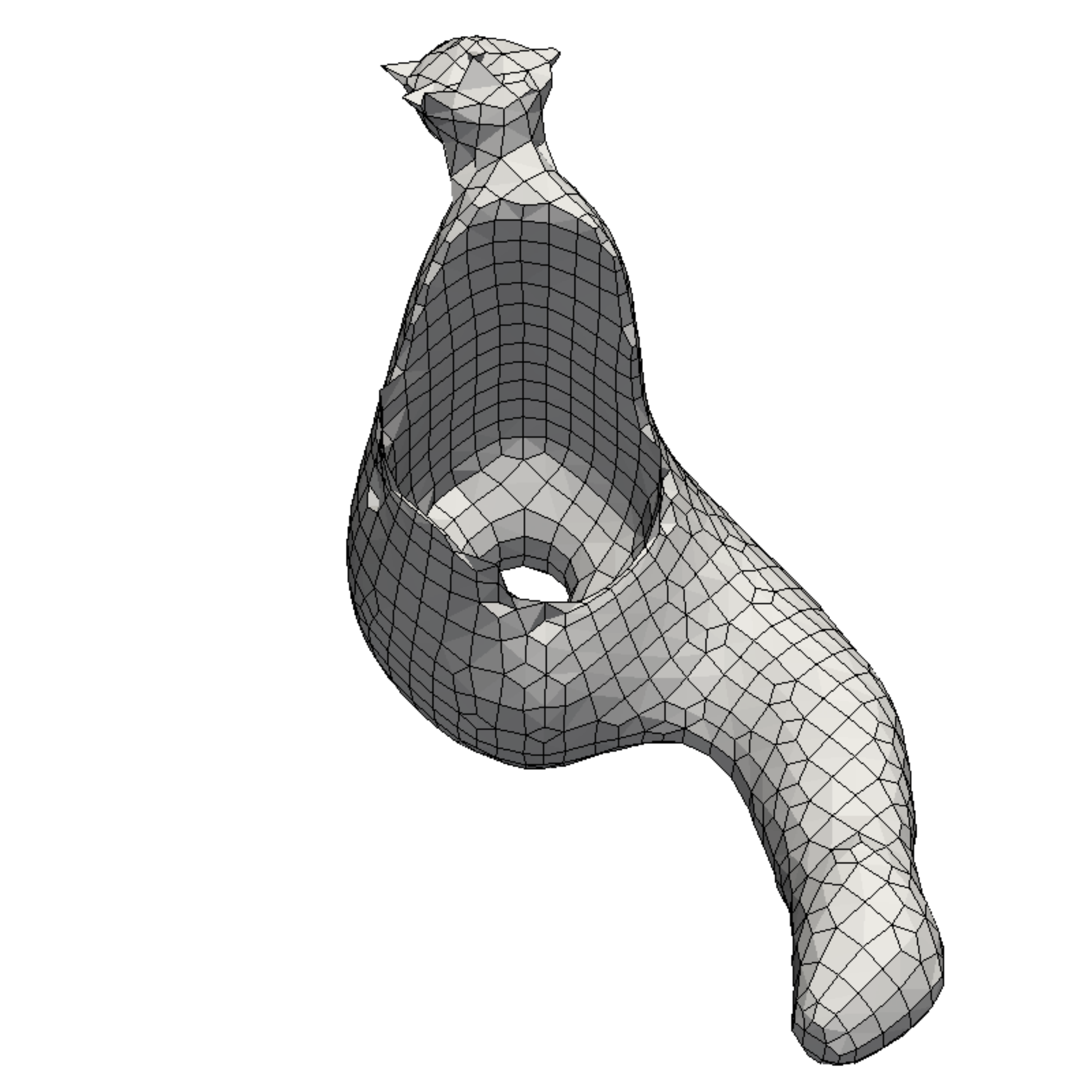} & \includegraphics[width=0.25\textwidth]{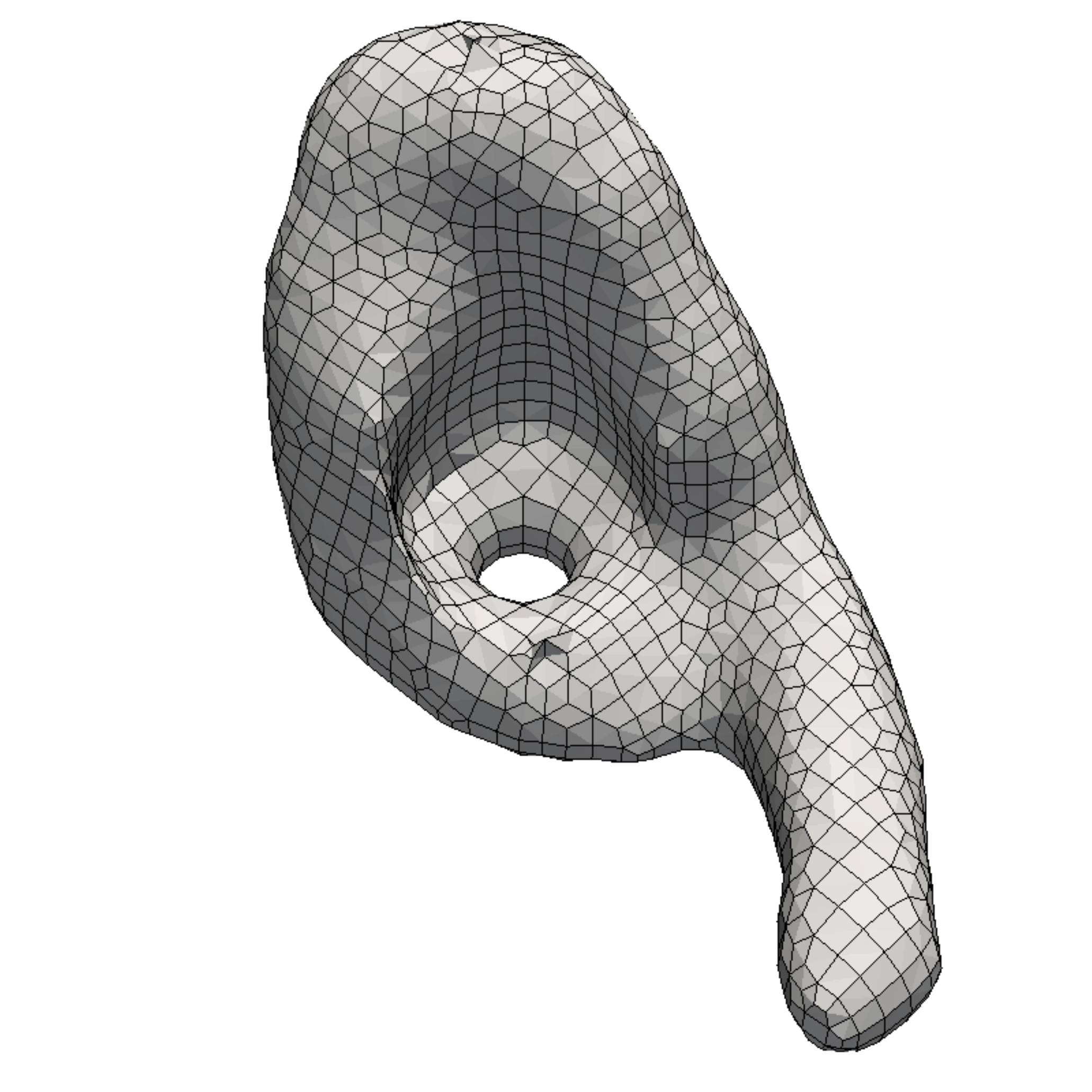} & \includegraphics[width=0.25\textwidth]{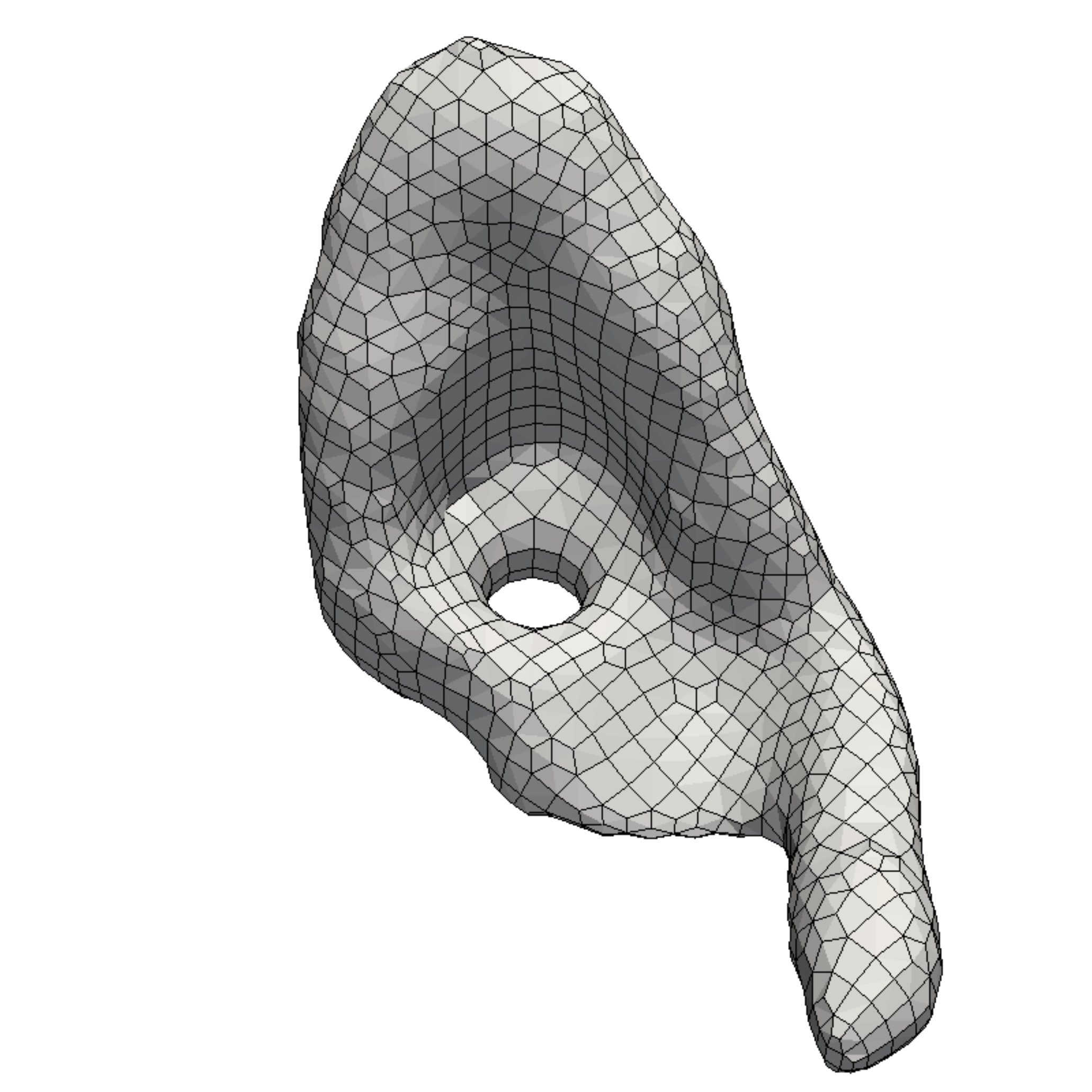}\\

\xrowht{20pt} 
\includegraphics[width=0.25\textwidth]{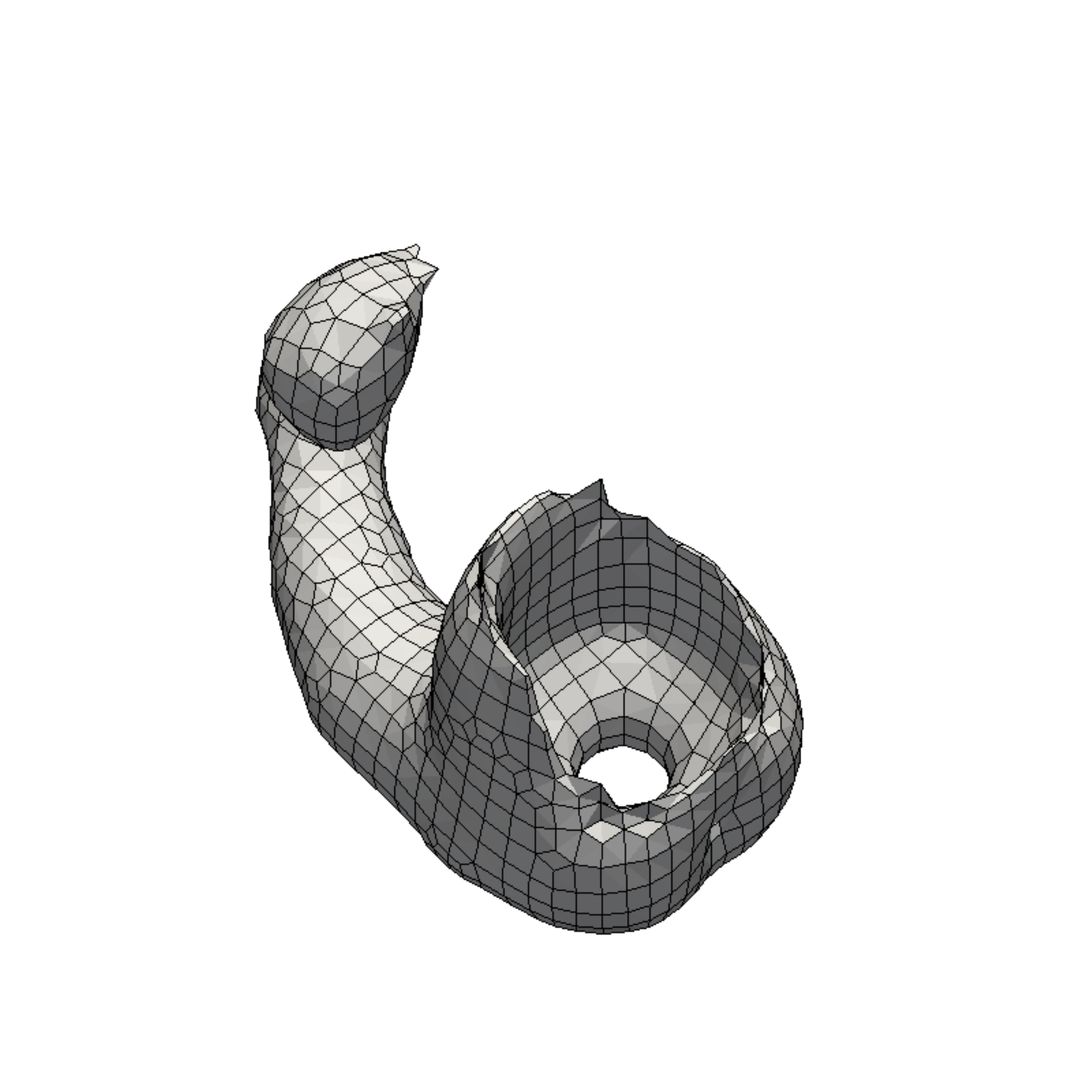} & \includegraphics[width=0.25\textwidth]{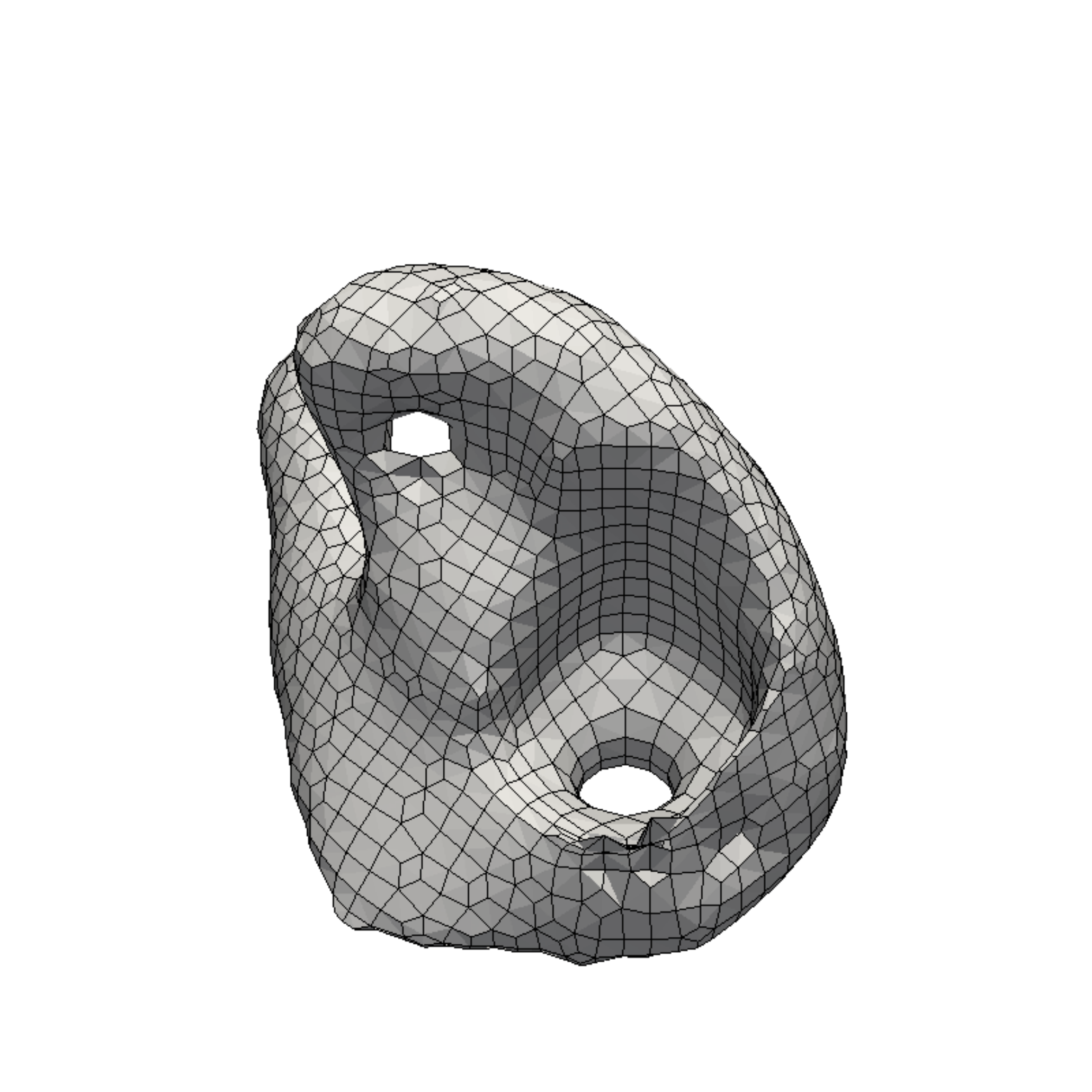} & \includegraphics[width=0.25\textwidth]{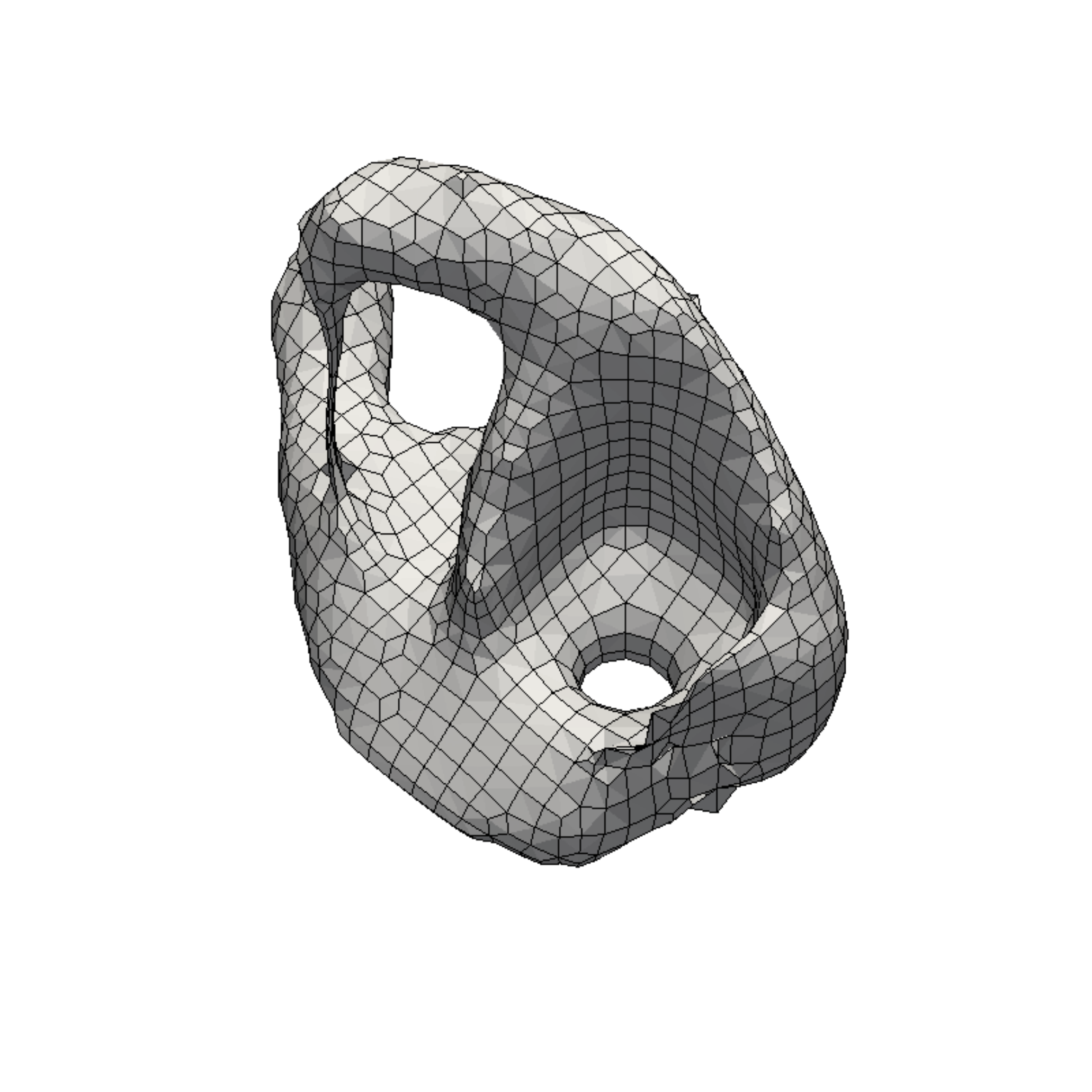}\\

\xrowht{20pt} 
\includegraphics[width=0.25\textwidth]{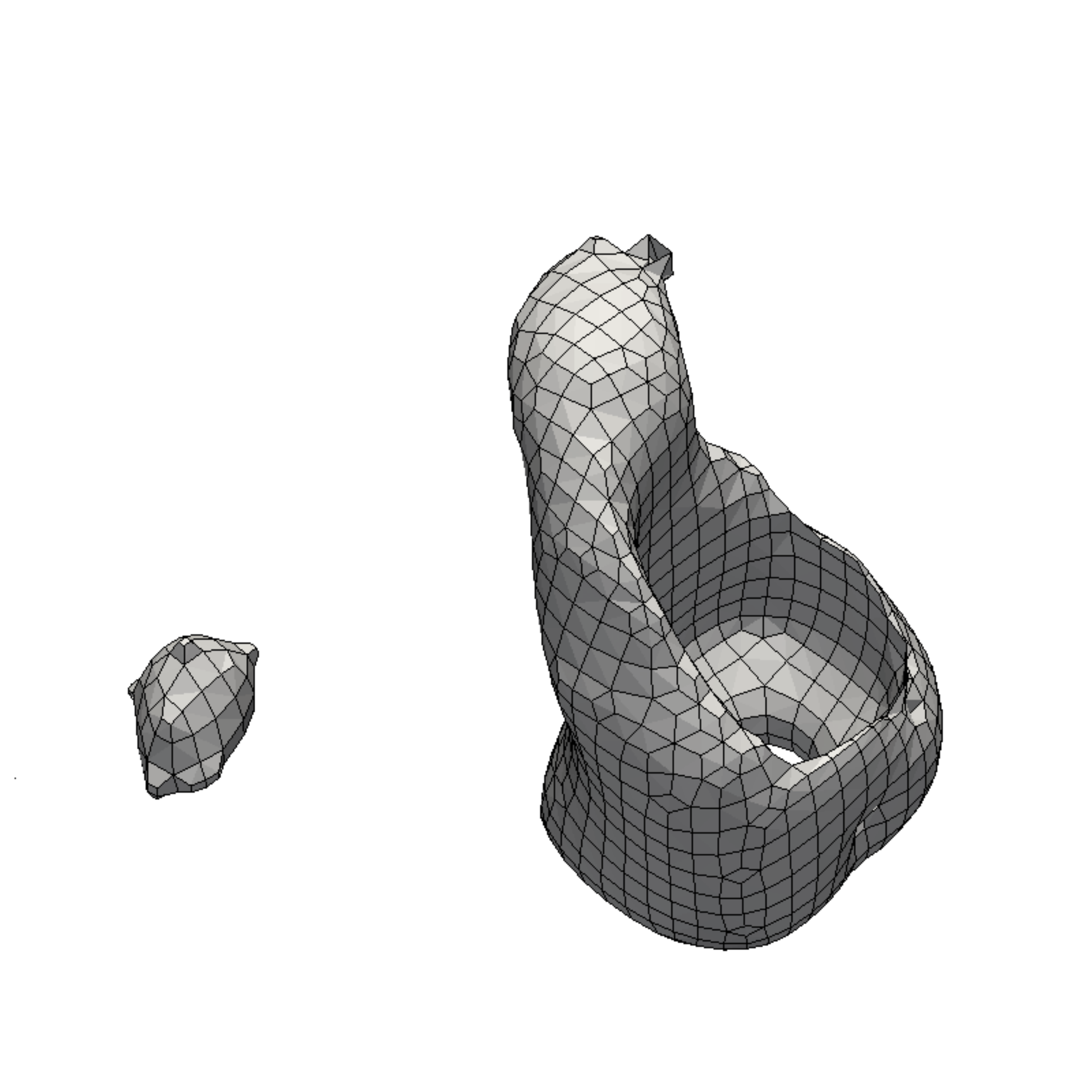} & \includegraphics[width=0.25\textwidth]{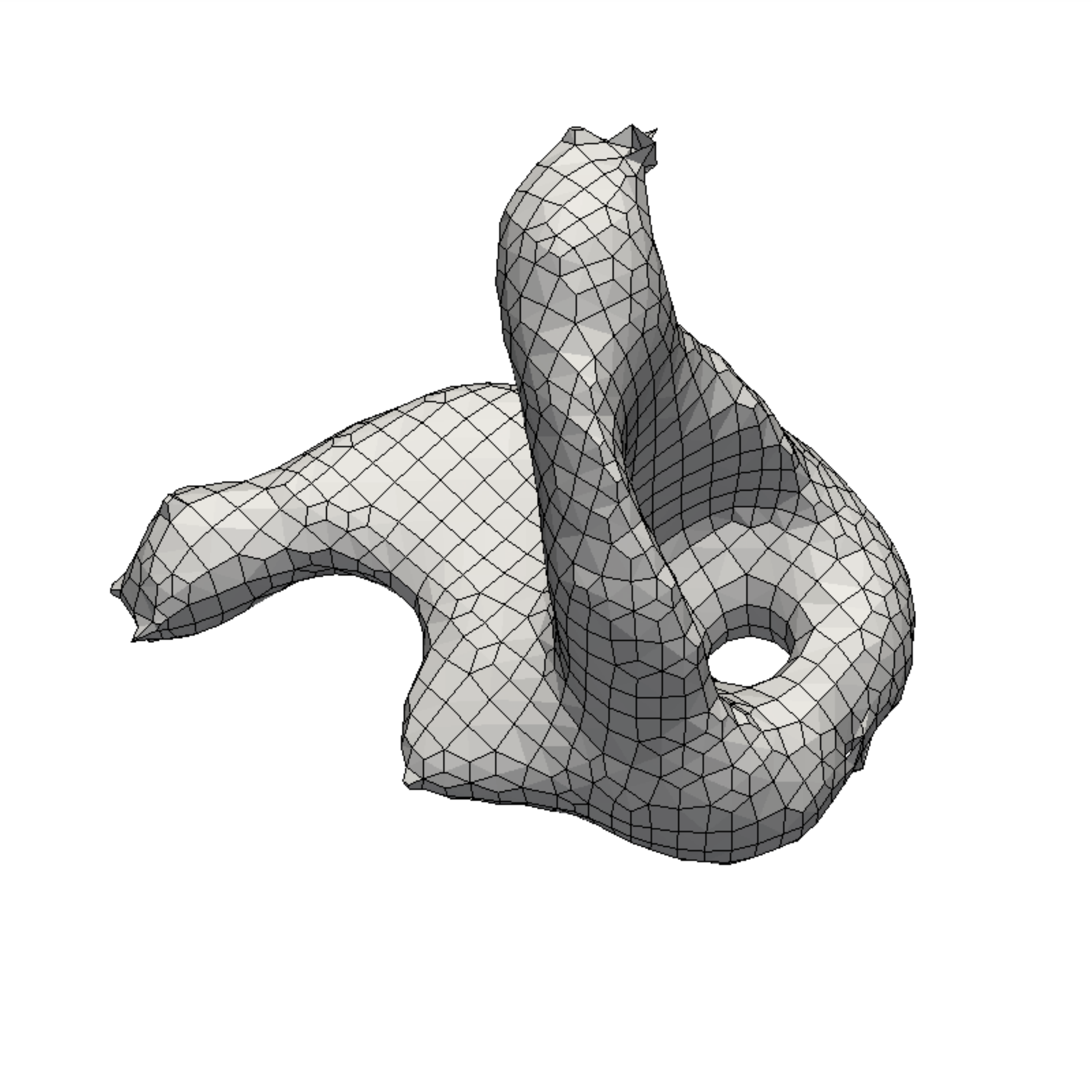} & \includegraphics[width=0.25\textwidth]{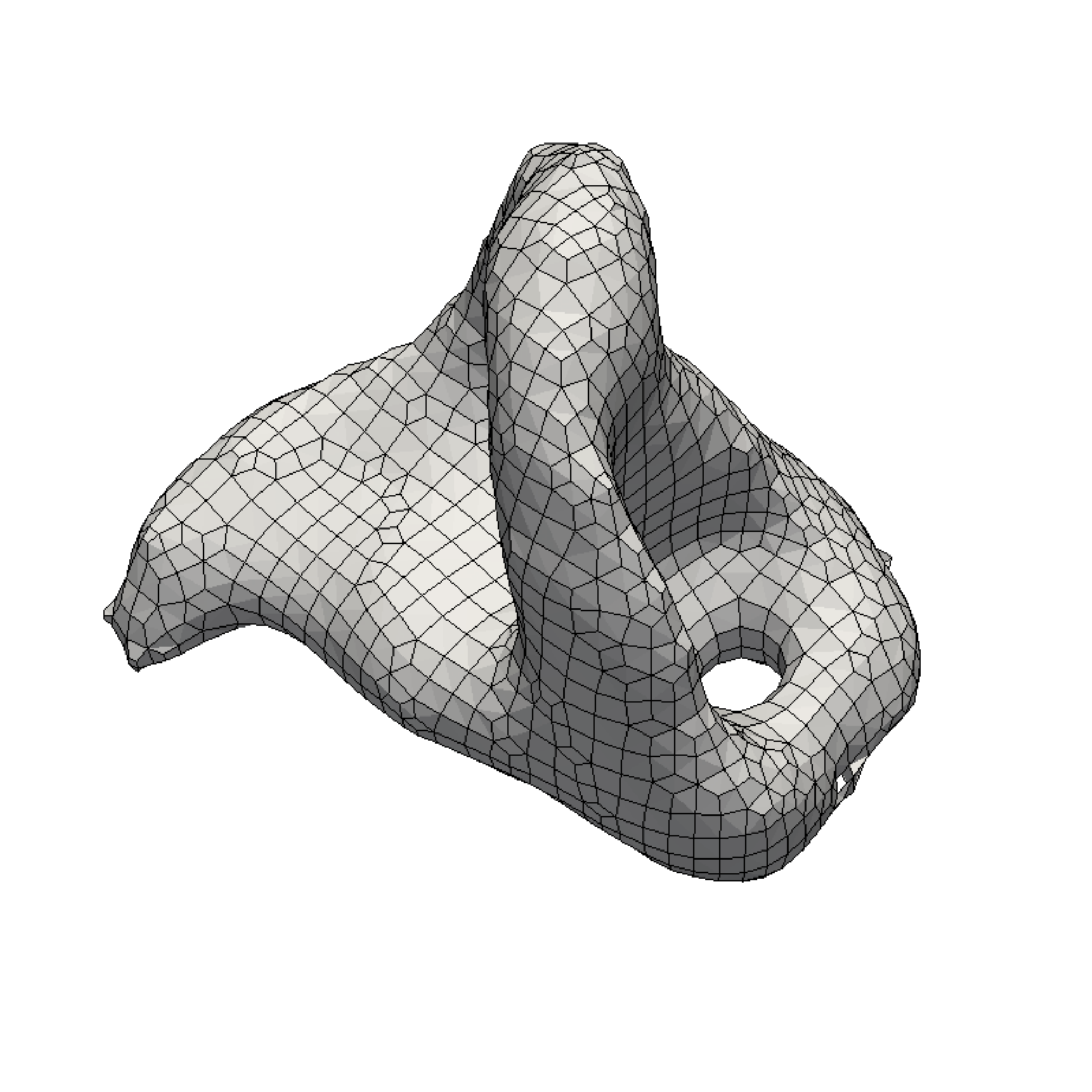}\\

\hline
\end{tabular}

}
        \subcaption{sphere complex}
    \end{subtable}
\end{table}

\end{document}